\PassOptionsToPackage{table}{xcolor}
\PassOptionsToPackage{dvipsnames}{xcolor}
\PassOptionsToPackage{xcdraw}{xcolor}

\documentclass[10pt,twocolumn,letterpaper]{article}
\usepackage{utils/cvpr}              %

\usepackage{pifont}

\usepackage{times}
\usepackage{epsfig}
\usepackage{graphicx}
\usepackage{bm}
\usepackage{amsmath}
\usepackage{amssymb}
\usepackage{amsthm}
\usepackage{booktabs}

\usepackage{xcolor}
\usepackage{caption,subcaption}

\usepackage[accsupp]{axessibility}

\ifdefined\siggraph
\usepackage{times}
\fi

\usepackage{color}
\usepackage{ifthen}
\usepackage{float}
\usepackage{alltt}
\usepackage{newlfont} %
\usepackage{wrapfig}
\usepackage{booktabs}
\usepackage{multirow}
\usepackage{makecell}
\usepackage{array}
\usepackage{appendix}
\usepackage{comment}
\usepackage{amsmath}
\usepackage{amssymb}
\usepackage{amsthm}
\usepackage{textcomp}
\usepackage{adjustbox}

\newcommand{\Caption}[2]{\caption[#1]{{\footnotesize #1} {\footnotesize #2}}}

\newcommand{\nothing}[1]{}

\definecolor{DeltaColor}{rgb}{0.039,0.73,0.71}
\definecolor{SetaColor}{rgb}{0.867, 0.0235, 0.376}
\definecolor{SigmaColor}{rgb}{0.98,0.45,0.0}
\definecolor{RedColor}{rgb}{0.8,0,0}
\definecolor{AlphaColor}{rgb}{0,0,0.8}
\definecolor{BetaColor}{rgb}{0.8,0,0.8}
\definecolor{GammaColor}{rgb}{0.5,0,0.7}
\definecolor{EpsilonColor}{rgb}{0.353,0.725,0.906}
\definecolor{TauColor}{rgb}{0.423,0.235,0.192}

\newcommand{\delete}[1]{}

\definecolor{AudioColor}{rgb}{0.56,0.34,0.62}

\definecolor{DeadlineColor}{rgb}{0.9,0.4,0} %

\definecolor{figred}{rgb}{1,0,0}
\definecolor{figgreen}{rgb}{0,0.6,0}
\definecolor{figblue}{rgb}{0,0,1}
\definecolor{figpink}{rgb}{1,0.63,0.63}

\newcolumntype{C}[1]{>{\centering}m{#1}}

\newcounter{pccount}
\floatstyle{plain}

\newcommand{\filename}[1]{\url{#1}}
\newcommand{\foldername}[1]{\url{#1}}

\DeclareMathOperator*{\argmin}{argmin}

\newcommand{\tabincell}[2]{\begin{tabular}
{@{}#1@{}}#2\end{tabular}}

\newcommand{\OurNetName}{NeUDF}

\usepackage{lipsum}

\usepackage{dsfont}
\usepackage{setspace}
\usepackage{threeparttable}
\usepackage{multirow}
\usepackage{cite}
\usepackage[ruled]{algorithm}
\usepackage{algpseudocode}
\usepackage{algorithmicx}
\usepackage{adjustbox}

\usepackage[pagebackref=false,breaklinks,colorlinks,citecolor=RoyalBlue,bookmarks=false]{hyperref}

\usepackage[capitalize]{cleveref}
\crefname{section}{Sec.}{Secs.}
\Crefname{section}{Section}{Sections}
\Crefname{table}{Table}{Tables}
\crefname{table}{Tab.}{Tabs.}

\def\cvprPaperID{xxxx} %
\def\confName{CVPR}
\def\confYear{2023}

\graphicspath{
	{figs/raster_pdf/}
	{figs/handdrawn/}
}

\begin{document}

    \title{\OurNetName: Leaning Neural Unsigned Distance Fields with Volume Rendering}
    
\author{
\textbf{Yu-Tao Liu}\textsuperscript{1,2}
\quad 
\textbf{Li Wang}\textsuperscript{1,2}
\quad 
\textbf{Jie Yang}\textsuperscript{1}
\quad 
\textbf{Weikai Chen}\textsuperscript{3}
\quad \\
\textbf{Xiaoxu Meng}\textsuperscript{3}
\quad 
\textbf{Bo Yang}\textsuperscript{3}
\quad 
\textbf{Lin Gao}\textsuperscript{1,2*} \\
\textsuperscript{1}Beijing Key Laboratory of Mobile Computing and Pervasive Device, \\Institute of Computing Technology, Chinese Academy of Sciences\\
\textsuperscript{2}University of Chinese Academy of Sciences\\
\textsuperscript{3}Digital Content Technology Center, Tencent Games\\
{\tt\small liuyutao17@mails.ucas.ac.cn} 
\quad
{\tt\small \{wangli20s, yangjie01\}@ict.ac.cn} 
\quad
{\tt\small chenwk891@gmail.com} 
\quad\\
{\tt\small \{xiaoxumeng, brandonyang\}@global.tencent.com} 
\quad
{\tt\small gaolin@ict.ac.cn} 
\quad
}

    \twocolumn[{%
    \renewcommand\twocolumn[1][]{#1}%
    \maketitle
    \begin{center}
    \centering
    \captionsetup{type=figure}
    \vspace{-8mm}

    \includegraphics[width=\linewidth]{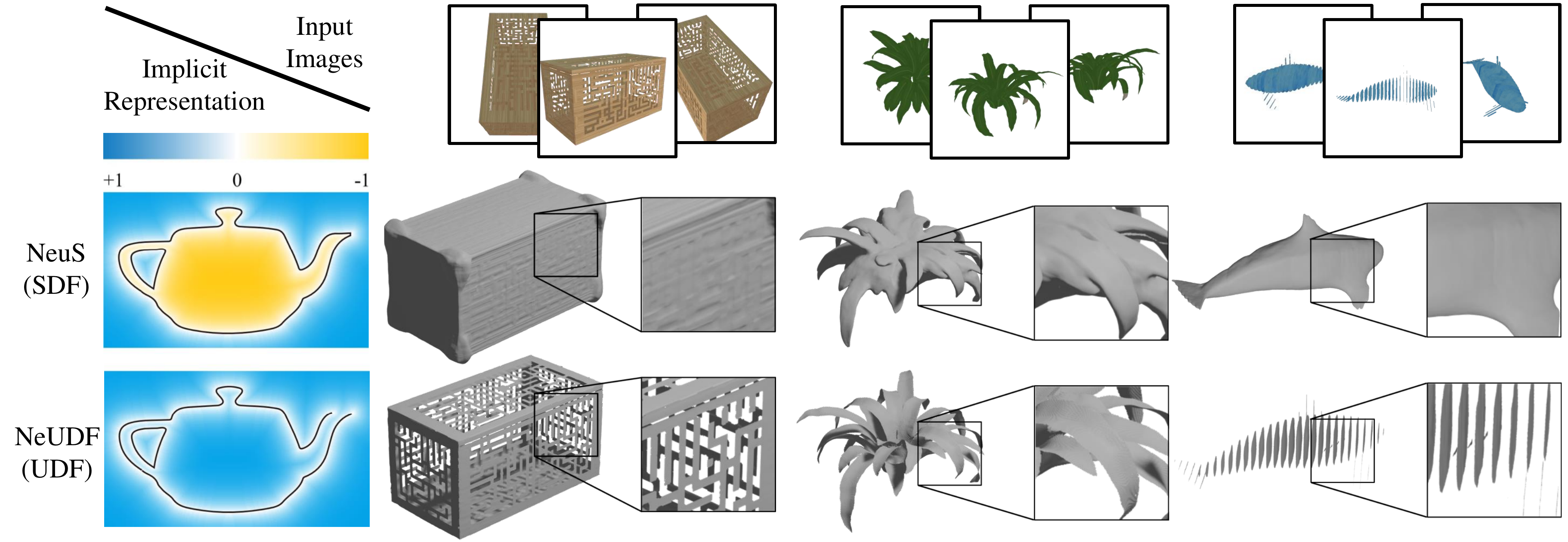}
    \vspace{-8mm}
    \captionof{figure}{
    We show comparisons of the input multi-view images (top), watertight surfaces (middle) reconstructed with state-of-the-art SDF-based volume rendering method NeuS~\cite{DBLP:conf/nips/WangLLTKW21}, and open surfaces (bottom) reconstructed with our method. Our method is capable of reconstructing high-fidelity shapes with both open and closed surfaces from multi-view images.
    }
    \label{fig:teaser}
    \end{center}%
    }]
    
    \if TT\insert\footins{\footnotesize{
*Corresponding Author is Lin Gao (gaolin@ict.ac.cn).}}\fi
    \begin{abstract}

Multi-view shape reconstruction has achieved impressive progresses thanks to the latest advances in neural implicit surface rendering.
However, existing methods based on signed distance function (SDF) are limited to closed surfaces, failing to reconstruct a wide range of real-world objects that contain open-surface structures.
In this work, we introduce a new neural rendering framework, coded \OurNetName{}\footnote{
Visit our project page at
\href{http://geometrylearning.com/neudf/}
{http://geometrylearning.com/neudf/}
}
, that can reconstruct surfaces with arbitrary topologies solely from multi-view supervision.
To gain the flexibility of representing arbitrary surfaces, \OurNetName{} leverages the unsigned distance function (UDF) as surface representation. 
While a naive extension of an SDF-based neural renderer cannot scale to UDF, we propose two new formulations of weight function specially tailored for UDF-based volume rendering.
Furthermore, to cope with open surface rendering, where the in/out test is no longer valid, we present a dedicated normal regularization strategy to resolve the surface orientation ambiguity.
We extensively evaluate our method over a number of challenging datasets, including DTU~\cite{jensen2014large}, MGN~\cite{DBLP:conf/iccv/BhatnagarTTP19}, and Deep Fashion 3D~\cite{zhu2020deep}.
Experimental results demonstrate that \OurNetName{} can significantly outperform the state-of-the-art method in the task of multi-view surface reconstruction, especially for complex shapes with open boundaries.

\end{abstract}

    \section{Introduction}
\label{sec:intro}

Multi-view surface reconstruction is a long-standing and fundamental problem in computer vision and computer graphics.
Conventional multi-view stereo based methods~\cite{DBLP:conf/cvpr/SchonbergerF16, DBLP:conf/eccv/SchonbergerZFP16} often underperform when the input images are sparse or appear textureless.
Recent advances in neural implicit representation~\cite{DBLP:conf/eccv/MildenhallSTBRN20,park2019deepsdf,chen2019learning,mescheder2019occupancy} have brought impressive progress in achieving high-quality reconstruction of intricate geometry even with sparse views.  
Specifically, they~\cite{DBLP:conf/nips/WangLLTKW21, zhu2022nice, long2022sparseneus,yariv2021volume,darmon2022improving,wang2022hfneus,Fu2022GeoNeus} leverage the volume rendering scheme to jointly learn the implicit geometry and color field by minimizing the discrepancy between the rendering results and the input images. 
However, since these methods represent surfaces using either signed distance function (SDF)~\cite{long2022sparseneus,DBLP:conf/nips/WangLLTKW21} or occupancy field~\cite{DBLP:conf/iccv/OechsleP021}, they can only reconstruct watertight shapes. 
This greatly limits their applications as shapes with open surfaces, such as garments, 3D-scanned scenes, \emph{etc}, are widely seen in the real world.
Recent works, such as NDF~\cite{DBLP:conf/nips/ChibaneMP20}, 3PSDF~\cite{chen_2022_3psdf}, and GIFS~\cite{GIFS_arxiv}, have proposed new neural implicit functions to represent surfaces with arbitrary topologies. 
Nonetheless, none of these methods is compatible with existing neural rendering frameworks.
Hence, how to leverage neural rendering to reconstruct non-watertight shapes, \emph{e.g.} open surfaces, remains an open question.

\begin{figure}
    \centering
    \begin{minipage}[c]{.13\linewidth}
    \includegraphics[width=0.9\linewidth]{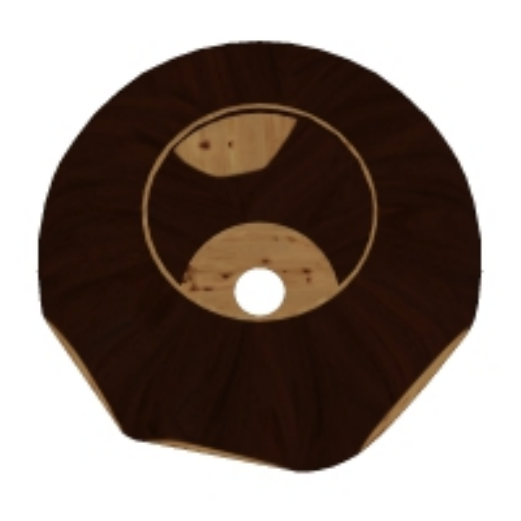}
    
    \includegraphics[width=0.9\linewidth]{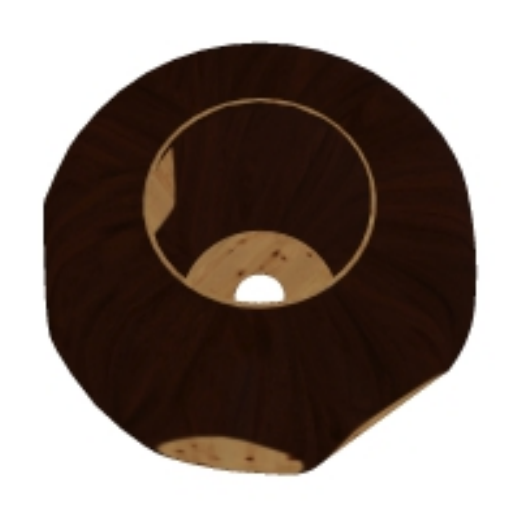}
    
    {\includegraphics[width=0.9\linewidth]{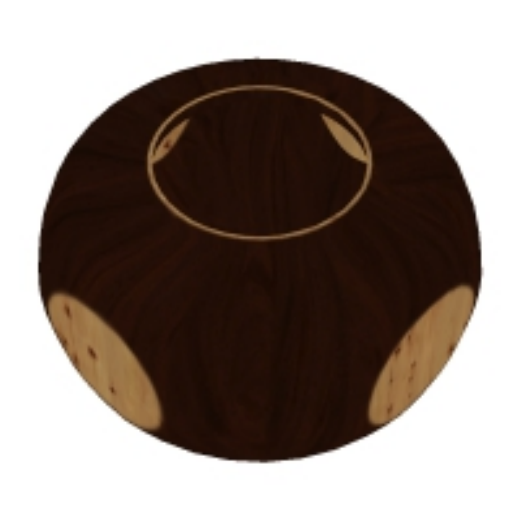}} 

    \end{minipage}
    \begin{minipage}[c]{.42\linewidth}
        {\includegraphics[width=\linewidth]{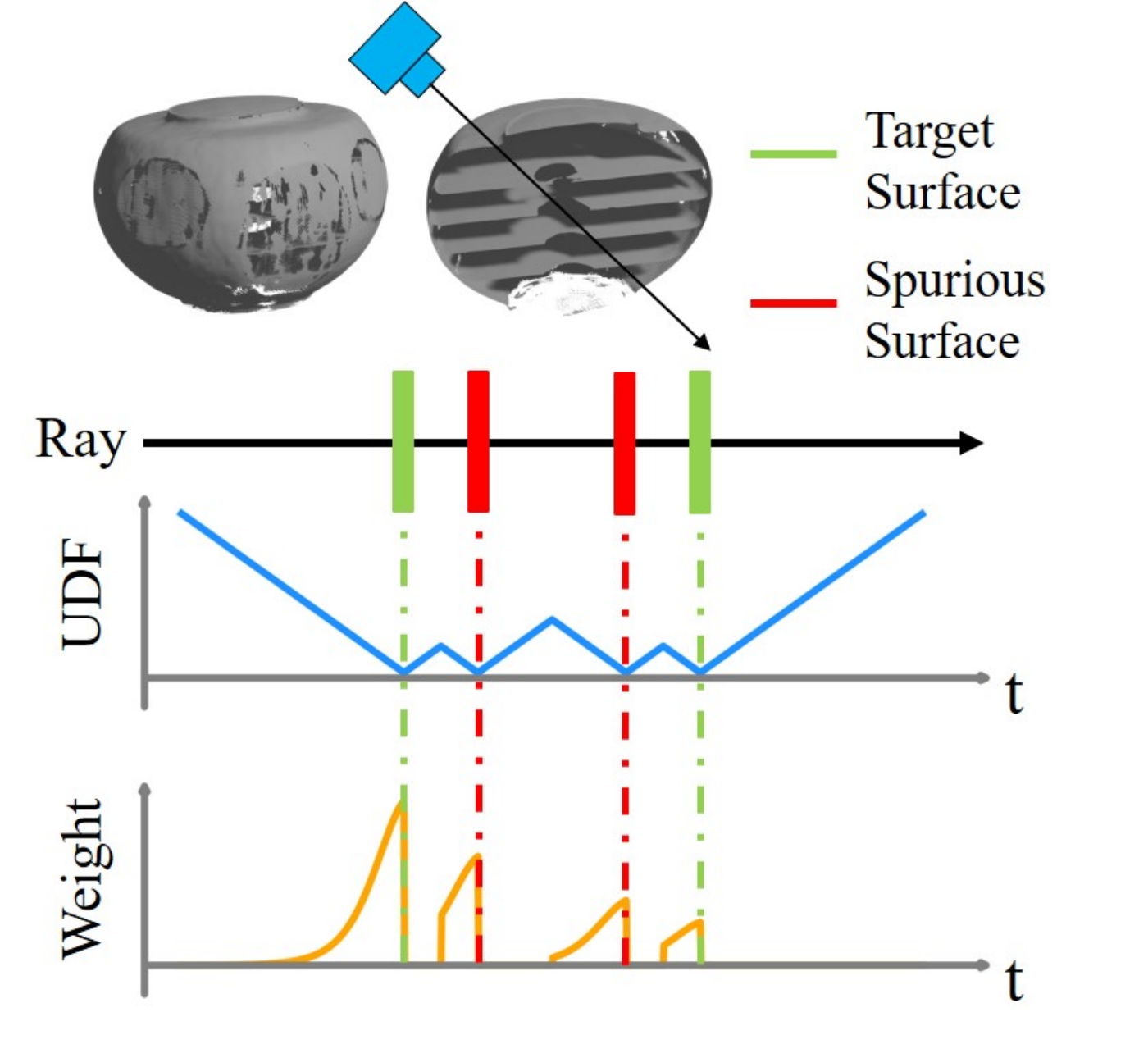}}
    \end{minipage}
    \begin{minipage}[c]{.42\linewidth}
        {\includegraphics[width=\linewidth]{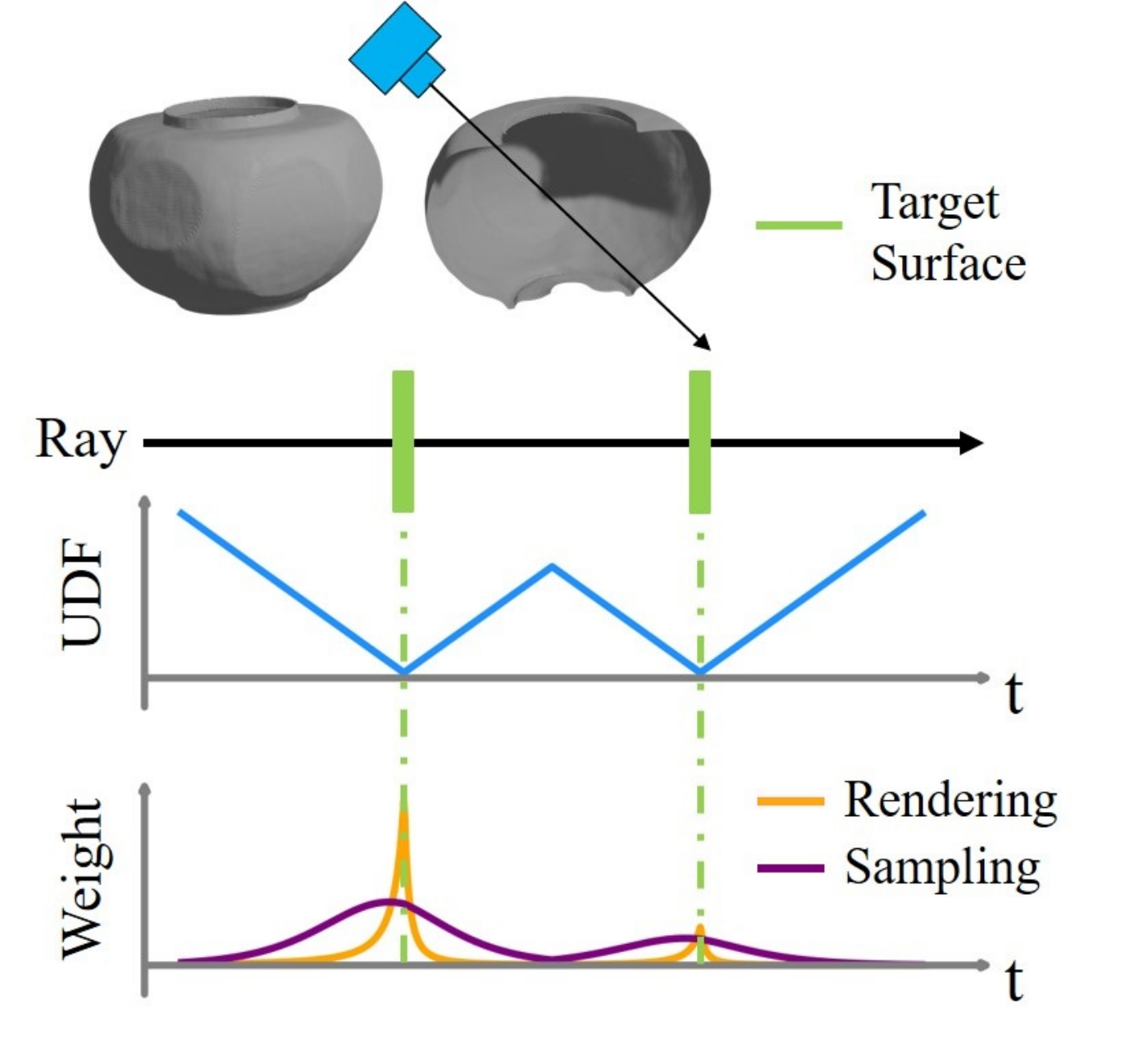}}
    \end{minipage}
    \vspace{-5mm}

    \begin{minipage}[c]{.13\linewidth}
        \centering
         \vspace{3mm}
        \subcaption{Input}
    \end{minipage}
    \begin{minipage}[c]{.42\linewidth}
        \centering
        \vspace{3mm}
        \subcaption{\tabincell{c}{Naive UDF solution based\\on SDF renderer}}
    \end{minipage}
    \begin{minipage}[c]{.42\linewidth}
        \centering
         \vspace{3mm}
        \subcaption{NeUDF}
    \end{minipage}
    \vspace{-3mm}
    
    \Caption{As shown in (b), the naive UDF solution based on the SDF renderer is biased, thus resulting in redundant surfaces in the reconstruction. NeUDF solves this problem by introducing a novel unbiased weighting mechanism as shown in (c). }

\label{fig:motivation}
\end{figure}

We fill this gap by introducing \OurNetName{}, a new volumetric rendering framework that can reconstruct shapes with arbitrary topologies only from multi-view image supervision.
\OurNetName{} is built upon the unsigned distance function (UDF), a straightforward implicit function that returns the absolute distance from a query point to the target surface.
Despite its simplicity, we show that naively extending the SDF-based neural rendering mechanism to unsigned distance fields cannot ensure unbiased rendering of non-watertight surfaces.
In particular, as shown in Figure~\ref{fig:motivation}, the SDF-based weighting function would generate spurious surfaces where the rendering weight triggers undesirable local maxima in the void region.
To resolve this issue, we propose a new unbiased weighting paradigm specially tailored for UDF while being aware of surface occlusions.
To accommodate the proposed weighting function, we further present a customized importance sampling strategy that ensure high-quality reconstruction of non-watertight surfaces.
Furthermore, to tackle the inconsistent gradients of UDFs near the zero iso-surface, we introduce a normal regularization method to enhance the gradient consistency by leveraging normal information in the surface neighborhood.

To the best of our knowledge, \OurNetName{} is the first attempt to reconstruct the surfaces with arbitrary topologies solely from 2D image supervision.
Extensive experiments on the public datasets, \eg MGN~\cite{DBLP:conf/iccv/BhatnagarTTP19}, Deep Fashion3D~\cite{zhu2020deep}, and BMVS~\cite{yao2020blendedmvs}, demonstrate that \OurNetName{}
can significantly outperform the state-of-the-art methods in the task of open surface reconstruction while achieving comparable results in recovering watertight surfaces.
We summarize our contributions as follows:
\begin{itemize}
    \item The first UDF-based neural volume rendering framework, dubbed \OurNetName{}, that can be used for multi-view reconstruction of shapes with arbitrary topologies, including complex shapes with open boundaries.
    \item A novel unbiased weighting function and importance sampling strategy specially tailored for UDF rendering.
    \item The new state-of-the-art performance in the task of multi-view surface reconstruction over a number of challenging datasets with non-watertight 3D shapes.
\end{itemize}

    \section{Related Works}\label{sec:related_work}

In this section, we first discuss classical implicit representations and neural rendering techniques. 
Next, we provide an overview of recent works on combining them to improve the performance of multi-view reconstruction tasks.

\vspace{-3mm}
\paragraph{Neural Implicit Representation}
Recent developments in neural implicit representation~\cite{chen2019learning,michalkiewicz2019deep,park2019deepsdf,remelli2020meshsdf, WANG202241} have surpassed the prior topological and resolution limit of explicit representations (\eg point clouds, voxels, and meshes), setting a new state of the art for 3D modeling and reconstruction. Complex shapes can be implicitly represented by classifying the query points into inside or outside the shape (binary occupancy)~\cite{mescheder2019occupancy,peng2020convolutional,chibane2020implicit1,genova2019deep,saito2019pifu,deng2020nasa,chibane2020implicit} or predicting the signed distance (SDF) to the  surface~\cite{chen2019learning,michalkiewicz2019deep,park2019deepsdf,jiang2020local,remelli2020meshsdf}. Due to the reliance on in/out the partition of 3D space, such methods can only model watertight objects. 
Methods based on unsigned distance function (UDF)~\cite{DBLP:conf/nips/ChibaneMP20,venkatesh2020dude,Venkatesh_2021_ICCV,Wang22HSDF,GIFS_arxiv} are proposed to overcome the limitation, enabling deep neural networks to properly represent and learn a much wider range of shapes with open surfaces. 
NDF~\cite{DBLP:conf/nips/ChibaneMP20} predicts an unsigned distance from an input query point and its position-aware shape feature which is encoded in a multi-scale manner. 
HSDF~\cite{Wang22HSDF} simultaneously predicts a UDF field and a sign field to achieve better mesh fidelity. 
But they~\cite{DBLP:conf/nips/ChibaneMP20,venkatesh2020dude,Venkatesh_2021_ICCV,Wang22HSDF,GIFS_arxiv} require 3D supervision for mesh reconstruction. 

\vspace{-3mm}
\paragraph{Neural Rendering}
Besides the geometry information, appearance information is also needed to faithfully depict a scene, especially when the input observations take the form of 2D pictures. Methods based on neural implicit surface rendering~\cite{DBLP:conf/nips/SitzmannZW19, DBLP:conf/cvpr/NiemeyerMOG20, DBLP:conf/cvpr/LiuZPSPC20, DBLP:conf/nips/LiuS0L19, DBLP:conf/nips/YarivKMGABL20, DBLP:conf/cvpr/KellnhoferJJSPW21, DBLP:conf/cvpr/TakikawaLYKLNJM21} find the intersection between a ray and the surface using differential sphere tracing~\cite{DBLP:journals/vc/Hart96} or its variants. They query the RGB color of the ray-surface intersection point using another network branch. 
Because the back-propagated gradients are influenced by the entire space, surface rendering methods like IDR~\cite{DBLP:conf/nips/YarivKMGABL20} and DVR~\cite{DBLP:conf/cvpr/NiemeyerMOG20} struggle in reconstructing complex shapes without additional 2D mask supervision.
In contrast, the methods based on neural volumetric rendering~\cite{DBLP:conf/eccv/MildenhallSTBRN20, DBLP:conf/cvpr/Martin-BruallaR21, DBLP:conf/cvpr/Niemeyer021, DBLP:conf/cvpr/PumarolaCPM21, DBLP:conf/cvpr/SrinivasanDZTMB21, DBLP:journals/corr/abs-2010-07492, DBLP:journals/cgf/NeffSPKMCKS21} imply that rather than a binary intersection case, rays can have a chance of interacting with the scene properties at every point in space. For machine learning pipelines that largely rely on the availability of well-behaved gradients for optimization, this continuous model performs well as a differentiable rendering framework.

\vspace{-3mm}
\paragraph{Multi-view Reconstruction}
Multi-view stereo approaches~\cite{DBLP:conf/cvpr/AgrawalD01, DBLP:conf/bmvc/BleyerRR11,de1999poxels,DBLP:conf/iccv/BroadhurstDC01,DBLP:conf/iccv/KutulakosS99,DBLP:conf/eccv/SchonbergerZFP16,DBLP:conf/cvpr/SeitzD97,DBLP:conf/cvpr/SeitzCDSS06} before  the advent of deep learning mainly rely on image feature matching~\cite{DBLP:conf/bmvc/BleyerRR11,DBLP:conf/eccv/SchonbergerZFP16} across viewpoints or volumetric representation like voxel grids~\cite{DBLP:conf/cvpr/AgrawalD01,de1999poxels,DBLP:conf/iccv/BroadhurstDC01,DBLP:conf/iccv/KutulakosS99,DBLP:conf/cvpr/SeitzD97}. The former, like the widely used method COLMAP~\cite{DBLP:conf/eccv/SchonbergerZFP16}, highly relies on rich texture information and classic meshing techniques from point clouds because it computes multi-view depth maps from correspondence between images and fuses them into dense point clouds, while the latter is limited to low resolution due to the cubic memory growth of voxel representation. 

Recent works~\cite{DBLP:conf/cvpr/NiemeyerMOG20,DBLP:conf/cvpr/KellnhoferJJSPW21,DBLP:conf/nips/YarivKMGABL20,DBLP:conf/cvpr/LiuZPSPC20,DBLP:conf/iccv/OechsleP021,long2022sparseneus,DBLP:conf/nips/WangLLTKW21,yariv2021volume,darmon2022improving,wang2022hfneus,Fu2022GeoNeus} combining implicit representation and neural rendering outperform previous approaches in reconstructing watertight surfaces with high fidelity. Since these methods represent surfaces using either occupancy values~\cite{DBLP:conf/iccv/OechsleP021} or signed distance function~\cite{long2022sparseneus,DBLP:conf/nips/WangLLTKW21} (SDF), their reconstruction results are limited to be watertight.
Our \OurNetName{} proposes a novel neural volume rendering algorithm for unsigned distance function (UDF) and thus can naturally extract the surface as the zero-level set of UDF, which is capable of representing complex shapes with open surfaces and thin structures.
    \section{Methodology}

Given a set of calibrated images $\{\mathcal{I}_k| 1\leq k \leq n\}$ of a object or scene, we aim to reconstruct arbitrary surfaces, including closed and open structures, only {using 2D image supervision.}
In our paper, a surface is represented as a zero-level set of unsigned distance functions (UDFs).
To learn the UDF representation of objects or scenes, we introduce a novel neural  rendering architecture that incorporates  unbiased formulation of weights for rendering.
We first define our scene representation based on UDF (Sec.~\ref{sec:udf}). Then 
we introduce \OurNetName{} with two key formulations of weight function specially tailored for UDF-based volume rendering (Sec.~\ref{sec:renderudf}). 
Finally, we illustrate our normal regularization (Sec.~\ref{sec:normal}) for alleviating the ambiguity from 2D images  and our loss configuration (Sec.~\ref{sec:loss}).

\subsection{Scene Representation}
\label{sec:udf}

{Different from signed distance function (SDF), unsigned distance function (UDF) is sign-less and capable of representing open surfaces with arbitrary topologies, in addition to watertight surfaces.}
Given a 3D object $\mathcal{O} = \{V, F\}$, where $V$ and $F$ are the collections of vertices and faces,
the UDF of an object $\mathcal{O}$ can be formulated as a function $d = \Psi_{\mathcal{O}}(x): \mathbb{R}^3 \mapsto \mathbb{R}^+$, which maps a point coordinate $x$ to the Euclidean distance $d$ to the surface. We define UDF$_\mathcal{O} = \{\Psi_{\mathcal{O}}(x) | d < \epsilon, d = \argmin_{f \in F }(\|x - f\|_2)\}$, where $\epsilon$ is a small threshold, and the surface of the object can be modulated by the zero-level set of UDF$_\mathcal{O}$.

We introduce a differentiable volume rendering framework to predict UDF from input images. The framework is approximated by a neural network $\psi$, that predicts a UDF value $d$ and the rendering color $c$ according to a spatial location $x$ along the sampling ray $v$:
\begin{equation}
    (d, c) = \psi(v, x): \mathbb{S}^2 \times \mathbb{R}^3 \mapsto (\mathbb{R}^+,\left[0,1\right]^3)
\end{equation}
{With the help of volume rendering, }the weights are optimized by minimizing the distance between the predicted images $\mathcal{I}_k^\prime$ and ground-truths $\mathcal{I}_k$.

The learned surface $\mathcal{S}_\mathcal{O}$ can be represented by the zero-level set of the predicted UDF:
\begin{equation}
    \mathcal{S}_\mathcal{O} = \{x\in\mathbb{R}^3 | d = 0, (d, c) = \psi(v, x)\}
\end{equation}

\subsection{\OurNetName{} Rendering}\label{sec:renderudf}

Rendering procedure is the key to learning an accurate UDF as it connects the output color and the UDF value via integration along ray $v$:
\begin{equation}
    C(o, v)=\int_{0}^{+\infty} w(t)c(p(t), v) dt,
\end{equation}
where $C(o, v)$ is the output pixel color from the camera origin $o$ along the view direction $v$, $w(t)$ is the weight function for the point $p(t)$, and $c(p(t), v)$ is the color at the point $p(t)$ along the view direction $v$.

To reconstruct UDFs via volume rendering, we first introduce a probability density function $\varsigma_r'(\Psi(x))$, called \textit{U-density}, where $\Psi(x)$ is the unsigned distance of $x$. The U-density function $\varsigma_r'(\Psi(x))$ maps UDF field to a probability density distribution which assumes prominently high values near the surface for accurate reconstruction. {Inspired by NeuS~\cite{DBLP:conf/nips/WangLLTKW21},} 
we derive an unbiased and occlusion-ware weight function $w_{r}(t)$ and its opaque density $\tau_r(t)$ using U-density function as:
\begin{gather}
    w_{r}(t)=\tau_r(t)e^{-\int_0^t\tau_r(u)du} \label{eq:weight_r}
    \\
    \tau_r(t)=\left|\frac{\frac{\partial (\varsigma_r \circ \Psi \circ p)}{\partial t}(t)}{\varsigma_r \circ \Psi \circ p(t)}\right| \label{eq:tau_r}
\end{gather}
where $\circ$ is the function composition operator, and $\varsigma_r(\cdot)$ must satisfy the following rules for valid UDF reconstruction: 
\begin{gather}
    \varsigma_r(0)=0, \lim_{d\to +\infty}\varsigma_r(d) = 1
    \label{eq:rule1}
    \\
    \varsigma_r'(d)>0; \varsigma_r''(d)<0, \forall d>0
    \label{eq:rule2}
\end{gather}

\setlength{\columnsep}{7pt}%
\setlength{\intextsep}{10pt}
\begin{wrapfigure}[13]{r}{0.18\textwidth}\vspace{-6mm}\hspace{-1mm}%
\centering
    \includegraphics[width=0.18\textwidth]{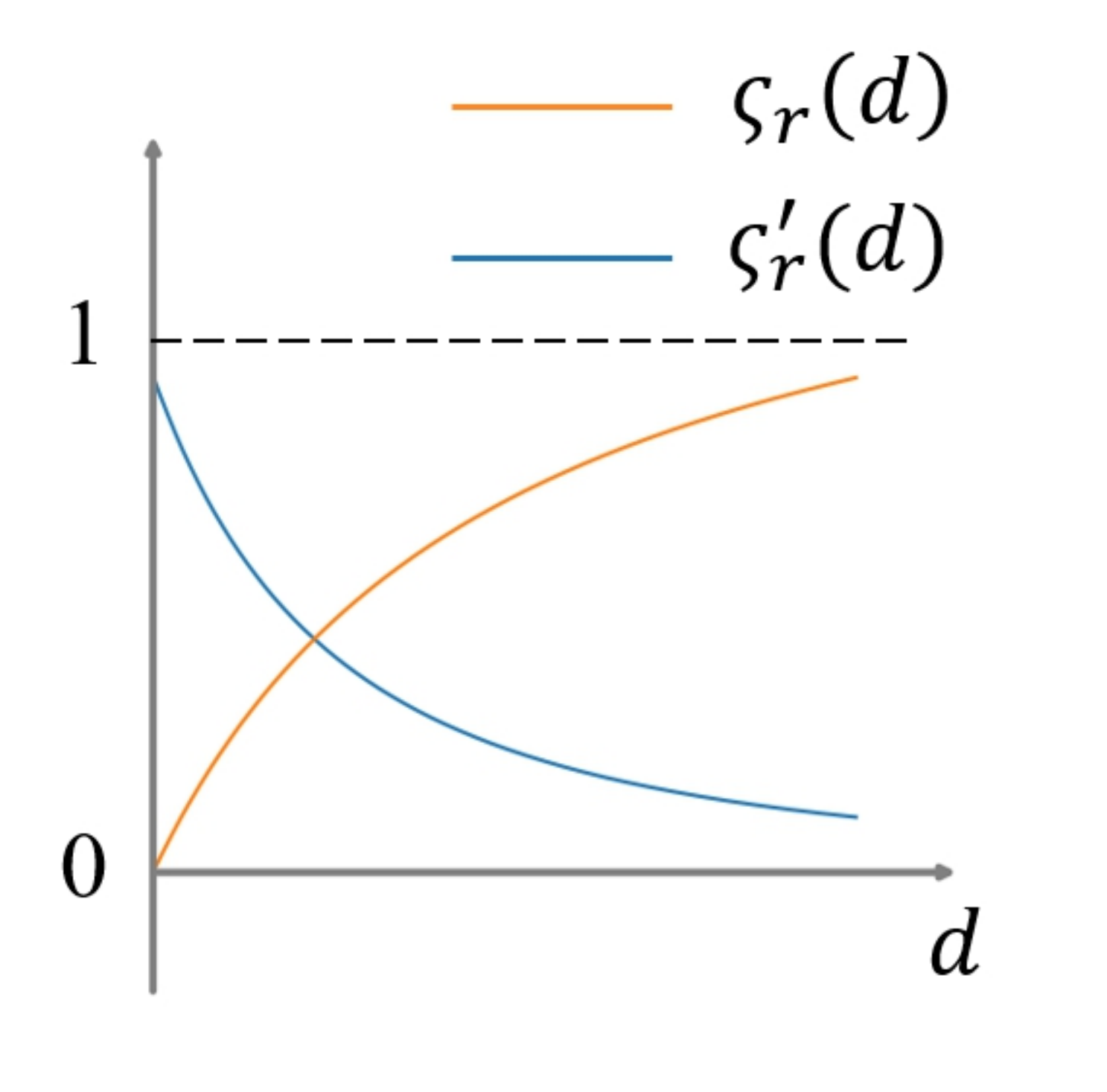}
    \vspace{-3mm}
    \caption{U-density function $\varsigma_r'(d)$ and its cumulative distribution function $\varsigma_r(d)$ satisfying the rules in Equ.~\ref{eq:rule1} and Equ.~\ref{eq:rule2}.}
    \label{fig:hrcurve}
\end{wrapfigure}

The $\varsigma_r(d)$ can be any function shaped in the right figure. Since $\varsigma_r(d)$ is the cumulative distribution function of U-density, $\varsigma_r(0)=0$ guarantees that there is no accumulated density from points with negative distances. Furthermore, $\varsigma_r'(d)>0$ and $\varsigma_r''(d)<0$ ensure U-density values are positive and prominently high for points near the surface.
The parameter $r$ in $\varsigma_r(d)$ is learnable and controls the distribution of the density.
This function structure addresses the volume-surface gap between the volume rendering and the surface reconstruction and guarantees global unbiased property. Please refer to our supplementary for detailed discussions.

We argue that a naive extension of SDF-based neural renderers would violate some of the above rules. For example, the cumulative distribution function of U-density in NeuS~\cite{DBLP:conf/nips/WangLLTKW21} is $\Phi_s$ (Sigmoid Function) and $\Phi_s(0)>0$ violates Equ.~\ref{eq:rule1}.
The violation would lead to bias in rendering weights and thus result in redundant floating faces and irregular noises shown in Fig.~\ref{fig:motivation}.
Note that the local maximal constraint proposed in NeuS cannot address this rendering bias in UDF. 
Please check out the detailed discussion of the unbiased property and the global/local maximal constraint in our supplemental materials. 

After extensive evaluations for different forms of $\varsigma_r(d)$ in the ablation study (Sec.~\ref{sec:abla}), we ultimately choose $\varsigma_r(d)=\frac{r d}{1+r d}$ with $r$ initialized to 0.05.
Further, we adopt the $\alpha$-compositing to discretize the weight function, which samples the points along the ray direction and accumulates the colors according to the weight integral.
For the detailed discretization and proofs of the unbiased and the occlusion-aware properties of Eqn.~\ref{eq:weight_r} and Eqn.~\ref{eq:tau_r}, please refer to our supplemental materials.

\paragraph{Importance points sampling.}
Points sampling that accommodates the rendering weight is an important step in volume rendering. 
Unlike SDF, to achieve unbiased rendering of UDF, the rendering function should distribute more weights before the intersection points (Fig.~\ref{fig:motivation}(c)). Hence, if
both the rendering and sampling functions employ the same
weights, the regularization (the Eikonal loss) on UDF gradients would lead to highly unbalanced gradient magnitudes
on the two sides of the surface. This could significantly
hamper the quality of the reconstructed UDF field. Therefore, we propose a specially-tailored sampling weight function (Fig. 2(c)) to achieve well-balanced regularization all over the space.
The importance sampling $w_{s}(t)$ is formulated as follows:
\begin{equation}
    w_{s}(t)=\tau_s(t)e^{-\int_0^t\tau_s(u)du}, \tau_s(t) = \zeta_s \circ \Psi \circ p(t)\label{eq:weight_s},
\end{equation}

where $\zeta_s(\cdot)$ satisfies the rules: $\zeta_s(d)>0 \text{ and } \zeta_s'(d)<0, \forall d > 0$.
Intuitively, $\zeta_s(\cdot)$ is a monotonically decreasing function in the first quadrant.
In our paper, we use $\zeta_s(d) = \frac{se^{-sd}}{(1+e^{-sd})^2}$,
where the parameter $b$ in $\zeta_s(d)$ controls the intensity at $x=0$.
$s$ starts from 0.05 and changes every sampling step $z$ with the rate set to $2^{z-1}$.
Any sampling function that can achieve balanced regularization with the rendering function is compatible with our framework.
For a detailed illustration of the above rules, please see our supplementary document.
Further, we evaluate the necessity of the $\zeta_s(d)$ qualitatively and quantitatively in the ablation study (Sec.~\ref{sec:abla}).

Overall, the weight functions are collaboratively used in rendering (Eqn.~\ref{eq:weight_r}) and sampling (Eqn.~\ref{eq:weight_s}) during volume rendering, which enables the high-fidelity open surface reconstruction with differentiable volume rendering.

\begin{figure}
    \centering
    \includegraphics[width = 0.9\linewidth]{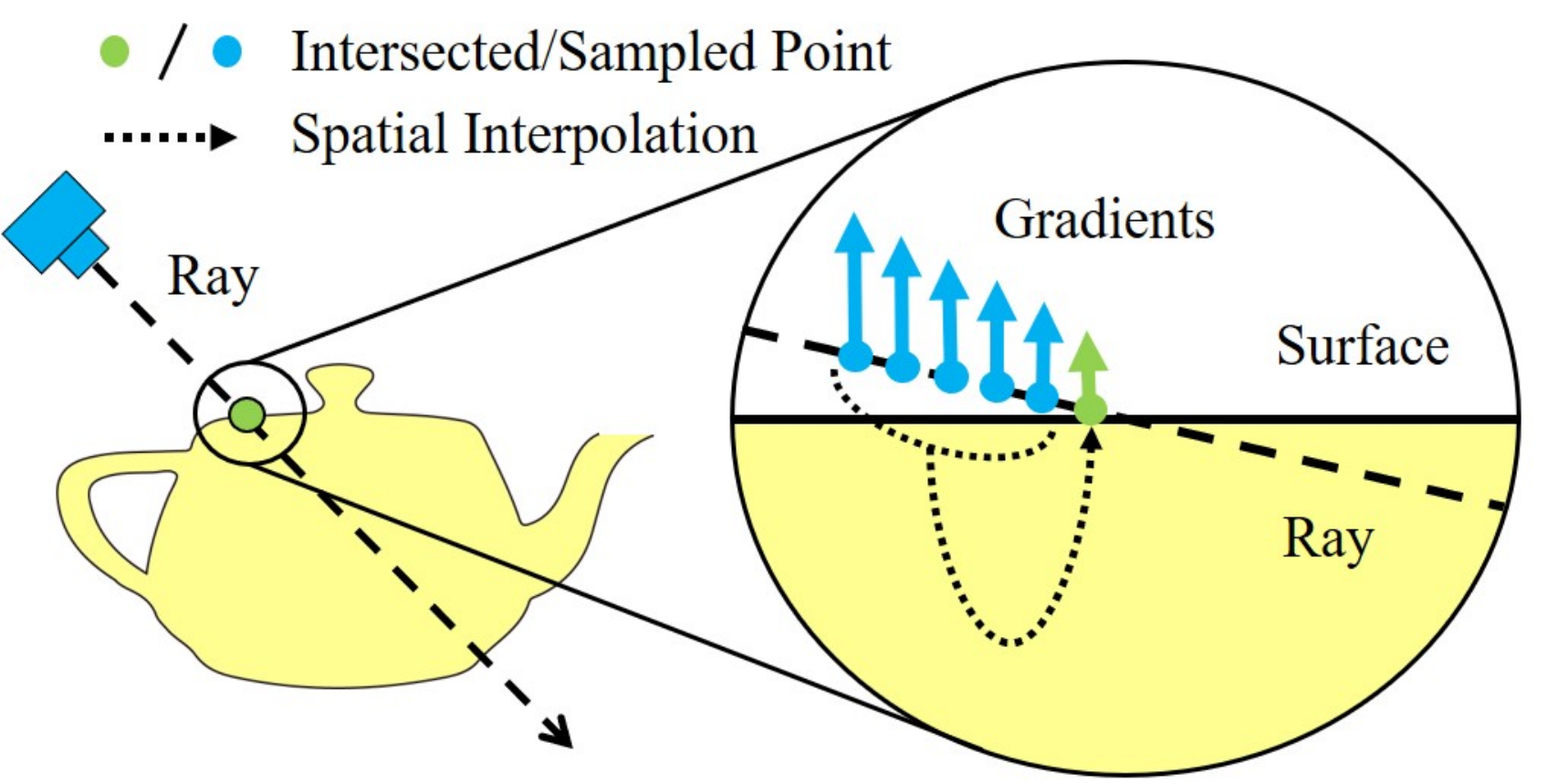}
    \caption{Normal Regularization Diagram. We use the gradients of points (in blue) with an offset from the surface to approximate the unstable surface normal (in green) of UDF representation.}
    \label{fig:normal}
\end{figure}
\subsection{Normal Regularization}\label{sec:normal}
Since points in UDF's zero-level set are cusps that are not first-order differentiable, the gradients of the sampled points in the vicinity of the learned surface are not numerically stable (jittered). 
As the rendering weight function takes as input the UDF gradient, unreliable gradients lead to inaccurate surface reconstruction. 
We introduce a normal regularization to perform spatial interpolation to alleviate this problem.
The normal regularization replaces the naively sampled surface normal with an interpolated normal from its neighborhood.
Figure~\ref{fig:normal} presents a detailed illustration. 
Since the unstable normal only exists near the surface, we use the point normal with an offset from the surface to approximate the unstable normal.
We discretely formulate it at point $p(t_i)$ as follows:
\begin{equation}
    \mathbf{n}(p(t_i)) = \frac{\sum_{k=1}^K w_{i-k} \Psi'(p(t_{i-k}))}{\sum_{k=1}^K w_{i-k}}
\end{equation}
where $w_{i-k} = \|p(i) - p(i-k)\|_2^2$ is distance from $p(i)$ to $p(i-k)$. $\Psi'(\cdot)$ is the derivative of UDF $\Psi(\cdot)$, which returns the gradients of UDF.
By leveraging normal regularization, our framework achieves smoother open surface reconstruction from the 2D images.
We can adjust the normal regularization weight to obtain a more detailed geometry.
Experiments show that normal regularization can prevent the highly bright and dark regions in 2D images from the high-quality reconstruction as shown in Fig.~\ref{fig:abl_brightdark}.

\subsection{Training}\label{sec:loss}

To learn the high-fidelity open surface reconstruction, we optimize the network by minimizing the difference between the rendered images and groundtruth images with known camera poses, without any 3D supervision.
Following NeuS~\cite{DBLP:conf/nips/WangLLTKW21}, we also apply the three loss terms used in SDF volumetric rendering: Color loss $\mathcal{L}_c$, Eikonal loss~\cite{DBLP:conf/nips/YarivKMGABL20} $\mathcal{L}_e$, and Mask loss $\mathcal{L}_m$. The color loss measures the difference between rendered image and input images under L1 loss. The Eikonal loss numerically regularizes the gradients of UDF on sampled points. If the masks are provided, the Mask loss also encourages the predicted mask to be close to the groundtruth mask under the BCE measurement.
Overall, we use a loss that is composed of three parts:
\begin{equation}
    \mathcal{L}=\mathcal{L}_c+\alpha\mathcal{L}_e+\beta\mathcal{L}_m%
\end{equation}
For detailed implementation and network architecture, please refer to our supplementary document.

\begin{figure*}[ht]

\centering

\begin{minipage}[c]{.12\linewidth}
    \centering
    \includegraphics[width=.9\linewidth]{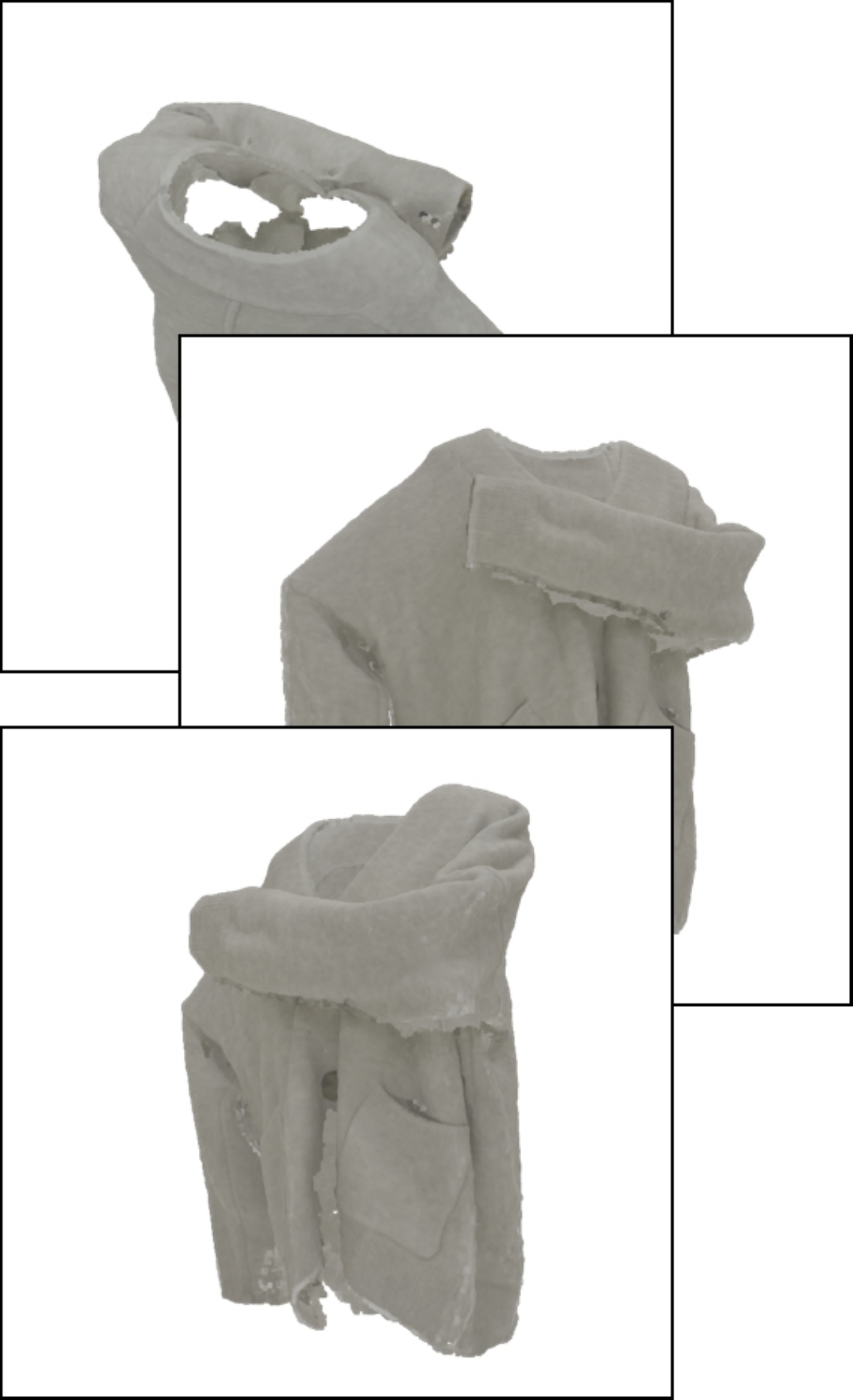}
\end{minipage}
\begin{minipage}[c]{.14\linewidth}
    \centering
    \includegraphics[width=.99\linewidth]{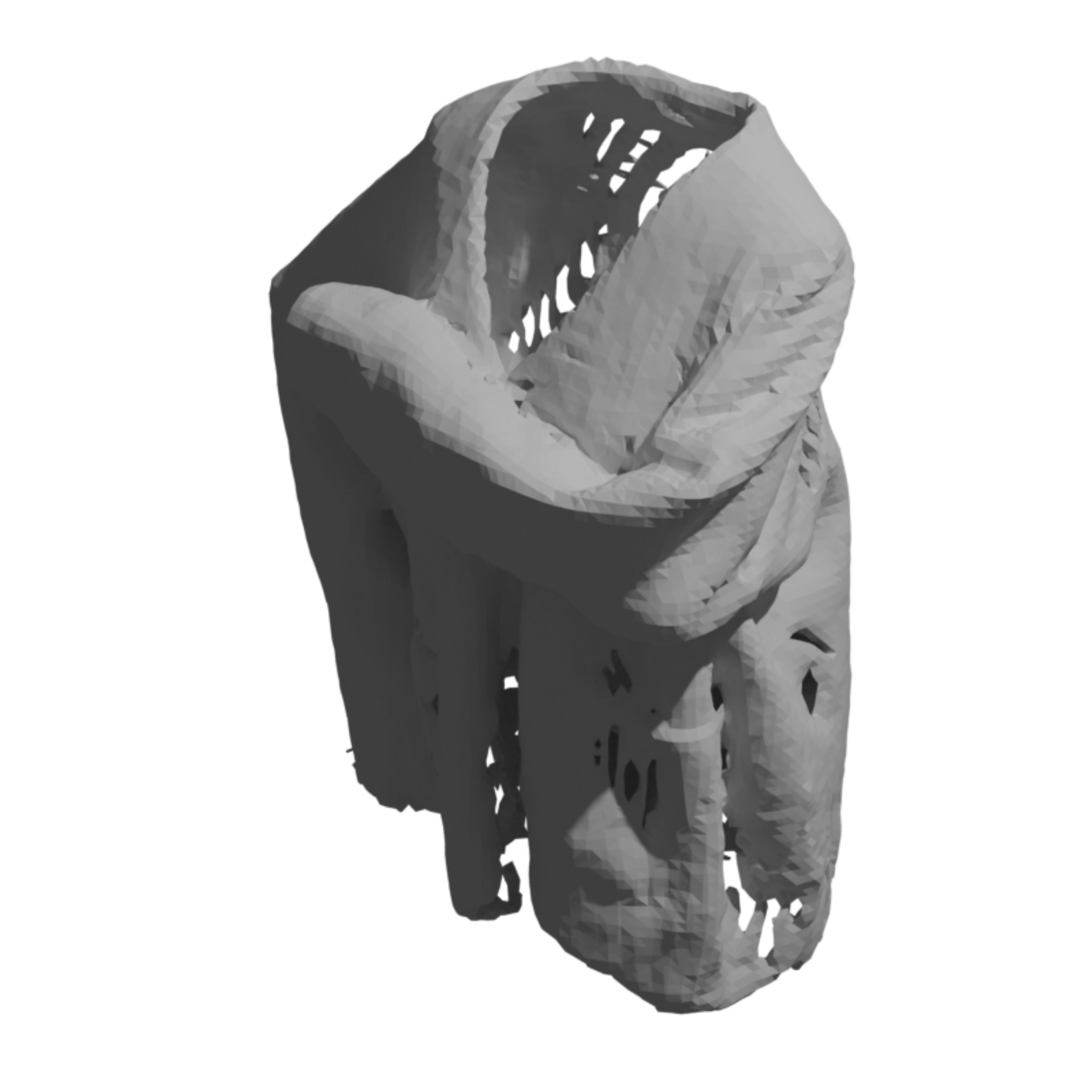}\\
    \includegraphics[width=.99\linewidth]{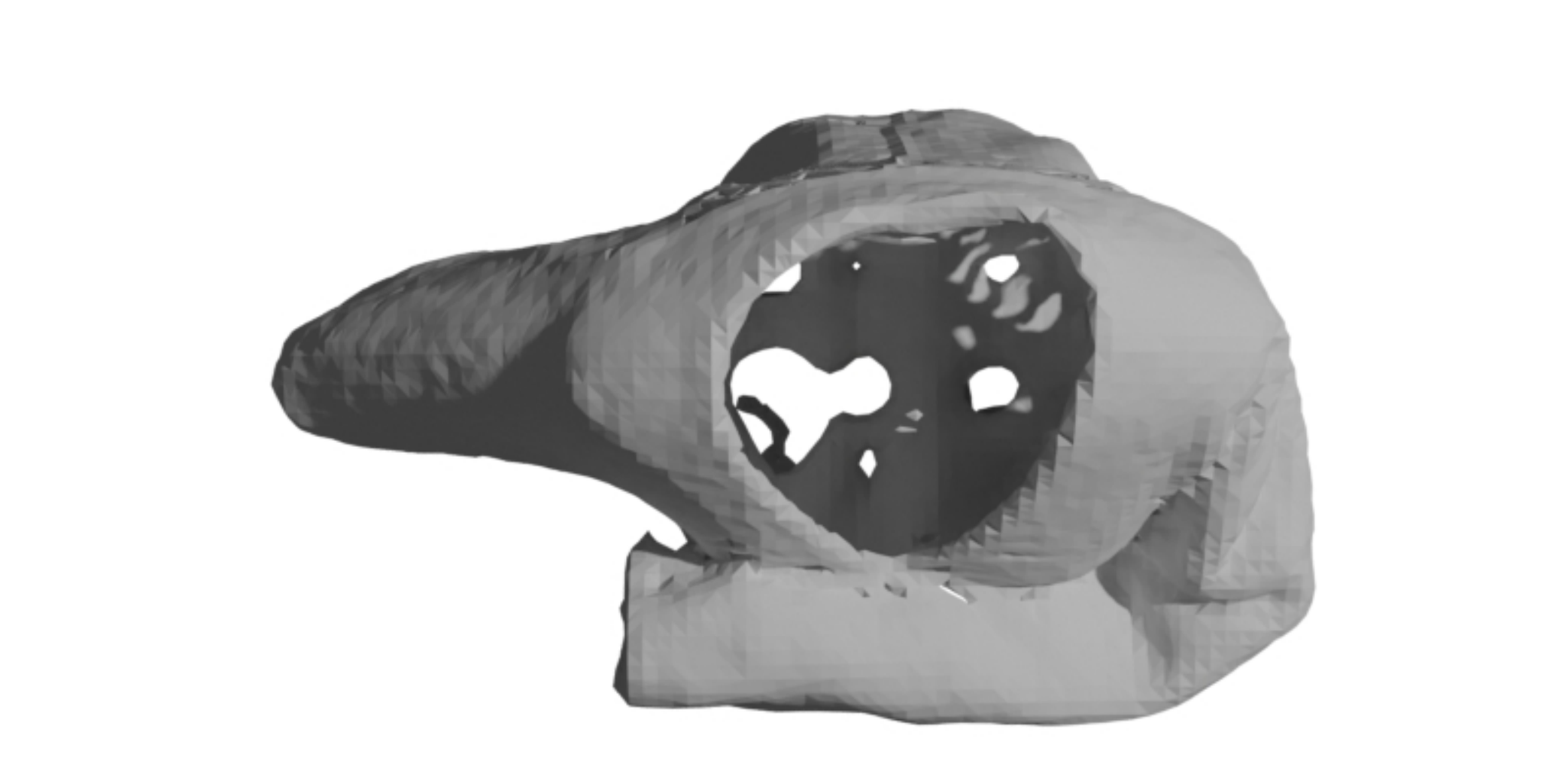}
\end{minipage}
\begin{minipage}[c]{.14\linewidth}
    \centering
    \includegraphics[width=.99\linewidth]{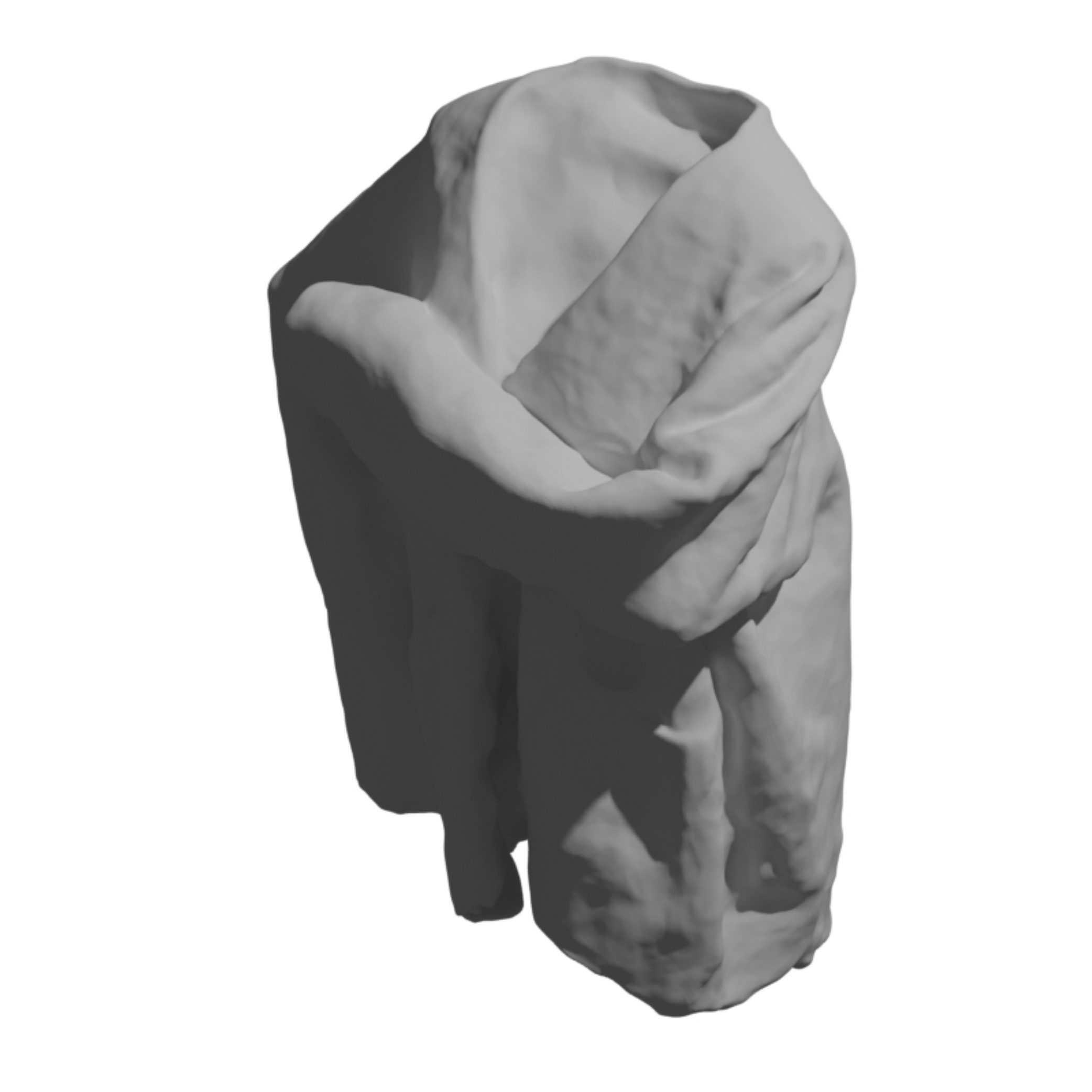}\\
    \includegraphics[width=.99\linewidth]{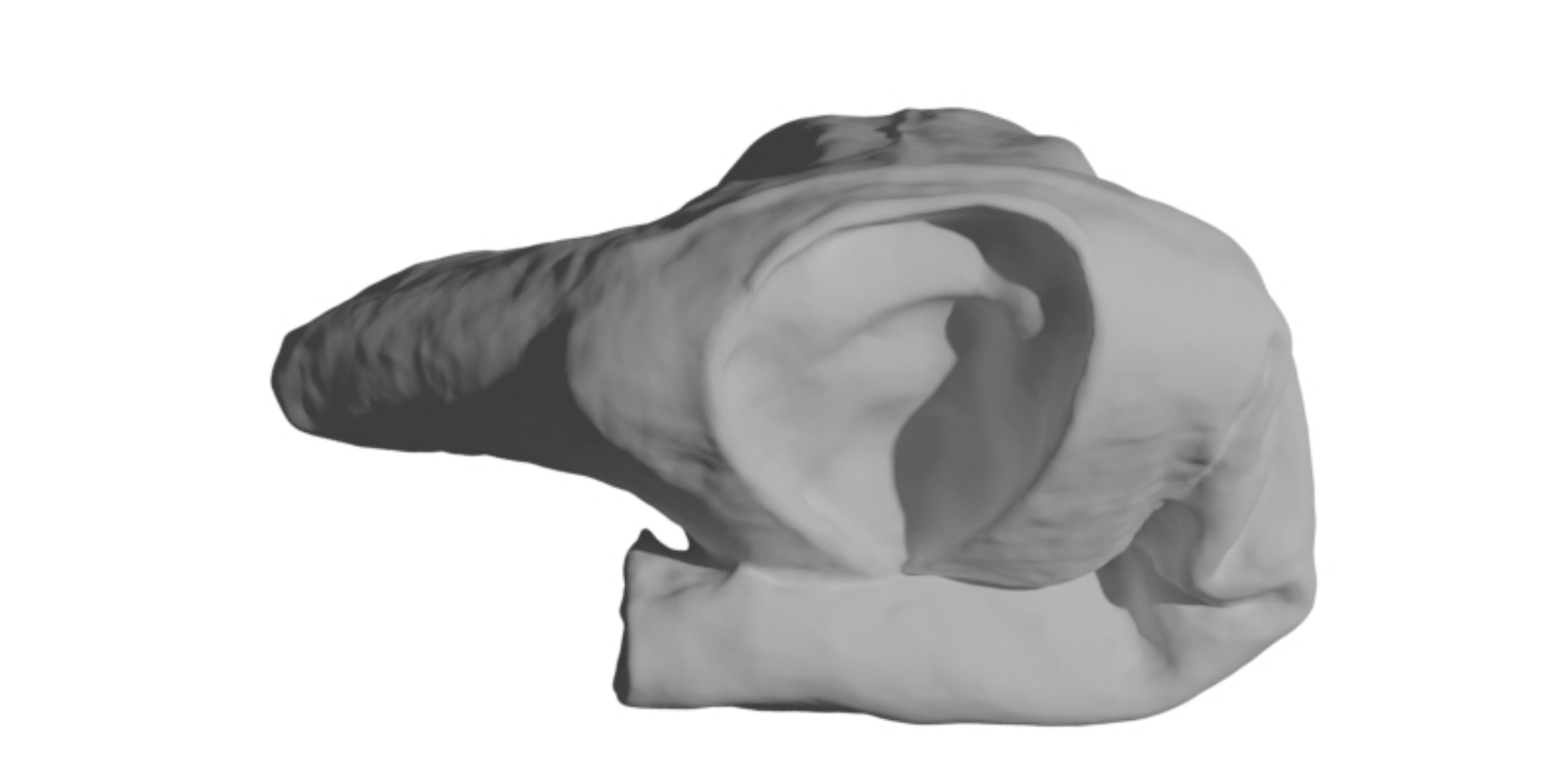}
\end{minipage}
\begin{minipage}[c]{.14\linewidth}
    \centering
    \includegraphics[width=.99\linewidth]{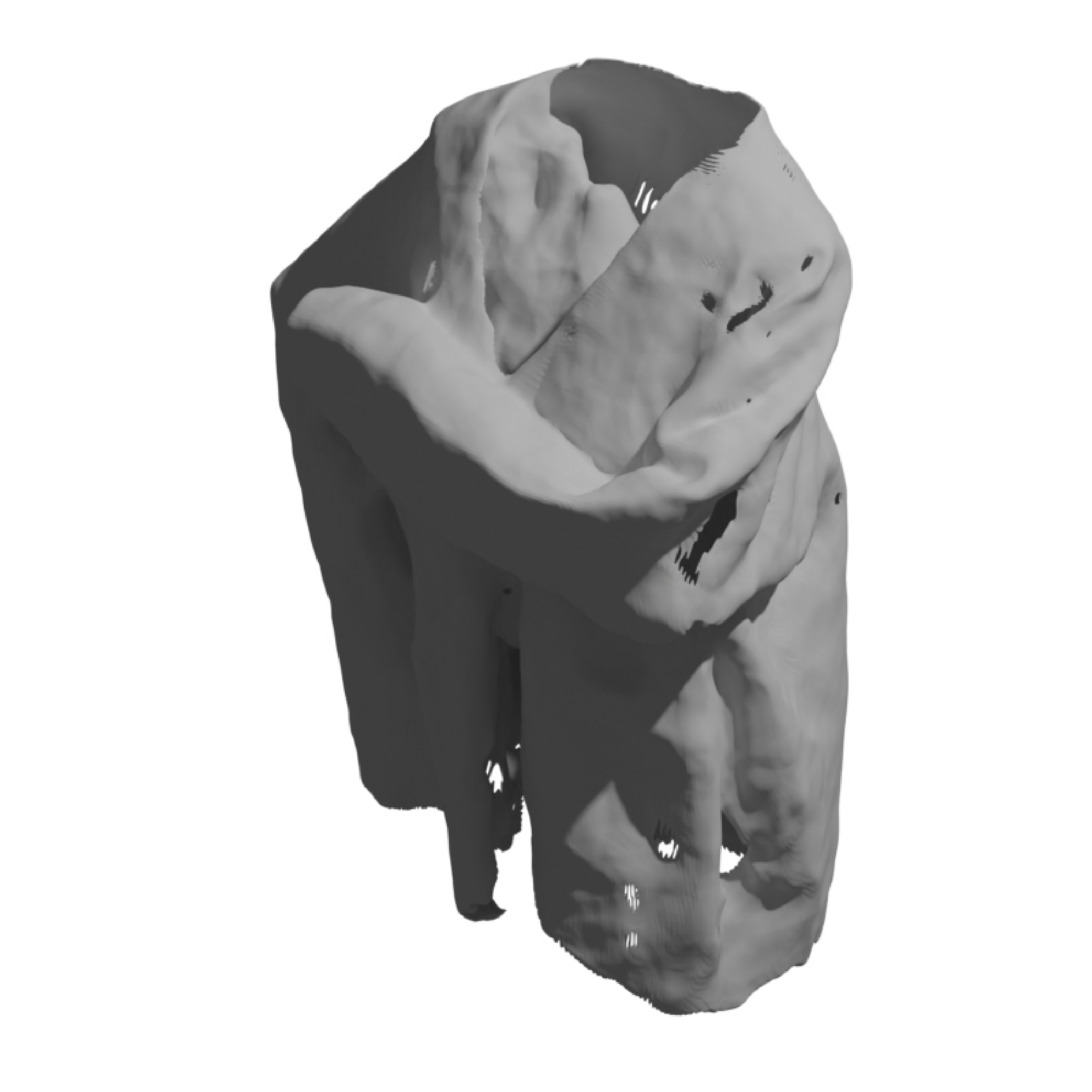}\\
    \includegraphics[width=.99\linewidth]{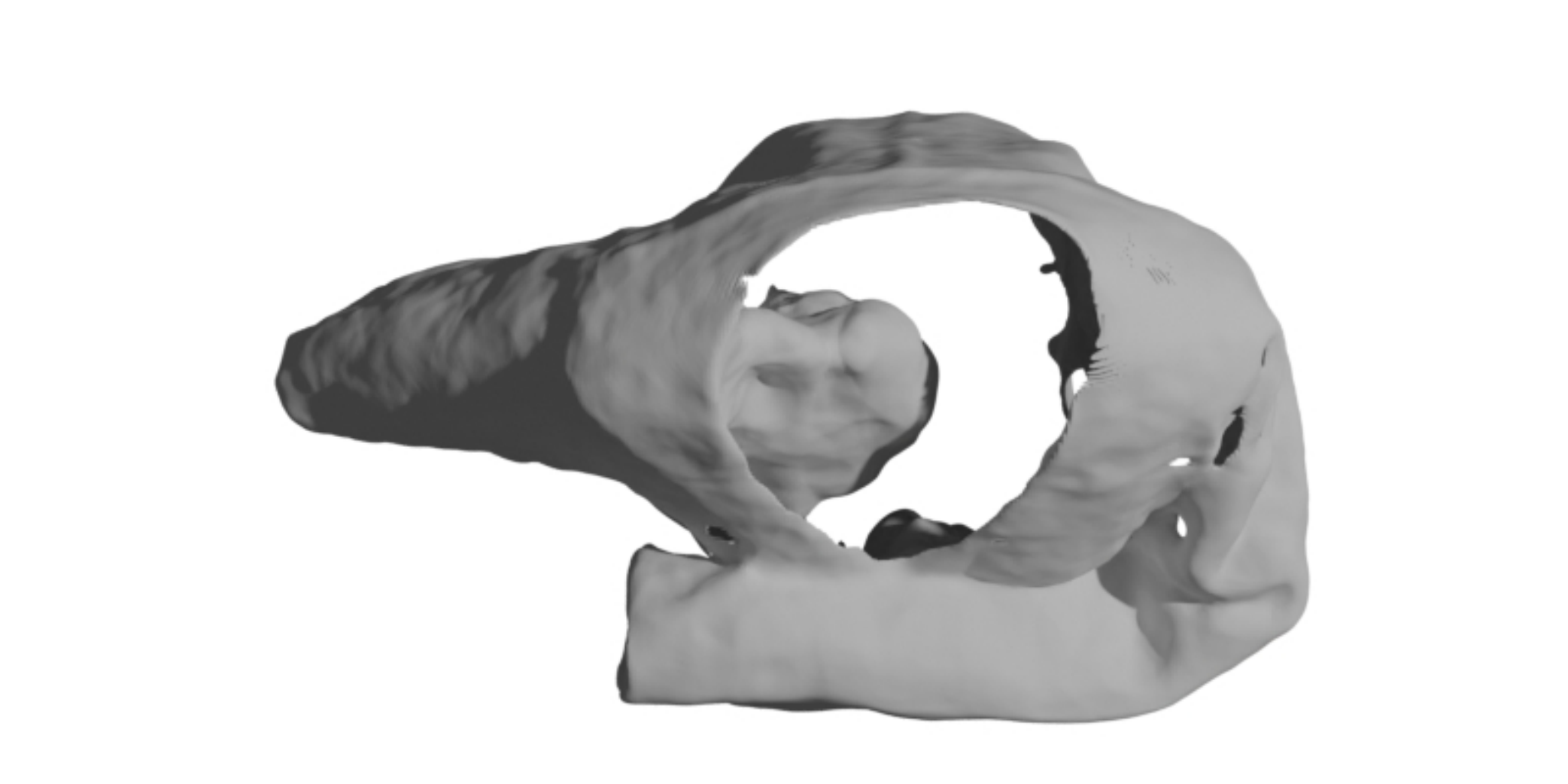}
\end{minipage}
\begin{minipage}[c]{.14\linewidth}
    \centering
    \includegraphics[width=.99\linewidth]{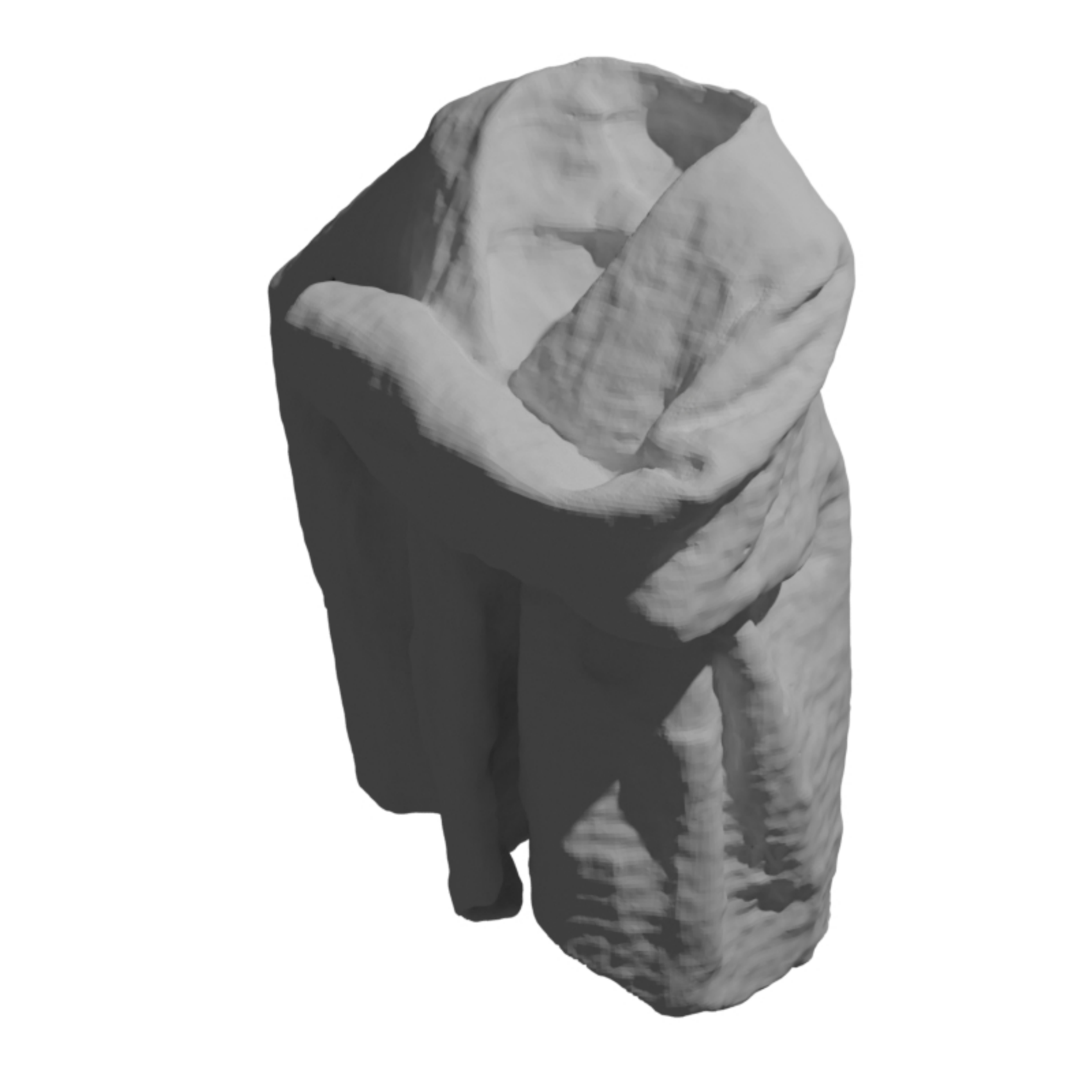}\\
    \includegraphics[width=.99\linewidth]{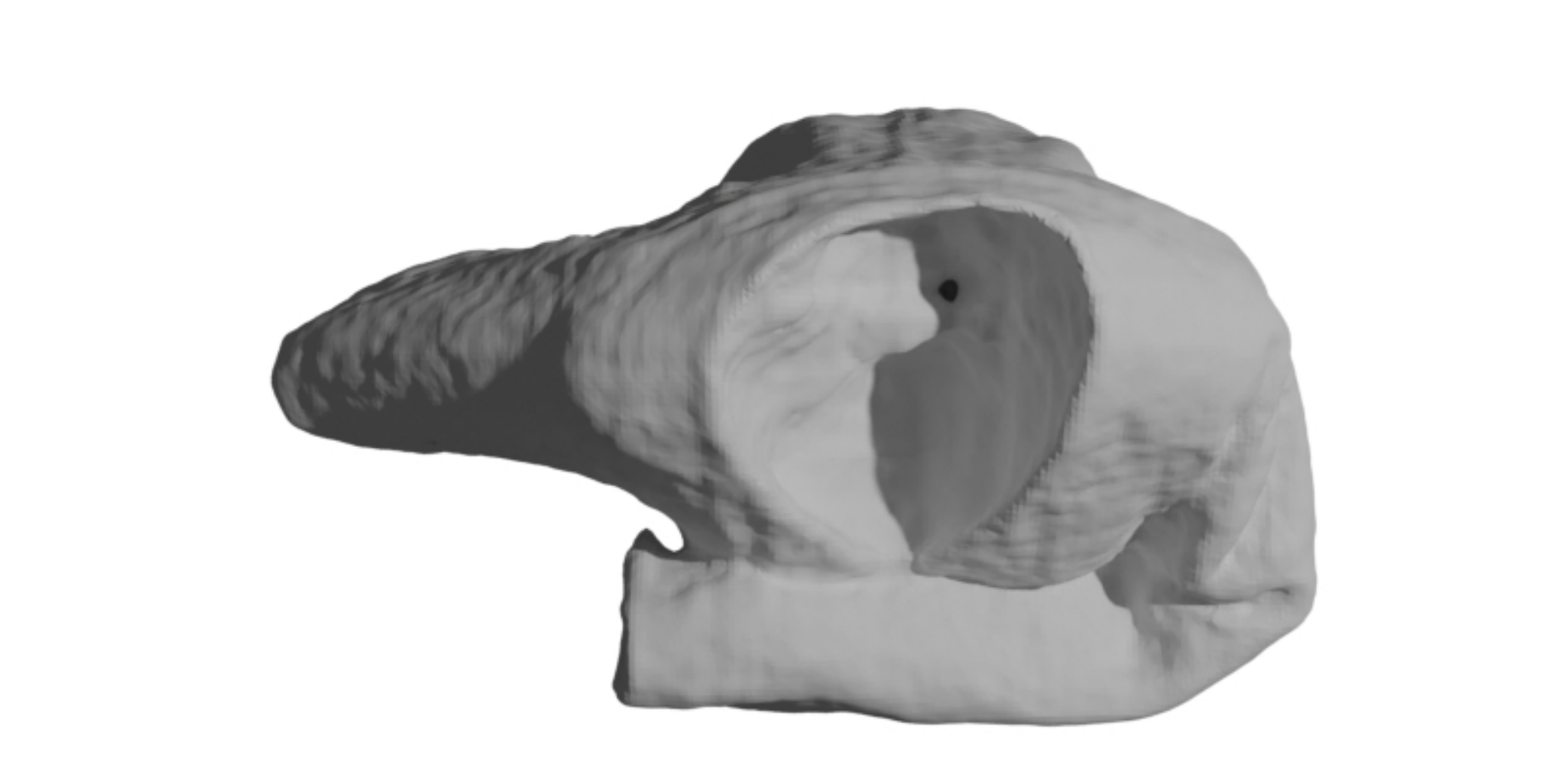}
\end{minipage}
\begin{minipage}[c]{.14\linewidth}
    \centering
    \includegraphics[width=.99\linewidth]{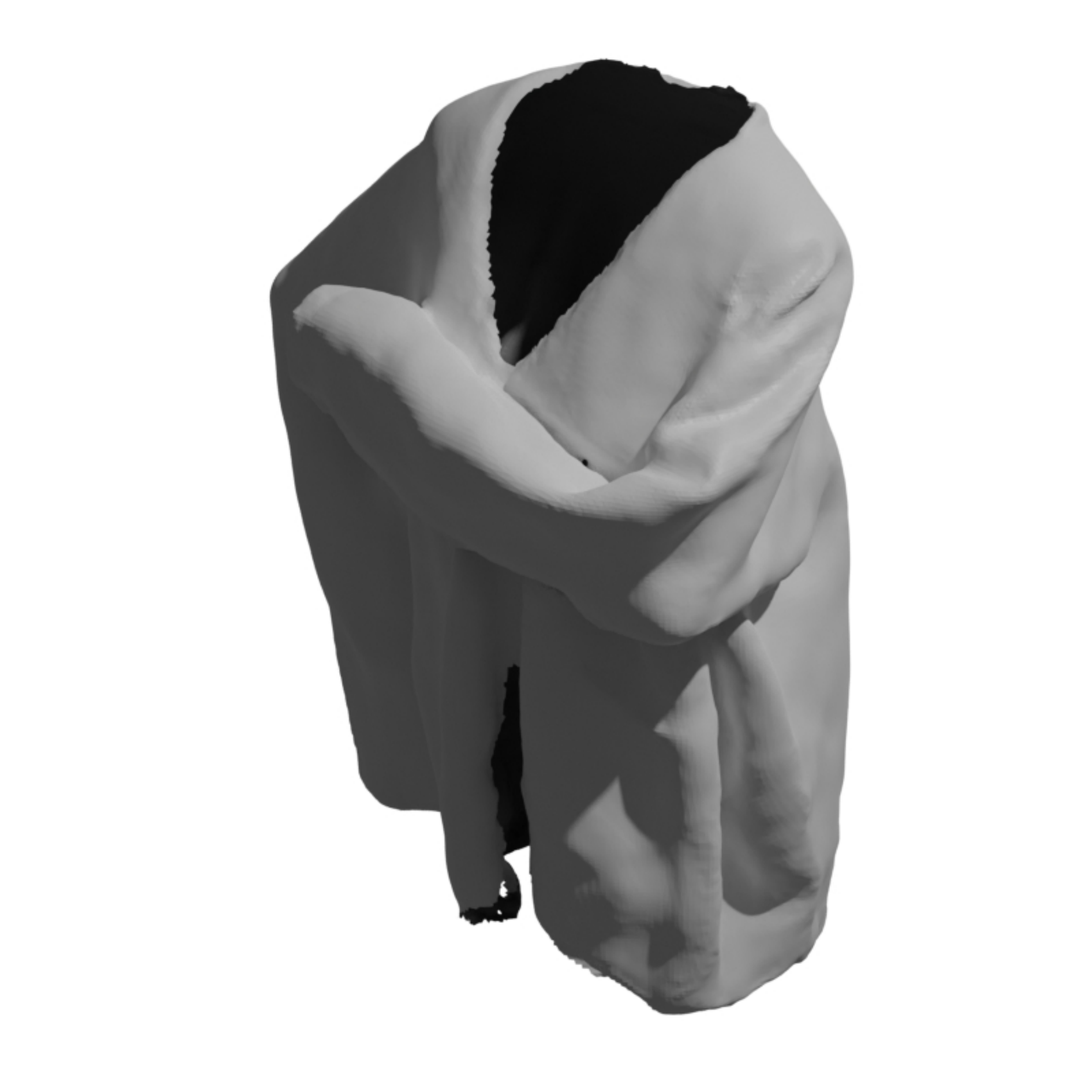}\\
    \includegraphics[width=.99\linewidth]{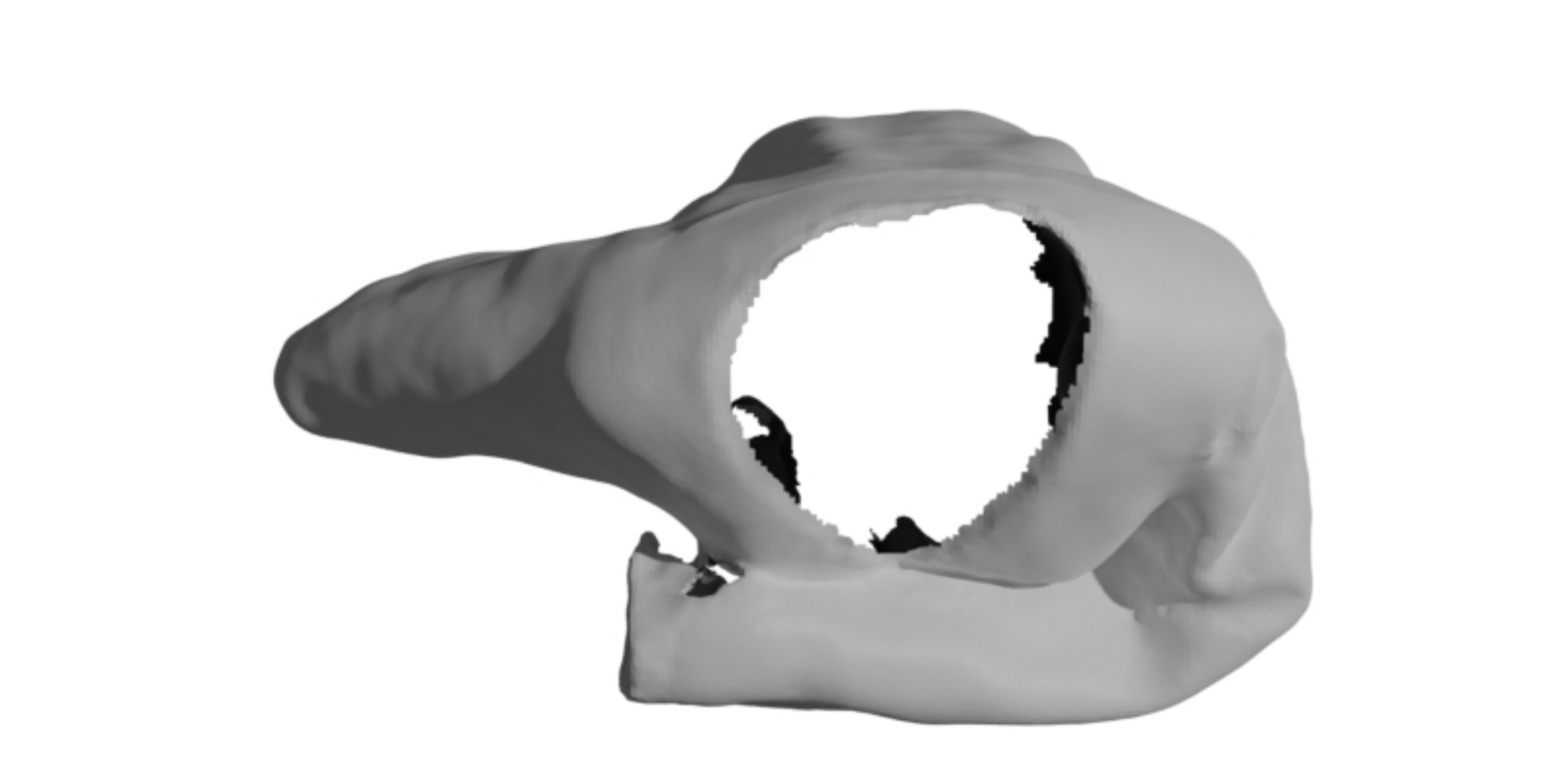}
\end{minipage}
\begin{minipage}[c]{.14\linewidth}
    \centering
    \includegraphics[width=.99\linewidth]{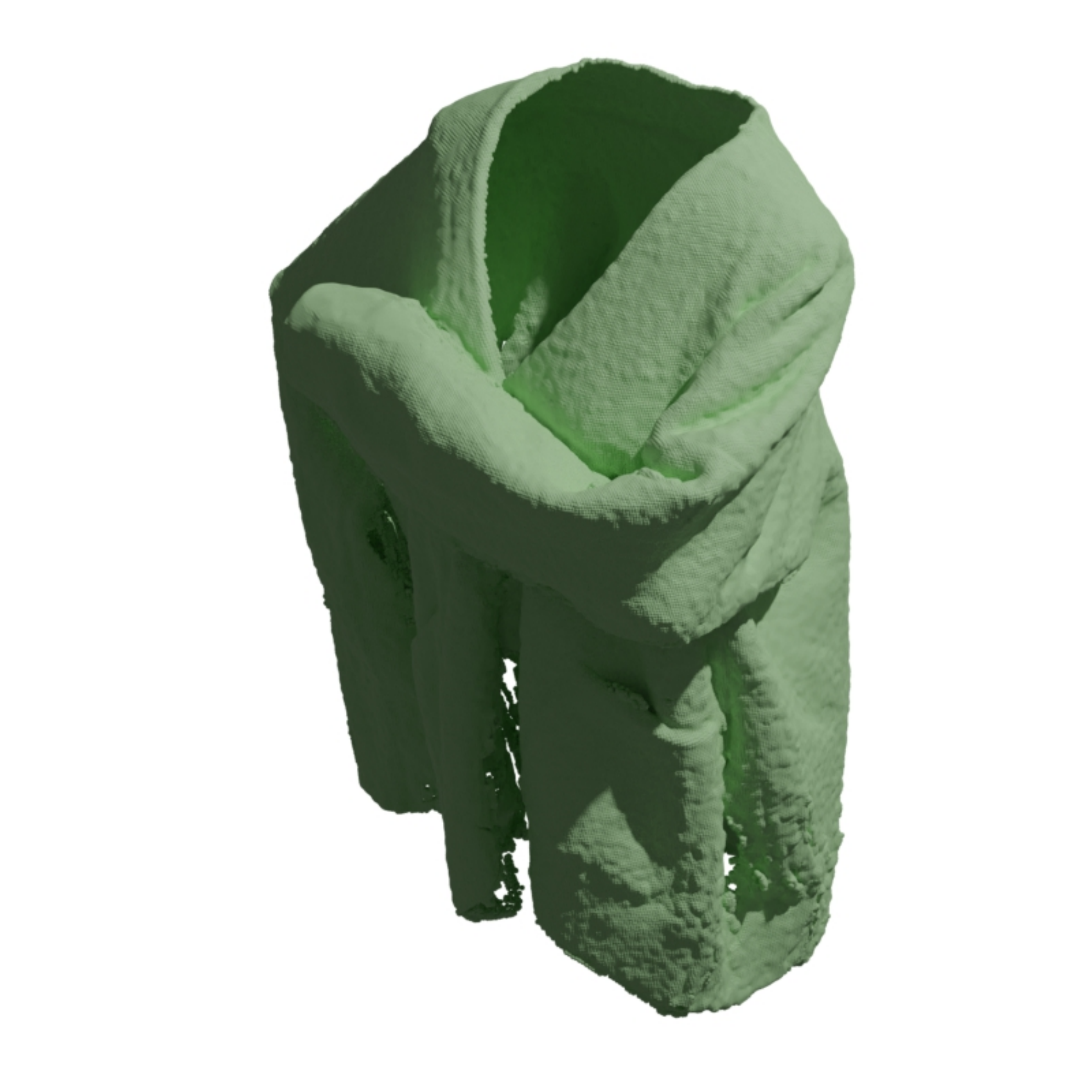}\\
    \includegraphics[width=.99\linewidth]{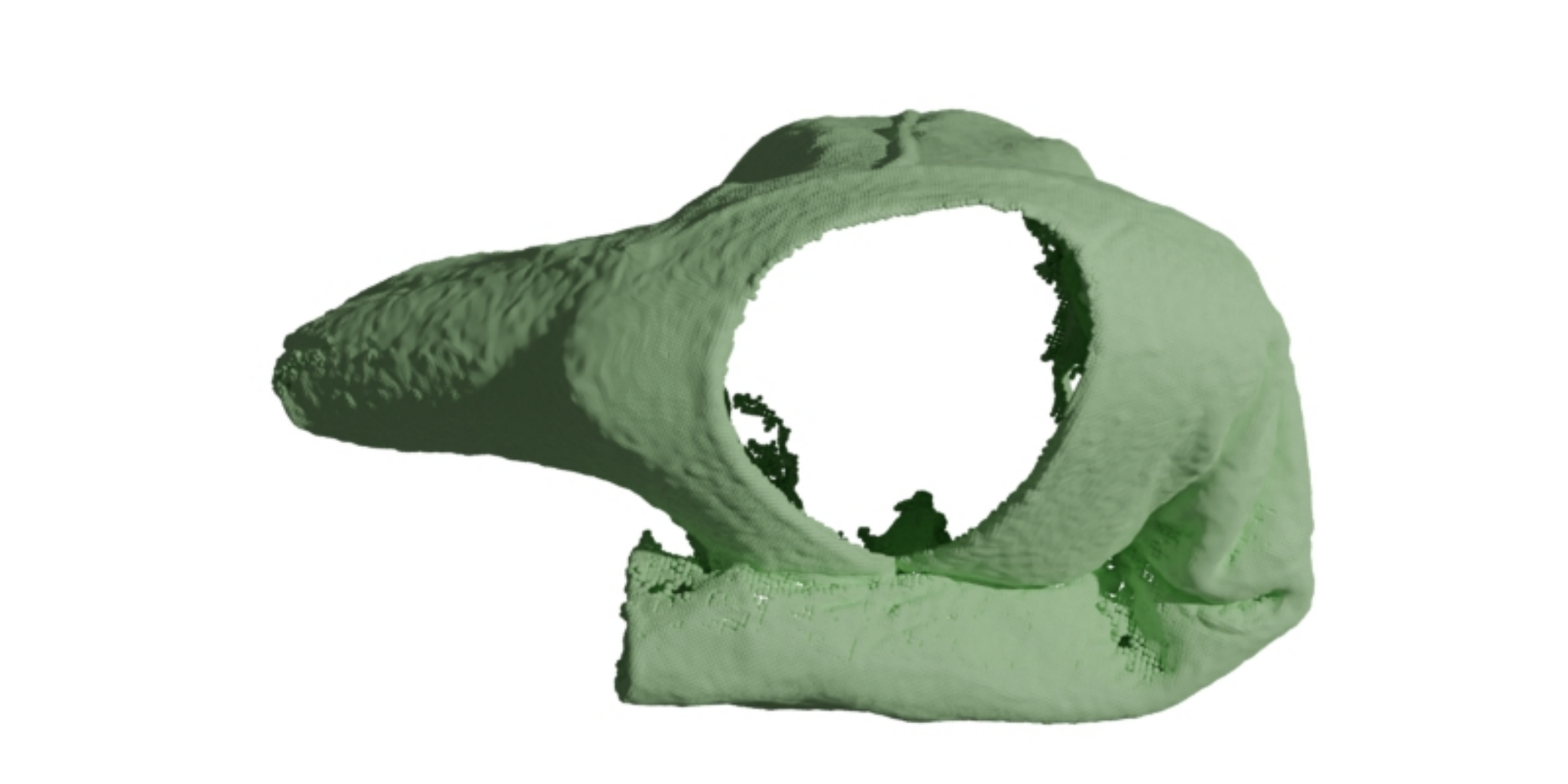}
\end{minipage}
\vspace{5mm}

\begin{minipage}[c]{.12\linewidth}
    \centering
    \includegraphics[width=.9\linewidth]{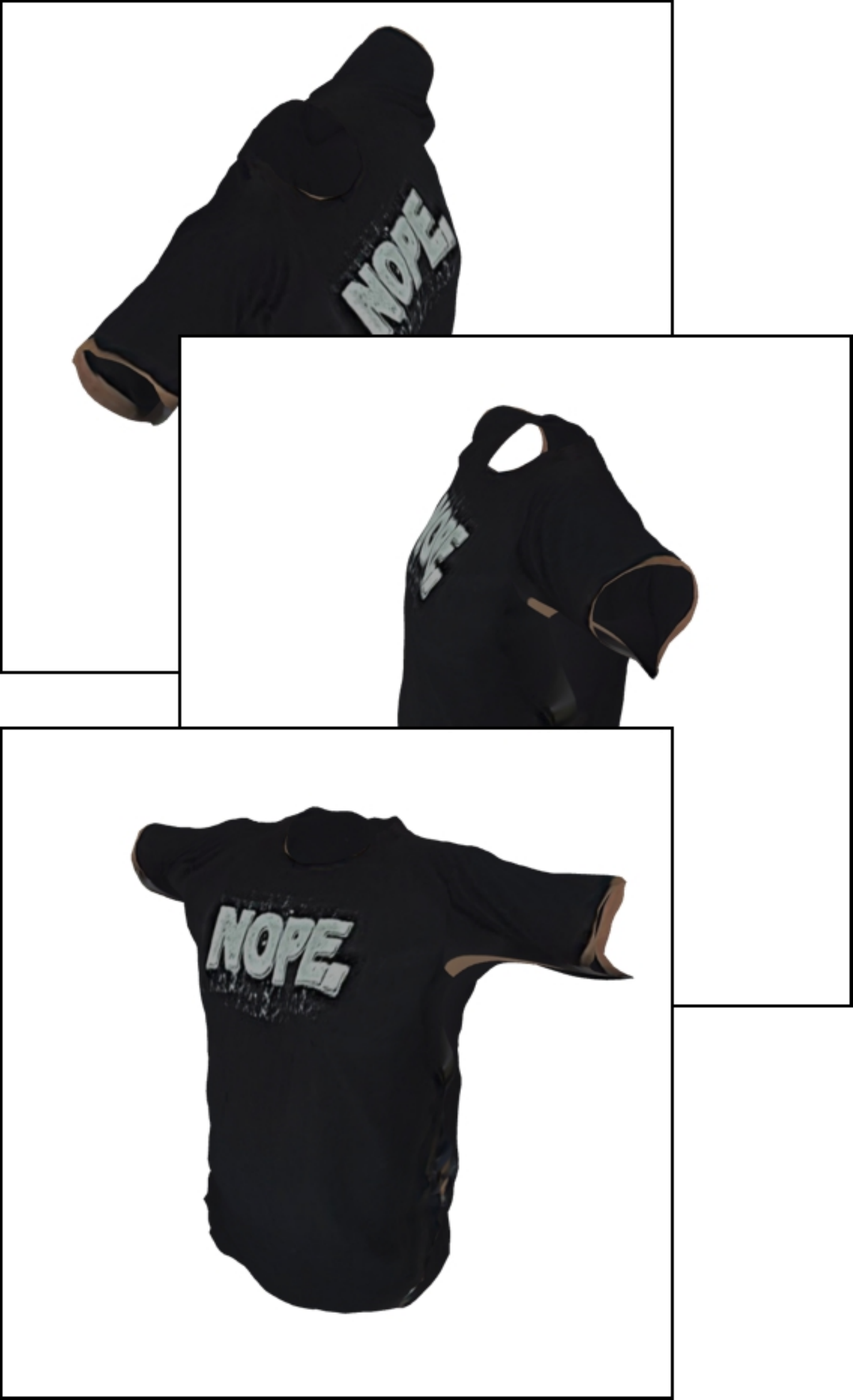}
\end{minipage}
\begin{minipage}[c]{.14\linewidth}
    \centering
    \includegraphics[width=.99\linewidth]{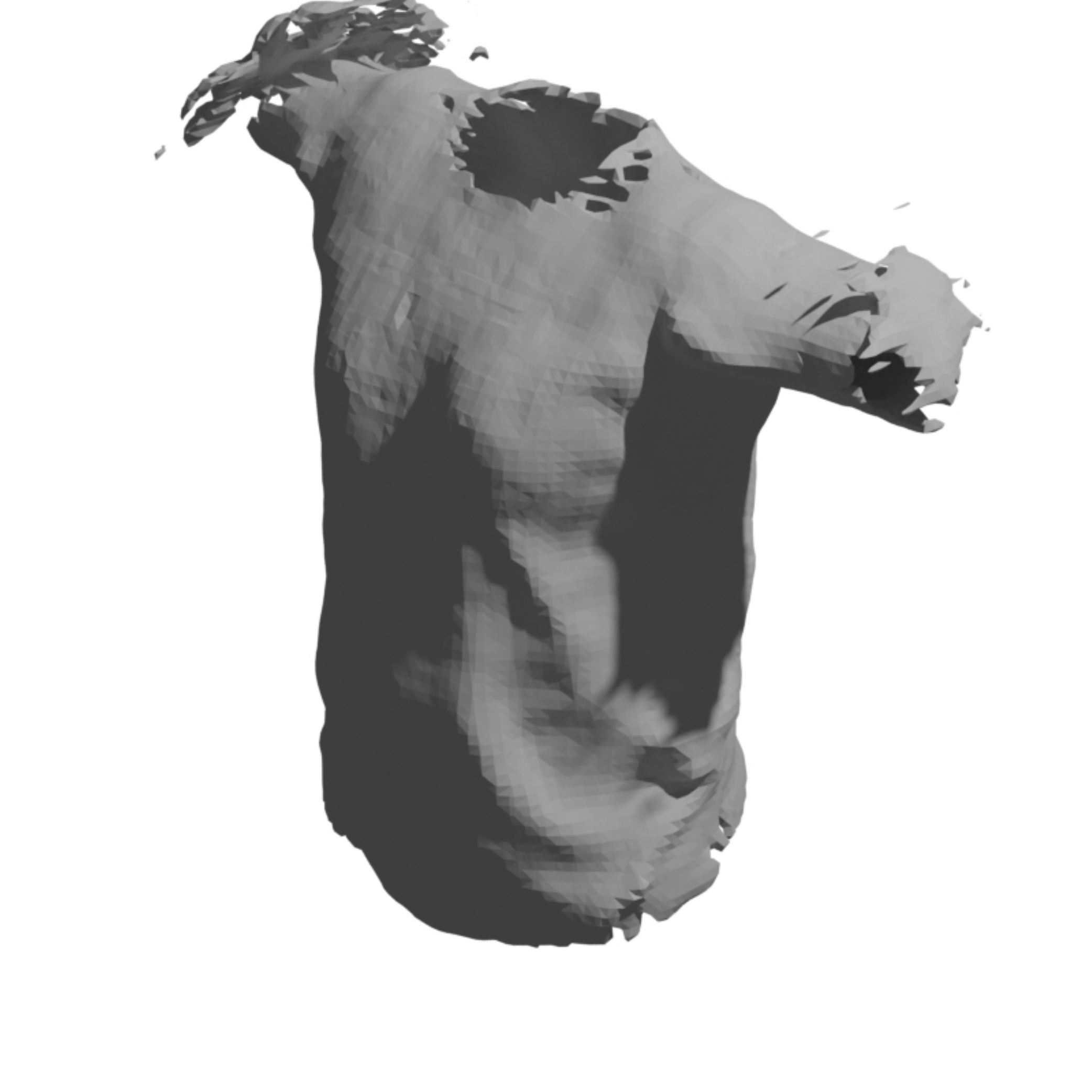}\\
    \includegraphics[width=.99\linewidth]{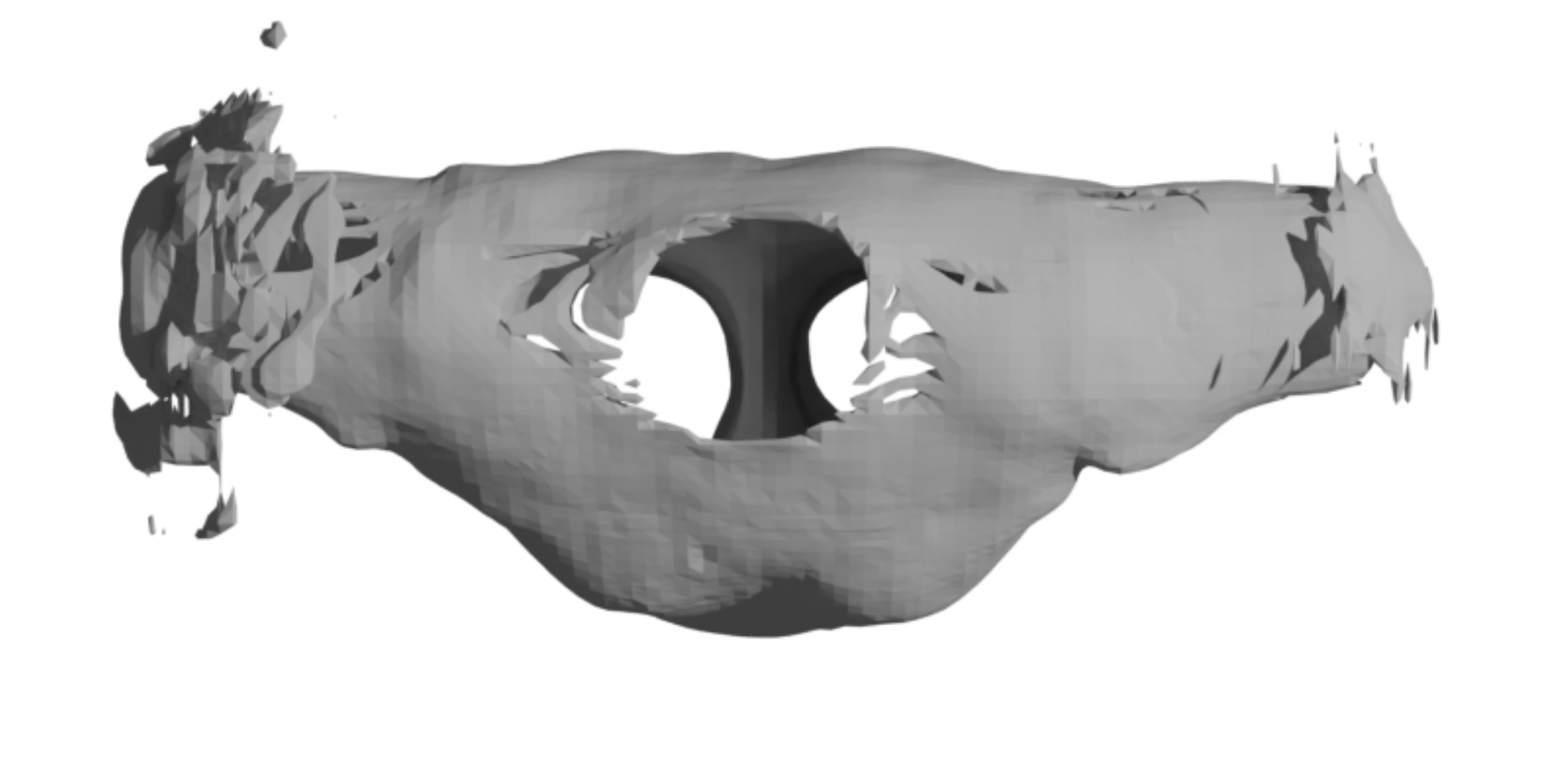}
\end{minipage}
\begin{minipage}[c]{.14\linewidth}
    \centering
    \includegraphics[width=.99\linewidth]{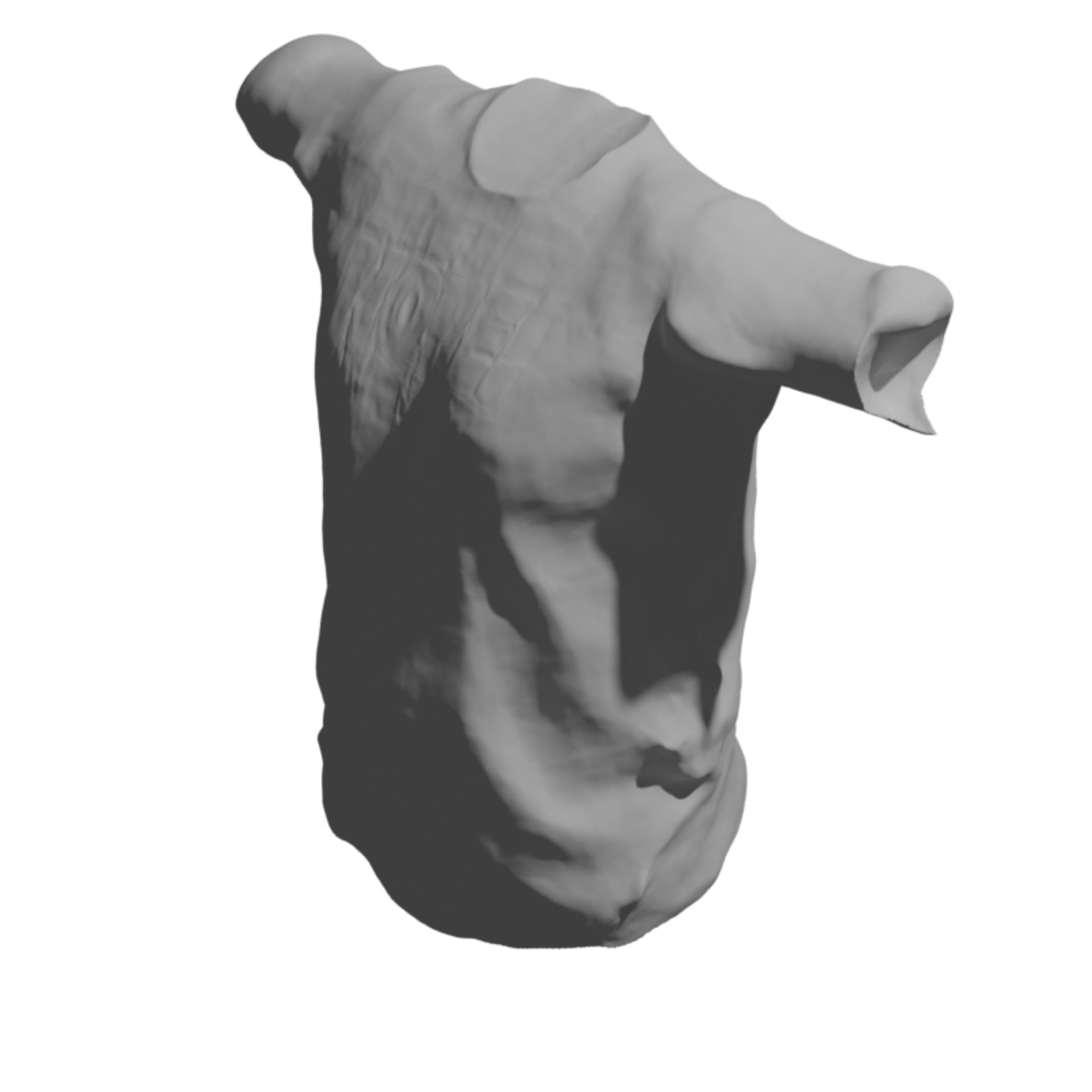}\\
    \includegraphics[width=.99\linewidth]{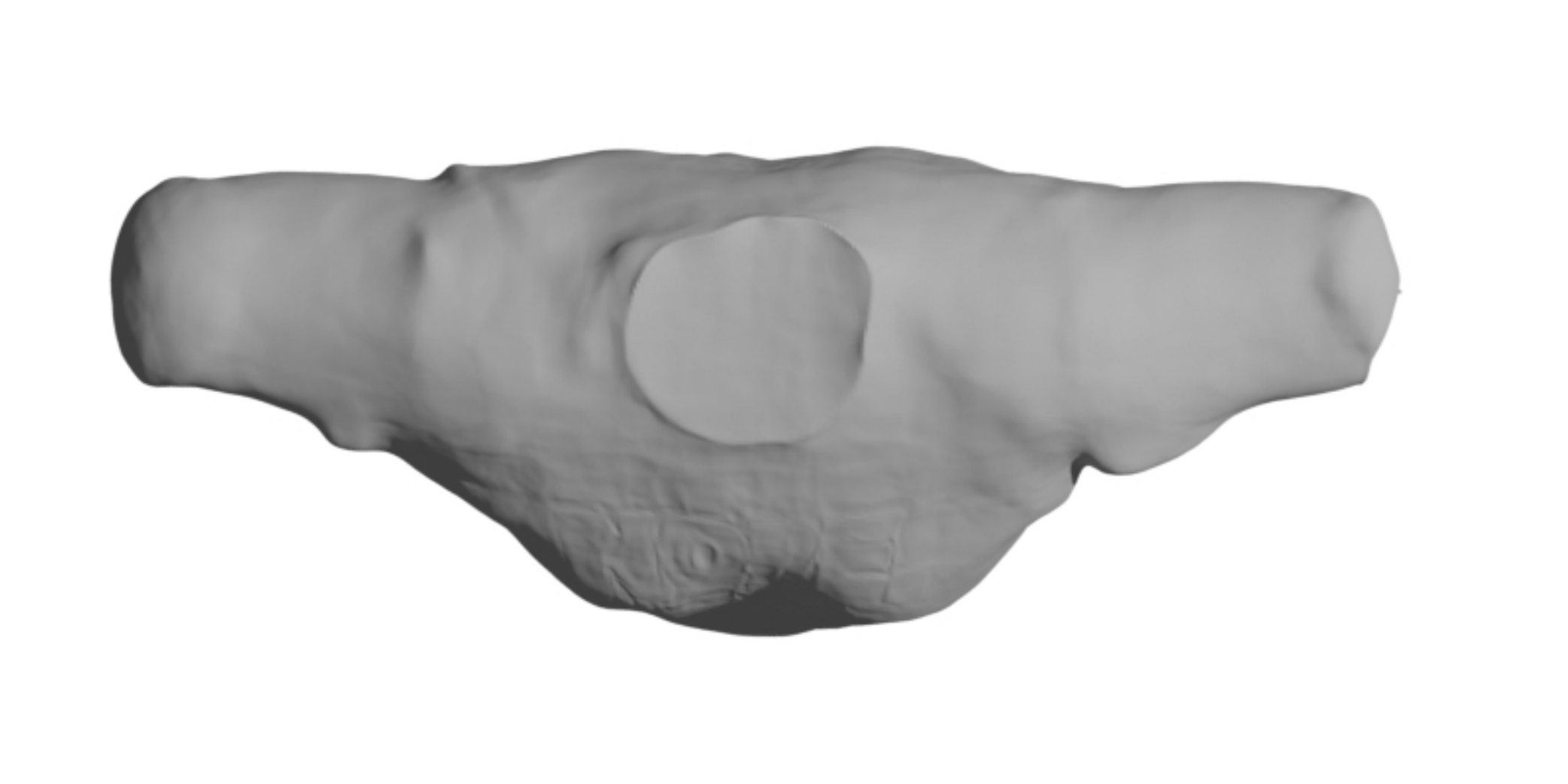}
\end{minipage}
\begin{minipage}[c]{.14\linewidth}
    \centering
    \includegraphics[width=.99\linewidth]{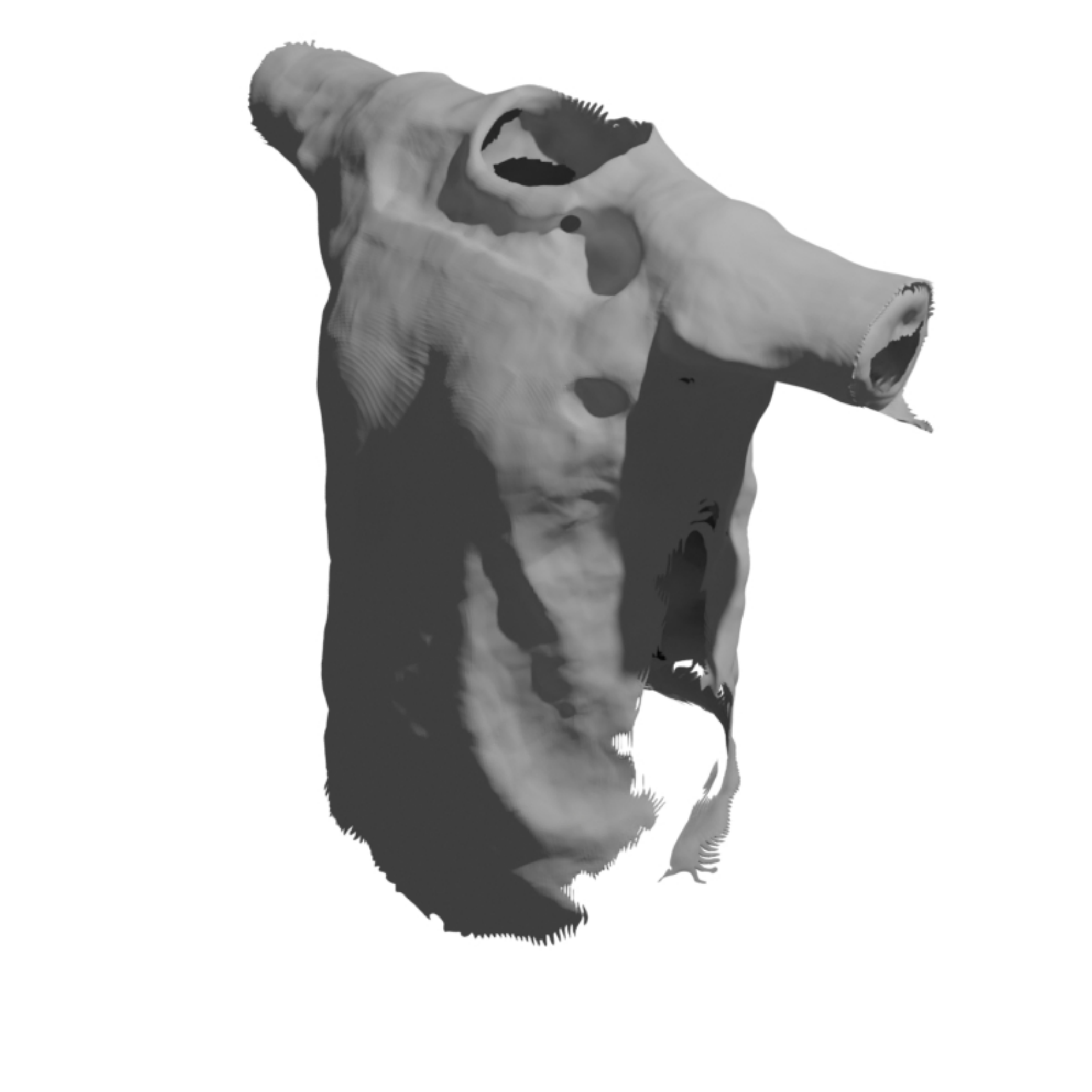}\\
    \includegraphics[width=.99\linewidth]{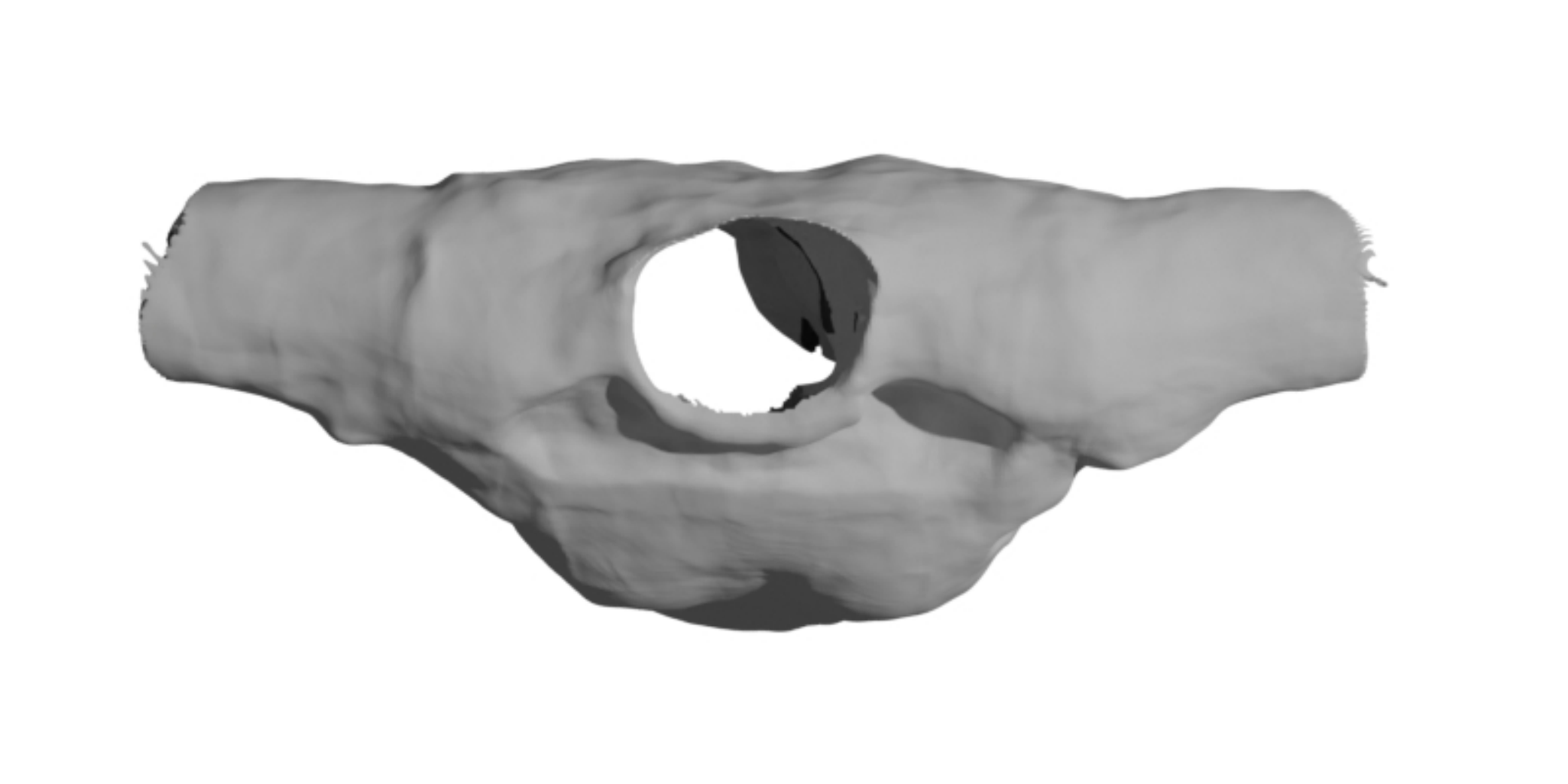}
\end{minipage}
\begin{minipage}[c]{.14\linewidth}
    \centering
    \includegraphics[width=.99\linewidth]{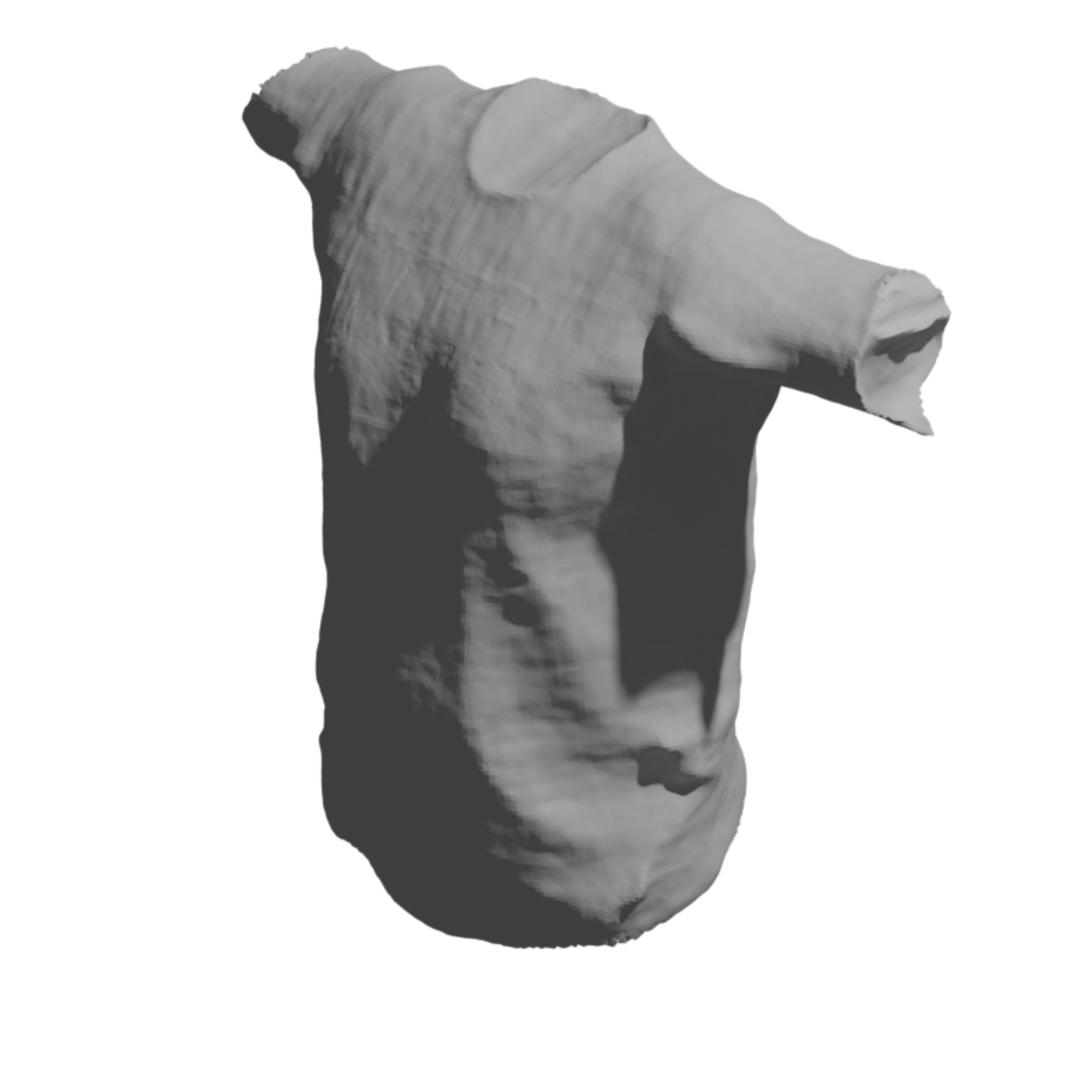}\\
    \includegraphics[width=.99\linewidth]{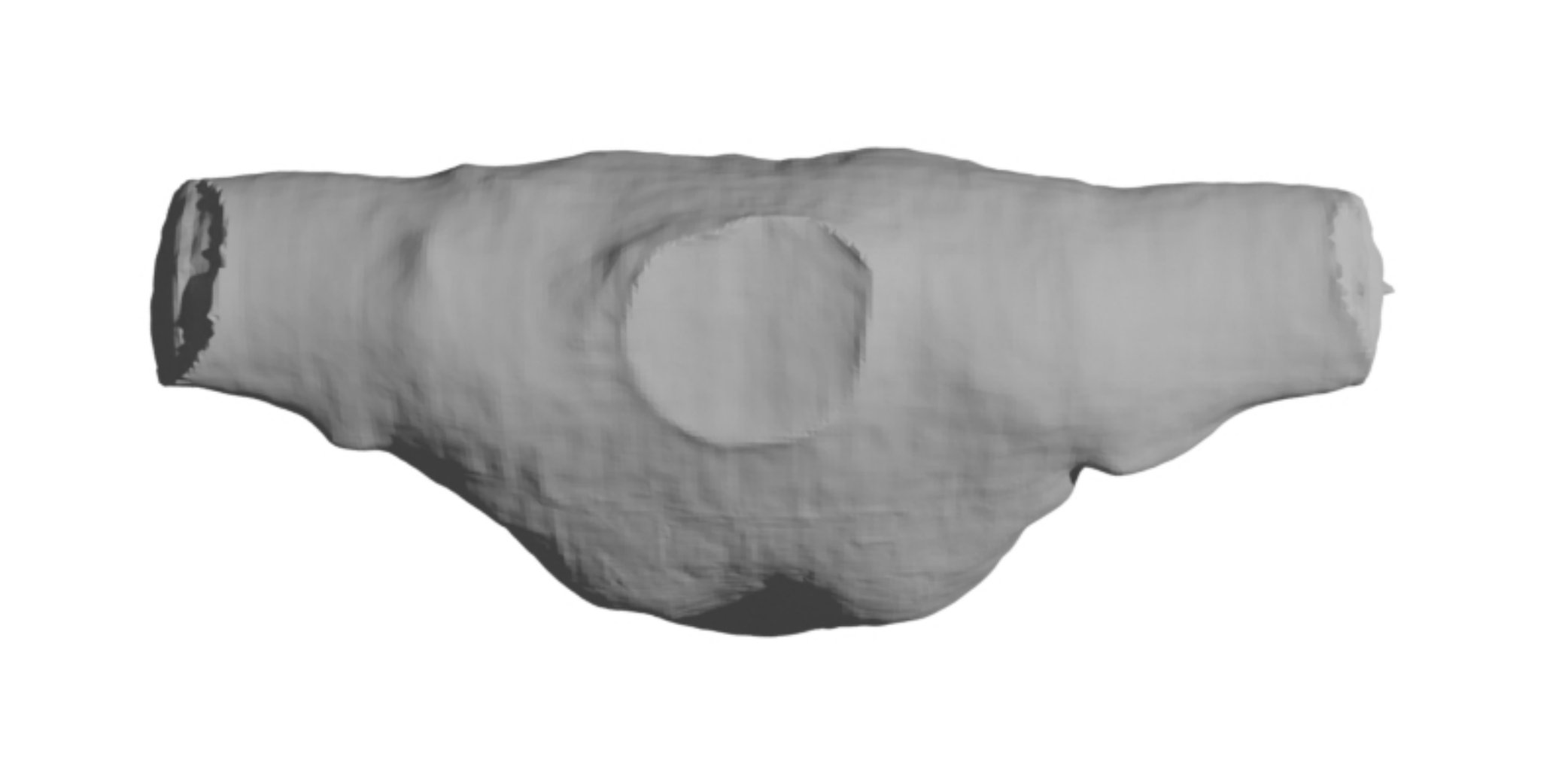}
\end{minipage}
\begin{minipage}[c]{.14\linewidth}
    \centering
    \includegraphics[width=.99\linewidth]{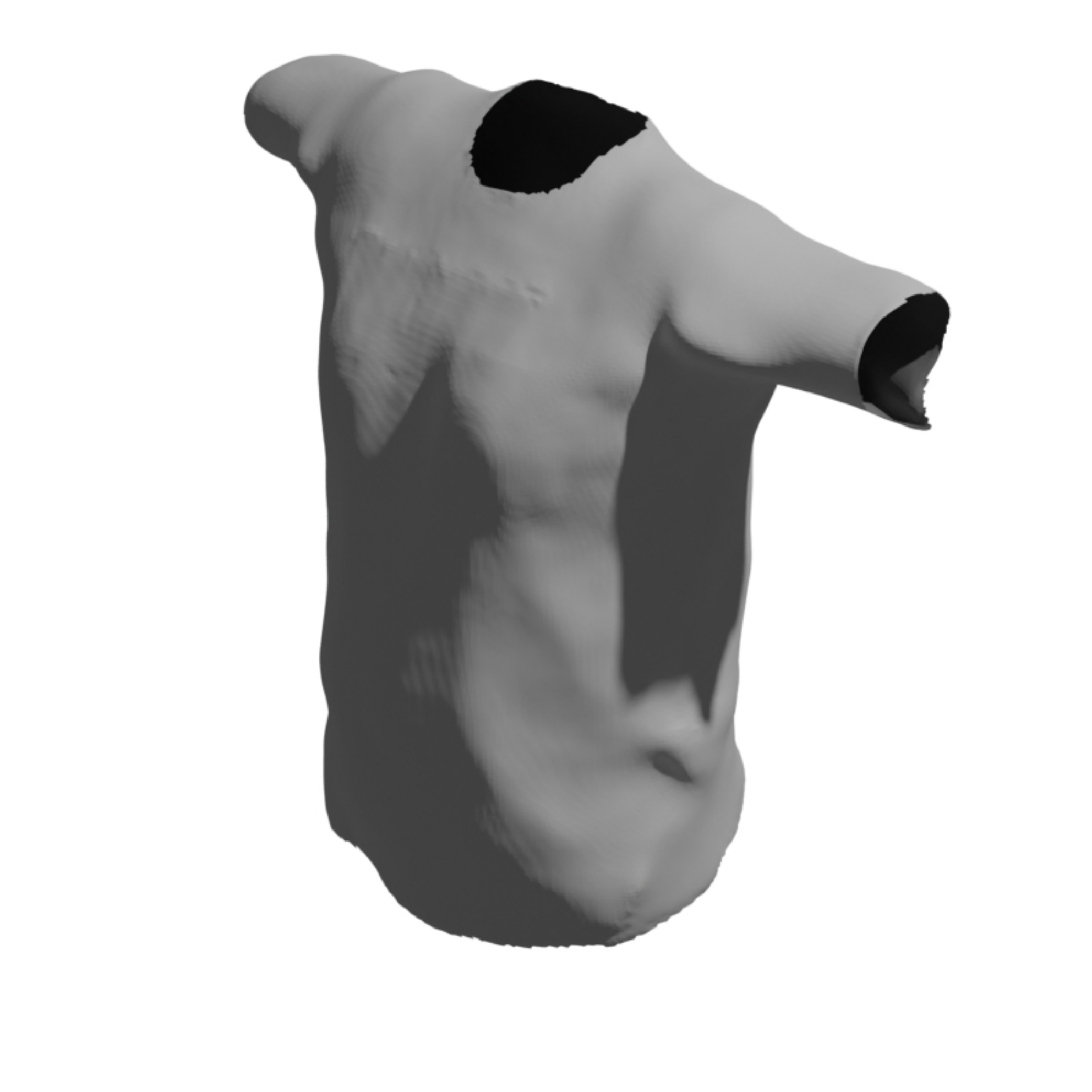}\\
    \includegraphics[width=.99\linewidth]{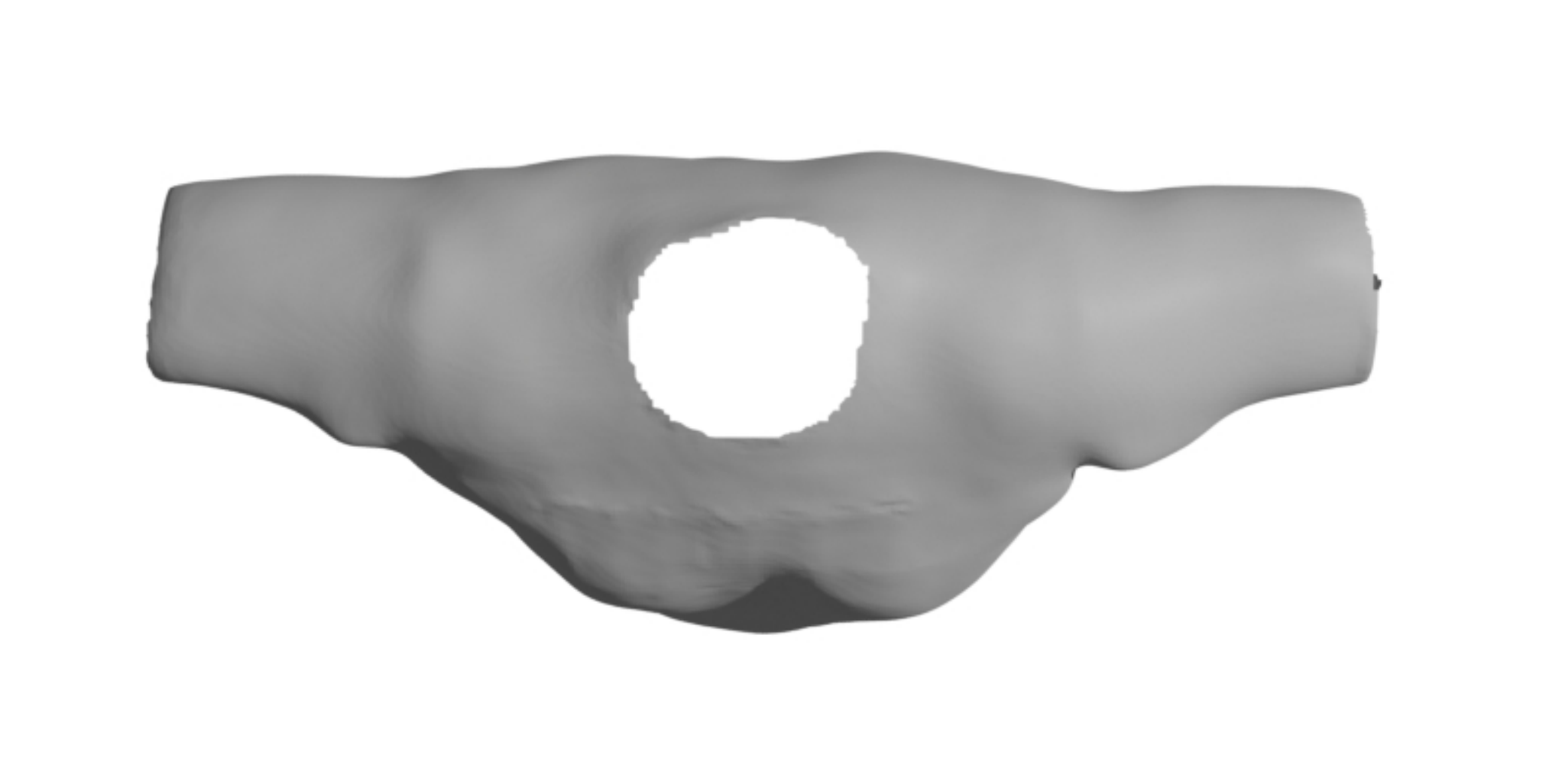}
\end{minipage}
\begin{minipage}[c]{.14\linewidth}
    \centering
    \includegraphics[width=.99\linewidth]{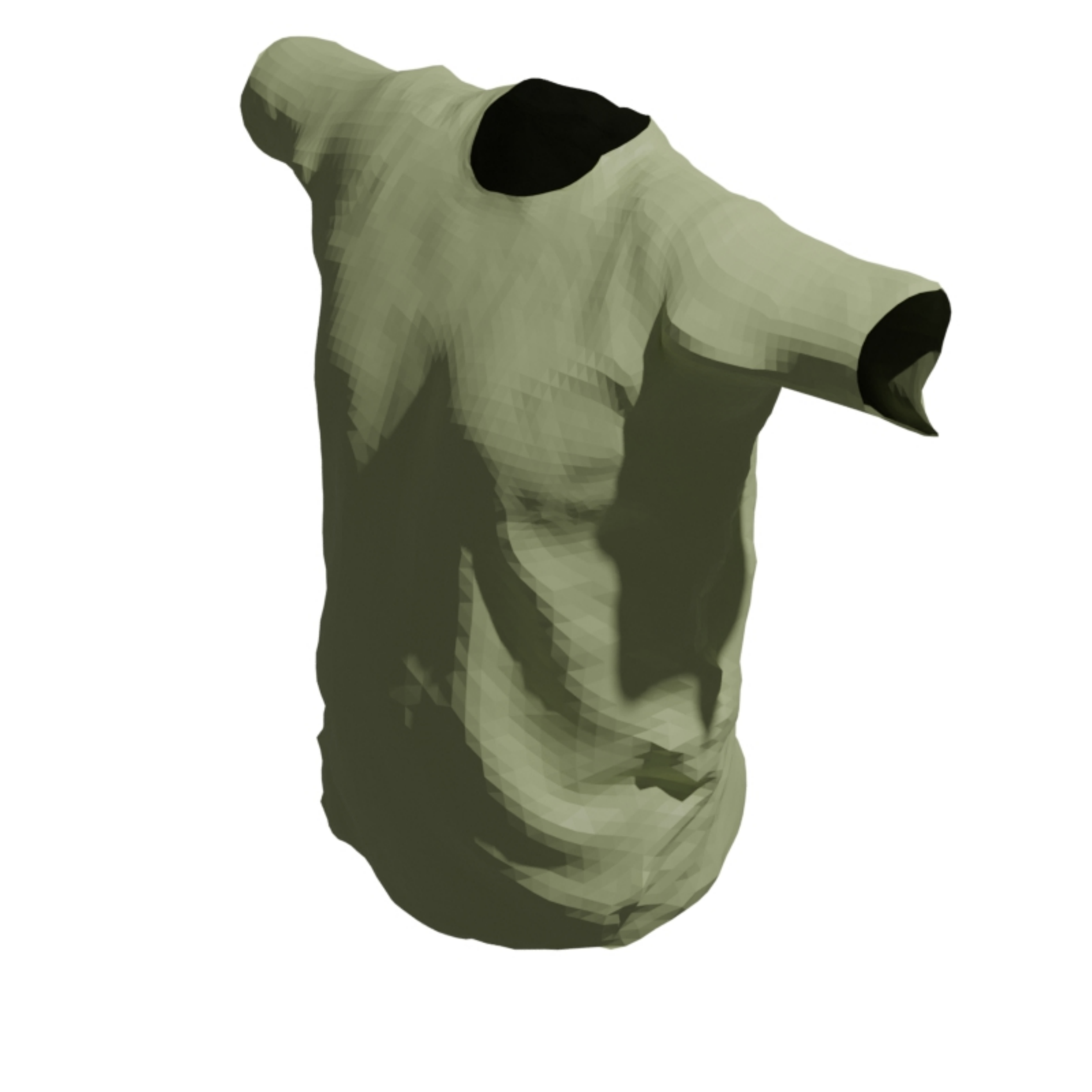}\\
    \includegraphics[width=.99\linewidth]{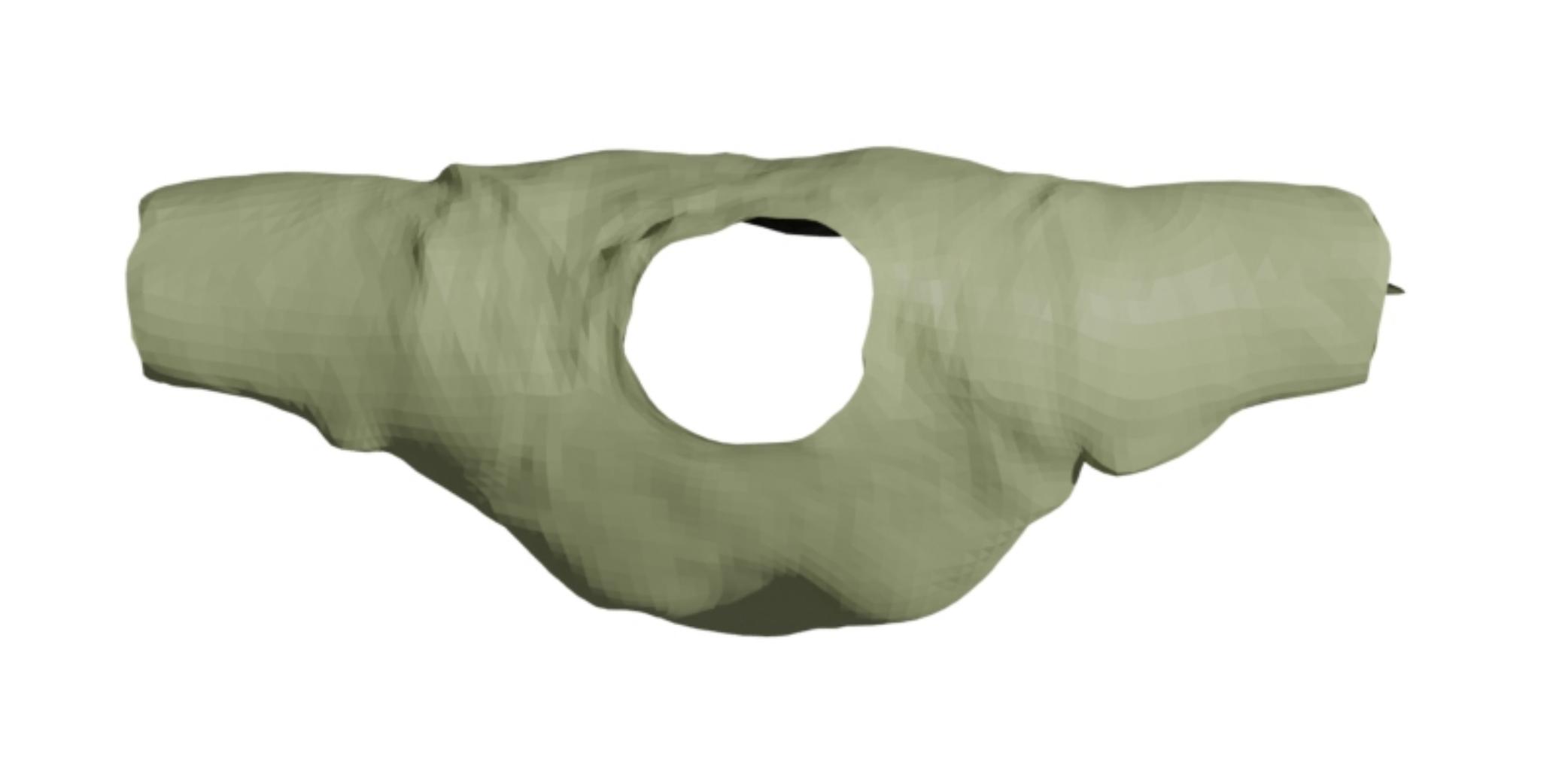}
\end{minipage}

\begin{minipage}[c]{.12\linewidth}
    \centering
    \subcaption{Input}
\end{minipage}
\begin{minipage}[c]{.14\linewidth}
    \centering
    \subcaption{IDR}
\end{minipage}
\begin{minipage}[c]{.14\linewidth}
    \centering
    \subcaption{NeuS}
\end{minipage}
\begin{minipage}[c]{.14\linewidth}
    \centering
    \subcaption{NeuralWarp}
\end{minipage}
\begin{minipage}[c]{.14\linewidth}
    \centering
    \subcaption{HF-NeuS}
\end{minipage}
\begin{minipage}[c]{.14\linewidth}
    \centering
    \subcaption{Ours}
\end{minipage}
\begin{minipage}[c]{.14\linewidth}
    \centering
    \subcaption{GT}
\end{minipage}
\caption{
Qualitative comparison with IDR~\cite{DBLP:conf/nips/YarivKMGABL20}, NeuS~\cite{DBLP:conf/nips/WangLLTKW21}, NeuralWarp~\cite{darmon2022improving} and HF-NeuS~\cite{wang2022hfneus}. 
The GT of DF3D~\cite{zhu2020deep} data is point cloud (green) and the GT of MGN~\cite{DBLP:conf/iccv/BhatnagarTTP19} data is open mesh (yellow). 
The back faces of the open surfaces are rendered in deep colors. 
The baselines are limited by the SDF representation and incorporate erroneous prior of closed surfaces. 
In contrast, our results can reconstruct complicated high-fidelity surfaces with open boundaries thanks to the UDF representation. 
}
\label{fig:comp-vis}
\end{figure*}

\section{Experiments \& Evaluations}\label{sec:exp}
In this section, 
we validate \OurNetName{} on multi-view reconstruction task qualitatively and quantitatively and further tested our method for real scenes. The experiments demonstrate that \OurNetName{} outperforms the state-of-the-art techniques and can successfully reconstruct complex shapes with open boundaries. 
Lastly, we perform ablation studies and further discussions to demonstrate the importance of each key design.

\subsection{Experimental Setup}

\noindent\textbf{Datasets.}
Since our method mainly focuses on open surface reconstruction under multi-view supervision, we perform our experiments on three commonly used datasets, including Multi-Garment Net dataset (MGN)~\cite{mgn}, Deep Fashion3D dataset (DF3D)~\cite{zhu2020deep}, and DTU MVS dataset(DTU)~\cite{jensen2014large}.
For DTU MVS dataset, Each scene contains 49 or 64 images at $1600 \times 1200$ resolution and masks are from IDR~\cite{DBLP:conf/nips/YarivKMGABL20}. 
And the DF3D and MGN contain some real-scanned garments with open boundaries, which are rendered as 200 colored images with $800 \times 800$ resolution for reconstruction. 
We respectively sampled 18 and 10 shapes from different categories for the two datasets.
For the detailed camera poses, please refer to the supplementary document.
Furthermore, we also collect some complex shapes with non-watertight structures\footnote{https://downloadfree3d.com/,~https://archive3d.net/} and rendered them to evaluate our framework.
These shapes contain more intricate structures, which are composed of surfaces with open boundaries, \eg plant leaves, and hollow structures (Fig.~\ref{fig:teaser}).
Some datasets with diverse shapes (\emph{e.g.} BMVS, Mixamo, and some real captured objects) are also tested. 

\vspace{-2mm}
\paragraph{Baselines.}
We compare \OurNetName{} with several baselines for multi-view reconstruction task, including COLMAP~\cite{DBLP:conf/cvpr/SchonbergerF16, DBLP:conf/eccv/SchonbergerZFP16}, IDR~\cite{DBLP:conf/nips/YarivKMGABL20}, NeuS~\cite{DBLP:conf/nips/WangLLTKW21}, NeuralWarp~\cite{darmon2022improving}, 
HF-NeuS~\cite{wang2022hfneus}. 
COLMAP is a widely used MVS approach, where it reconstructs the point cloud from multi-vies images and extracts the explicit open surface by Ball-Pivoting Algorithm (BPA)~\cite{Bernardini1999ballpivoting}.
IDR is the state-of-the-art surface rendering method, which can reconstruct high-quality meshes under mask supervision for training.
NeuS is a pioneering work on surface reconstruction via SDF-based volume rendering, which achieves impressive results in surface reconstruction. 
The latest works, NeuralWarp, HF-NeuS achieve better performance for watertight shapes with improved high-frequency details or geometry consistency.
However, they fail to model arbitrary surfaces with open boundaries.
The straightforward solution mentioned in Sec.~\ref{sec:intro}, which is a naive extension of NeuS renderer by adding an absolute operation on predicted SDF values and keeping all other configurations the same for UDF reconstruction, is also evaluated. 

\vspace{-2mm}
\paragraph{Metrics.}
To measure the accuracy of reconstructed shapes with regards to the ground truth, we adopt the commonly used metric -- Chamfer Distance~\cite{barrow1977parametric} (CD) for quantitative comparisons to state-of-the-art methods. 
We use a masked Poisson method for UDF mesh extraction, where we first sample one million points in the UDF and adopt SPSR~\cite{kazhdan2013screened} to extract a watertight mesh, and then mask out the spurious surfaces with non-zero UDF values. 
We scale all the meshes of different datasets into a unit sphere for a fair comparison.
For the detailed calculation of the metrics, please refer to Fan~\etal~\cite{fan2017point}.

\subsection{Comparisons on Multi-view Reconstruction}
To demonstrate our reconstruction ability on diverse datasets (especially for the open surfaces), we perform quantitative and qualitative comparison to the SOTA methods on the above three datasets, including the open surface datasets varying in topology and geometry, as well as the watertight surface used in previous work.
Note that IDR uses mask supervision for DTU~\cite{jensen2014large}, DF3D~\cite{zhu2020deep} and MGN~\cite{DBLP:conf/iccv/BhatnagarTTP19} datasets, and ours uses mask supervision for DTU~\cite{jensen2014large} dataset.

\vspace{-2mm}
\paragraph{Quantitative Results.}
We report the average Chamfer Distance in Tab.~\ref{tab:quan-comp}.
The results show that our method outperforms these baselines on the two open surface datasets (DF3D~\cite{zhu2020deep} and MGN~\cite{mgn}) by a large margin. 
Our method is the only one which is able to reconstruct high-fidelity open surfaces, while the baselines are subject to watertight shapes. 
For the watertight dataset (DTU~\cite{jensen2014large}), our method is comparable with baselines.
We also provide the evaluation of the naive extension of NeuS renderer. 
The naive extension results in a large Chamfer Distance on open surface samples (naive extension: 9.53 vs ours: \textbf{1.49}) due to the noisy surfaces and sometimes fails to converge (DTU\_scan65).

\begin{table}[!t]
  \centering
  \caption{Quantitative comparison with the baselines on DF3D~\cite{zhu2020deep}, MGN~\cite{mgn} and DTU~\cite{jensen2014large} datasets. We split the open surface dataset (\ie MGN, DF3D) into some sub-categories. 
For each dataset, we mark the evaluated number of scenes in the subcategories.
  In this table, we report the average score of each category under the metric -- Chamfer Distance ($\times 10^{-3}$).
  From the results, we can see that our method outperforms the IDR and NeuS on the two open surface datasets (MGN, DF3D) by a large margin.
  }
  \vspace{-2mm}
  \begin{adjustbox}{width={\linewidth},keepaspectratio}
    \begin{tabular}{c|ccccccc}
    \toprule[1.1pt]
    DataSet & COLMAP & IDR   & NeuS  & NeuralWarp & HF-NeuS  & Ours \\
    \midrule
    MGN-upper (6) & 12.32 & 19.68 & 11.65 & 15.40 & 9.16 & \textbf{6.78} \\
    MGN-pants (4) & 30.62 & 23.70 & 17.95 & 22.26 & 24.02 & \textbf{16.43} \\
    DF3D-upper (6) & \textbf{8.60} & 14.46 & 15.29 & 10.27 & 23.31 & 8.72 \\
    DF3D-pants (4) & 25.91 & 16.91 & 16.00 & 7.99 & 12.29 & \textbf{5.77} \\
    DF3D-dress (8) & 9.77 & 14.27 & 11.75 & 7.79 & 12.03 & \textbf{7.39} \\
    DTU (15) & \textbf{3.75} & 4.92 & 4.46 & 3.78 & 5.60 & 4.98 \\
    \midrule
    Mean(open surface) & 15.35 & 17.19 & 13.98 & 12.05 & 15.58 & \textbf{8.60} \\
    \midrule
    Mean(all) & 11.31 & 12.91 & 10.80 & 9.16 & 12.10 & \textbf{7.34} \\
    \bottomrule[1.1pt]
    \end{tabular}%
     \vspace{-2mm}
    \end{adjustbox}
  \label{tab:quan-comp}%
\end{table}%

\begin{figure}
    \centering
    \begin{minipage}[c]{.24\linewidth}
    \centering
        \includegraphics[width=\linewidth]{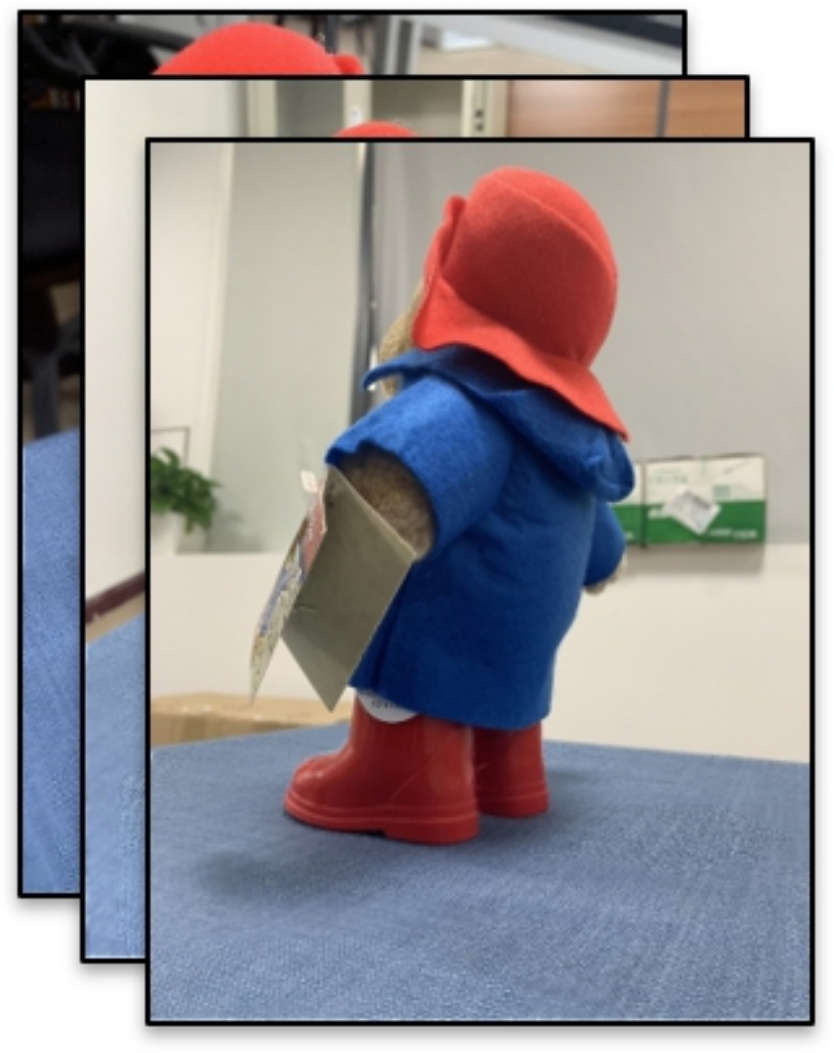}
    \end{minipage}
    \begin{minipage}[c]{.75\linewidth}
    \centering
        \includegraphics[width=.49\linewidth]{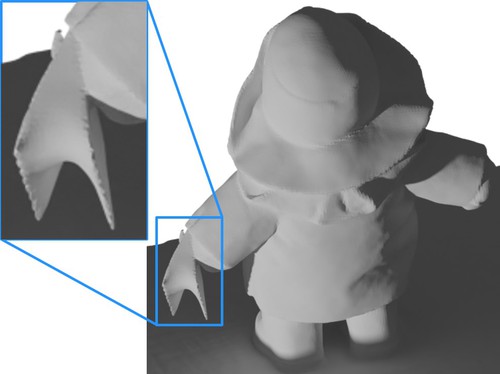}
        \includegraphics[width=.49\linewidth]{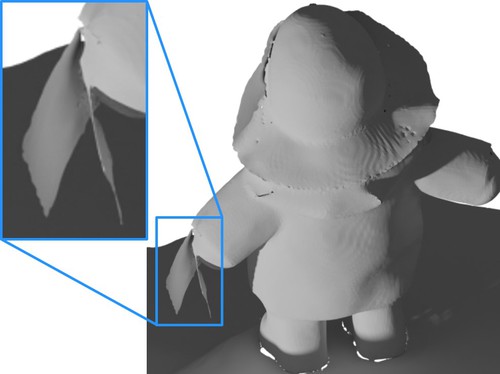} 
    \end{minipage}

    \begin{minipage}[c]{.24\linewidth}
    \centering
        {\includegraphics[width=\linewidth]{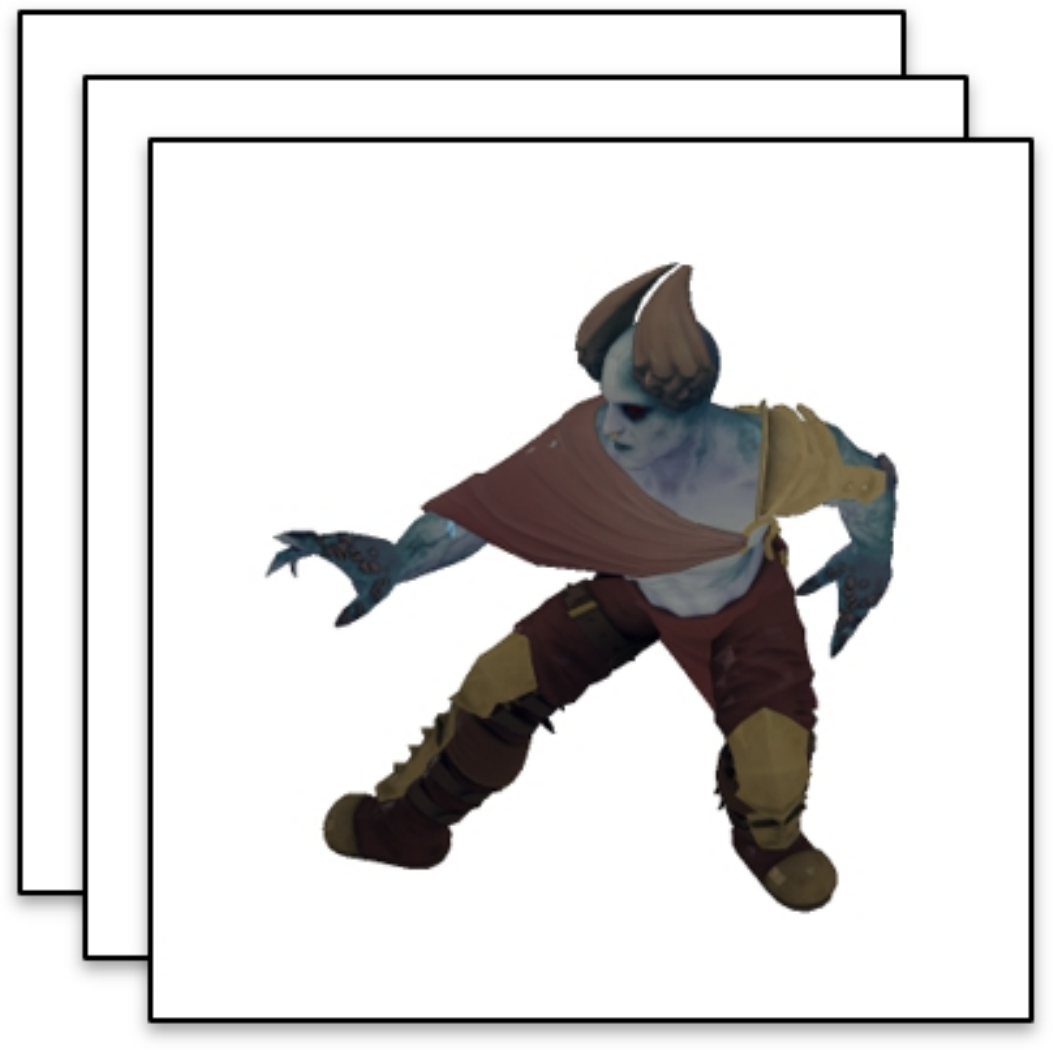}}
    \end{minipage}
    \begin{minipage}[c]{.75\linewidth}
    \centering
        {\includegraphics[width=.49\linewidth]{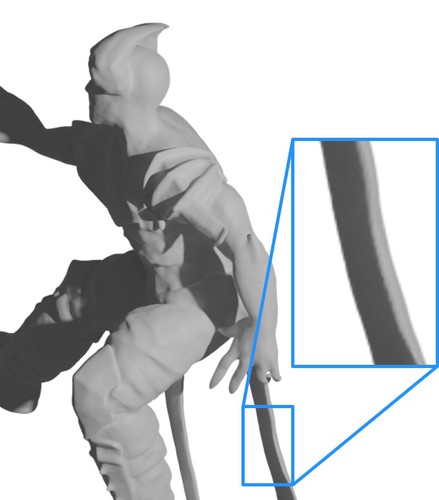}}
        {\includegraphics[width=.49\linewidth]{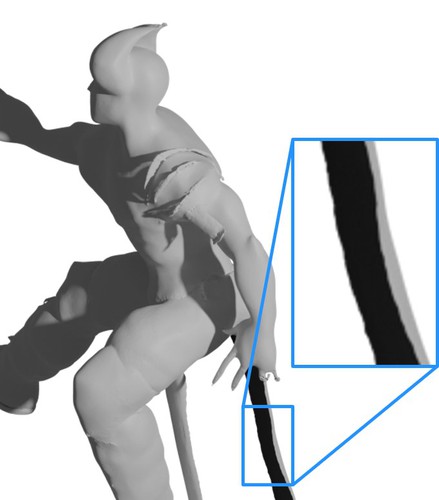}}
    \end{minipage}
    
    \begin{minipage}[c]{.24\linewidth}
        \centering
        \subcaption{Input}
    \end{minipage}
    \begin{minipage}[c]{.37\linewidth}
        \centering
        \subcaption{NeuS}
    \end{minipage}
    \begin{minipage}[c]{.37\linewidth}
        \centering
        \subcaption{Ours}
    \end{minipage}

    \vspace{-2mm}
    \caption{Comparisons with NeuS~\cite{DBLP:conf/nips/WangLLTKW21} on BMVS~\cite{yao2020blendedmvs} dataset and Mixamo~\cite{mixamo} dataset. Our \OurNetName{} can reconstruct geometries with open boundaries (\eg greeting card at the bear hand, clothes on the human character) while SDF-based method NeuS~\cite{DBLP:conf/nips/WangLLTKW21} is unable to properly represent. Further, our method also maintain comparable reconstruction quality for watertight parts such as bodies of the bear and the human character.}
    \label{fig:dtu}
    \vspace{-5mm}
\end{figure}

\vspace{-2mm}
\paragraph{Qualitative Results.}
The quantitative comparisons on the DF3D and MGN datasets are visualized in Fig.~\ref{fig:comp-vis}.
As shown in Fig.~\ref{fig:comp-vis} (b) (c) (d) (e), the SDF-based methods (IDR~\cite{DBLP:conf/nips/YarivKMGABL20},  NeuS~\cite{DBLP:conf/nips/WangLLTKW21}, NeuralWarp~\cite{darmon2022improving}, HF-NeuS~\cite{wang2022hfneus}) are subject to watertight shapes, and underperform with surfaces with open boundaries.
In comparison, \OurNetName{} can reconstruct high-fidelity meshes with open boundaries (such as the sleeves, collars and waists)
without mask as shown in Fig.~\ref{fig:comp-vis} (f).

We further conduct comparisons with NeuS~\cite{DBLP:conf/nips/WangLLTKW21} on the Mixamo~\cite{mixamo} and BMVS~\cite{yao2020blendedmvs} datasets. As shown in Figure~\ref{fig:dtu} (\ie mixamo-demon and bmvs-bear), our method is able to reconstruct geometries with open boundaries, such as the greeting card in the bear's hand and the single-layer cloak.
It is clear to obverse that NeuS fails to synthesis the surface with open boundaries. 
In comparison, our reconstructed shape geometries are accurate, and the complex open structures are preserved.
More qualitative results are presented in our supplementary document.

We additionally show some challenging cases with the complex structure in Fig.~\ref{fig:teaser}, such as the hollowed box, plant leaves, and patch-based fish.
From the results on these objects with complex open boundaries, we see both detailed geometry and complex open-surface structures clearly, which validates that \OurNetName{} learns a better UDF for multi-view reconstruction.

\begin{figure}[h]
    \centering
    \begin{minipage}[c]{.3\linewidth}
    {\includegraphics[width=\linewidth]{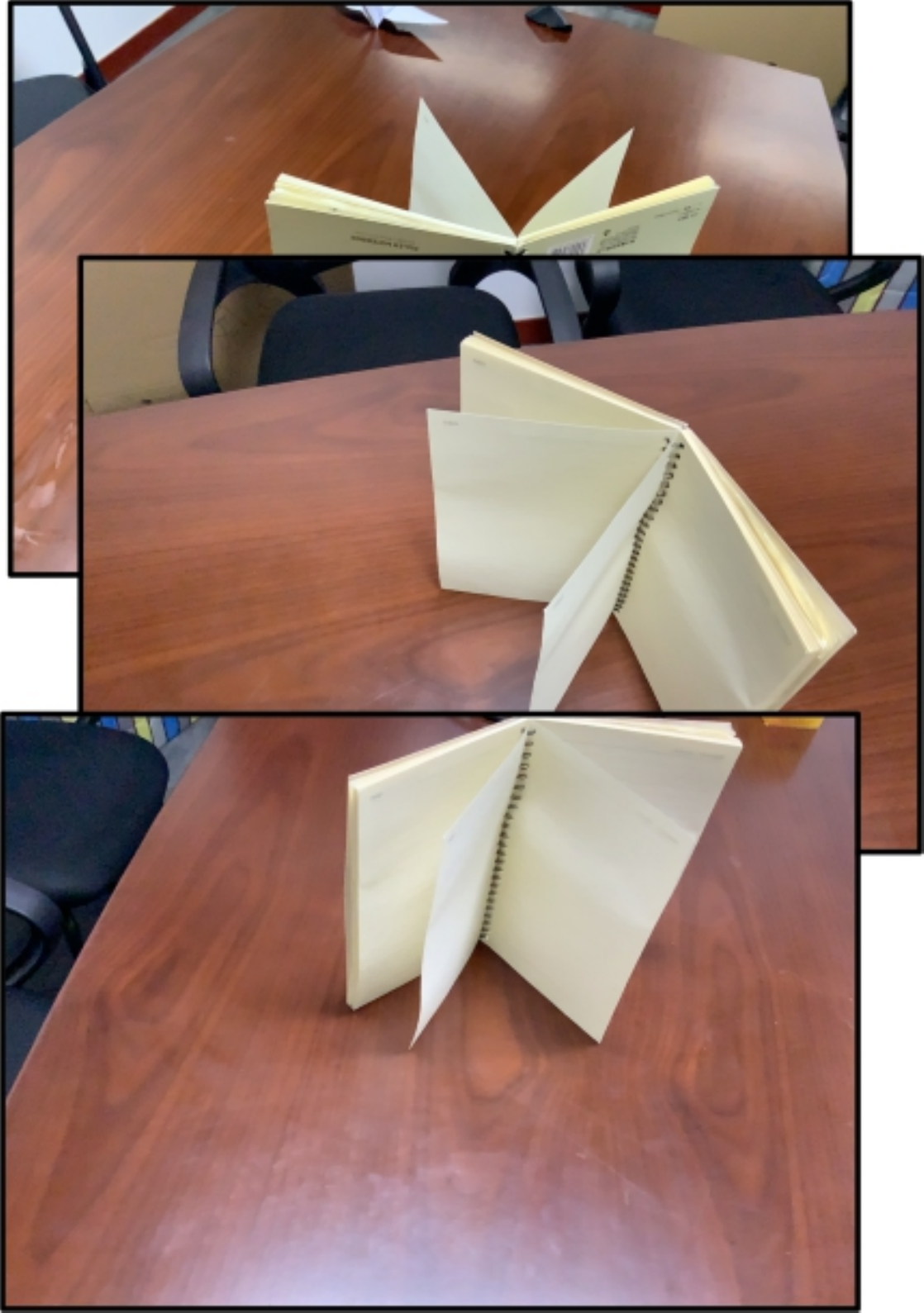}}
    \end{minipage}
    \begin{minipage}[c]{.6\linewidth}
    \centering
        \includegraphics[width=.49\linewidth]{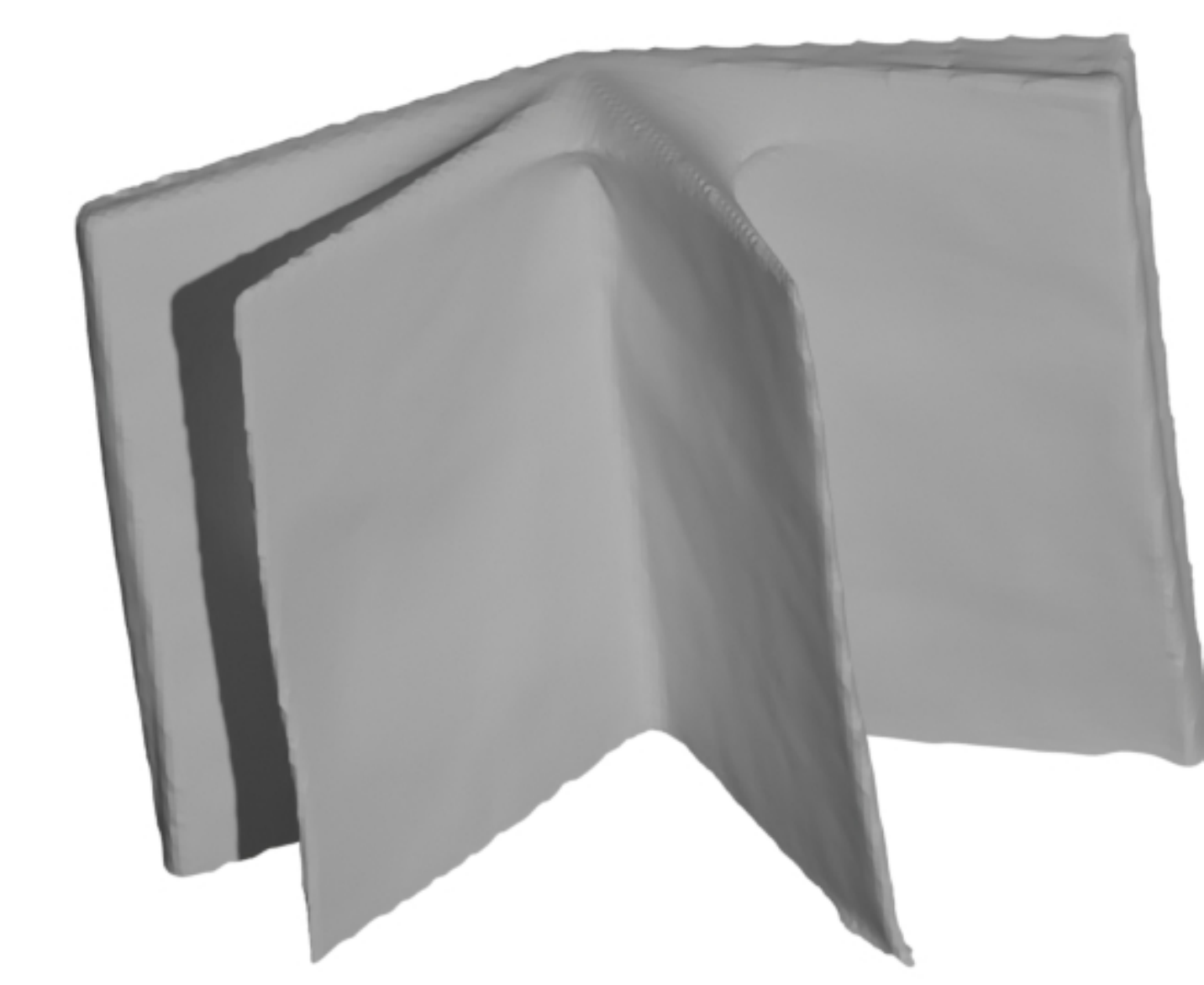}
        \includegraphics[width=.49\linewidth]{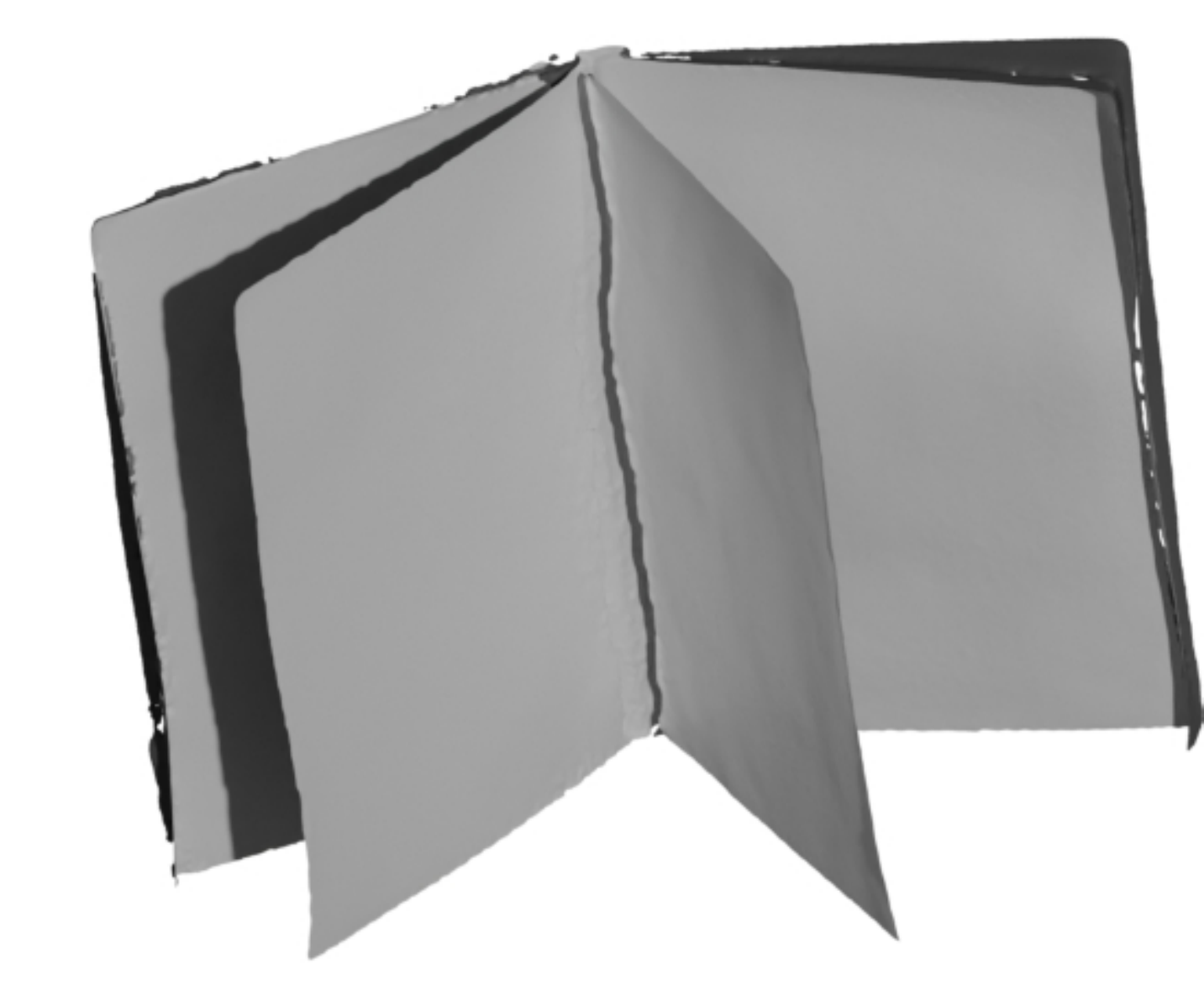} \\       
        {\includegraphics[width=.49\linewidth]{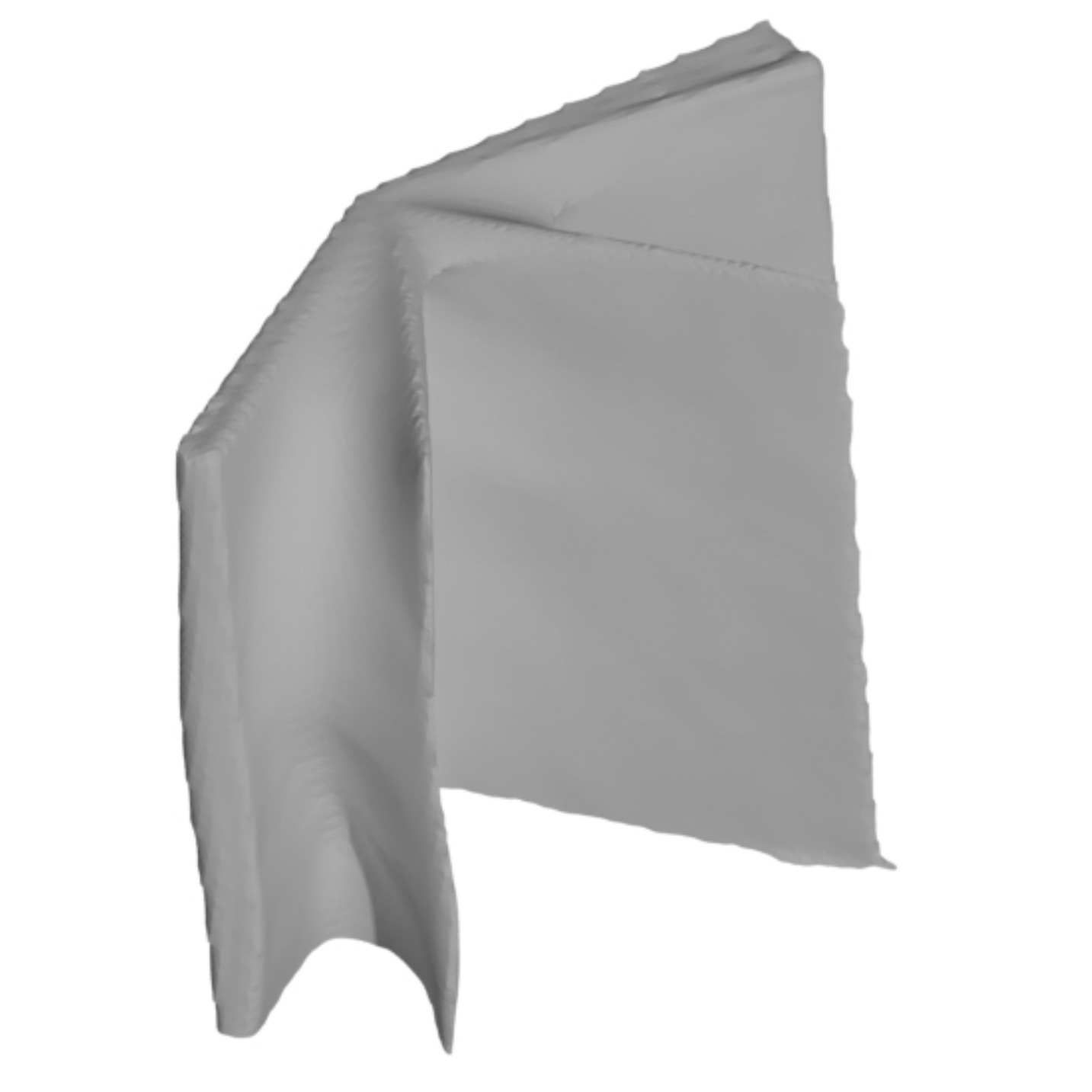}}
        {\includegraphics[width=.49\linewidth]{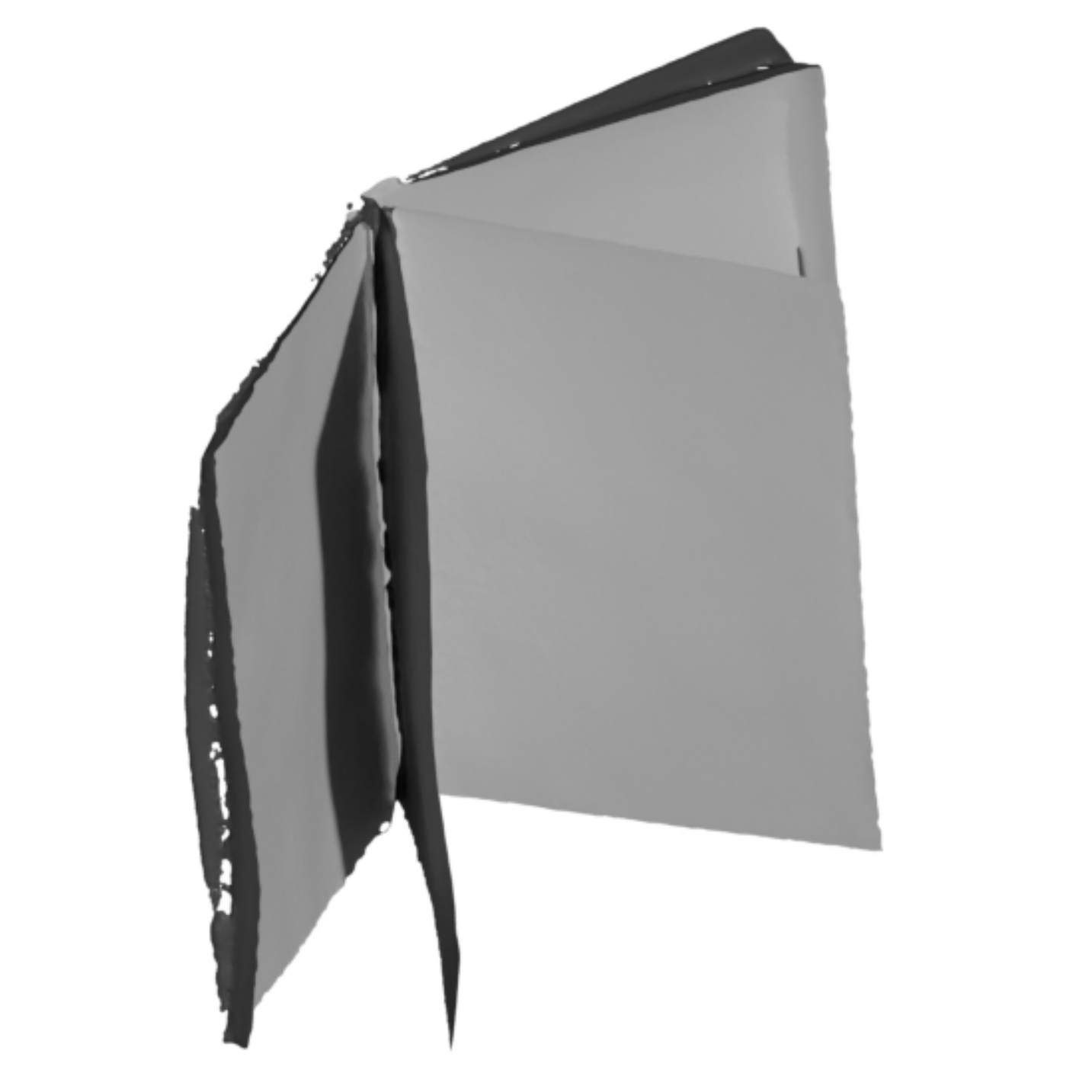}}
    \end{minipage}
    
    \begin{minipage}[c]{.3\linewidth}
        \centering
        \subcaption{Input}
    \end{minipage}
    \begin{minipage}[c]{.3\linewidth}
        \centering
        \subcaption{NeuS}
    \end{minipage}
    \begin{minipage}[c]{.3\linewidth}
        \centering
        \subcaption{Ours}
    \end{minipage}
        
    \vspace{-2mm}
    \caption{The evaluation on the captured real scenes. We conduct a comparison with NeuS on the real scenes. From the results, we can see that our method is capable of complex surface reconstruction with open boundaries, while NeuS encourages to merge the book pages together.}
    \label{fig:real}
\end{figure}
\noindent\textbf{Captured Real Scenes.}
We further evaluate our method on the captured data from real-world objects, including book pages, fan blades and plant leaves.
For each scene, we use the mobile phone to capture a video surrounding the object and extract about 200 frames from the video.
Then, we use COLMAP to estimate the camera poses and take the calibrated images as input to optimize the network parameters without mask supervision.
Fig.~\ref{fig:real} presents the reconstructed shape of book pages, and more captured real scenes are presented in our supplementary document.
The results show that NeuS encourages merging the book pages together and causing unrealistic geometry, while ours achieves accurate surface reconstruction with open boundaries.

\begin{figure}
    \centering
        \subfloat[Input]{\includegraphics[width=.22\linewidth]{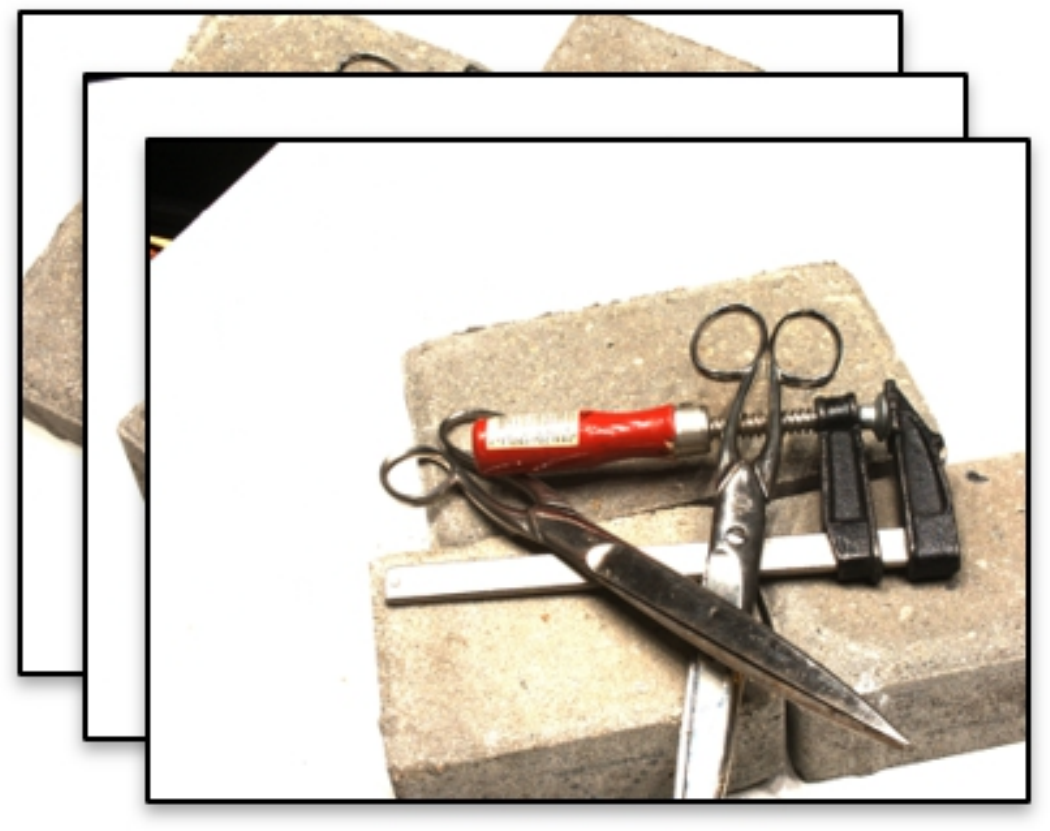}}
        \subfloat[w/o $w_s(t)$]{\includegraphics[width=.19\linewidth]{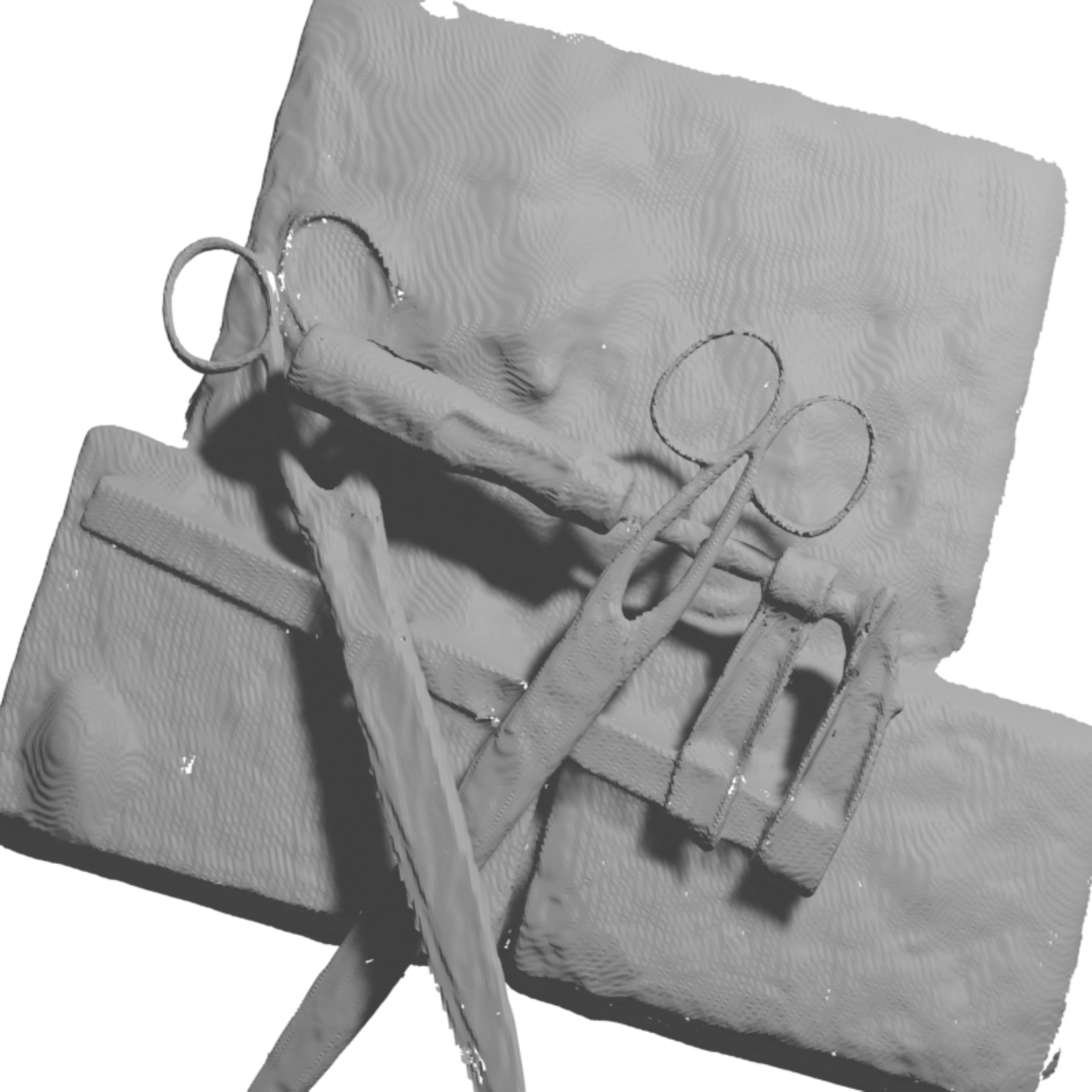}}
        \subfloat[w/o Normal Reg]{\includegraphics[width=.19\linewidth]{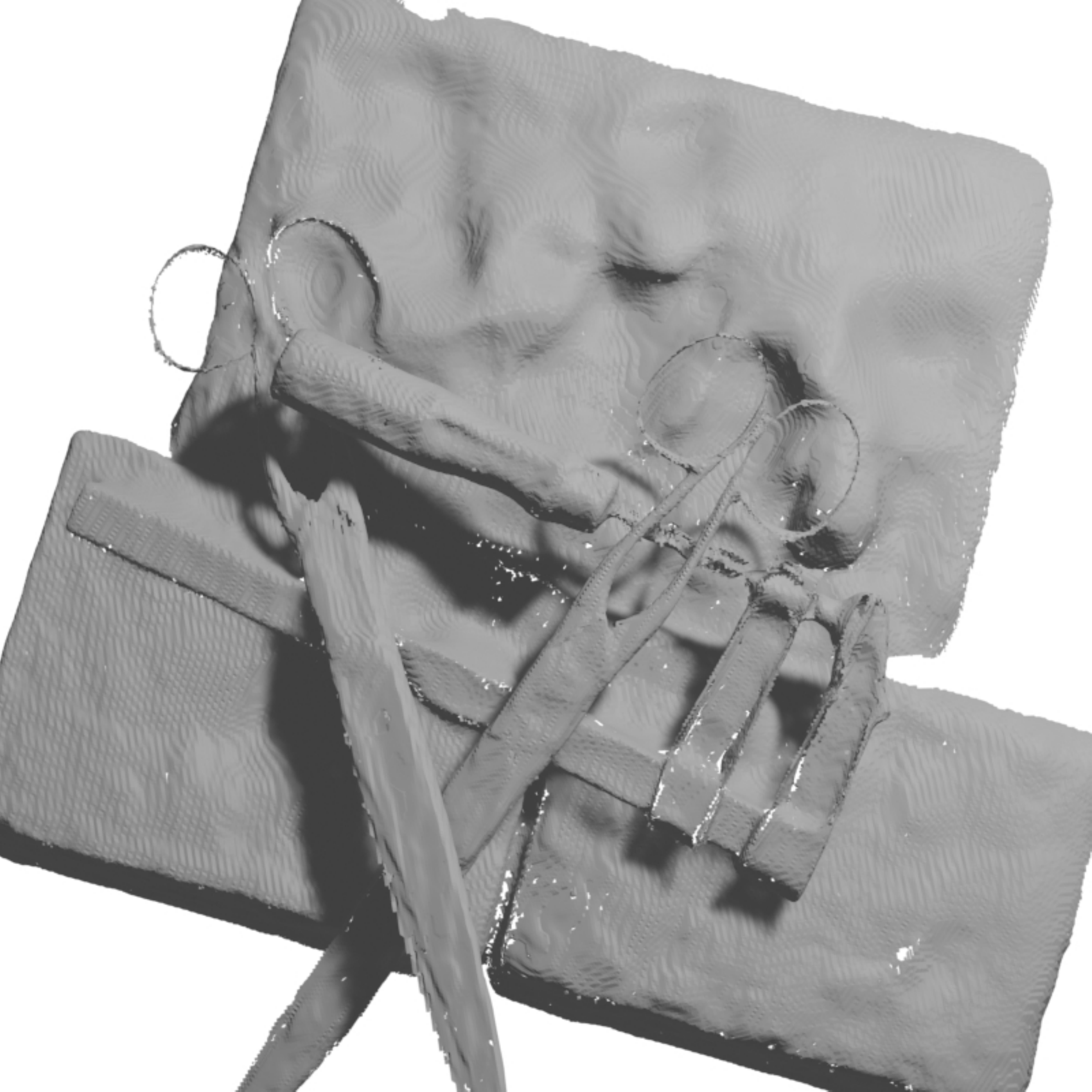}}
        \subfloat[w/o $\mathcal{L}_e$]{\includegraphics[width=.19\linewidth]{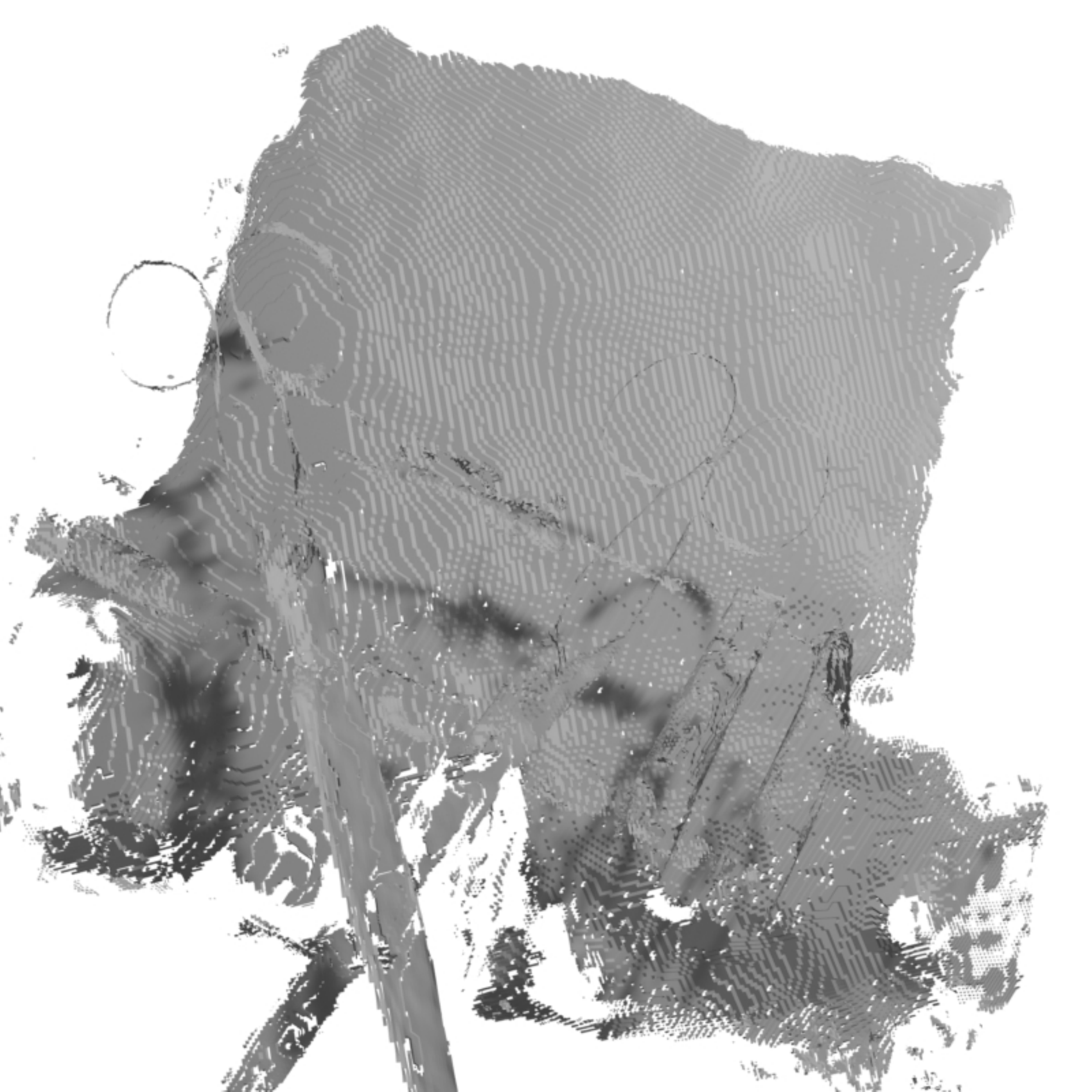}}
        \subfloat[Ours]{\includegraphics[width=.19\linewidth]{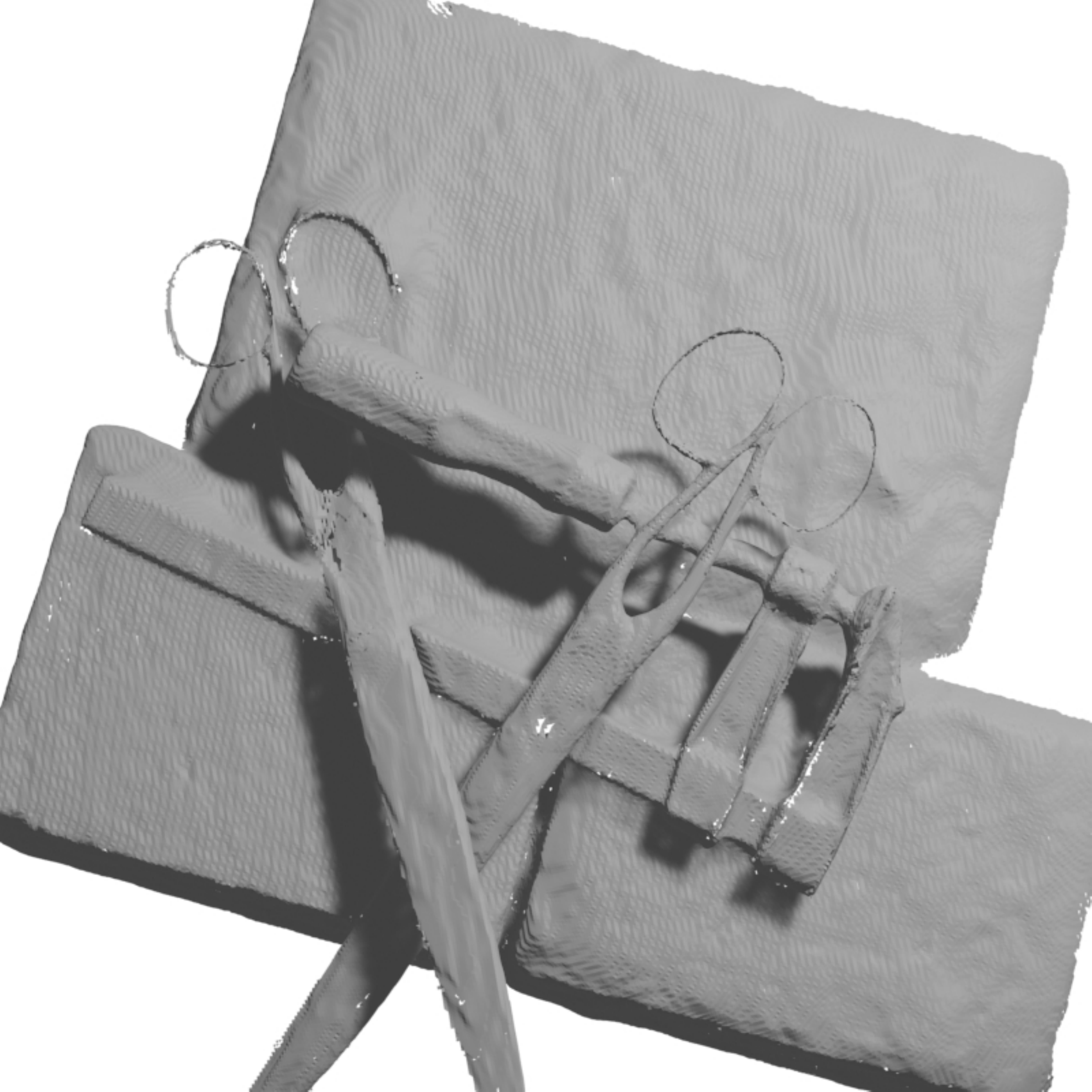}}
        \hrule\vspace{1mm}
        CD $\times 10^{-4}$\hspace{7mm}1.29\hspace{11mm}1.95\hspace{11mm}6.48\hspace{9mm}{0.85\hspace{3mm}}
    \caption{Ablation of different components in our \OurNetName{}. (a) the input multi-view images; (b) w/o our importance points sampling, namely adopting color weight function for sampling instead; (c) w/o normal regularization; (d) w/o Eikonal loss; (e) Ours with full components. From all the results, it is clearly observed that all key designs are critical for our high-fidelity complex surface reconstruction.}
    \label{fig:abl}
    \vspace{-3mm}
\end{figure}
\begin{figure}[h]
    \centering
        \subfloat[Input]{\includegraphics[width=.27\linewidth]{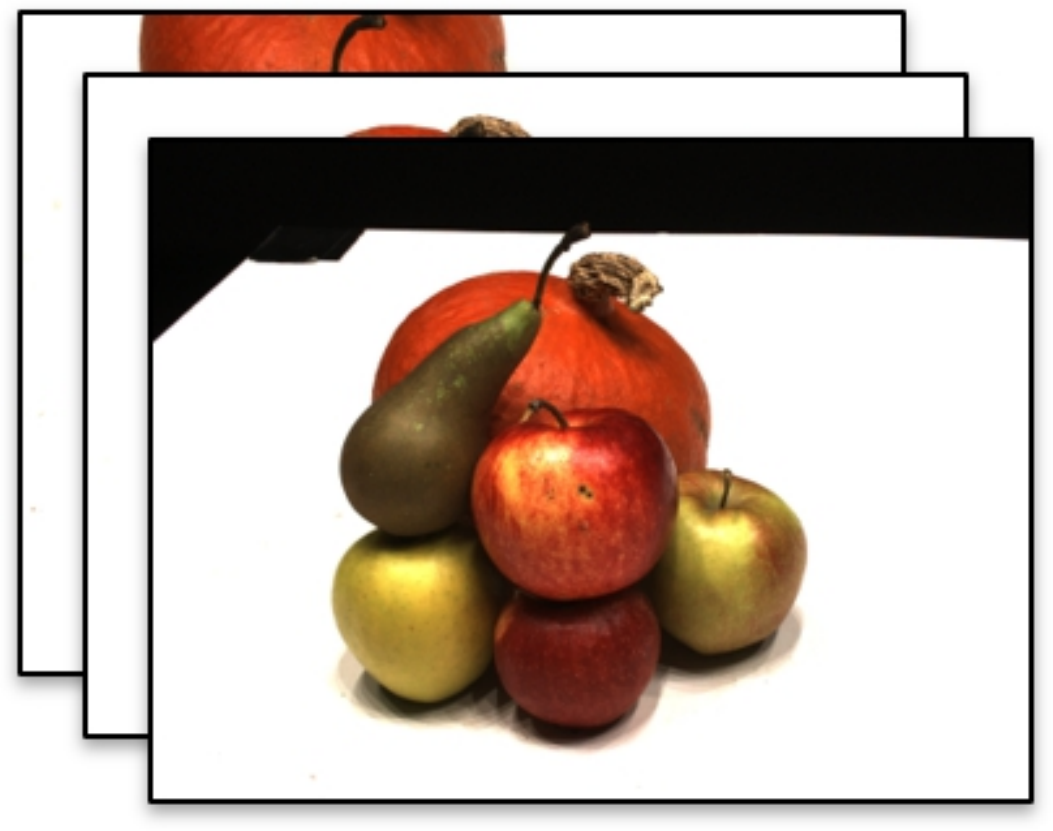}}
        \subfloat[$1-e^{-x}$]{\includegraphics[width=.24\linewidth]{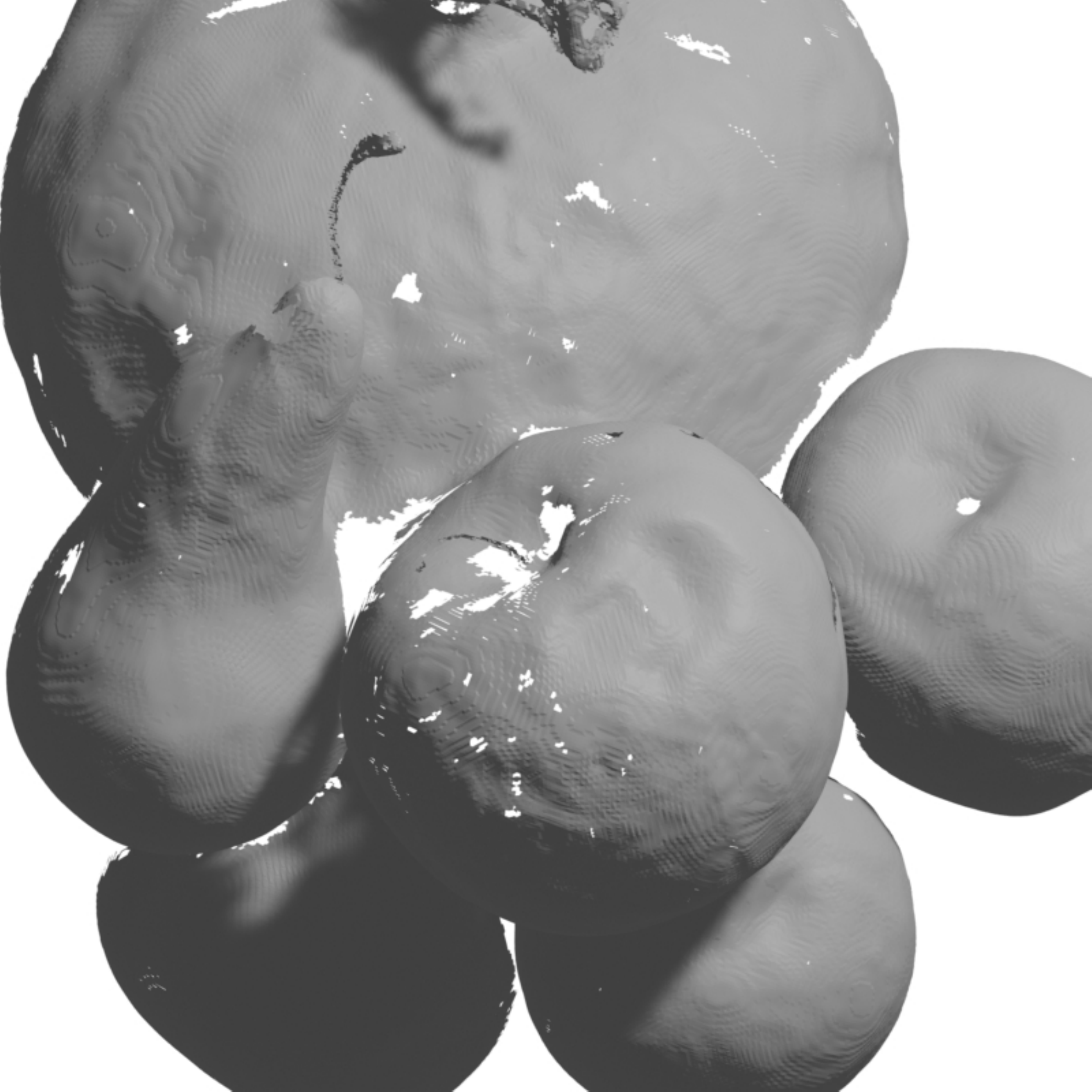}}
        \subfloat[$\frac{2arctan(x)}{\pi}$]{\includegraphics[width=.24\linewidth]{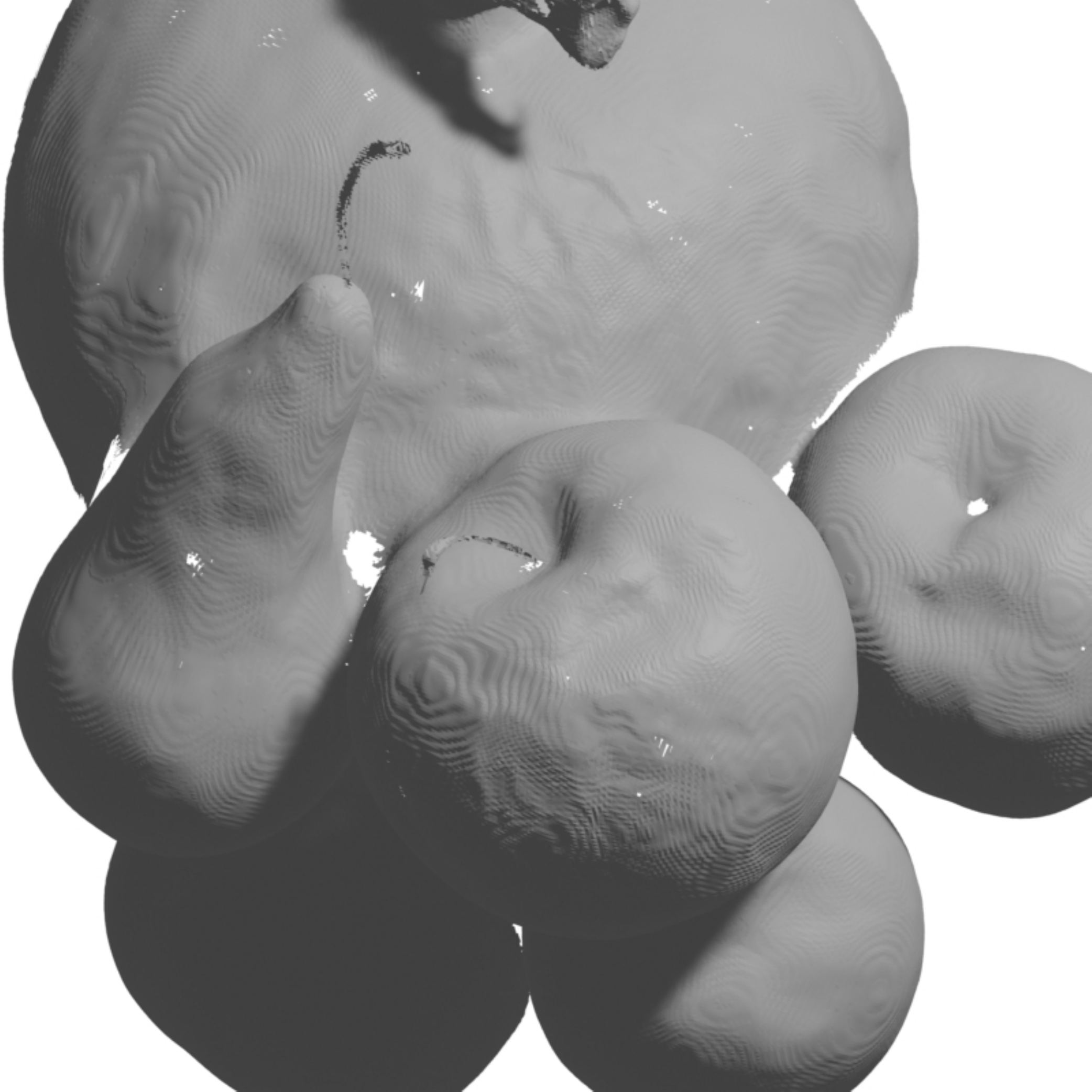}}
        \subfloat[$x/(1+x)$]{\includegraphics[width=.24\linewidth]{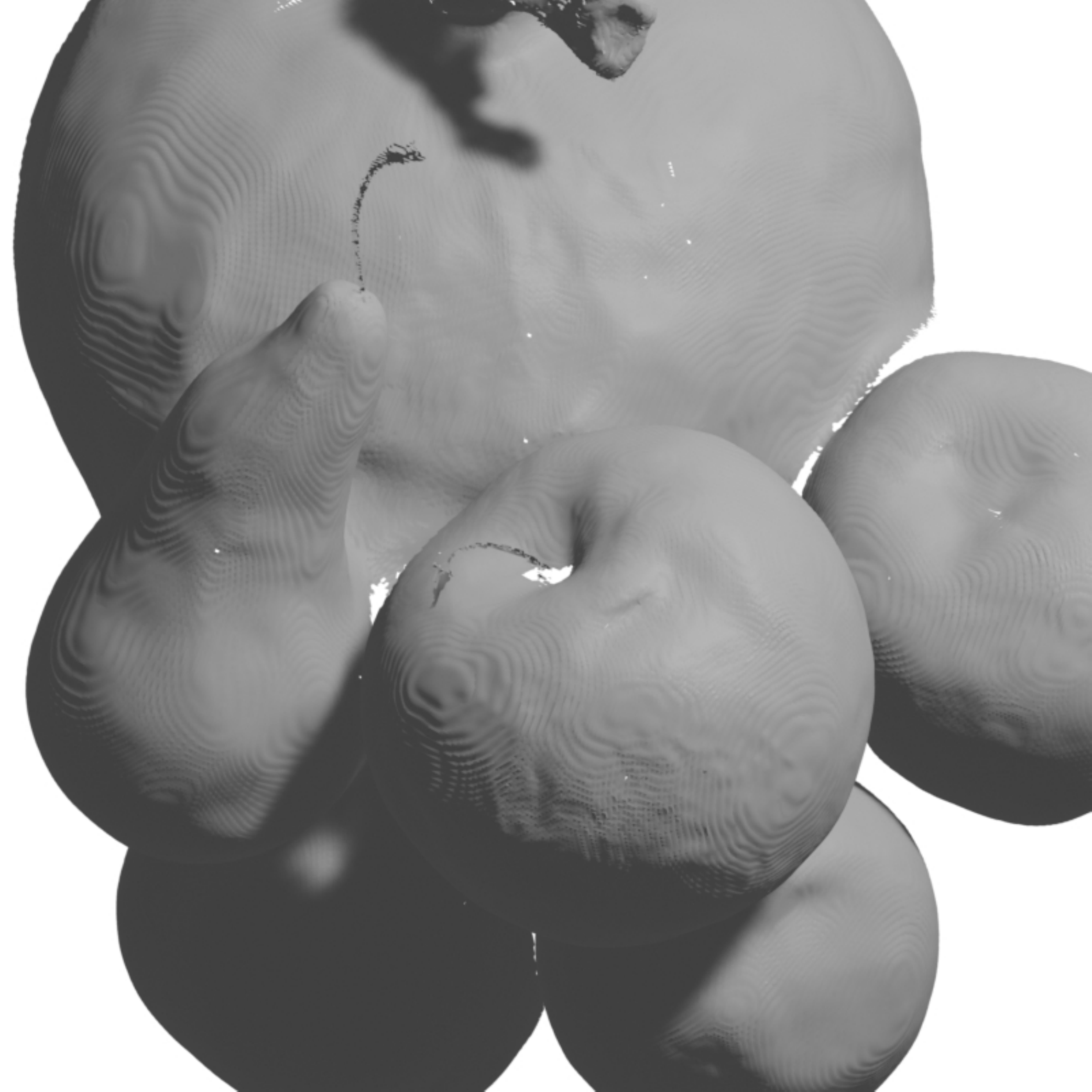}}
        \hrule\vspace{1mm}
        CD $\times10^{-4}$\qquad\quad1.10\qquad\quad1.09\quad\qquad\quad1.01
    \caption{Ablation on different choice of $\varsigma_r$ in $\tau_r$. According to the given rules (Eqn.~\ref{eq:rule1}, Eqn.~\ref{eq:rule2}), we have evaluated some similar function (\eg, $1-e^{-x}$, $2arctan(x)/(\pi)$, $x/(1+x)$). From the results, we find the different $\varsigma_r$ only affects the speed of converge. The reconstructed shapes are similar until converge. In the figure, we present the visual results when ours converges. We can see that other ablated versions do not  converge with several holes.}
    \vspace{-3mm}
    \label{fig:abl_h}
\end{figure}

\subsection{Further Discussions and Analysis}\label{sec:abla}
We conduct three ablation studies to validate our individual designs of our methods. 
First, evaluating different choices of $\varsigma_r$ in $\tau_r$ shows its effectiveness for UDF learning.
Then, we also validate the necessity of our designed importance sampling and normal regularization for the accurate open surface reconstruction.
All the ablation studies are conducted on multiple sample objects with diverse shapes.

\paragraph{The choice of $\varsigma_r$ in $\tau_r$.}%
Although we have given the rules (Eqn.~\ref{eq:rule1}, Eqn.~\ref{eq:rule2}) that $\varsigma_r$ should satisfy, there is a family of functions that satisfy the rules. 
All the functions in the family are suitable for UDF volume rendering, so we conduct validations on several different candidate functions to check the convergence aptitude of each function for network optimization, \ie, with which function the network converges to the best results in a given training iterations. 
Fig~\ref{fig:abl_h} shows the visual results of three candidate functions ($1-e^{-x}$, $\frac{2arctan(x)}{\pi}$ and $\frac{x}{1+x}$) that follow the rules. 
After the given iterations (300k), the network using the function $\frac{x}{1+x}$ converges to the best result both qualitatively and quantitatively, while the other functions are not fully convergent and cause incomplete surfaces and slightly higher Chamfer-Distances. 
Evaluation on diverse shapes also demonstrates that all the functions work well and the chosen one ($\frac{x}{1+x}$) works the best (ours: \textbf{1.11} vs candidates: 1.13/1.18) in our setting.

\paragraph{Necessity of Importance Points Sampling $w_s(t)$ (Eqn.~\ref{eq:weight_s}).}
To demonstrate the necessity of Eqn.~\ref{eq:weight_s}, we design an ablated version that removes the importance of point sampling and uses the Eqn.~\ref{eq:weight_r} to sample the points for training.
Fig.~\ref{fig:abl} (b) shows that the output surfaces using the same weight for both rendering and sampling are less smooth and with larger CD errors as expected.
The errors come from the sampling points distribution is not balanced on both sides of the surface, and the network is not well regularized. 
Fig.~\ref{fig:abl} (e) shows that the network is well regularized with the importance sampling and produces better results. 
 
\begin{figure}[t!]
    \centering
        \includegraphics[width=.25\linewidth]{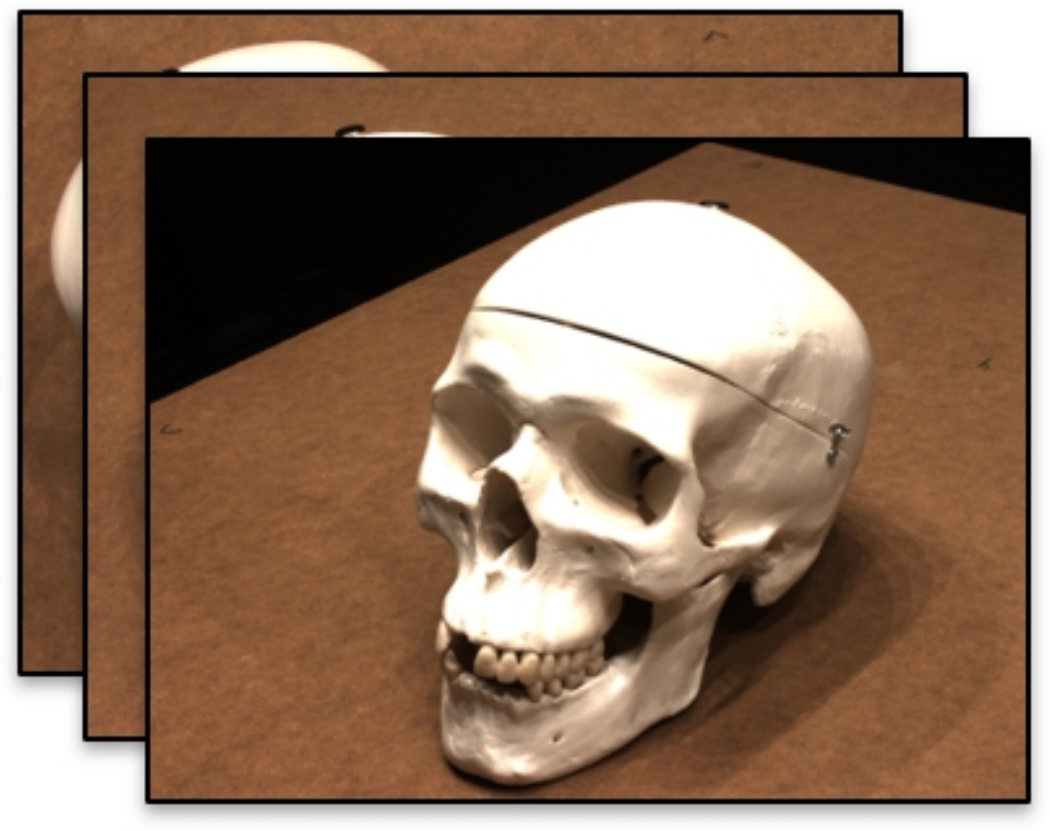}
        \includegraphics[width=.35\linewidth]{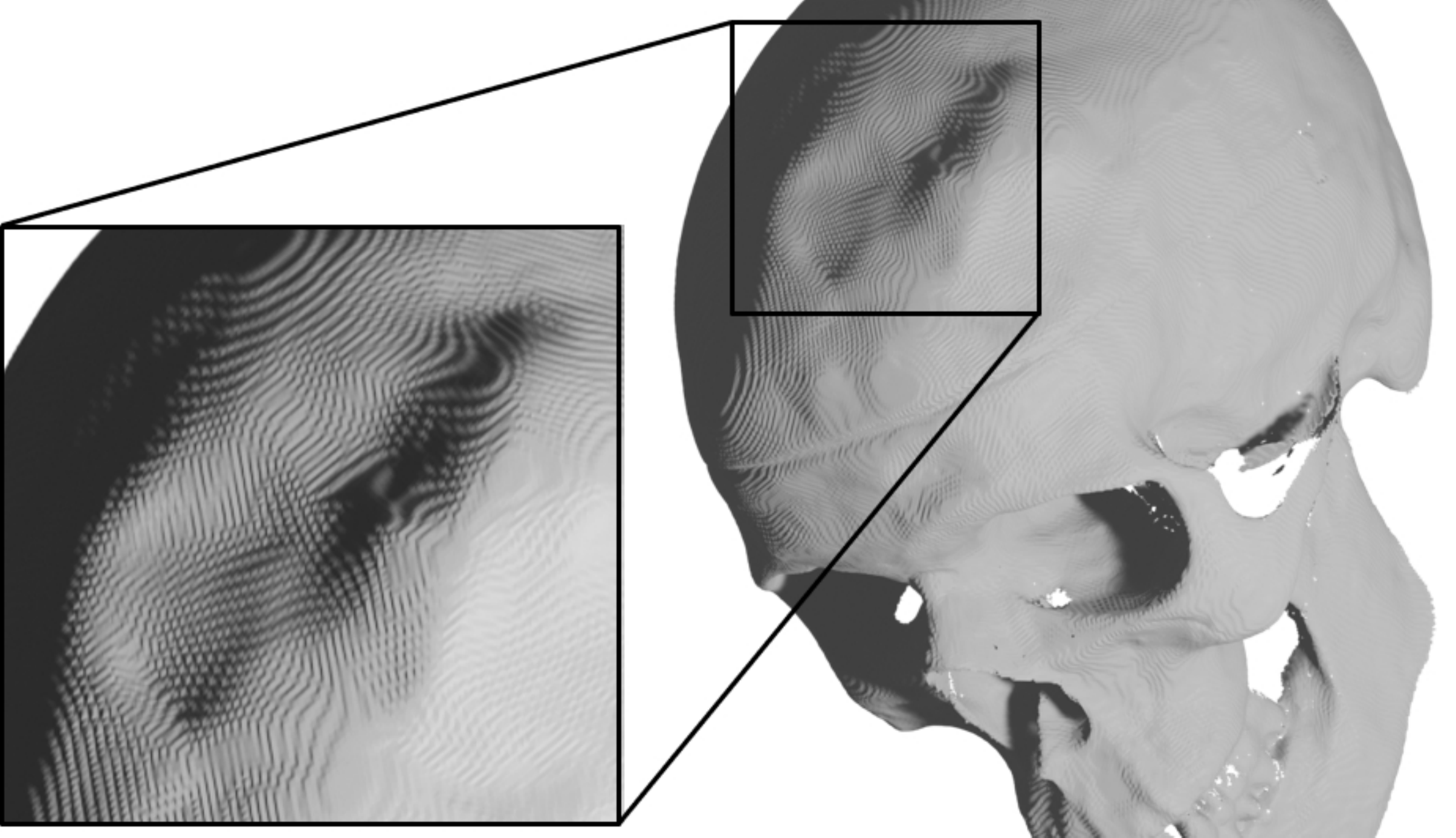}
        \includegraphics[width=.35\linewidth]{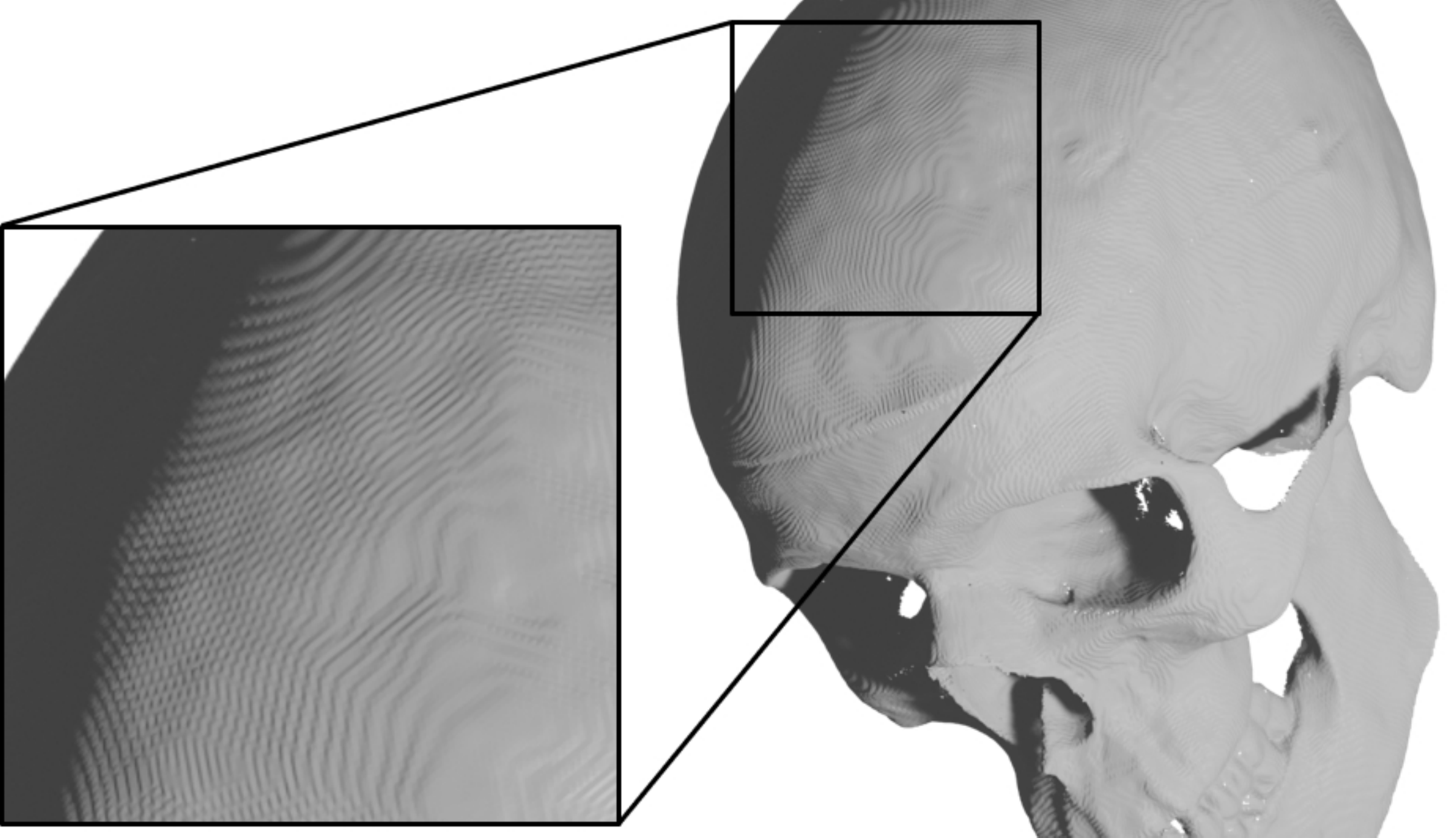}\\
        \subfloat[Input]{\includegraphics[width=.25\linewidth]{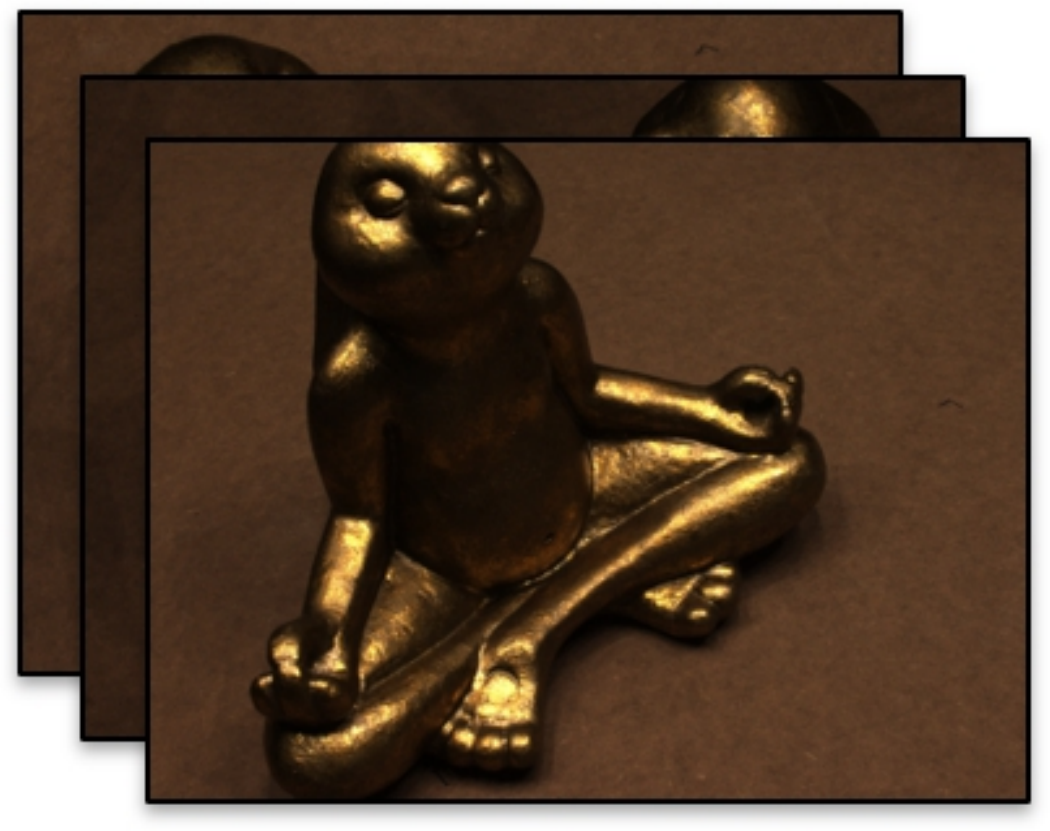}}
        \subfloat[w/o Normal Reg]{\includegraphics[width=.35\linewidth]{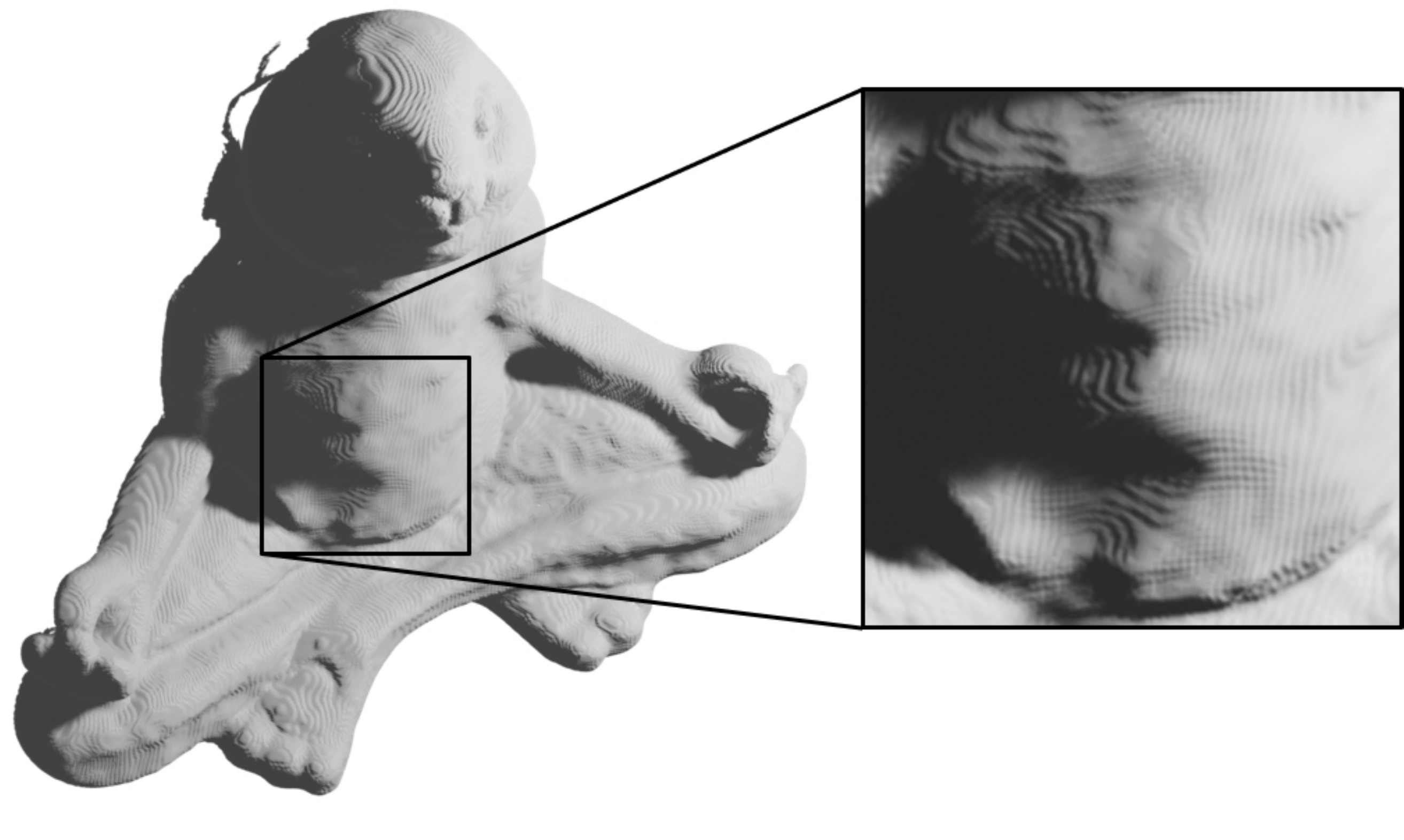}}
        \subfloat[Ours]{\includegraphics[width=.35\linewidth]{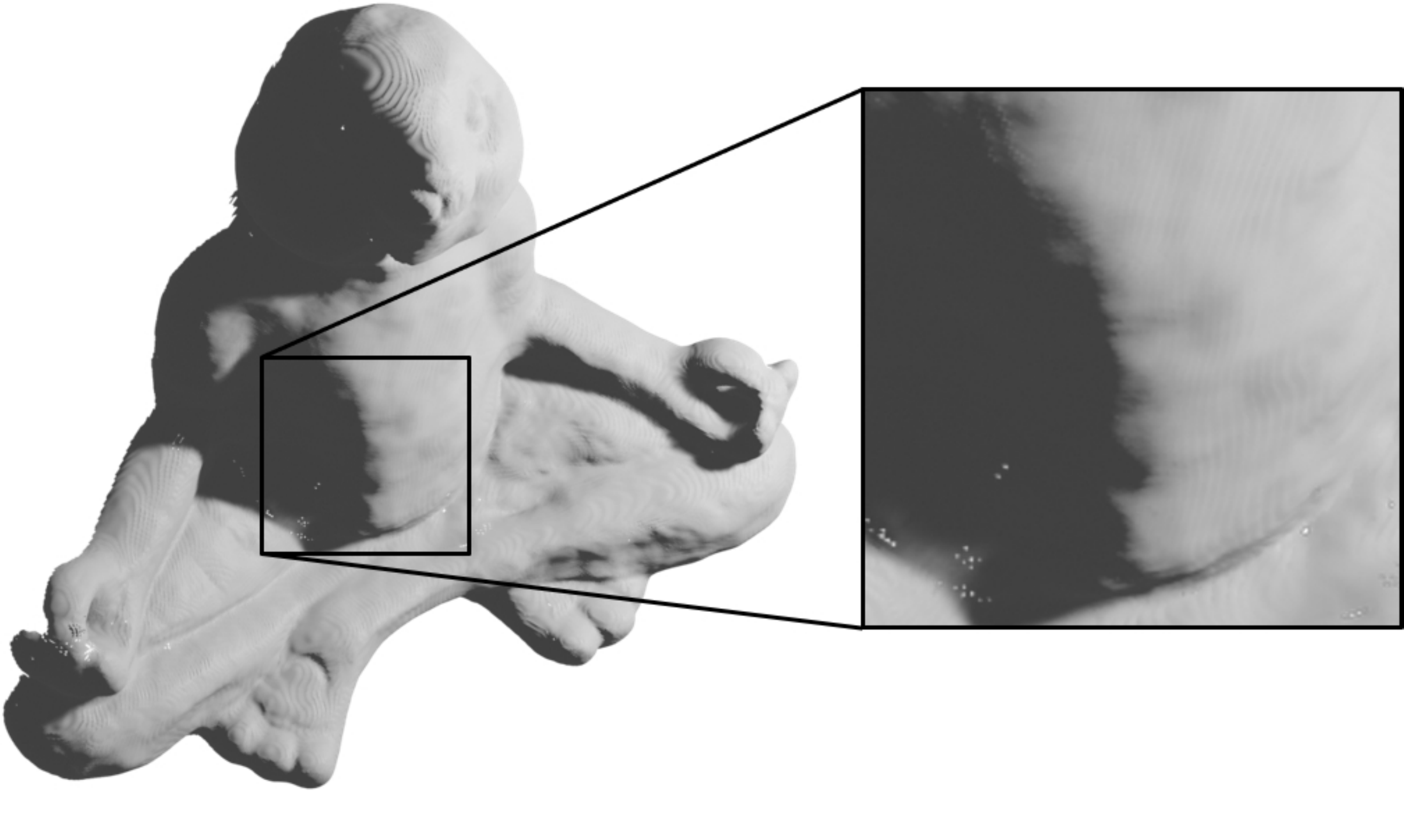}}
    \vspace{-2mm}
    \caption{Evaluations on two extreme cases for effectiveness of normal regularization. (b): Without normal regularization, artifacts can be observed on the reconstructed skull with high lights and metal rabbit with dark shadow. (c): Our full method with normal regularization can produce more visually pleasing results.
    }
    \label{fig:abl_brightdark}
    \vspace{-4mm}
\end{figure}

\paragraph{Necessity of Normal Regularization.}
We use the Normal Regularization (Sec~\ref{sec:normal}) to address the unstable gradients at the zero level set of UDF, and we conduct a validation on the necessity of the Normal Regularization. 
As shown in Fig.~\ref{fig:abl}(c), without the Normal Regularization the result suffers from rough surfaces and large Chamfer distance due to unstable gradient calculating.
Further, the normal regularization benefits the surface reconstruction even with extreme cases, \eg extremely bright or dark regions. 
Figure~\ref{fig:abl_brightdark} shows two cases in extremely bright and dark light conditions. The visualization results indicate that normal regularization is critical to alleviating the geometric error induced by the ambiguity of light conditions (\eg the artifacts on the DTU-skull and the DTU-metal-rabbit).

\begin{figure}
    \centering
    \subfloat[Input]{\includegraphics[width=0.3\linewidth]{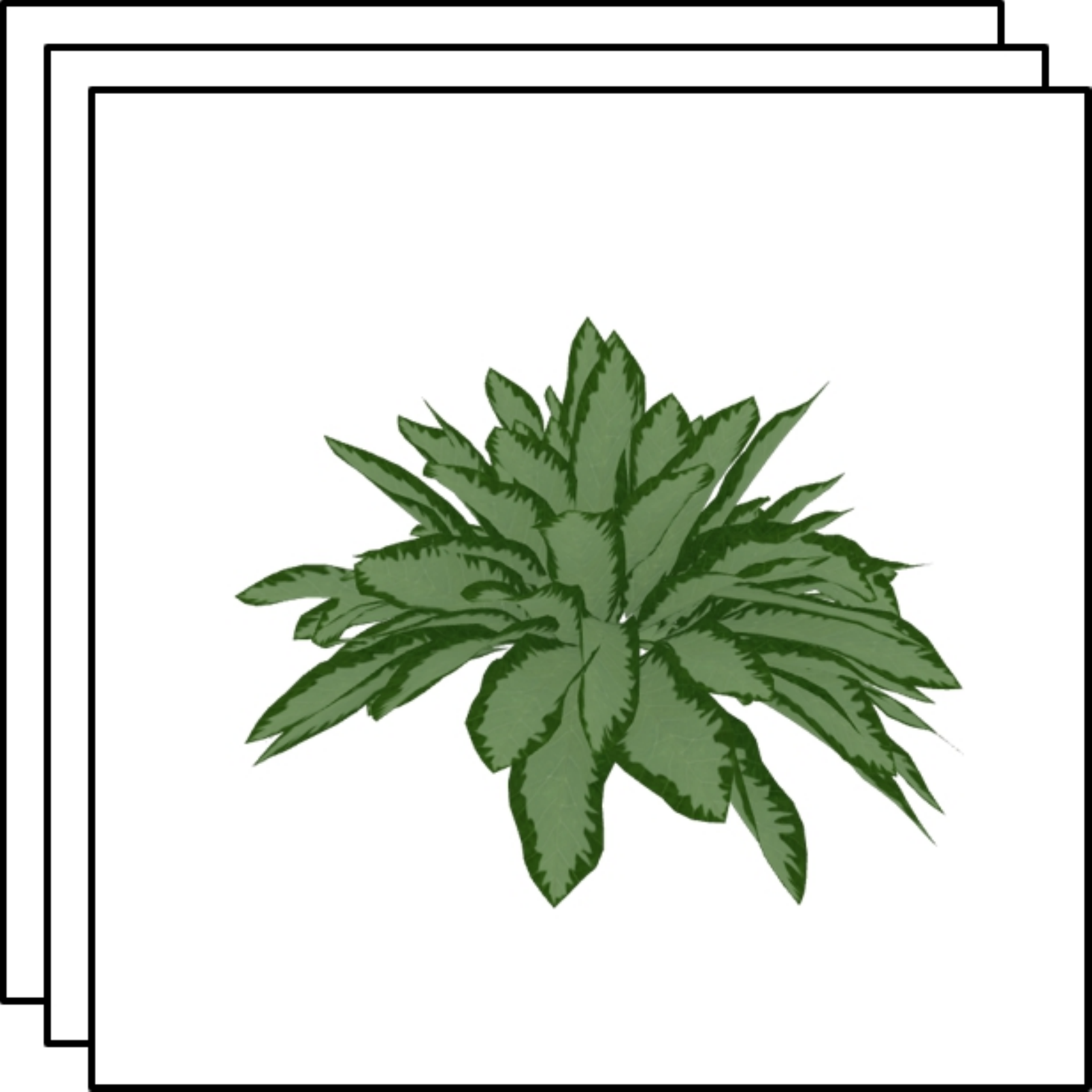}}
    \subfloat[Ours]{\includegraphics[width=0.33\linewidth]{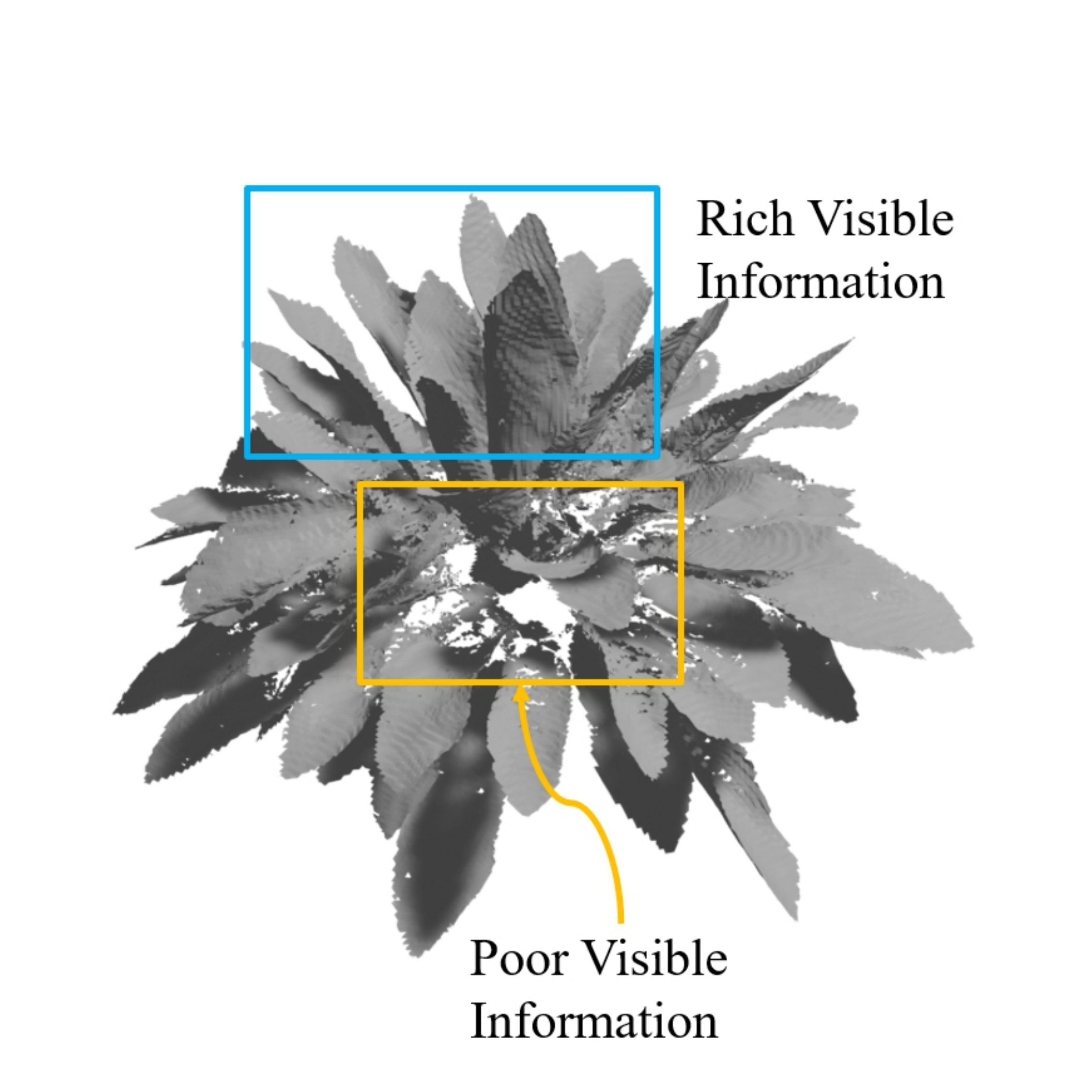}}
    \subfloat[GT]{\includegraphics[width=0.33\linewidth]{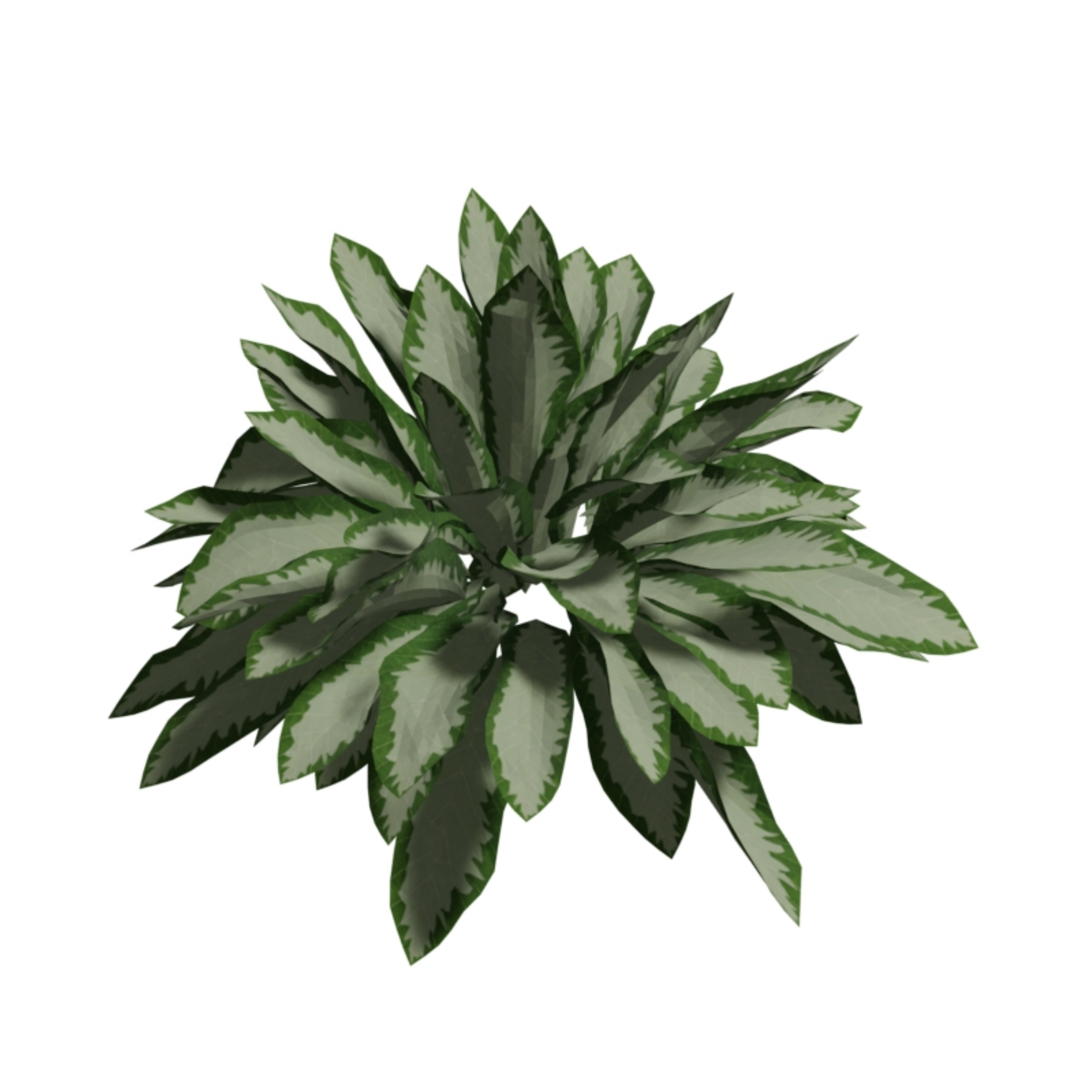}}
    \caption{Failure Case. The severely occluded parts in this plant can't be well reconstructed.}
    \label{fig:failure}
    \vspace{-4mm}
\end{figure}

    \section{Discussions \& Conclusions}
\label{sec:conclu}
\paragraph{Limitation.}
While our method can successfully reconstruct arbitrary surfaces with open boundaries, it still has several limitations. First, it is difficult to model transparent surfaces with our formulation. The reconstruction quality degrades when there are not enough visible information from input images (\eg sparse in viewpoints or severely occluded) and an example of failure cases is given in Figure~\ref{fig:failure}. 
There also exist trade-offs between the smoothness and high-frequency details due to the normal regularization, since it accumulates the vicinity information to alleviate the surface normal ambiguity.
Further, since we introduce UDF for better representation ability, we need additional meshing tools like MeshUDF~\cite{DBLP:journals/corr/abs-2111-14549} or SPSR~\cite{kazhdan2013screened} which may introduce more reconstruction errors. 

\vspace{-3mm}
\paragraph{Conclusions.}
We propose \OurNetName{}, a novel UDF-based volume rendering approach to achieve high-fidelity multi-view reconstruction for arbitrary shapes with both open and closed surfaces from 2D images with or without masks. NeUDF outperforms the state-of-the-art methods both qualitatively and quantitatively, especially on complex surfaces with open boundaries. Therefore, our \OurNetName{} can play a crucial role in real-world 3D applications. In future work, we can extend our formulation for better reconstruction of transparent surfaces. Enhancing our \OurNetName{} to support sparse input images is also an interesting future direction.

    \section*{Acknowledgment}
This work was supported by CCF-Tencent Open Fund, the Beijing Municipal Natural Science Foundation for Distinguished Young Scholars (No. JQ21013), the National Natural Science Foundation of China (No. 62061136007) and the Youth Innovation Promotion Association CAS.
    
    {
        \small
        \bibliographystyle{utils/ieee_fullname}
        \bibliography{reference}
    }

\newpage
\appendix
\appendixpage

\section{Overview}

In the main paper, we introduce a novel UDF-based volume rendering approach to achieve high-fidelity multi-view reconstruction for arbitrary shapes with both open and closed surfaces. This supplemental material consists of detailed proofs, implementation details and additional results of multi-view reconstruction.
All the sections are organized as follows:
\begin{itemize}
    \item Section~\ref{sec:bias} analyzes the inherent bias in color rendering of the naive UDF solution based on SDF renderer. 
    \item Section~\ref{sec:proofs} provides detailed proofs of the unbiased and occlusion-aware properties of our proposed \OurNetName{}.
    \item Section~\ref{sec:implementation_details} provides implementation details on network architecture (Section~\ref{subsec:network_architecture}), training details (Section~\ref{subsec:training_details}) and data preparation (Section~\ref{subsec:data_preparation}).
    \item Section~\ref{sec:results} provides additional qualitative results of multi-view reconstruction.
\end{itemize}

\section{Bias in Naive UDF solution based on SDF renderer}
\label{sec:bias}
In this section we illustrate the bias of color rendering introduced by the naive UDF solution based on SDF renderer, 
which directly extends the weight of NeuS to UDF. 
The bias causes inherent geometric error like redundant surfaces and floating noises.

To apply the naive UDF solution based on the SDF renderer of NeuS, we denote the rendered color $C(o, v)$:
\begin{equation}\label{eq:color_neus}
    C(o, v)=\int_{0}^{+\infty} w_n(t)c(p(t), v) \mathrm{d}t,
\end{equation}
where $(o, v)$ are the origin and view direction of the sample ray, $c(x, v)$ the color at position $x$ along the view direction $v$, and $w_n(t)$ the rendering weight of NeuS:
\begin{gather}
    w_{n}(t)=\rho_s(t)e^{-\int_0^t\rho_s(u)\mathrm{d}u} \label{eq:weight_neus}
    \\
    \rho_s(t)=\max\{{-\frac{\frac{\partial (\Phi_s \circ \Psi \circ p)}{\partial t}(t)}{\Phi_s \circ \Psi \circ p(t)}, 0}\} \label{eq:rho}
\end{gather}
where $\rho_s(t)$ denotes the opaque density of NeuS, $\Phi_s(d)$ the Sigmoid function, and $\Psi(x)$ the UDF value at position $x$. The learnable parameter $s$ controls the distribution of the Sigmoid function, which is expected to increase to infinity during training. 

Assume that the ray linearly crosses the open surface in its local neighbor, \eg, there exists an interval $(t^l, t^r)$, the intersection point $t^* \in (t^l, t^r)$, which satisfies:
\begin{equation}
    \Psi \circ p(t)=\left|\cos{\theta}\right|\cdot\left|t-t^*\right|, \forall t\in (t^l, t^r), 
\end{equation}
where $\theta$ is the angle between the view direction and the surface normal. 

In UDF, the color $C(o, v)$ rendered based on SDF renderer, 
Equ. \ref{eq:color_neus}, 
consists of inherent bias and inconsistency of the geometry. 
Denote the first intersection point $t_0^*$ and its corresponding interval $(t_0^l, t_0^r)$, the bias can be formularized as below:
\begin{equation}\label{equ:c0.5}
    \lim_{s\to\infty}C(o, v)=0.5c(p(t_0^*), v)+\frac{2^{k}-1}{2^{k+1}}c_m + \frac{1}{2^{k+1}}c_n,
\end{equation}
where $k$ is the number of intersection points along the ray, $c_m$ the undesired mixture of colors from invisible surfaces and $c_n$ the colors from floating noise induced by the rendering bias. 
The parameter $s$ decides the weight distribution of colors along the ray, and is supposed to increase towards infinity during training. 

Note that the weight distribution corresponding to Equ.~\ref{equ:c0.5} satisfies the local maximal constraint discussed in NeuS, \emph{i.e.} the weight attains local maxima at each intersection point (locally unbiased). 
But the local maximal constraint is not sufficient for an unbiased rendering for open surfaces due to the volume-surface representation discrepancy. 
The volume rendering relies on the volume-level color fusion for optimization, while the ground-truth color is exactly the surface color at the intersection point of the sample ray and the first intersected surface. 
A self-consistent rendering procedure should be able to address this volume-surface discrepancy, \emph{i.e.} the color fusion range should be limited as close to the first intersection point as possible (globally unbiased).
Otherwise the network is not able to converge to a surface representation through volume rendering. 
Note that the weight of NeuS is globally and locally unbiased for SDF, but not globally unbiased for UDF, and this difference comes from the difference of the value domains of SDF and UDF. 

To illustrate the detailed causation of $c_m$ and $c_n$, we first prove that:
\begin{gather}
    \lim_{s\to\infty}\int_0^{t_0^l}w_n(t)\mathrm{d}t=0\label{limit_bl}
    \\
    \lim_{s\to\infty}\int_{t_0^l}^{t_0^r}w_n(t)\mathrm{d}t=0.5\label{limit_l},
\end{gather}
which means that the output color consists of undesired bias whose weight sums to 0.5, and the bias cannot be corrected through training. 
Then we show the detailed distribution of the bias $c_m$ and $c_n$ for corroboration. 

\paragraph{Proof of Equ.~\ref{limit_bl}.}
Specifically, to prove Equ.~\ref{limit_bl}, we have:
\begin{equation}\label{w_neus_before_l}
    \begin{split}
    &\int_0^{t_0^l}w_n(t)\mathrm{d}t
    \\
    =&\int_0^{t_0^l}\rho_s(t)e^{-\int_0^t\rho_s(u)\mathrm{d}u}\mathrm{d}t
    \\
    =&\int_0^{t_0^l}-\frac{\partial}{\partial t}e^{-\int_0^t\rho_s(u)\mathrm{u}}\mathrm{d}t
    \\
    =&-e^{-\int_0^t\rho_s(u)\mathrm{d}u}|_0^{t_0^l}
    \\
    =&-e^{-\int_0^{t_0^l}\rho_s(u)\mathrm{d}u}+1
    \\
    =&-e^{-\int_0^{t_0^l}\max\{{\frac{\frac{\partial (\Phi_s \circ \Psi \circ p)}{\partial u}(u)}{\Phi_s \circ \Psi \circ p(u)}, 0}\}\mathrm{d}u}+1
    \end{split}
\end{equation}

It follows that:
\begin{equation}
    \begin{split}
    &\int_0^{t_0^l}w_n(t)\mathrm{d}t
    \\
    \leqslant&-e^{-\int_0^{t_0^l}\left|{\frac{\frac{\partial (\Phi_s \circ \Psi \circ p)}{\partial u}(u)}{\Phi_s \circ \Psi \circ p(u)}}\right|\mathrm{d}u}+1
    \\
    =&-e^{-\int_0^{t_0^l}\left|{\frac{\frac{\partial \Phi_s \circ \Psi \circ p(u)}{\partial \Psi \circ p(u)}\cdot\frac{\partial \Psi \circ p(u)}{\partial u}}{\Phi_s \circ \Psi \circ p(u)}}\right|\mathrm{d}u}+1
    \\
    =&-e^{-\int_0^{t_0^l}\left|{\frac{{\Phi_s' \circ \Psi \circ p(u)}\cdot\frac{\partial \Psi \circ p(u)}{\partial u}}{\Phi_s \circ \Psi \circ p(u)}}\right|\mathrm{d}u}+1
    \\
    =&-e^{-\int_0^{t_0^l}{\frac{\left|{\Phi_s' \circ \Psi \circ p(u)}\right|\cdot\left|\frac{\partial \Psi \circ p(u)}{\partial u}\right|}{\left|\Phi_s \circ \Psi \circ p(u)\right|}}\mathrm{d}u}+1
    \end{split}
\end{equation}

Denote that:
\begin{align}
    A &= \left| \Phi_s' \circ \Psi \circ p(u) \right|
    \\
    B &= \left| \frac{\partial \Psi \circ p(u)}{\partial u} \right|
    \\
    C &= \left| \Phi_s \circ \Psi \circ p(u) \right|
\end{align}

We have:
\begin{equation}
    \int_0^{t_0^l}w_n(t)\mathrm{d}t=-e^{-\int_0^{t_0^l}{\frac{A \cdot B}{C}}\mathrm{d}u}+1
\end{equation}

Because $t_0^*$ is the first zero point of $\Psi\circ p(t)$ and $\Psi(x)$ is a continuous function, there is:
$$\exists \Psi_{min}>0, s.t., \Psi\circ p(t)>\Psi_{min}, \forall t\in(0, t_0^l).$$

Note that $\Phi_s(x)$ is the Sigmoid function $\Phi_s(x)=(1+e^{-s*x})^{-1}$, and $\frac{\partial \Psi \circ p(u)}{\partial u}$ is the gradient of the UDF along the ray. 
We have:
\begin{align}
    C &= \left| \Phi_s \circ \Psi \circ p(u) \right| 
    \\
    &= (1+e^{-s\cdot\Psi\circ p(u)})^{-1}
    \\
    &> (1+e^{-s\cdot\Psi_{min}})^{-1}
    \\
    &>0.5
    \\
    B &= \left| \frac{\partial \Psi \circ p(u)}{\partial u} \right| < 1
\end{align}
and $\forall \epsilon>0, \exists S=\max\{1, \frac{-4t_0^l}{\ln{(1-\epsilon)}\cdot\Psi_{min}^2}\}, s.t., \forall s>S,$ there is:
\begin{equation}
\begin{split}
    A &= \left| \Phi_s' \circ \Psi \circ p(u) \right| 
    \\
    &= \frac{s\cdot e^{-s\cdot\Psi\circ p(t)}}{(1+s\cdot e^{-s\cdot\Psi\circ p(t))^2}} 
    \\
    &\leqslant \frac{2}{\Psi^2\circ p(t) \cdot s}
    \\
    &\leqslant \frac{2}{\Psi_{min}^2 \cdot s}
    \\
    &\leqslant \frac{2}{\Psi_{min}^2\cdot\frac{-4t_0^l}{\ln{(1-\epsilon)}\cdot\Psi_{min}^2}}
    \\
    &= \frac{-0.5\ln{(1-\epsilon)}}{t_0^l}
\end{split}
\end{equation}

It follows that:
\begin{equation}\label{limit_bl_upper}
\begin{split}
    \int_0^{t_0^l}w_n(t)\mathrm{d}t
    &=-e^{-\int_0^{t_0^l}{\frac{A \cdot B}{C}}\mathrm{d}u}+1
    \\
    &<-e^{-\int_0^{t_0^l}{\frac{\frac{0.5\ln{(1-\epsilon)}}{t_0^l} \cdot 1}{0.5}}\mathrm{d}u}+1
    \\
    &=-e^{-\int_0^{t_0^l}{\frac{\ln{(1-\epsilon)}}{t_0^l}}\mathrm{d}u}+1
    \\
    &=-e^{-t_0^l\cdot \frac{\ln{(1-\epsilon)}}{t_0^l}}+1
    \\
    &=-e^{\ln{(1-\epsilon)}}+1
    \\
    &=-(1-\epsilon)+1
    \\
    &=\epsilon
\end{split}
\end{equation}

This leads to:
\begin{equation}\label{limit_w_neus_before_l}
\begin{split}
    &\lim_{s\to\infty}\int_0^{t_0^l}w_n(t)\mathrm{d}t
    \\
    =&\lim_{s\to\infty}(-e^{-\int_0^{t_0^l}{\frac{A \cdot B}{C}}\mathrm{d}u}+1)
    \\
    =&0
\end{split}
\end{equation}
    
The Equ.~\ref{limit_w_neus_before_l} means that the weight before the first intersection of the ray converges against zero during training, so the output color composites no color before the first intersected surface. 
This completes the proof of Equ.~\ref{limit_bl}.

\paragraph{Proof of Equ.~\ref{limit_l}.}
Then we give the proof of Equ.~\ref{limit_l}.
Same as the derivation of Equ. \ref{w_neus_before_l}, we have:
\begin{equation}\label{sum_w_n_l}
\begin{split}
    &\int_{t_0^l}^{t_0^*}w_n(t)\mathrm{d}t
    \\
    =&\int_{t_0^*}^{t_0^*}\rho_s(t)e^{-\int_0^t\rho_s(u)\mathrm{d}u}\mathrm{d}t
    \\
    =&\int_{t_)^l}^{t_0^*}-\frac{\partial}{\partial t}e^{-\int_0^t\rho_s(u)\mathrm{u}}\mathrm{d}t
    \\
    =&-e^{-\int_0^t\rho_s(u)\mathrm{d}u}|_{t_0^l}^{t_0^*}
    \\
    =&-e^{-\int_{0}^{t_0^*}\rho_s(u)\mathrm{d}u}+e^{-\int_{0}^{t_0^l}\rho_s(u)\mathrm{d}u}
    \\
    =&-e^{-\int_{0}^{t_0^l}\rho_s(u)\mathrm{d}u-\int_{t_0^l}^{t_0^*}\rho_s(u)\mathrm{d}u}+e^{-\int_{0}^{t_0^l}\rho_s(u)\mathrm{d}u}
    \\
    =&e^{-\int_{0}^{t_0^l}\rho_s(u)\mathrm{d}u}(-e^{{-\int_{t_0^l}^{t_0^*}\rho_s(u)\mathrm{d}u}}+1)
\end{split}
\end{equation}

Note that when $t\in (t_0^l, t_0^*)$, we have:
\begin{equation}
    \frac{\partial\Psi\circ p(t)}{\partial t}=-\left|\cos{\theta}\right|. 
\end{equation}

It follows that:
\begin{equation}
\begin{split}
    & -e^{{-\int_{t_0^l}^{t_0^*}\rho_s(u)\mathrm{d}u}}+1
    \\
    =& -e^{{-\int_{t_0^l}^{t_0^*}\max\{{\frac{\frac{\partial (\Phi_s \circ \Psi \circ p)}{\partial u}(u)}{\Phi_s \circ \Psi \circ p(u)}, 0}\}\mathrm{d}u}}+1
    \\
    =& -e^{{-\int_{t_0^l}^{t_0^*}\left|\frac{\frac{\partial (\Phi_s \circ \Psi \circ p)}{\partial u}(u)}{\Phi_s \circ \Psi \circ p(u)}\right|\mathrm{d}u}}+1
    \\
    =& -e^{{-\int_{t_0^l}^{t_0^*}\left|\frac{\partial}{\partial u}\ln{\Phi_s\circ\Psi\circ p(u)}\mathrm{d}u\right|}}+1
    \\
    =& -e^{{-\int_{t_0^l}^{t_0^*}-\frac{\partial}{\partial u}\ln{\Phi_s\circ\Psi\circ p(u)}\mathrm{d}u}}+1
    \\
    =& -e^{\ln\Phi_s\circ\Psi\circ p(t_0^*)-\ln\Phi_s\circ\Psi\circ p(t_0^l)}+1
    \\
    =& -\frac{e^{\ln\Phi_s\circ\Psi\circ p(t_0^*)}}{e^{\ln\Phi_s\circ\Psi\circ p(t_0^l)}}+1
    \\
    =& -\frac{\Phi_s\circ\Psi\circ p(t_0^*)}{\Phi_s\circ\Psi\circ p(t_0^l)}+1
\end{split}
\end{equation}

Since $t_0^*$ is the intersection point, we have $\Psi\circ p(t_0^*)=0$ and $\Phi_s\circ\Psi\circ p(t_0^*)=0.5$.
It follows that:
\begin{equation}\label{limit_l_upper}
\begin{split}
    & -e^{{-\int_{t_0^l}^{t_0^*}\rho_s(u)\mathrm{d}u}}+1
    \\
    =& -\frac{\Phi_s\circ\Psi\circ p(t_0^*)}{\Phi_s\circ\Psi\circ p(t_0^l)}+1
    \\
    =& -\frac{0.5}{\Phi_s\circ\Psi\circ p(t_0^l)}+1
    \\
    \leqslant& -\frac{0.5}{1}+1=0.5
\end{split}
\end{equation}

$\forall \epsilon>0, \exists S=\frac{-\ln2\epsilon}{\Psi\circ p(t_0^l)}, s.t., \forall s>S,$
\begin{equation}\label{limit_l_lower}
\begin{split}
    & -e^{{-\int_{t_0^l}^{t_0^*}\rho_s(u)\mathrm{d}u}}+1
    \\
    =& -\frac{0.5}{\Phi_s\circ\Psi\circ p(t_0^l)}+1
    \\
    =& -\frac{0.5}{(1+e^{-s\cdot\Psi\circ p(t_0^l)})^{-1}}+1
    \\
    \geqslant& -\frac{0.5}{(1+e^{-\frac{-\ln2\epsilon}{\Psi\circ p(t_0^l)} \Psi\circ p(t_0^l)})^{-1}}+1
    \\
    =& -\frac{0.5}{(1+e^{\ln2\epsilon})^{-1}}+1
    \\
    =& -\frac{0.5}{(1+2\epsilon)^{-1}}+1
    \\
    =& -\epsilon+0.5
\end{split}
\end{equation}

The Equ.~\ref{limit_l_upper} and \ref{limit_l_lower} derive that:
\begin{equation}\label{limit_l_r}
    \lim_{s\to\infty}(-e^{{-\int_{t_0^l}^{t_0^*}\rho_s(u)\mathrm{d}u}}+1)=0.5
\end{equation}

It has been proved in Equ.~\ref{limit_bl_upper} that: 
\begin{equation}
    \lim_{s\to\infty}(-e^{-\int_0^{t_0^l}{\rho(u)}\mathrm{d}u}+1)=0, i.e.,
\end{equation}
\begin{equation}\label{limit_l_l}
    \lim_{s\to\infty}(e^{-\int_0^{t_0^l}{\rho(u)}\mathrm{d}u})=1
\end{equation}

The equations \ref{sum_w_n_l}, \ref{limit_l_r} and \ref{limit_l_l} together derive that:
\begin{equation}\label{limit_w_neus_l}
\begin{split}
    &\lim_{s\to\infty}\int_{t_0^l}^{t_0^r}w_n(t)\mathrm{d}t
    \\
    =&\lim_{s\to\infty}(e^{-\int_{0}^{t_0^l}\rho_s(u)\mathrm{d}u}(-e^{{-\int_{t_0^l}^{t_0^*}\rho_s(u)\mathrm{d}u}}+1))
    \\
    =&0.5
\end{split}
\end{equation}

The Equ.~\ref{limit_w_neus_l} determines that the rendered color $C(o, v)$ of NeuS in UDF cannot converge to the ground-truth color $c(p(t_0^*), v)$ as up to half of the weight is not constrained, which causes the mixed rendering color with undesired bias and inherent geometric error. This completes the proof of Equ.~\ref{limit_l}.

\paragraph{Distribution of Bias.}
Further, we illustrate the components of the bias, \eg, $c_m$ and $c_n$, and show the corresponding distribution. 

For $t\in (t_0^*,t_1^*)$, where $t_0^*$ and $t_1^*$ denotes the first and second intersection points along the ray $p(t)$.
Consider that:
\begin{equation}
\begin{split}
    w_n(t)&=\rho_s(t)e^{-\int_0^t\rho_s(u)\mathrm{d}u}
    \\
    &=\rho_s(t)e^{-\int_{t_0^*}^t\rho_s(u)\mathrm{d}u}\cdot e^{-\int_0^{t_0^*}\rho_s(u)\mathrm{d}u}
\end{split}
\end{equation}

As is proved, $\lim_{s\to\infty}e^{-\int_0^{t_0^*}\rho_s(u)\mathrm{d}u}=0.5$, there is:
\begin{equation}
    w_n(t_1^*)=0.5\rho_s(t)e^{-\int_{t_0^*}^{t_1^*}\rho_s(u)\mathrm{d}u}
\end{equation}

According to the assumption that $\exists(t_1^l,t_1^r)\ni t_1^*$, the UDF value $\Psi(t)$ along the ray is linear for $t\in(t_1^l,t_1^r)$.
So similarly we can prove that:
\begin{equation}
\begin{split}
    &\lim_{s\to\infty}\int_0^{t_1^*}w_n(t)\mathrm{d}t
    \\
    =&0.5\lim_{s\to\infty}\int_0^{t_1^*}\rho_s(t)e^{-\int_{t_0^*}^t\rho_s(u)\mathrm{d}u}\mathrm{d}t
    \\
    =&0.25
\end{split}
\end{equation}

Consequently, for any given $k>0$, we have:
\begin{equation}
    \lim_{s\to\infty}\int_0^{t_k^*}w_n(t)\mathrm{d}t=\frac{1}{2^{k+1}}
\end{equation}

The colors of the $k$ invisible surfaces are mixed to the output color $C(o, v)$, whose weight sums to $\frac{2^{k}-1}{2^{k+1}}$. 
The mixed colors integral $c_m$ leads to the undesired bias $\frac{2^{k}-1}{2^{k+1}}c_m$, which cannot be corrected during training. 
The last weight $1-0.5-\frac{2^{k}-1}{2^{k+1}}=\frac{1}{2^{k+1}}$ comes from the disturbance besides the neighborhood of surfaces, and leads to new redundant surfaces during training. 
The bias $c_m$ and $c_n$ case inherent geometric error like redundant surfaces and floating noises in invisible space.

\section{Proofs of Unbiased and Occlusion-aware properties of NeUDF}
\label{sec:proofs}

In this subsection we illustrate the capability of NeUDF for UDF learning from three aspects. 
First we show that different from NeuS, NeUDF avoids the $c_m$ and $c_n$ which cause the biased rendering color and inherent geometric error in UDF. 
Then we give the proofs of the unbiased and occlusion-aware properties of NeUDF respectively. 

\subsection{Avoidance of $c_m$ and $c_n$ in NeUDF.}

Before providing the detailed proofs of the unbiased and occlusion-aware properties of NeUDF, we briefly show that NeUDF is free from the undesired colors $c_m$ and $c_n$ by introducing the new rendering weight function:
\begin{equation}
    w_r(t)=\tau_r(t)e^{-\int_0^t\tau_r(u)\mathrm{d}u},
\end{equation}
\begin{equation}
    \tau_r(t)=\left|\frac{\frac{\partial \varsigma_r\circ\Psi\circ p}{\partial t}(t)}{\varsigma_r\circ\Psi\circ p(t)}\right|,
\end{equation}
where $\varsigma_r(d)$ satisfies that:
\begin{gather}
    \varsigma_r(0)=0, \lim_{d\to\infty}=1,
    \\
    \forall d>0, \varsigma_r'(d)>0, \varsigma_r''(d)<0.
\end{gather}

Similar to the derivation in \ref{sec:bias}, there is:
\begin{equation}
    \lim_{r\to\infty}\int_0^{t_0^l}w_r(t)\mathrm{d}t=0
\end{equation}
and
\begin{equation}
\begin{split}
    &\lim_{r\to\infty}\int_{t_0^l}^{t_0^*}w_n(t)\mathrm{d}t
    \\
    =&\lim_{r\to\infty}e^{-\int_{0}^{t_0^l}\tau_r(u)\mathrm{d}u}(-e^{{-\int_{t_0^l}^{t_0^*}\tau_r(u)\mathrm{d}u}}+1)
    \\
    =&\lim_{r\to\infty}-e^{{-\int_{t_0^l}^{t_0^*}\tau_r(u)\mathrm{d}u}}+1
\end{split}
\end{equation}

When $t\in(t_0^l,t_0^r)$, there is:
\begin{equation}
    \frac{\partial\Psi\circ p(t)}{\partial t}=-\left|\cos\theta\right|<0
\end{equation}
We have:
\begin{equation}
\begin{split}
    &-e^{{-\int_{t_0^l}^{t_0^*}\tau_r(u)\mathrm{d}u}}+1
    \\
    =&-e^{{-\int_{t_0^l}^{t_0^*}\left|\frac{\frac{\partial \varsigma_r\circ\Psi\circ p}{\partial u}(u)}{\varsigma_r\circ\Psi\circ p(u)}\right|\mathrm{d}u}}+1
    \\
    =&-e^{{-\int_{t_0^l}^{t_0^*}\left|\frac{\partial}{\partial u}\ln\varsigma_r\circ\Psi\circ p(u)\right|\mathrm{d}u}}+1
    \\
    =&-e^{{\int_{t_0^l}^{t_0^*}\frac{\partial}{\partial u}\ln\varsigma_r\circ\Psi\circ p(u)\mathrm{d}u}}+1
    \\
    =&-e^{\ln\varsigma_r\circ\Psi\circ p(t_0^*)-\ln\varsigma\circ\Psi\circ p(t_l)}+1
    \\
    =&-\frac{\varsigma_r\circ\Psi\circ p(t_0^*)}{\varsigma\circ\Psi\circ p(t_l)}+1
    \\
    =&-0+1
    \\
    =&1
\end{split}
\end{equation}

So we have:
\begin{equation}
    \lim_{r\to\infty}\int_{t_0^l}^{t_0^*}w_n(t)\mathrm{d}t=1
\end{equation}

It follows that:
\begin{equation}
\begin{split}
    \lim_{r\to\infty}C(o, v) =& \lim_{r\to\infty}\int_{t_0^l}^{t_0^*}w_n(t)\mathrm{d}t\cdot c(p(t_0^*), v) 
    \\
    &+ (1-\lim_{r\to\infty}\int_{t_0^l}^{t_0^*}w_n(t)\mathrm{d}t)\cdot c_m
    \\
    =&c(p(t_0^*), v) 
\end{split}
\end{equation}

It indicates that NeUDF avoids the limitation introduced by the undesired mixture $c_m$ (and $c_n$). 
The detailed proof of unbiased property of NeUDF is provided in the next section.

\subsection{Proof of Unbiased Property in NeUDF.}
Intuitively, the rendering weight function should be unbiased, \ie, more contribution should come from the intersection point than its neighbor. 
In this subsection we prove that NeUDF is unbiased:
\begin{itemize}
    \item Given the ray $p(t)$ and the UDF $\Psi(x)$, the weight of rendering $w_r(t)$ in NeUDF attains a locally maximum value at a intersection point $t^*$. 
\end{itemize}

Assume that the weight $w_r(t)$ is a linear function within the local neighborhood $(t^l, t^r)$ of the zero point $t^*\in(t^l, t^r)$. 
We consider the intervals $(t^l, t^*)$ and $(t^*, t^r)$ respectively. 
For $t\in(t^l, t^*)$, we have:
\begin{equation}
\begin{split}
    w_r(t)&=\tau_r(t)e^{-\int_0^t\tau(u)\mathrm{d}u}
    \\
    &=\tau_r(t)e^{-\int_0^{t^l}\tau(u)\mathrm{d}u}e^{-\int_{t^l}^t\tau(u)\mathrm{d}u}
    \\
    &=\tau_r(t)e^{-\int_0^{t^l}\tau(u)\mathrm{d}u}e^{-\int_{t^l}^t\left|\frac{\frac{\partial \varsigma_r\circ\Psi\circ p}{\partial u}(u)}{\varsigma_r\circ\Psi\circ p(t)}\right|\mathrm{d}u}
    \\
    &=\tau_r(t)e^{-\int_0^{t^l}\tau(u)\mathrm{d}u}e^{-\int_{t^l}^t\left|\frac{\partial}{\partial u}\ln\varsigma_r\circ\Psi\circ p(u)\right|\mathrm{d}u}
    \\
    &=\tau_r(t)e^{-\int_0^{t^l}\tau(u)\mathrm{d}u}e^{-\int_{t^l}^t\frac{\partial}{\partial u}\ln\varsigma_r\circ\Psi\circ p(u)\mathrm{d}u}
    \\
    &=\tau_r(t)e^{-\int_0^{t^l}\tau(u)\mathrm{d}u}e^{\ln\varsigma_r\circ\Psi\circ p(t)-\ln\varsigma_r\circ\Psi\circ p(t^l)}
    \\
    &=\tau_r(t)e^{-\int_0^{t^l}\tau(u)\mathrm{d}u}\frac{e^{\ln\varsigma_r\circ\Psi\circ p(t)}}{e^{\ln\varsigma_r\circ\Psi\circ p(t^l)}}
    \\
    &=\tau_r(t)e^{-\int_0^{t^l}\tau(u)\mathrm{d}u}\frac{\varsigma_r\circ\Psi\circ p(t)}{\varsigma_r\circ\Psi\circ p(t^l)}
    \\
    &=\left|\frac{\frac{\partial \varsigma_r\circ\Psi\circ p}{\partial u}(u)}{\varsigma_r\circ\Psi\circ p(t)}\right|e^{-\int_0^{t^l}\tau(u)\mathrm{d}u}\frac{\varsigma_r\circ\Psi\circ p(t)}{\varsigma_r\circ\Psi\circ p(t^l)}
    \\
    &=\frac{\left|\frac{\partial\varsigma_r\circ\Psi\circ p(t)}{\partial\Psi\circ p(t)}\right|\cdot\left|\frac{\partial\Psi\circ p(t)}{\partial t}\right|}{\left|\varsigma_r\circ\Psi\circ p(t)\right|}e^{-\int_0^{t^l}\tau(u)\mathrm{d}u}\frac{\varsigma_r\circ\Psi\circ p(t)}{\varsigma_r\circ\Psi\circ p(t^l)}
    \\
    &=\frac{\left|\varsigma_r'\circ\Psi\circ p(t)\right|\cdot\left|\cos\theta\right|}{\left|\varsigma_r\circ\Psi\circ p(t)\right|}e^{-\int_0^{t^l}\tau(u)\mathrm{d}u}\frac{\varsigma_r\circ\Psi\circ p(t)}{\varsigma_r\circ\Psi\circ p(t^l)}
    \\
    &=\frac{\varsigma_r'\circ\Psi\circ p(t)\cdot\left|\cos\theta\right|}{\varsigma_r\circ\Psi\circ p(t)}e^{-\int_0^{t^l}\tau(u)\mathrm{d}u}\frac{\varsigma_r\circ\Psi\circ p(t)}{\varsigma_r\circ\Psi\circ p(t^l)}
    \\
    &=\frac{\varsigma_r'\circ\Psi\circ p(t)\cdot\left|\cos\theta\right|\cdot e^{-\int_0^{t^l}\tau_r(u)\mathrm{d}u}}{\varsigma_r\circ\Psi\circ p(t^l)}
\end{split}
\end{equation}

For a given parameter $r$, $\varsigma_r\circ\Psi\circ p(t^l)$, $e^{-\int_0^{t^l}\tau_r(u)\mathrm{d}u}$ and $\left|\cos\theta\right|$ are all constant. 
So we have:
\begin{equation}
    w_r(t)=A\cdot\varsigma_r'\circ\Psi\circ p(t), A=\frac{\left|\cos\theta\right|\cdot e^{-\int_0^{t^l}\tau_r(u)\mathrm{d}u}}{\varsigma_r\circ\Psi\circ p(t^l)},
\end{equation}
where A is a fixed positive number for any given $r$. 

Note that $\varsigma_r'(d)>0, \varsigma_r''(d)<0$, it follows that:
\begin{equation}\label{unbiased_l}
    w_r(t_1)>w_r(t_2), \forall t_1>t_2, t_1,t_2\in(t^l,t^*).
\end{equation}

For $t\in(t^*,t^r)$, we have:
\begin{equation}
    \tau_r(t)=\left|\frac{\frac{\partial \varsigma_r\circ\Psi\circ p}{\partial t}(t)}{\varsigma_r\circ\Psi\circ p(t)}\right|=\frac{\varsigma_r'\circ\Psi\circ p(t)\cdot\left|\cos\theta\right|}{\varsigma_r\circ\Psi\circ p(t)}
\end{equation}

$\forall t_1>t_2, t_1,t_2\in(t^*,t^r)$, there is:
\begin{equation}
    \tau_r(t_1)<\tau_r(t_2)
\end{equation}
\begin{equation}
    e^{-\int_0^{t_1}\tau_r(u)\mathrm{d}u}<e^{-\int_0^{t_2}\tau_r(u)\mathrm{d}u}
\end{equation}
It follows that:
\begin{equation}\label{unbiased_r}
    w_r(t_1)<w_r(t_2), \forall t_1>t_2, t_1,t_2\in(t^l,t^*).
\end{equation}

The Equ.~\ref{unbiased_l} and \ref{unbiased_r} indicates that the point closer to the zero point is with higher weight value. 
Note that the proof does not require a strict zero point $t^*$, \ie, the property holds true when there is a small perturbation $\Delta$ to the zero point $t^*$: $\Psi\circ p(t^*)=\Delta>0$.

Empirically, the zero point of the UDF is encoded as a small positive number, so the weight function $w_r(t)$ is continuous along the ray. 
Therefore we have:
\begin{equation}
    w_r(t^*)>w_r(t), \forall t\in(t^l,t^r), t\neq t^*
\end{equation}
This completes the proof.

\subsection{Proof of Occlusion-aware Property in NeUDF.}

In this subsection we prove that NeUDF is occlusion-aware. 
Intuitively, for two parts of the sample ray with the same UDF value, we hope that more contribution of the output colors is from the part closer to the camera. 
That is, the closer surfaces are more likely to have higher weight. 

Specifically, given two surfaces $\mathrm{S}_1$ and $\mathrm{S}_2$ such that $\mathrm{S}_1$ is closer to the camera, for two corresponding points $p(t_1)$ and $p(t_2)$ with the same UDF value, we have:
\begin{equation}
    \int_{t_1}^{t_1+\delta}w_r(t)\mathrm{d}d_1(t)>\int_{t_2}^{t_2+\delta}w_r(t)\mathrm{d}d_2(t),
\end{equation}
where $d_i(t)$ denotes the distance between the location $p(t)$ and the surface $\mathrm{S}_i$, and $\delta$ denotes the small step length. 

\begin{equation}
    \tau_r(t)=\left|\frac{\frac{\partial\varsigma_r\circ\Psi\circ p}{\partial t}(t)}{\varsigma_r\circ\Psi\circ p(t)}\right|=\frac{\left|\varsigma_r'\circ\Psi\circ p(t)\right|\cdot\left|\cos\theta\right|}{\varsigma_r\circ\Psi\circ p(t)}
\end{equation}

For $t_1<t_2, \Psi(t_1)=\Psi(t_2), w_r(t_1), w_r(t_2)>0$, we have:
\begin{equation}
    \frac{\tau_r(t_1)}{\left|\cos\theta_1\right|}=\frac{\left|\varsigma_r'\circ\Psi\circ p(t_1)\right|}{\varsigma_r\circ\Psi\circ p(t_1)}=\frac{\left|\varsigma_r'\circ\Psi\circ p(t_2)\right|}{\varsigma_r\circ\Psi\circ p(t_2)}=\frac{\tau_r(t_2)}{\left|\cos\theta_2\right|}
\end{equation}

\begin{equation}
    e^{-\int_0^{t_1}\tau_r(u)\mathrm{d}u}>e^{-\int_0^{t_2}\tau_r(u)\mathrm{d}u}
\end{equation}

There is:
\begin{equation}
\begin{split}
    \frac{w_r(t_1)}{\left|\cos\theta\right|}&=\frac{\tau_r(t_1)e^{-\int_0^{t_1}\tau_r(t_u)\mathrm{d}u}}{\left|\cos\theta\right|}
    \\
    &>\frac{\tau_r(t_2)e^{-\int_0^{t_2}\tau_r(t_u)\mathrm{d}u}}{\left|\cos\theta\right|}=\frac{w_r(t_2)}{\left|\cos\theta\right|}
\end{split}
\end{equation}

It follows that:
\begin{equation}
    \int_{t_1}^{t_1+\delta}w_r(t)\mathrm{d}d_1(t)=\int_{t_1}^{t_1+\delta}\frac{w_r(t)}{\left|\cos\theta\right|}\mathrm{d}t
\end{equation}
\begin{equation}
    \int_{t_2}^{t_2+\delta}w_r(t)\mathrm{d}d_2(t)=\int_{t_2}^{t_2+\delta}\frac{w_r(t)}{\left|\cos\theta\right|}\mathrm{d}t
\end{equation}
\begin{equation}\label{w1w2}
    \int_{t_1}^{t_1+\delta}w_r(t)\mathrm{d}d_1(t)>\int_{t_2}^{t_2+\delta}w_r(t)\mathrm{d}d_2(t),
\end{equation}
where $d_i(t)$ denotes the distance between the location $p(t_i)$ and the surface $\mathrm{S}_i$. 

The Equ.~\ref{w1w2} indicates that the cumulative weight near the the first intersected surface are higher that the second one. 
This means that more concentration are on the former surface. 
Note that no prior assumption of the existence of other intersected surfaces is required, \ie, the property of occlusion-aware holds true for more than two surface intersections along the ray. 
This completes the proof of the occlusion-aware property.

\section{Implementation Details}
\label{sec:implementation_details}

\subsection{Network Architecture}
\label{subsec:network_architecture}
Similar to IDR~\cite{DBLP:conf/nips/WangLLTKW21} and NeuS~\cite{DBLP:conf/nips/WangLLTKW21}, we use two MLP networks to respectively encode the UDF and the color. 
The input of the UDF network is the spatial location $p(t)$ and the output is the corresponding UDF value along with a 256-dimensional feature vector. 
The UDF network $\Psi(x)$ consists of 8 hidden layers with hidden size of 256, and the activation function is chosen as the Softplus with $\beta=100$ for all hidden layers and the output layer. 
A skip connection is also used to connect the input with the output of the fourth layer. 
The inputs of the color network are the spatial location $p(t)$, the view direction $v$, the gradient $n$ of the UDF network at the spatial location $p(t)$ and the corresponding feature vector derived by the UDF network. 
The color network $c(x, v)$ consists of 4 hidden layers with hidden size of 256. 
Normal regularization is applied before the gradient $n$ of the UDF network is used as the input of the color network. 
Same positional encoding and weight normalization are adopted as in Neus.

\subsection{Training Details}
\label{subsec:training_details}

\paragraph{Discretization.}
We adopt the $\alpha$-compositing to discretize the weight function, 
which divides the sample ray into bins by sampling $n$ points ${p(t_i) = o + t_i|i =1, ..., n, t_i < t_{i+1}}$ and accumulate colors within each bin according to the weight integral:

\begin{equation}
\label{alpha}
    \begin{split}
    \alpha_i&=1-e^{-\int_{t_i}^{t_{i+1}}\tau_r(t)\mathrm{d}t} \\ 
    &=\frac{\left|\varsigma_r\circ\Psi\circ p(t_i)-\varsigma_r\circ\Psi\circ p(t_{i+1})\right|}{\varsigma_r\circ\Psi\circ p(t_i)}.
    \end{split}
\end{equation}

We slightly modify Equ.~\ref{alpha} by:
\begin{equation}
    \alpha_i=\frac{\varsigma_i^{max}-\varsigma_i^{min}}{\varsigma_i^{max}},
\end{equation}
where $\varsigma_i^{max}$ and $\varsigma_i^{min}$ is the maximum and minimum of the set $\{\varsigma_r\circ\Psi\circ p(t_i), \varsigma_r\circ\Psi\circ p(t_{i+1})\}$.

\paragraph{Up Sampling.}
We first formally sample 64 points per ray, and then hierarchically conduct importance sampling on top of the sampling weight $w_s(t)$ for another 64 points:

\begin{equation}
    w_{s}(t)=\tau_s(t)e^{-\int_0^t\tau_s(u)du}, \tau_s(t) = \zeta_s \circ \Psi \circ p(t)
\end{equation}

And $\zeta_s(\cdot)$ satisfies the rules: $\zeta_s(d)>0 \text{ and } \zeta_s'(d)<0, \forall d > 0$.
Intuitively, the $\tau_s(t)$ derived by the monotonically decreasing function is a view-invariant sampling density, and the density has positive correlation with the UDF value. 
To derive the sampling weight $w_s(t)$, the classical volume rendering scheme is applied. 

The weight of the $i^{th}$ sample point $w_s(t_i)$ is slightly modified by:
\begin{equation}
    w'_s(t_i) = \max\{w_s(t_{i+k}), k=-1,0,1\}
\end{equation}
And then the weight $w'_s(t)$ is normalized so that the integral equals to one:
\begin{equation}
    w''_s(t) = \frac{w'_s(t)}{\sum_{i=0}^{n-1} w'_s(t_i)}
\end{equation}

For each iteration we hierarchically conduct the importance sampling for two times, and each time 32 points are sampled. 
The total number of sampling points are 128. 
If no masks are provided, 32 points are randomly sampled in addition outside the unit sphere per ray to represent the outside scene. 
The outside scene is represented with NeRF++~\cite{DBLP:journals/corr/abs-2010-07492}, as used in NeuS~\cite{DBLP:conf/nips/WangLLTKW21}.

\paragraph{Platform.}
The network is trained with ADAM optimizer, and the learning rate warms up to $2\times10^{-4}$ in the first 5k iterations, and decreases to $1*10^{-5}$ by the end of training. 
For each iteration, 512 random rays are sampled from 8 input camera poses randomly selected. 
We train each model for 400k iterations in total for 9 hours for the setting of with mask, and 11 hours for the setting of without mask on a single Nvidia 3090 GPU.

\subsection{Data Preparation
}
\label{subsec:data_preparation}

\begin{figure}[h]

\includegraphics[width=.45\textwidth]{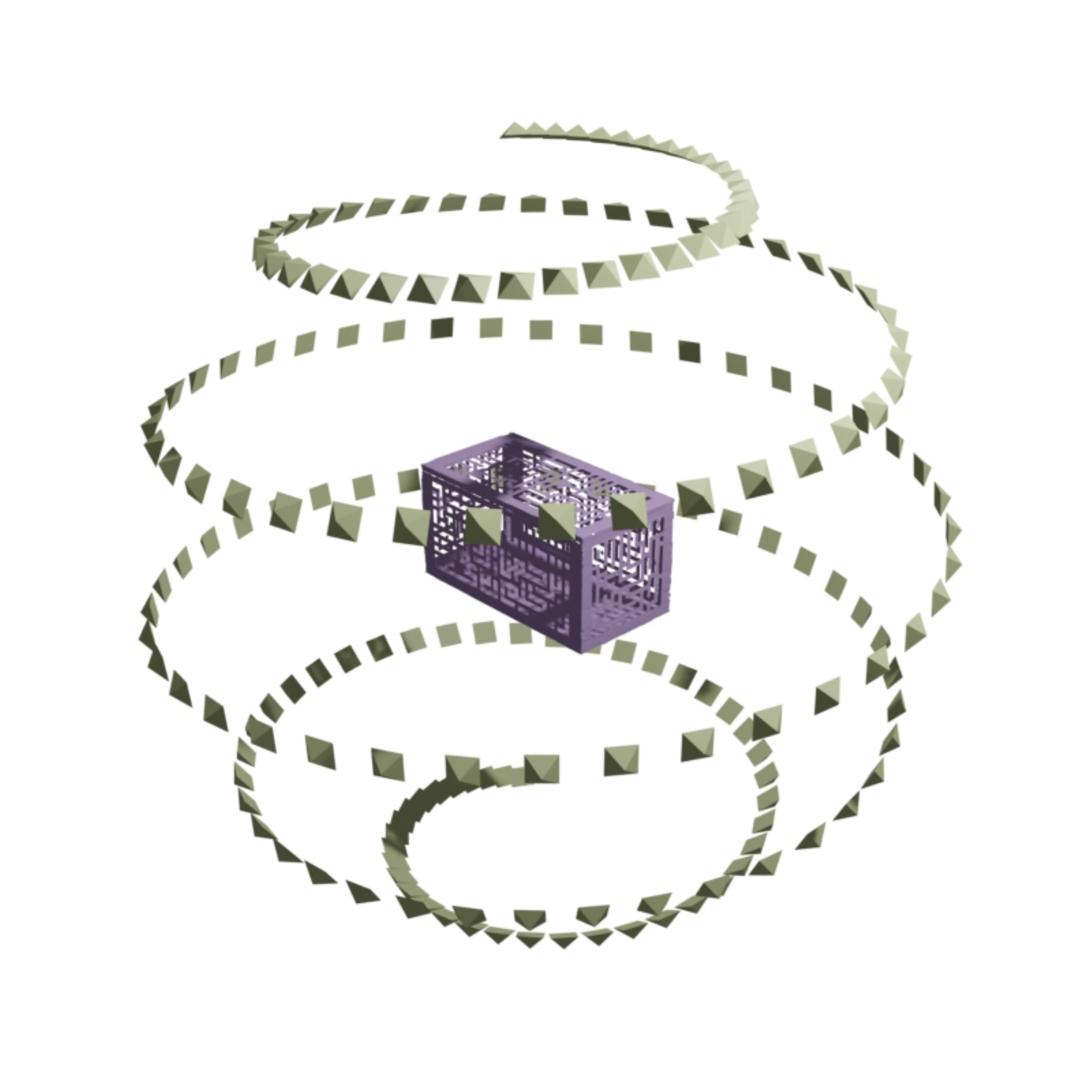}

\caption{Poses of the camera. The camera poses are represented as the yellow pyramid, and the object to reconstruct is represented in purple. }
\label{supp_pose}
\end{figure}

\paragraph{Rendered Data.}
To generate the customized data, we use the pyrender package to render images from the ground-truth objects. 
We rendered 200 views at $800 \times 800$ pixels for each textured mesh or colored point cloud. 
Fig \ref{supp_pose} visualizes the camera poses.
Corresponding masks with black background are provided optionally. 
Only the rendered images and the masks are used as inputs of the network.

\paragraph{Captured Data.}
We additionally captured several real-world objects using the mobile phone. 
The captured images are extracted from the captured videos around the object. 
For the book object we captured 200 images at the resolution of $1920\times1440$.
For the fan object we captured 59 images at the resolution of $3456\times4608$.
For the plant object we captured 200 images at the resolution of $720\times1280$.
All the camera poses are estimated by COLMAP~\cite{DBLP:conf/cvpr/SchonbergerF16, DBLP:conf/eccv/SchonbergerZFP16} and no masks are provided.

\section{Additional Results}
\label{sec:results}

\begin{figure*}[h]

\begin{minipage}[c]{.13\textwidth}
    \centering
    \includegraphics[width=1\linewidth]{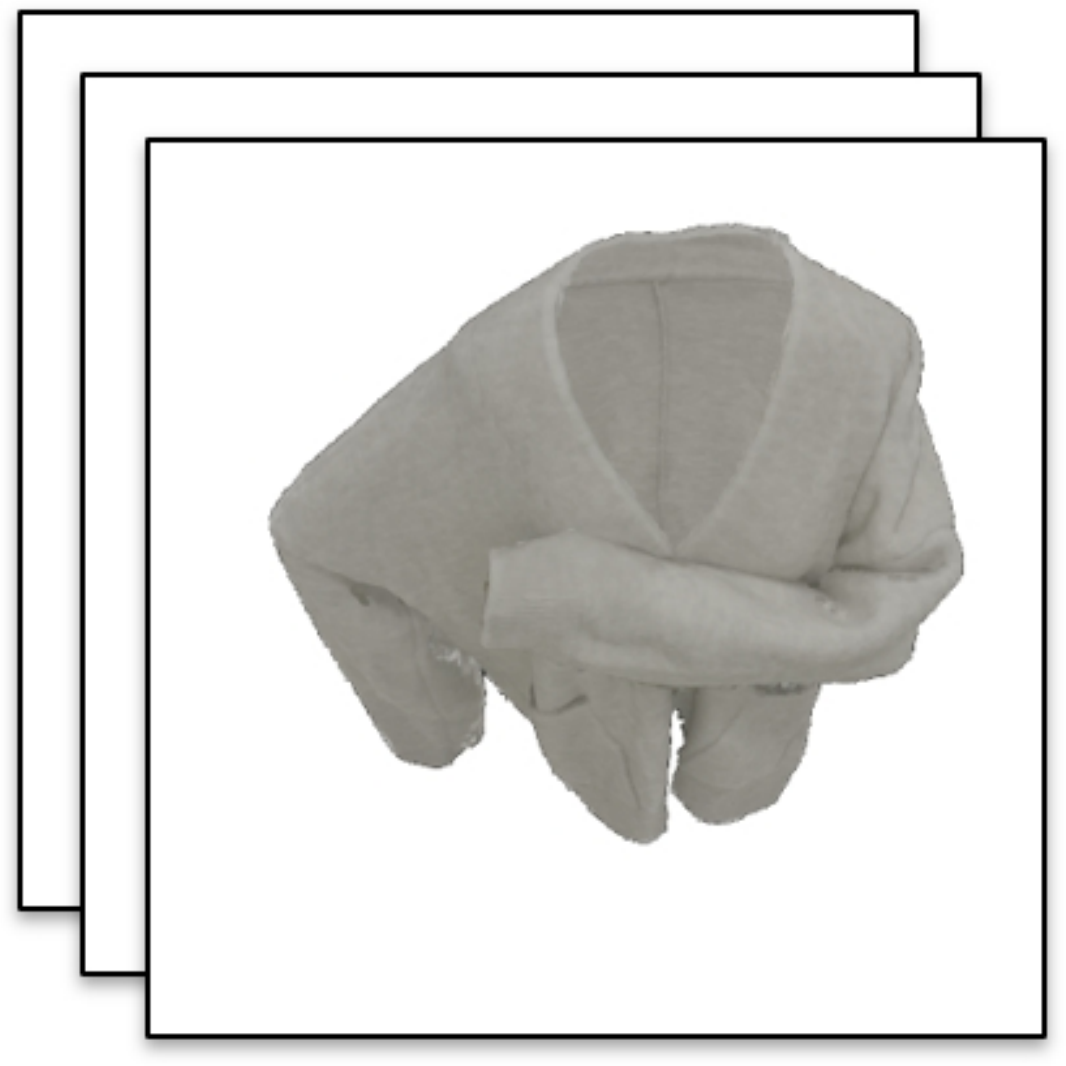}
\end{minipage}
\begin{minipage}[c]{.28\textwidth}
    \centering
    \includegraphics[width=.45\linewidth]{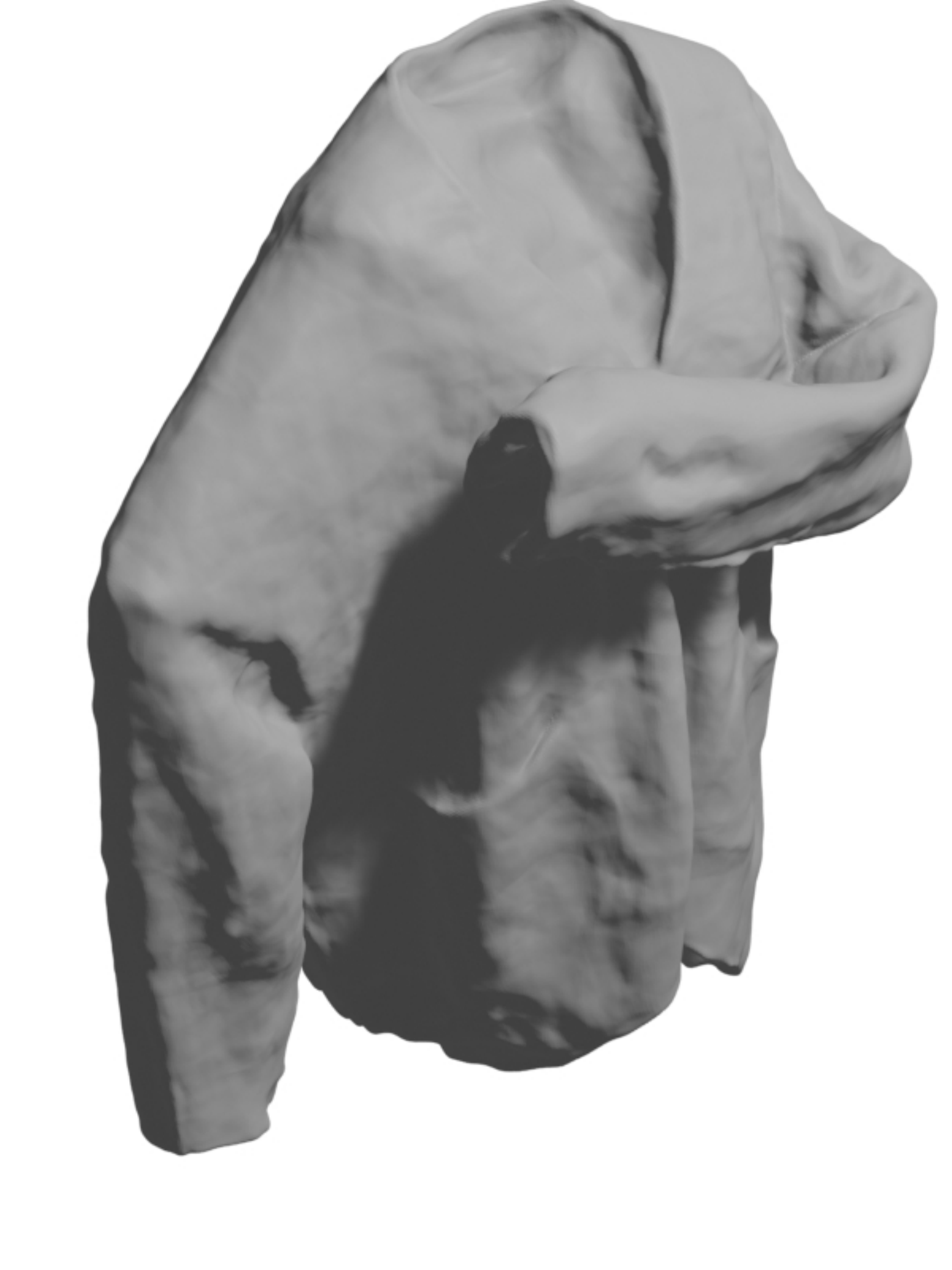}
    \includegraphics[width=.45\linewidth]{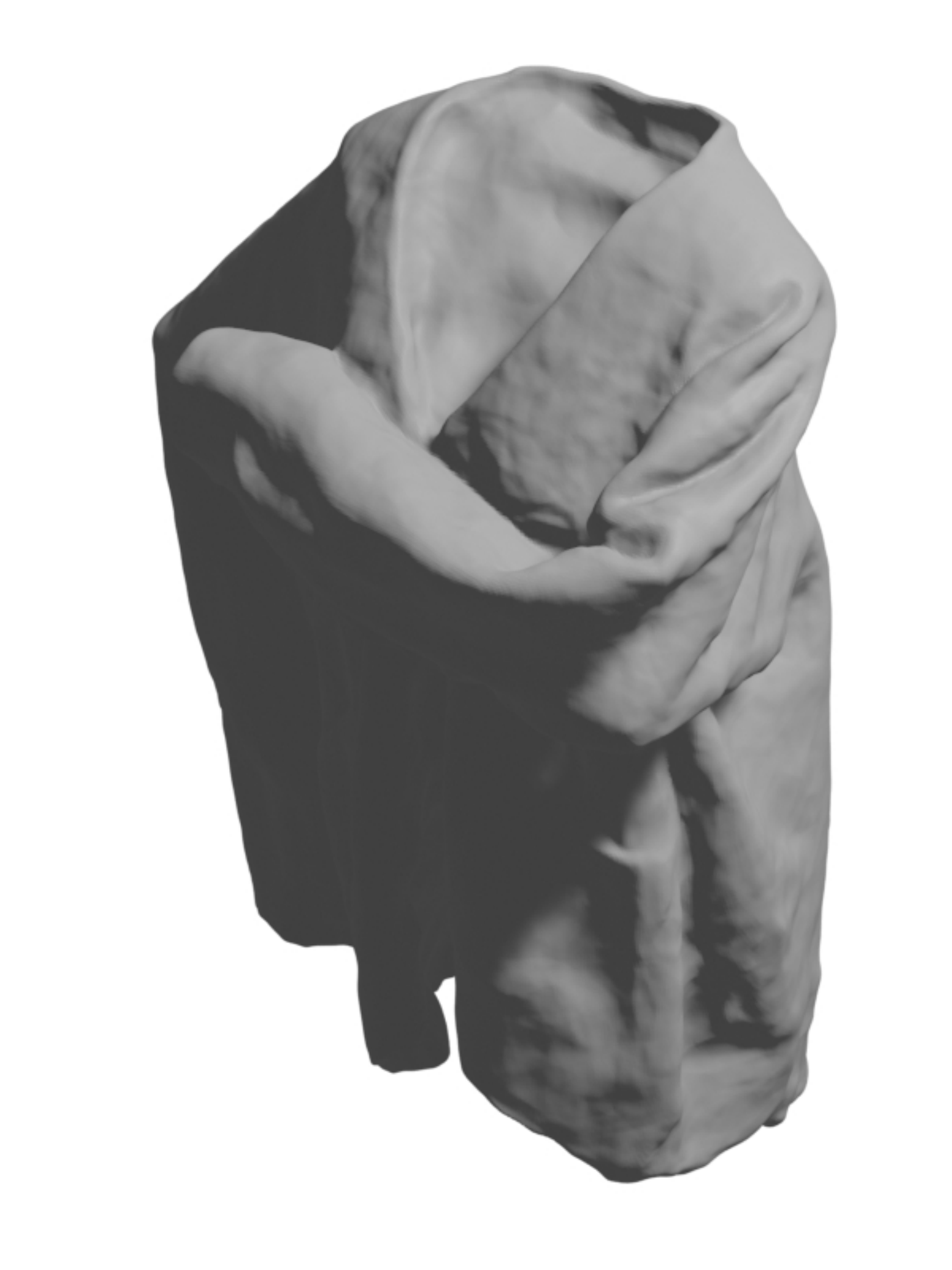}\\
    \includegraphics[width=.45\linewidth]{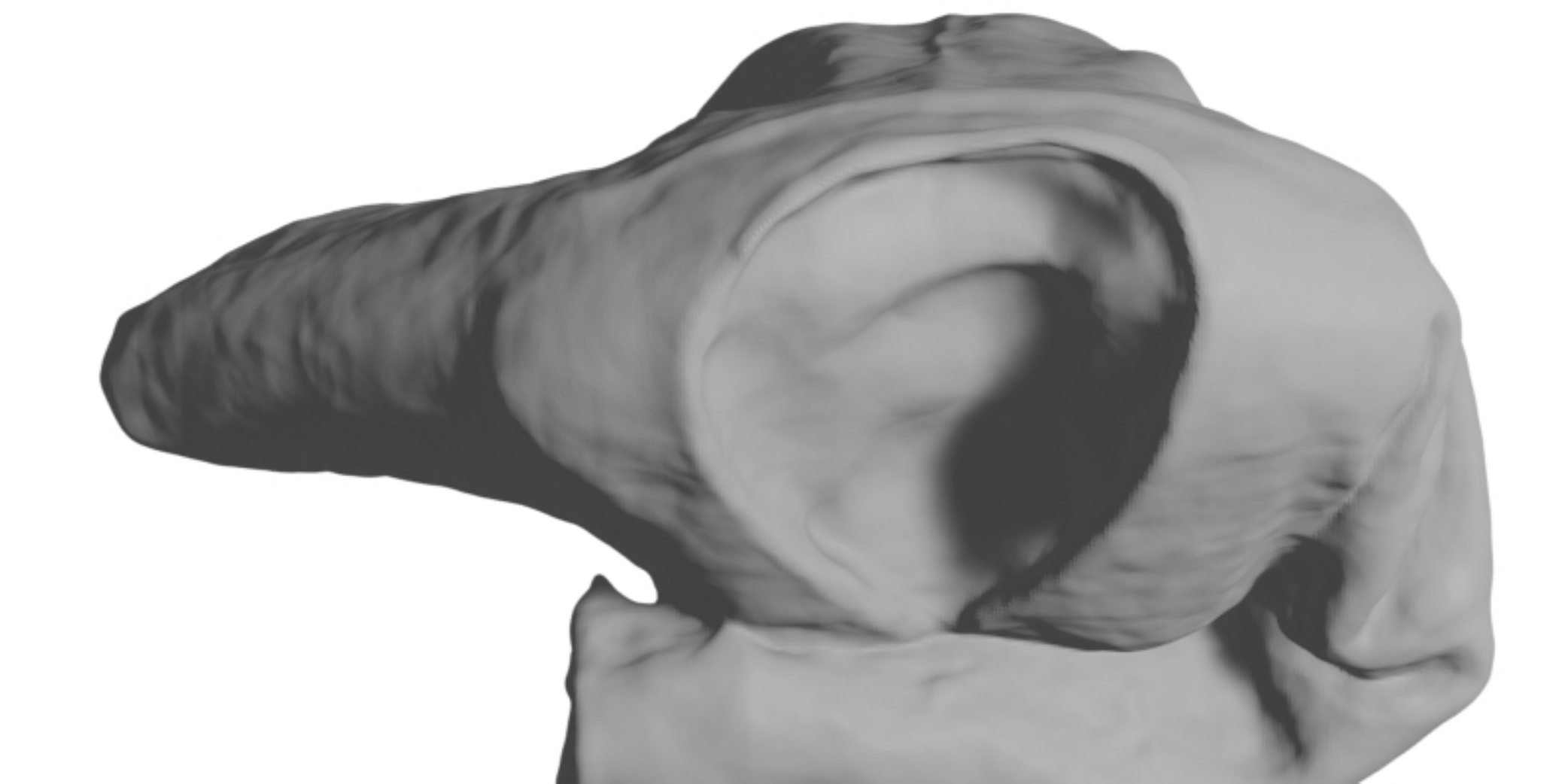}
\end{minipage}
\begin{minipage}[c]{.28\textwidth}
    \centering
    \includegraphics[width=.45\linewidth]{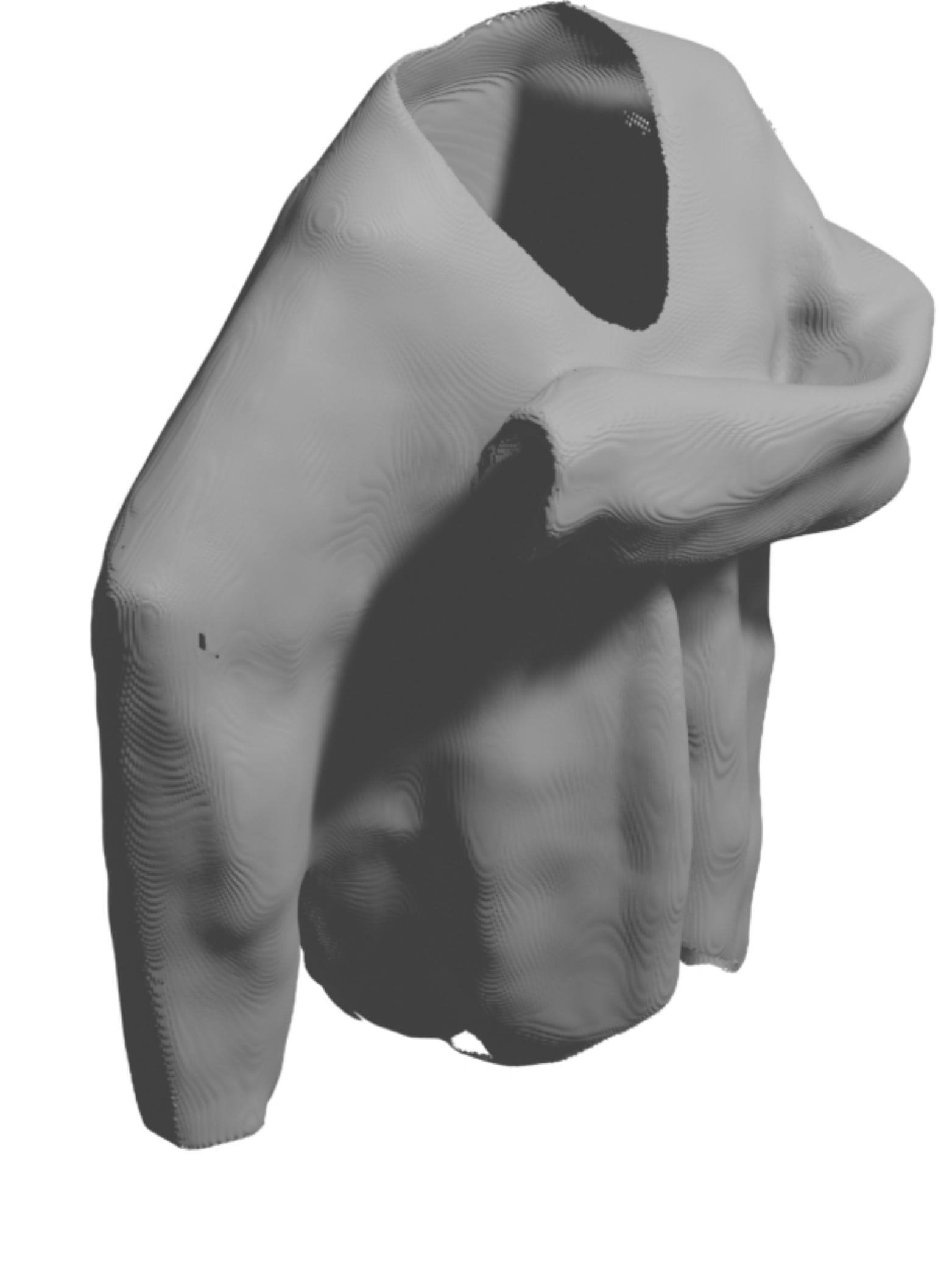}
    \includegraphics[width=.45\linewidth]{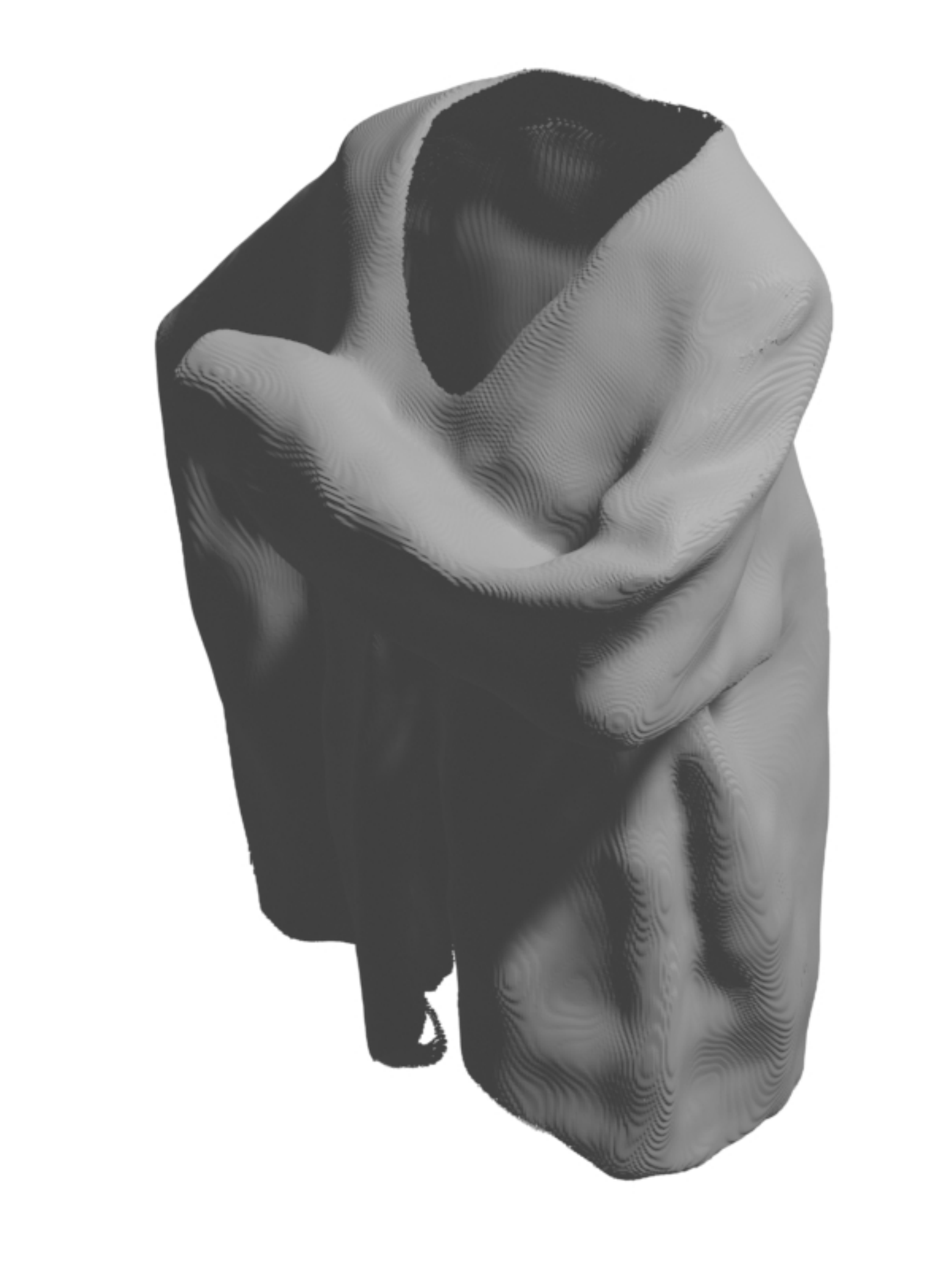}\\
    \includegraphics[width=.45\linewidth]{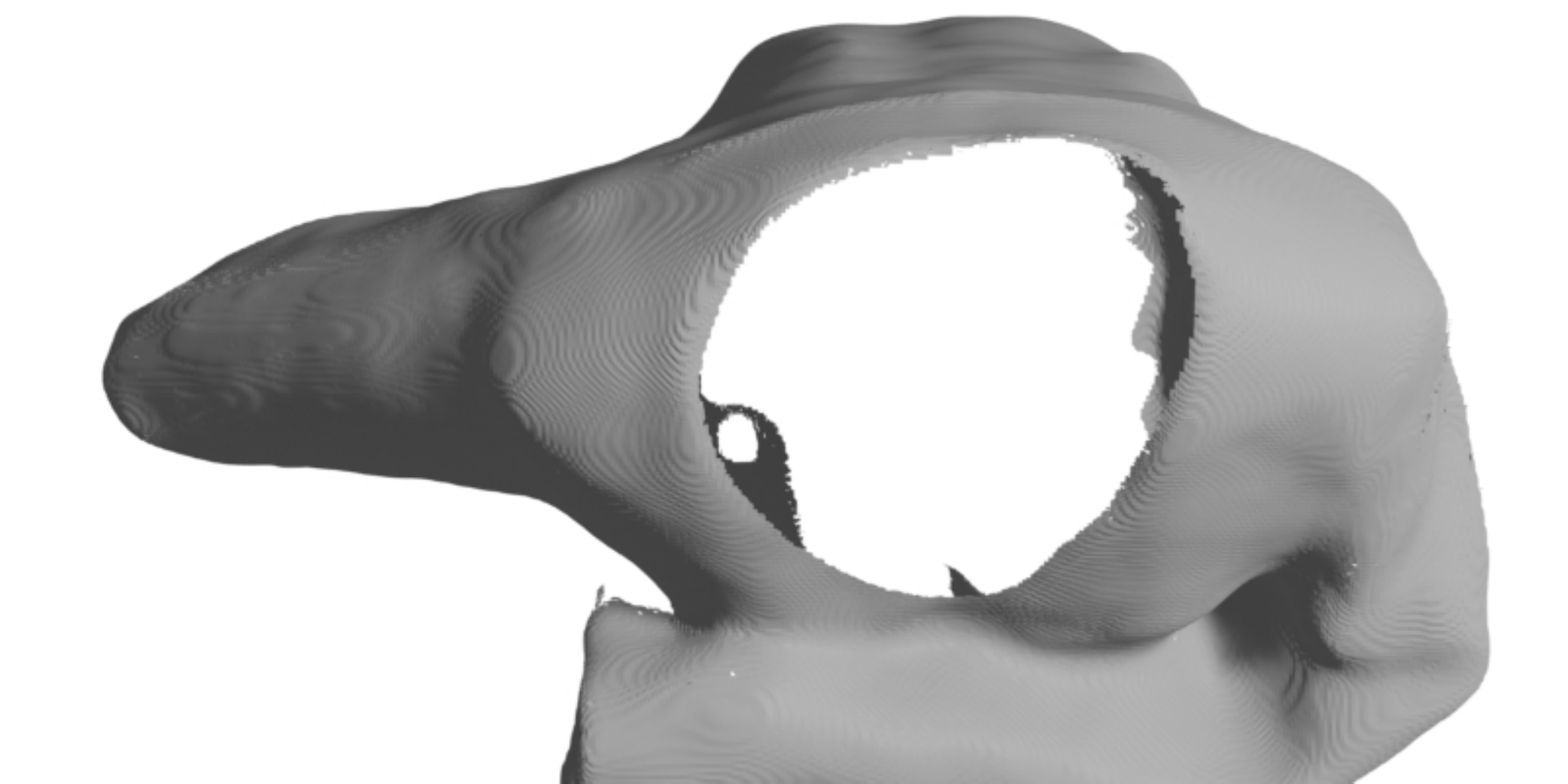}
\end{minipage}
\begin{minipage}[c]{.28\textwidth}
    \centering
    \includegraphics[width=.45\linewidth]{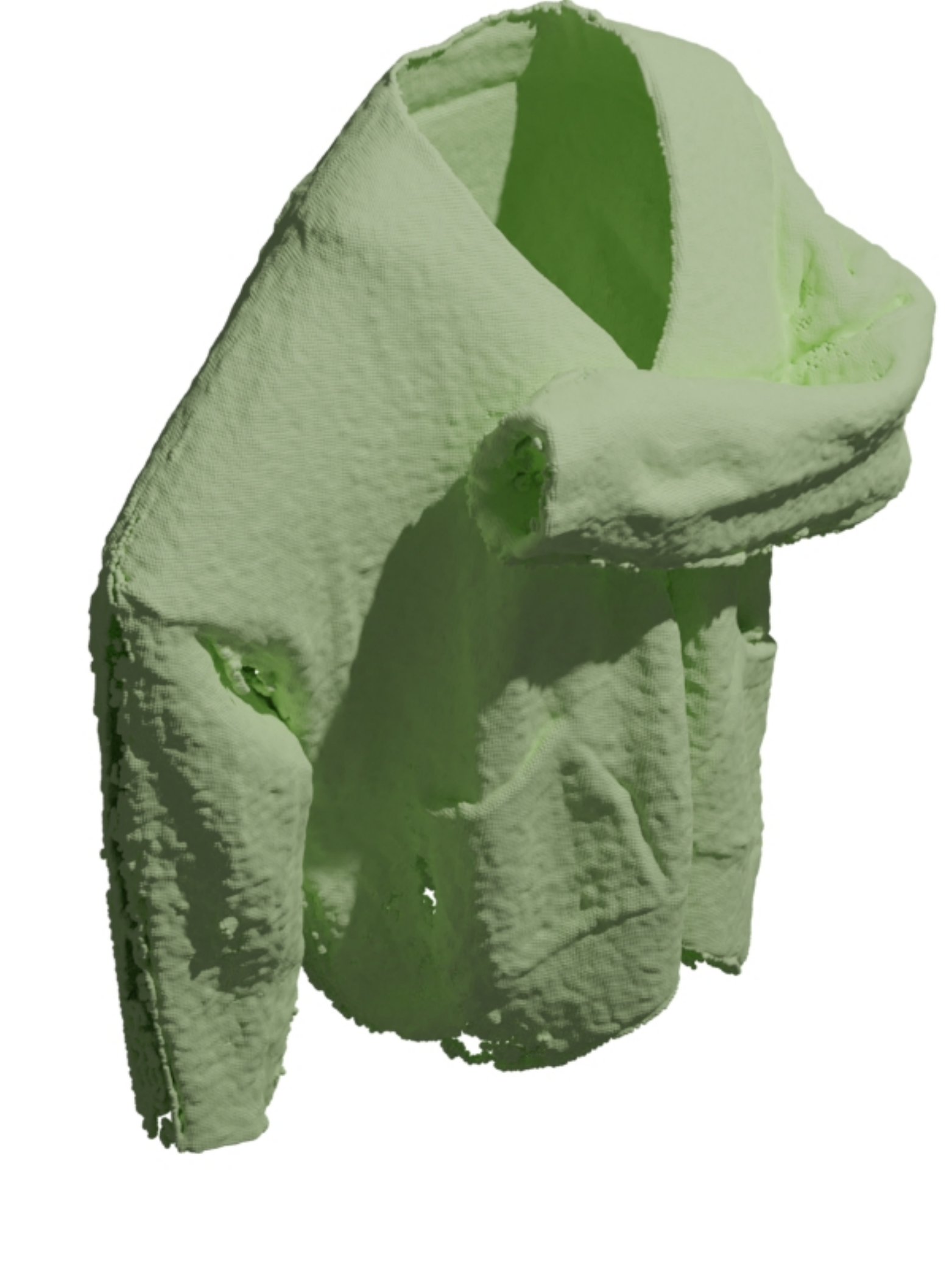}
    \includegraphics[width=.45\linewidth]{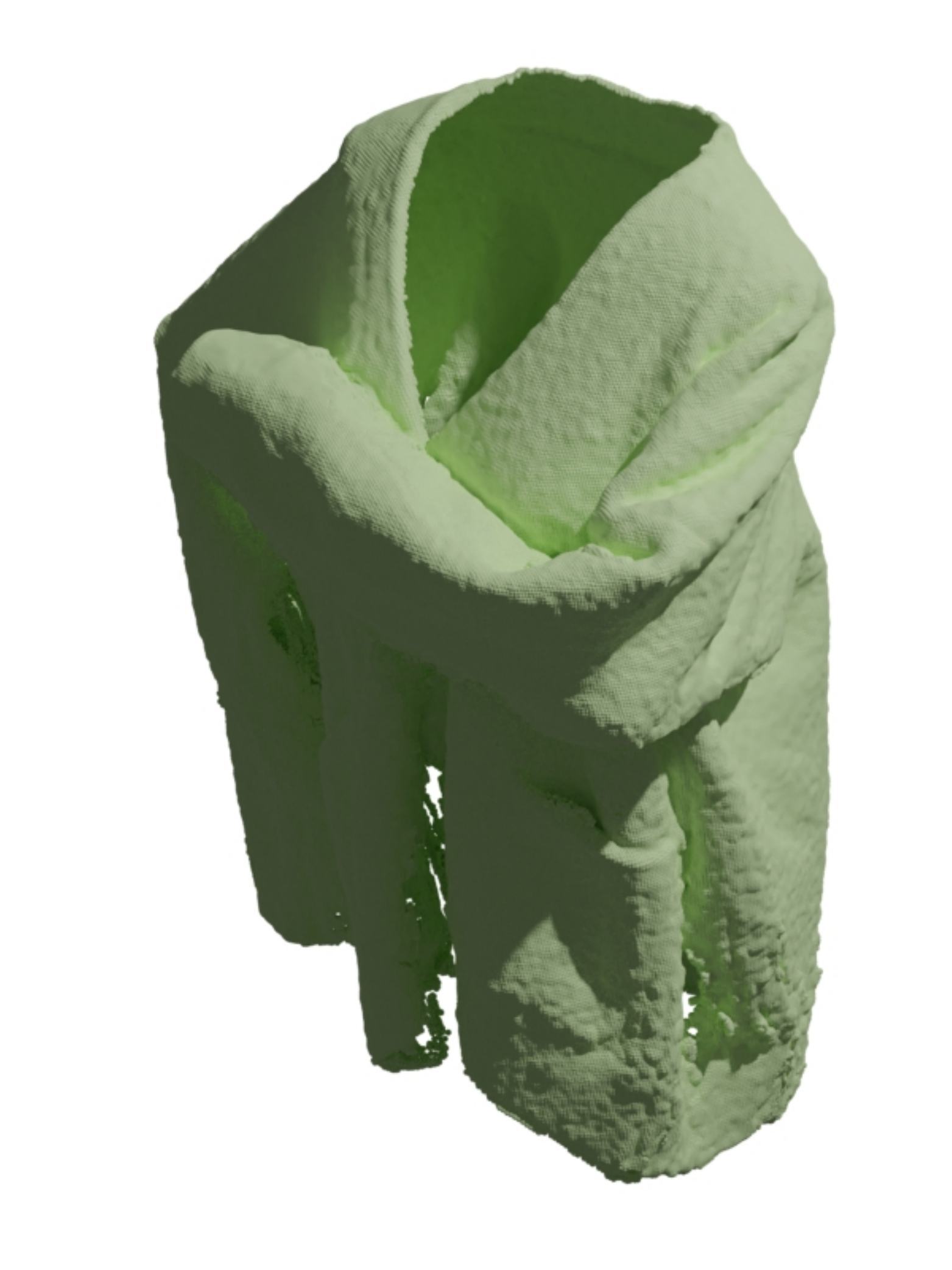}\\
    \includegraphics[width=.45\linewidth]{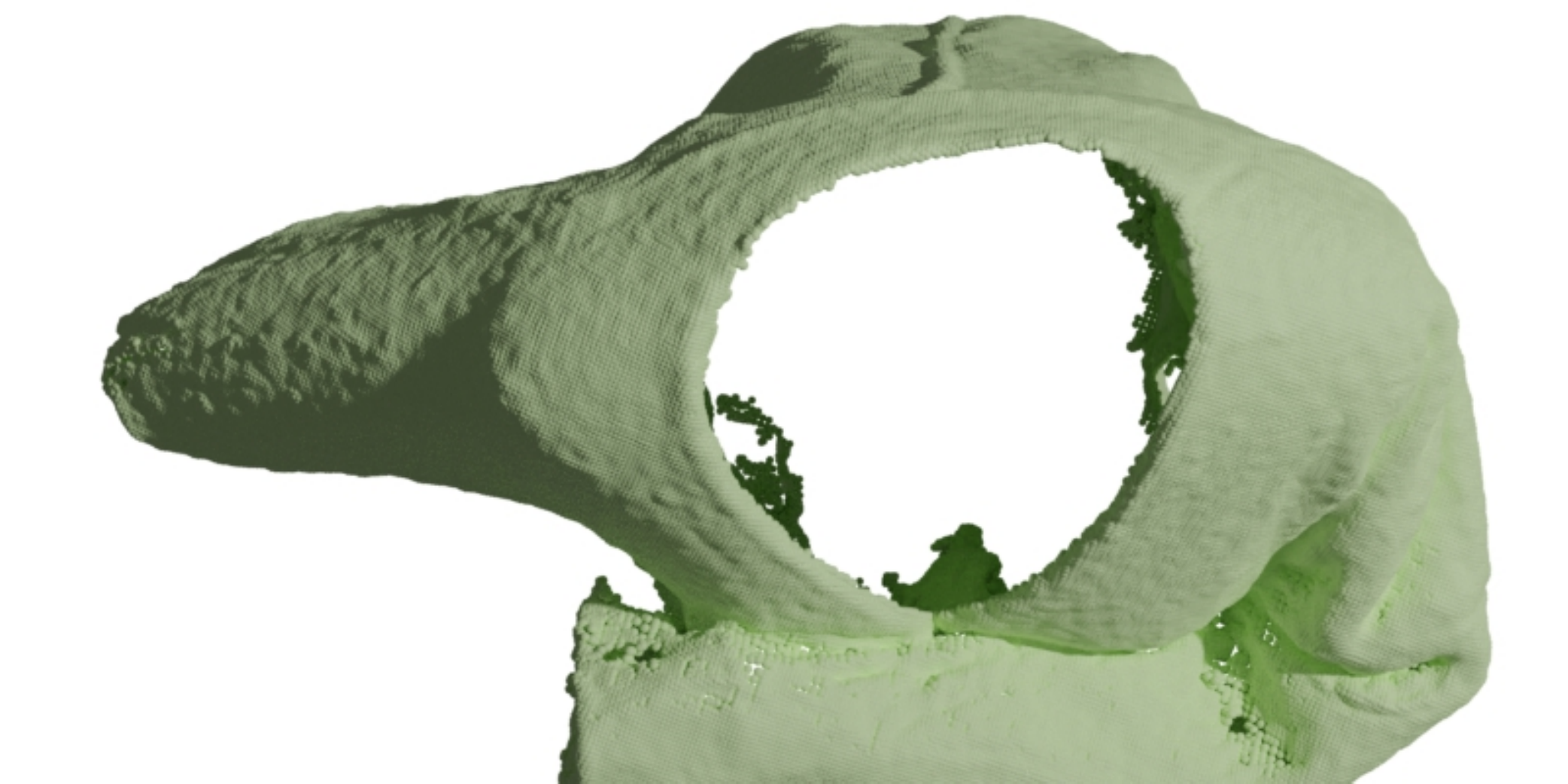}
\end{minipage}

\begin{minipage}[c]{.13\textwidth}
    \centering
    \includegraphics[width=1\linewidth]{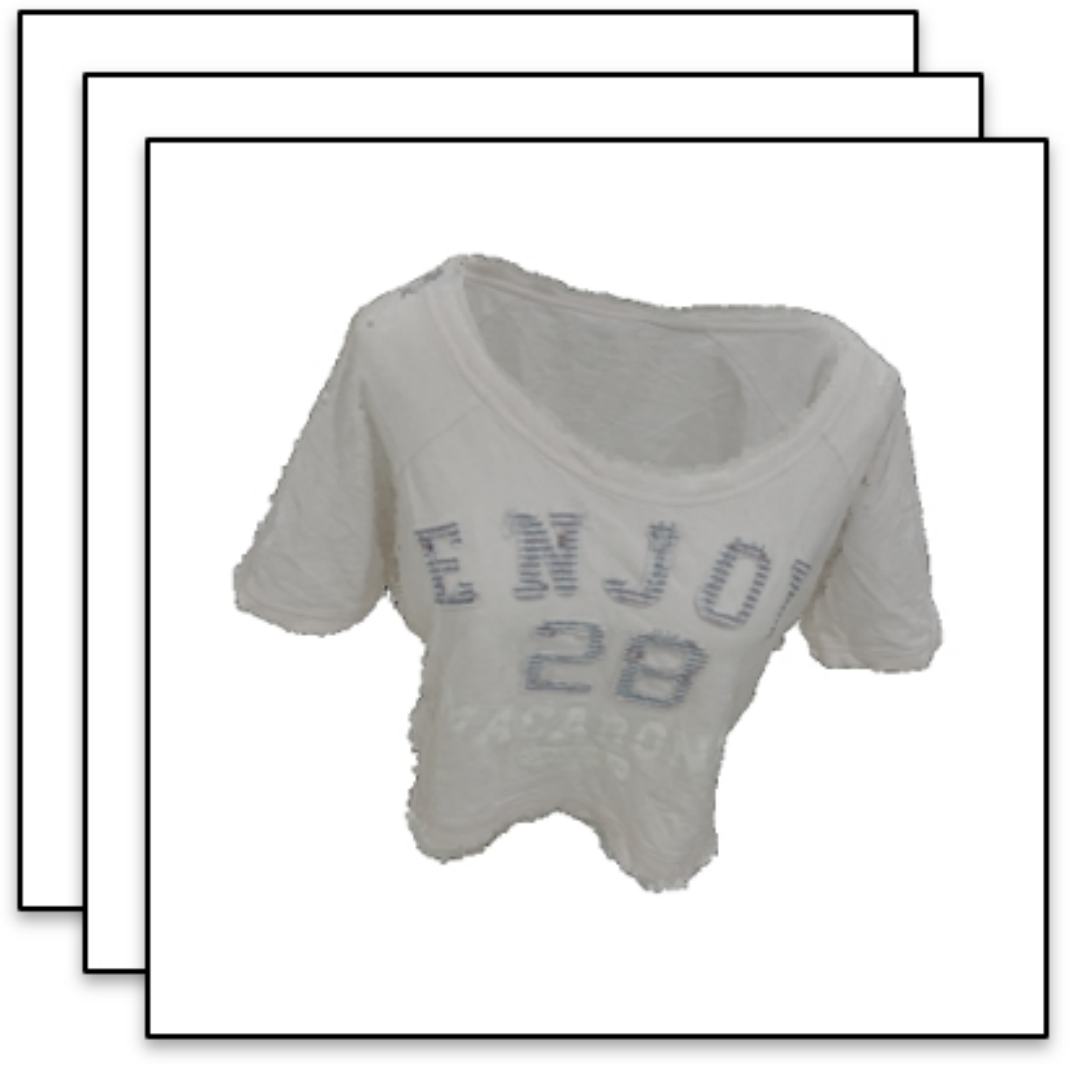}
\end{minipage}
\begin{minipage}[c]{.28\textwidth}
    \centering
    \includegraphics[width=.45\linewidth]{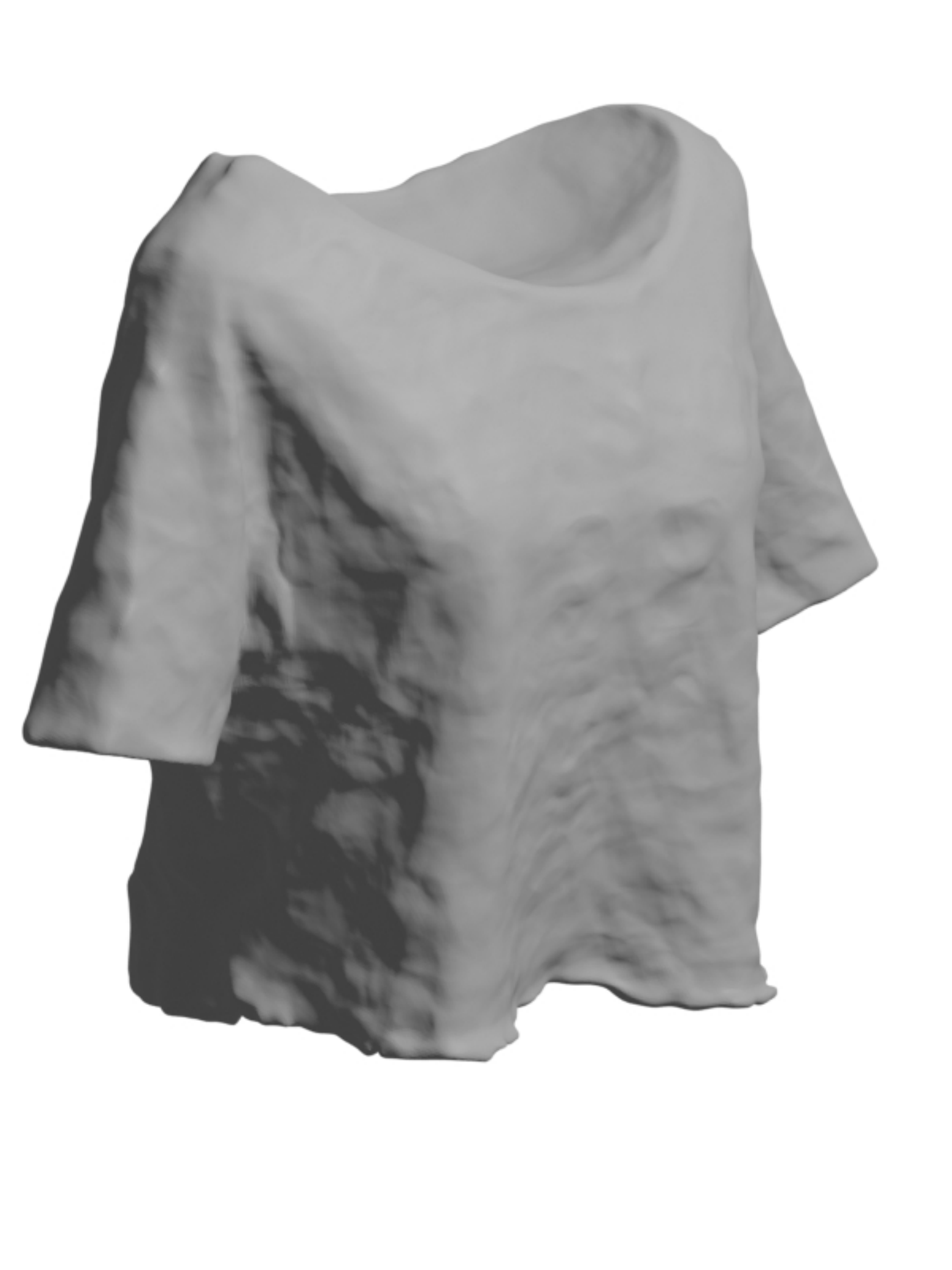}
    \includegraphics[width=.45\linewidth]{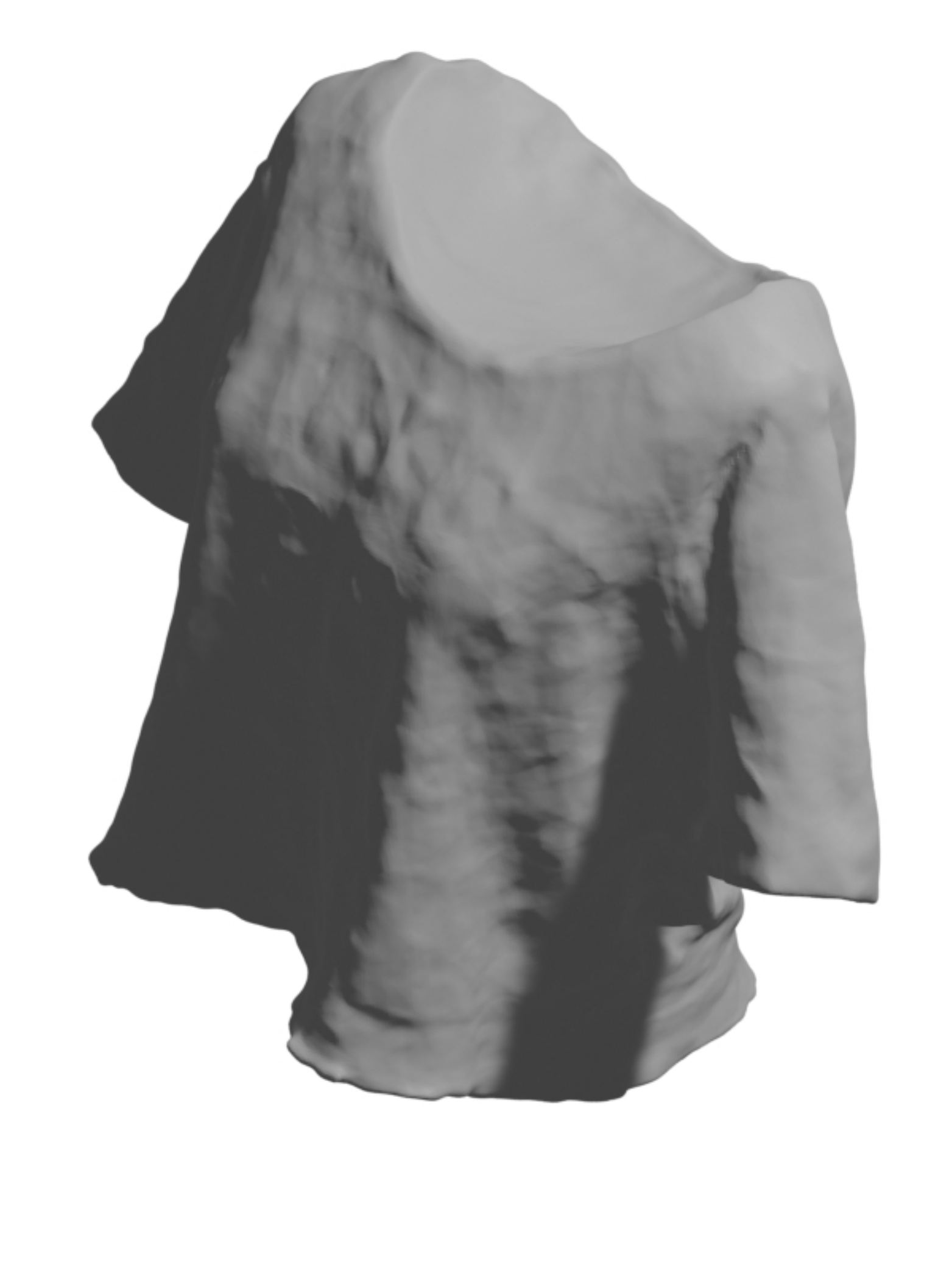}\\
    \includegraphics[width=.45\linewidth]{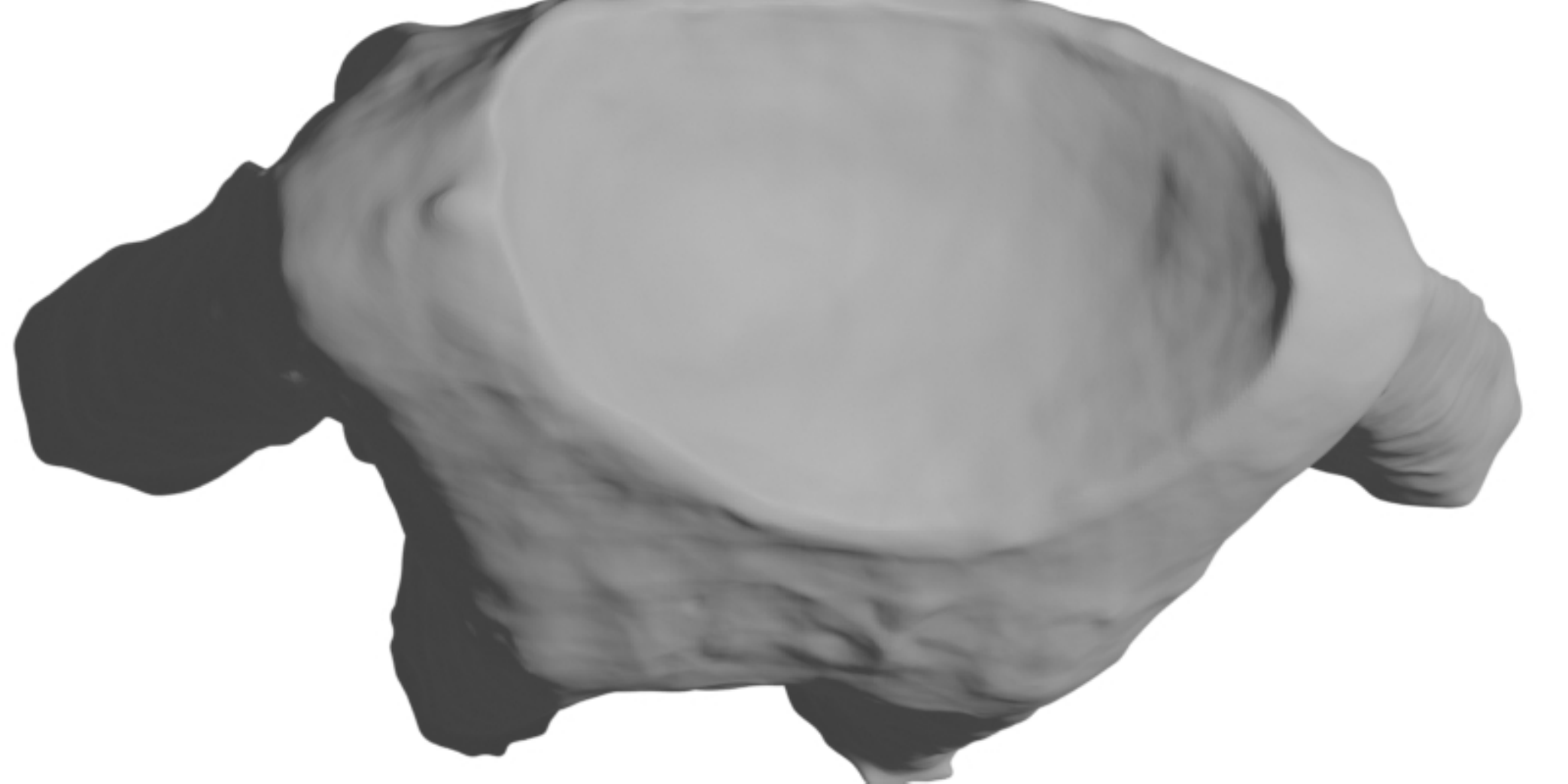}
\end{minipage}
\begin{minipage}[c]{.28\textwidth}
    \centering
    \includegraphics[width=.45\linewidth]{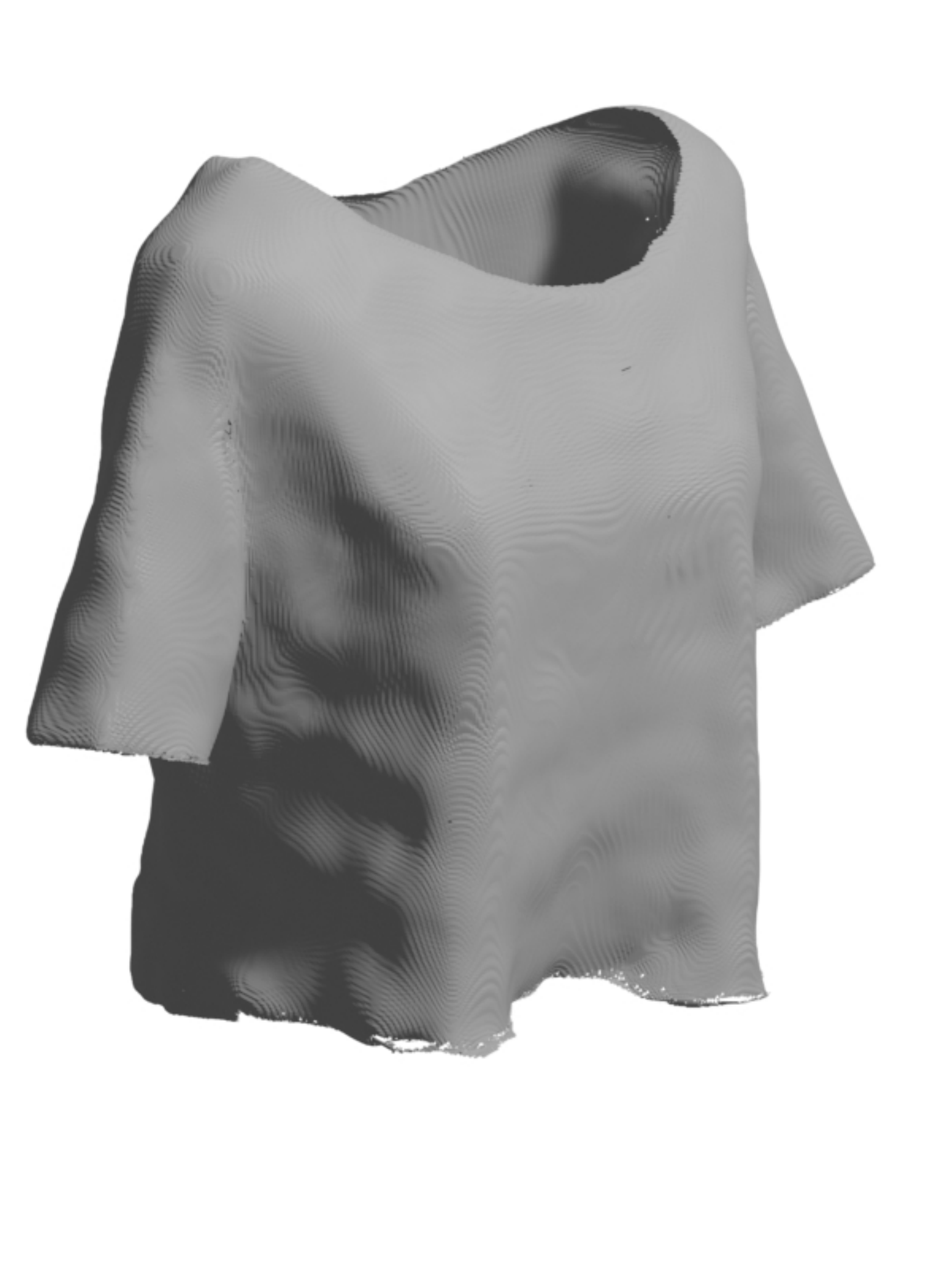}
    \includegraphics[width=.45\linewidth]{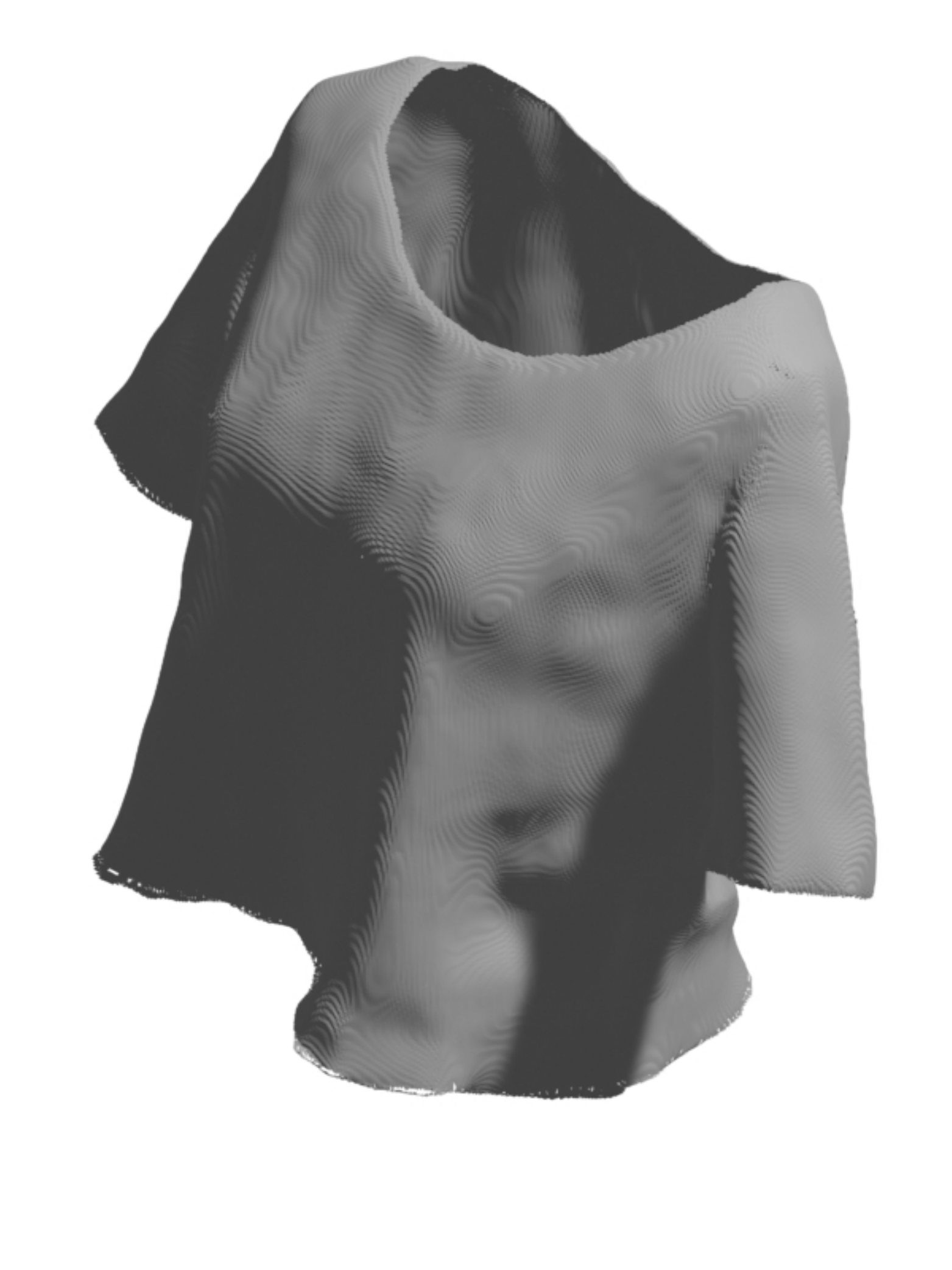}\\
    \includegraphics[width=.45\linewidth]{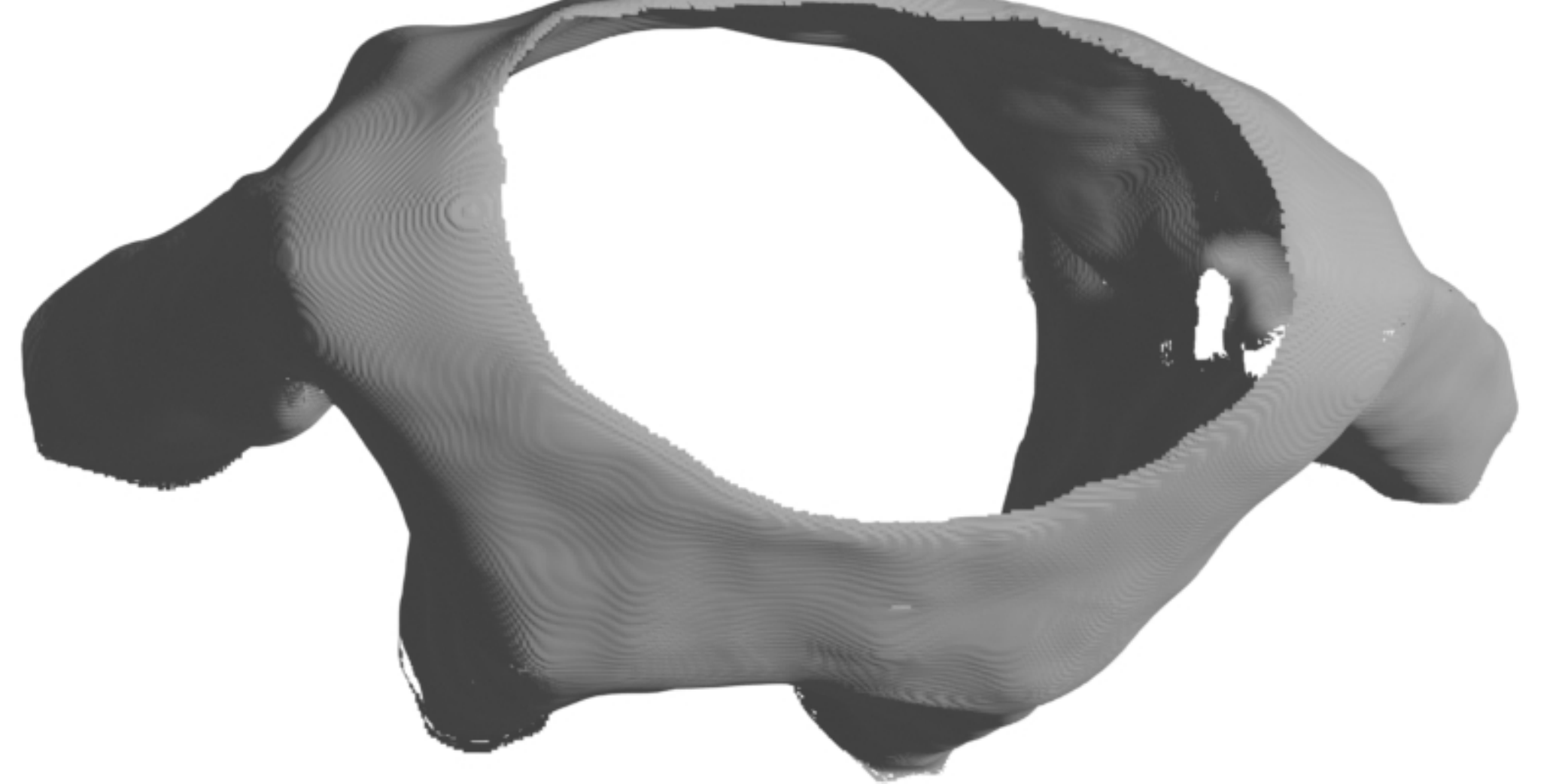}
\end{minipage}
\begin{minipage}[c]{.28\textwidth}
    \centering
    \includegraphics[width=.45\linewidth]{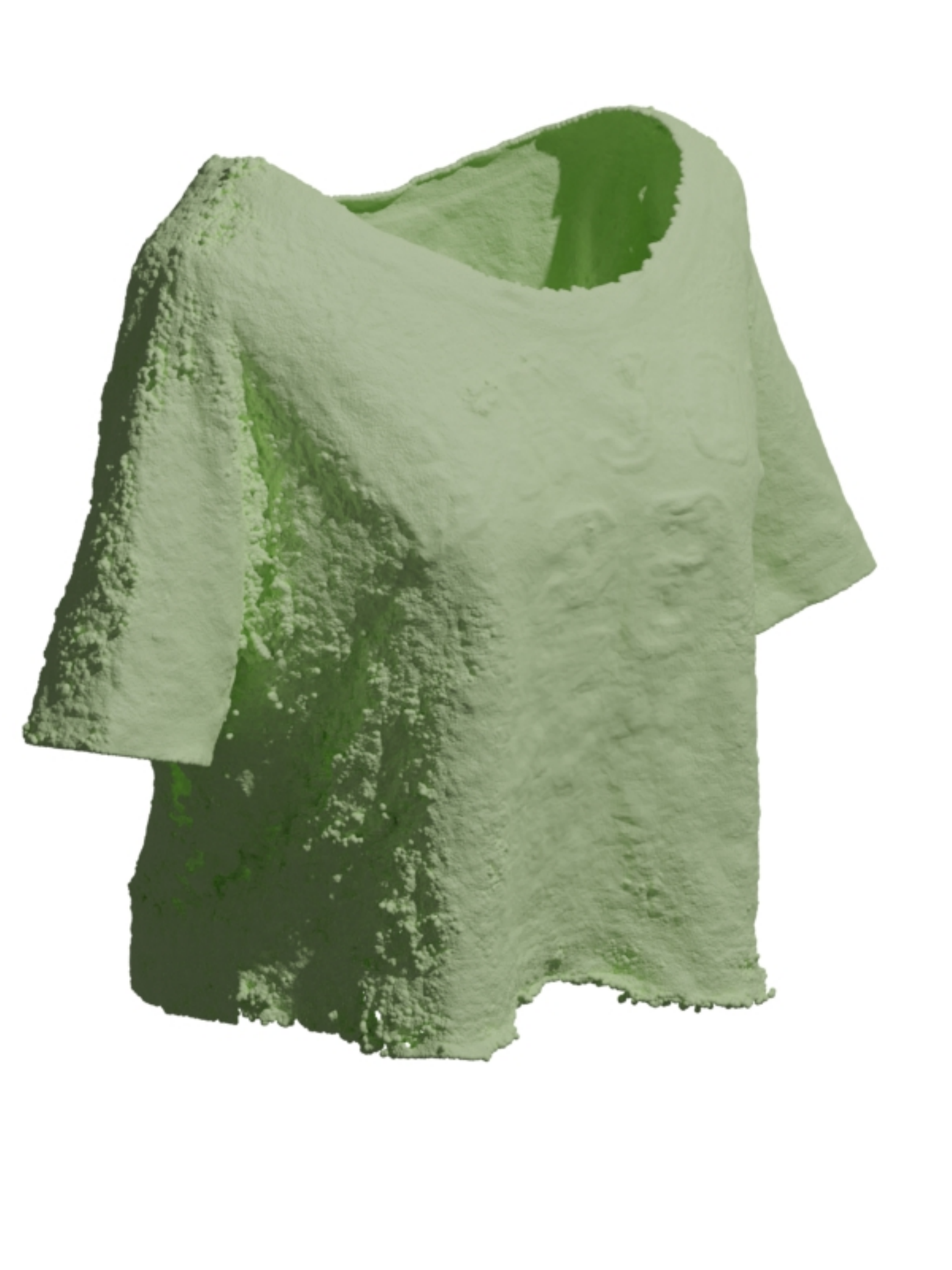}
    \includegraphics[width=.45\linewidth]{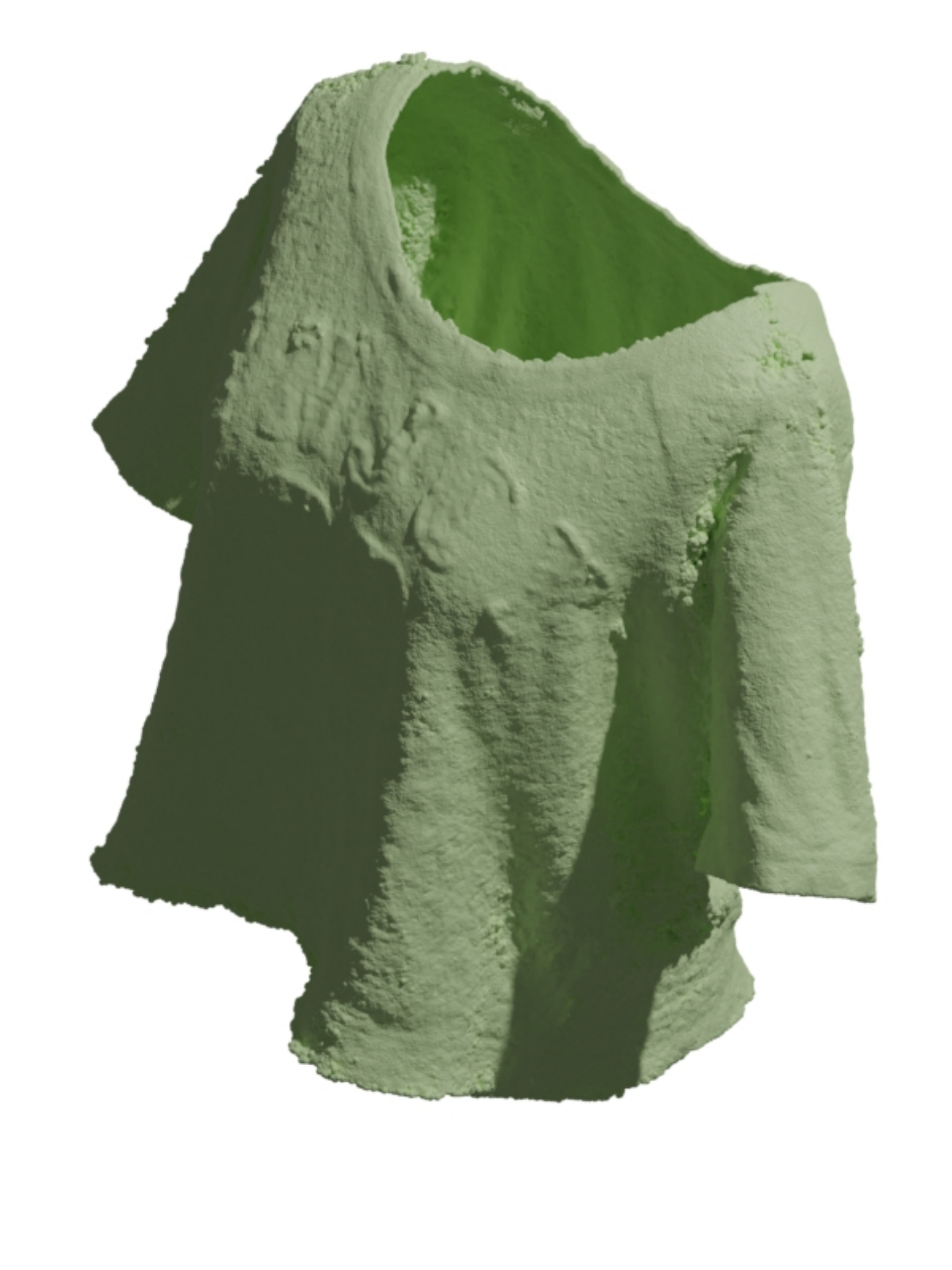}\\
    \includegraphics[width=.45\linewidth]{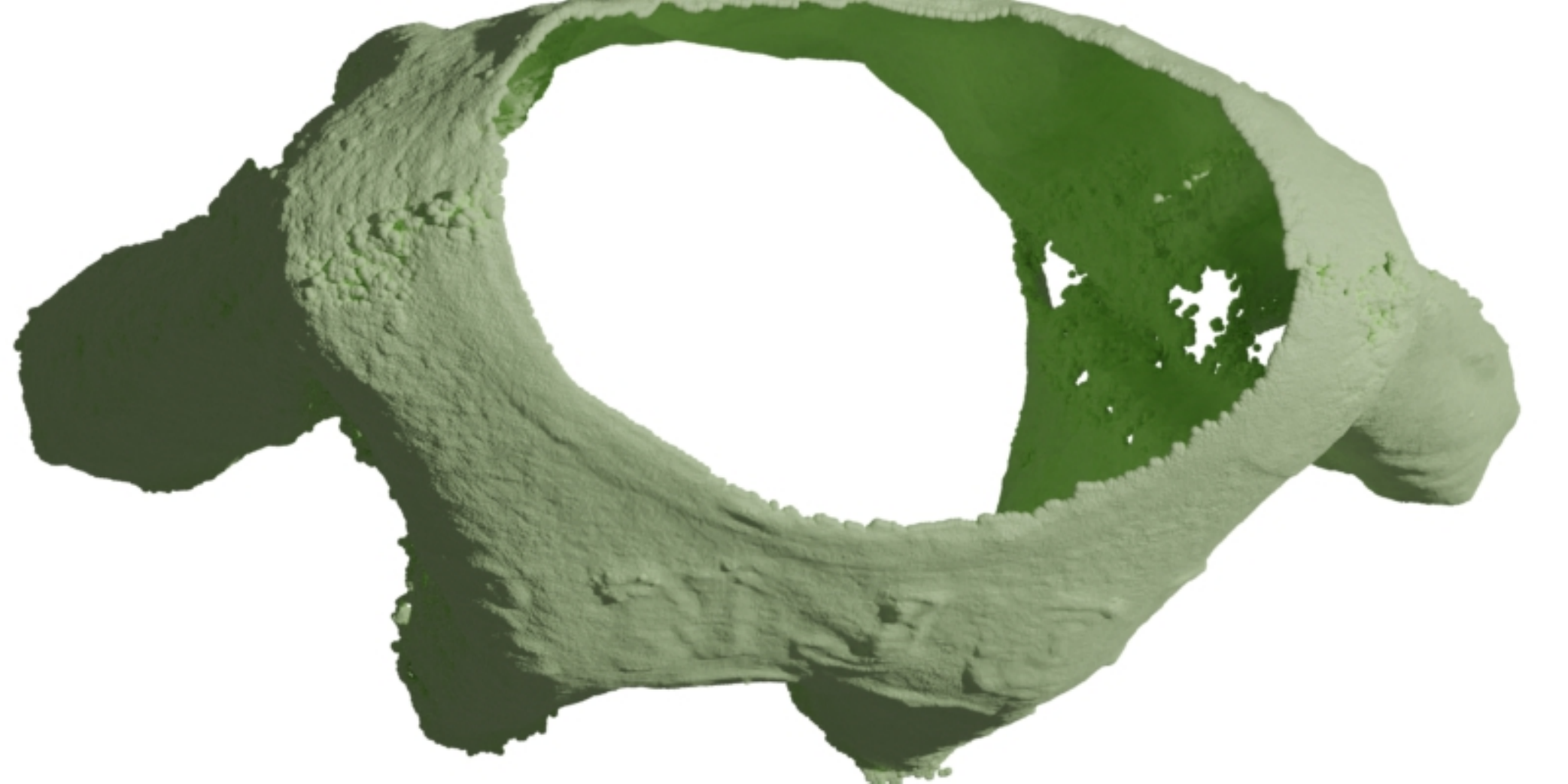}
\end{minipage}

\begin{minipage}[c]{.13\textwidth}
    \centering
    \includegraphics[width=1\linewidth]{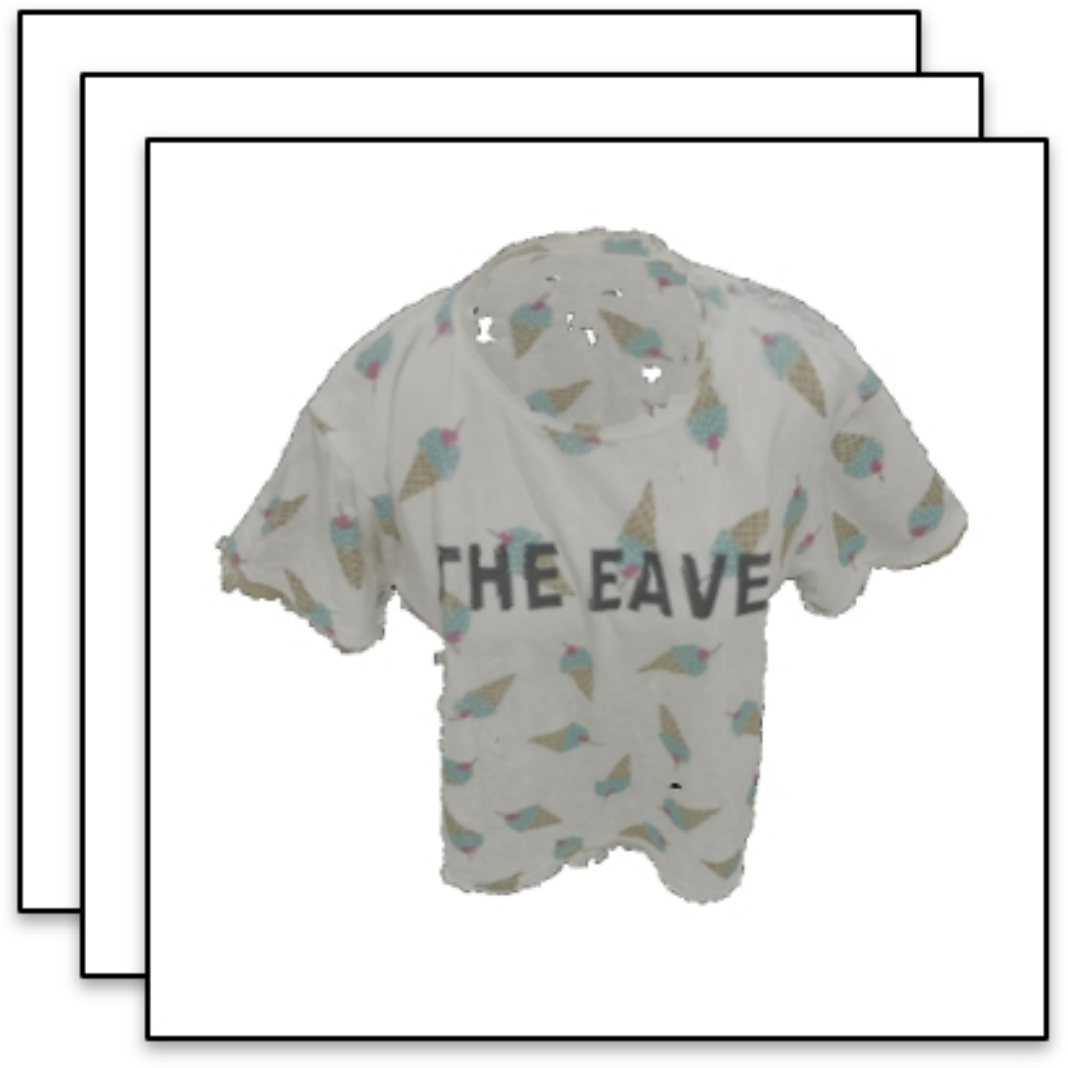}
\end{minipage}
\begin{minipage}[c]{.28\textwidth}
    \centering
    \includegraphics[width=.45\linewidth]{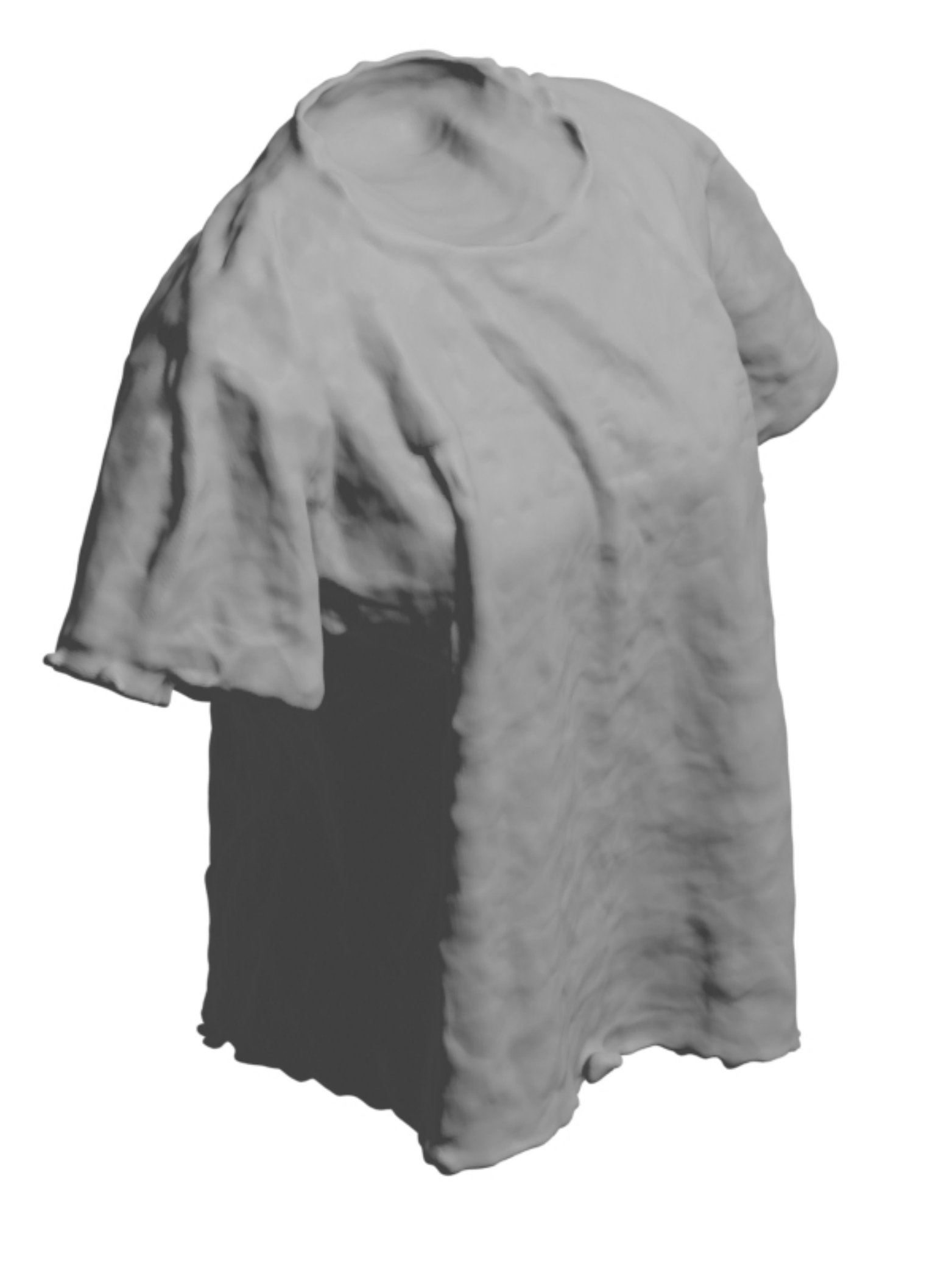}
    \includegraphics[width=.45\linewidth]{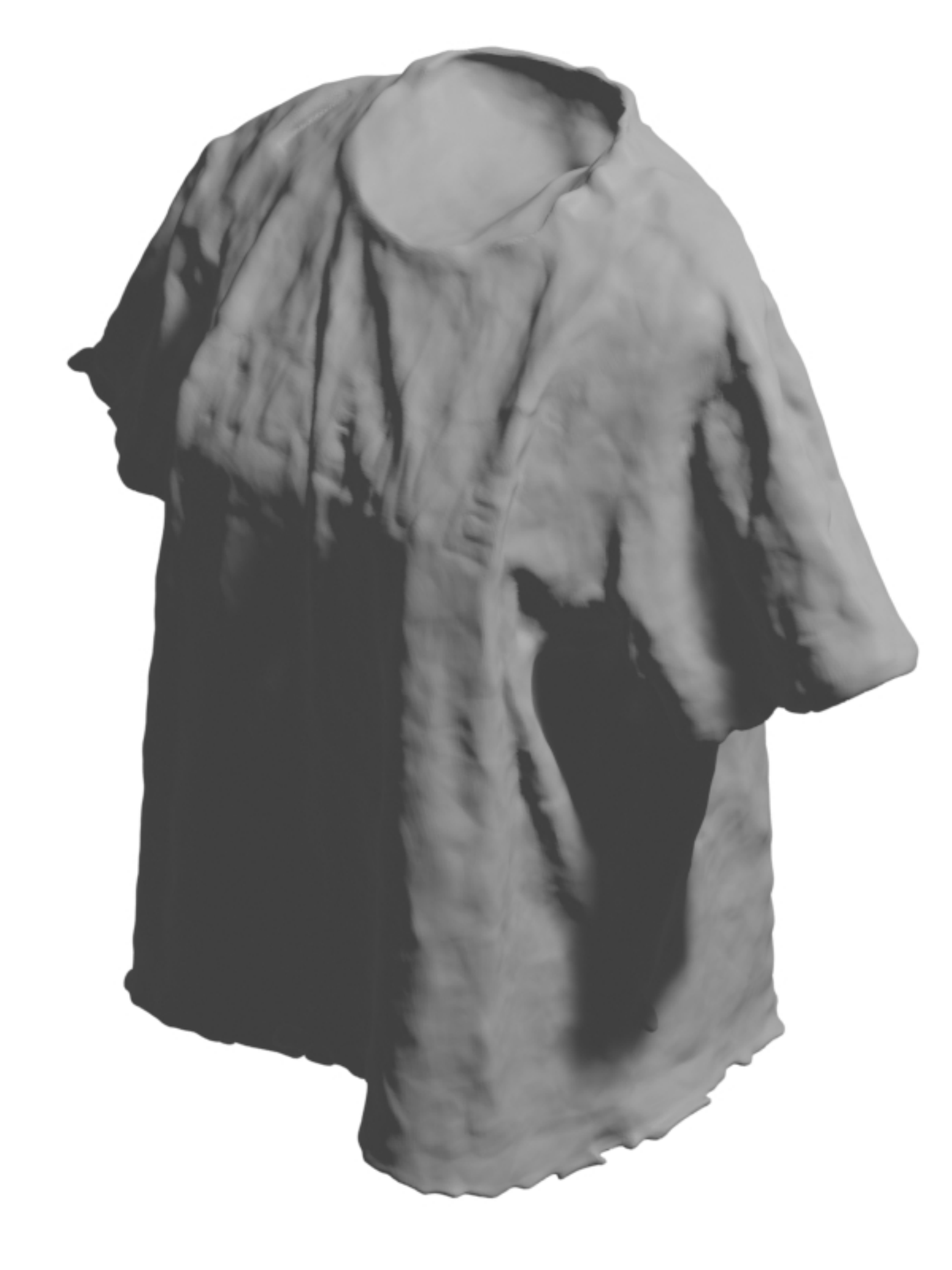}\\
    \includegraphics[width=.45\linewidth]{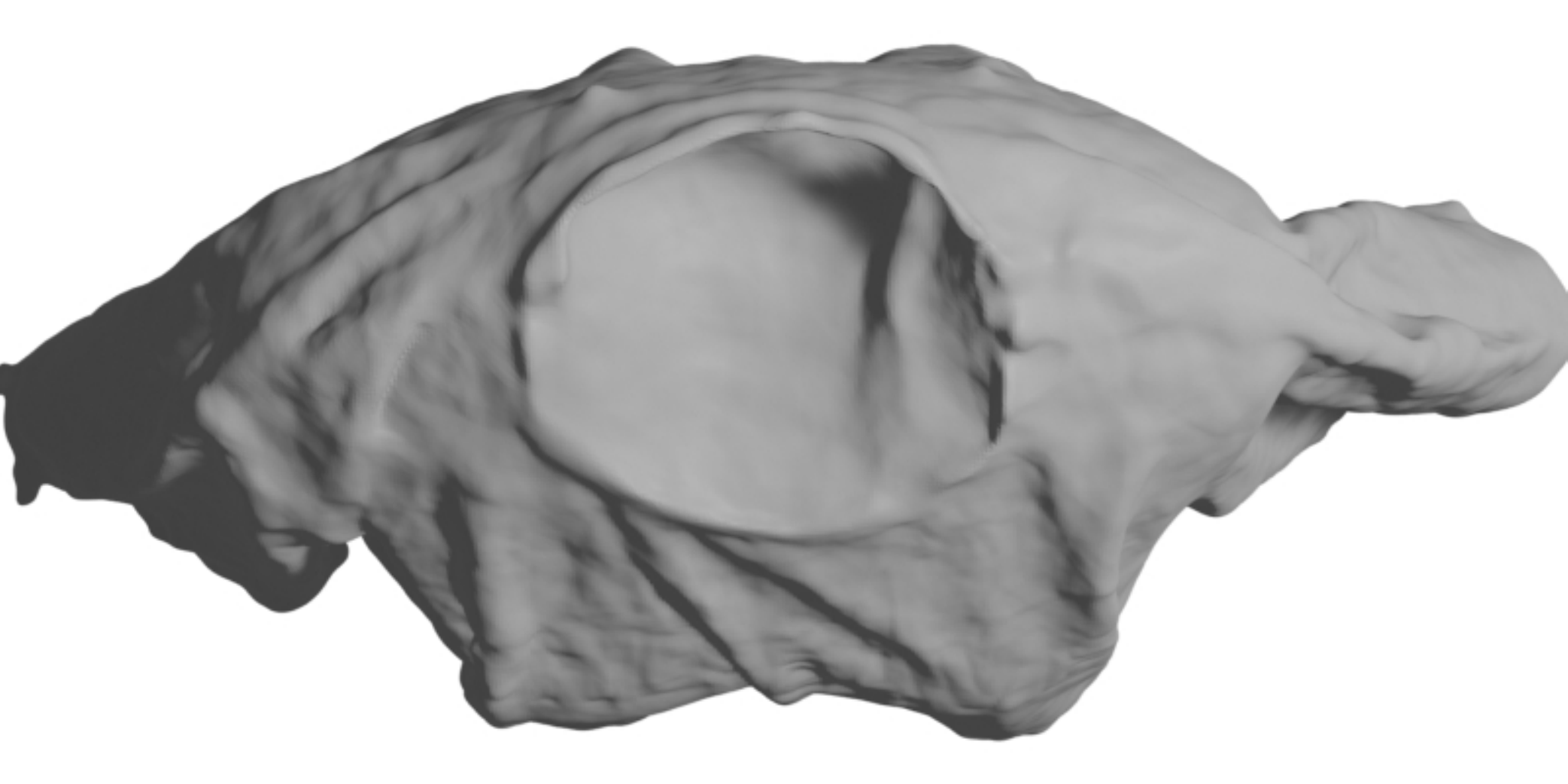}
\end{minipage}
\begin{minipage}[c]{.28\textwidth}
    \centering
    \includegraphics[width=.45\linewidth]{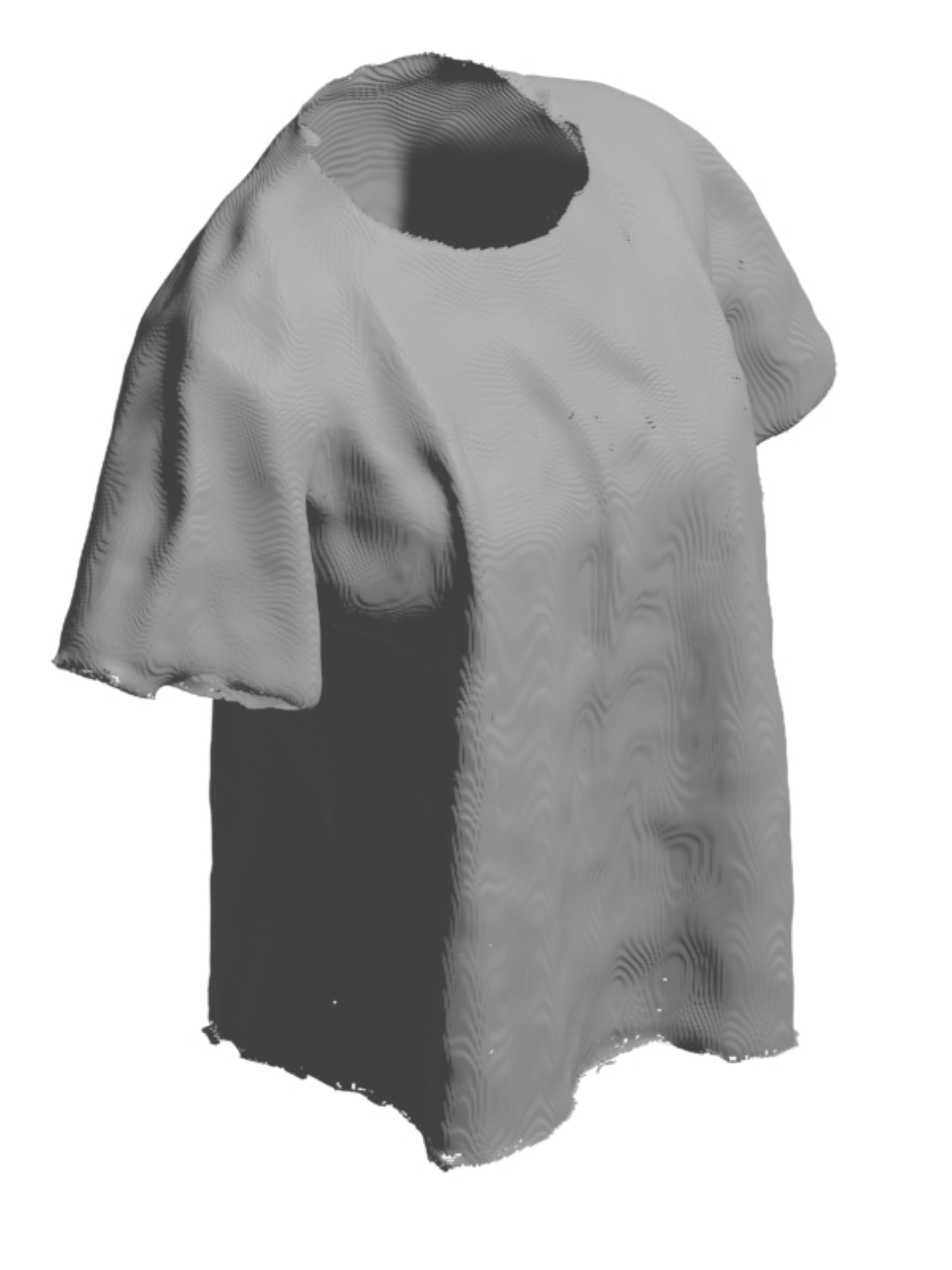}
    \includegraphics[width=.45\linewidth]{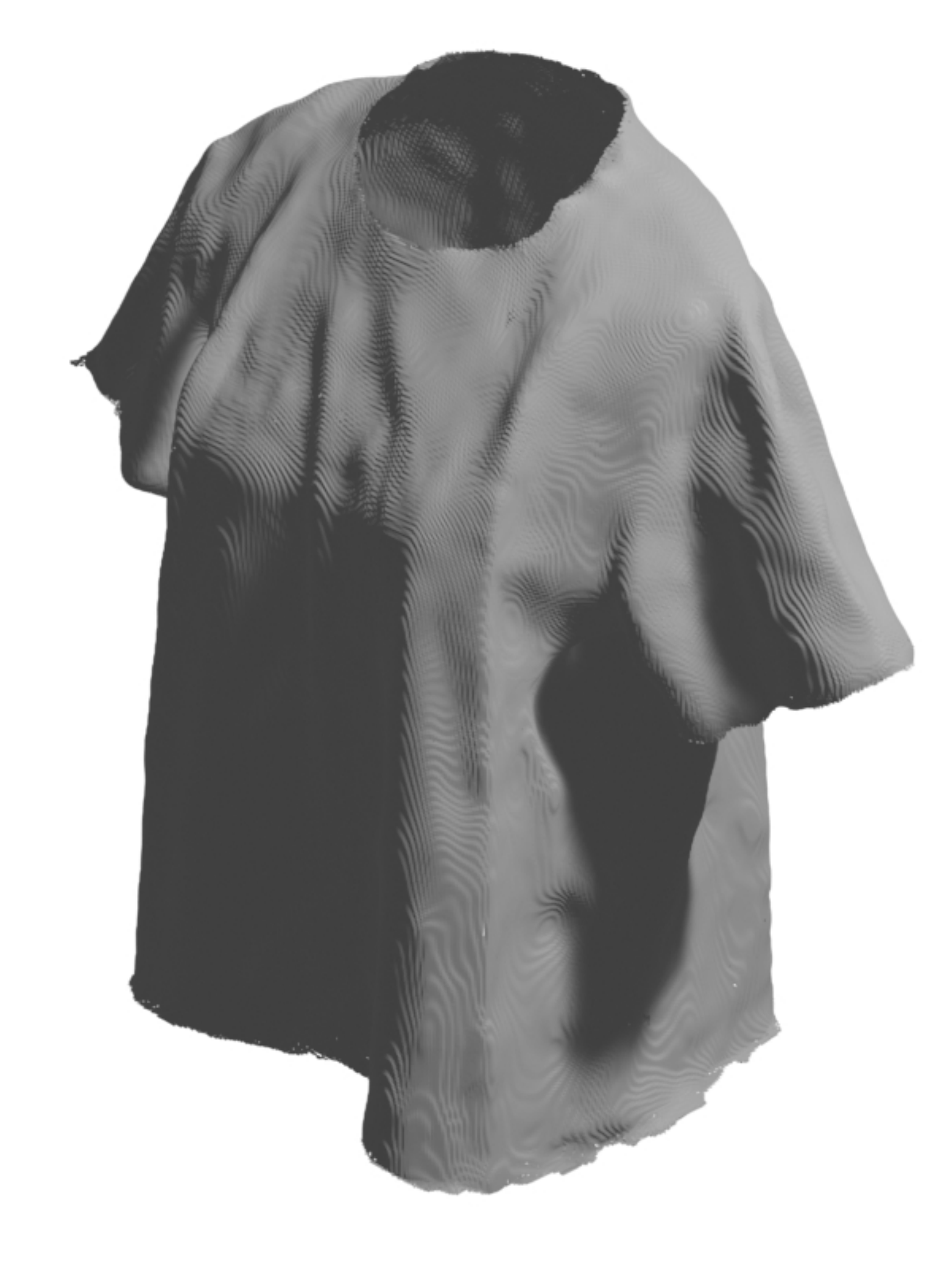}\\
    \includegraphics[width=.45\linewidth]{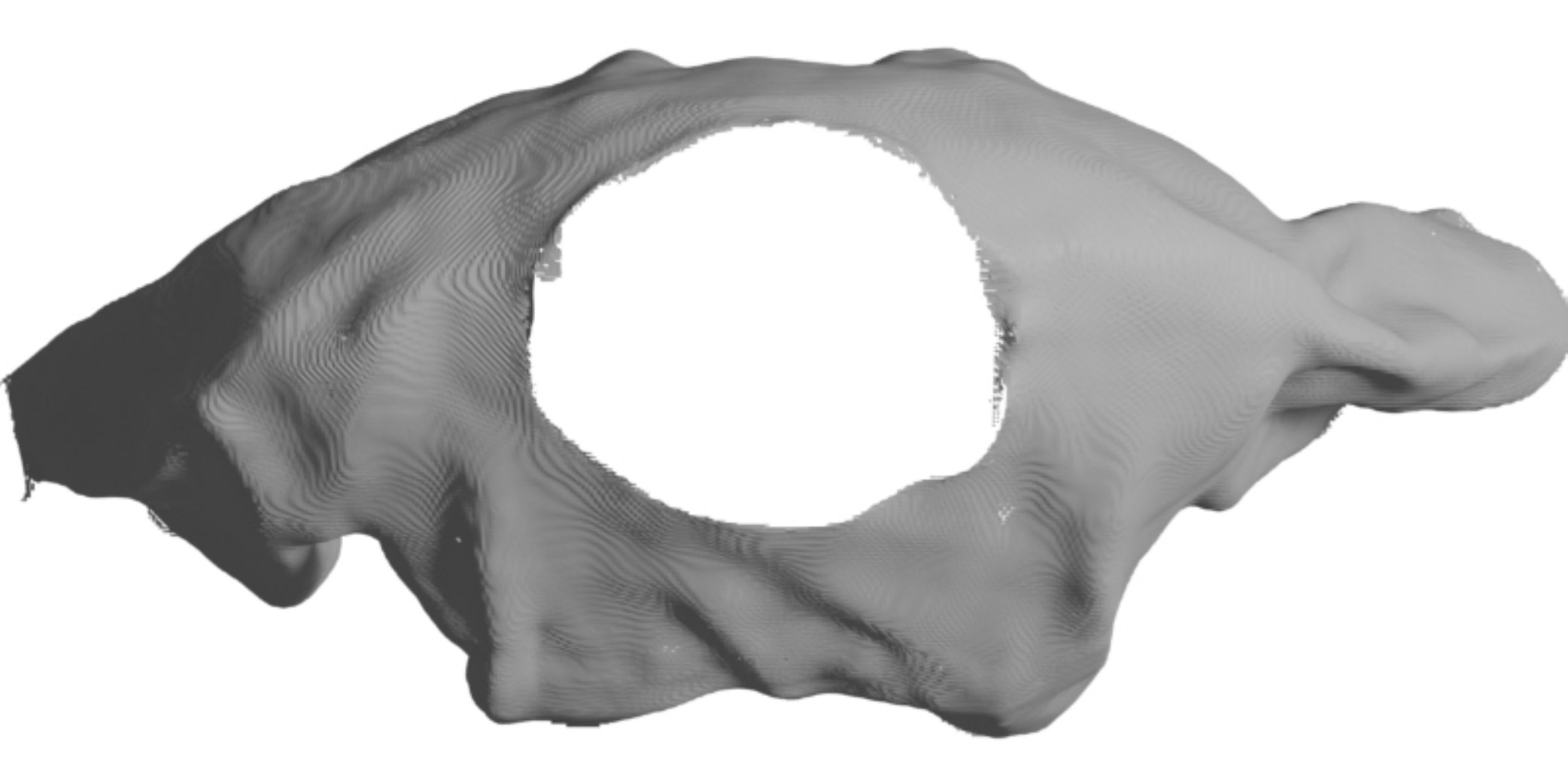}
\end{minipage}
\begin{minipage}[c]{.28\textwidth}
    \centering
    \includegraphics[width=.45\linewidth]{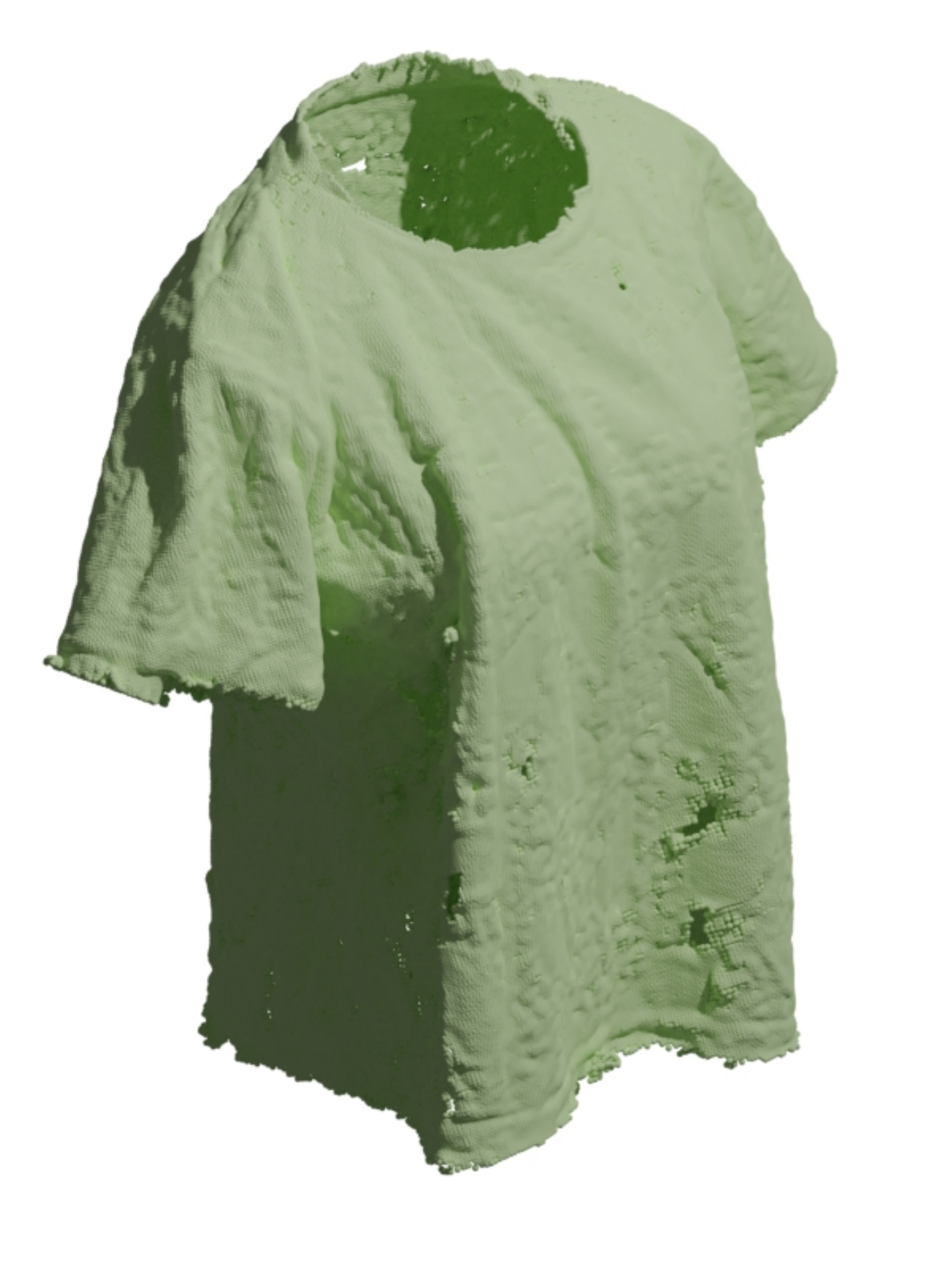}
    \includegraphics[width=.45\linewidth]{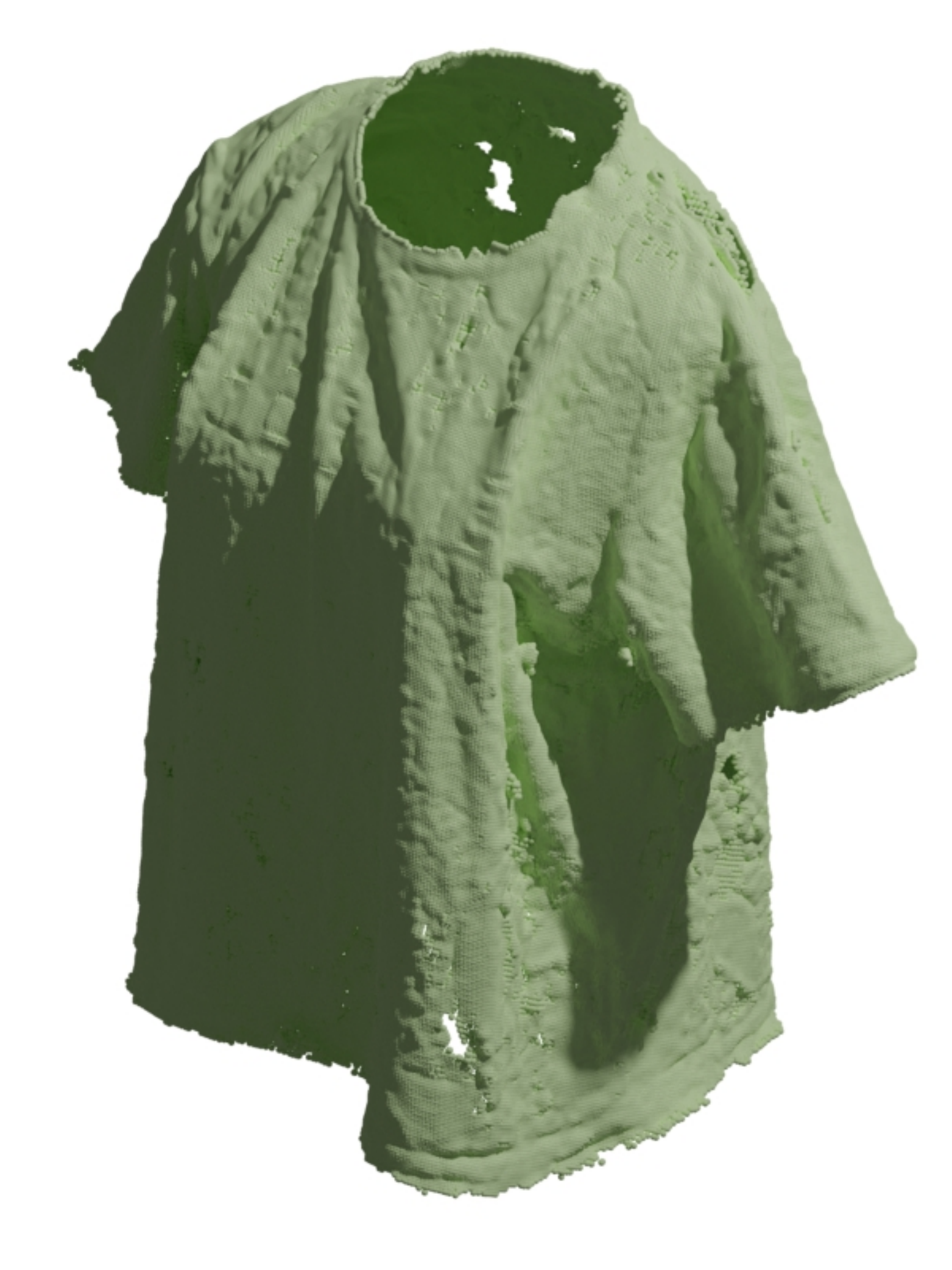}\\
    \includegraphics[width=.45\linewidth]{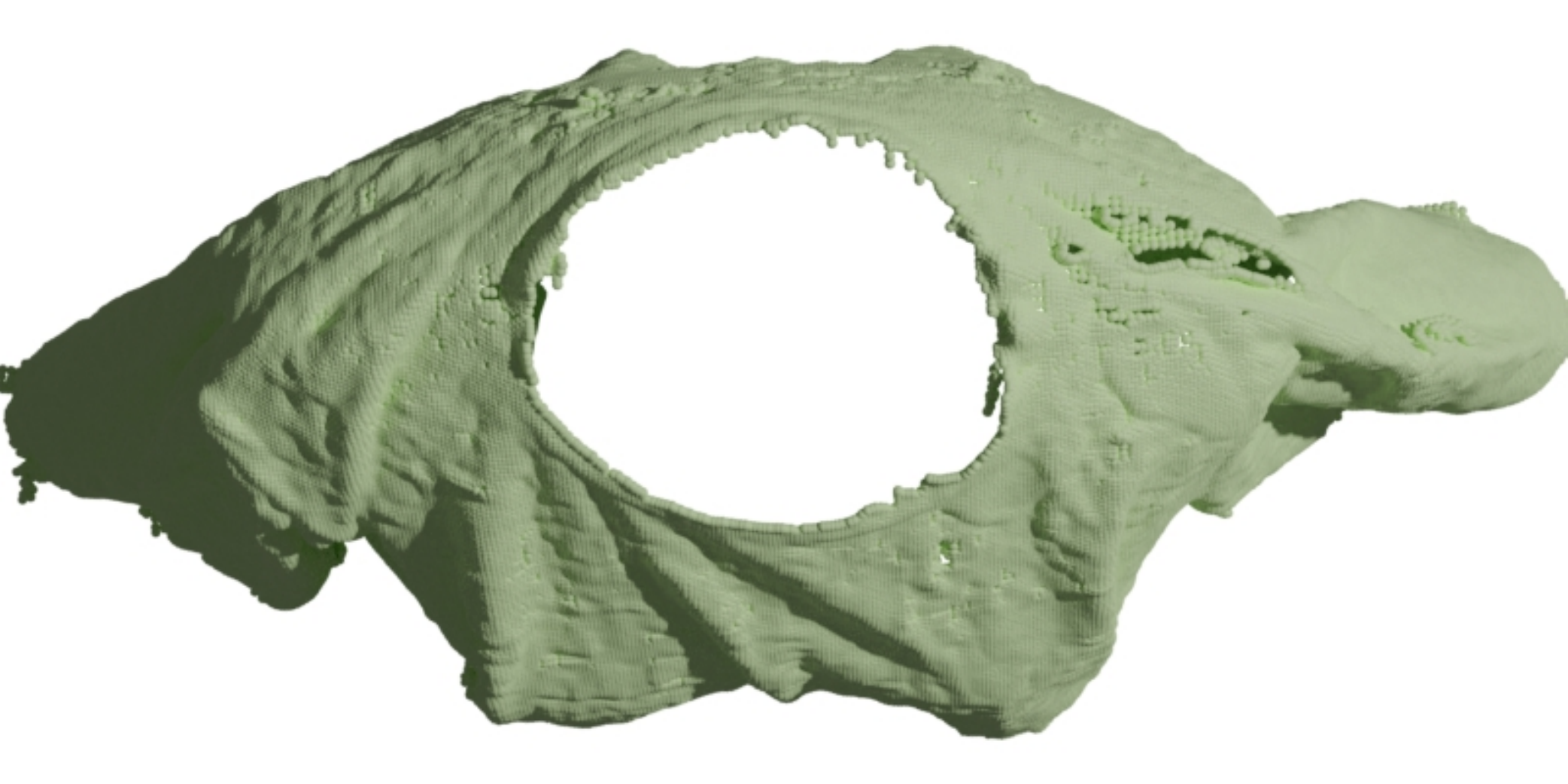}
\end{minipage}

\begin{minipage}[c]{.13\textwidth}
    \centering
    \includegraphics[width=1\linewidth]{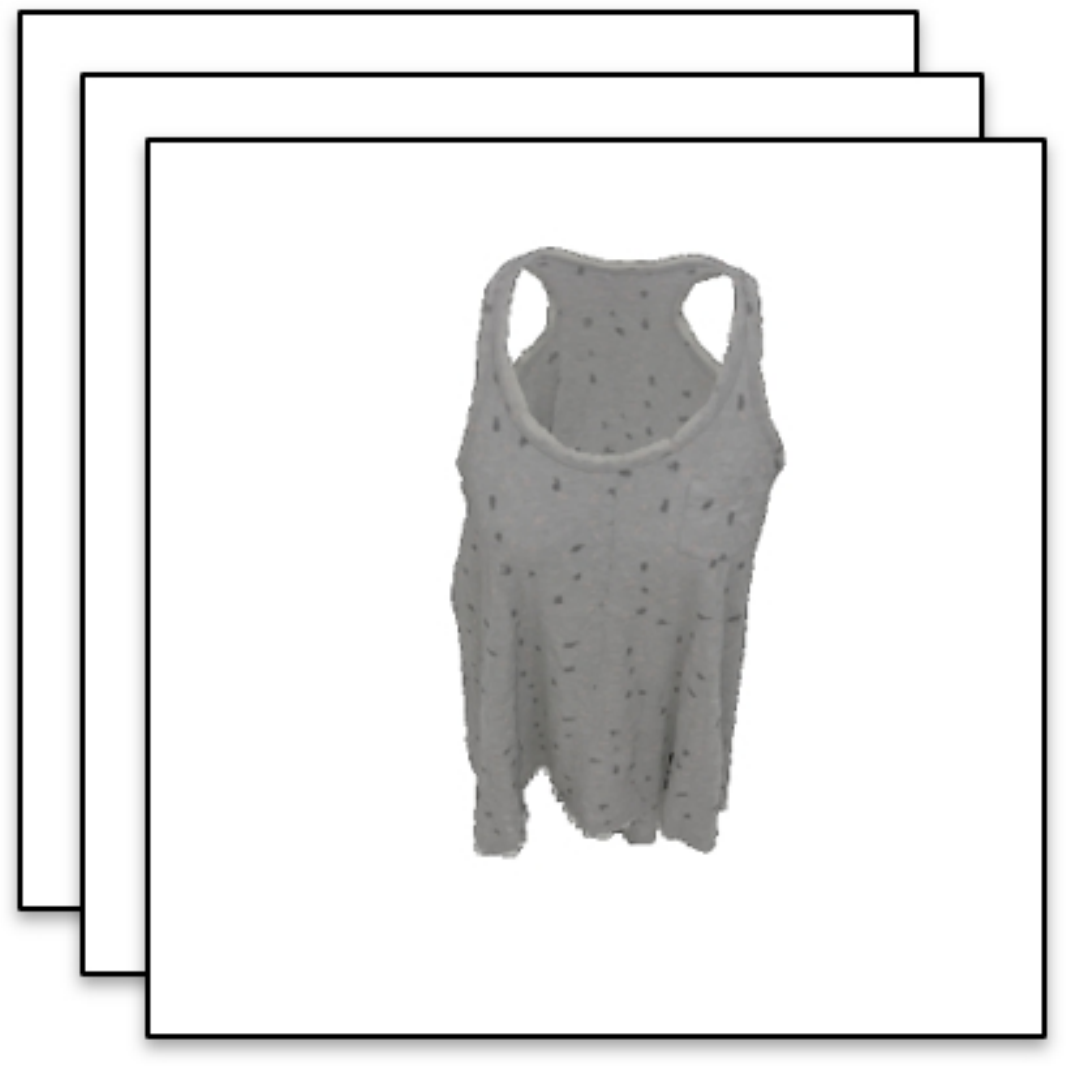}
\end{minipage}
\begin{minipage}[c]{.28\textwidth}
    \centering
    \includegraphics[width=.45\linewidth]{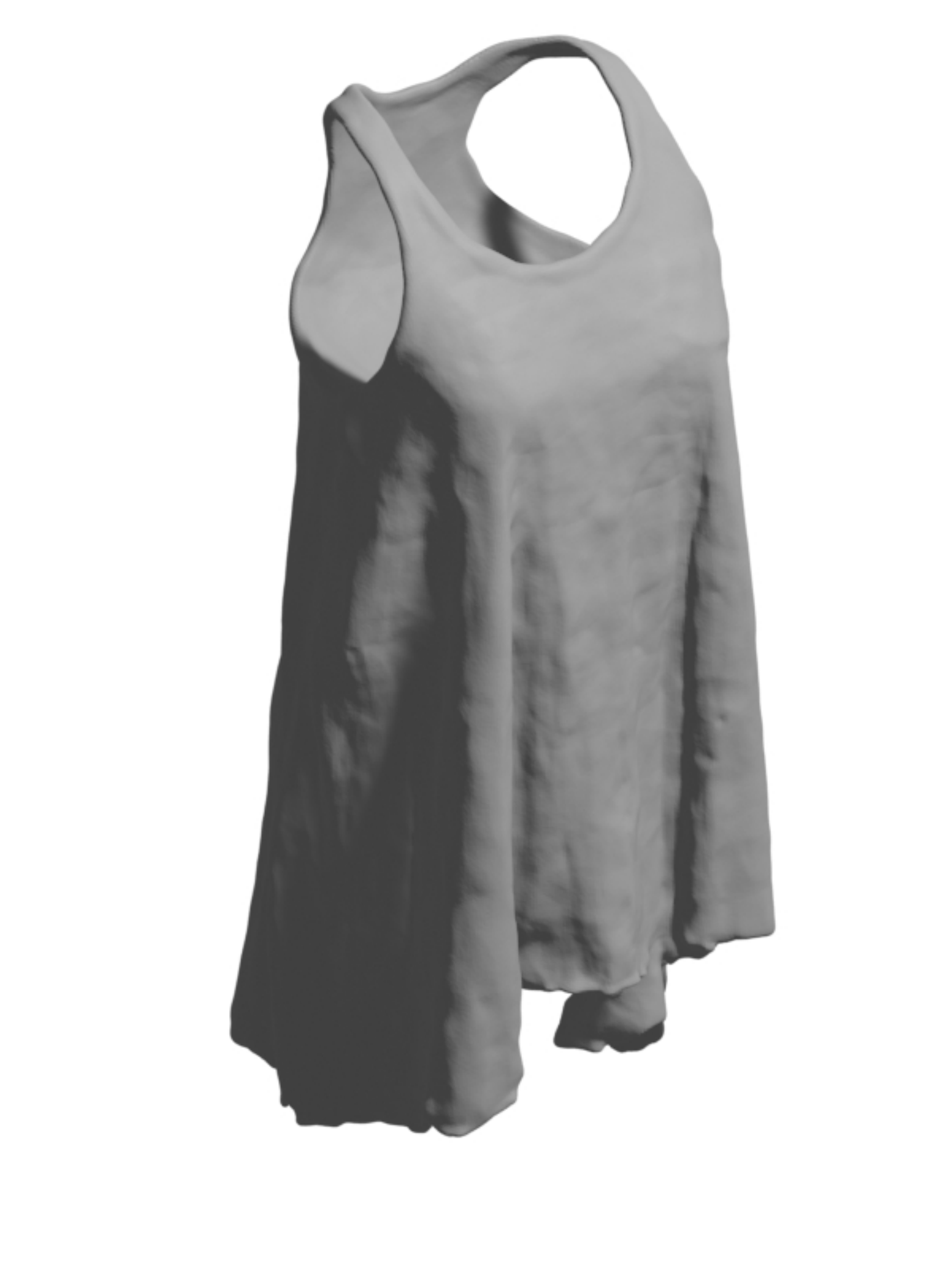}
    \includegraphics[width=.45\linewidth]{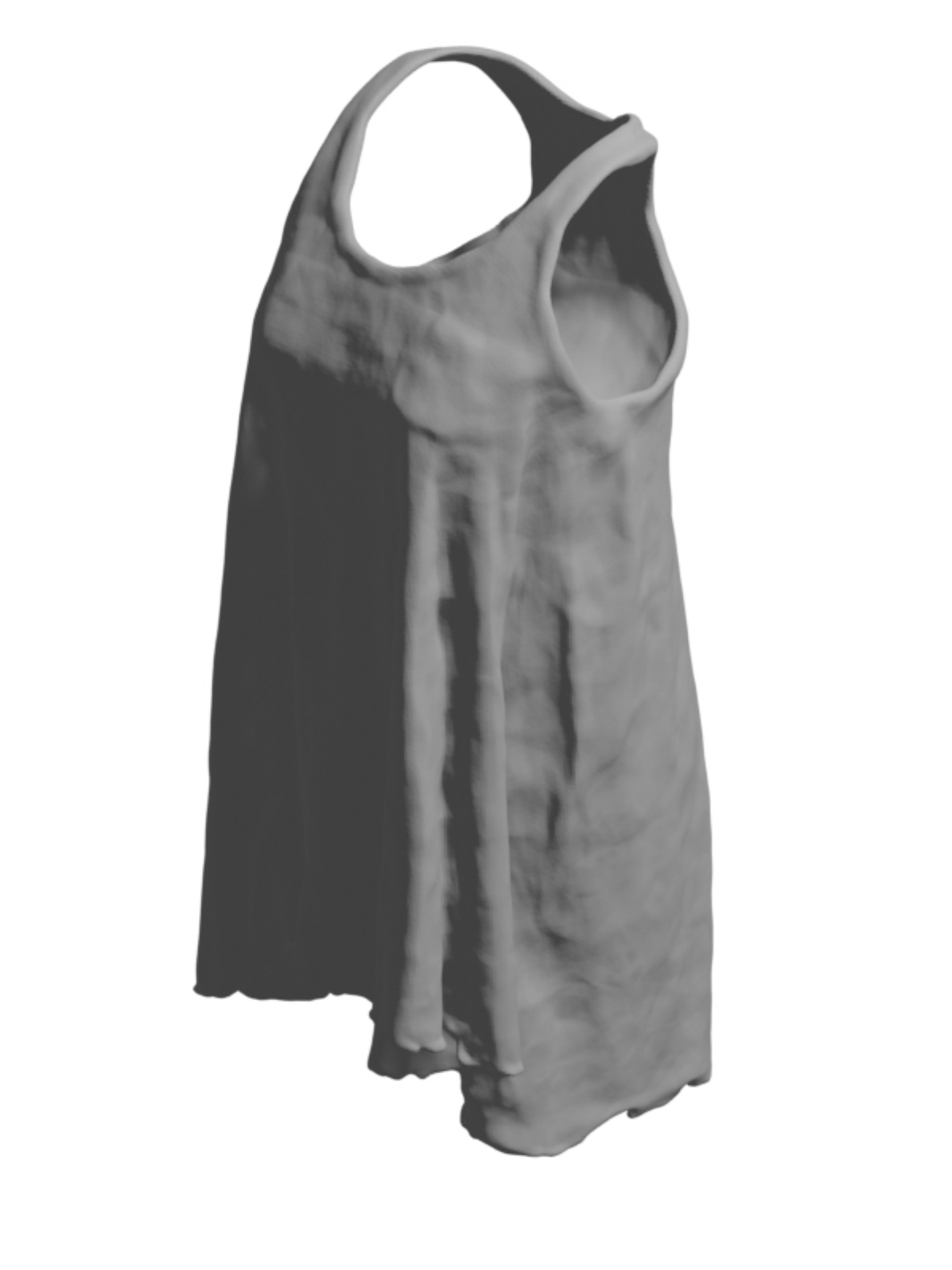}\\
    \includegraphics[width=.45\linewidth]{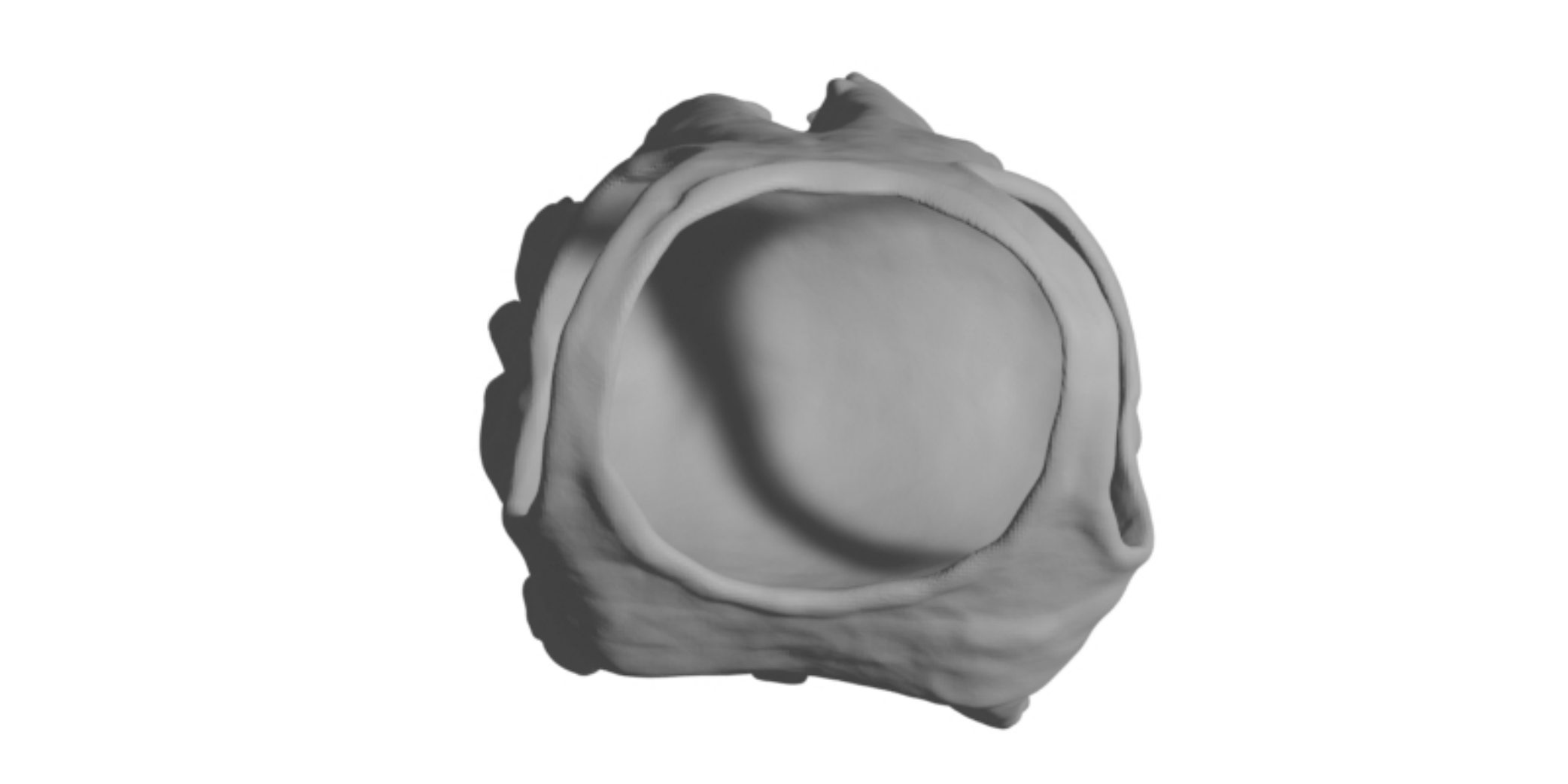}
\end{minipage}
\begin{minipage}[c]{.28\textwidth}
    \centering
    \includegraphics[width=.45\linewidth]{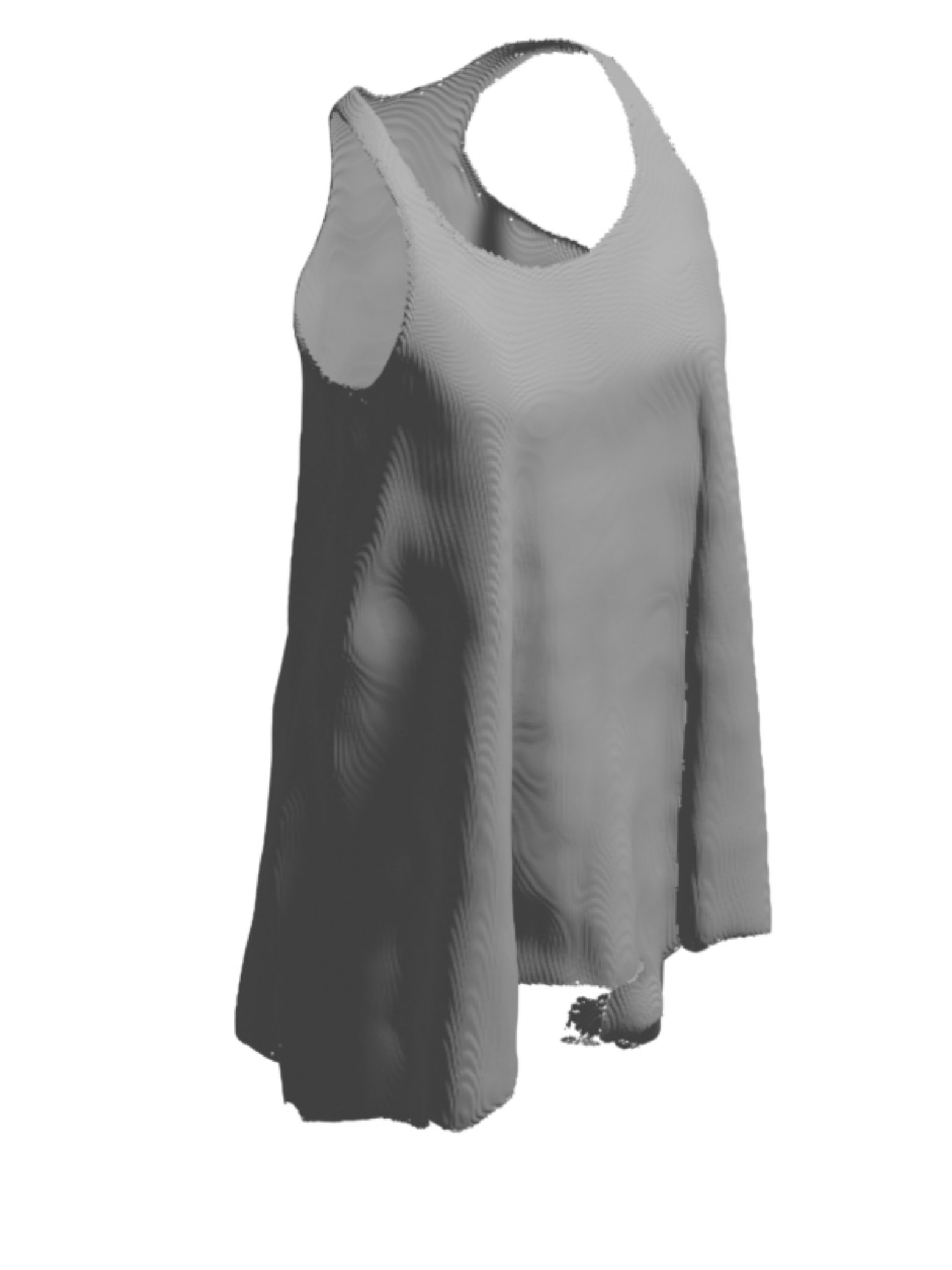}
    \includegraphics[width=.45\linewidth]{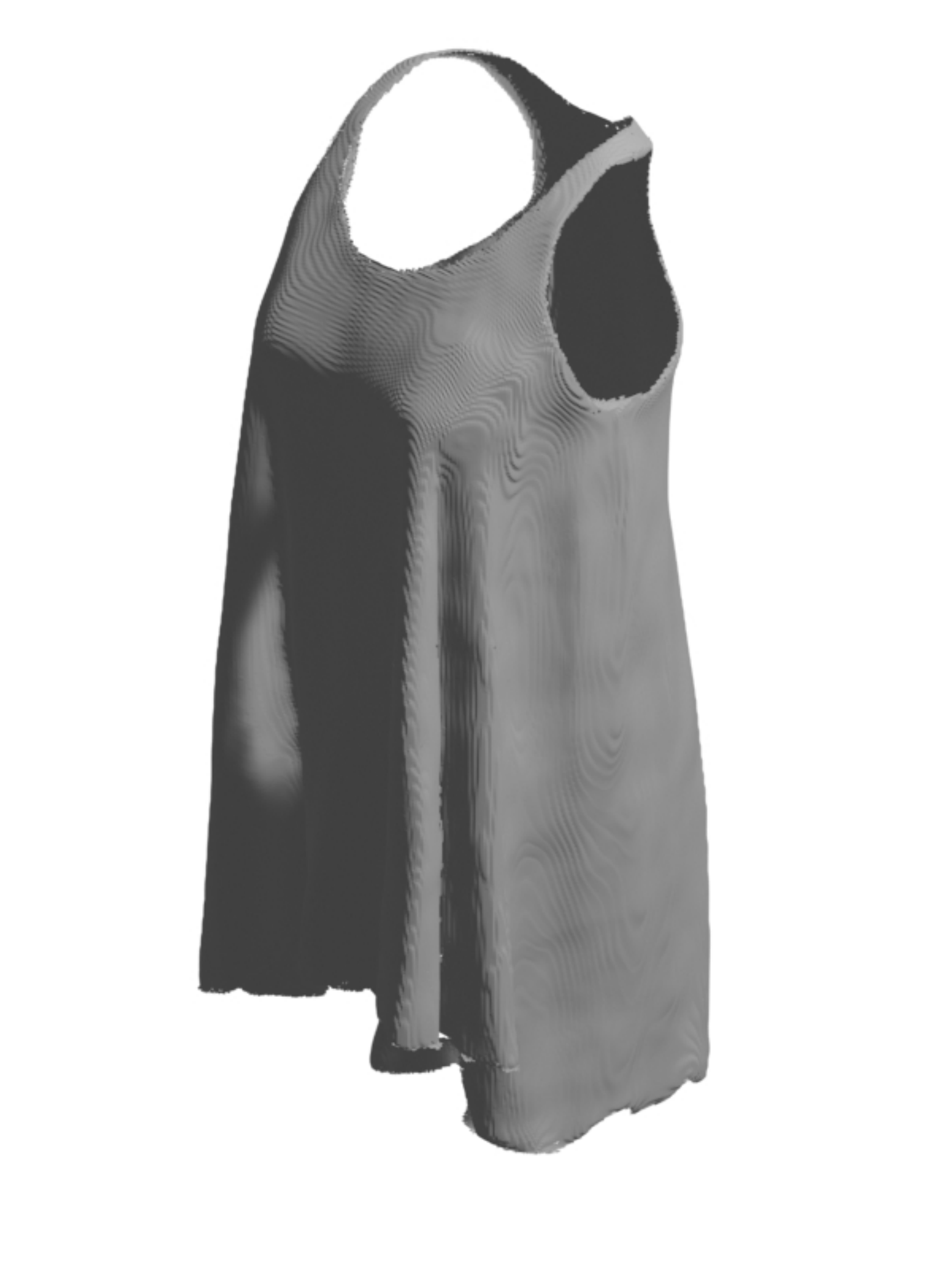}\\
    \includegraphics[width=.45\linewidth]{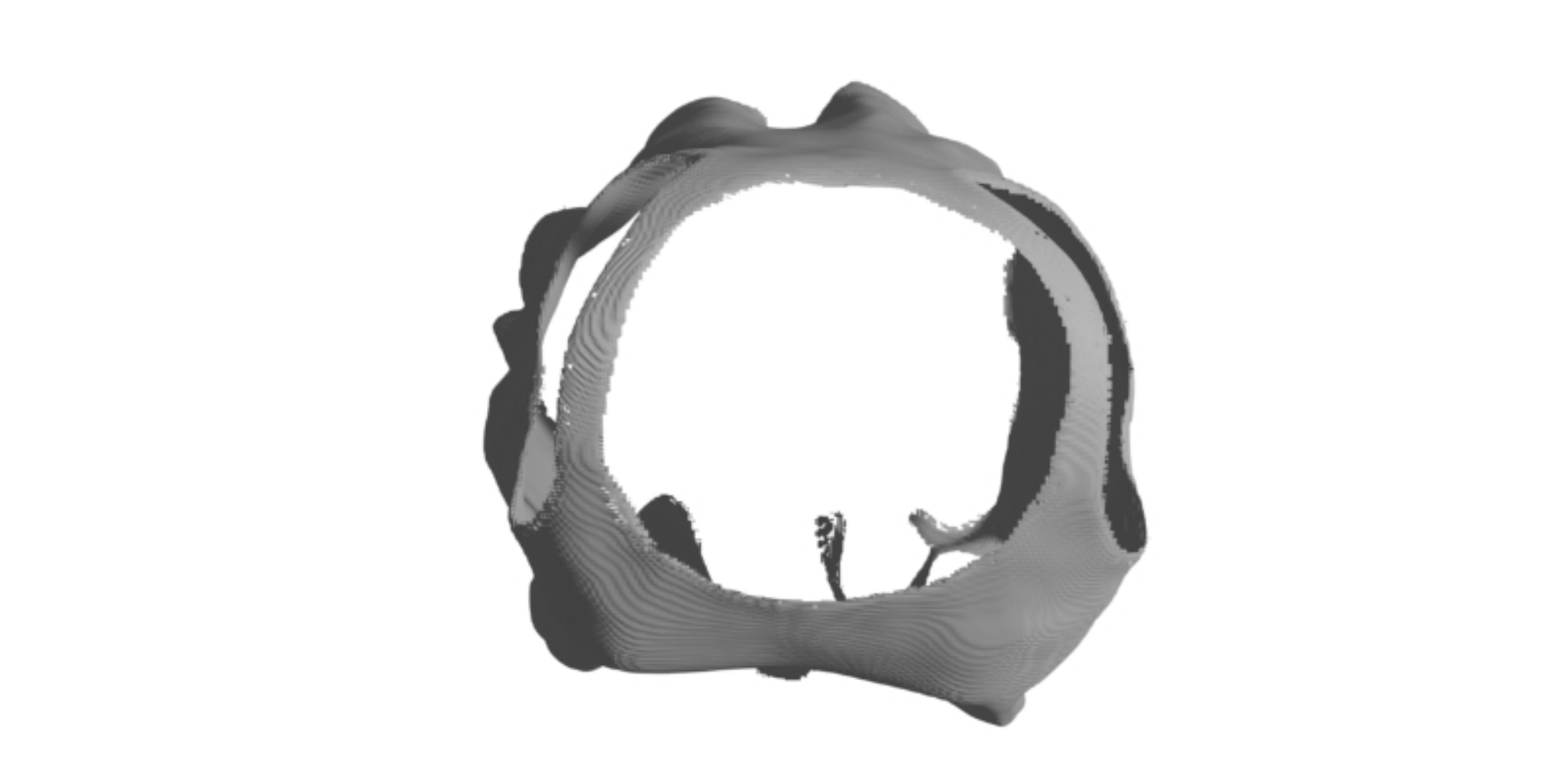}
\end{minipage}
\begin{minipage}[c]{.28\textwidth}
    \centering
    \includegraphics[width=.45\linewidth]{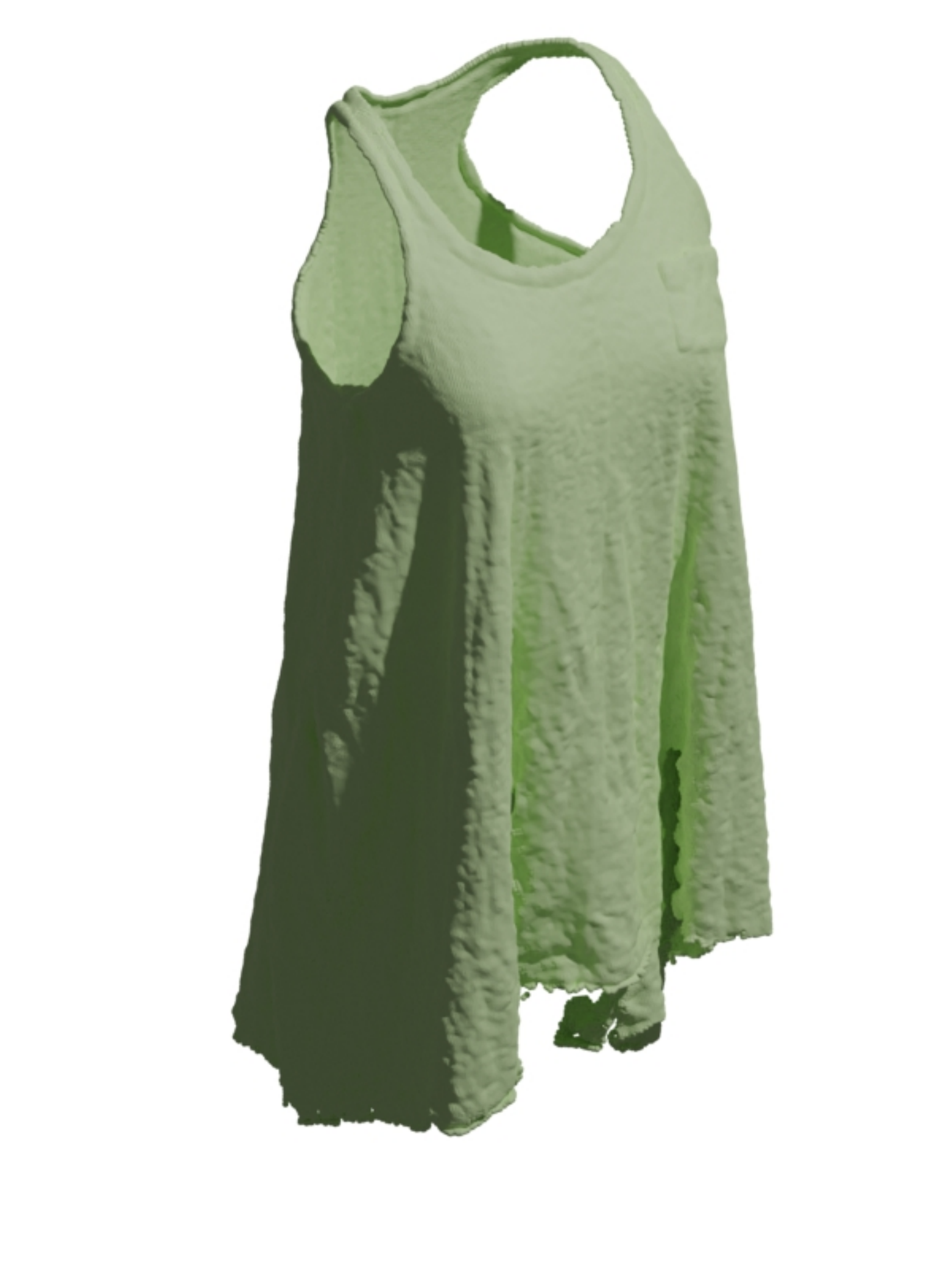}
    \includegraphics[width=.45\linewidth]{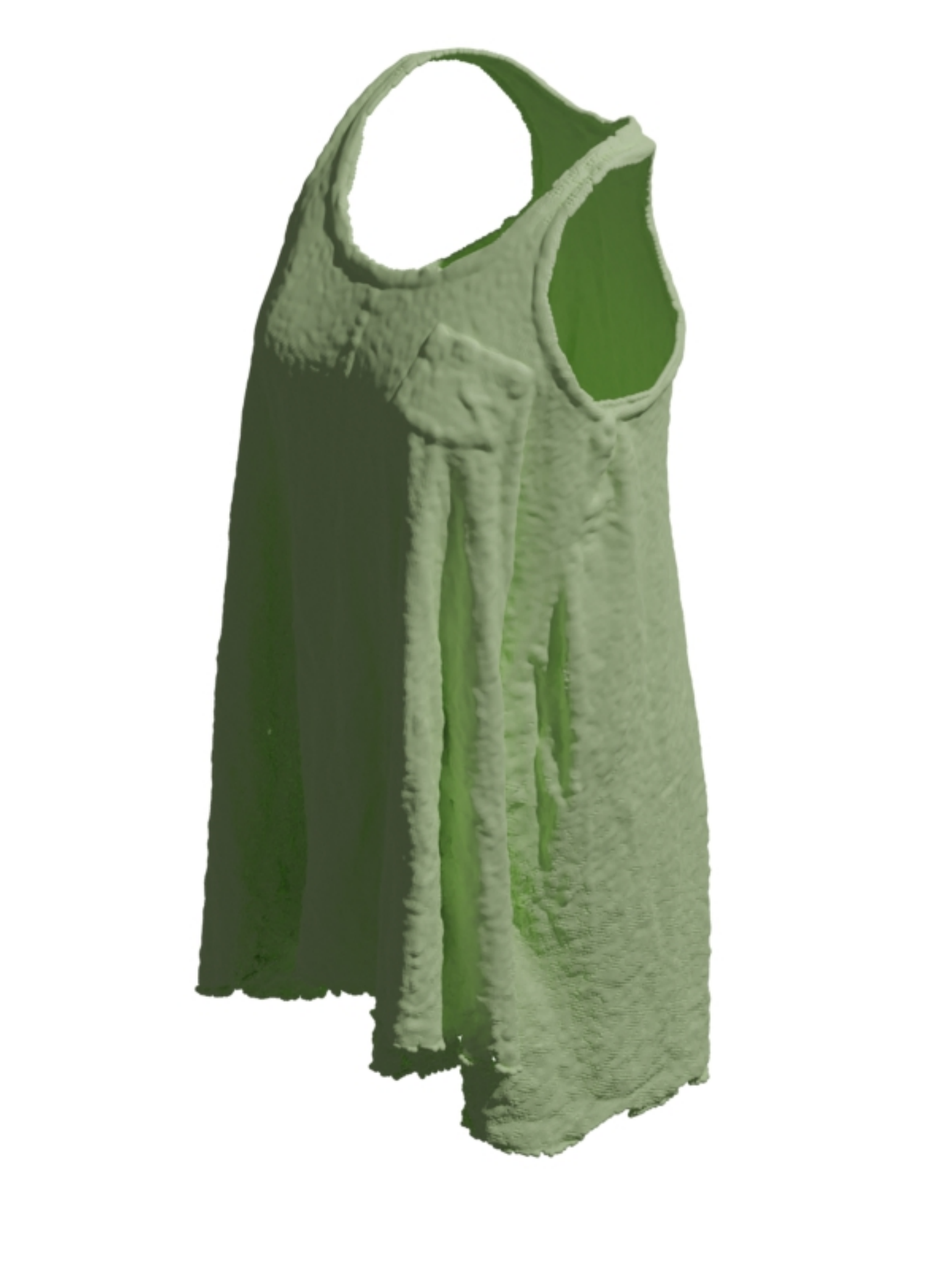}\\
    \includegraphics[width=.45\linewidth]{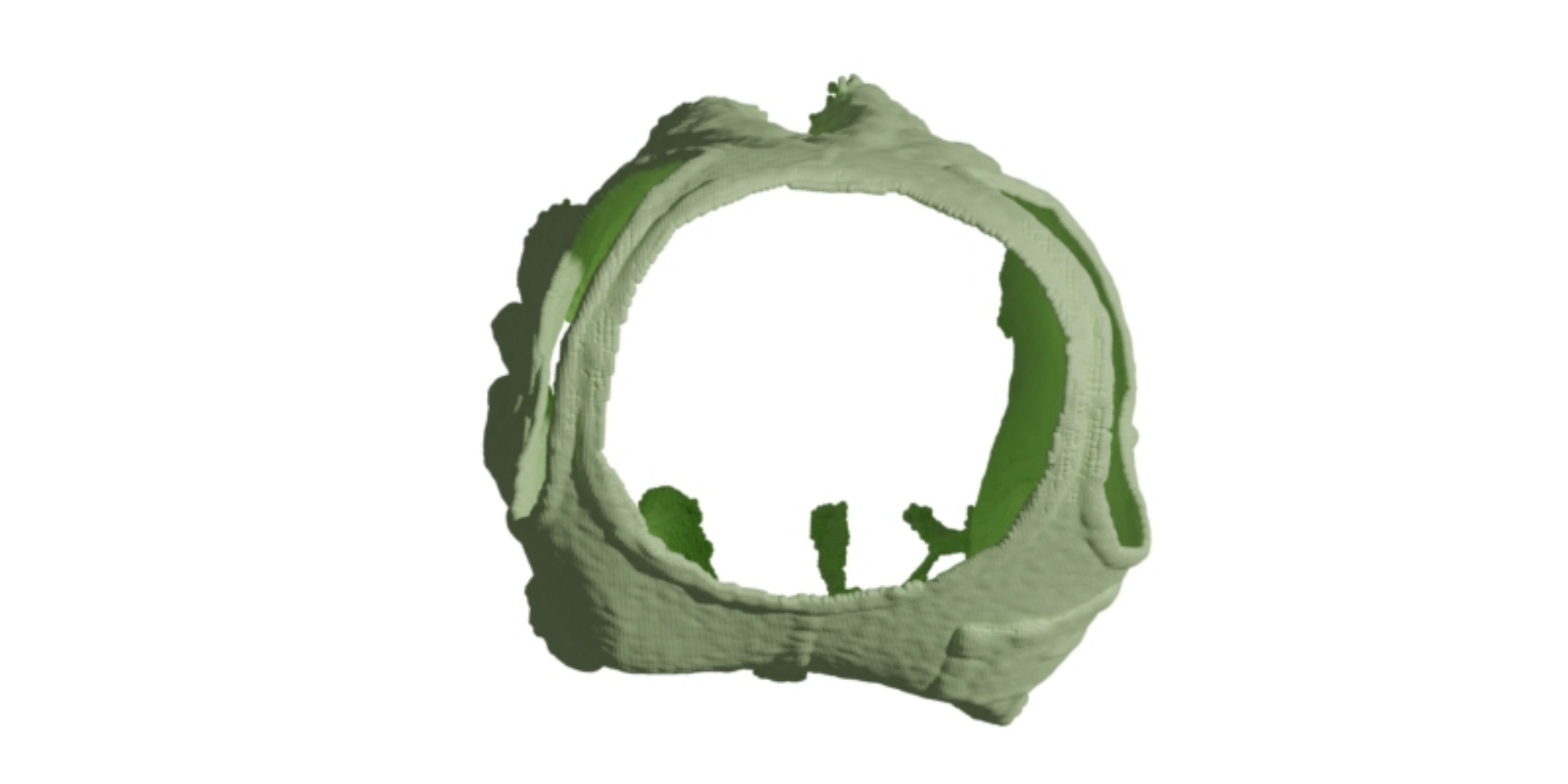}
\end{minipage}

\begin{minipage}[c]{.13\textwidth}
    \centering
    \includegraphics[width=1\linewidth]{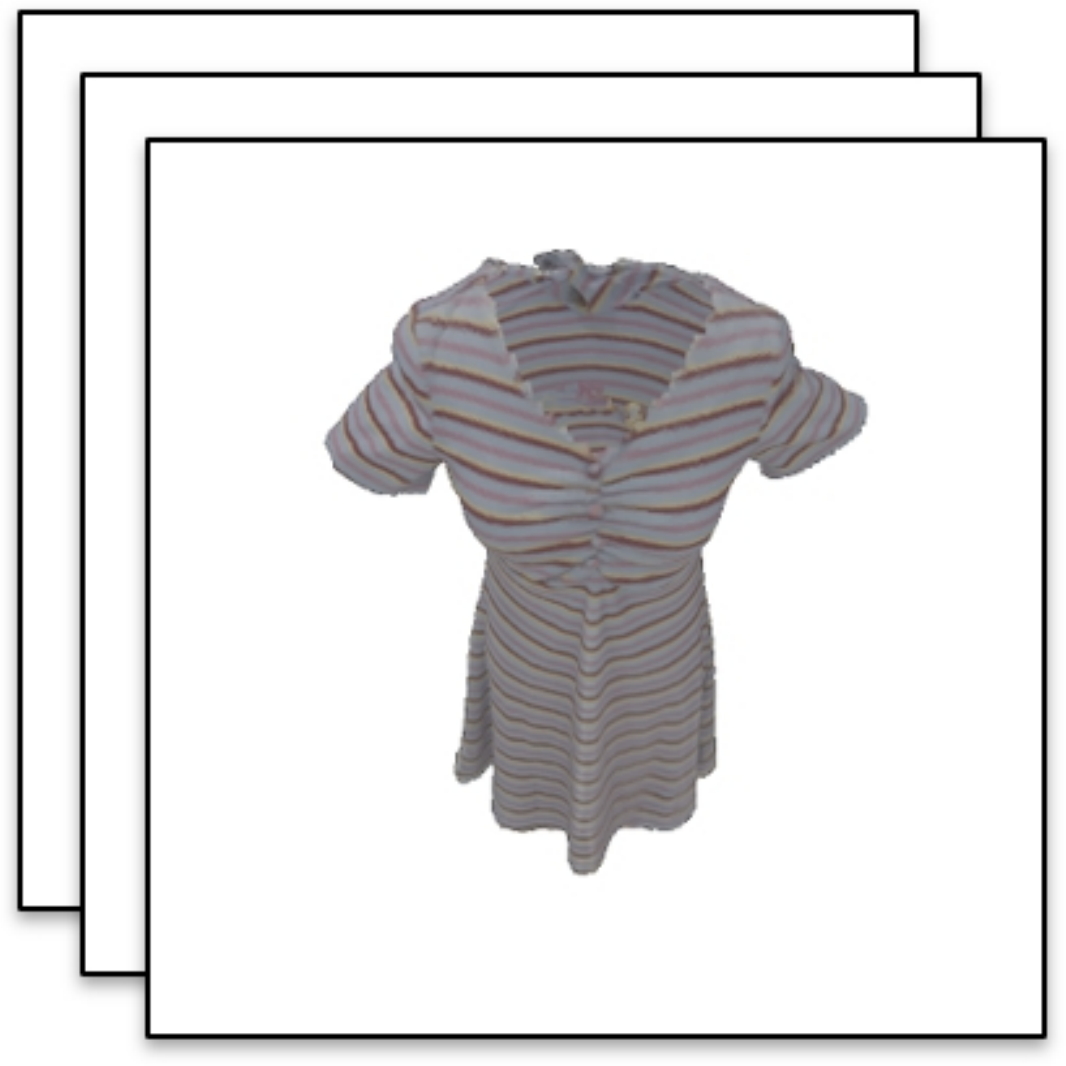}
\end{minipage}
\begin{minipage}[c]{.28\textwidth}
    \centering
    \includegraphics[width=.45\linewidth]{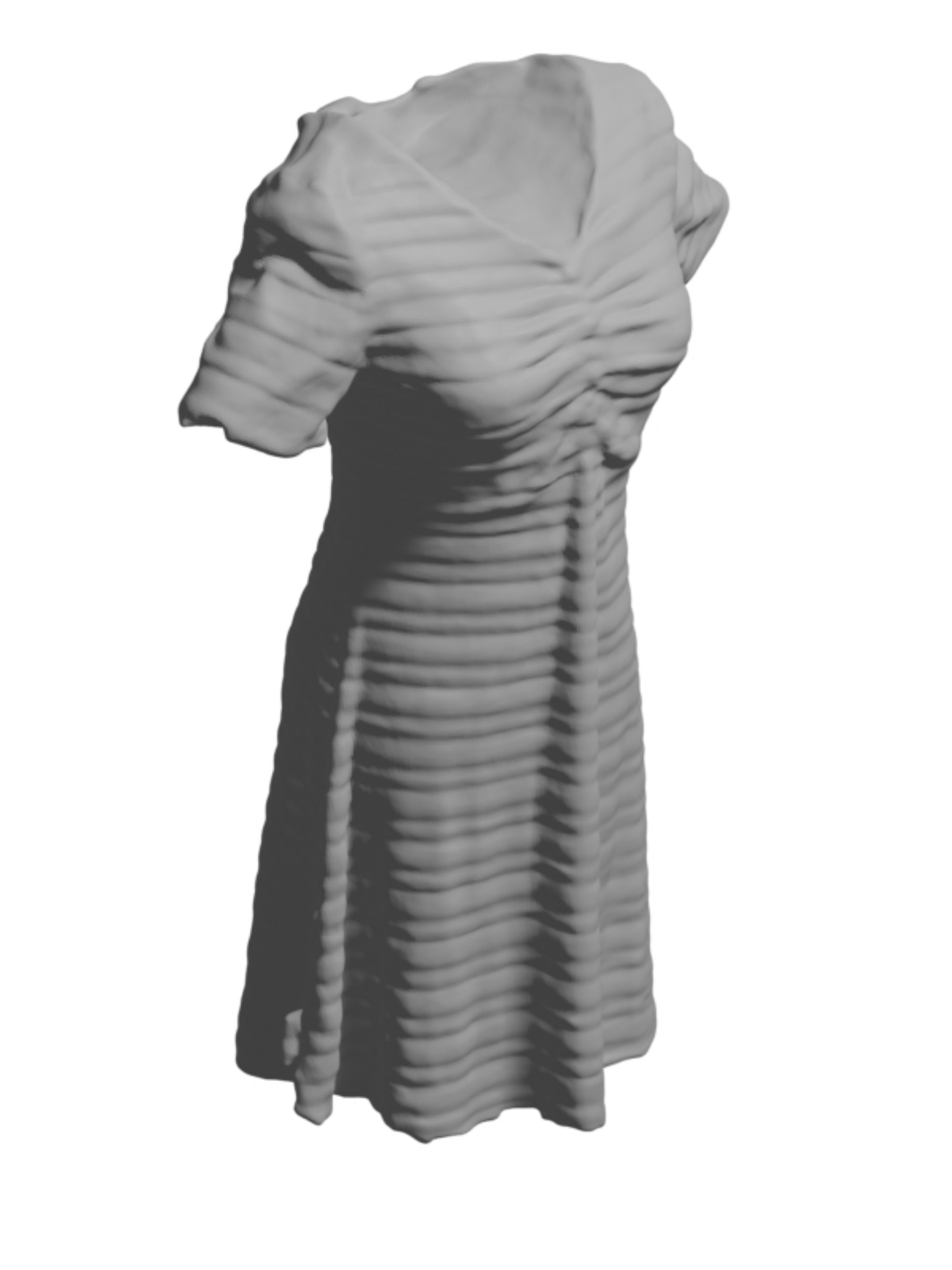}
    \includegraphics[width=.45\linewidth]{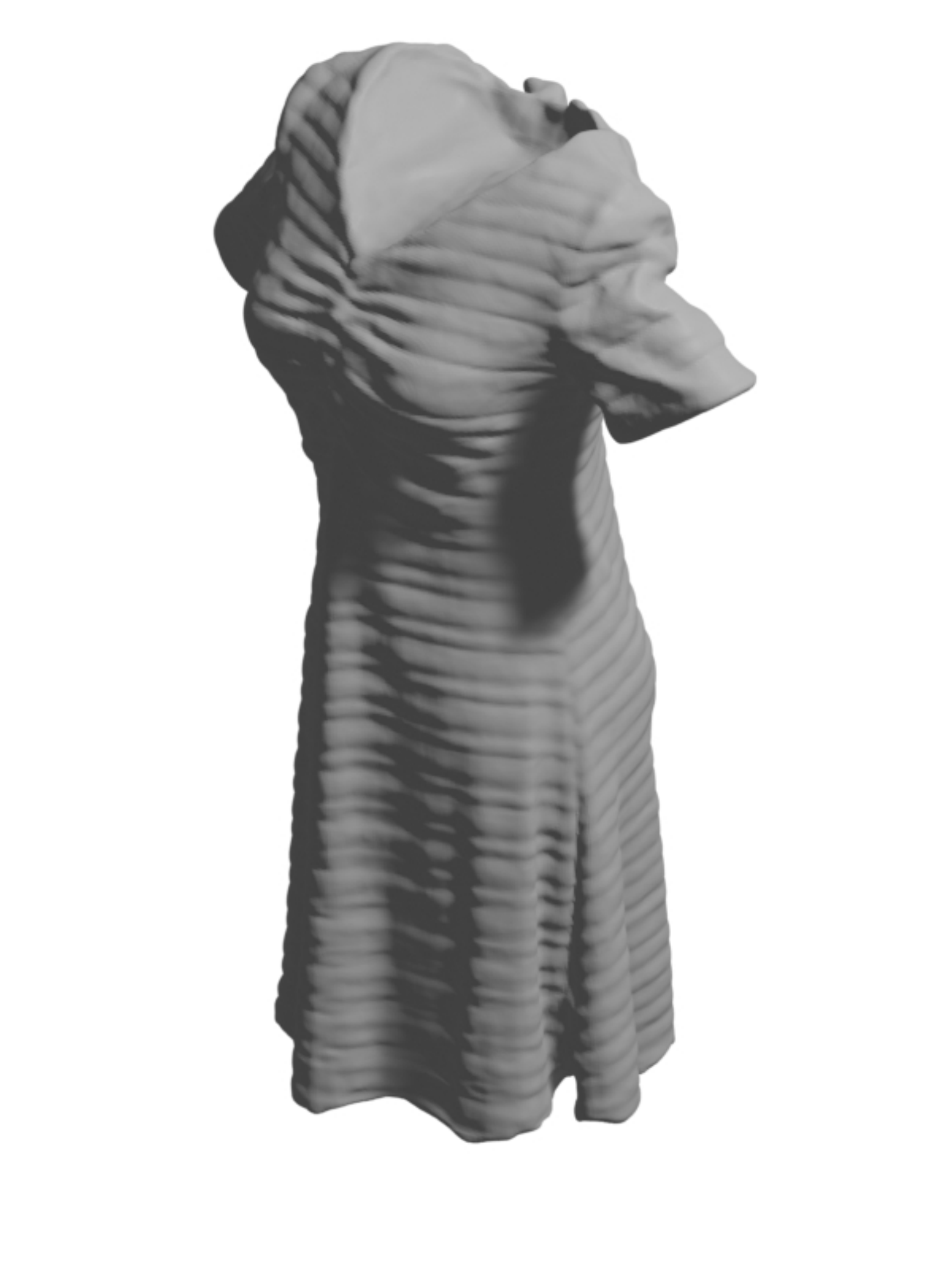}\\
    \includegraphics[width=.45\linewidth]{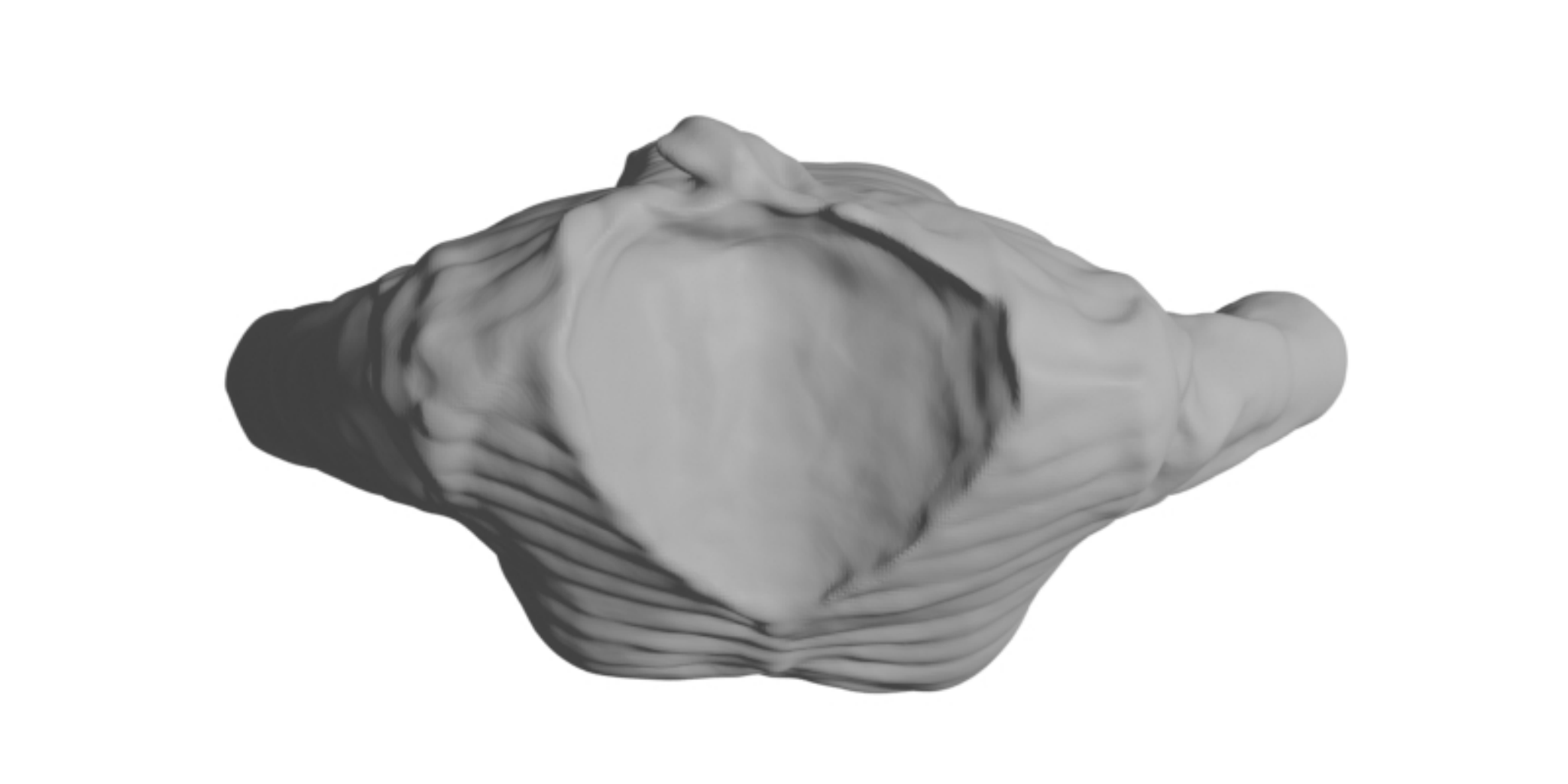}
\end{minipage}
\begin{minipage}[c]{.28\textwidth}
    \centering
    \includegraphics[width=.45\linewidth]{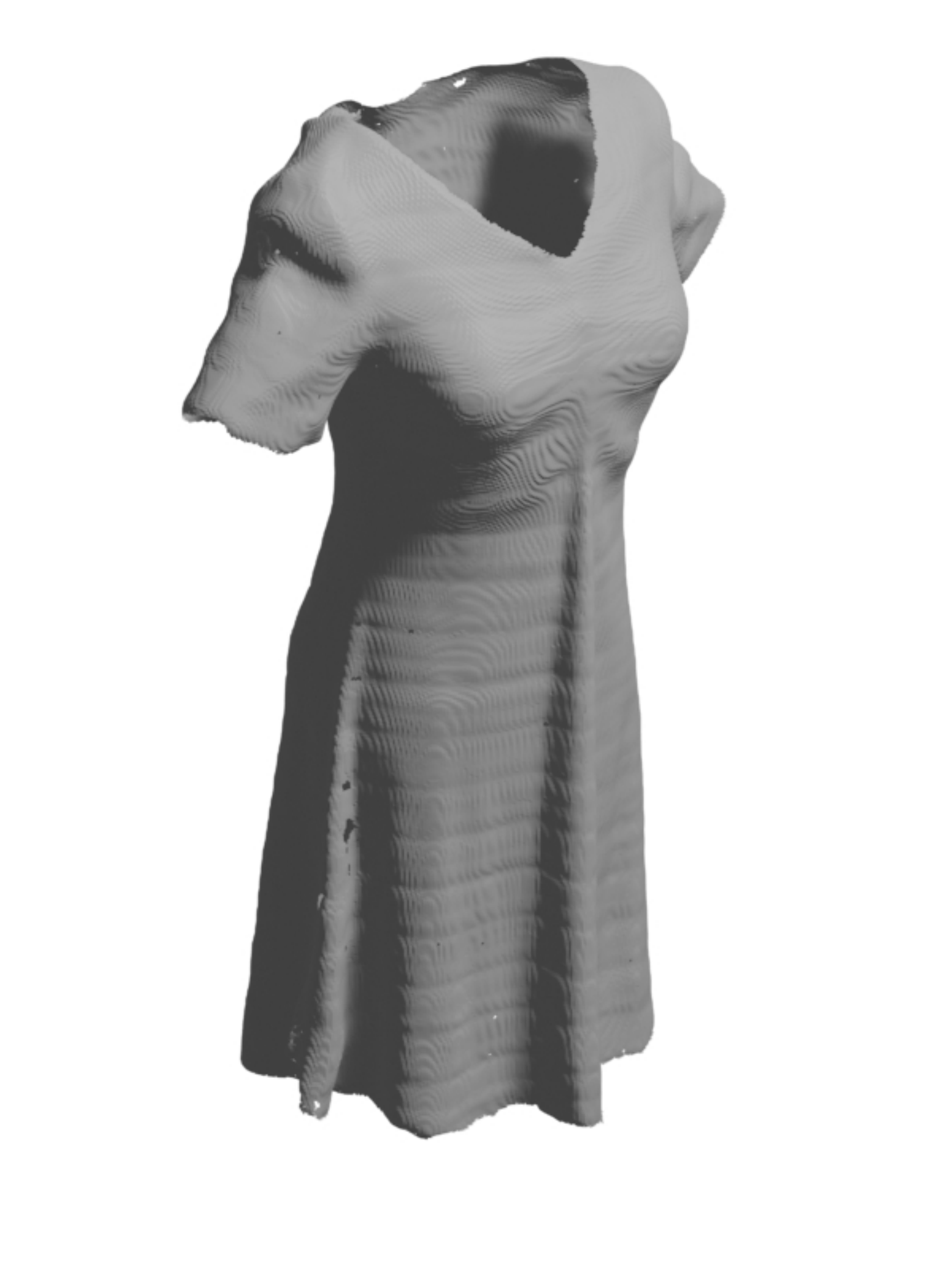}
    \includegraphics[width=.45\linewidth]{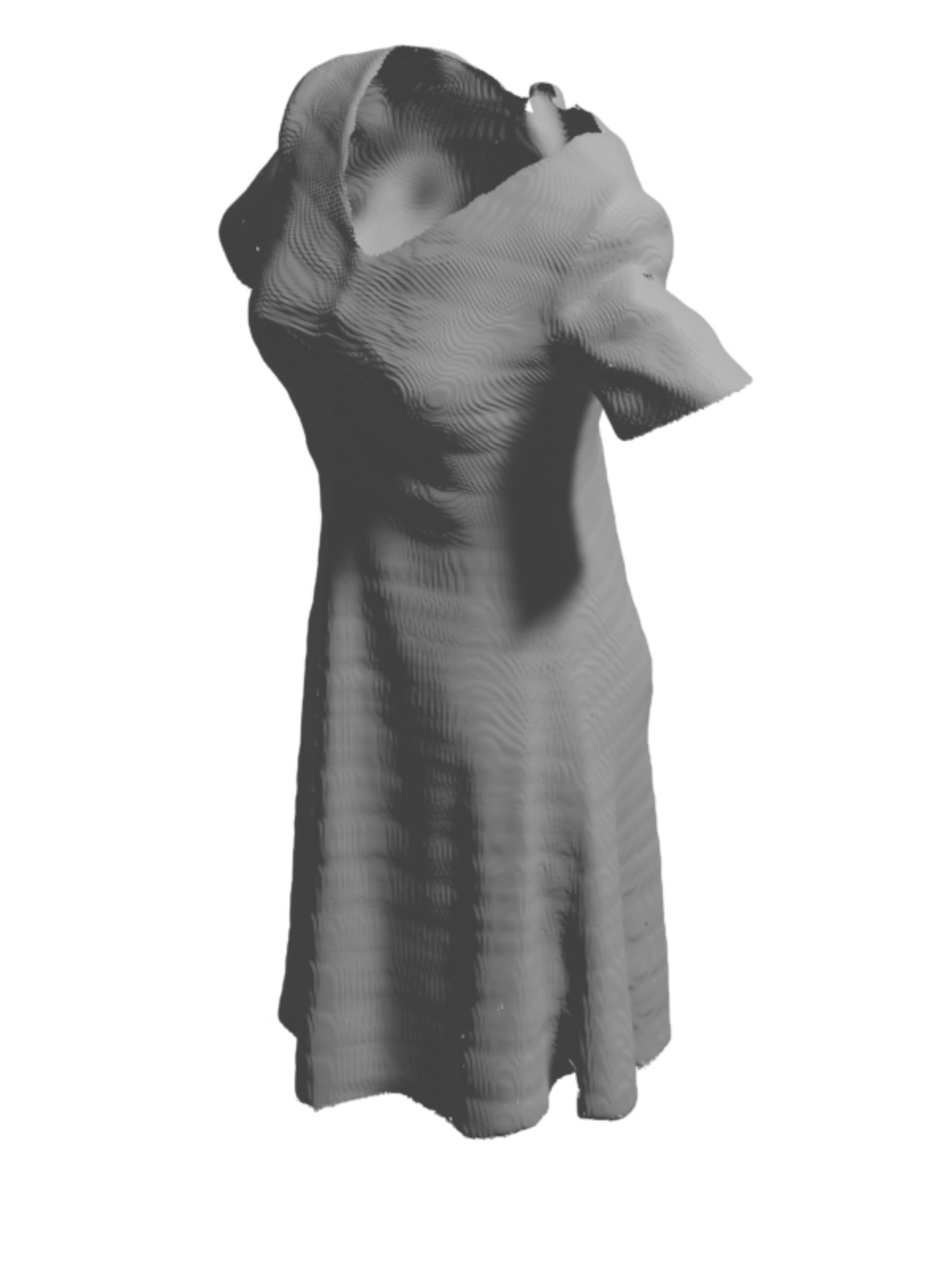}\\
    \includegraphics[width=.45\linewidth]{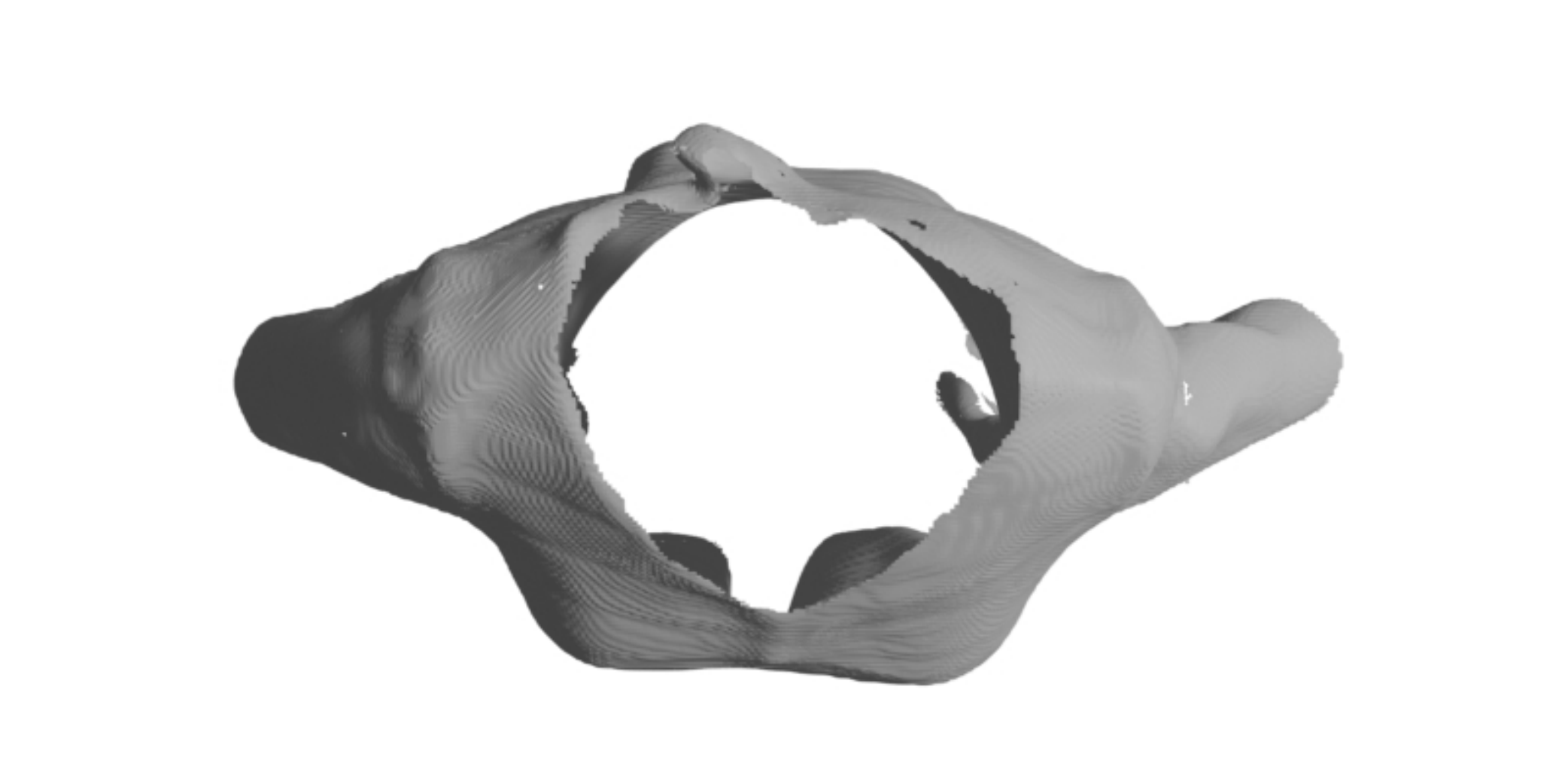}
\end{minipage}
\begin{minipage}[c]{.28\textwidth}
    \centering
    \includegraphics[width=.45\linewidth]{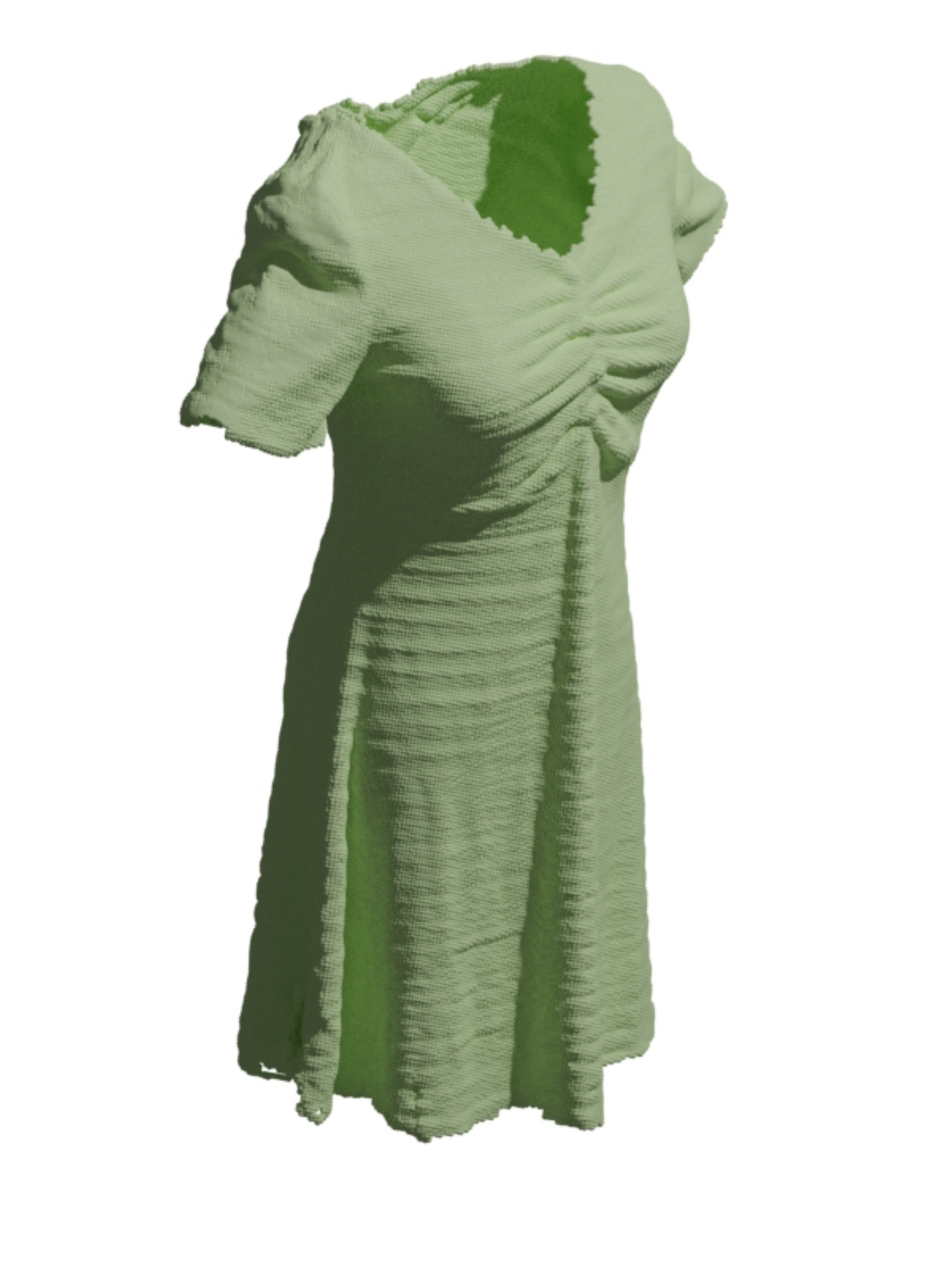}
    \includegraphics[width=.45\linewidth]{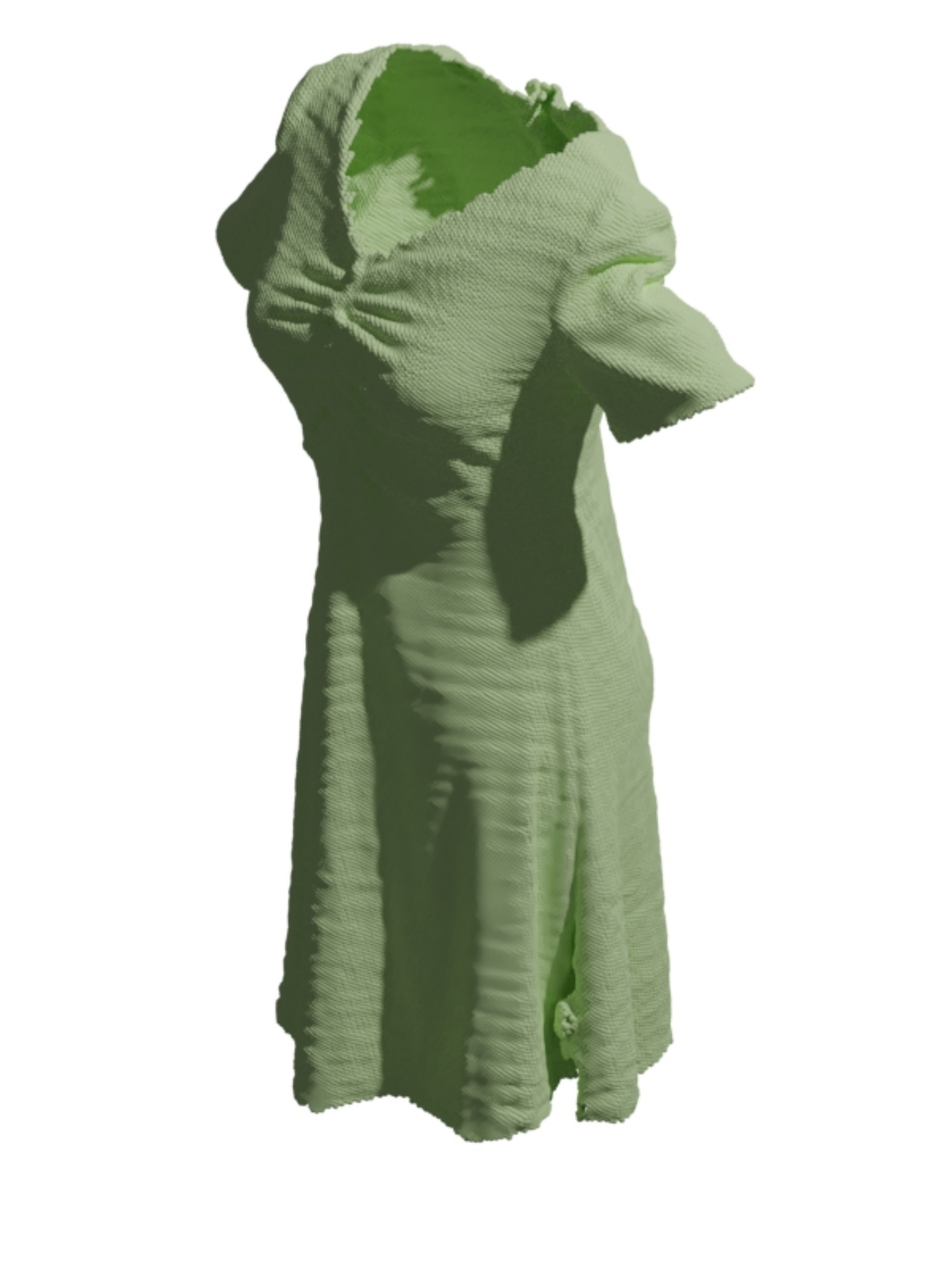}\\
    \includegraphics[width=.45\linewidth]{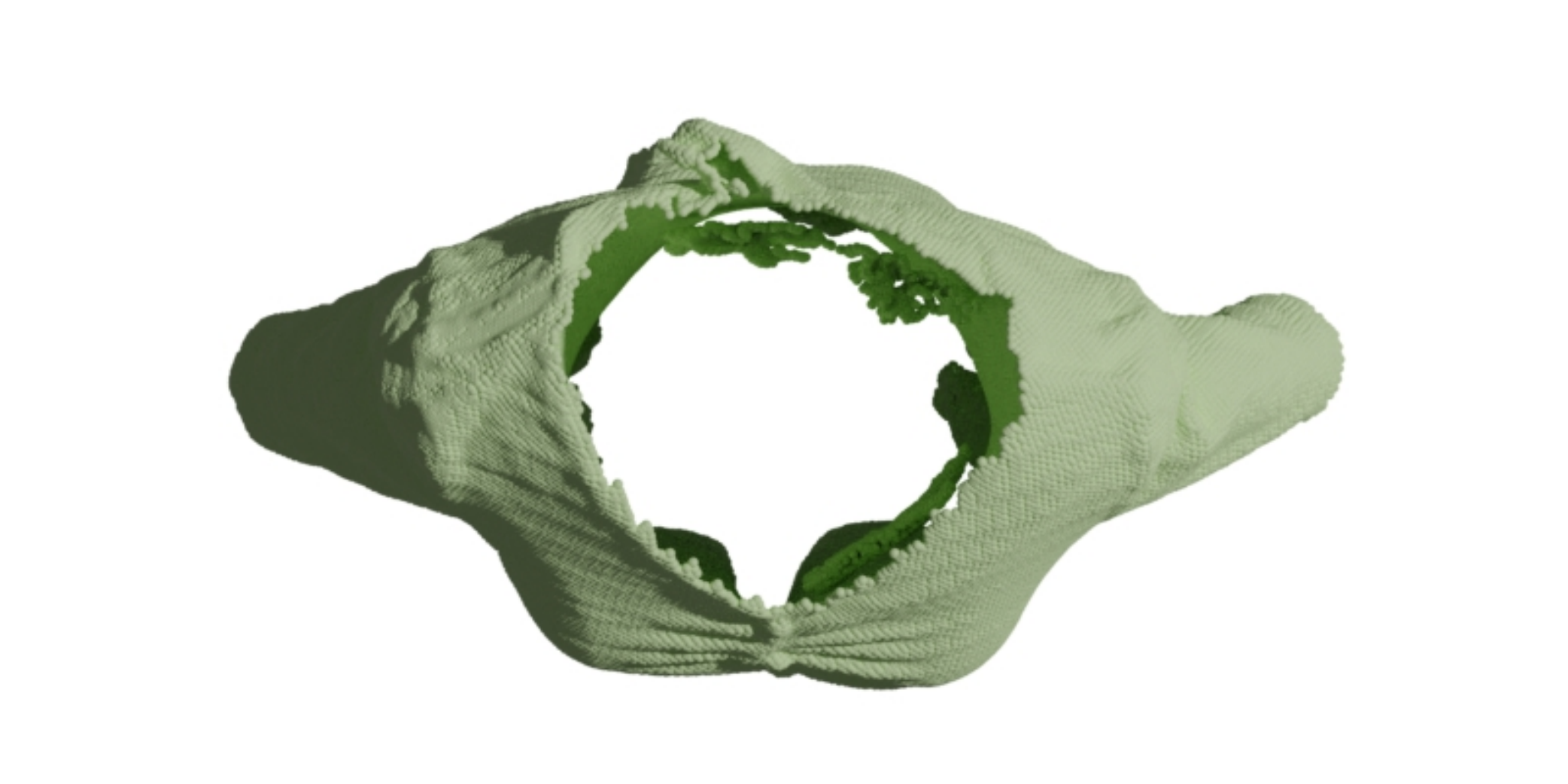}
\end{minipage}

\begin{minipage}[c]{.13\textwidth}
    \centering
    Input
\end{minipage}
\begin{minipage}[c]{.28\textwidth}
    \centering
    NeuS
\end{minipage}
\begin{minipage}[c]{.28\textwidth}
    \centering
    Ours
\end{minipage}
\begin{minipage}[c]{.28\textwidth}
    \centering
    Ground-truth
\end{minipage}

\caption{Additional results on the DF3D~\cite{zhu2020deep} dataset without mask supervision.}
\label{supp_df3d_wo}
\end{figure*}

\begin{figure*}[h]

\begin{minipage}[c]{.13\textwidth}
    \centering
    \includegraphics[width=1\linewidth]{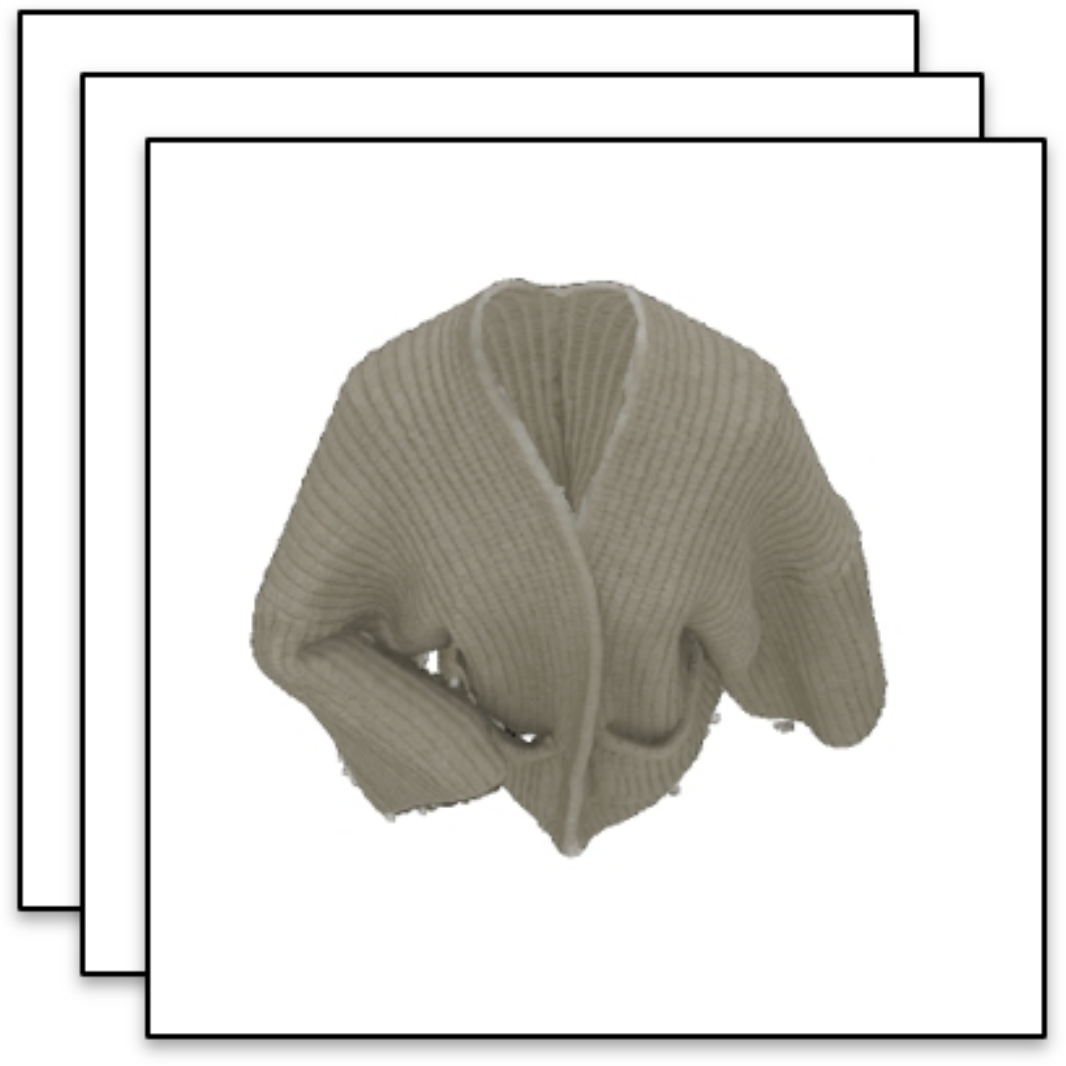}
\end{minipage}
\begin{minipage}[c]{.28\textwidth}
    \centering
    \includegraphics[width=.45\linewidth]{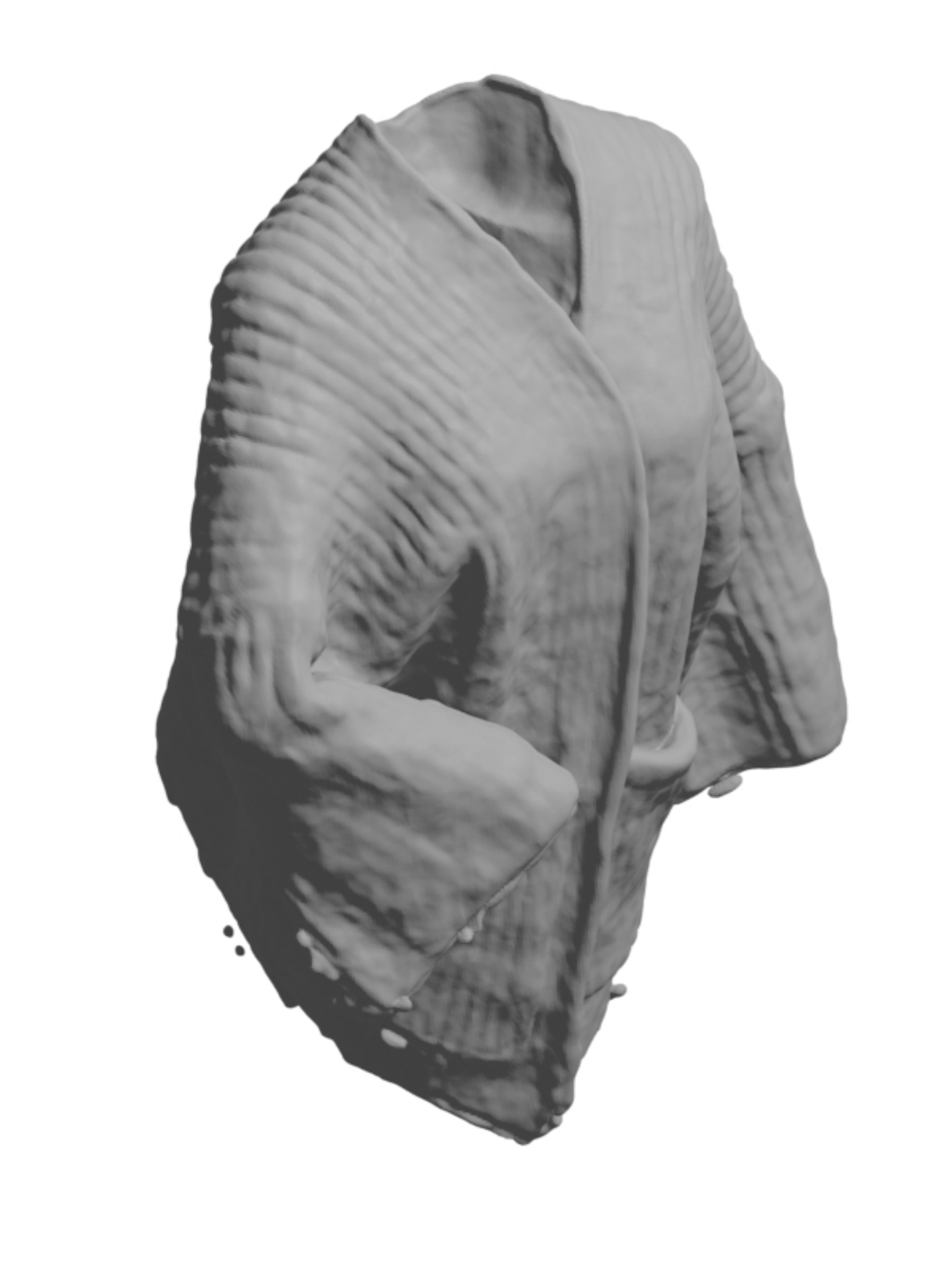}
    \includegraphics[width=.45\linewidth]{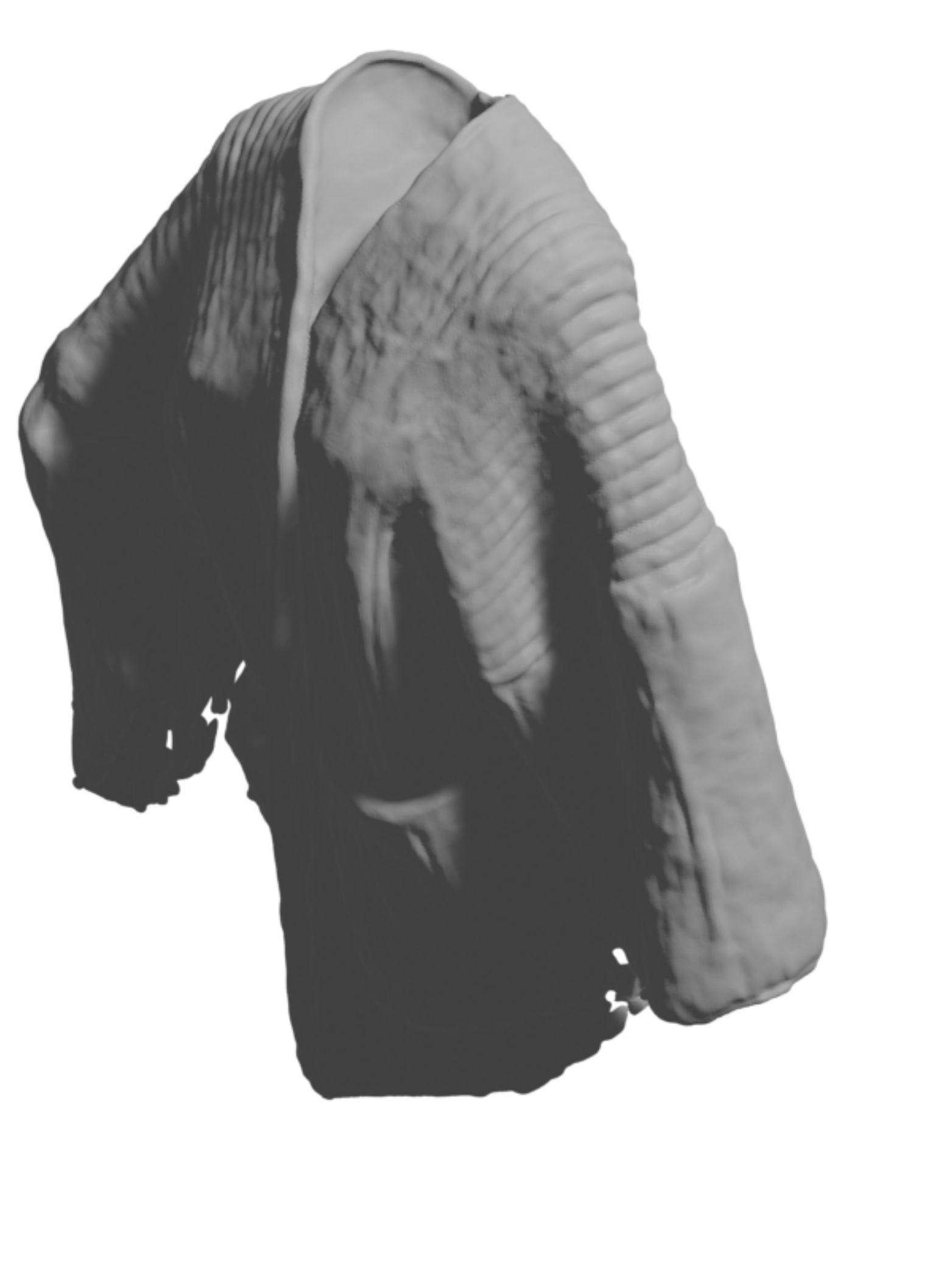}\\
    \includegraphics[width=.45\linewidth]{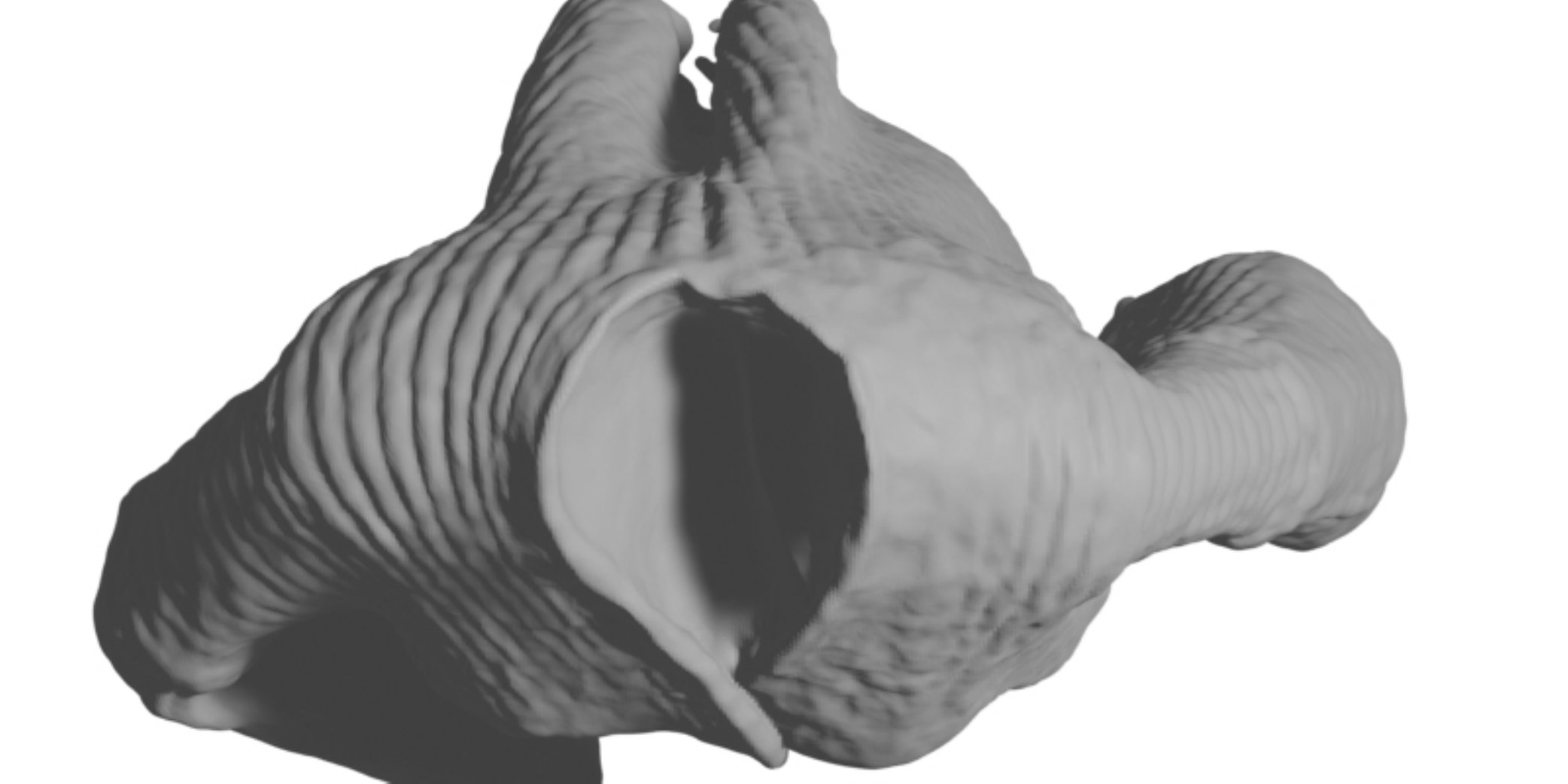}
\end{minipage}
\begin{minipage}[c]{.28\textwidth}
    \centering
    \includegraphics[width=.45\linewidth]{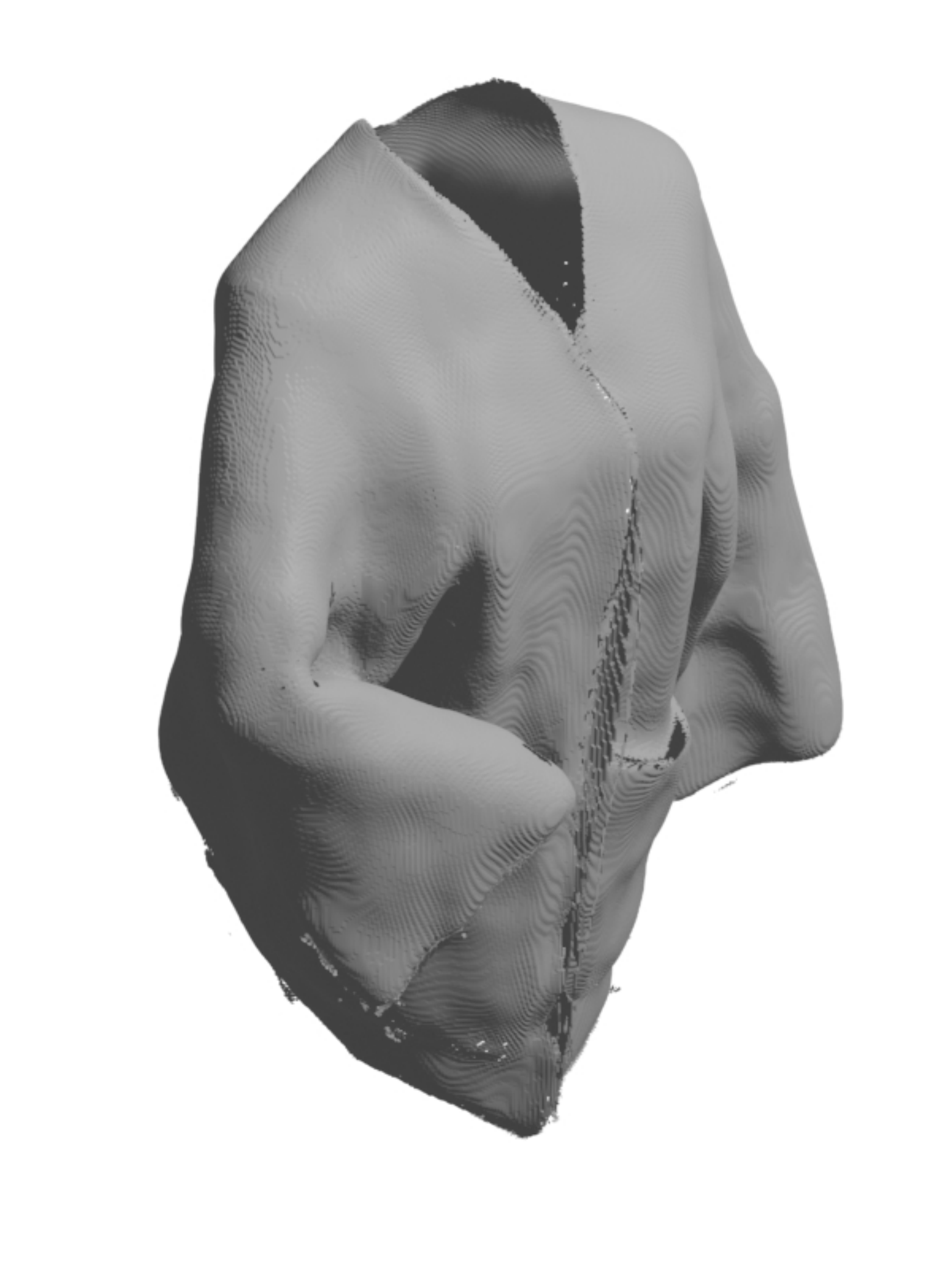}
    \includegraphics[width=.45\linewidth]{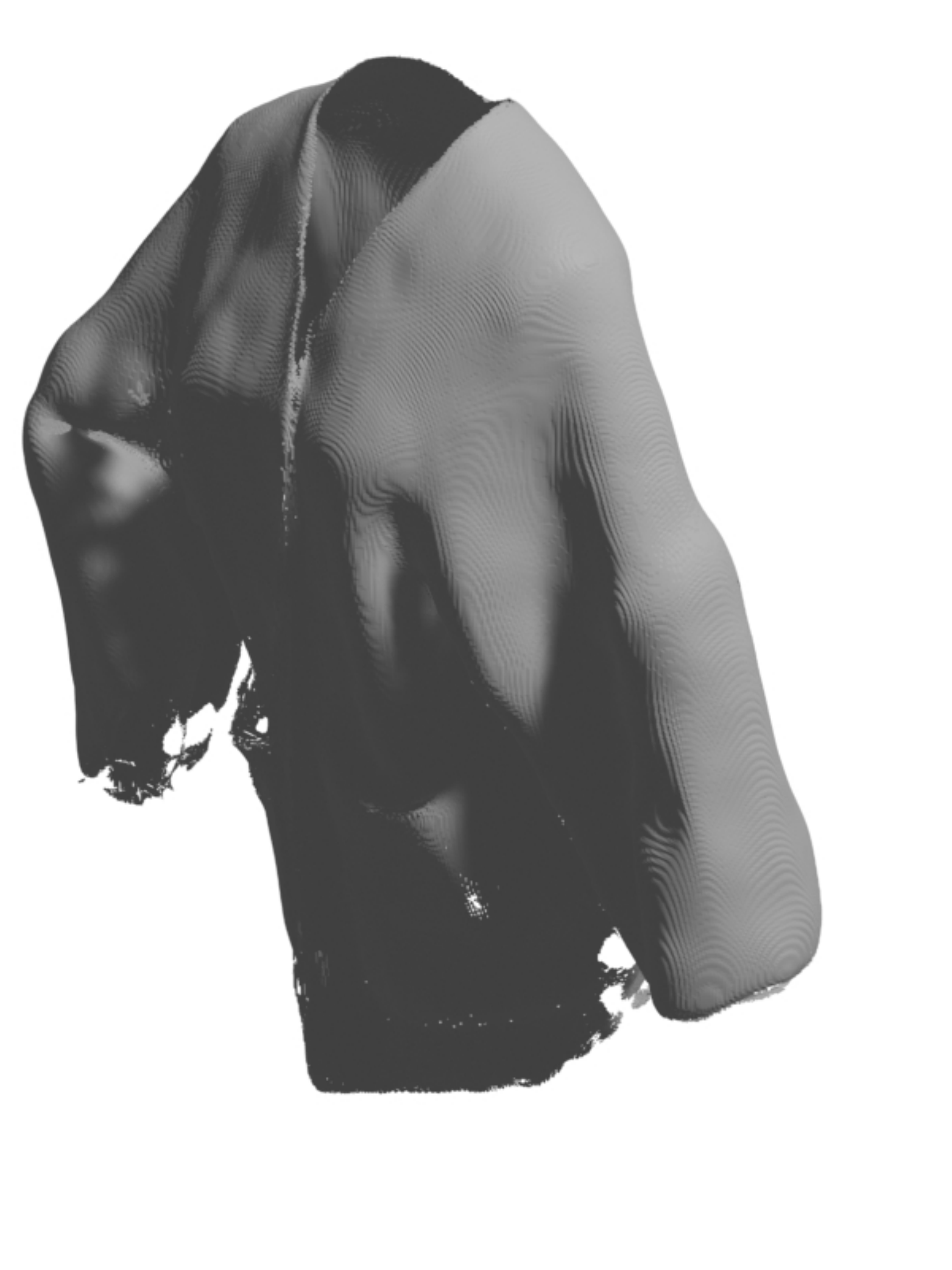}\\
    \includegraphics[width=.45\linewidth]{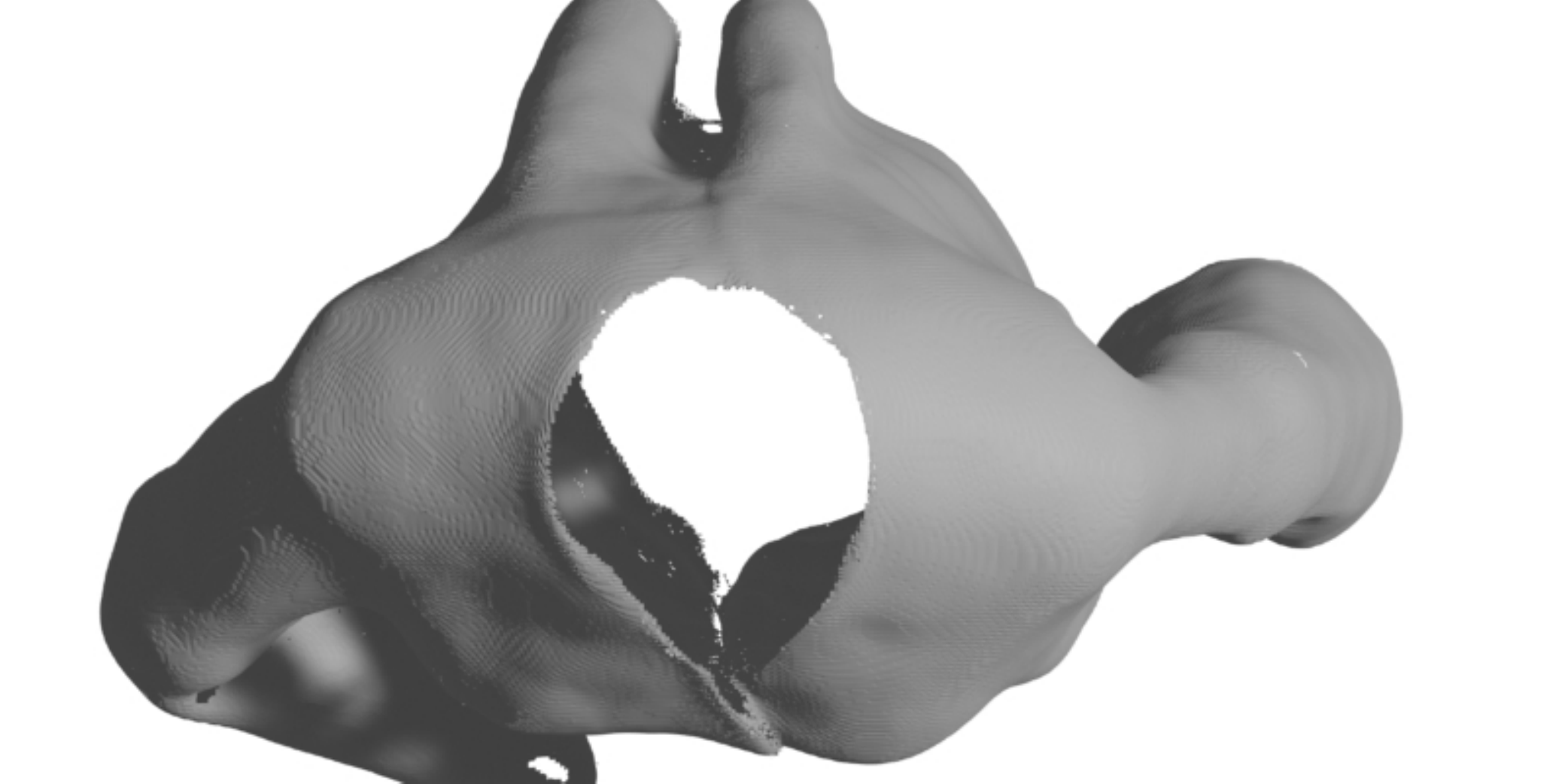}
\end{minipage}
\begin{minipage}[c]{.28\textwidth}
    \centering
    \includegraphics[width=.45\linewidth]{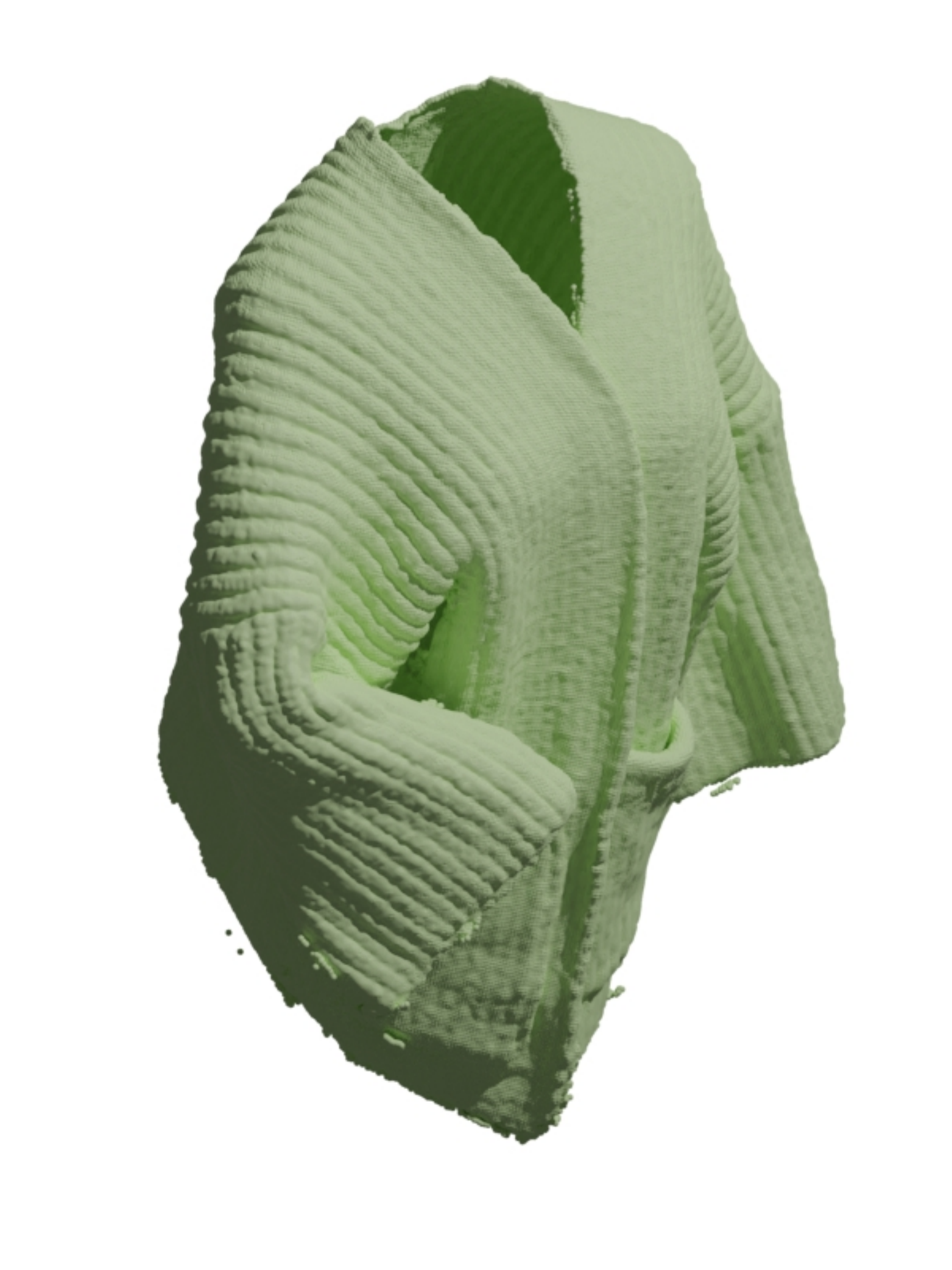}
    \includegraphics[width=.45\linewidth]{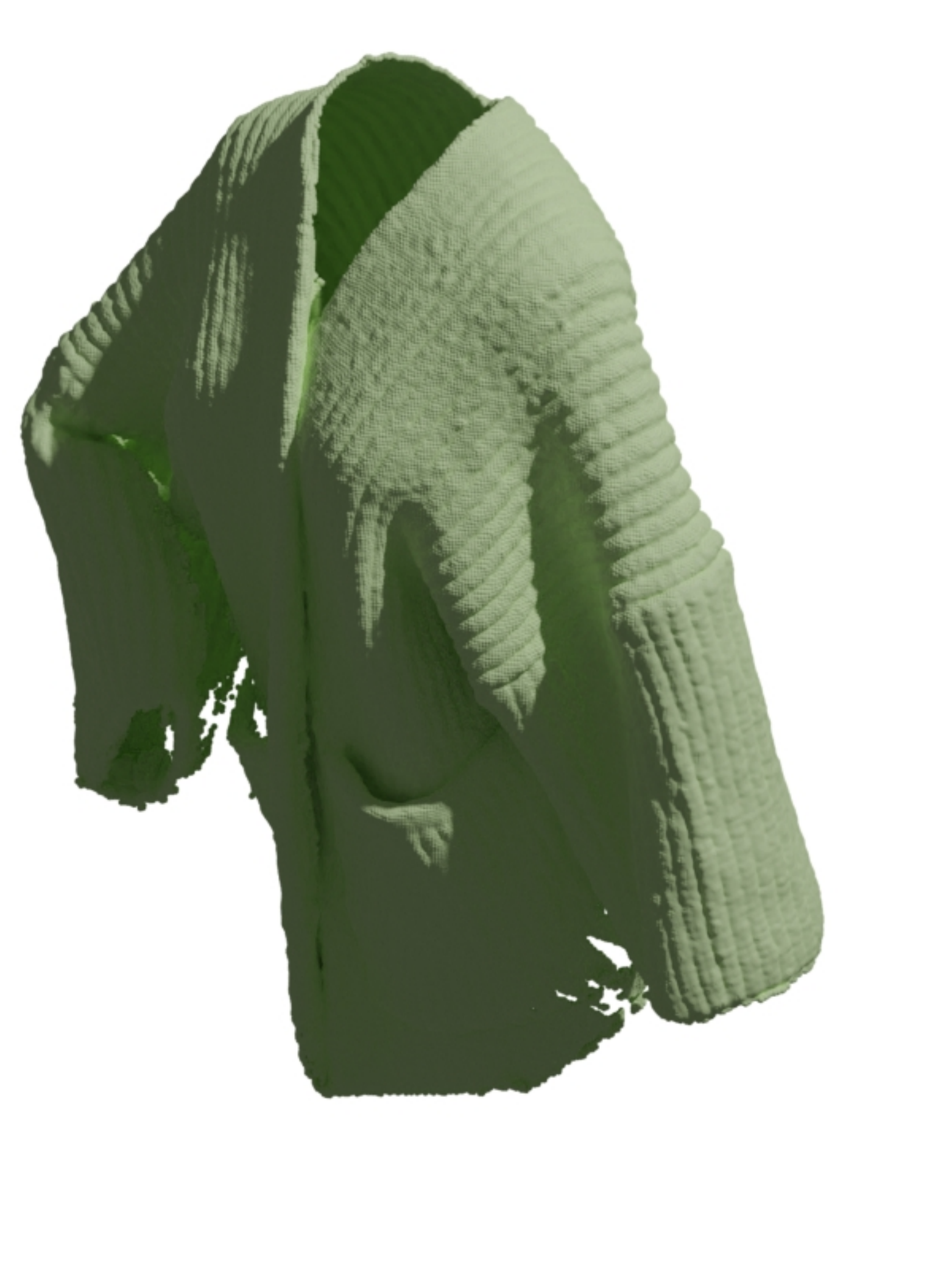}\\
    \includegraphics[width=.45\linewidth]{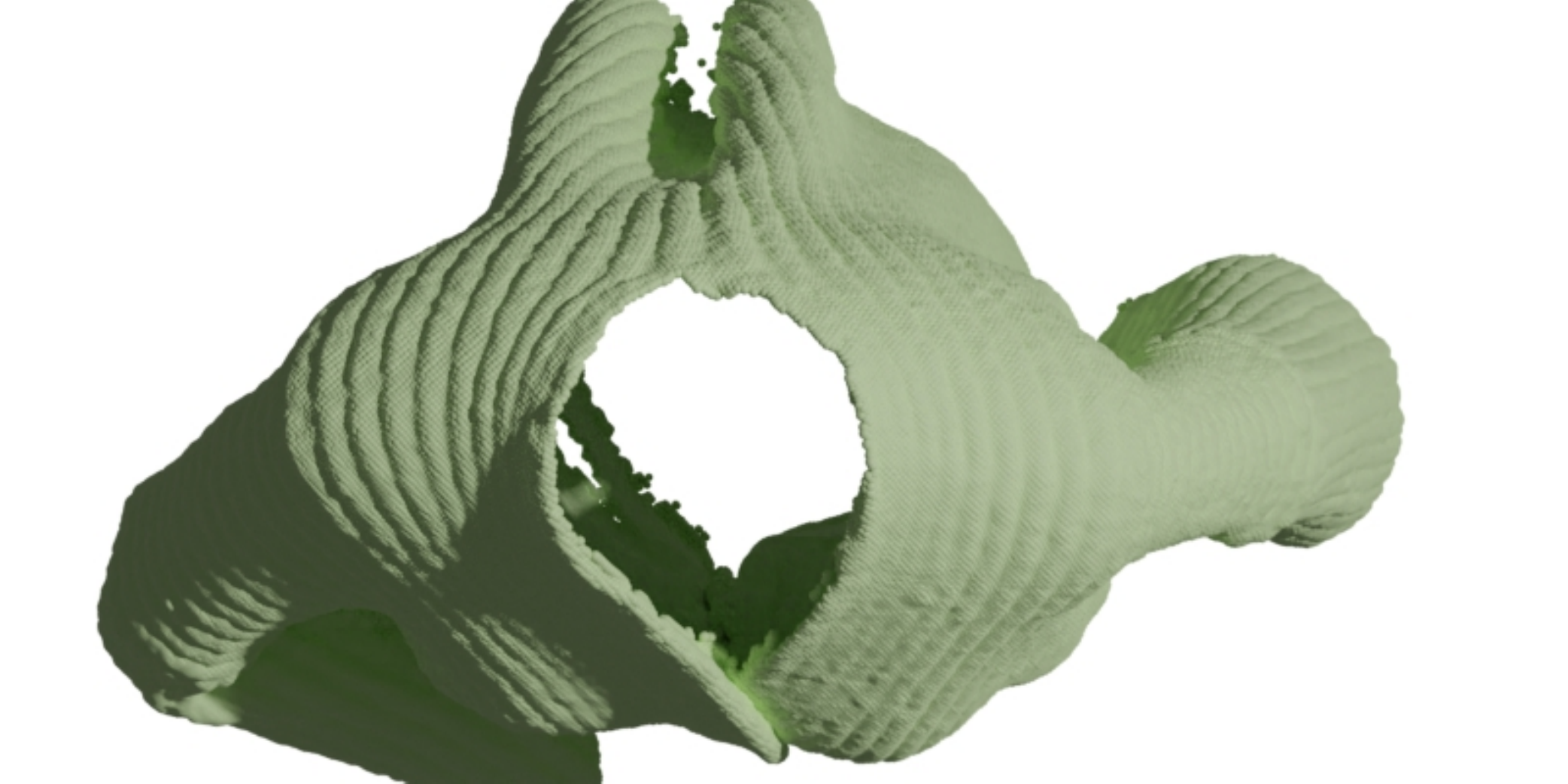}
\end{minipage}

\begin{minipage}[c]{.13\textwidth}
    \centering
    \includegraphics[width=1\linewidth]{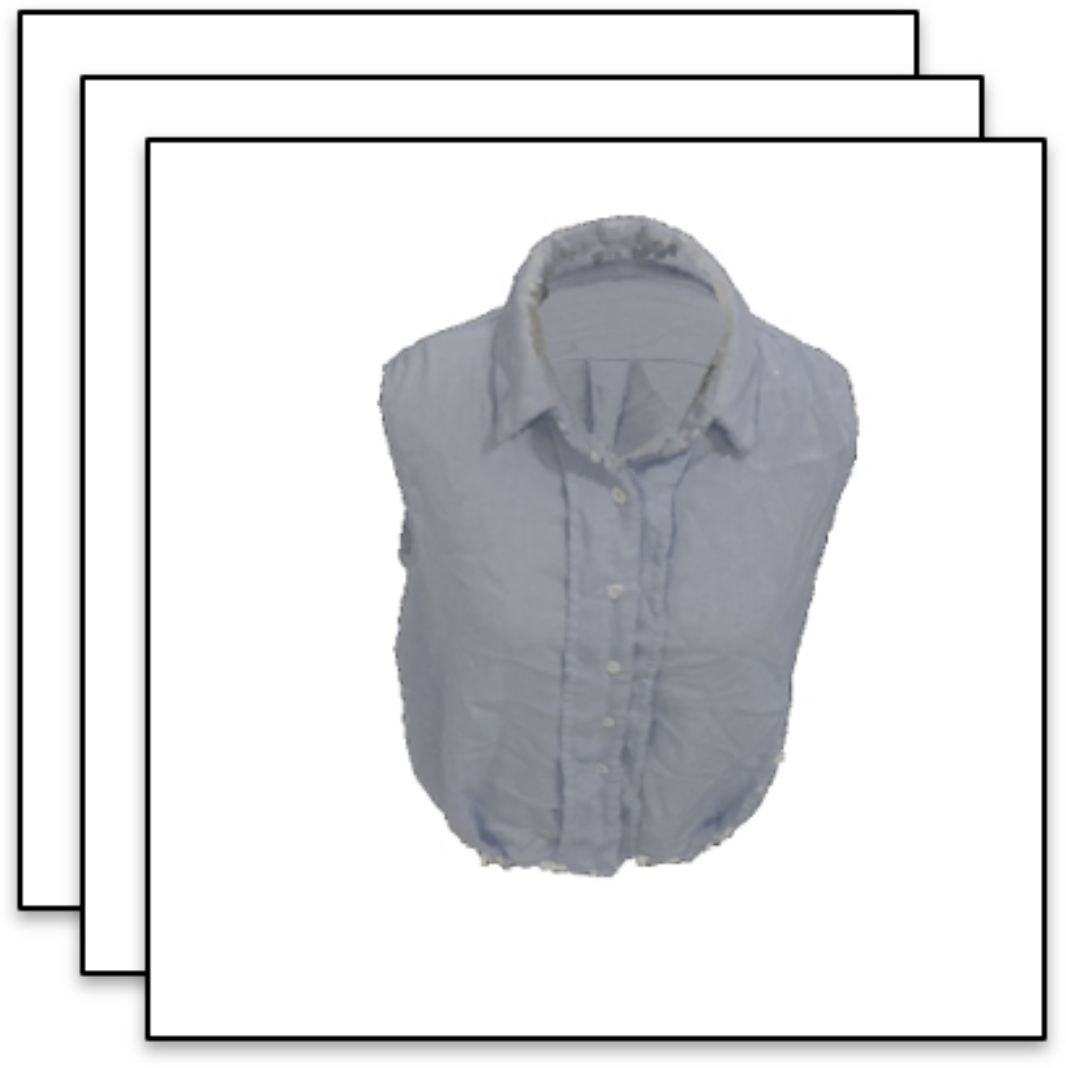}
\end{minipage}
\begin{minipage}[c]{.28\textwidth}
    \centering
    \includegraphics[width=.45\linewidth]{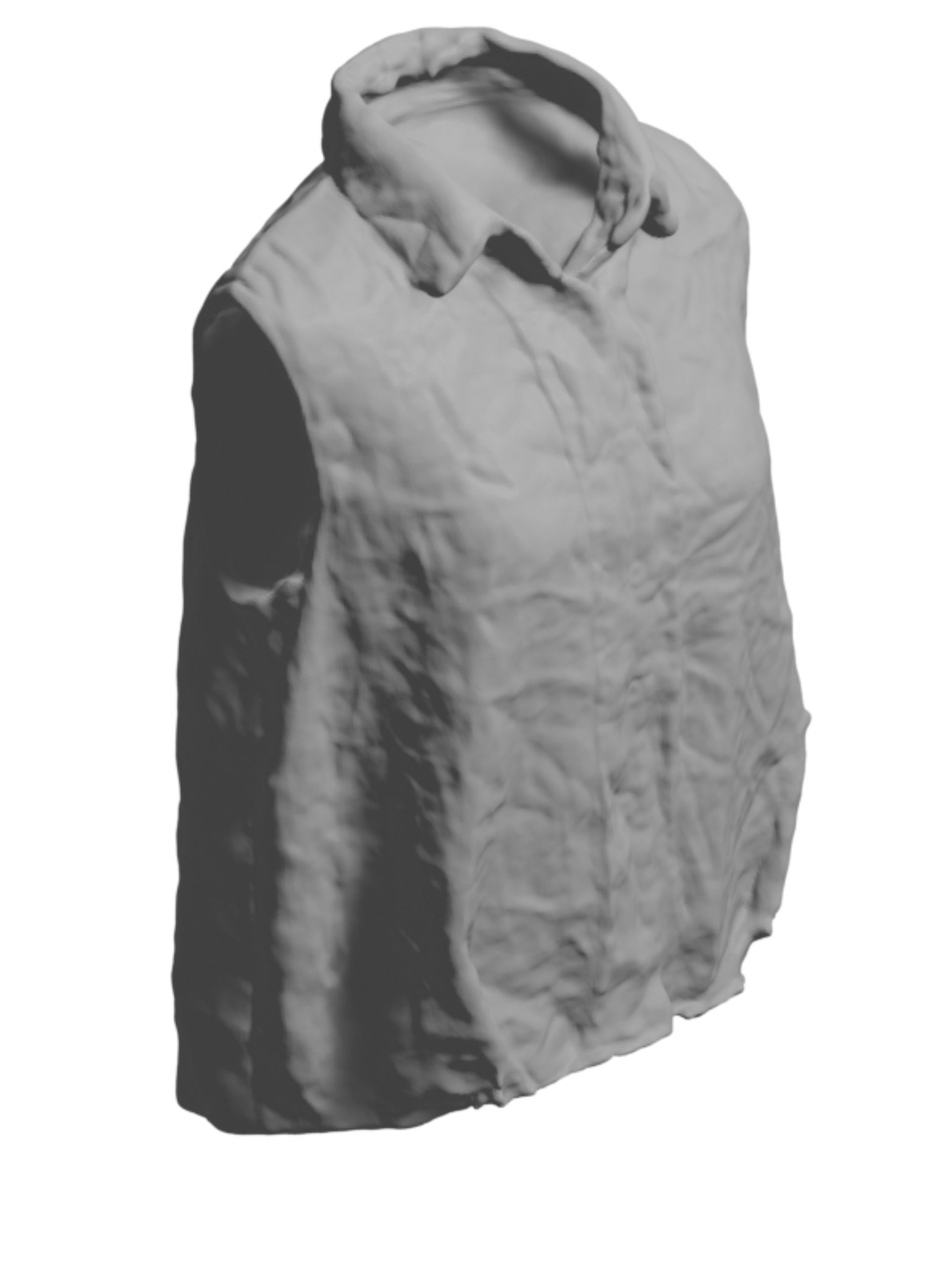}
    \includegraphics[width=.45\linewidth]{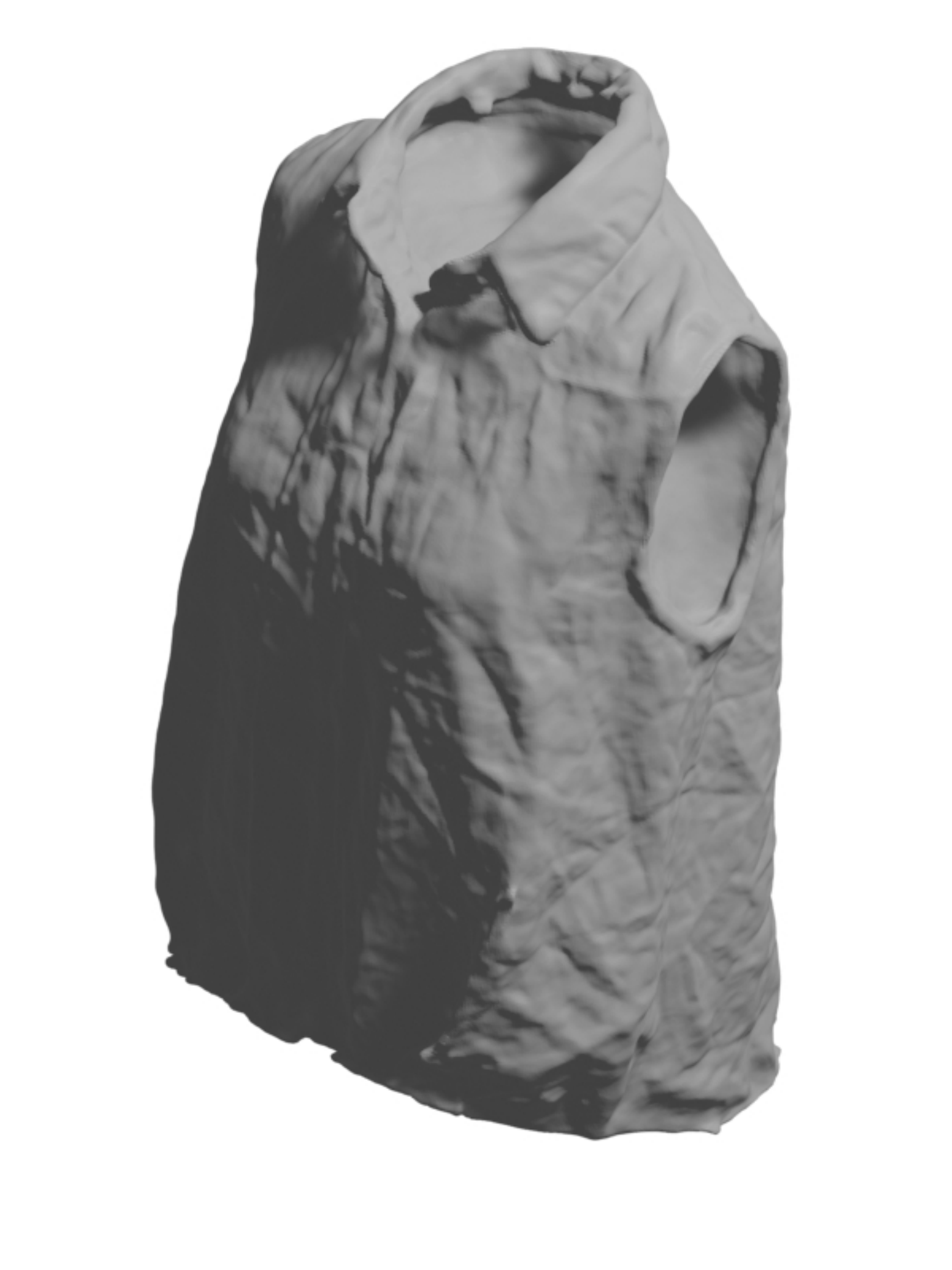}\\
    \includegraphics[width=.45\linewidth]{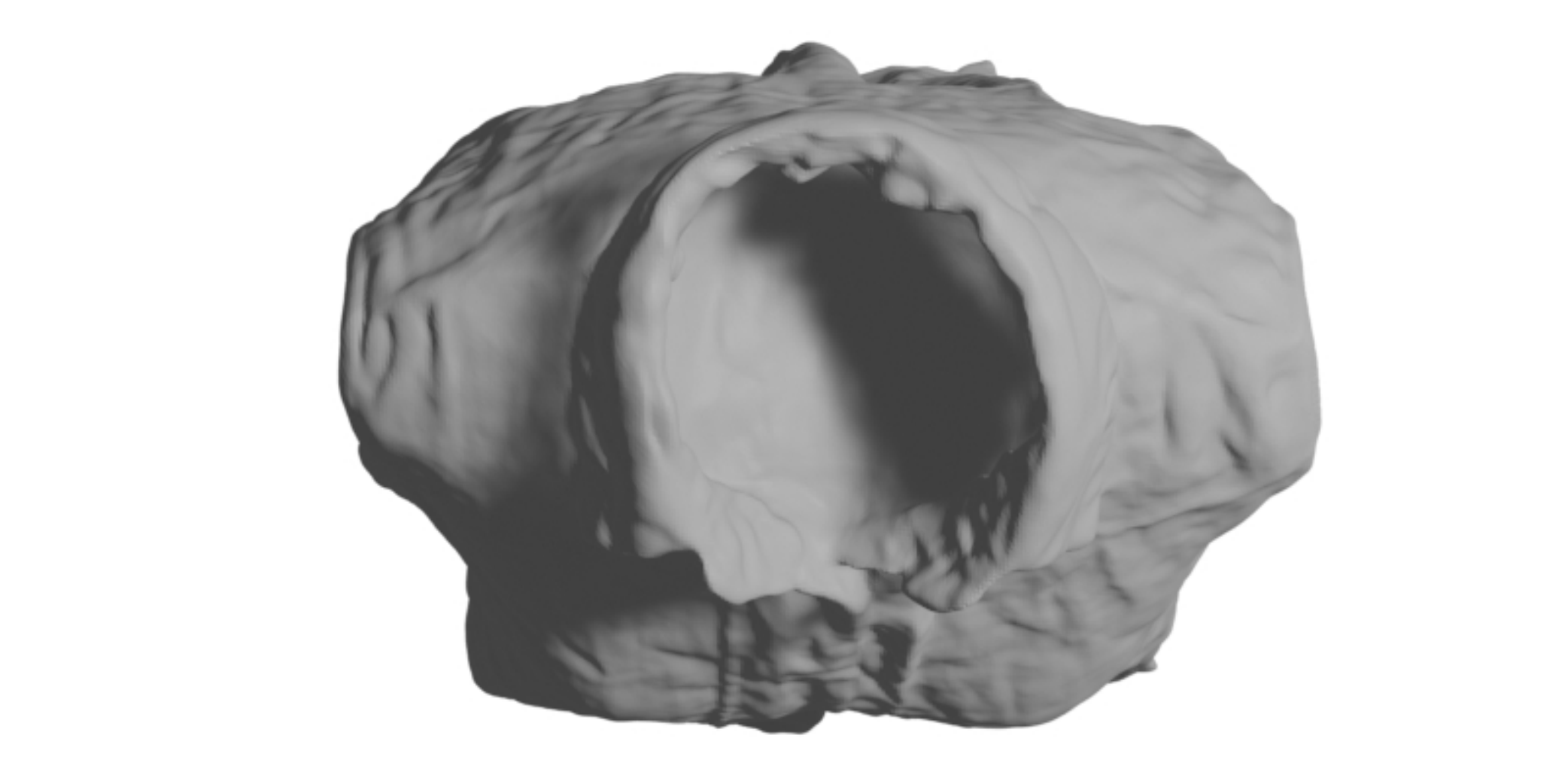}
\end{minipage}
\begin{minipage}[c]{.28\textwidth}
    \centering
    \includegraphics[width=.45\linewidth]{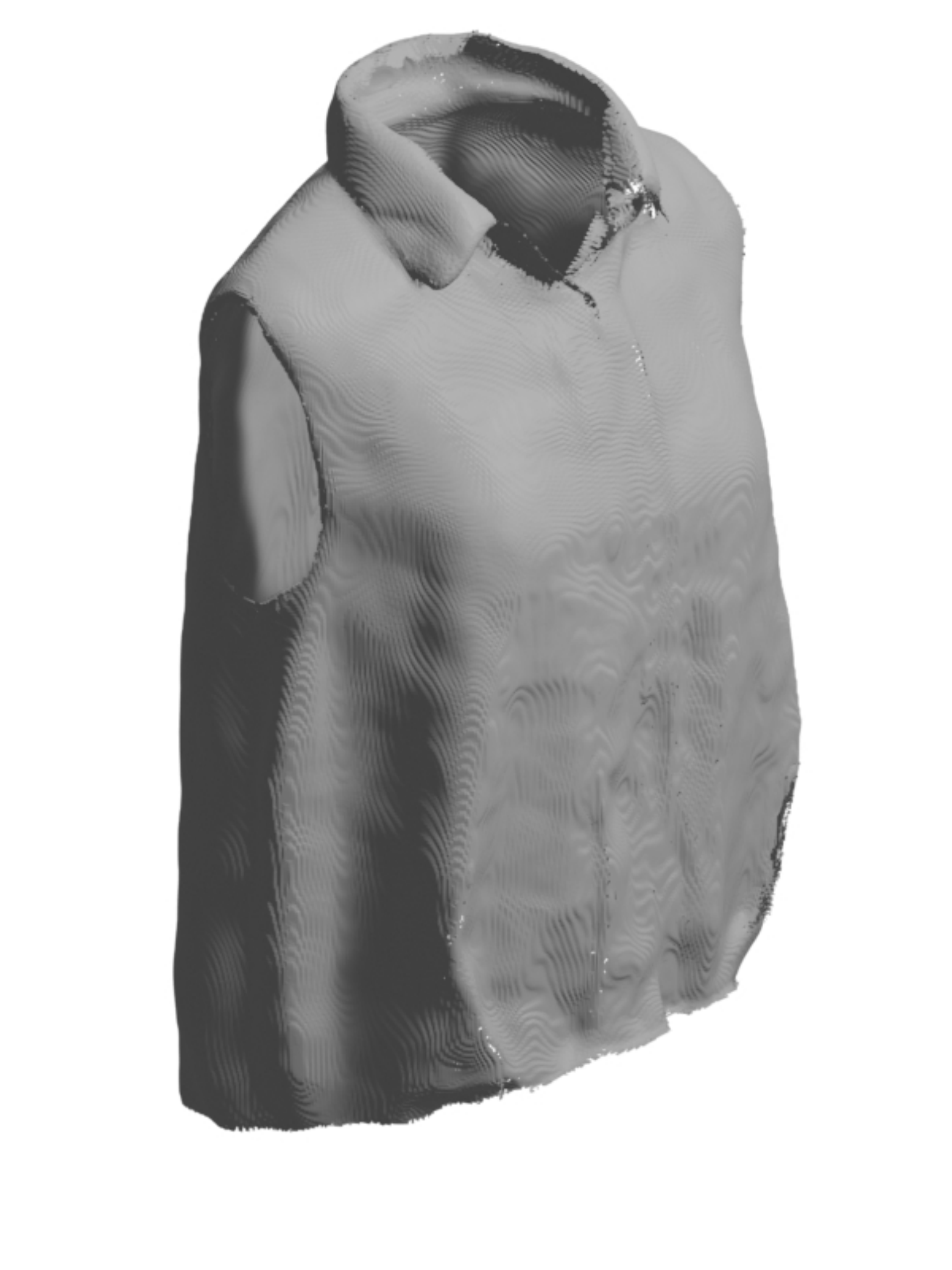}
    \includegraphics[width=.45\linewidth]{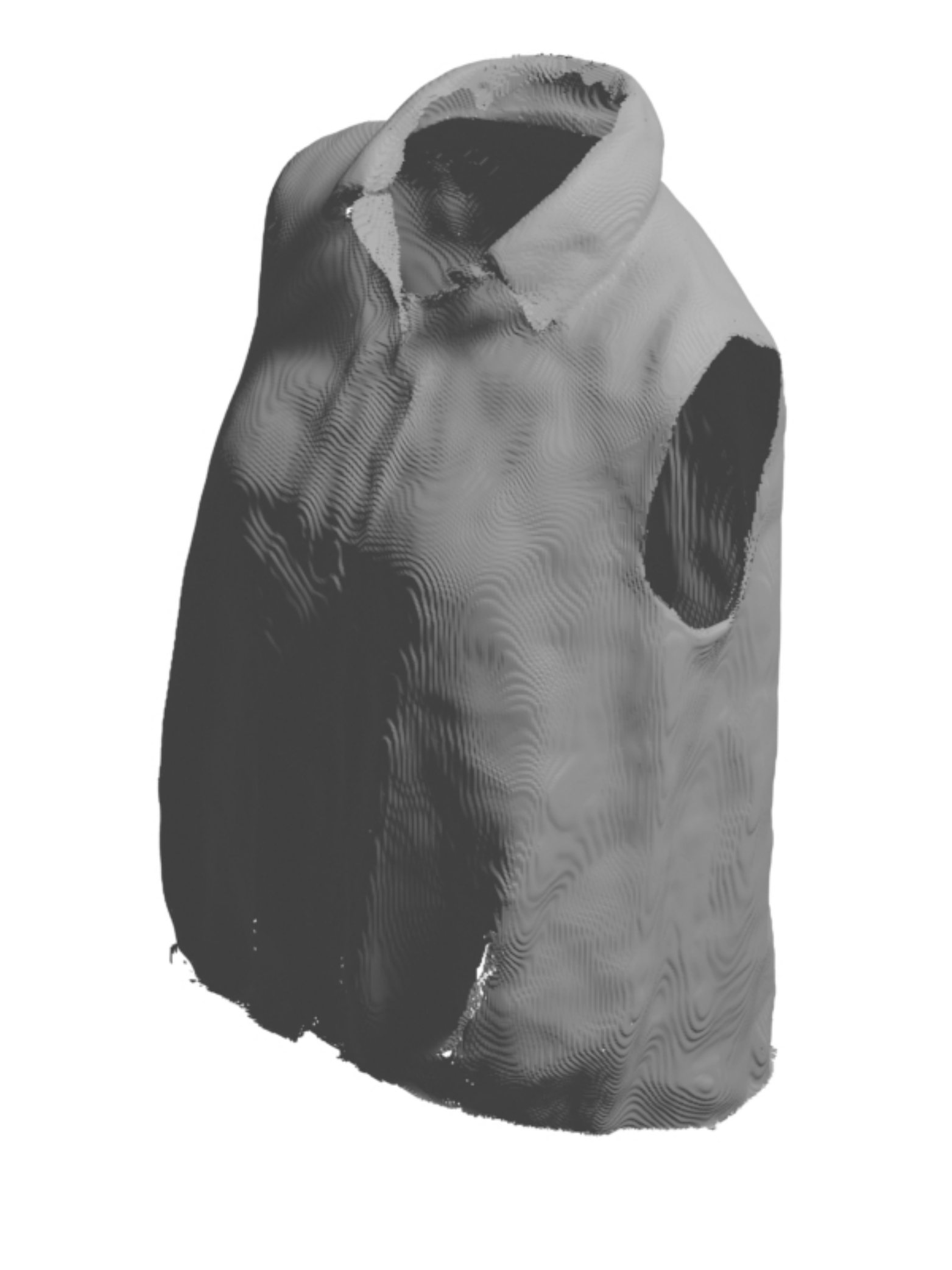}\\
    \includegraphics[width=.45\linewidth]{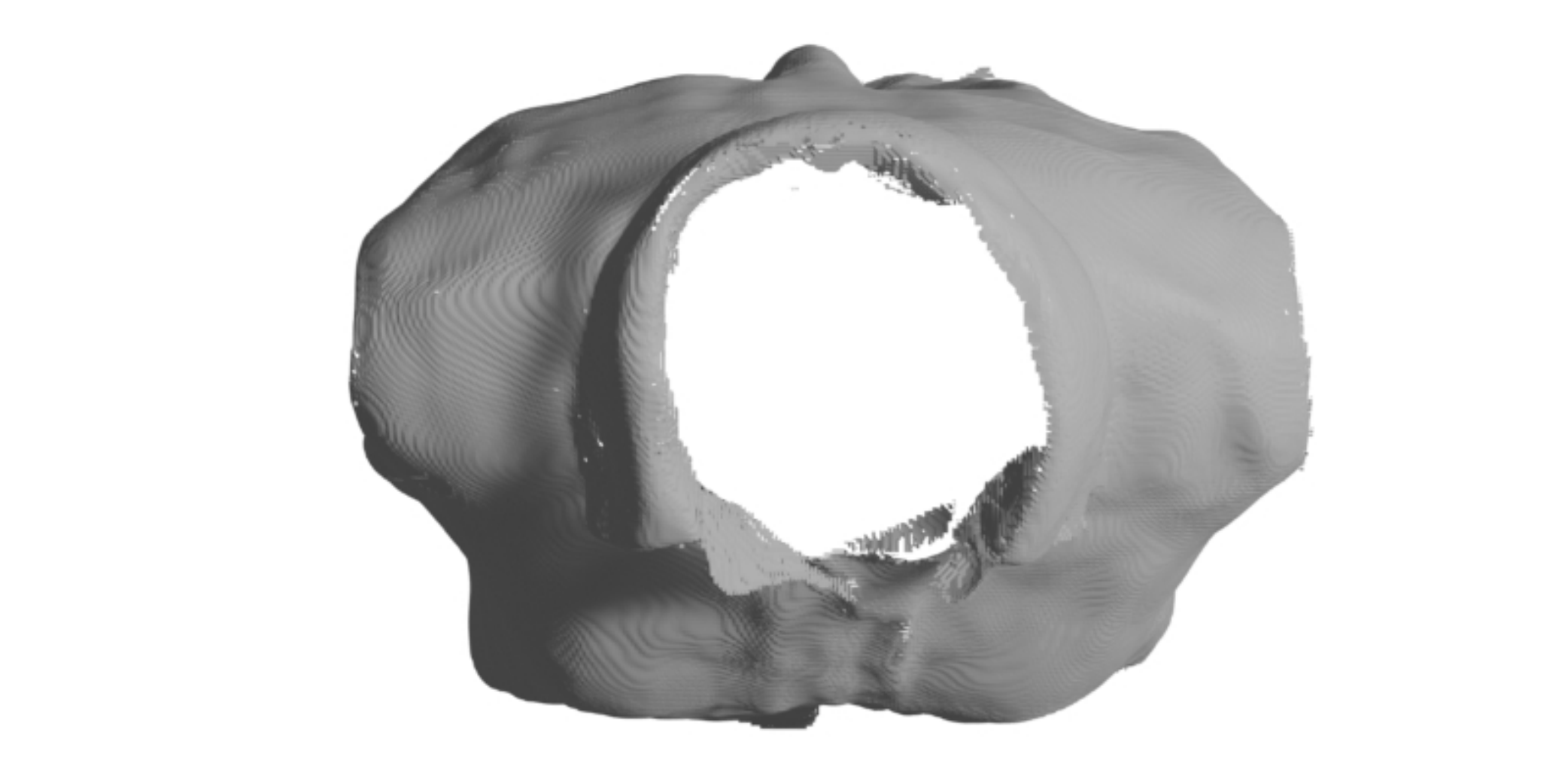}
\end{minipage}
\begin{minipage}[c]{.28\textwidth}
    \centering
    \includegraphics[width=.45\linewidth]{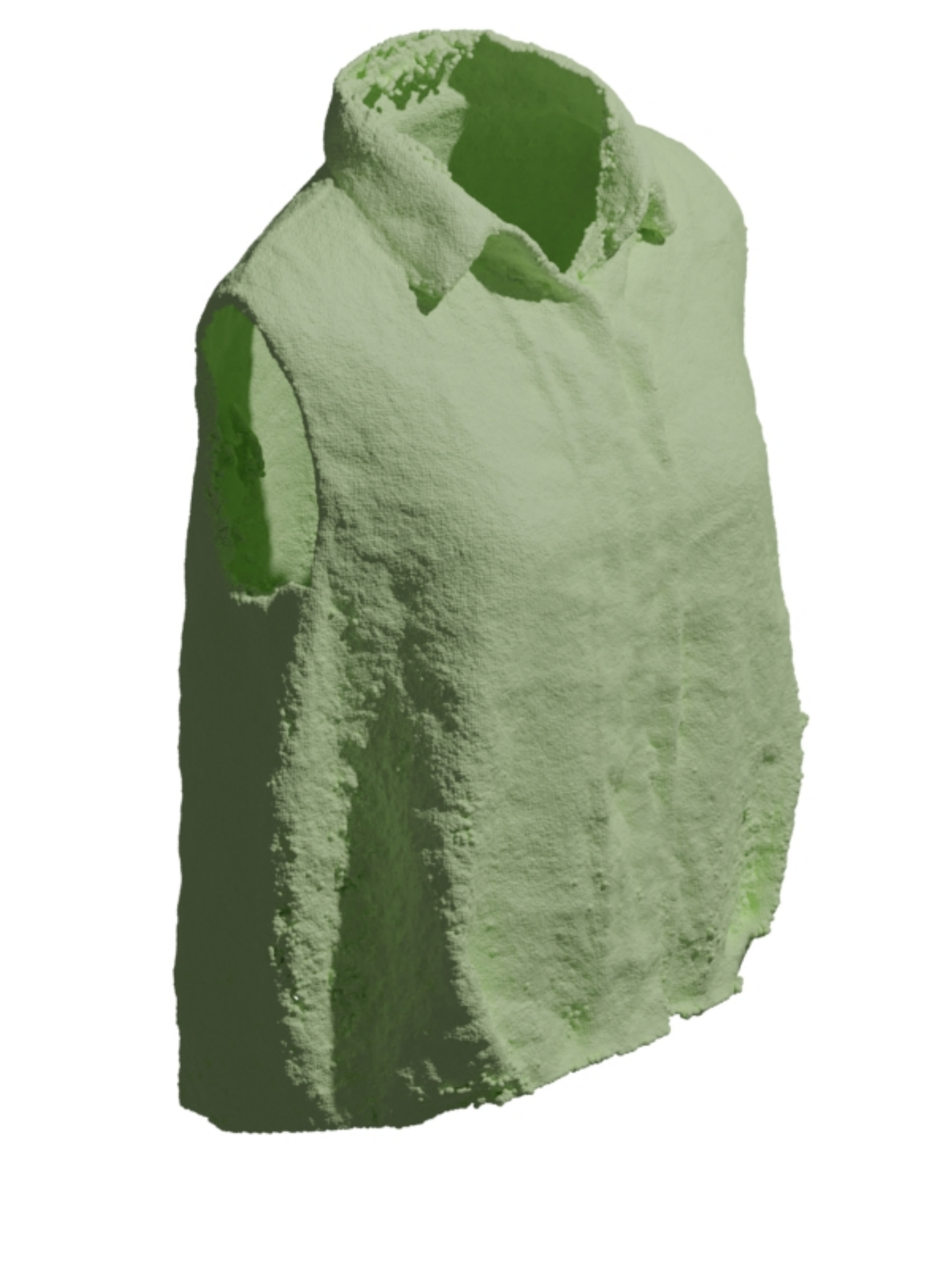}
    \includegraphics[width=.45\linewidth]{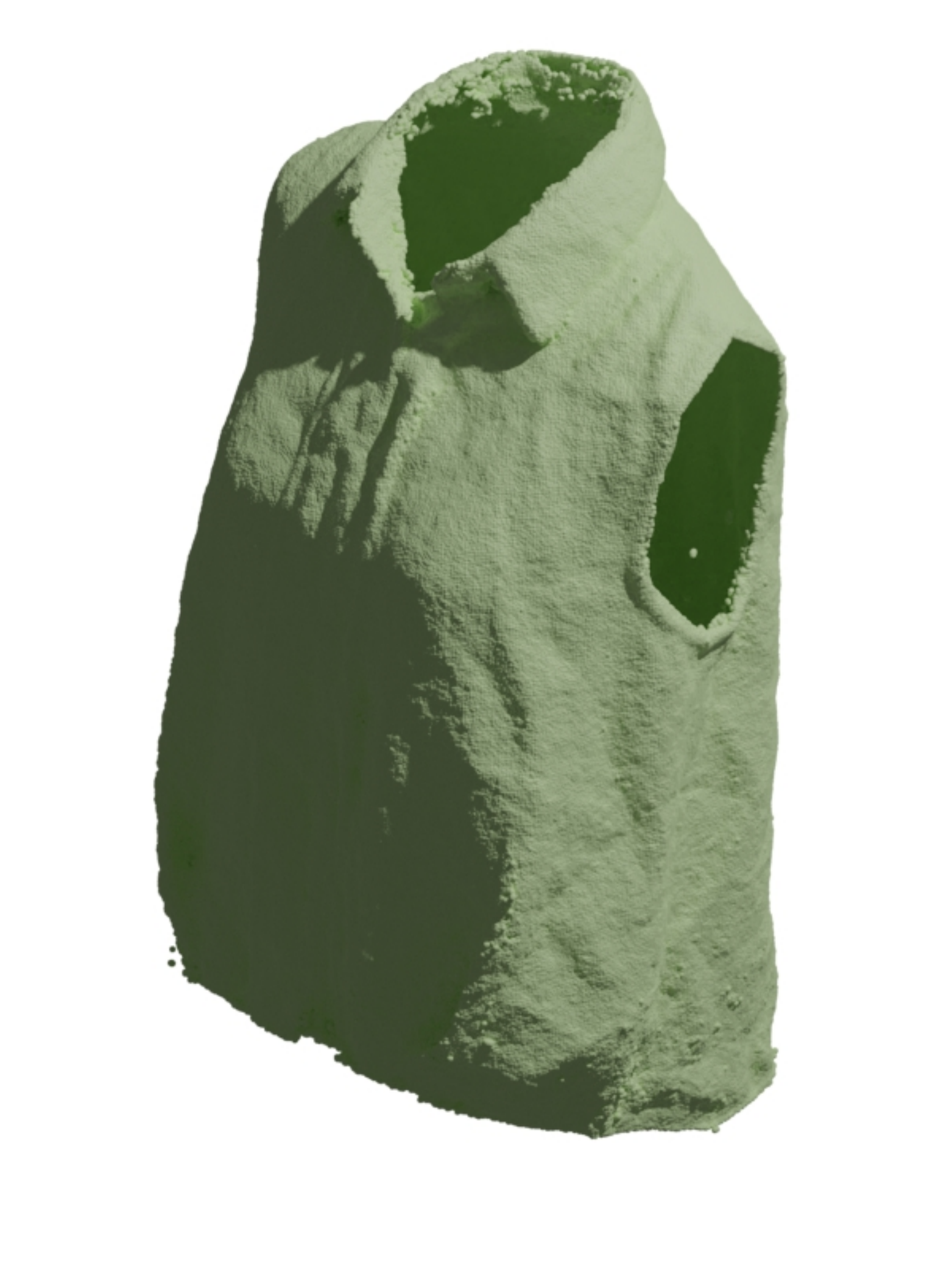}\\
    \includegraphics[width=.45\linewidth]{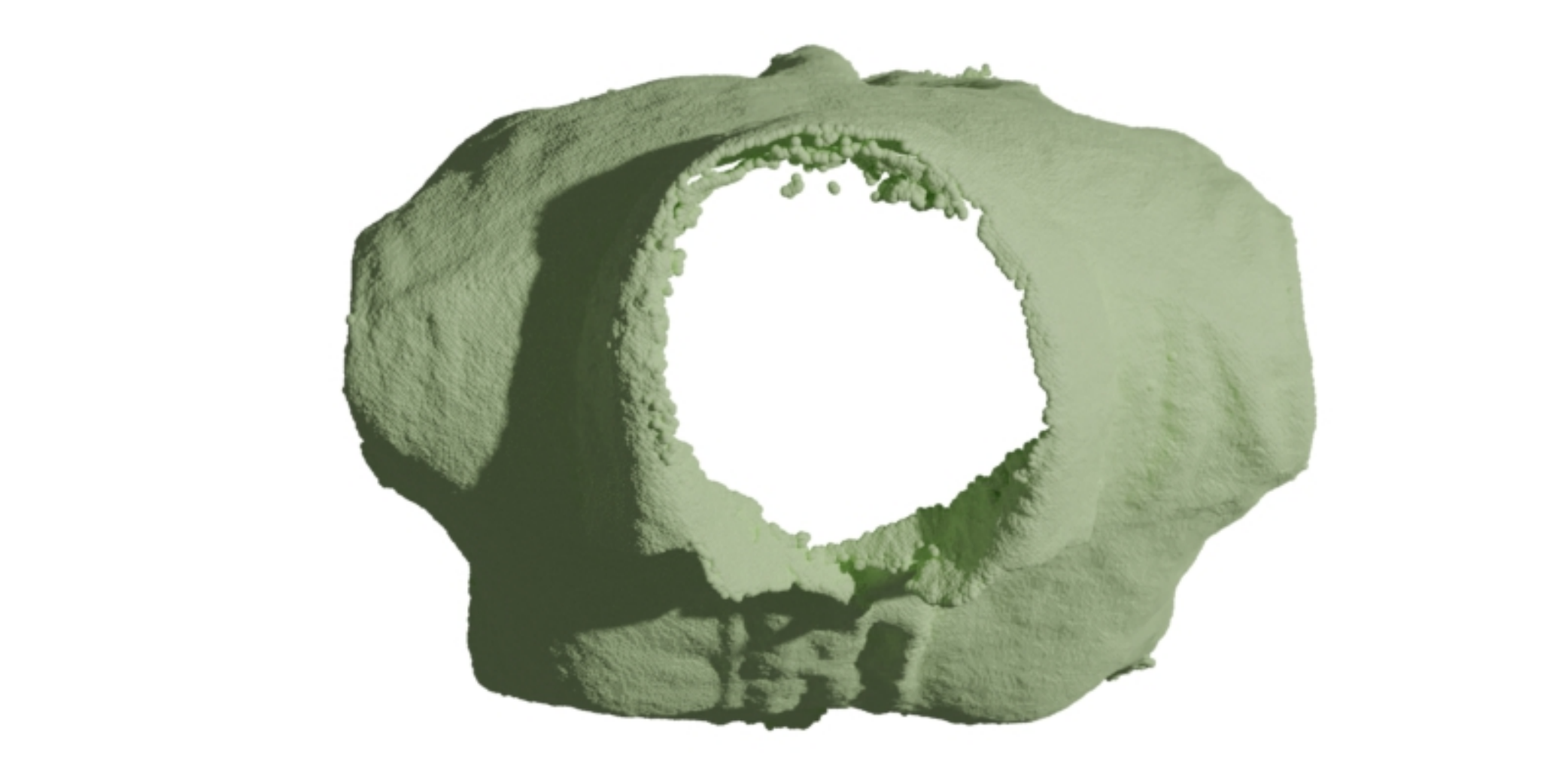}
\end{minipage}

\begin{minipage}[c]{.13\textwidth}
    \centering
    \includegraphics[width=1\linewidth]{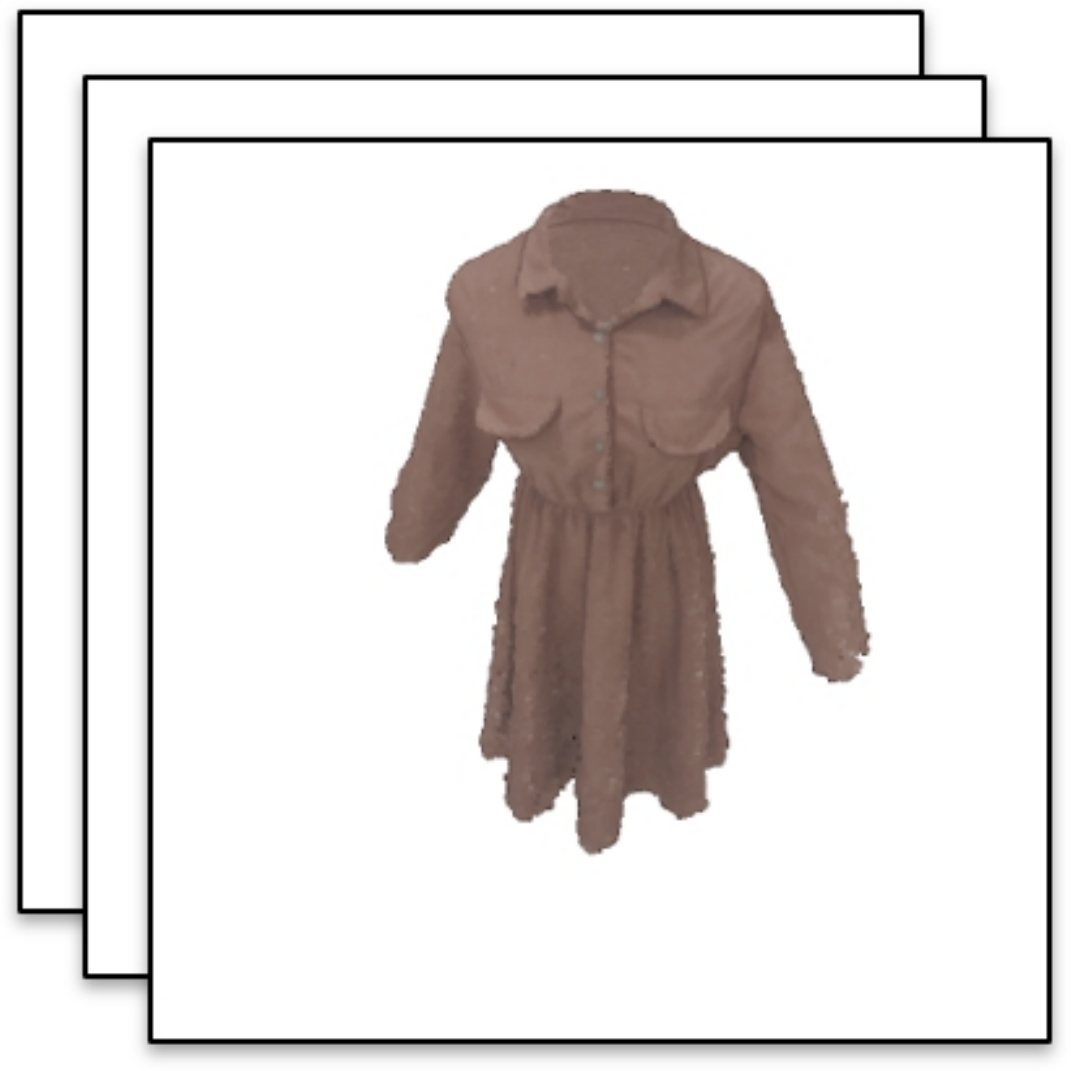}
\end{minipage}
\begin{minipage}[c]{.28\textwidth}
    \centering
    \includegraphics[width=.45\linewidth]{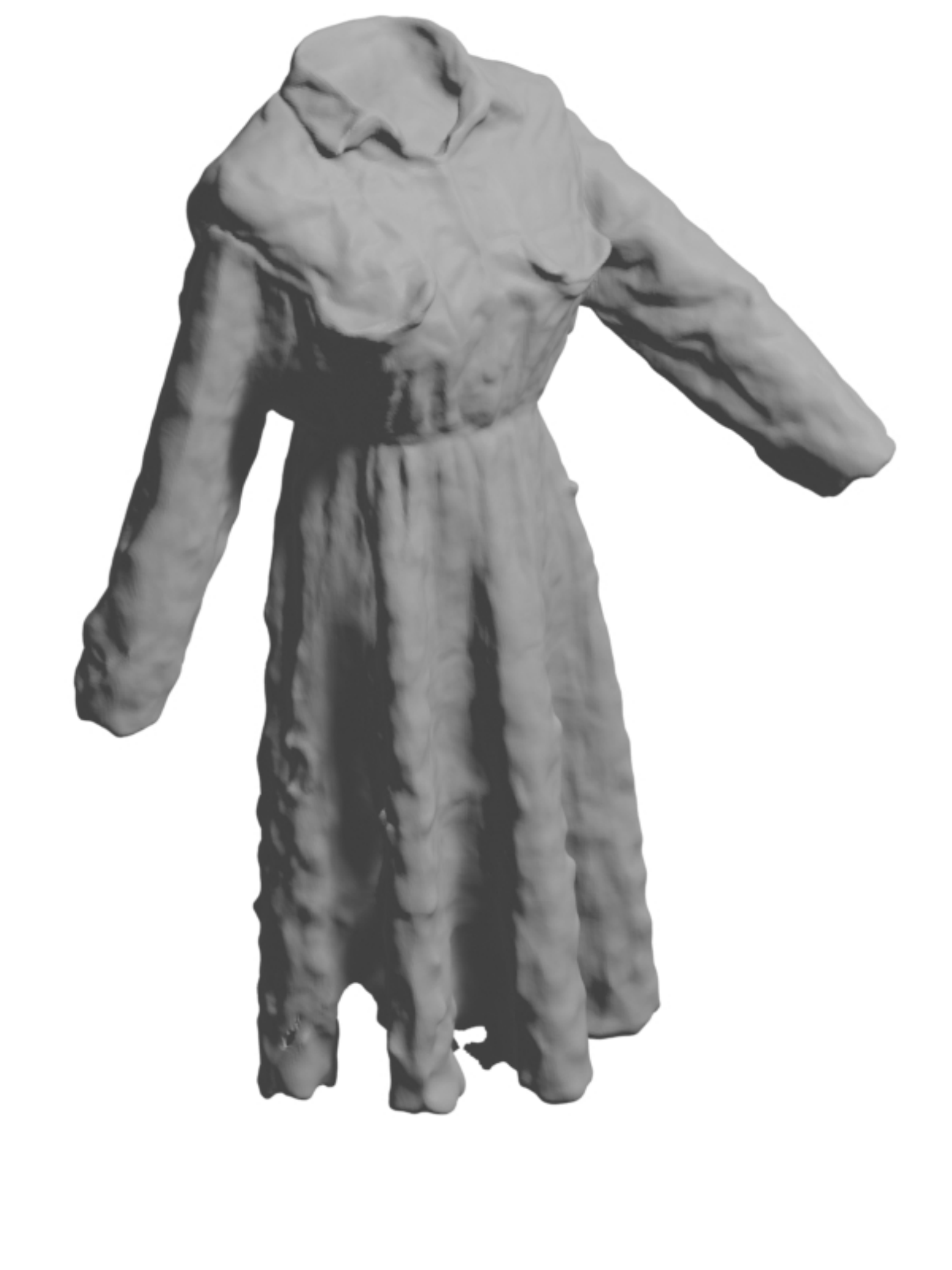}
    \includegraphics[width=.45\linewidth]{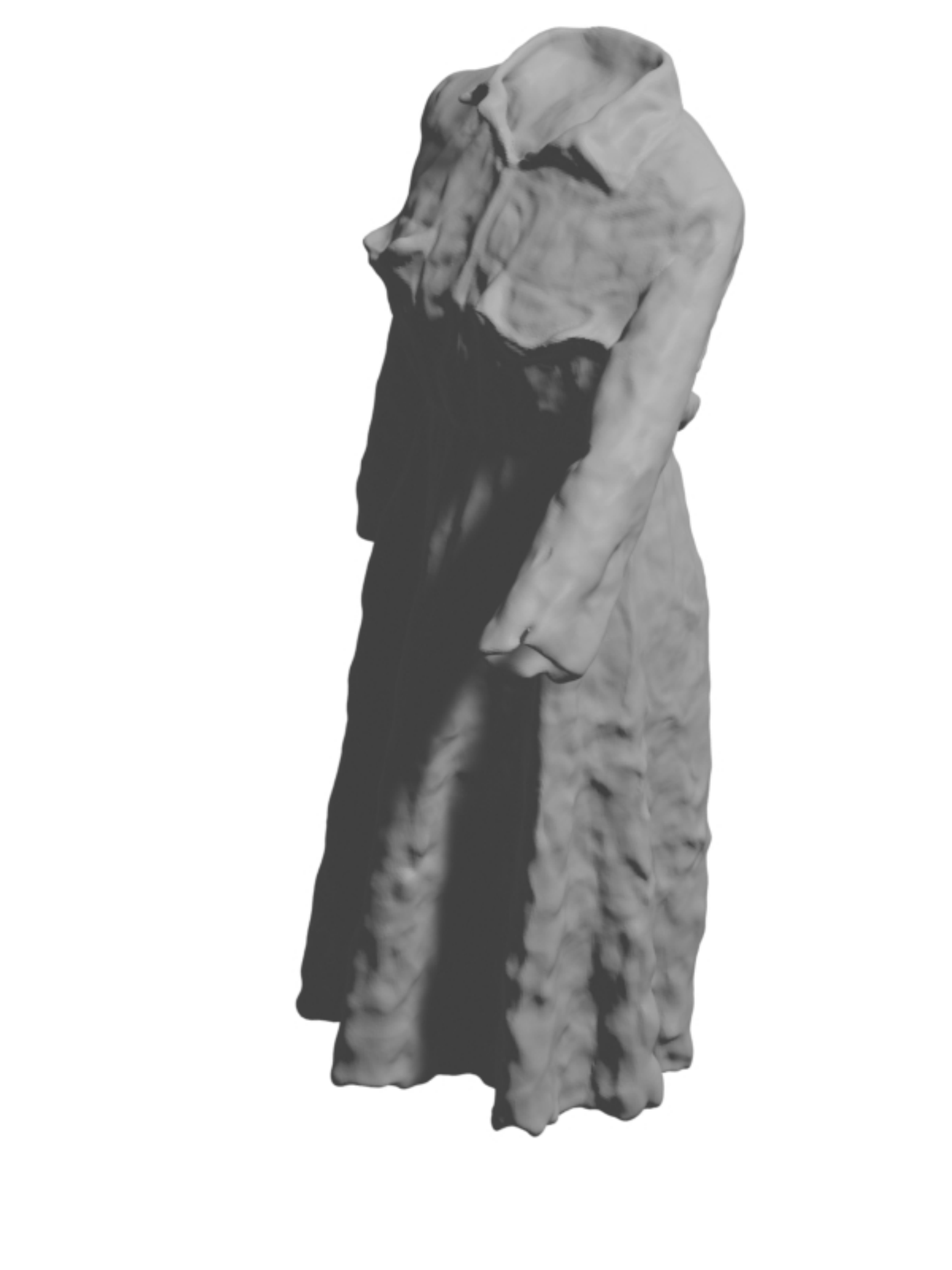}\\
    \includegraphics[width=.45\linewidth]{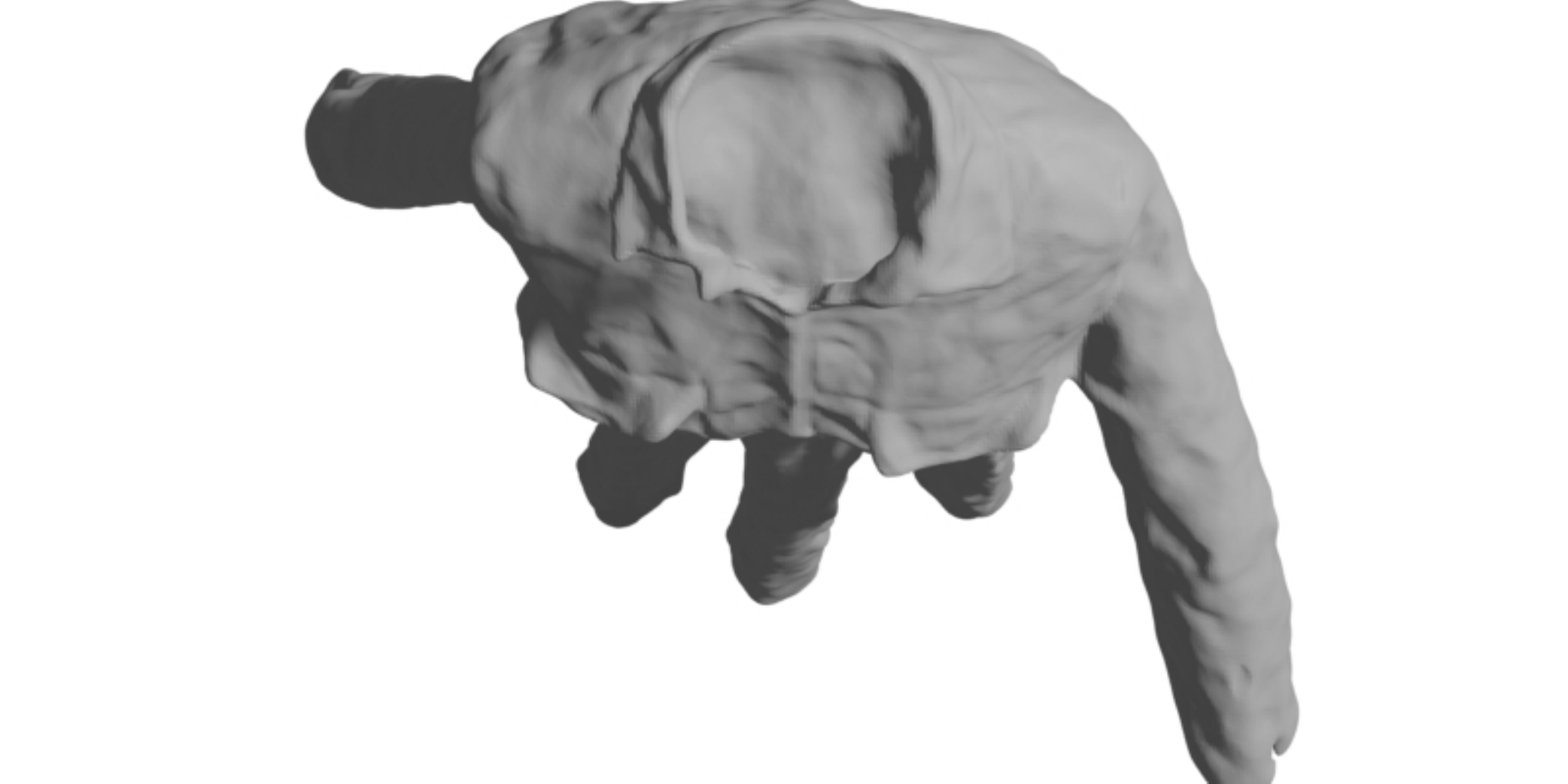}
\end{minipage}
\begin{minipage}[c]{.28\textwidth}
    \centering
    \includegraphics[width=.45\linewidth]{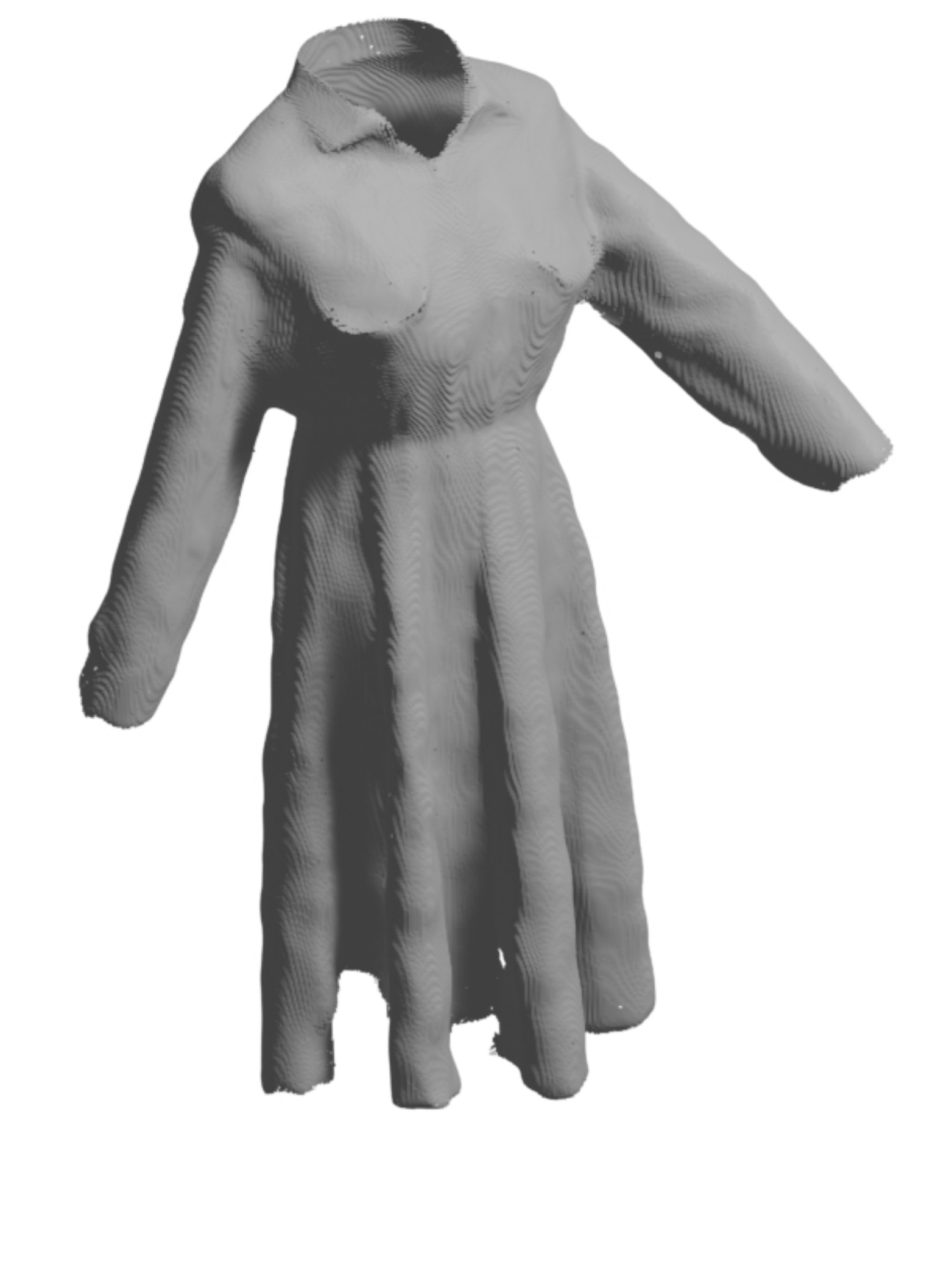}
    \includegraphics[width=.45\linewidth]{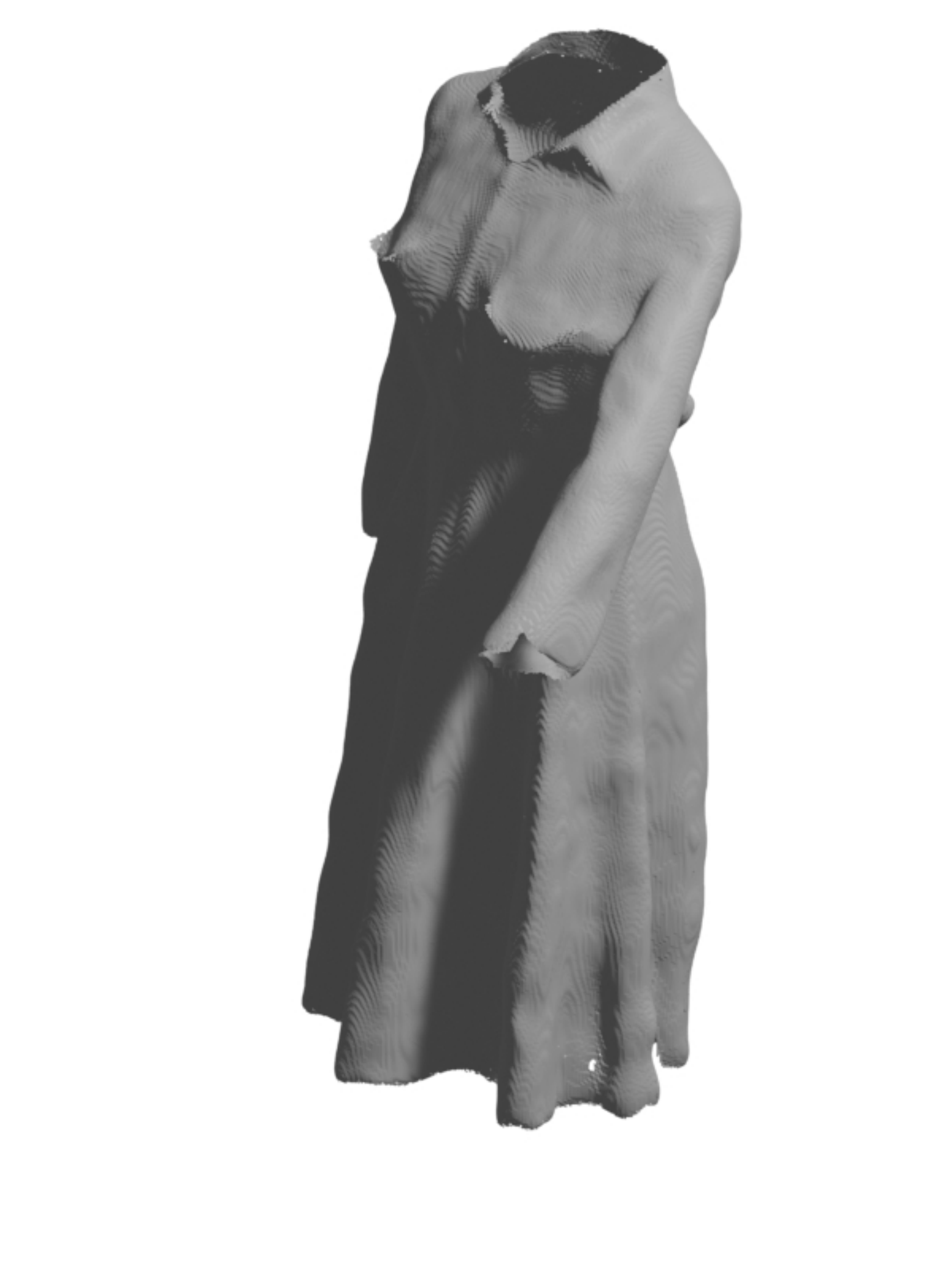}\\
    \includegraphics[width=.45\linewidth]{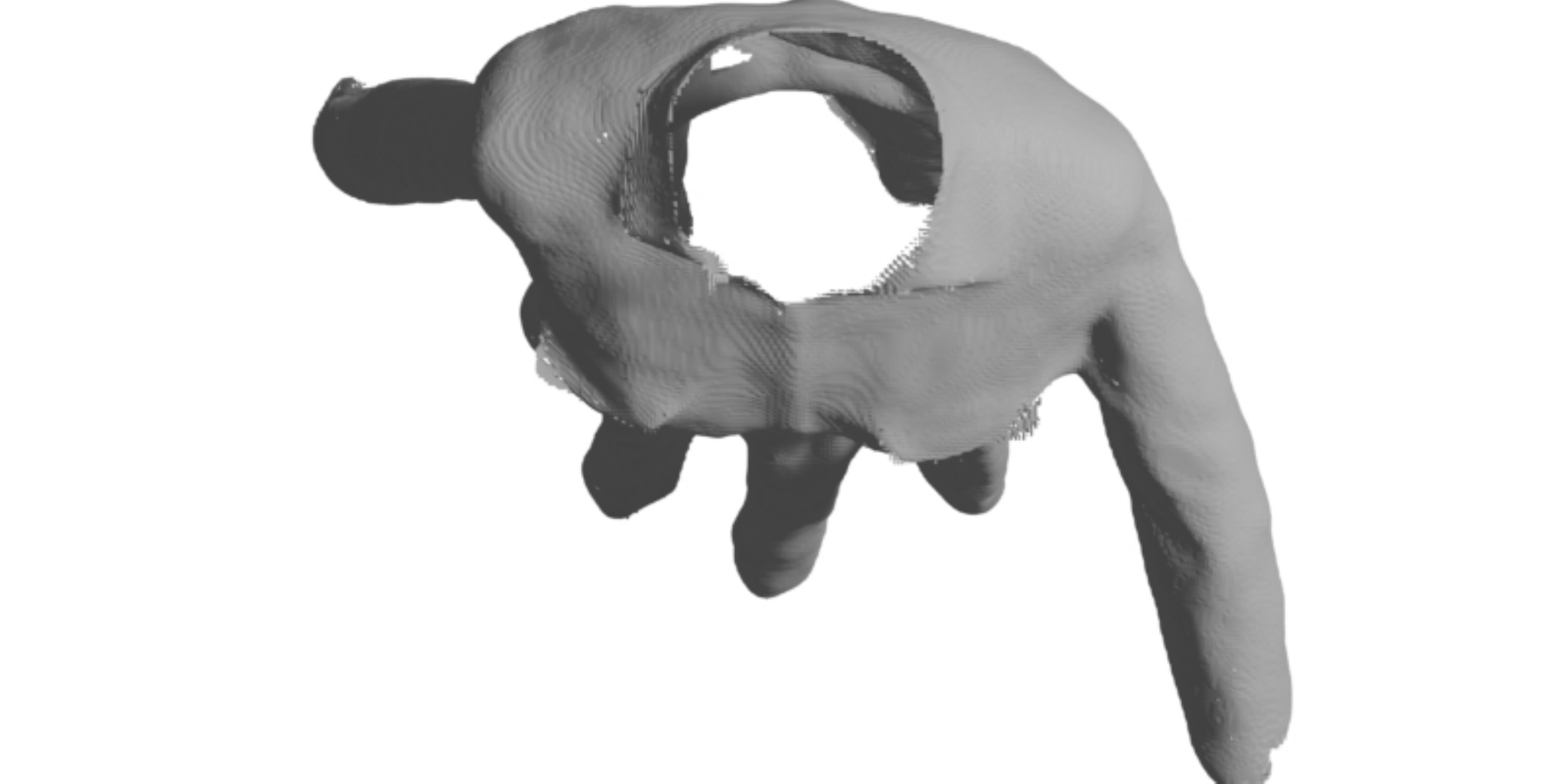}
\end{minipage}
\begin{minipage}[c]{.28\textwidth}
    \centering
    \includegraphics[width=.45\linewidth]{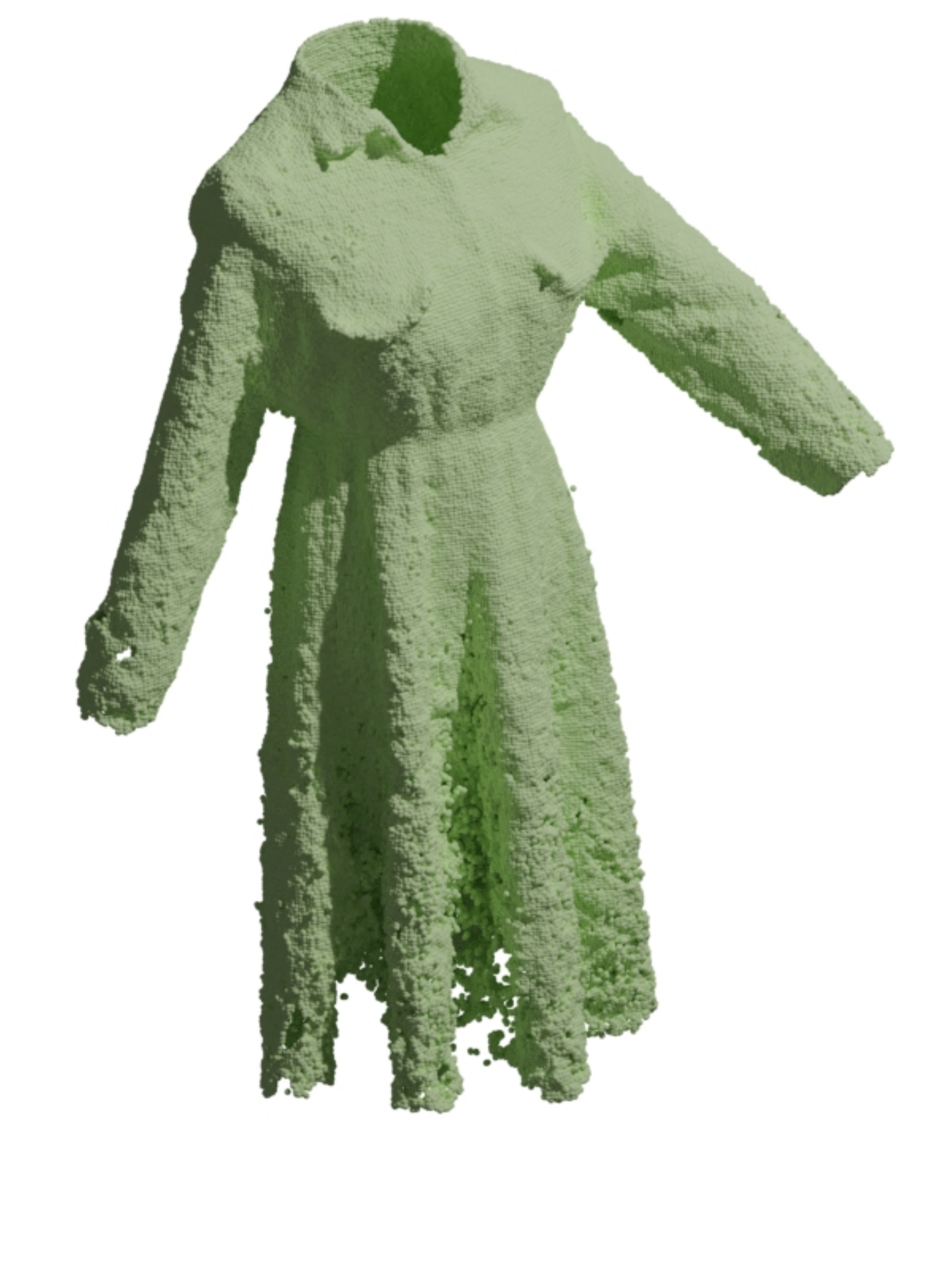}
    \includegraphics[width=.45\linewidth]{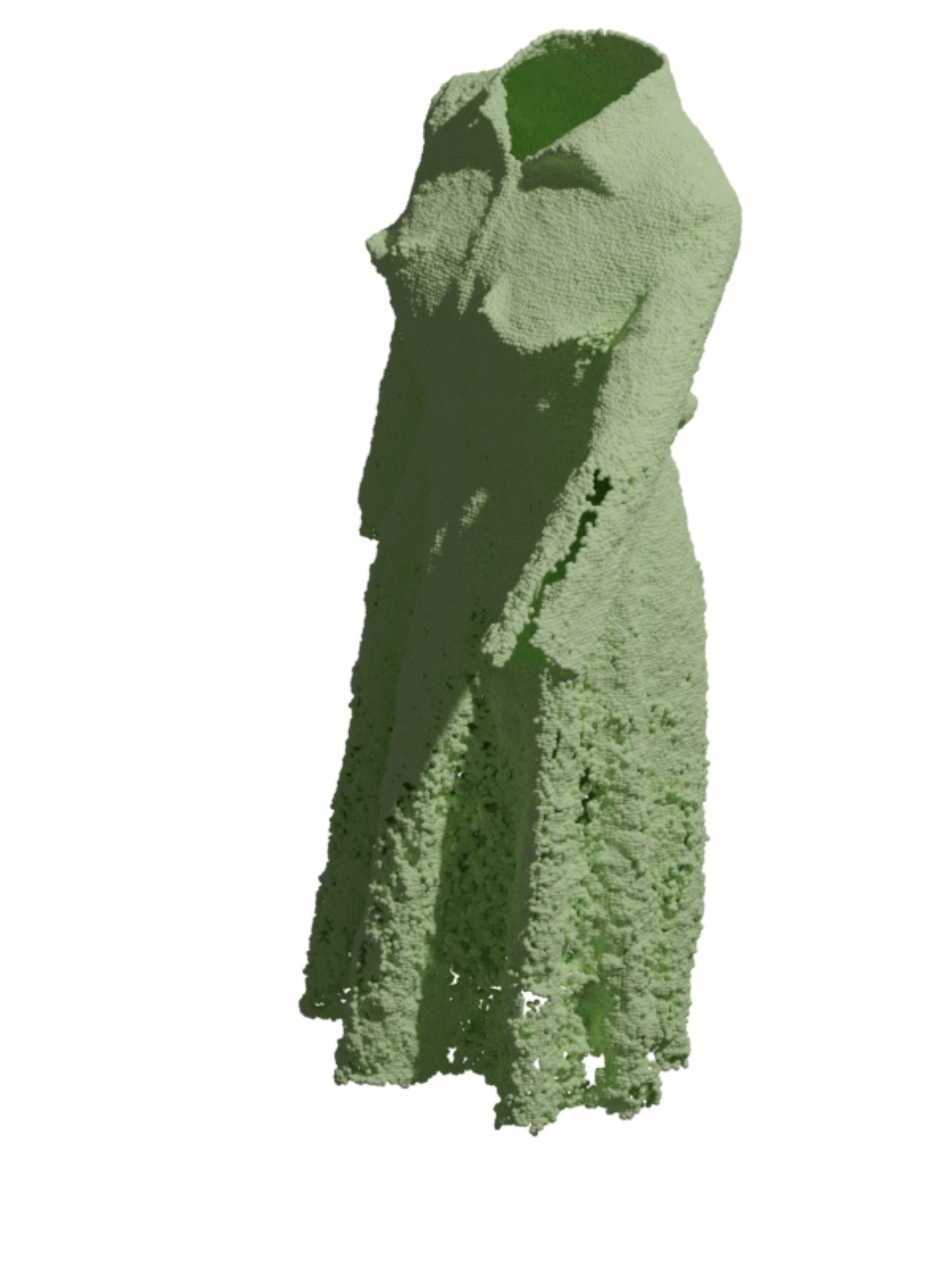}\\
    \includegraphics[width=.45\linewidth]{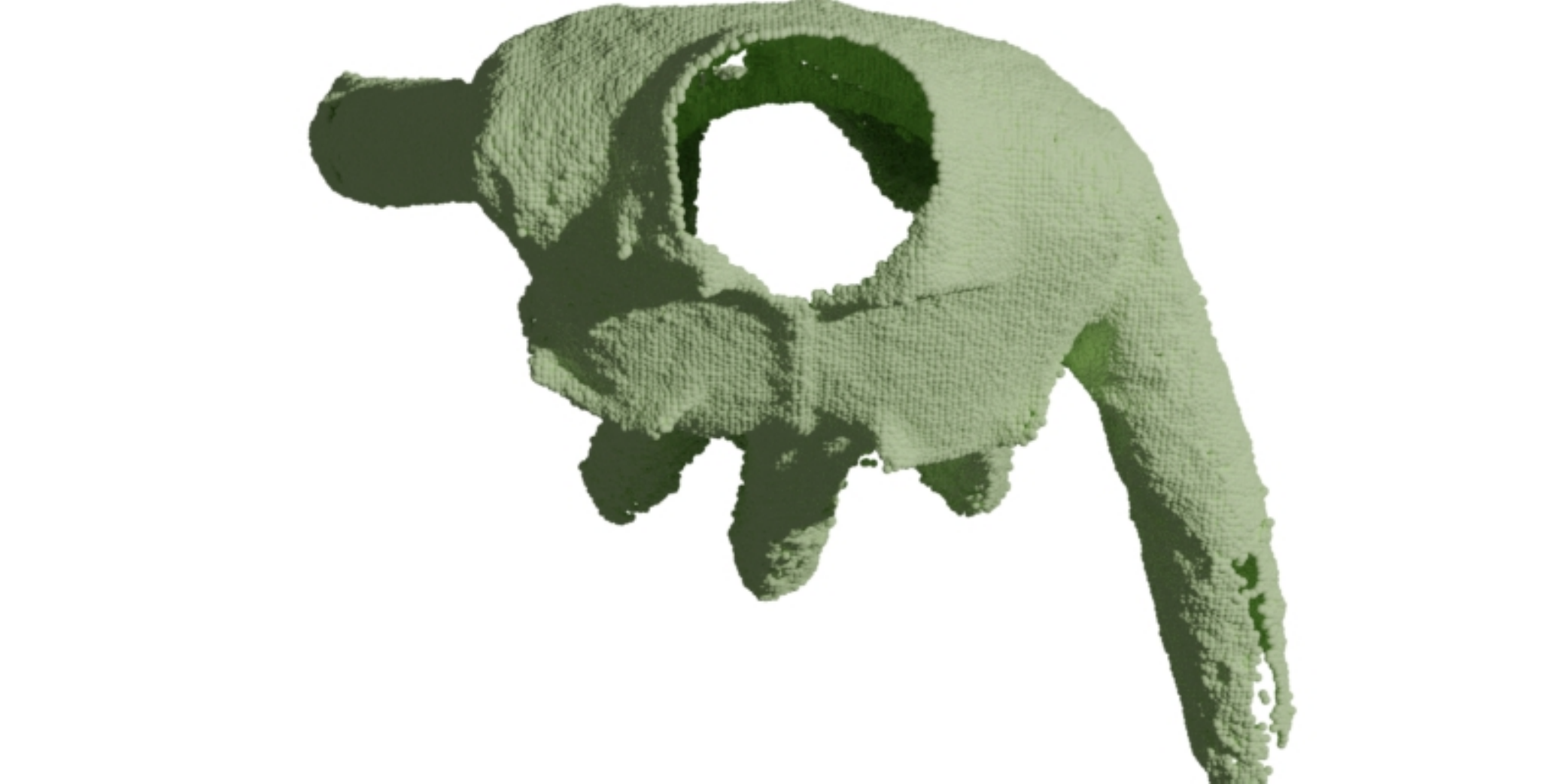}
\end{minipage}

\begin{minipage}[c]{.13\textwidth}
    \centering
    \includegraphics[width=1\linewidth]{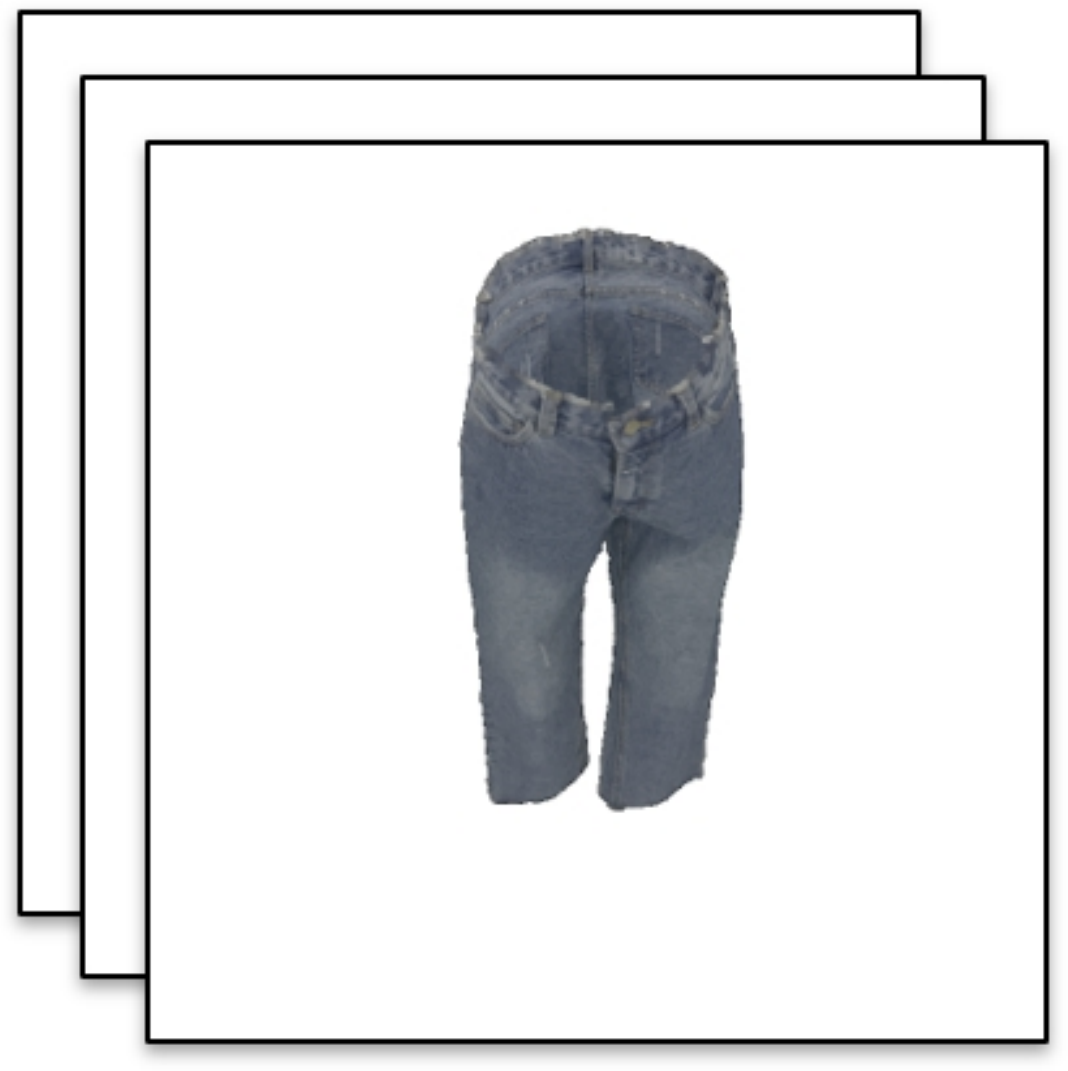}
\end{minipage}
\begin{minipage}[c]{.28\textwidth}
    \centering
    \includegraphics[width=.45\linewidth]{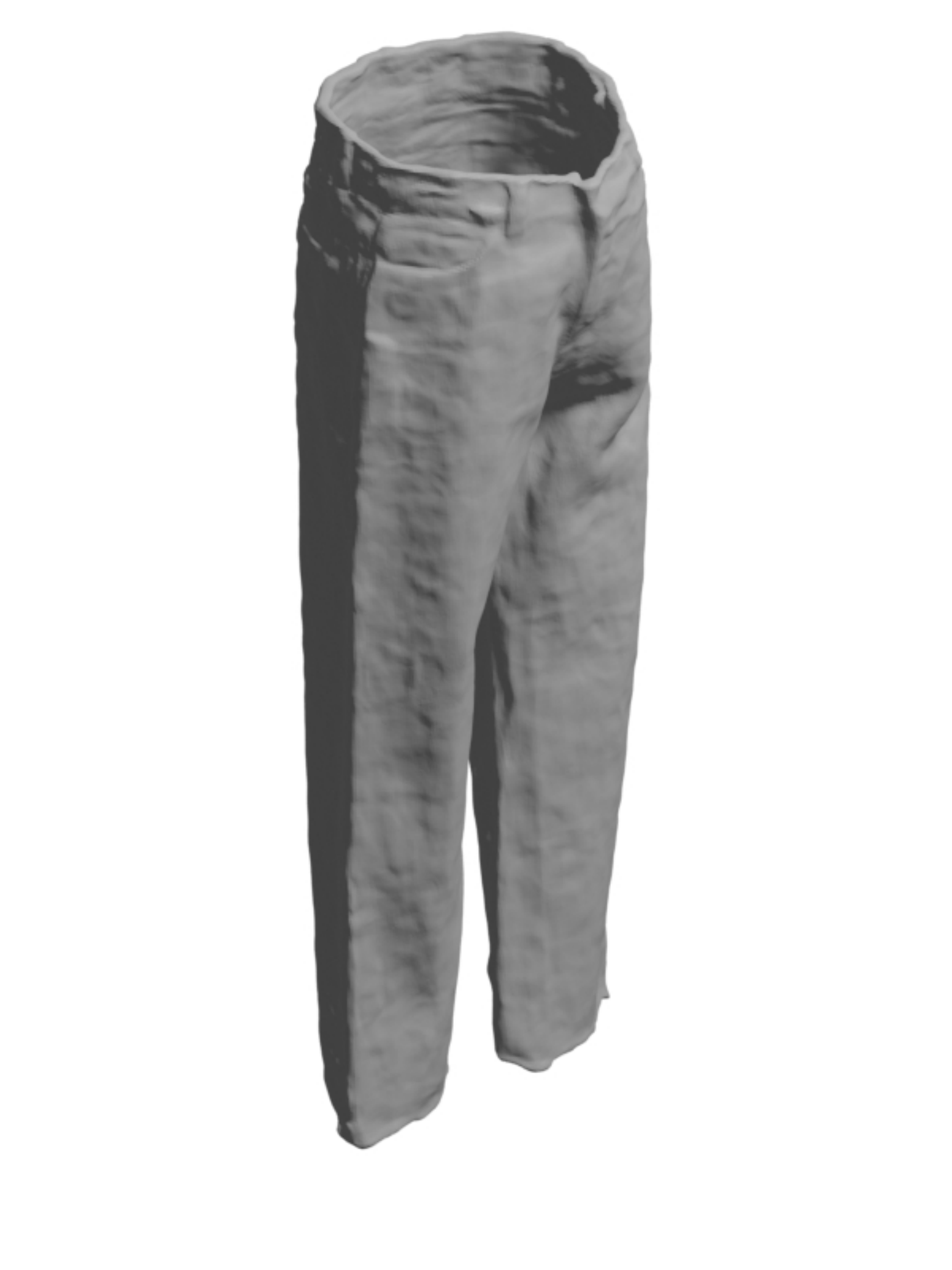}
    \includegraphics[width=.45\linewidth]{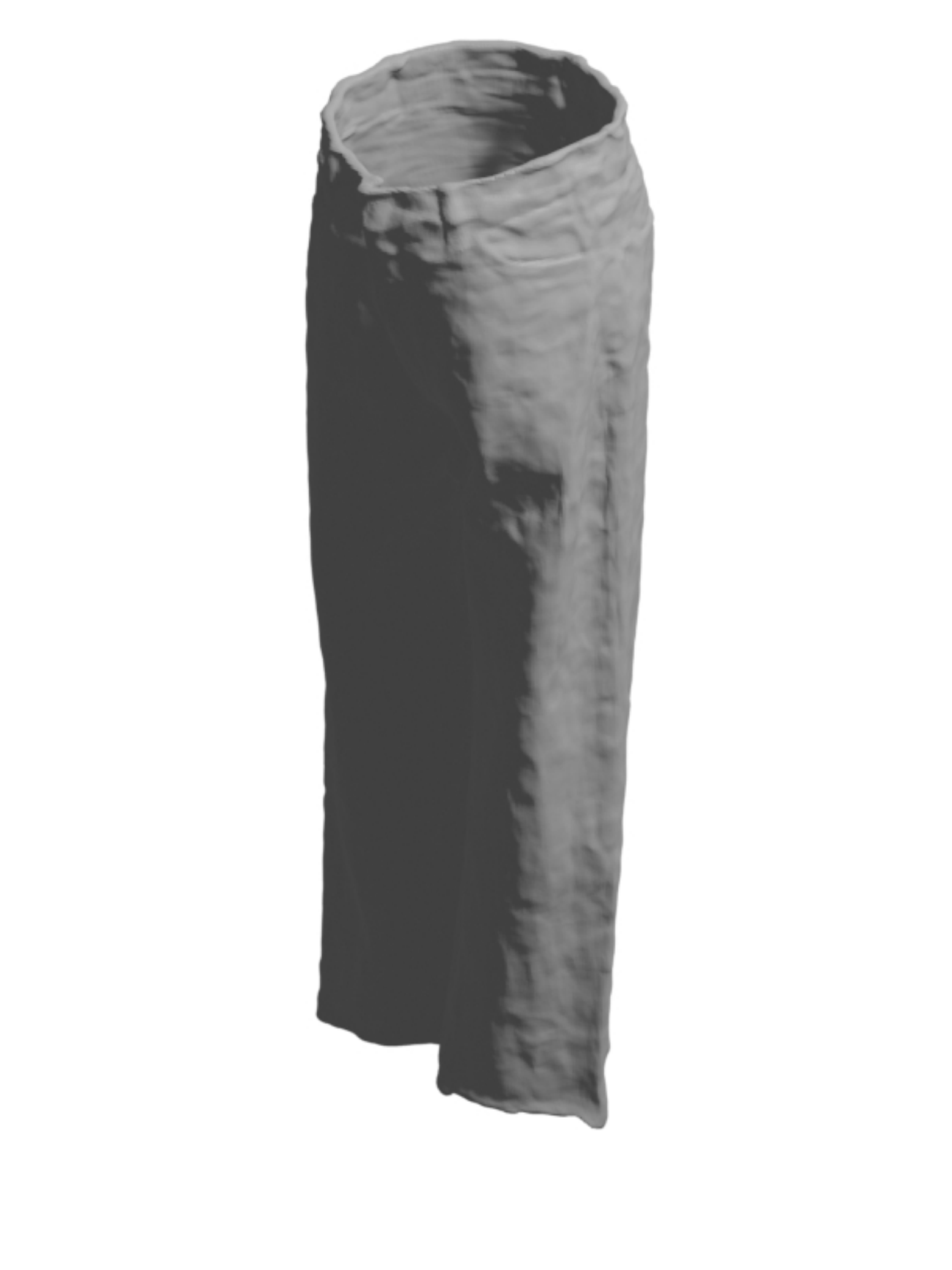}\\
    \includegraphics[width=.45\linewidth]{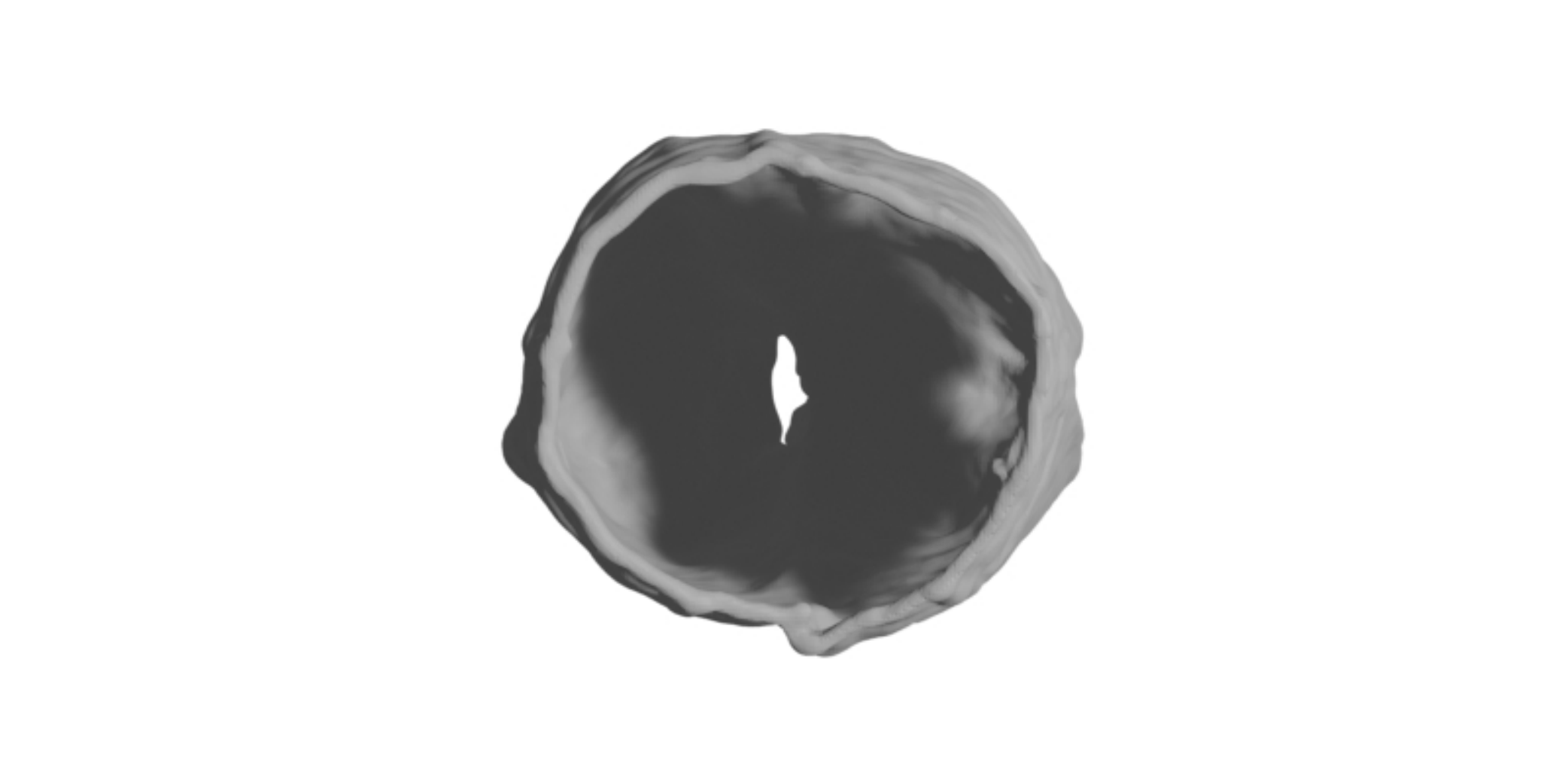}
\end{minipage}
\begin{minipage}[c]{.28\textwidth}
    \centering
    \includegraphics[width=.45\linewidth]{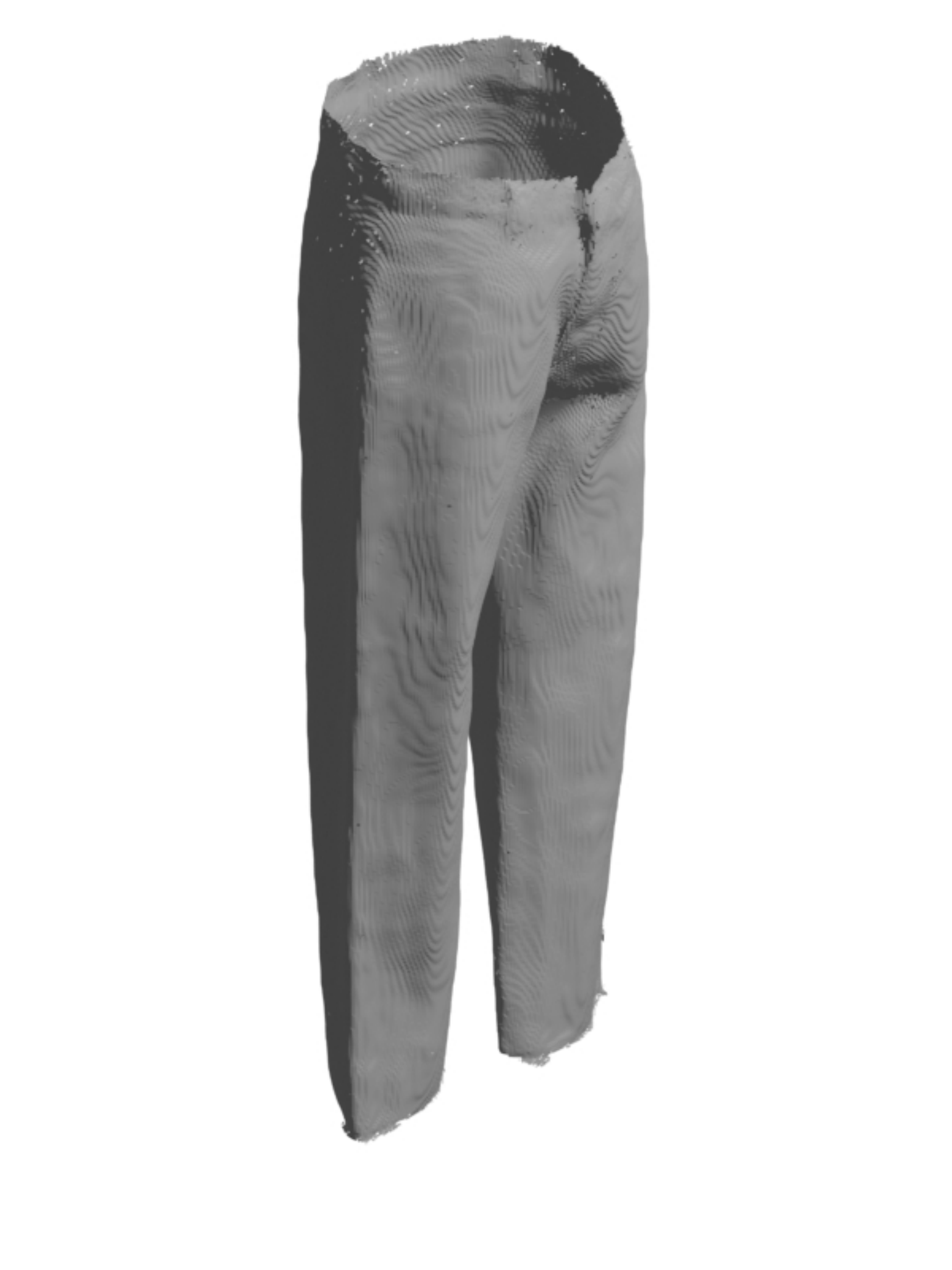}
    \includegraphics[width=.45\linewidth]{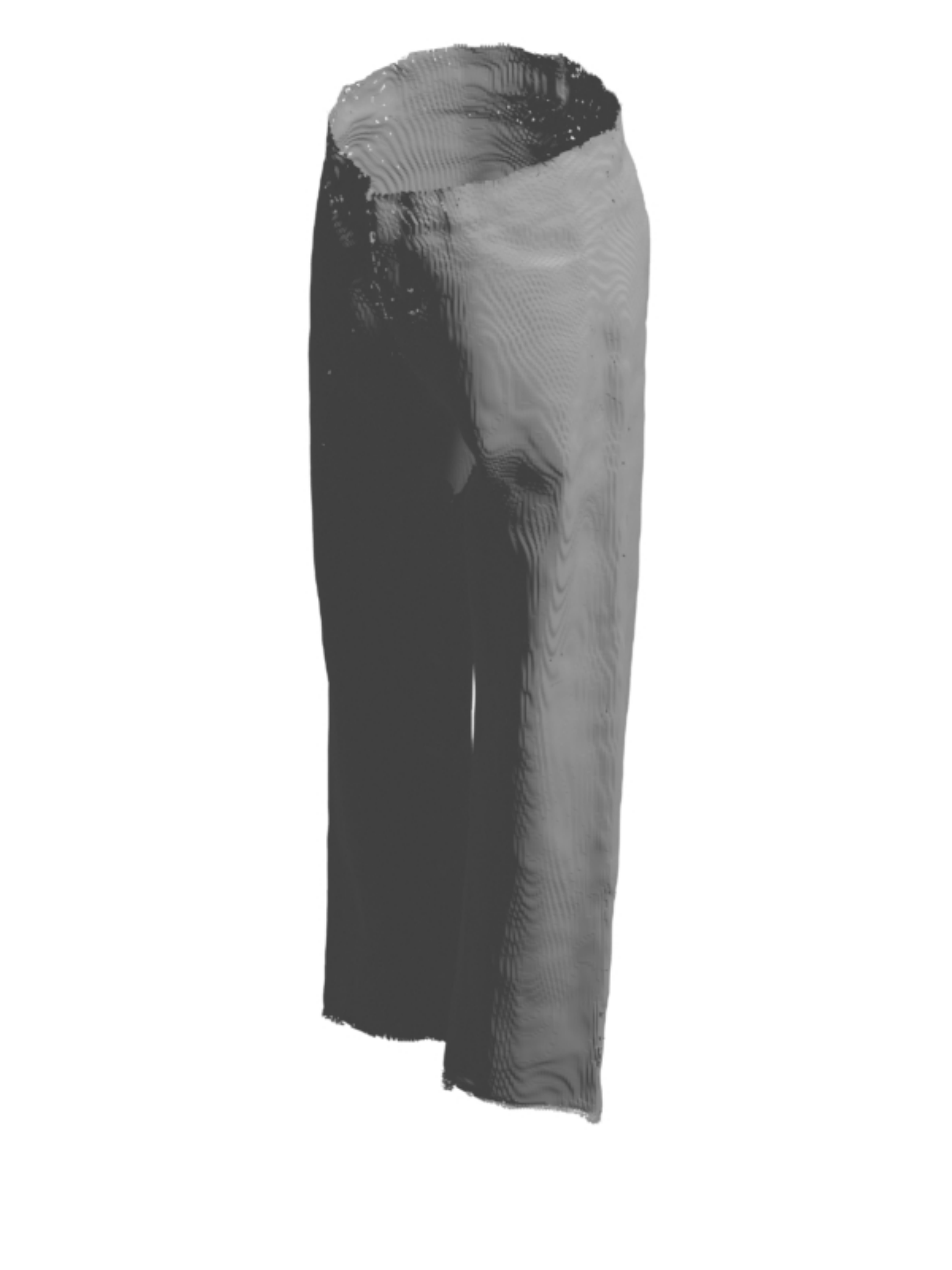}\\
    \includegraphics[width=.45\linewidth]{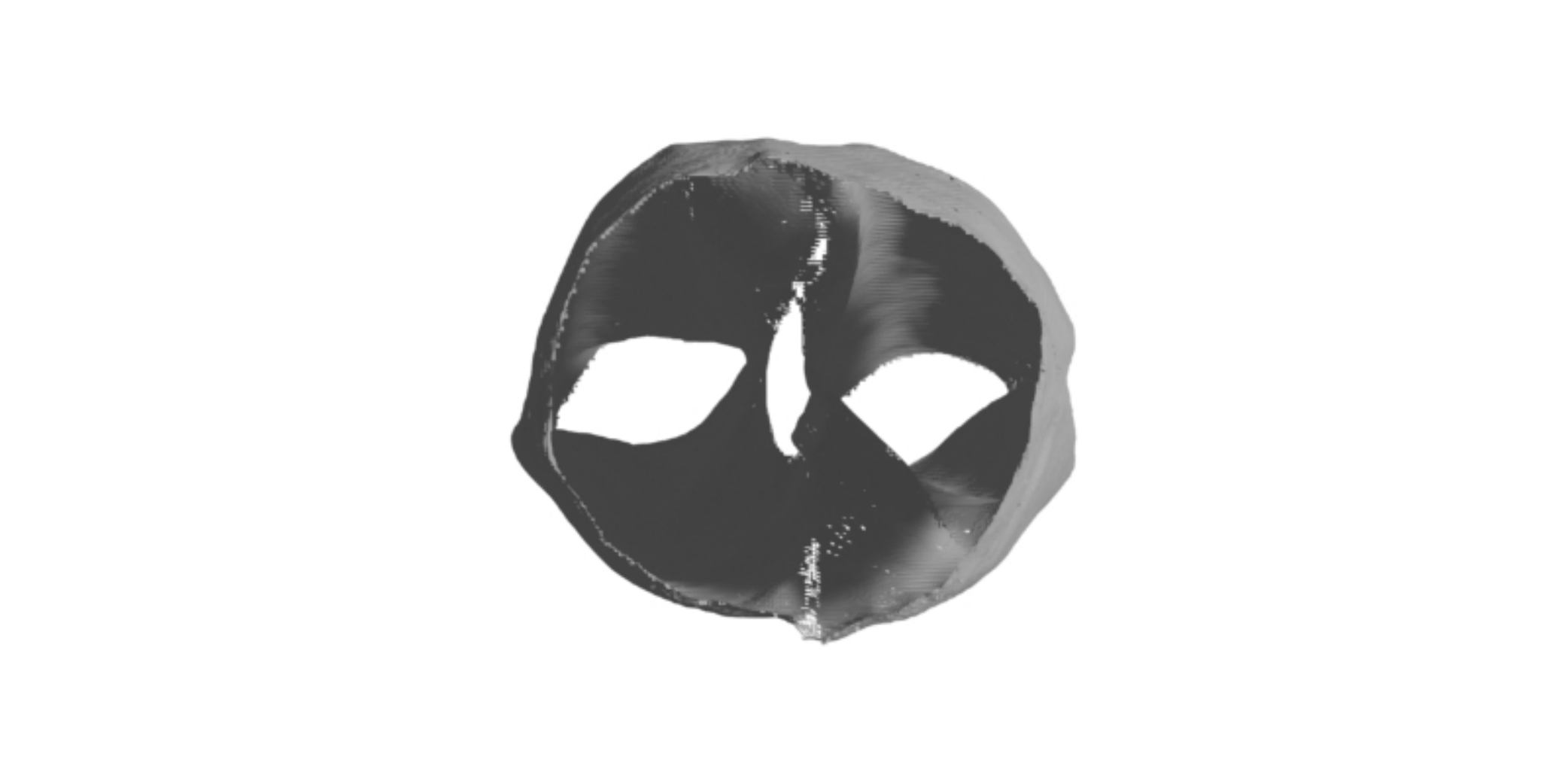}
\end{minipage}
\begin{minipage}[c]{.28\textwidth}
    \centering
    \includegraphics[width=.45\linewidth]{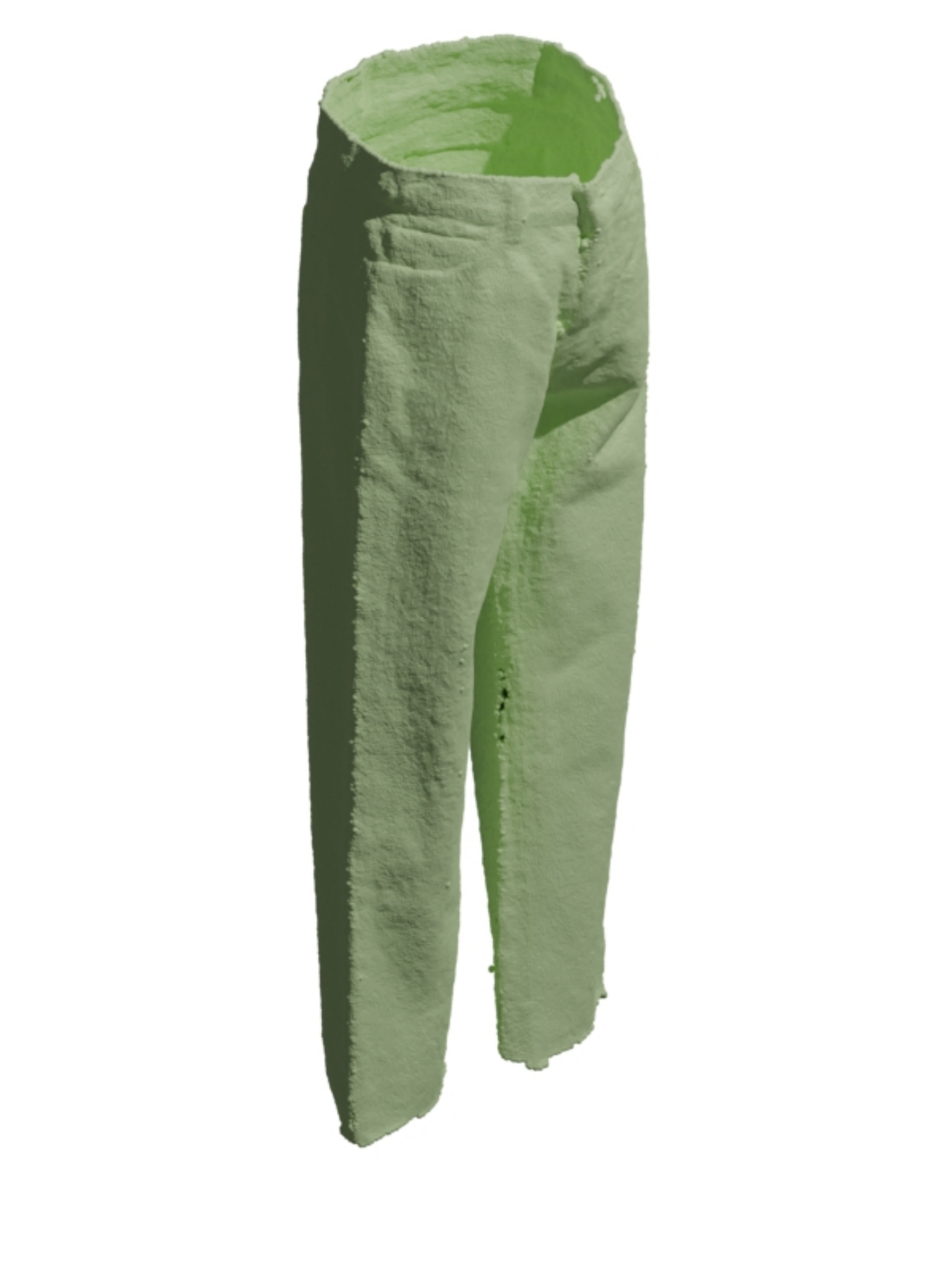}
    \includegraphics[width=.45\linewidth]{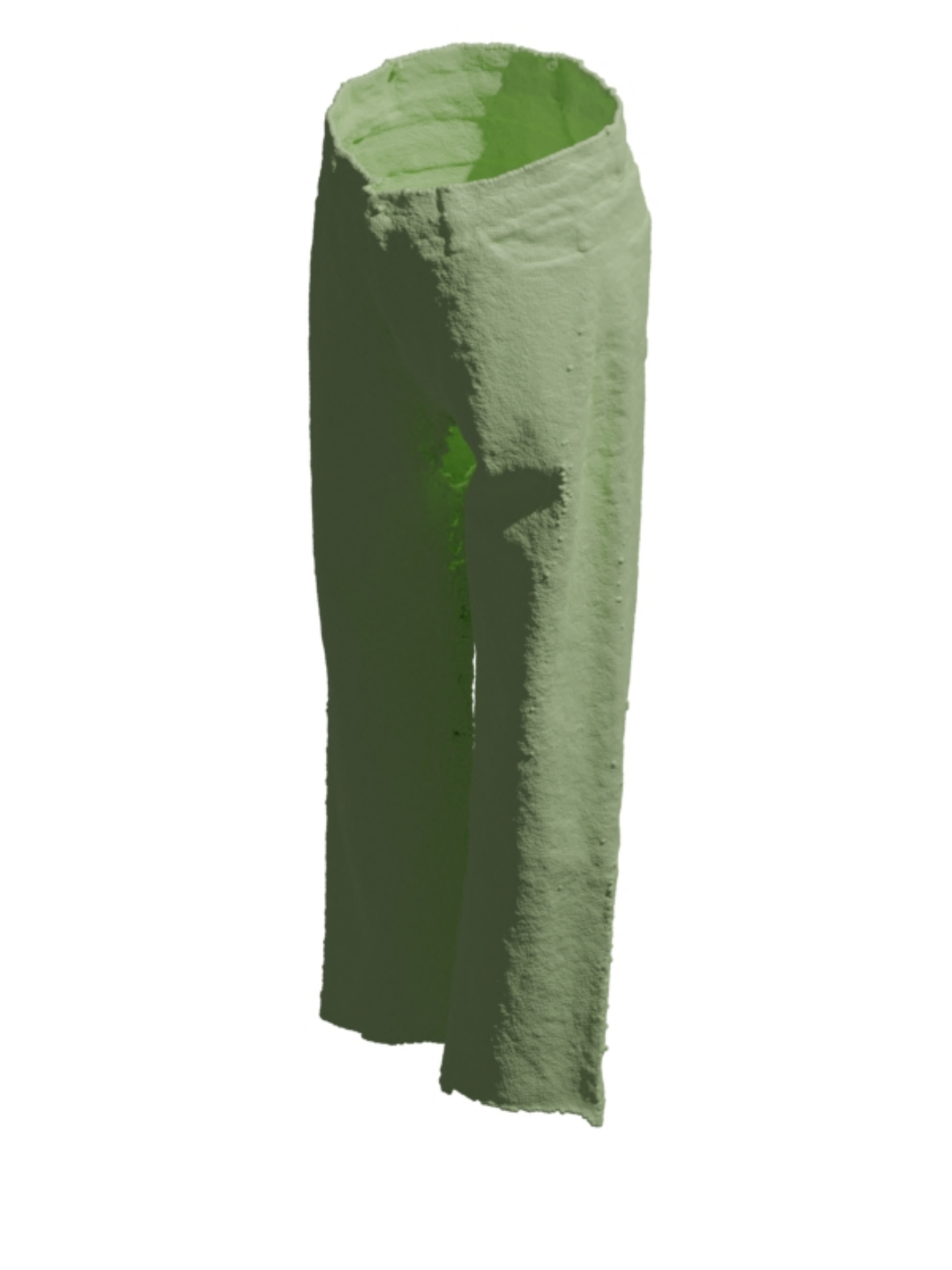}\\
    \includegraphics[width=.45\linewidth]{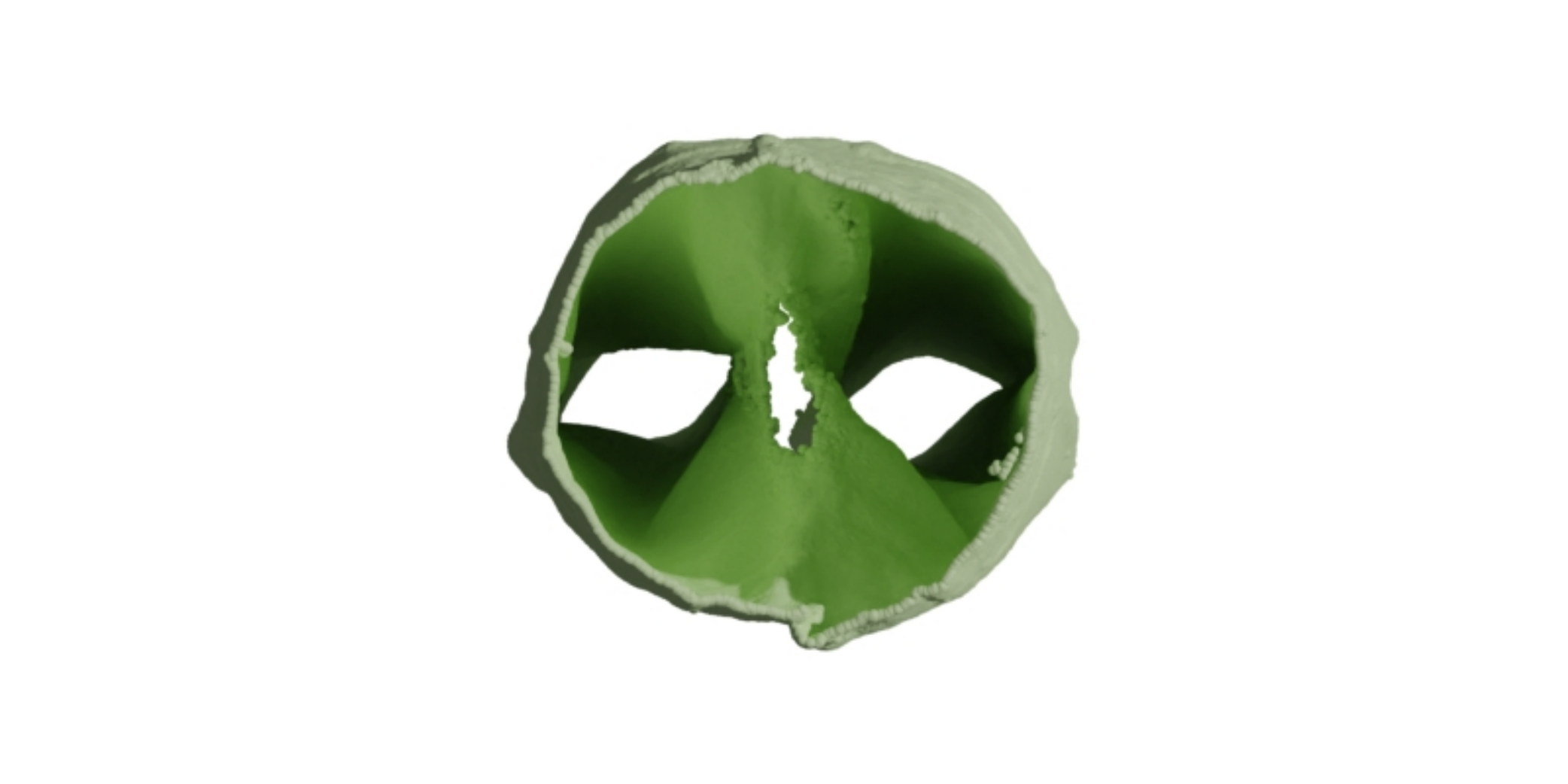}
\end{minipage}

\begin{minipage}[c]{.13\textwidth}
    \centering
    \includegraphics[width=1\linewidth]{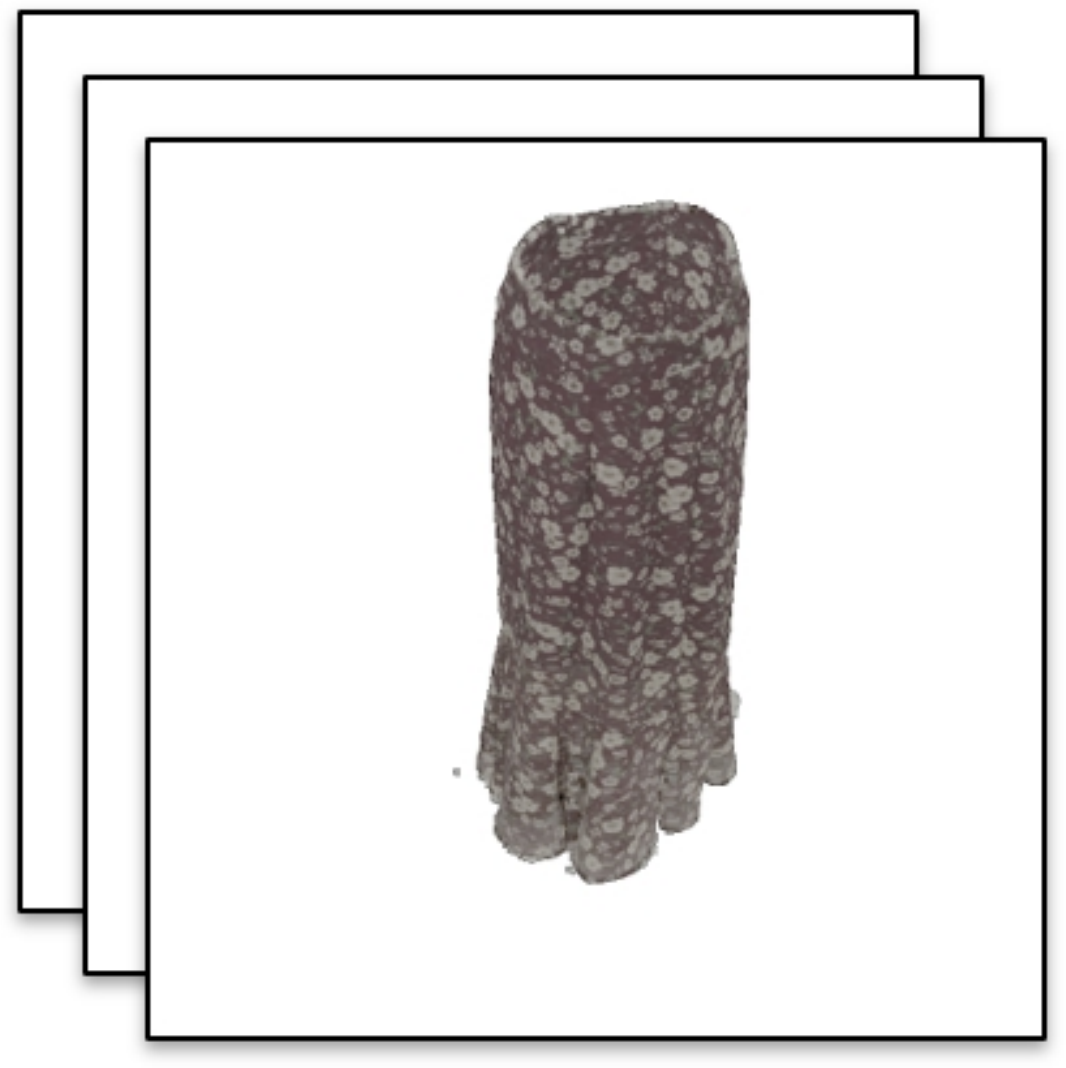}
\end{minipage}
\begin{minipage}[c]{.28\textwidth}
    \centering
    \includegraphics[width=.45\linewidth]{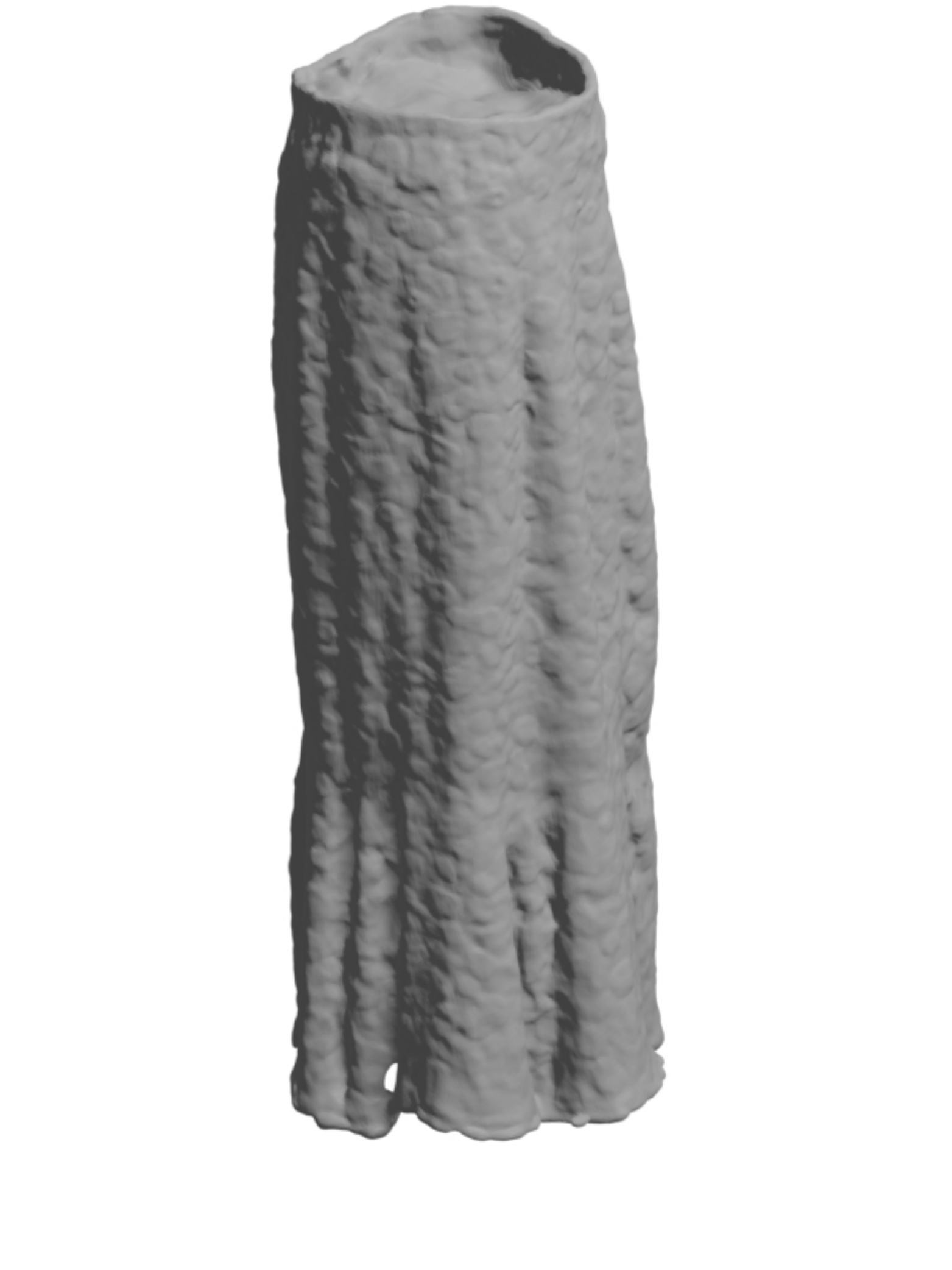}
    \includegraphics[width=.45\linewidth]{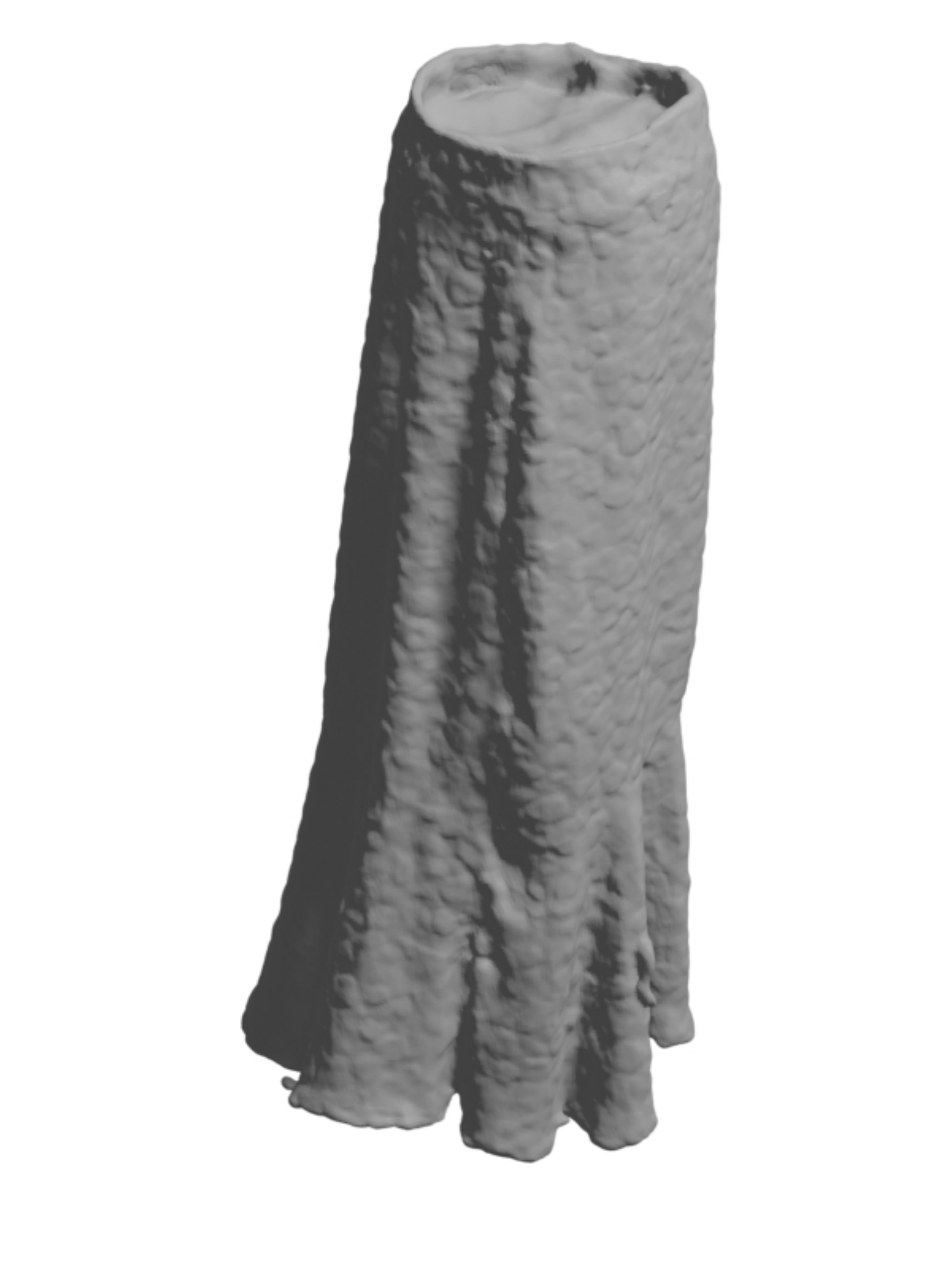}\\
    \includegraphics[width=.45\linewidth]{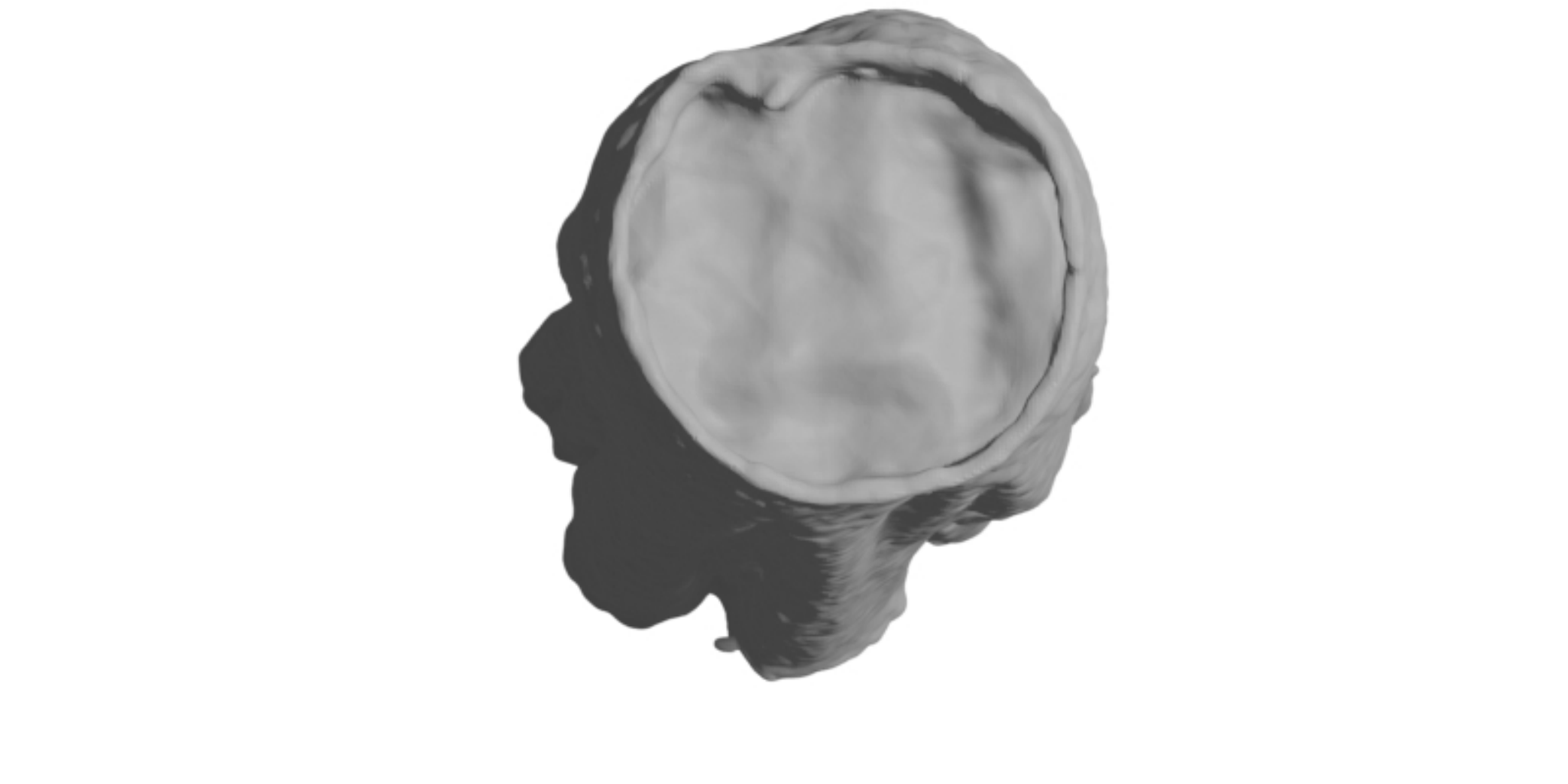}
\end{minipage}
\begin{minipage}[c]{.28\textwidth}
    \centering
    \includegraphics[width=.45\linewidth]{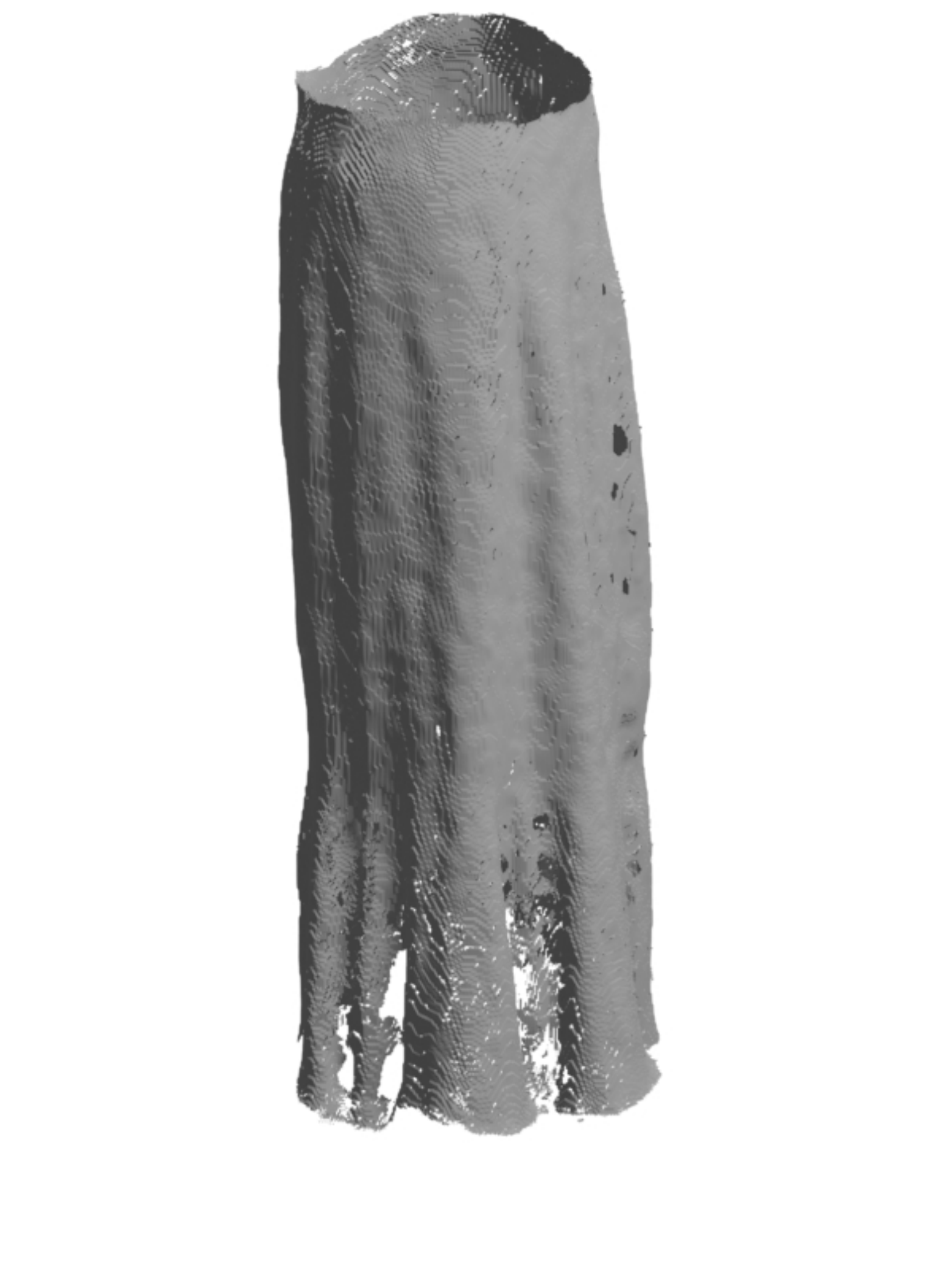}
    \includegraphics[width=.45\linewidth]{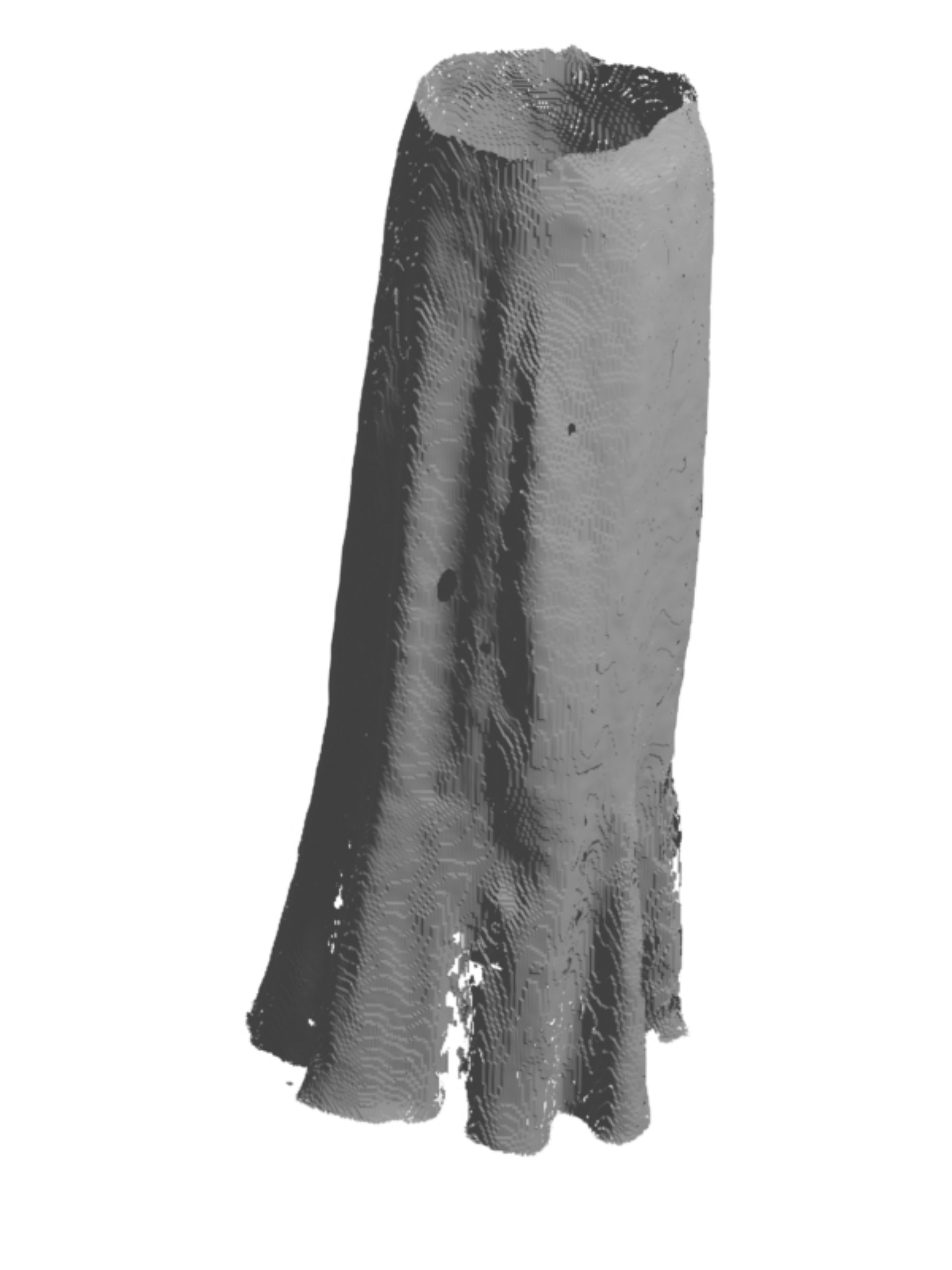}\\
    \includegraphics[width=.45\linewidth]{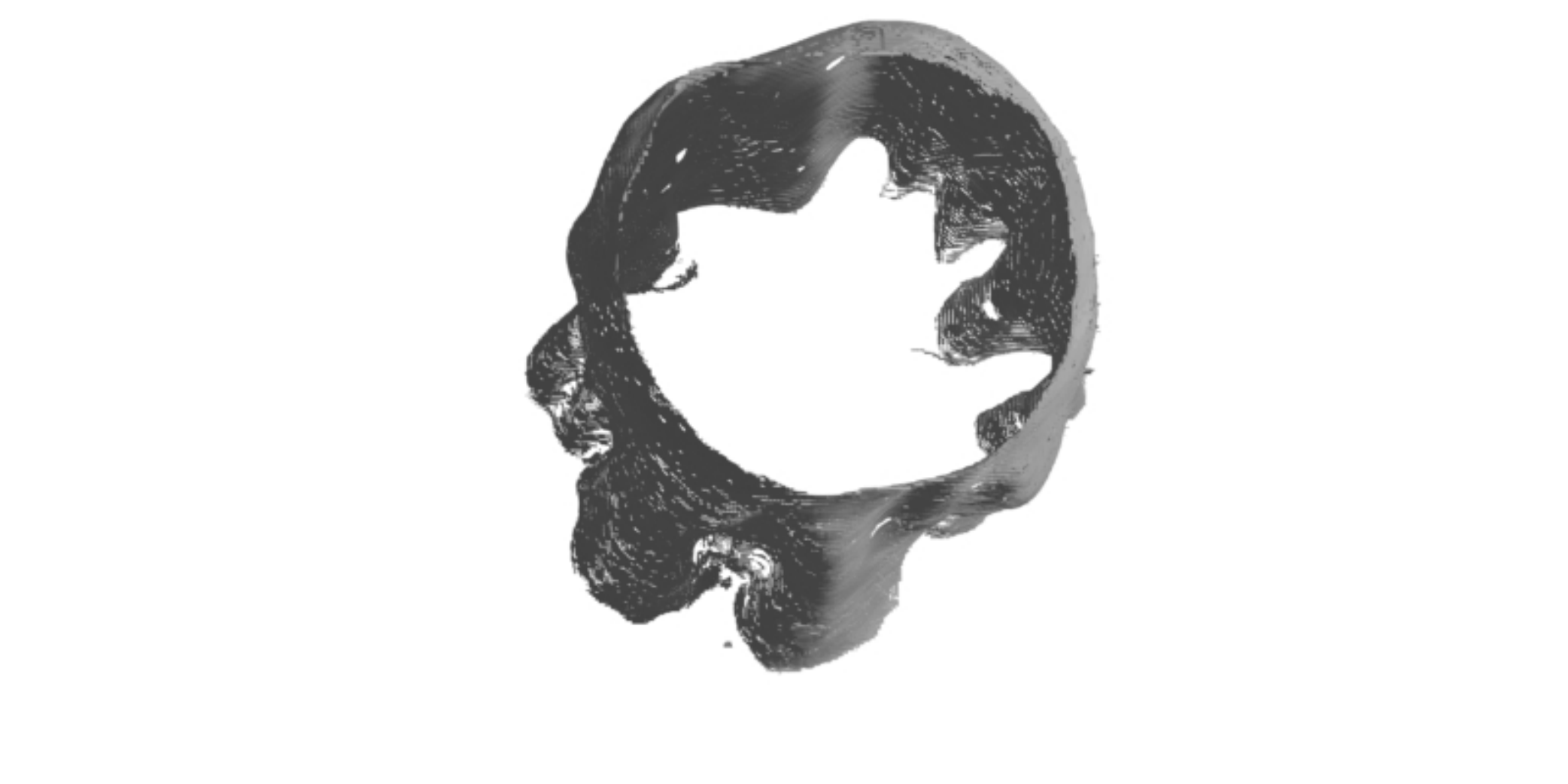}
\end{minipage}
\begin{minipage}[c]{.28\textwidth}
    \centering
    \includegraphics[width=.45\linewidth]{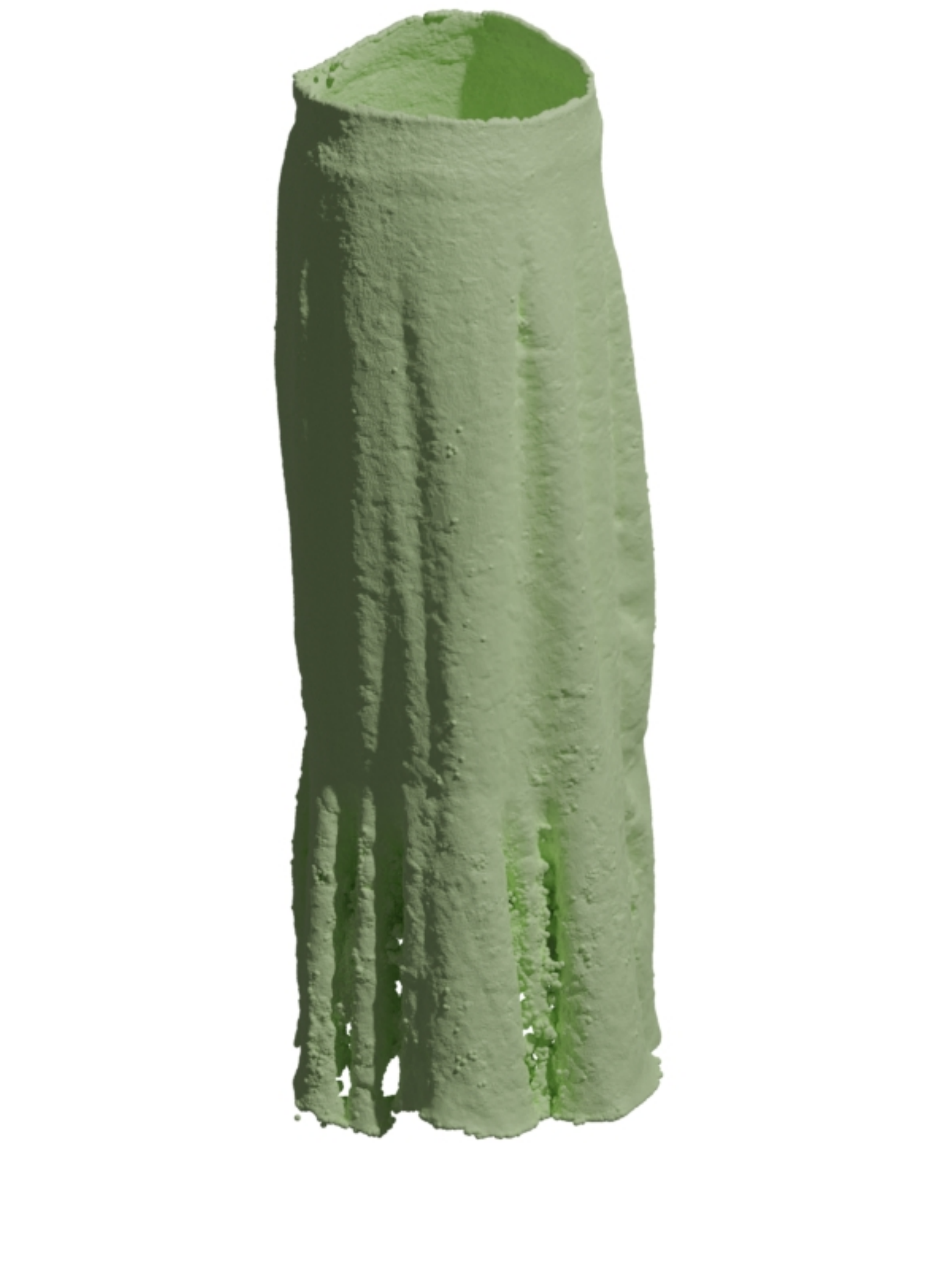}
    \includegraphics[width=.45\linewidth]{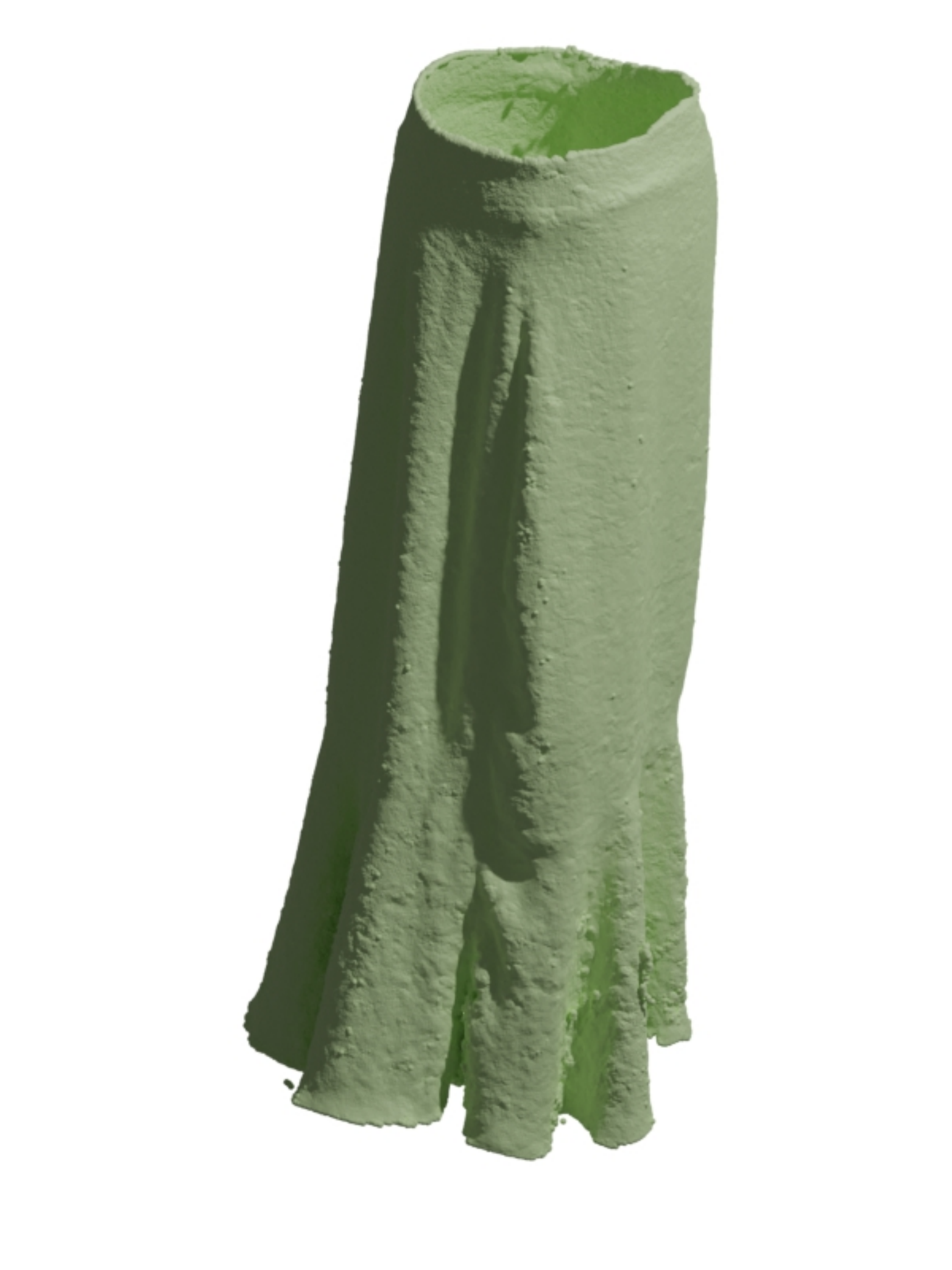}\\
    \includegraphics[width=.45\linewidth]{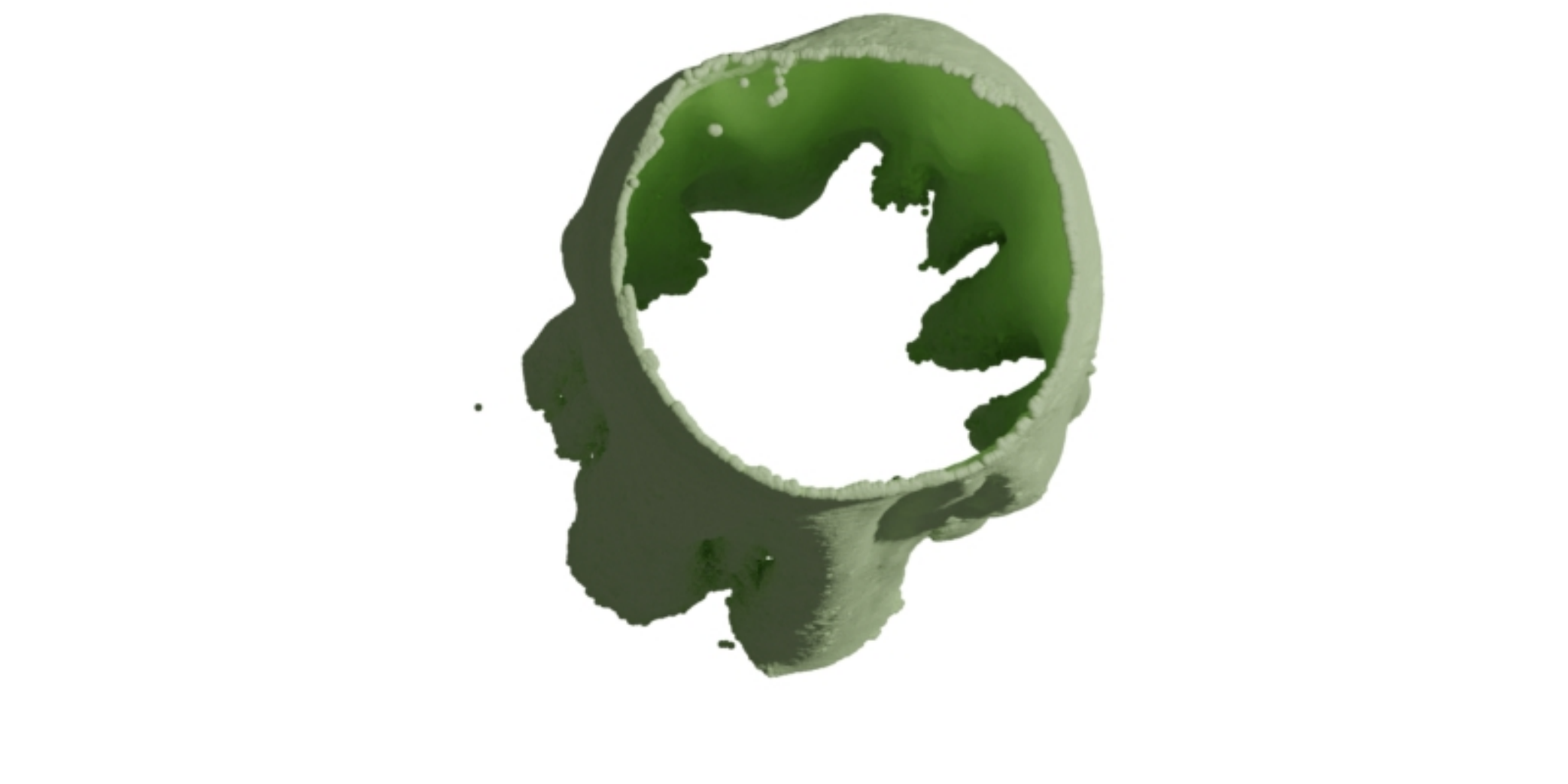}
\end{minipage}

\begin{minipage}[c]{.13\textwidth}
    \centering
    Input
\end{minipage}
\begin{minipage}[c]{.28\textwidth}
    \centering
    NeuS
\end{minipage}
\begin{minipage}[c]{.28\textwidth}
    \centering
    Ours
\end{minipage}
\begin{minipage}[c]{.28\textwidth}
    \centering
    Ground-truth
\end{minipage}

\caption{Additional results on the DF3D~\cite{zhu2020deep} dataset with mask supervision.}
\label{supp_df3d}
\end{figure*}

\begin{figure*}[h]

\begin{minipage}[c]{.13\textwidth}
    \centering
    \includegraphics[width=1\linewidth]{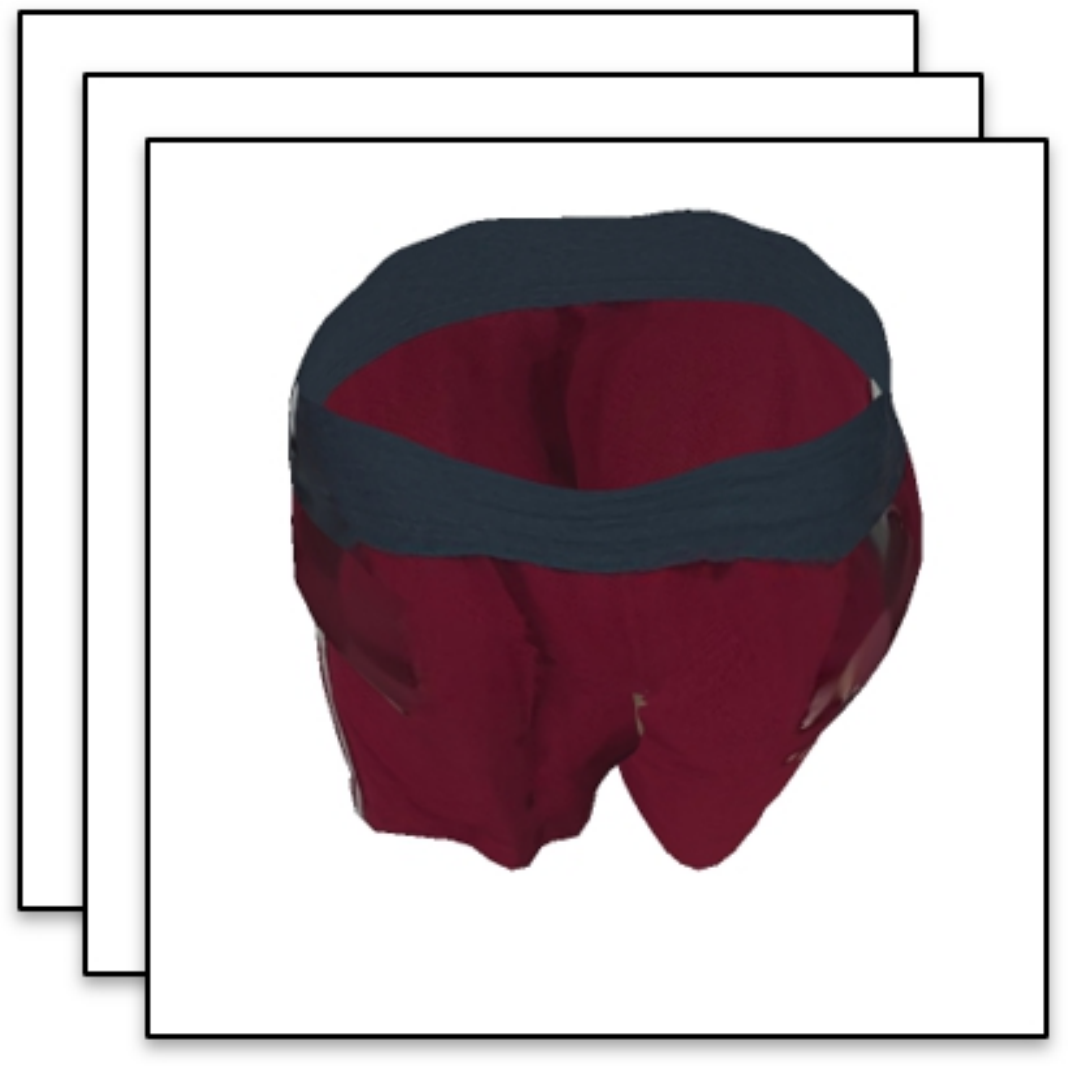}
\end{minipage}
\begin{minipage}[c]{.28\textwidth}
    \centering
    \includegraphics[width=.45\linewidth]{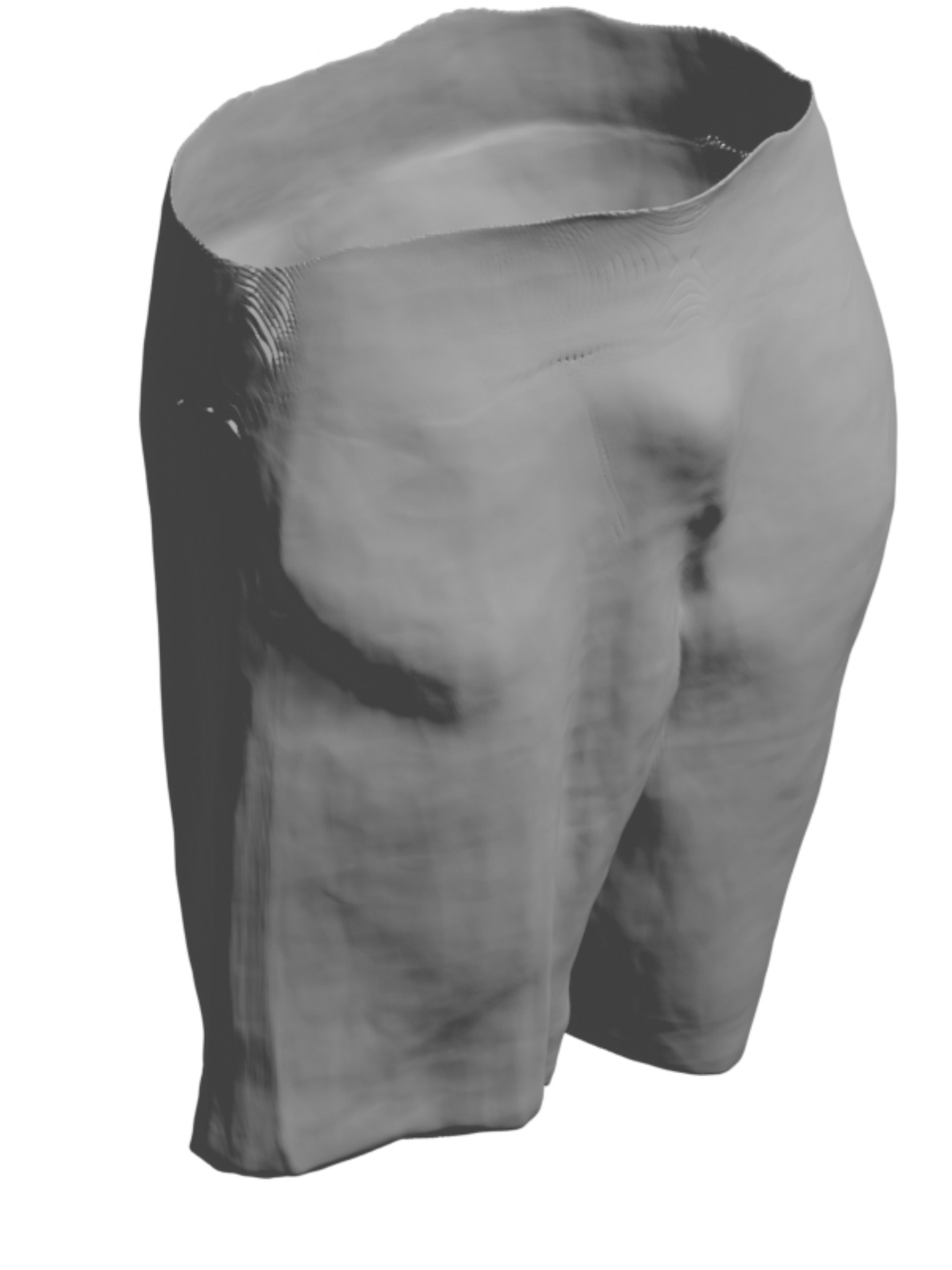}
    \includegraphics[width=.45\linewidth]{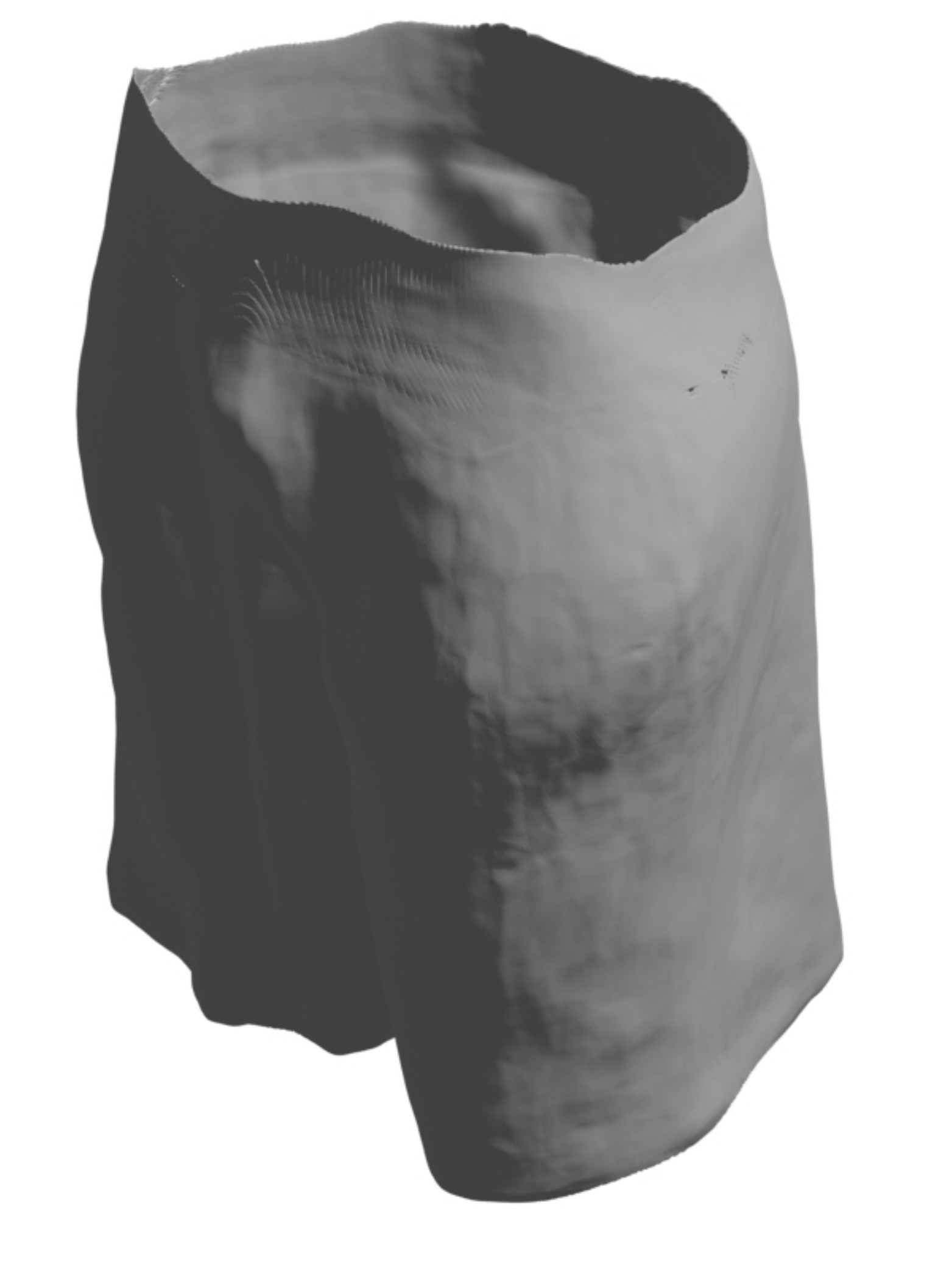}\\
    \includegraphics[width=.45\linewidth]{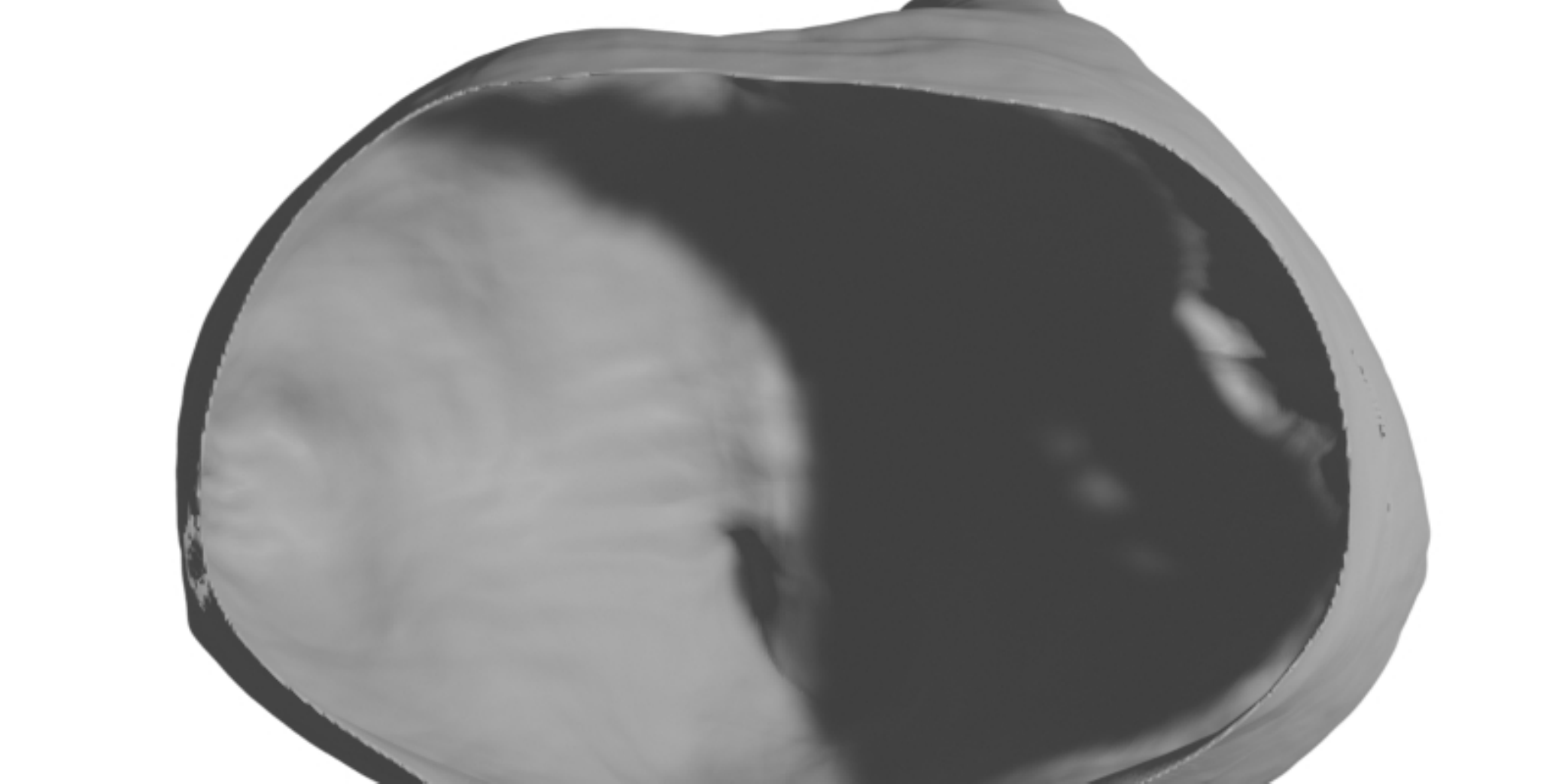}
\end{minipage}
\begin{minipage}[c]{.28\textwidth}
    \centering
    \includegraphics[width=.45\linewidth]{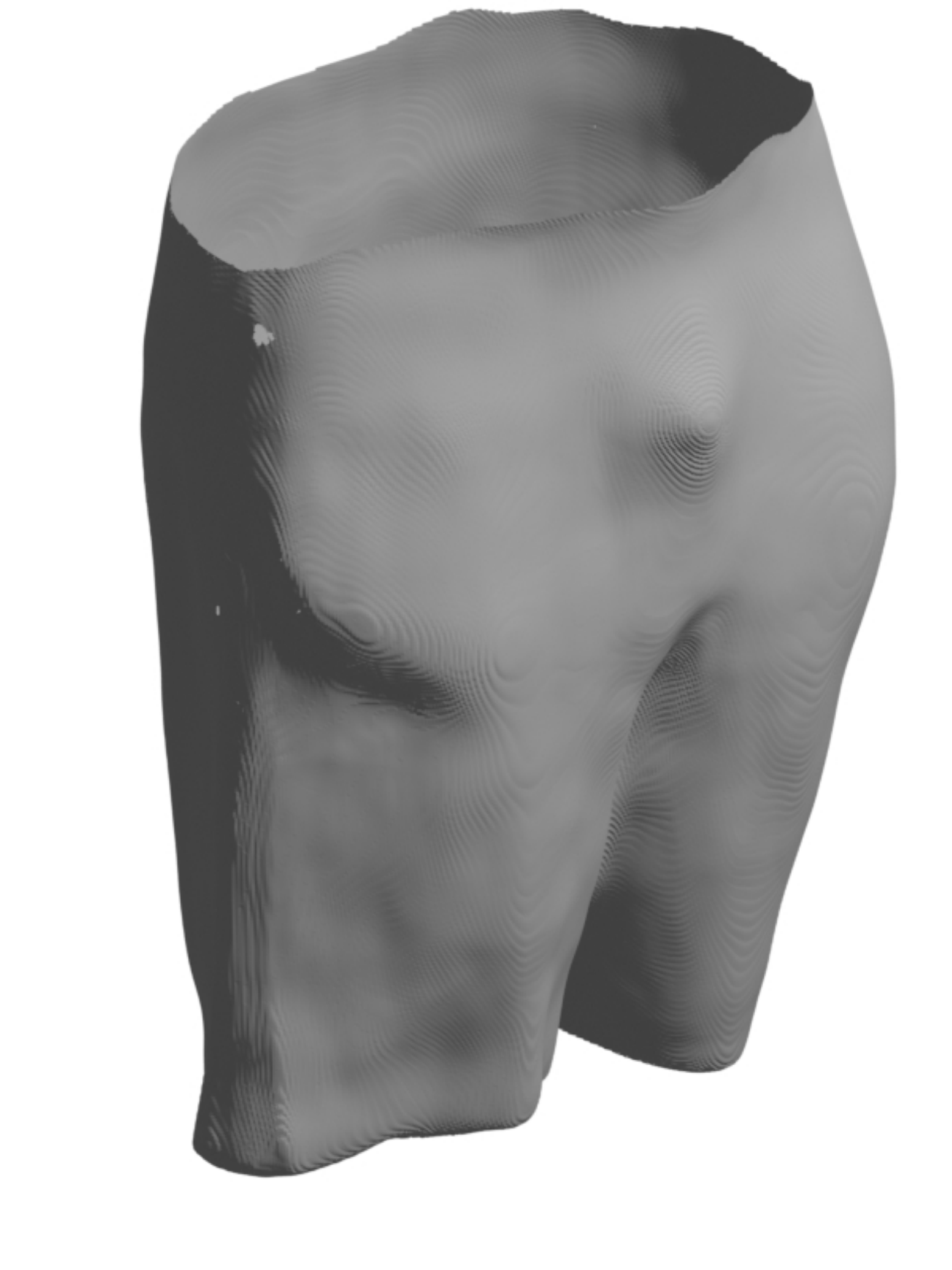}
    \includegraphics[width=.45\linewidth]{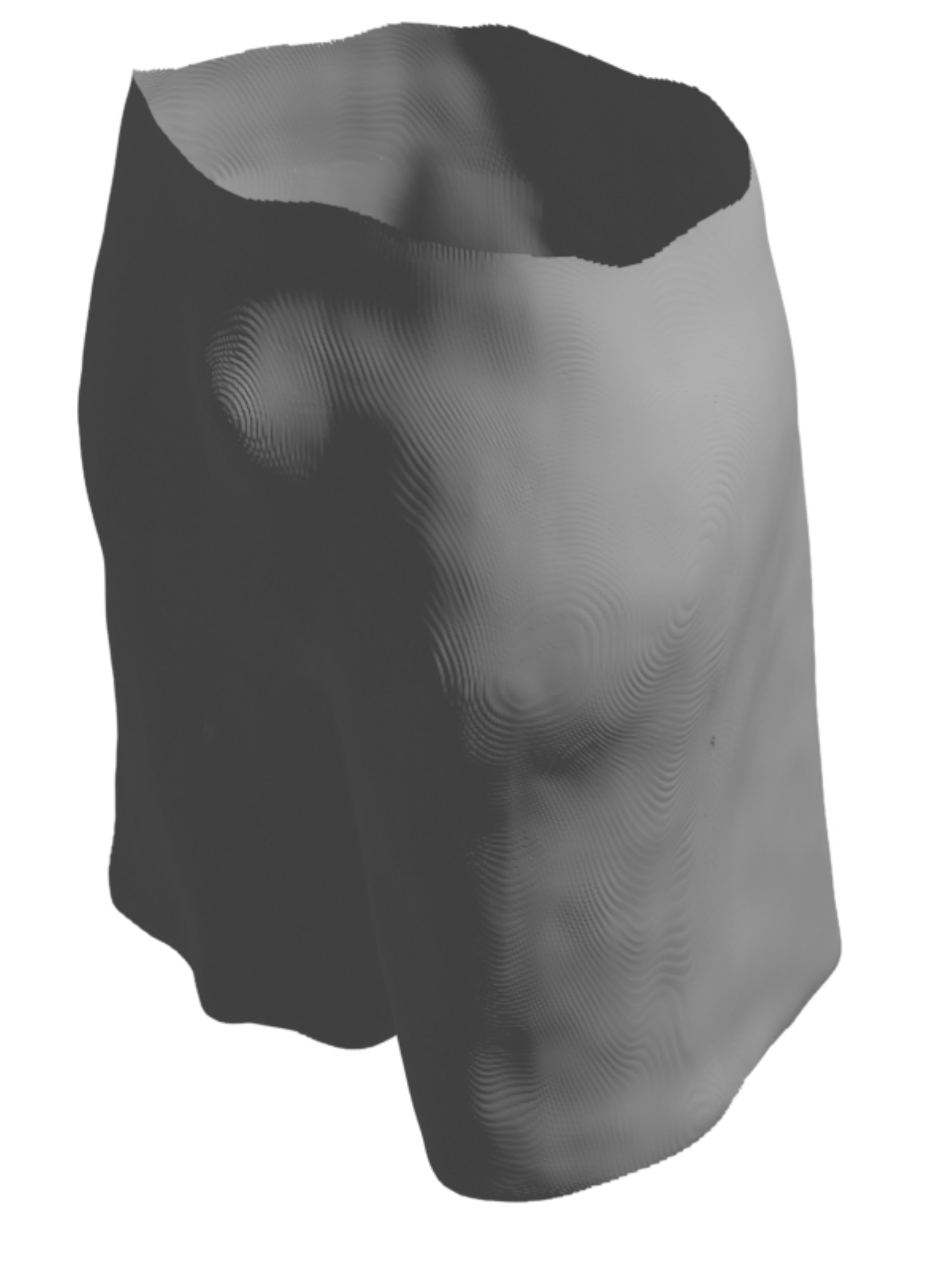}\\
    \includegraphics[width=.45\linewidth]{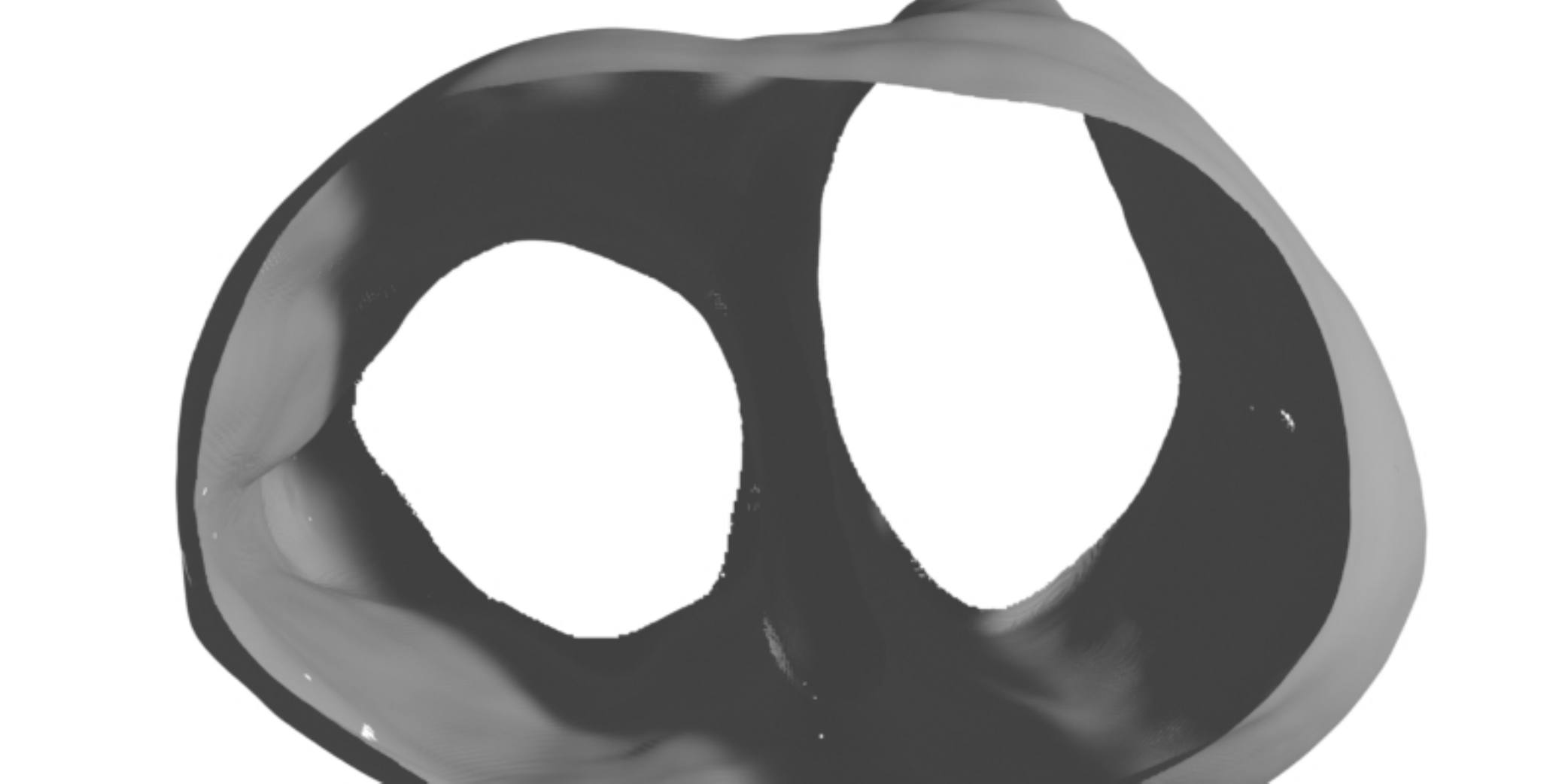}
\end{minipage}
\begin{minipage}[c]{.28\textwidth}
    \centering
    \includegraphics[width=.45\linewidth]{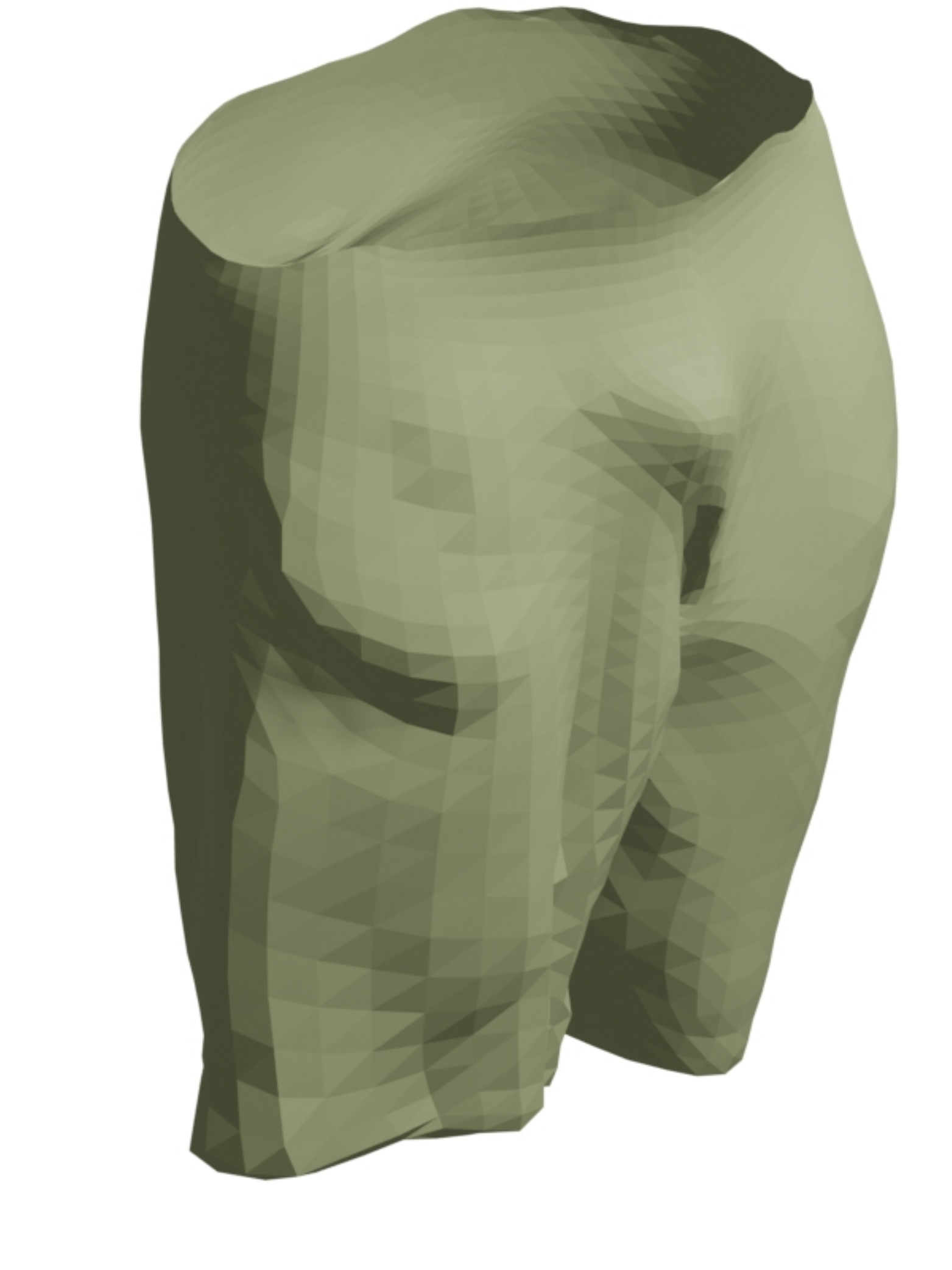}
    \includegraphics[width=.45\linewidth]{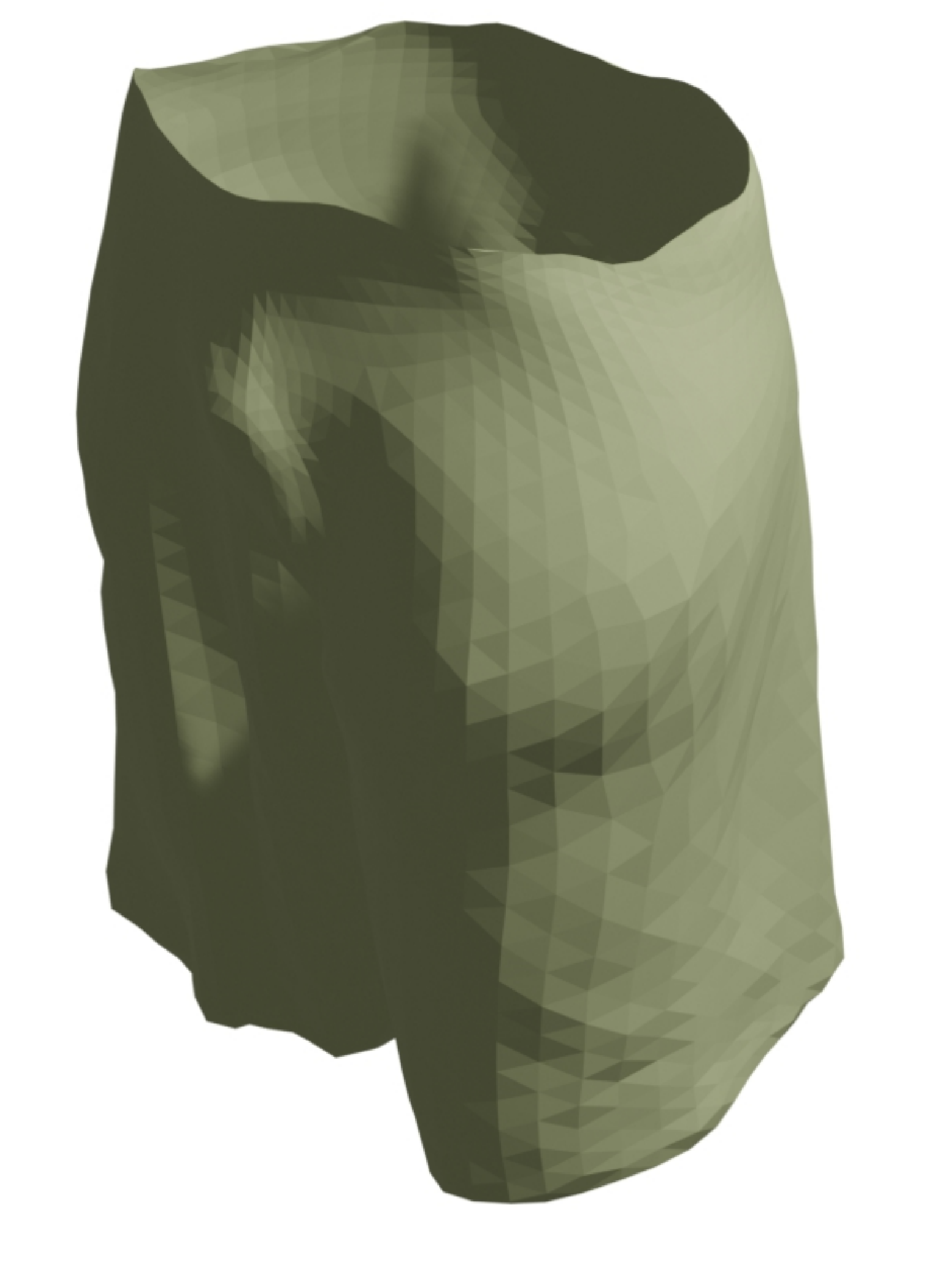}\\
    \includegraphics[width=.45\linewidth]{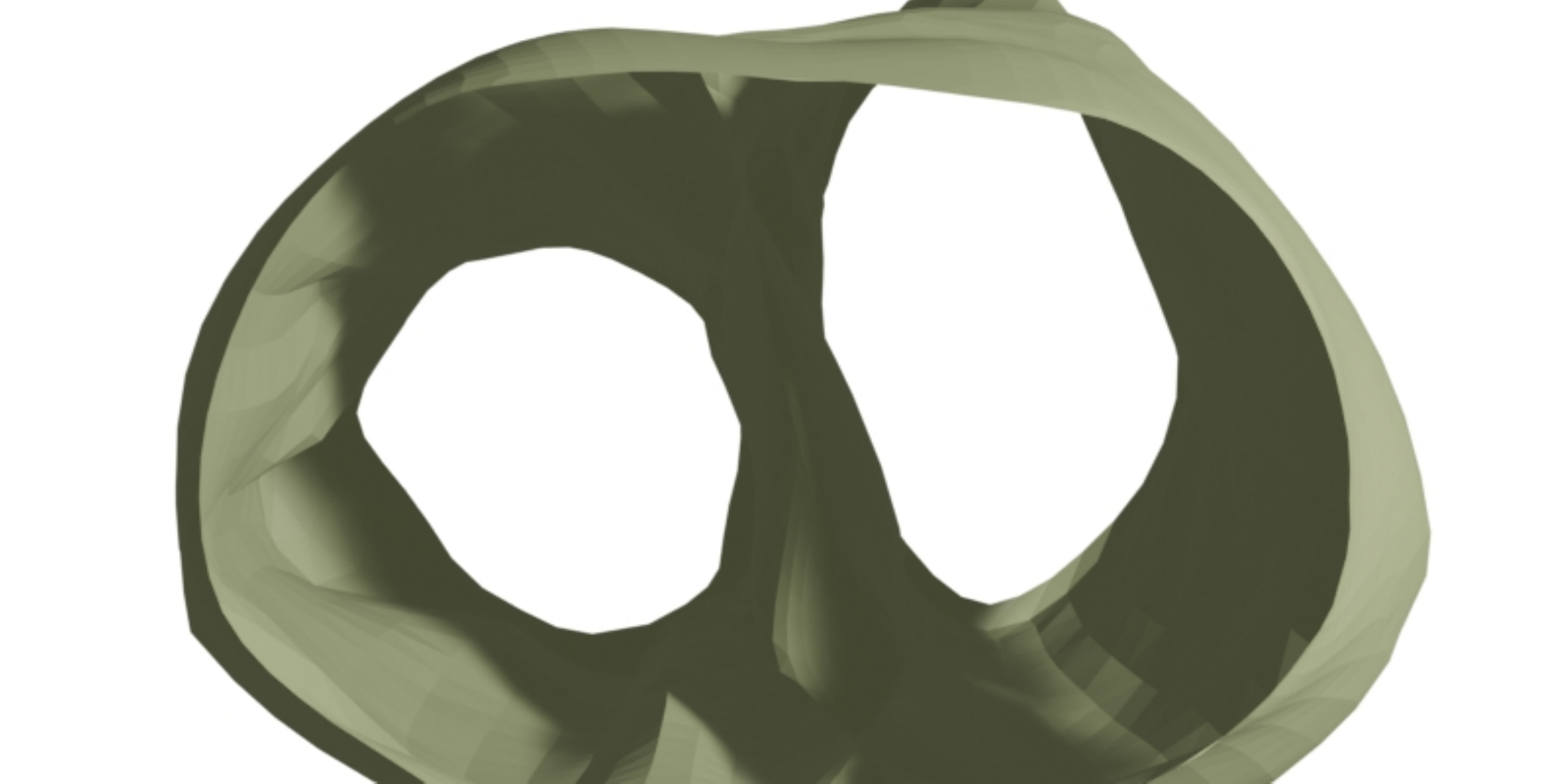}
\end{minipage}

\begin{minipage}[c]{.13\textwidth}
    \centering
    \includegraphics[width=1\linewidth]{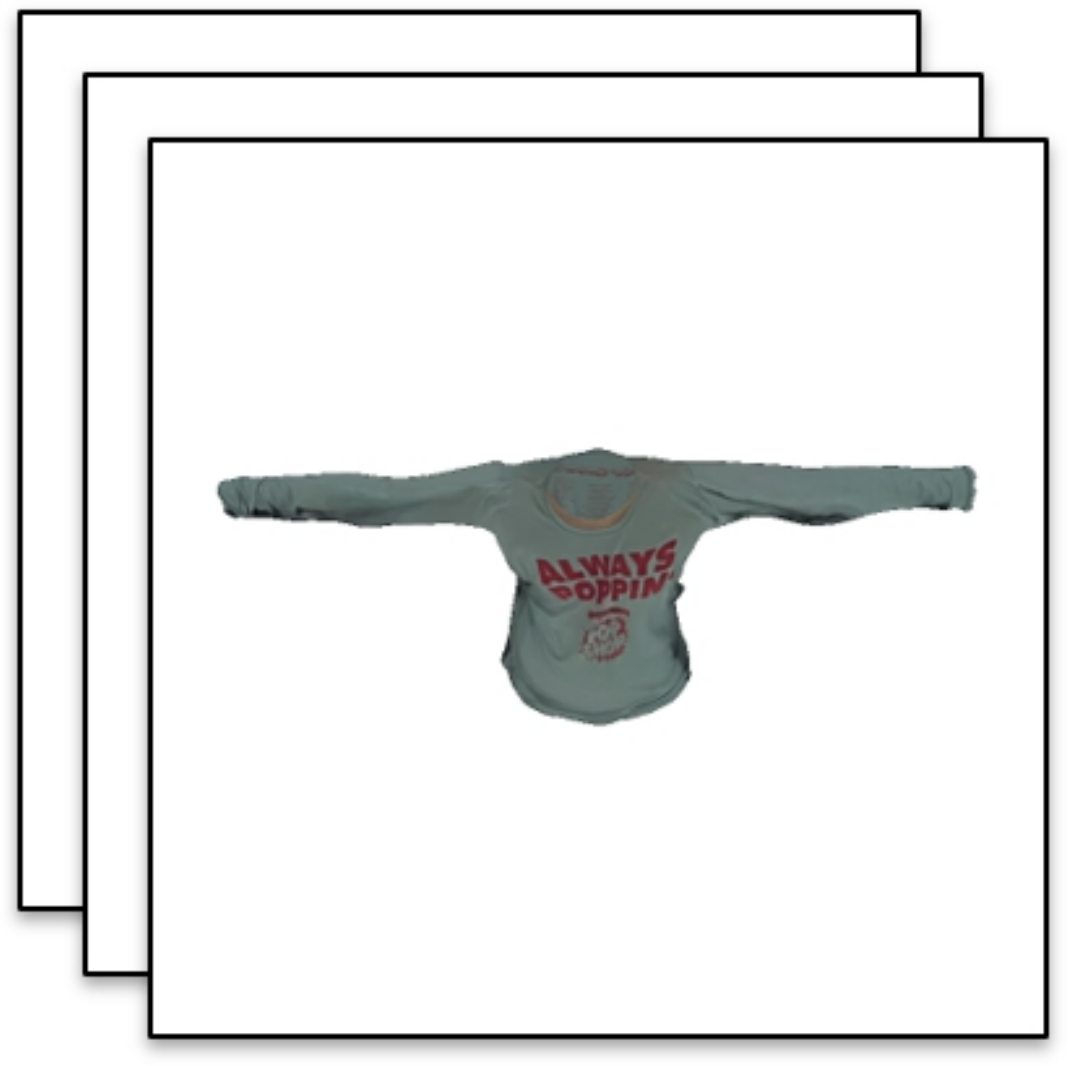}
\end{minipage}
\begin{minipage}[c]{.28\textwidth}
    \centering
    \includegraphics[width=.45\linewidth]{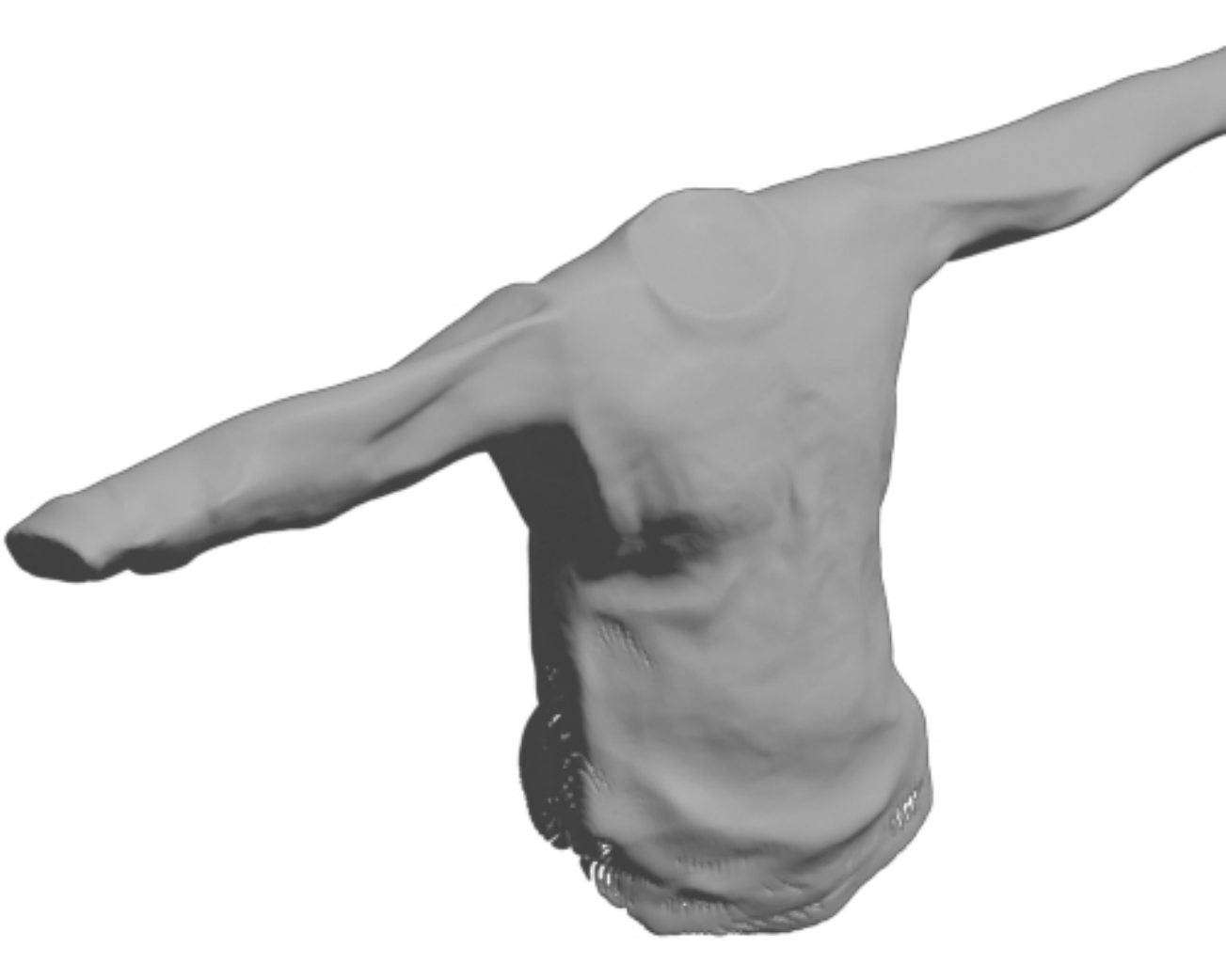}
    \includegraphics[width=.45\linewidth]{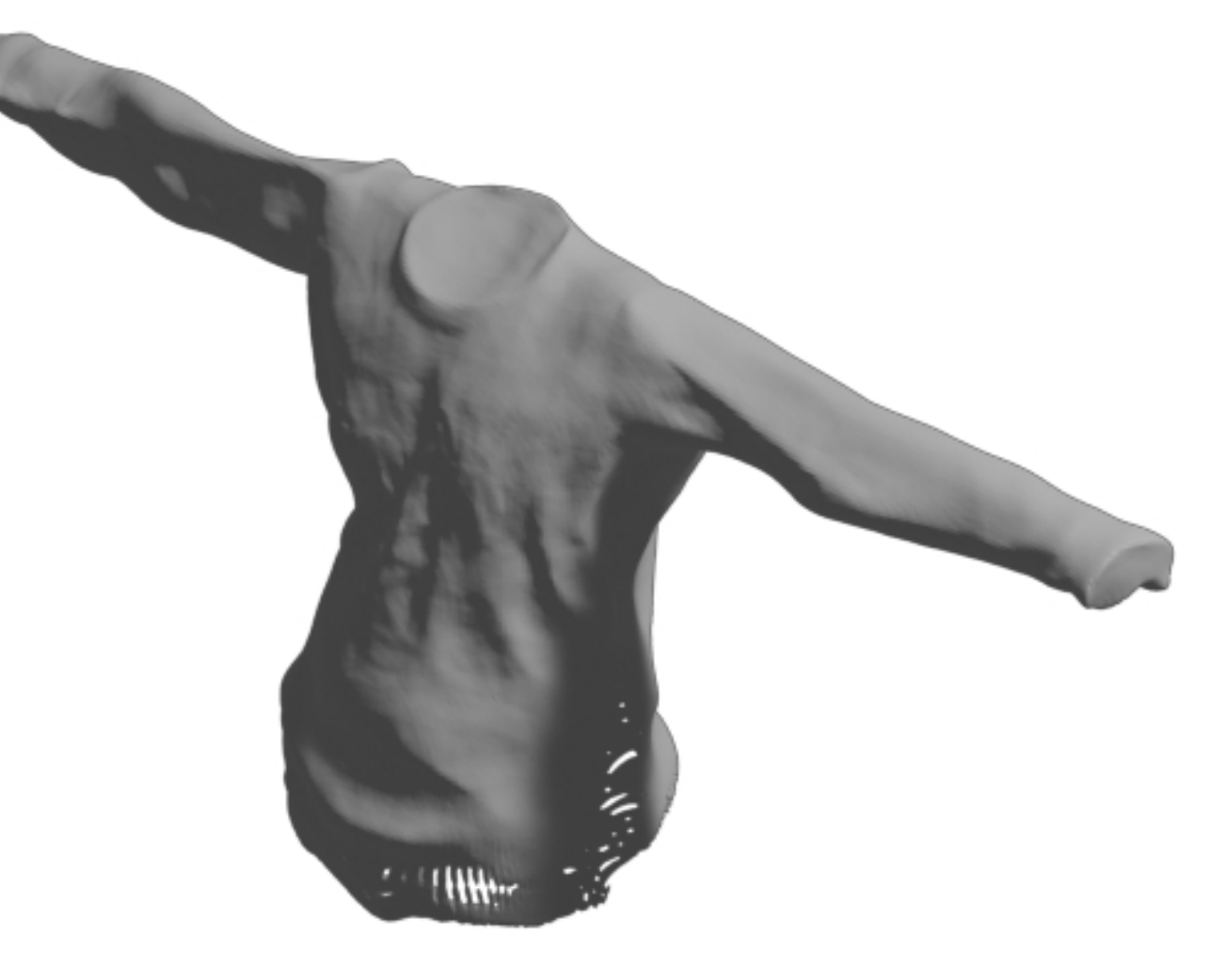}\\
    \includegraphics[width=.6\linewidth]{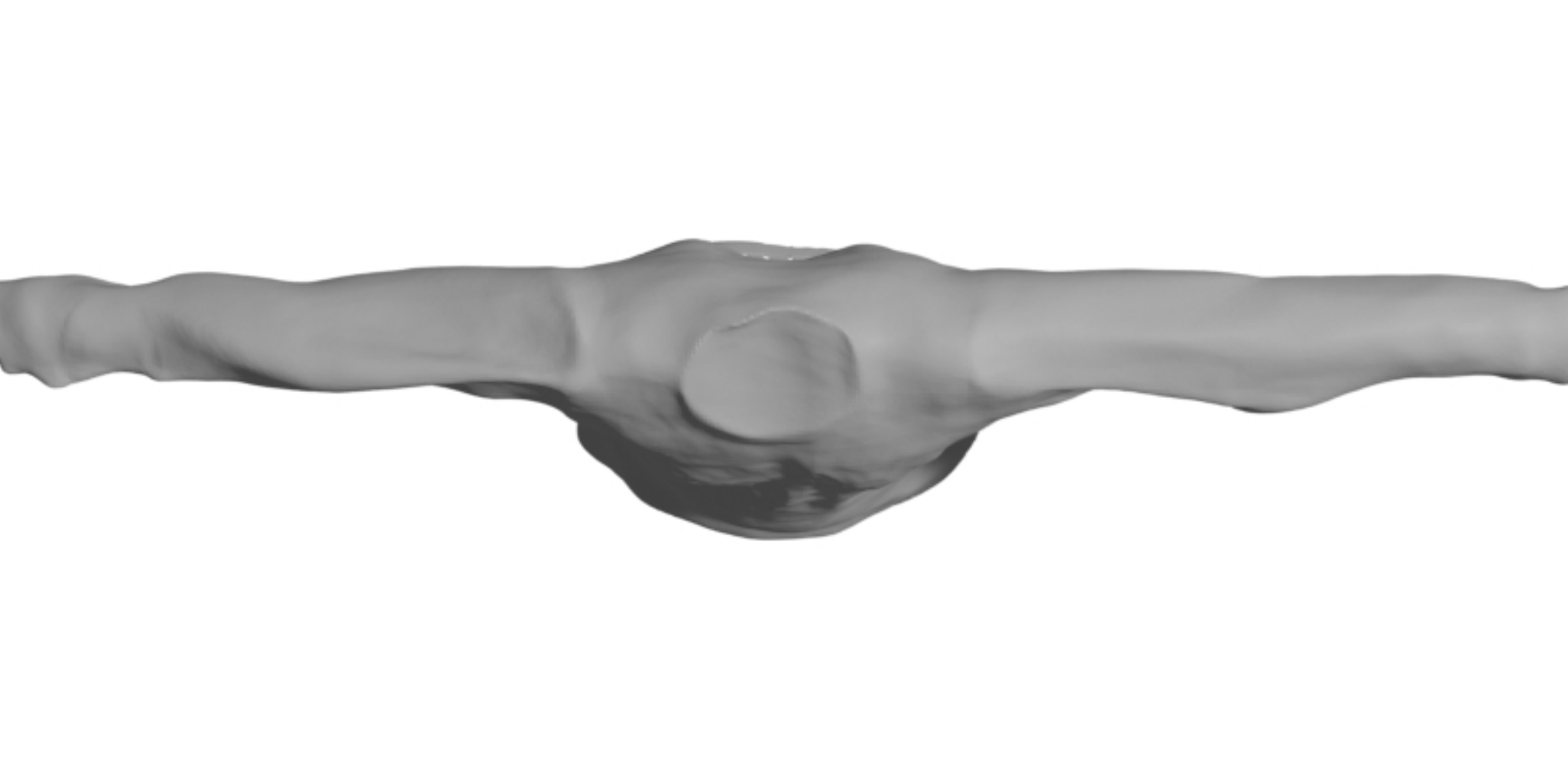}
\end{minipage}
\begin{minipage}[c]{.28\textwidth}
    \centering
    \includegraphics[width=.45\linewidth]{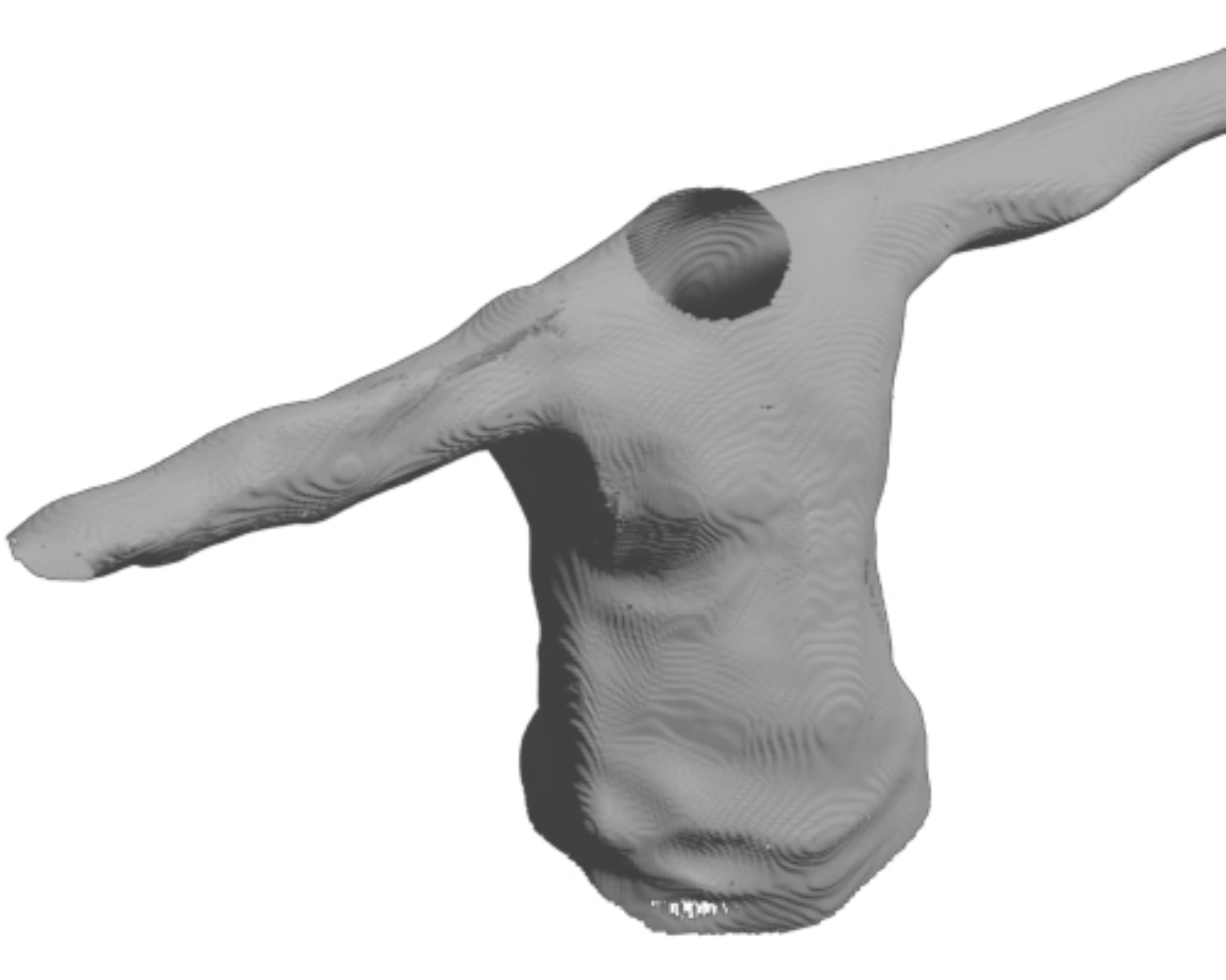}
    \includegraphics[width=.45\linewidth]{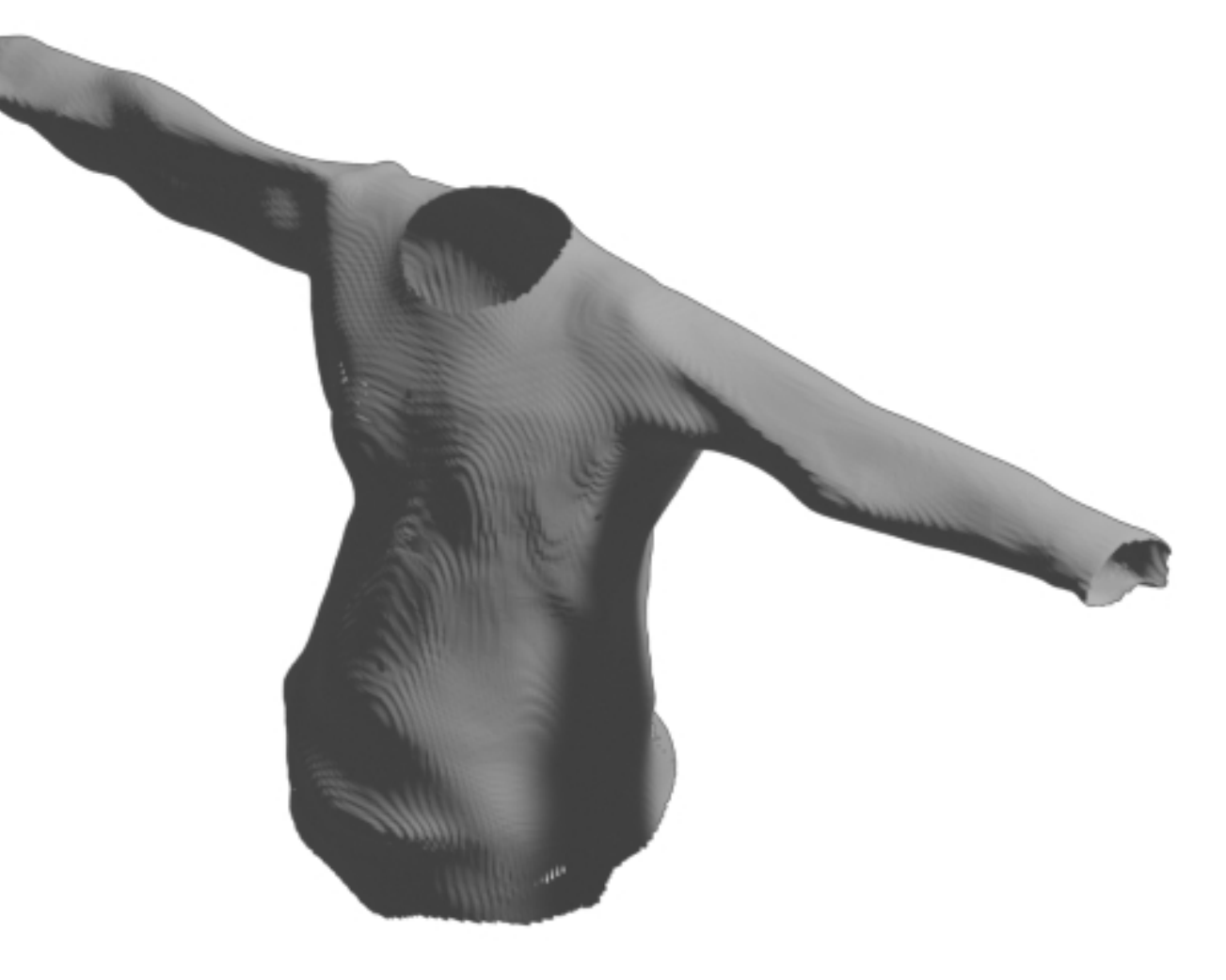}\\
    \includegraphics[width=.6\linewidth]{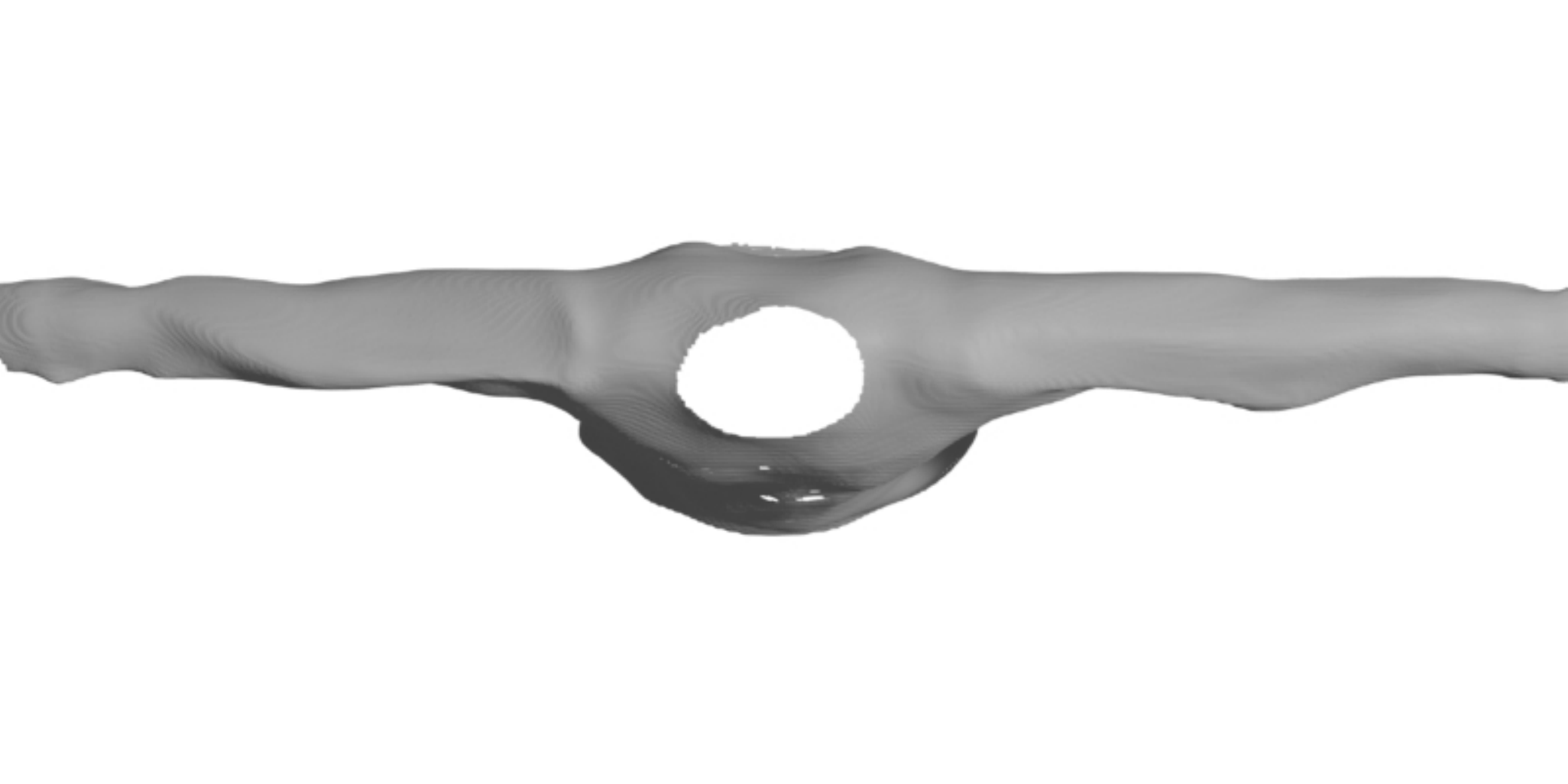}
\end{minipage}
\begin{minipage}[c]{.28\textwidth}
    \centering
    \includegraphics[width=.45\linewidth]{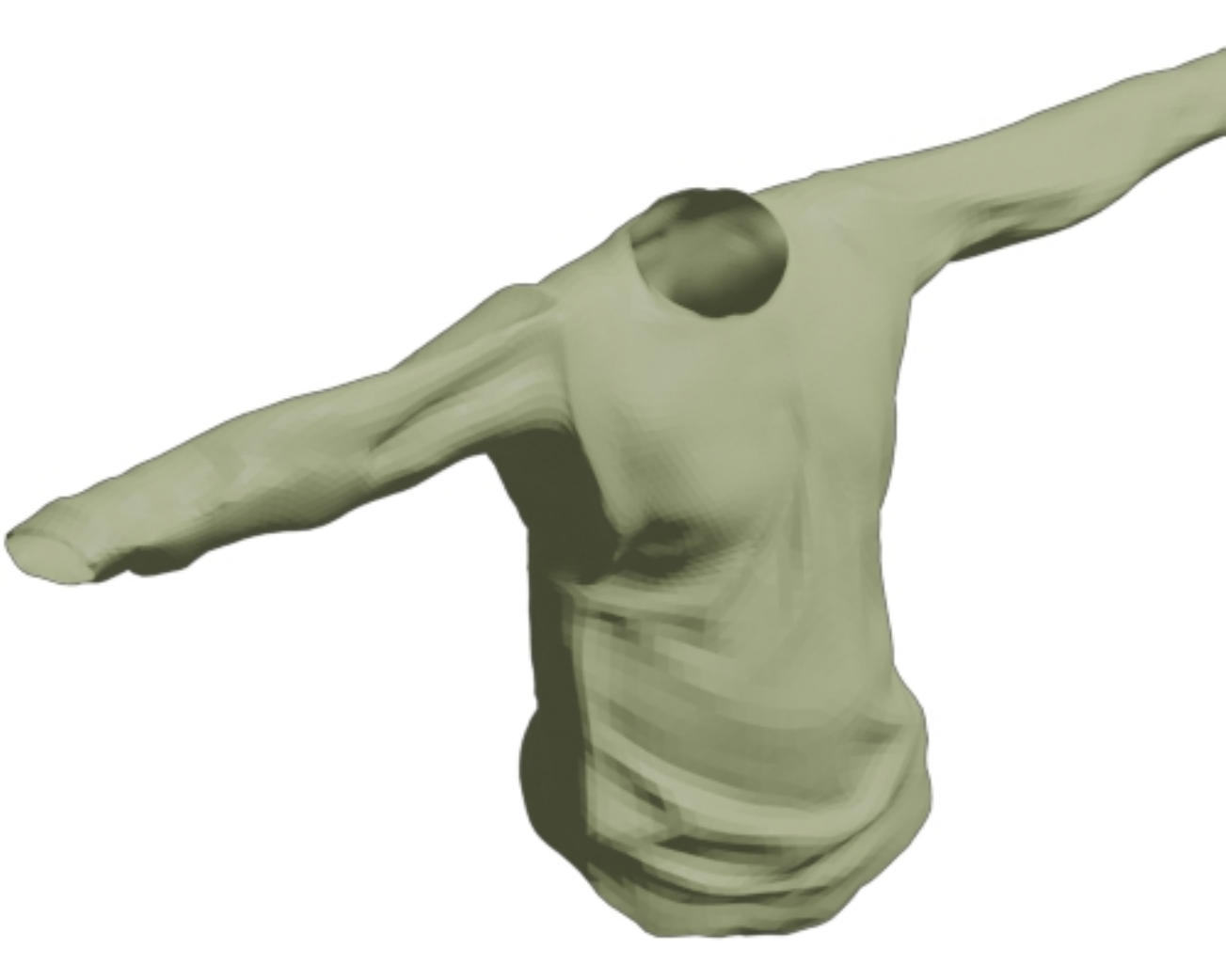}
    \includegraphics[width=.45\linewidth]{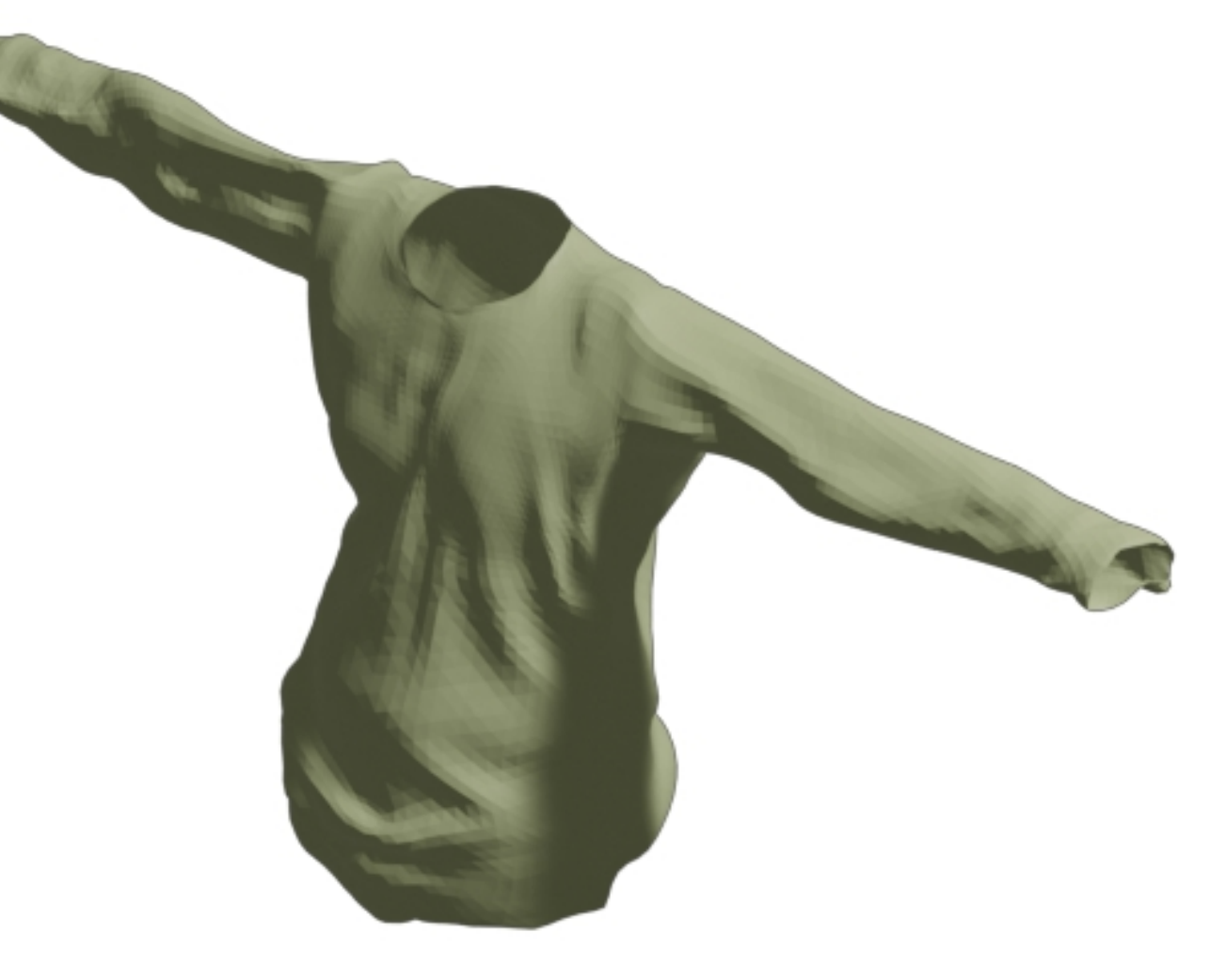}\\
    \includegraphics[width=.6\linewidth]{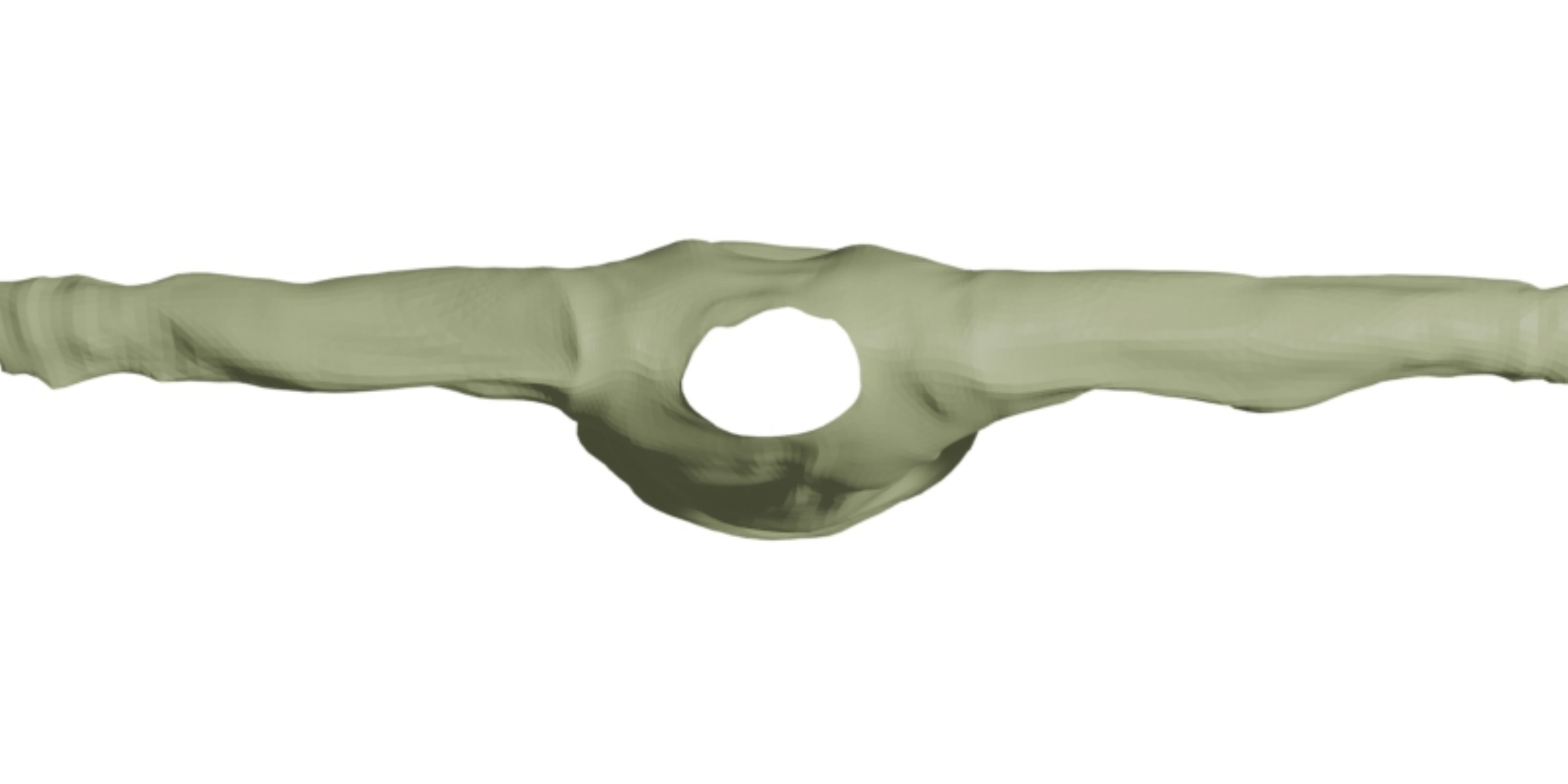}
\end{minipage}

\begin{minipage}[c]{.13\textwidth}
    \centering
    \includegraphics[width=1\linewidth]{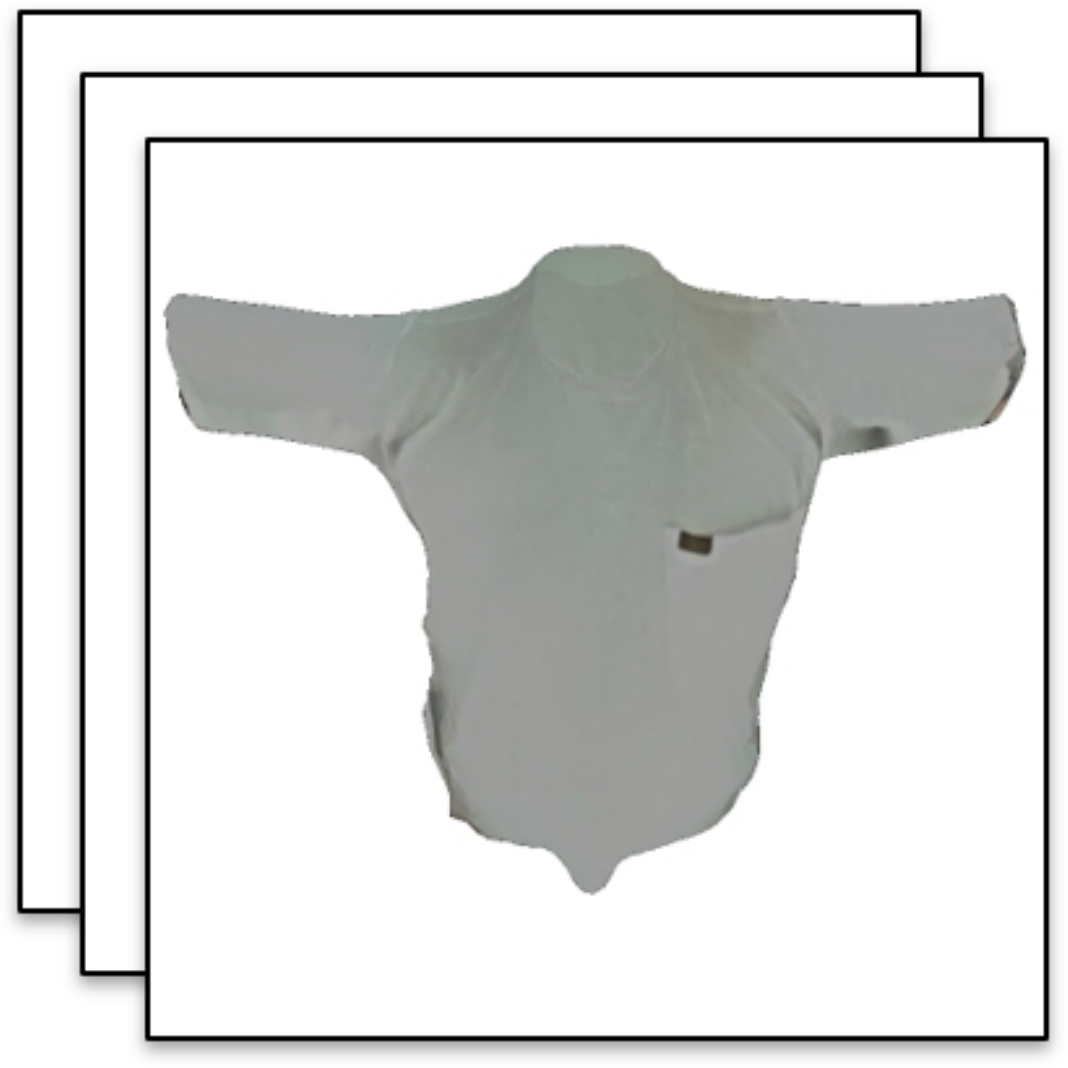}
\end{minipage}
\begin{minipage}[c]{.28\textwidth}
    \centering
    \includegraphics[width=.45\linewidth]{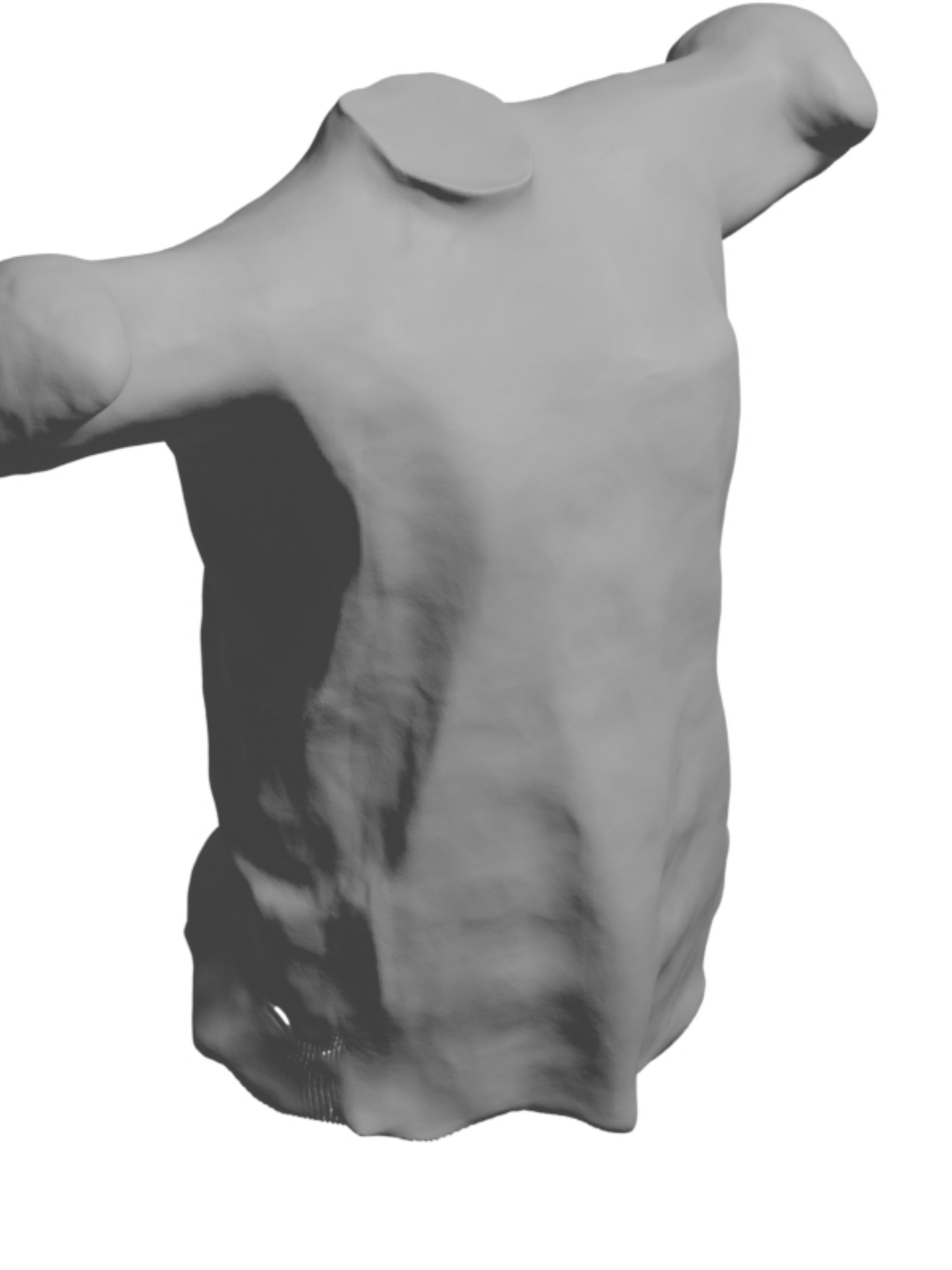}
    \includegraphics[width=.45\linewidth]{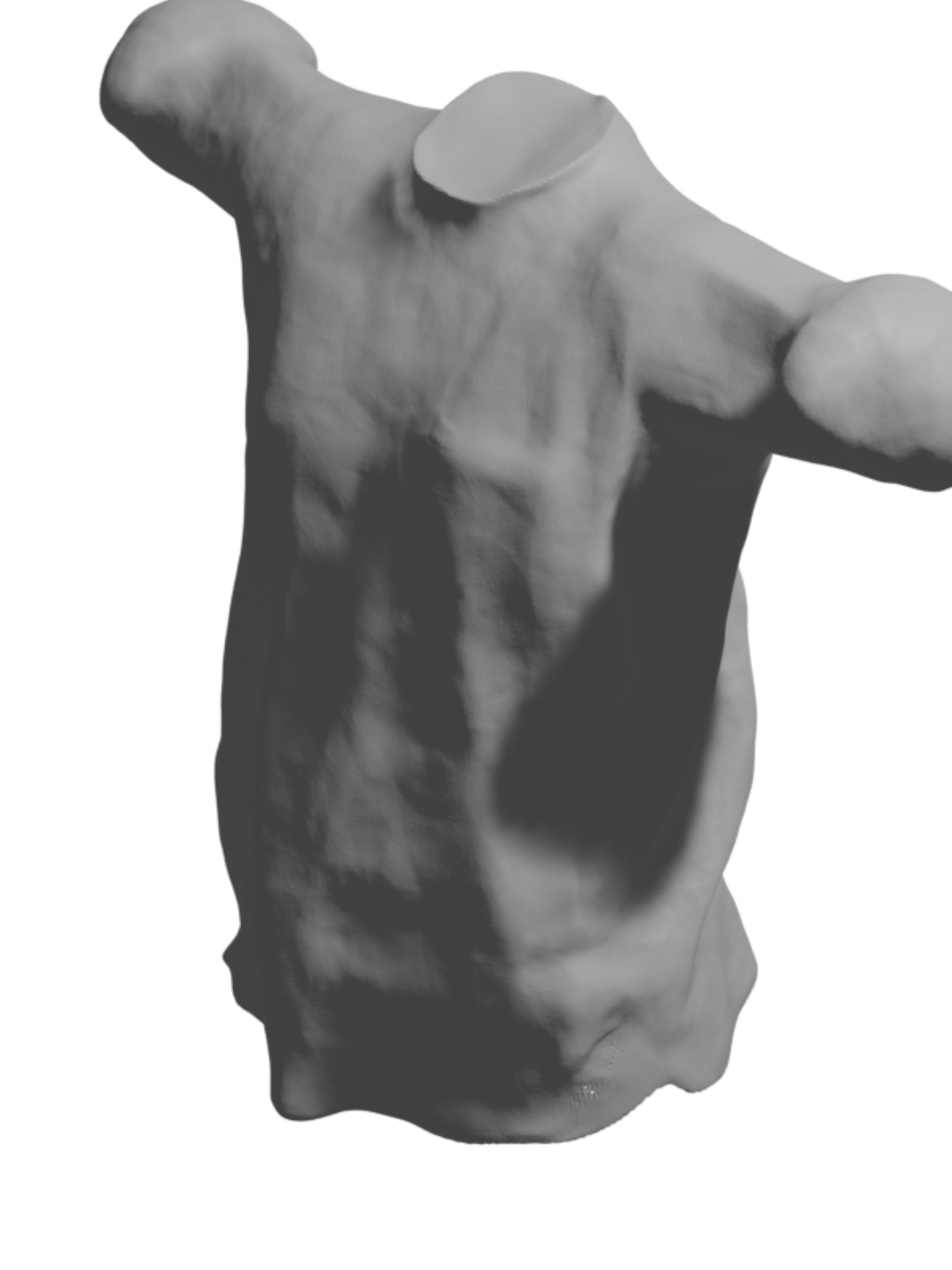}\\
    \includegraphics[width=.45\linewidth]{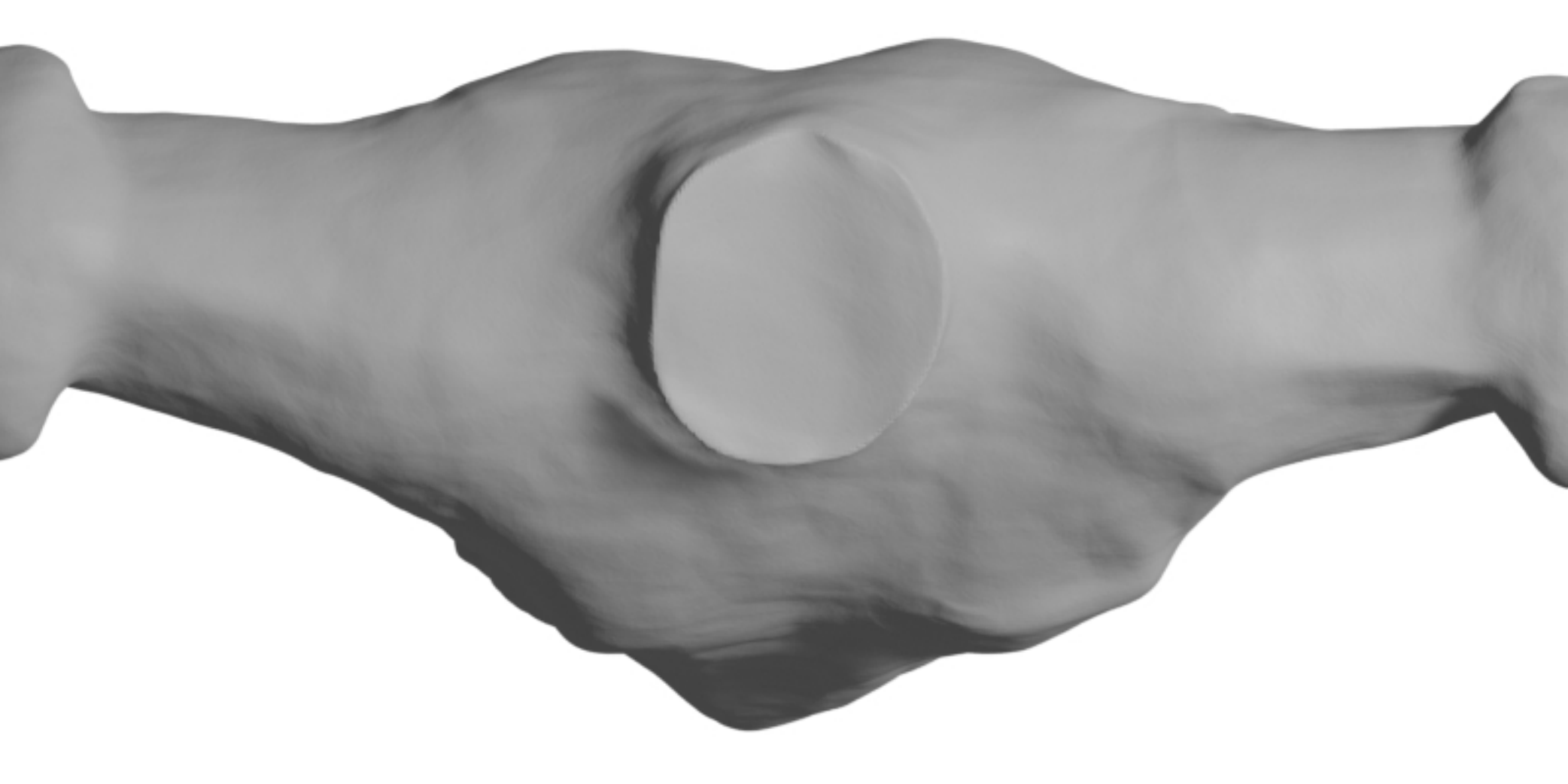}
\end{minipage}
\begin{minipage}[c]{.28\textwidth}
    \centering
    \includegraphics[width=.45\linewidth]{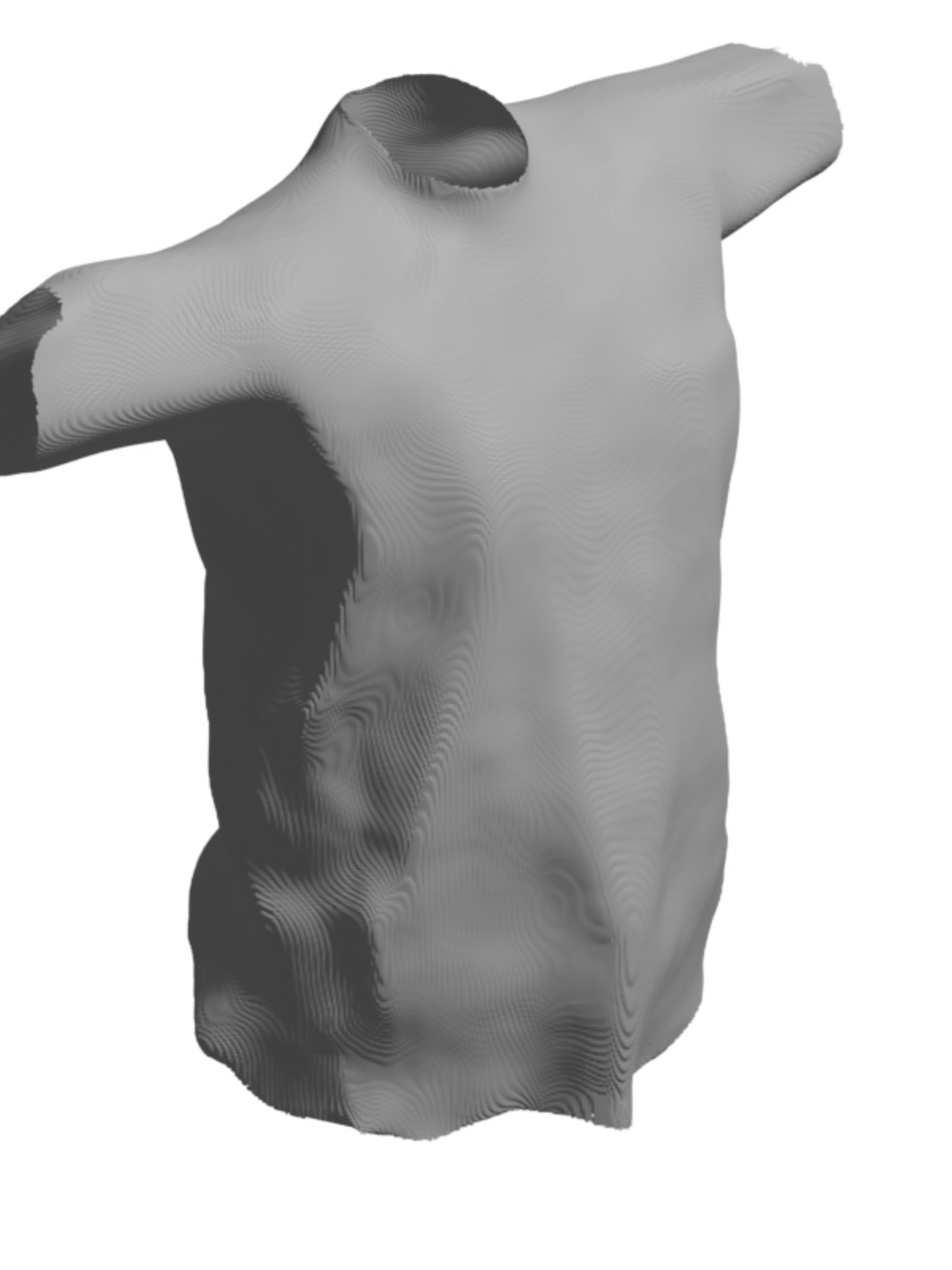}
    \includegraphics[width=.45\linewidth]{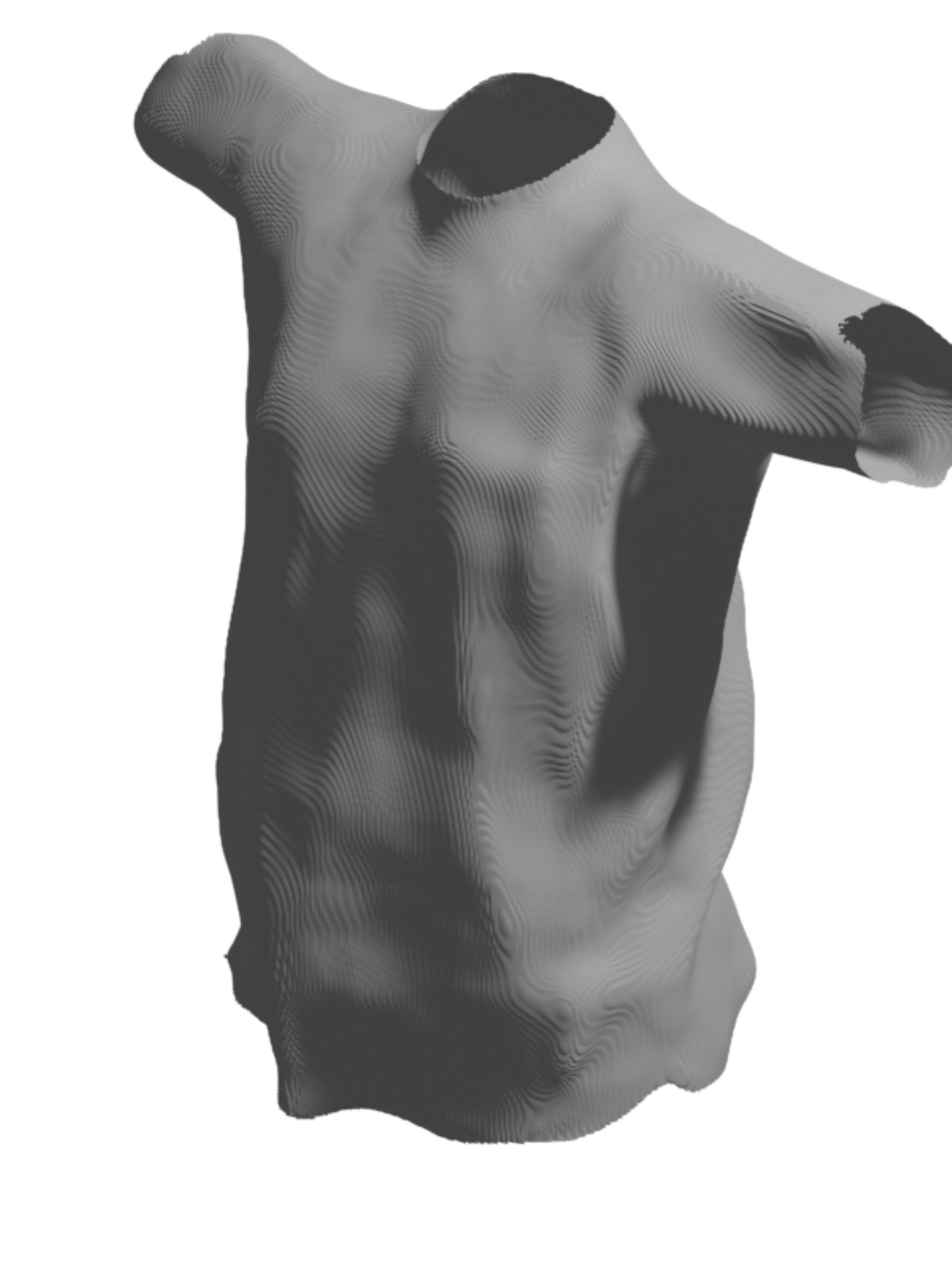}\\
    \includegraphics[width=.45\linewidth]{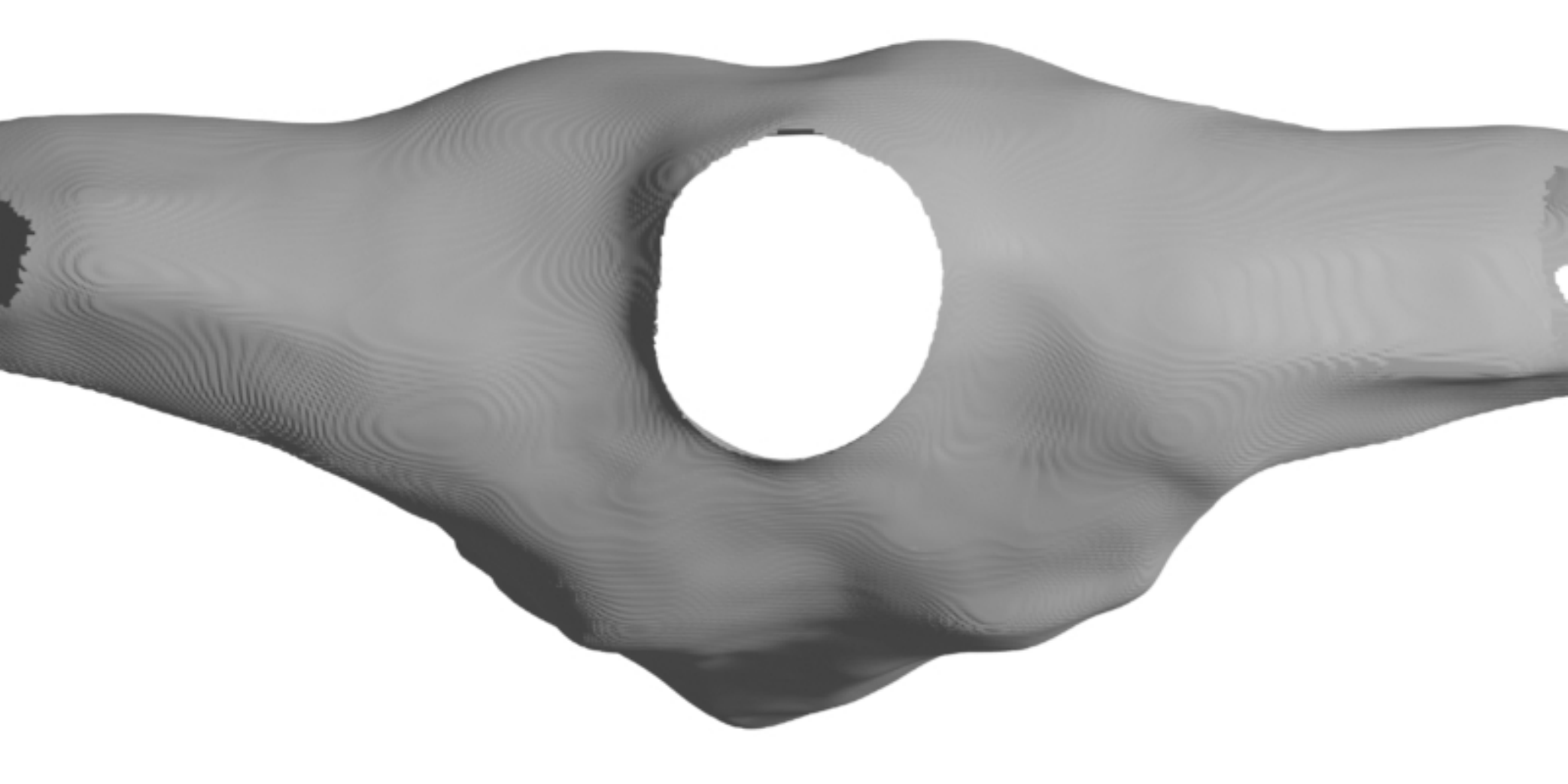}
\end{minipage}
\begin{minipage}[c]{.28\textwidth}
    \centering
    \includegraphics[width=.45\linewidth]{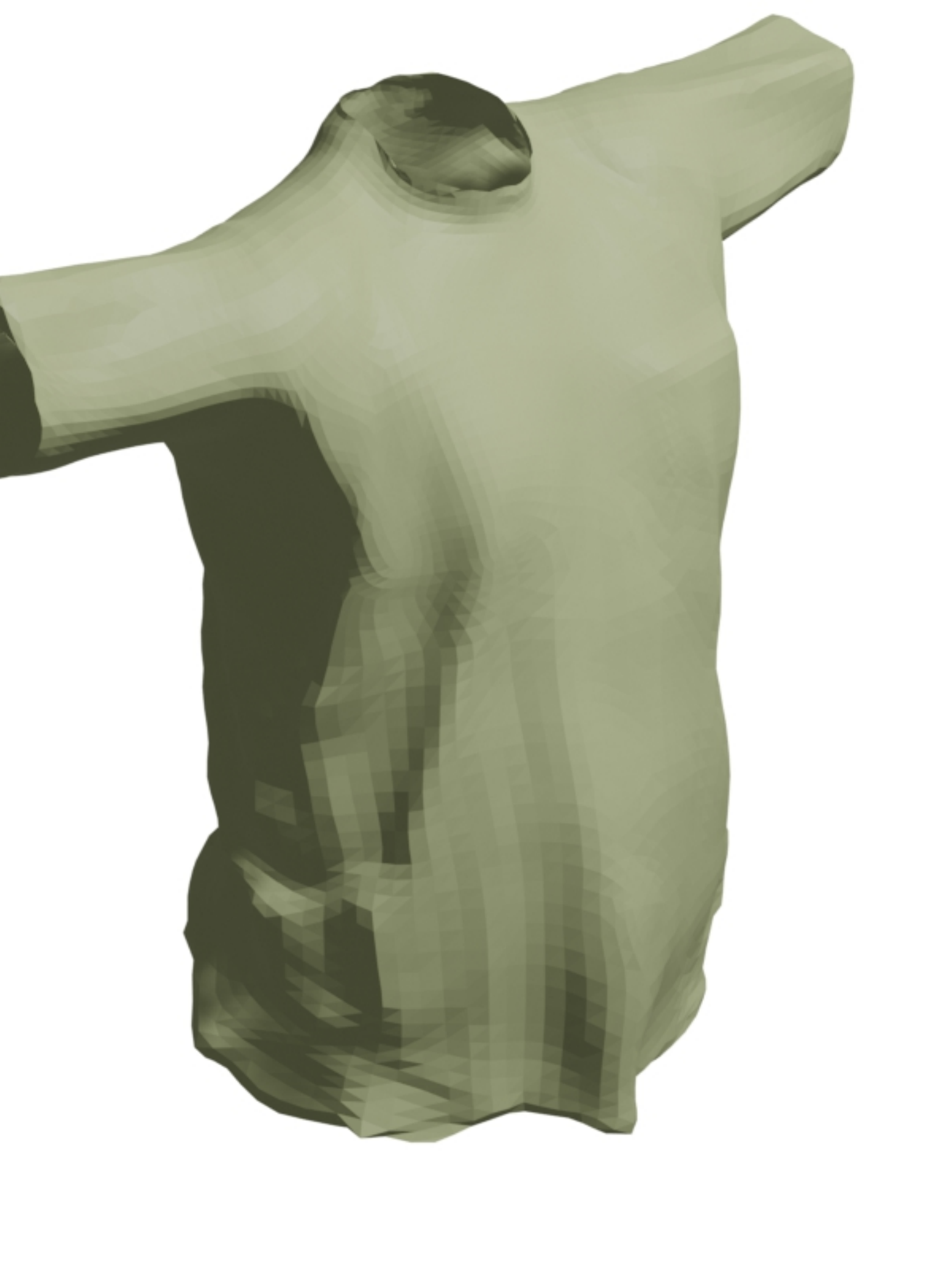}
    \includegraphics[width=.45\linewidth]{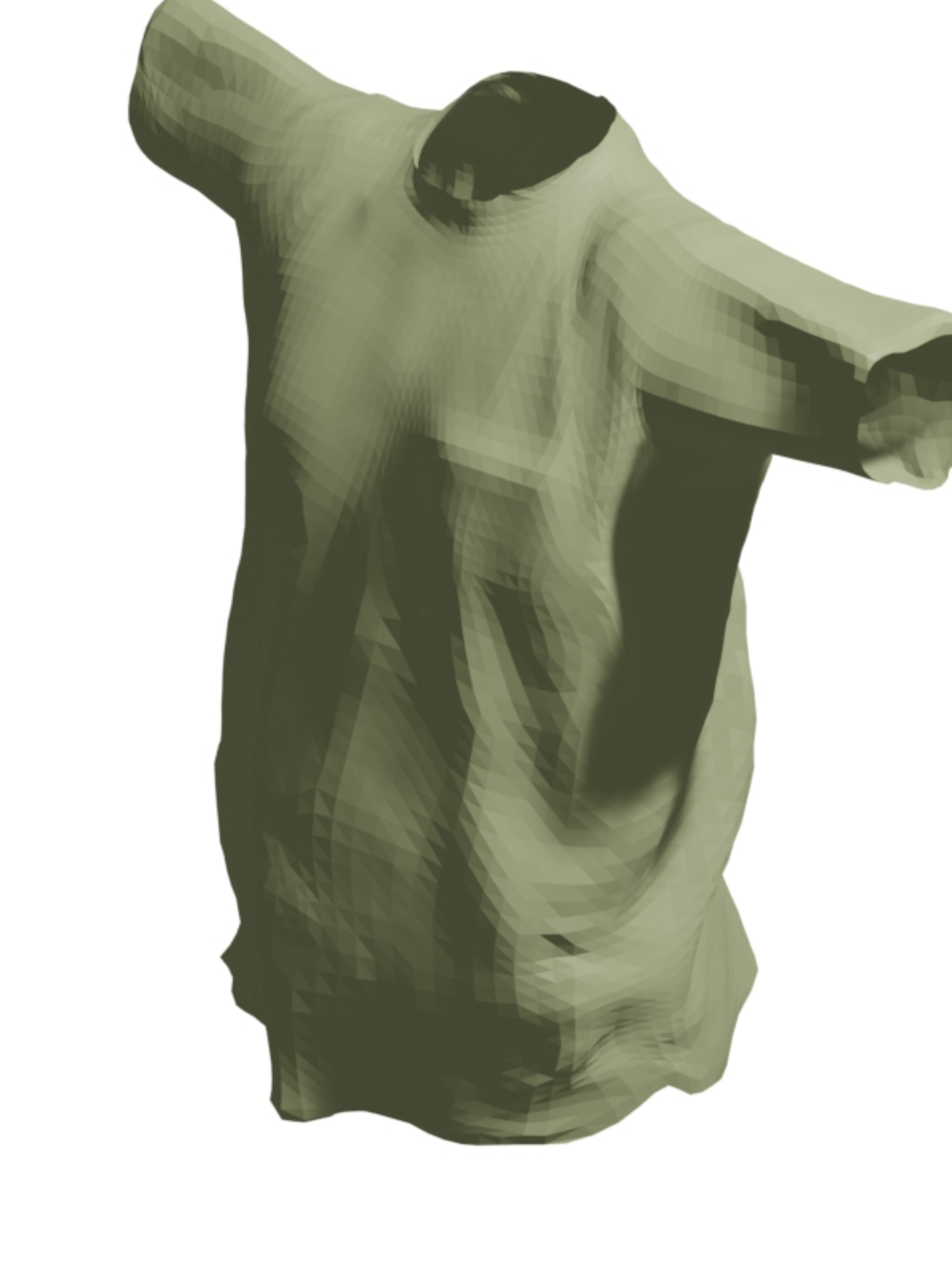}\\
    \includegraphics[width=.45\linewidth]{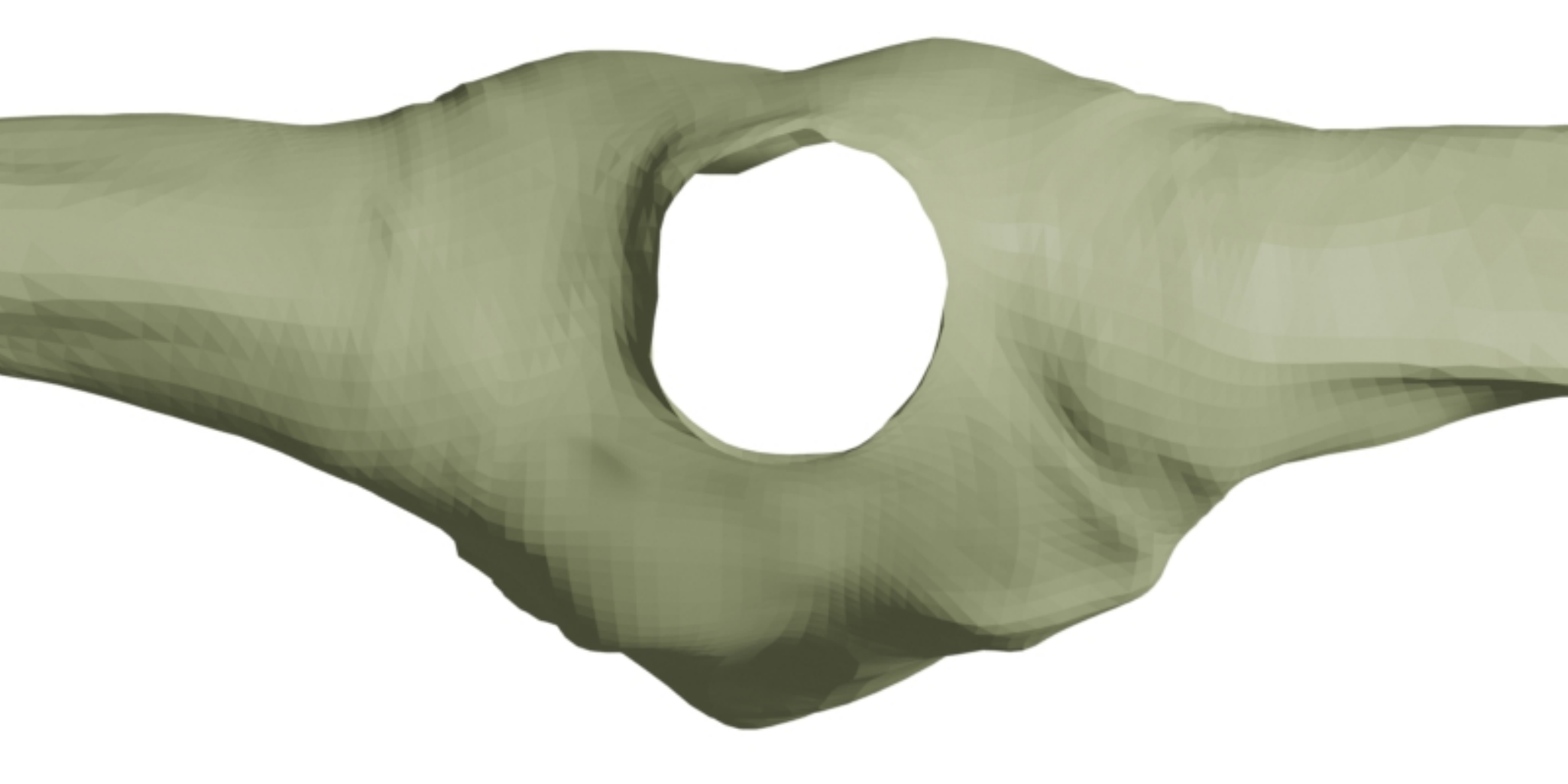}
\end{minipage}

\begin{minipage}[c]{.13\textwidth}
    \centering
    Input
\end{minipage}
\begin{minipage}[c]{.28\textwidth}
    \centering
    NeuS
\end{minipage}
\begin{minipage}[c]{.28\textwidth}
    \centering
    Ours
\end{minipage}
\begin{minipage}[c]{.28\textwidth}
    \centering
    Ground-truth
\end{minipage}

\caption{Additional results on the MGN~\cite{mgn} dataset without mask supervision.}
\label{supp_mgn_wo}
\end{figure*}

\begin{figure*}[h]

\begin{minipage}[c]{.14\textwidth}
    \centering
    \includegraphics[width=1\linewidth]{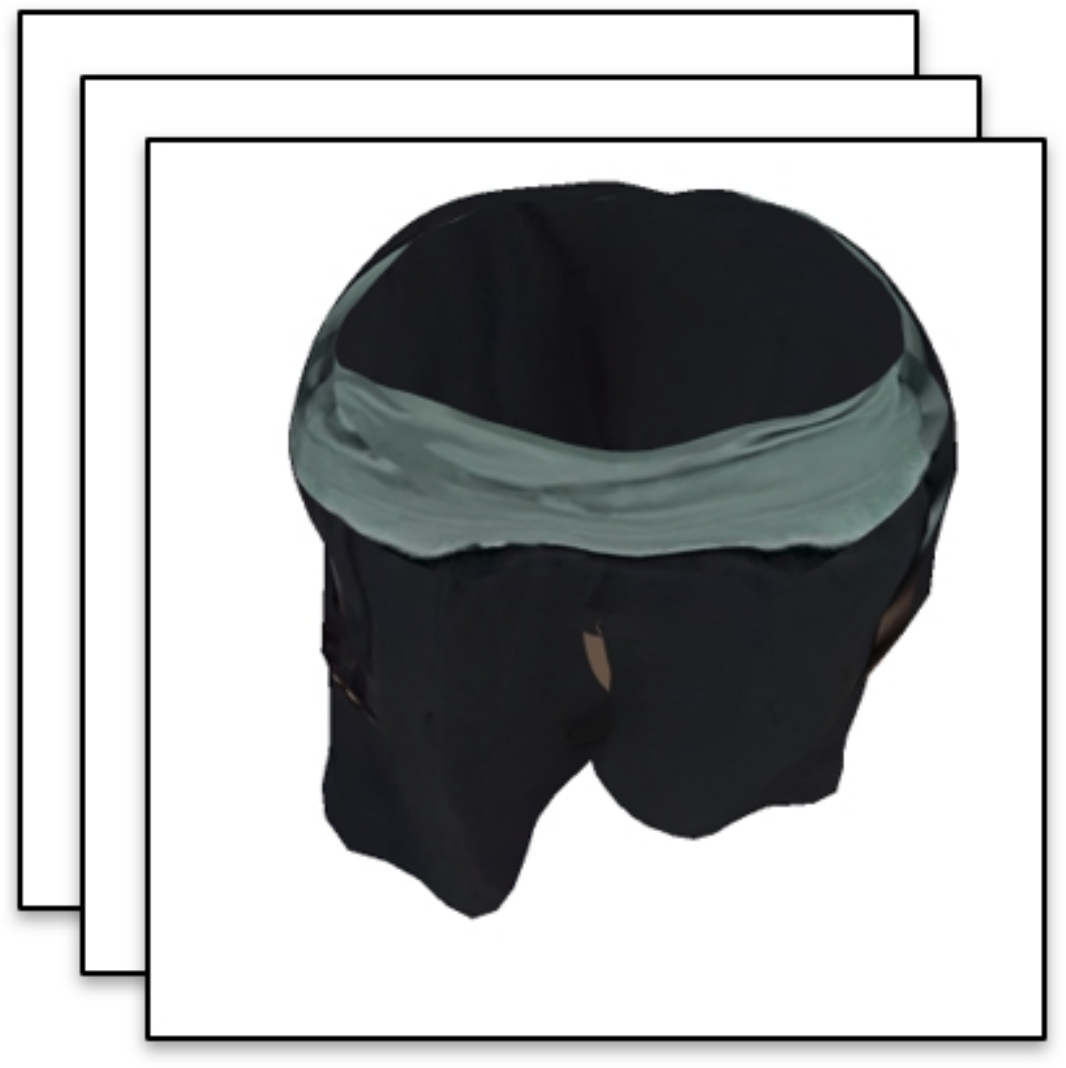}
\end{minipage}
\begin{minipage}[c]{.28\textwidth}
    \centering
    \includegraphics[width=.45\linewidth]{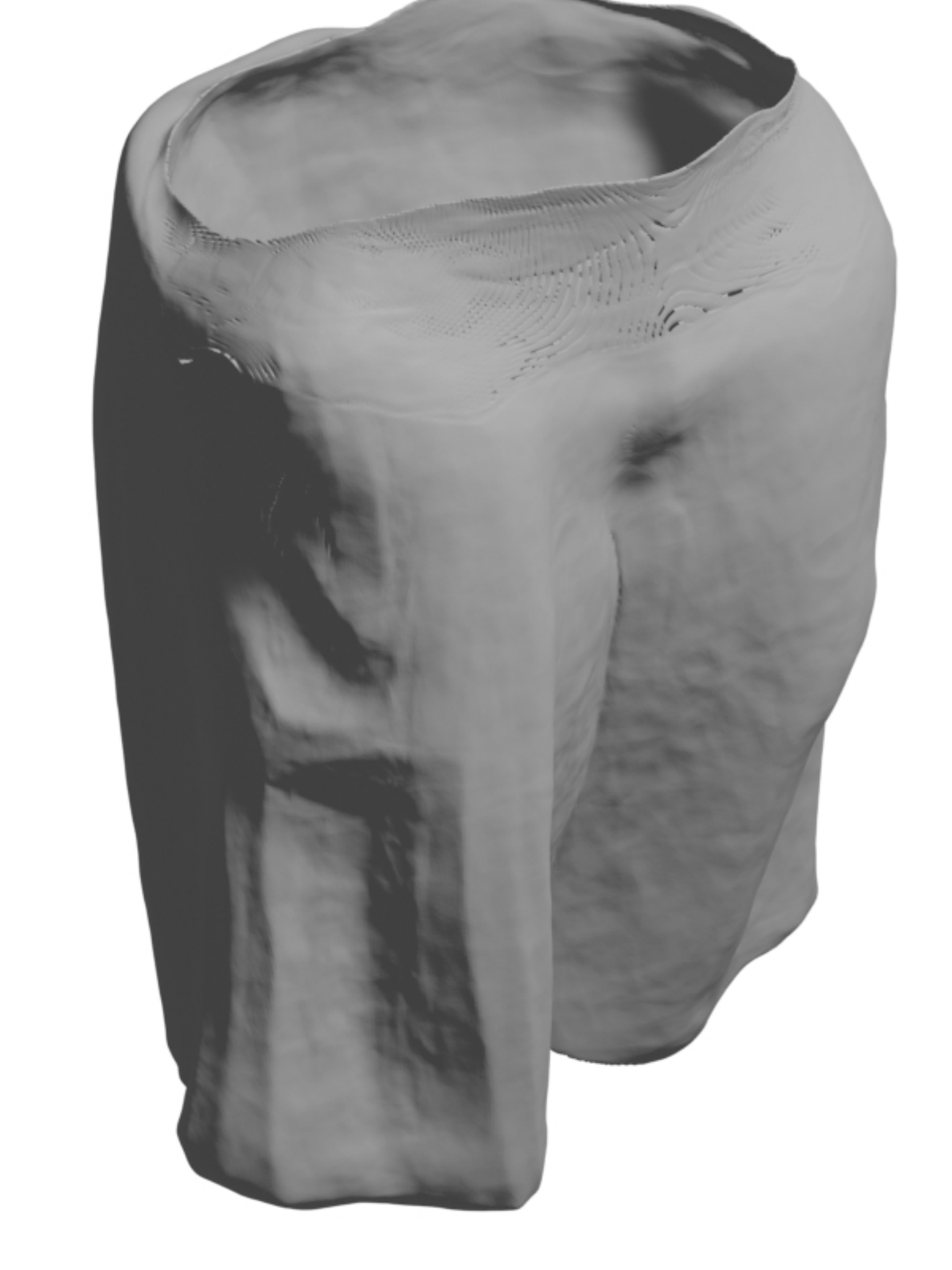}
    \includegraphics[width=.45\linewidth]{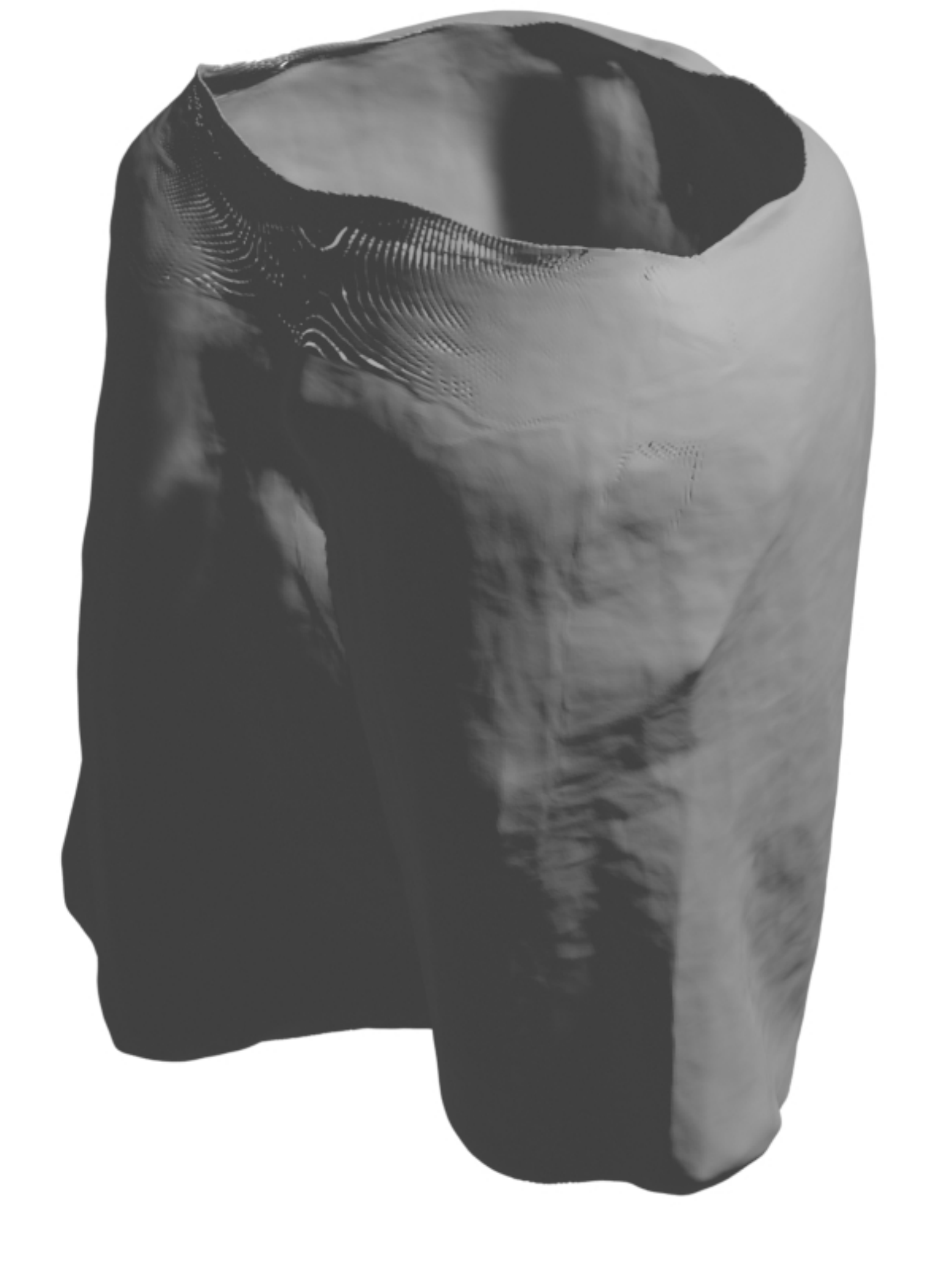}\\
    \includegraphics[width=.45\linewidth]{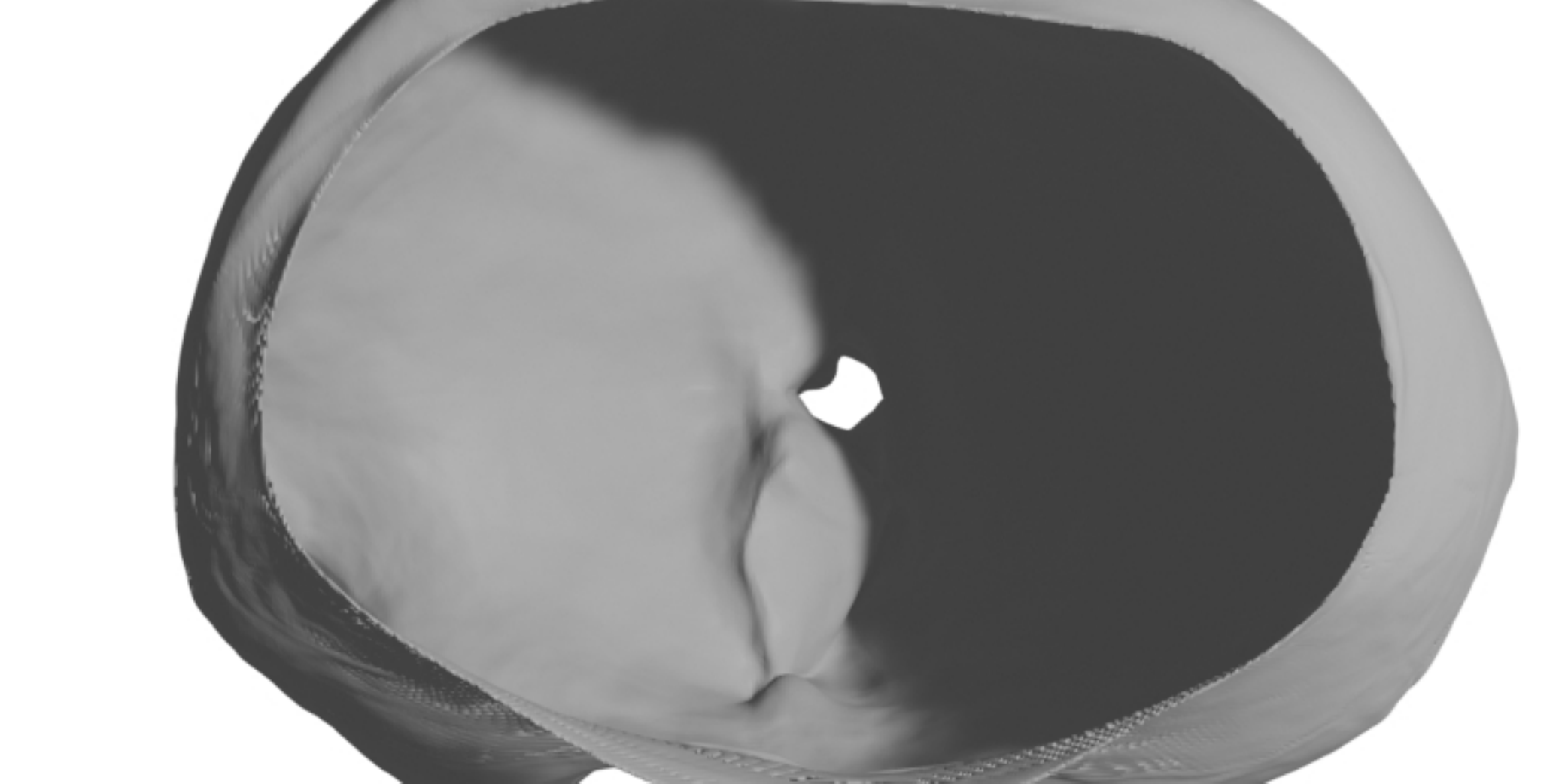}
\end{minipage}
\begin{minipage}[c]{.28\textwidth}
    \centering
    \includegraphics[width=.45\linewidth]{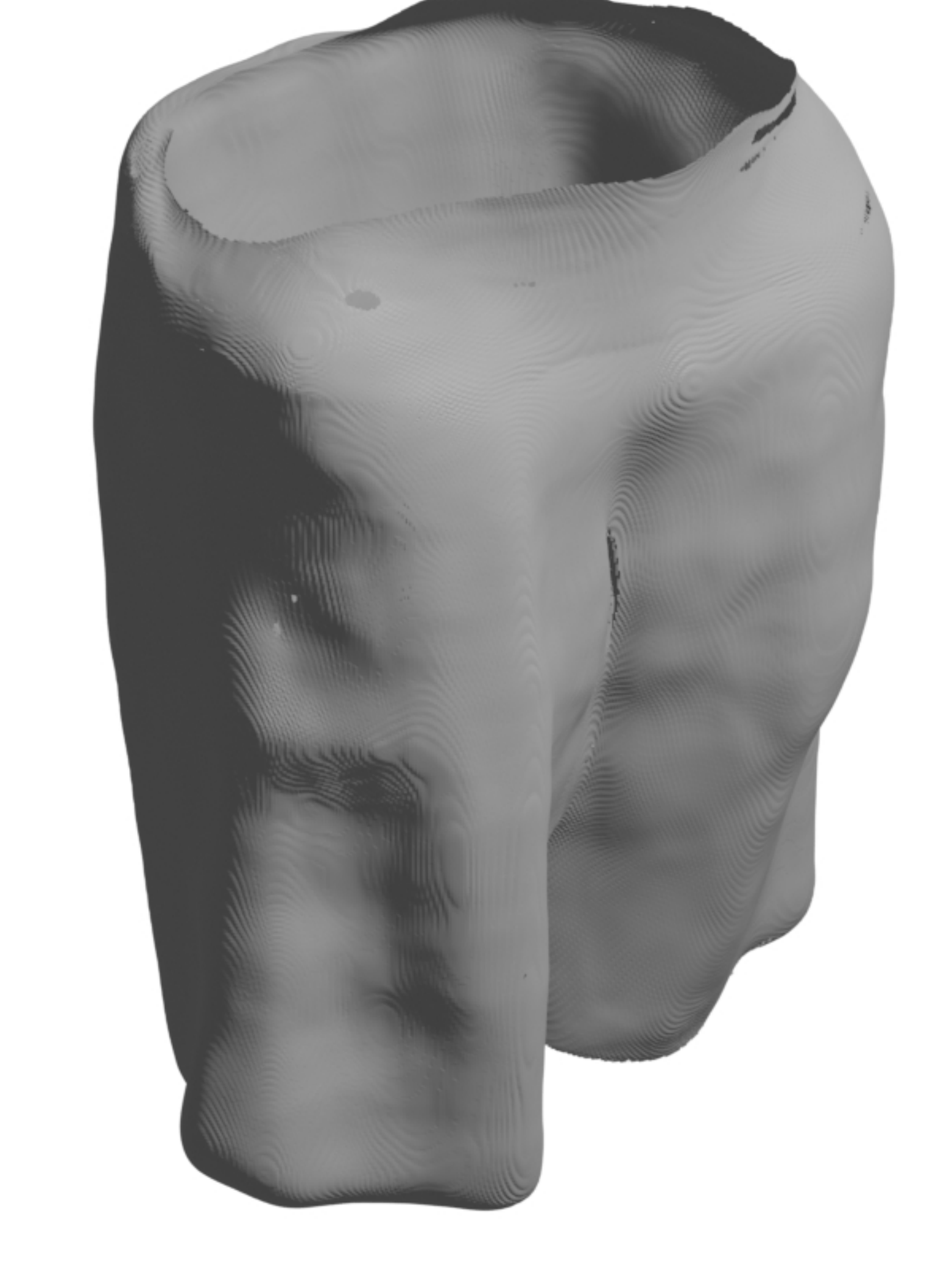}
    \includegraphics[width=.45\linewidth]{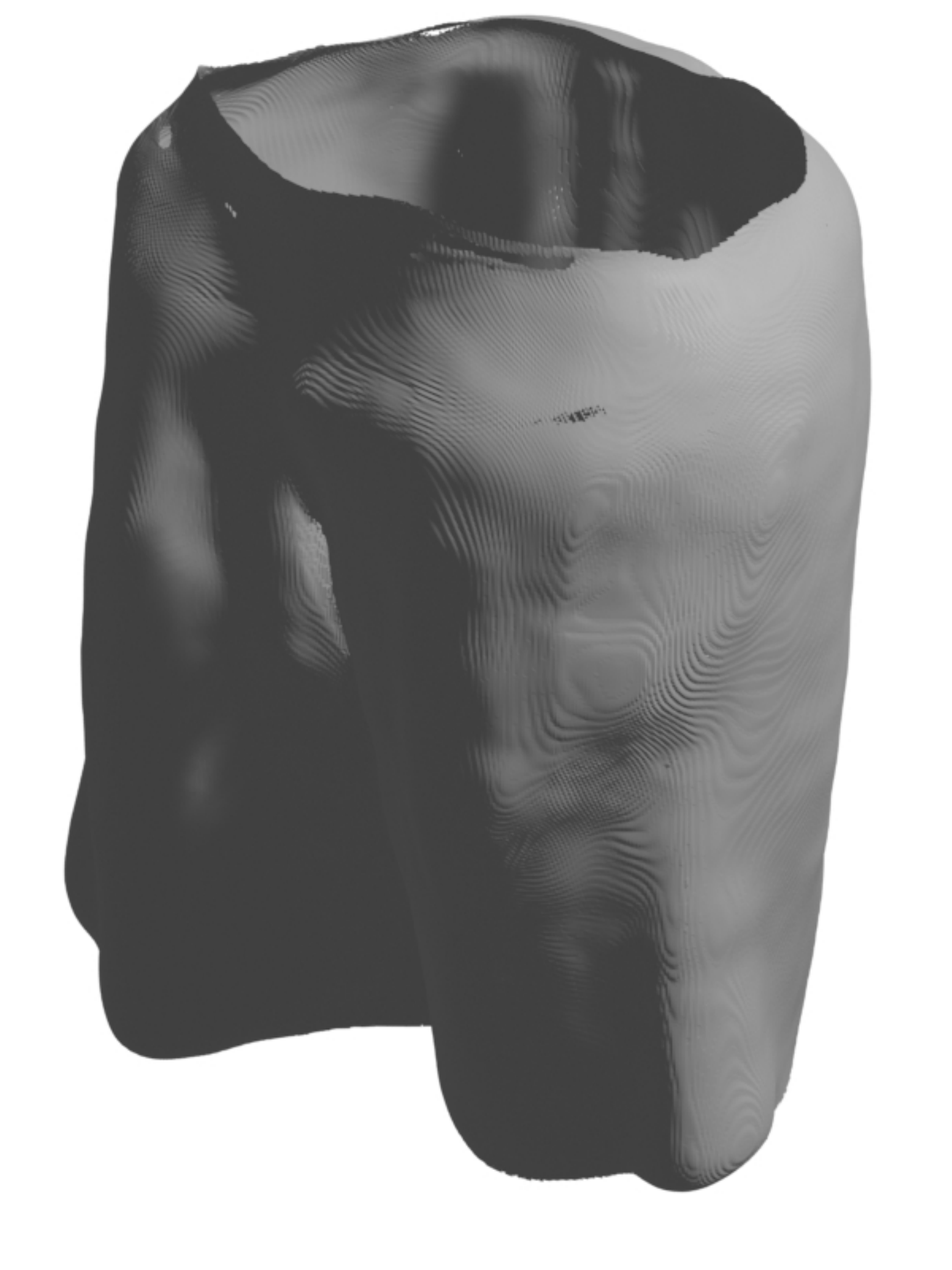}\\
    \includegraphics[width=.45\linewidth]{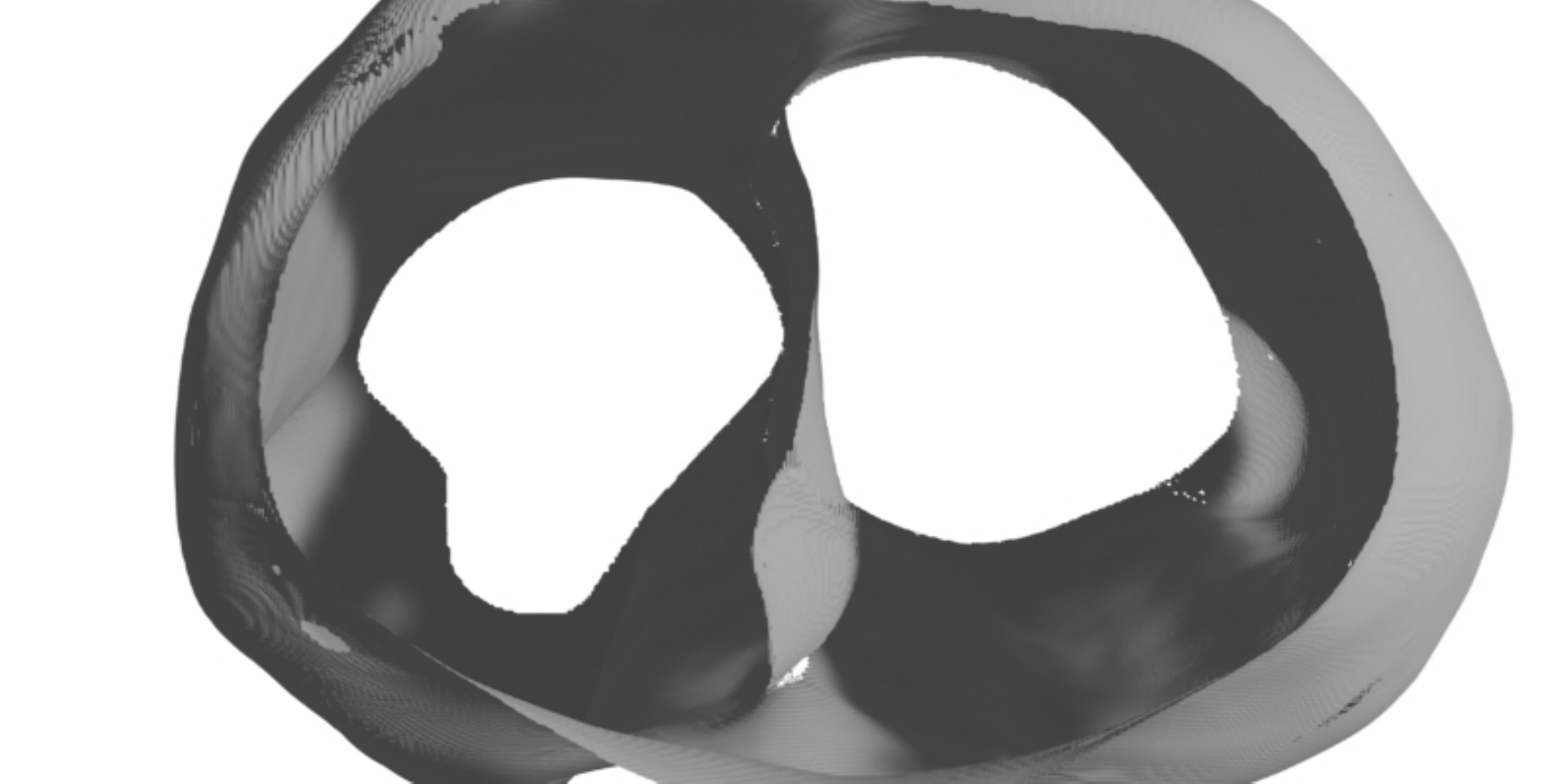}
\end{minipage}
\begin{minipage}[c]{.28\textwidth}
    \centering
    \includegraphics[width=.45\linewidth]{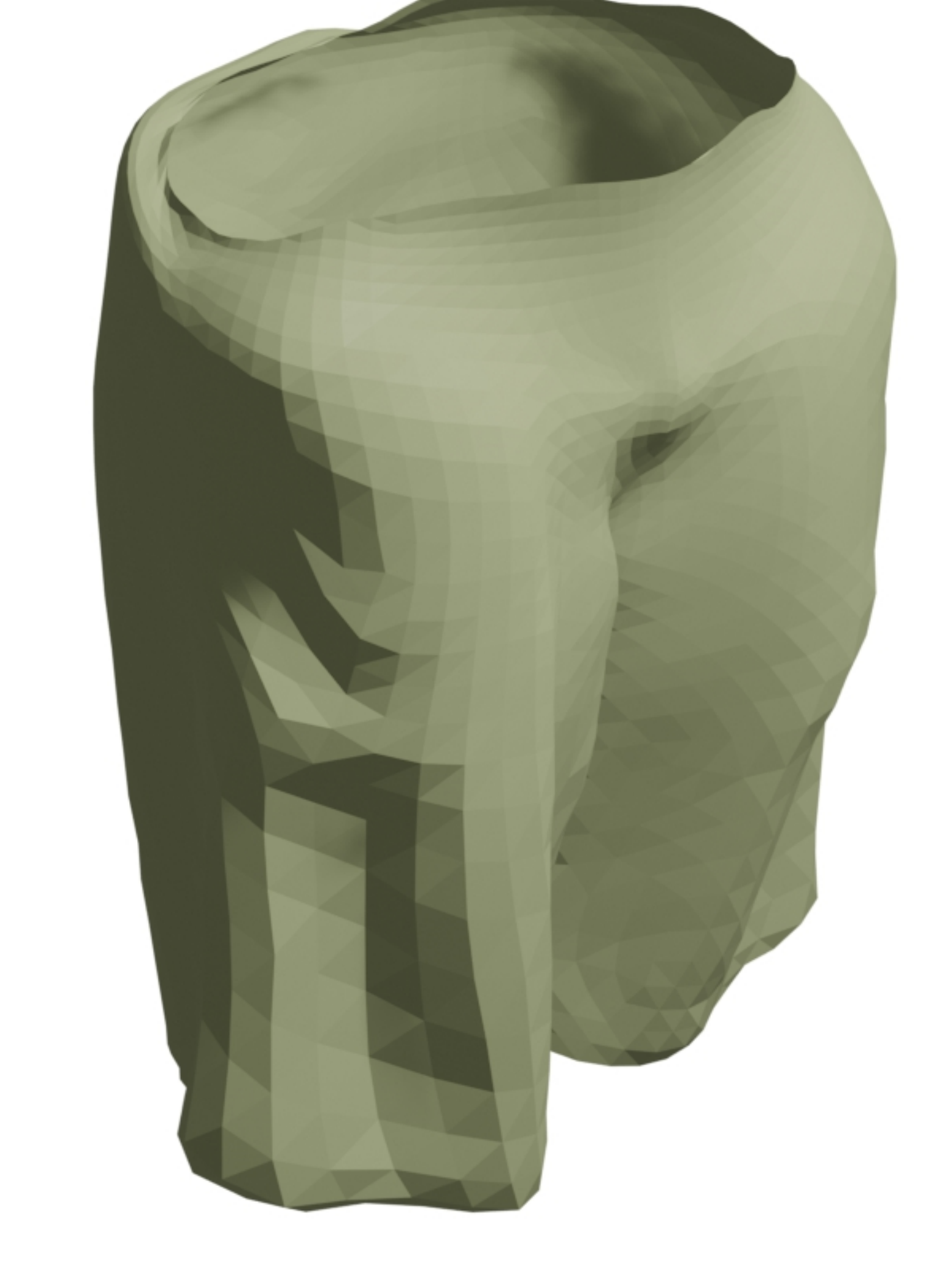}
    \includegraphics[width=.45\linewidth]{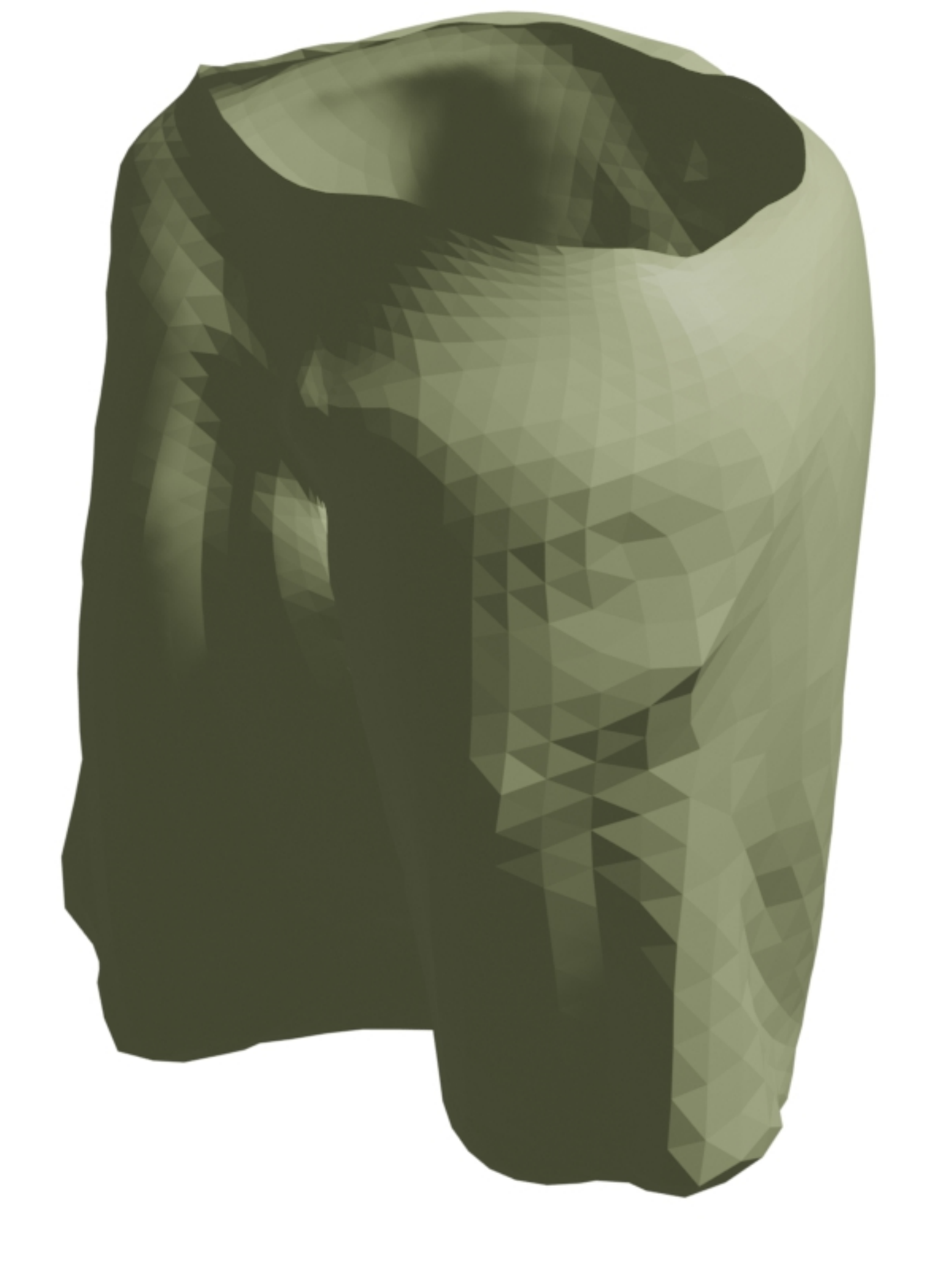}\\
    \includegraphics[width=.45\linewidth]{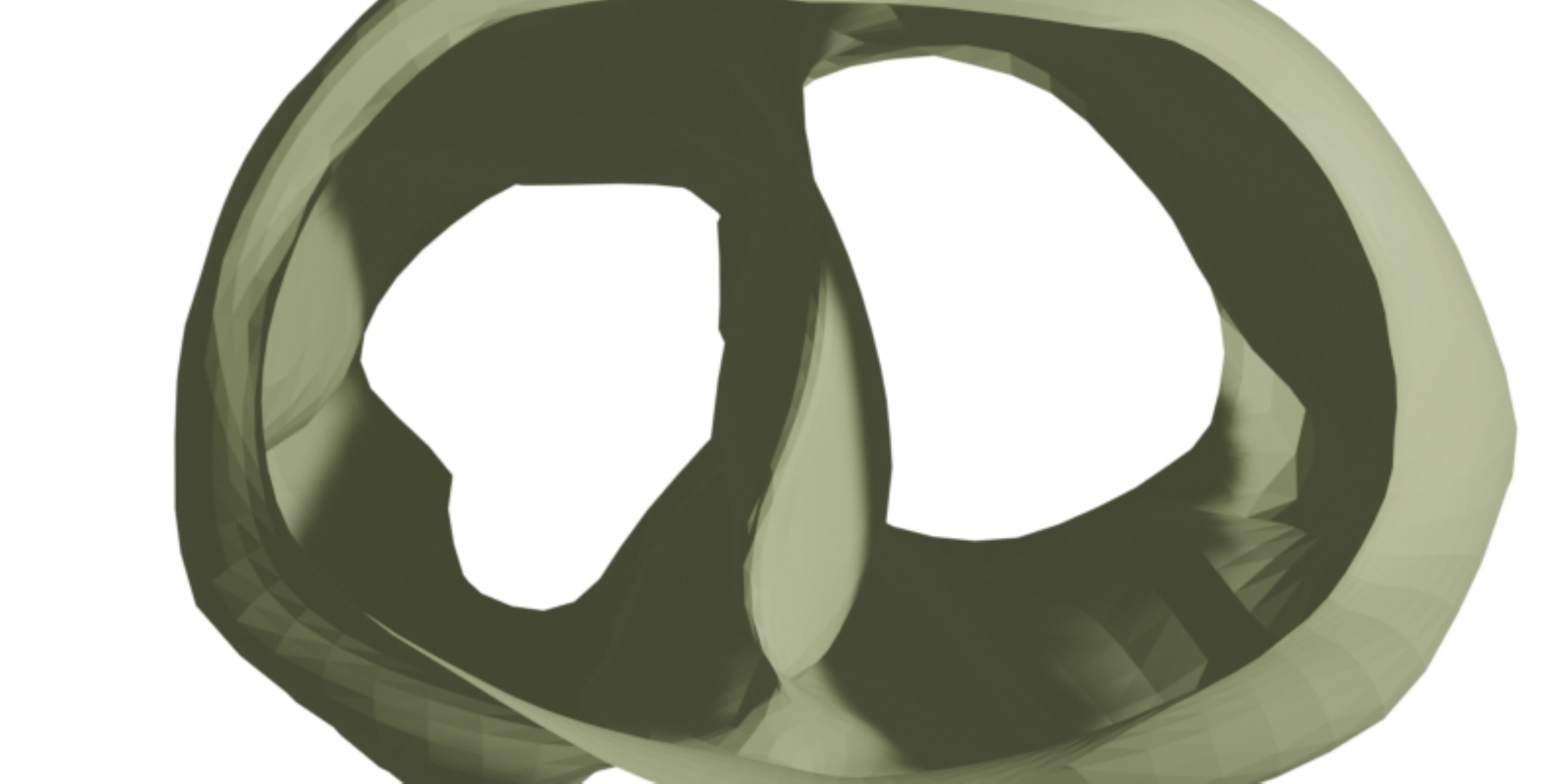}
\end{minipage}

\begin{minipage}[c]{.14\textwidth}
    \centering
    \includegraphics[width=1\linewidth]{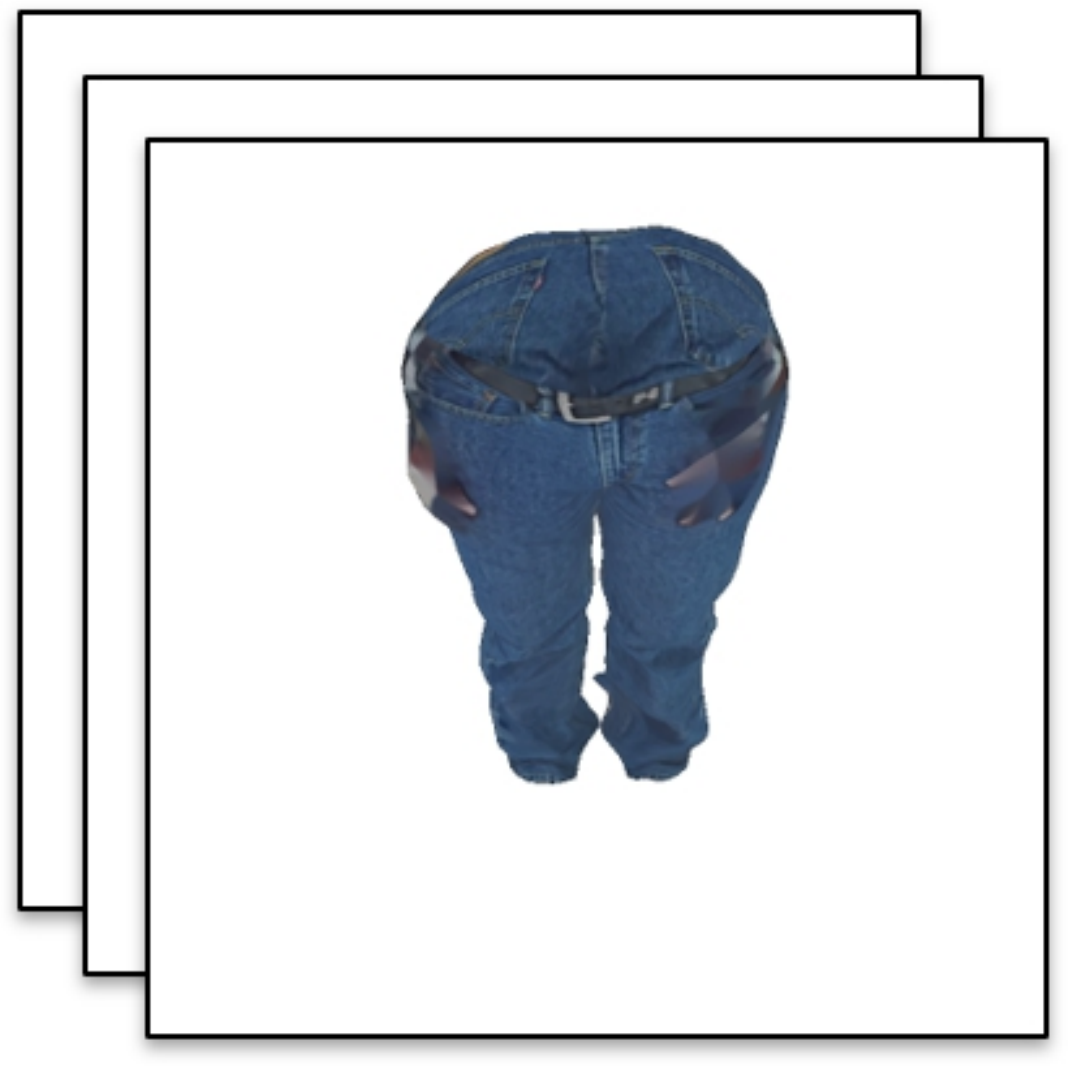}
\end{minipage}
\begin{minipage}[c]{.28\textwidth}
    \centering
    \includegraphics[width=.45\linewidth]{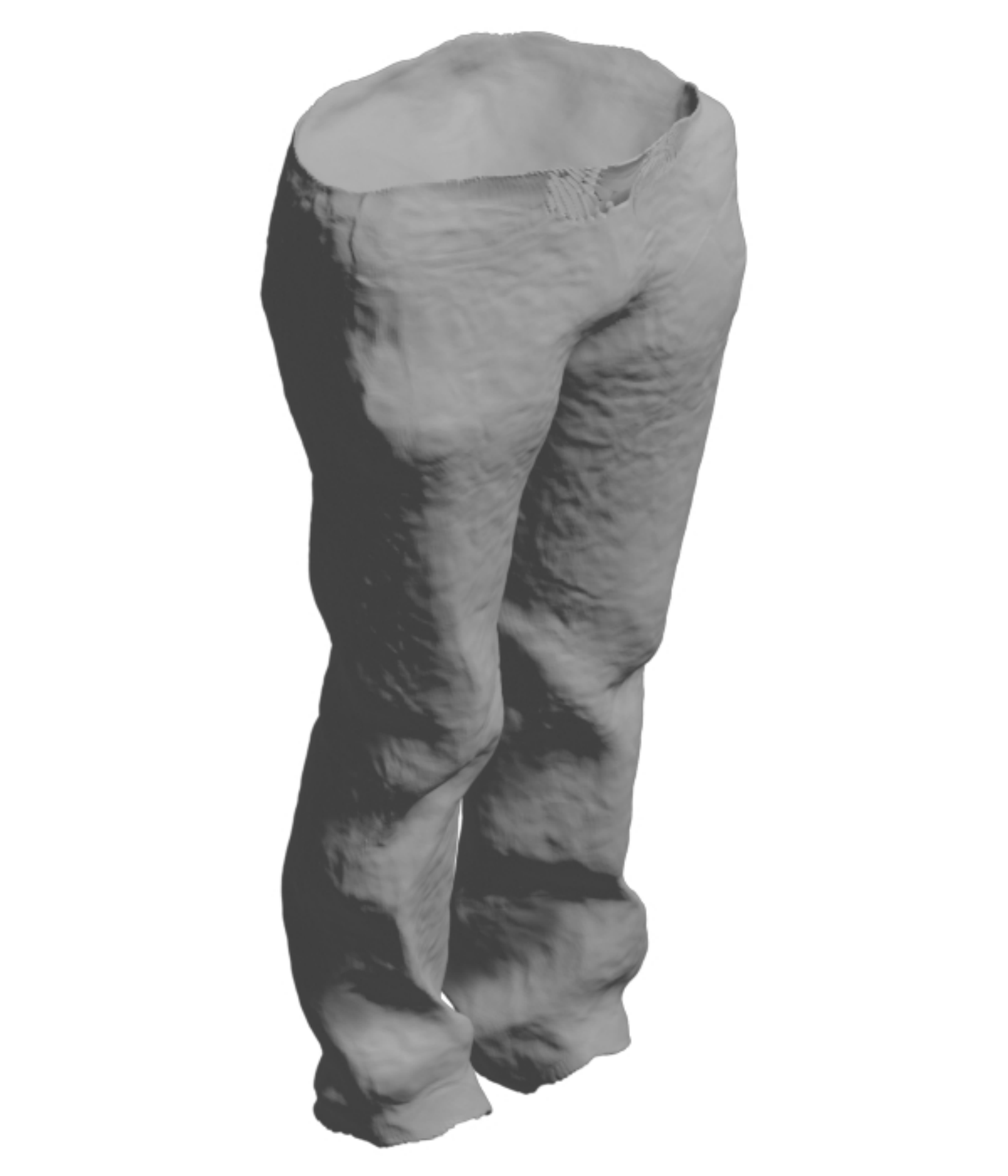}
    \includegraphics[width=.45\linewidth]{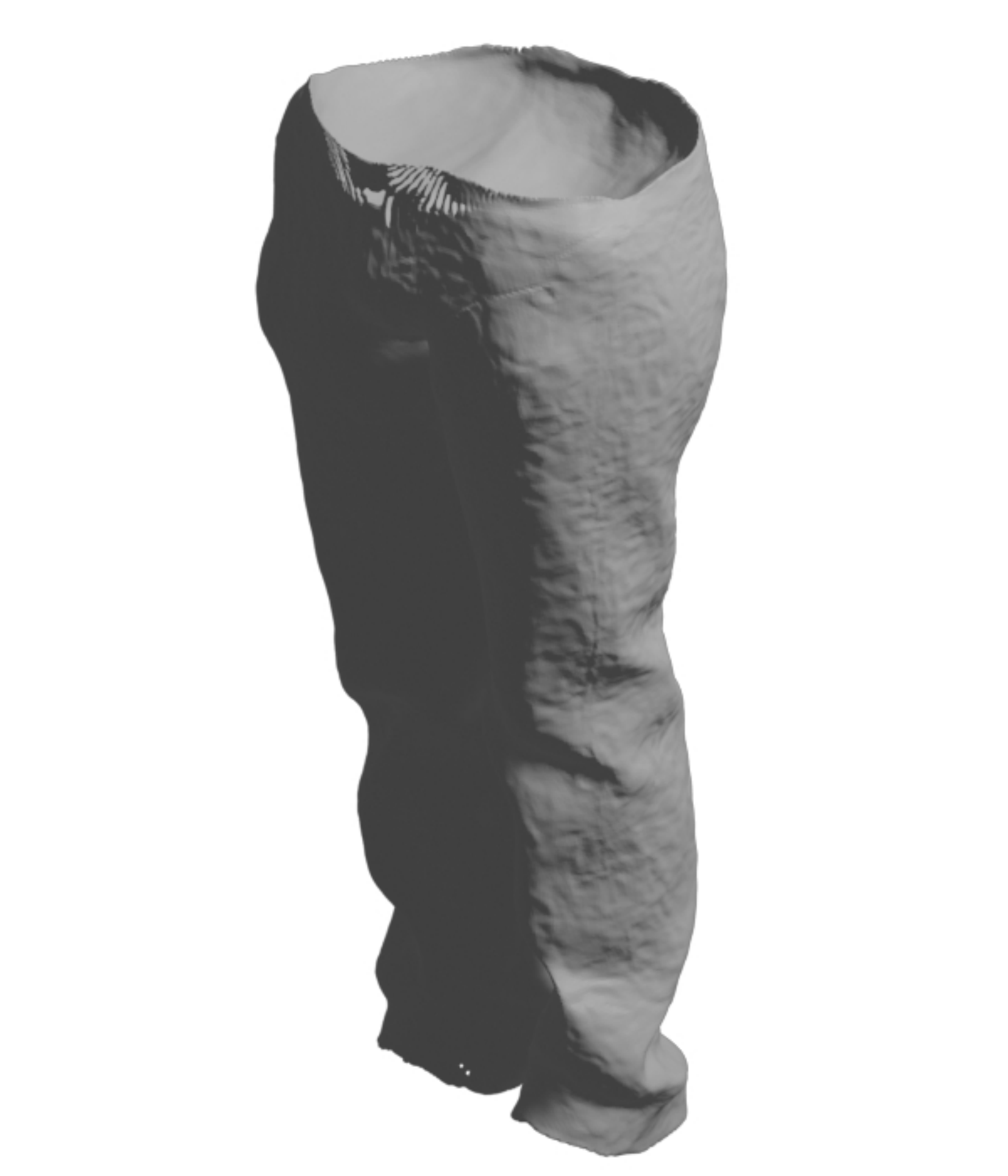}\\
    \includegraphics[width=.6\linewidth]{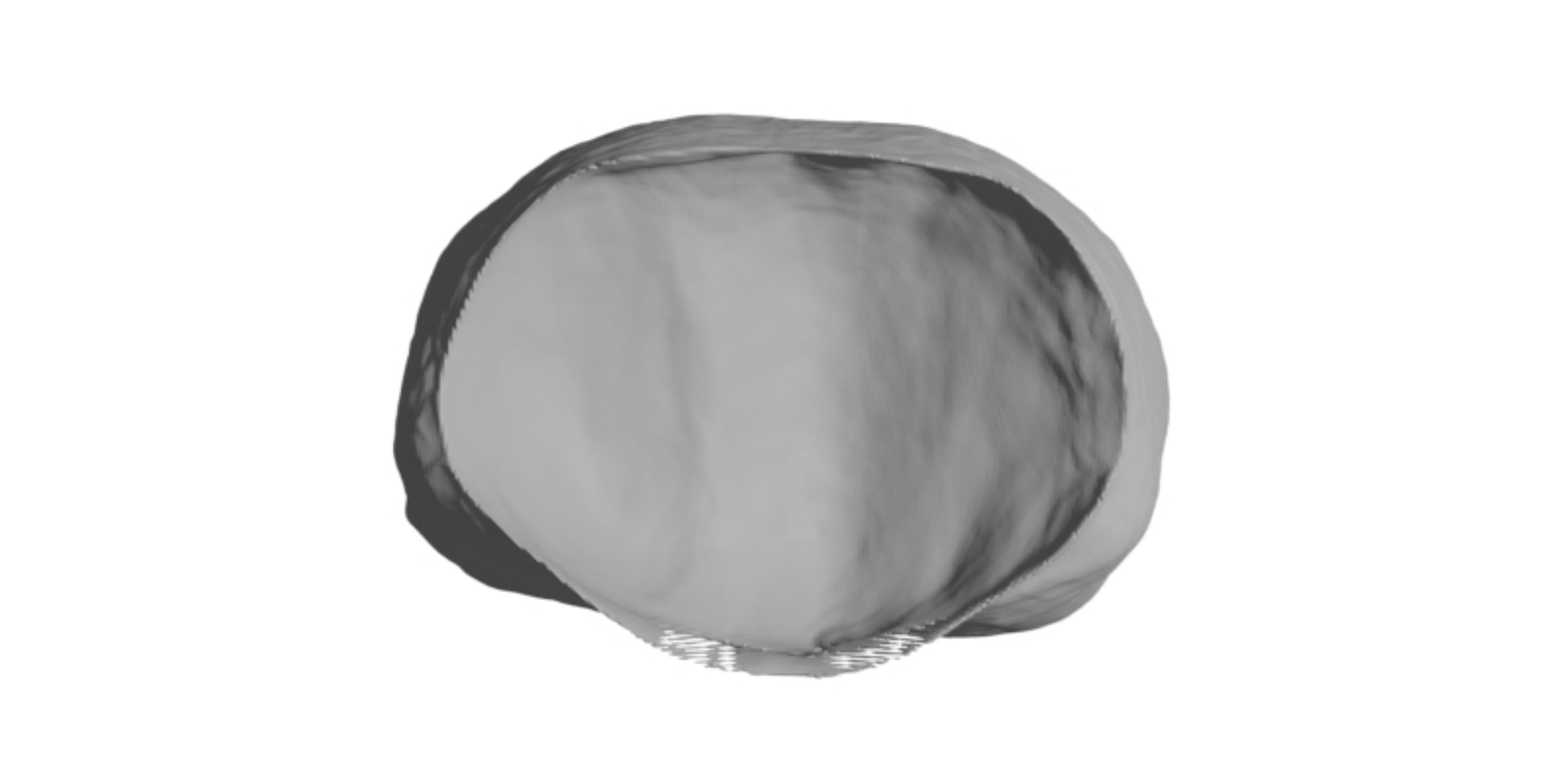}
\end{minipage}
\begin{minipage}[c]{.28\textwidth}
    \centering
    \includegraphics[width=.45\linewidth]{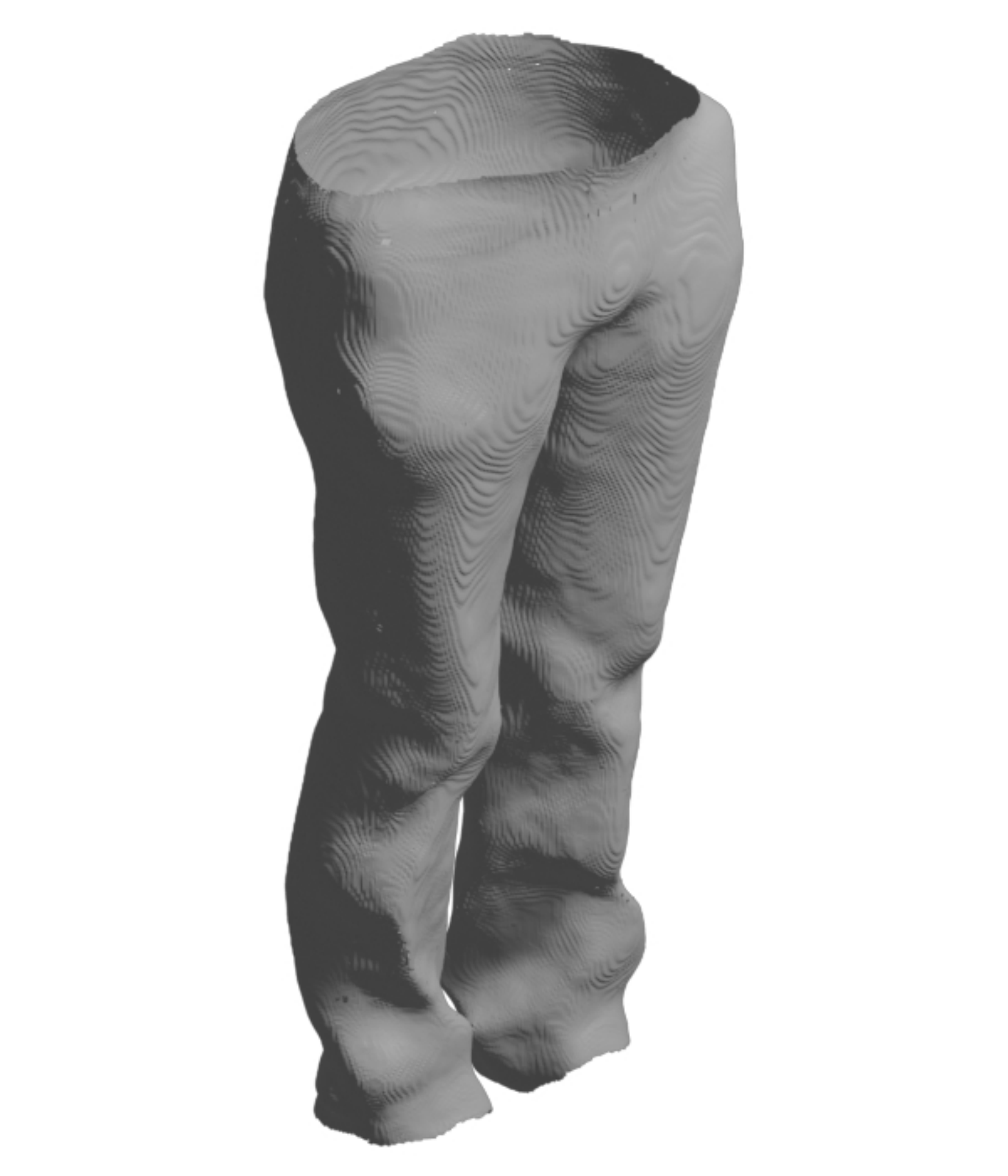}
    \includegraphics[width=.45\linewidth]{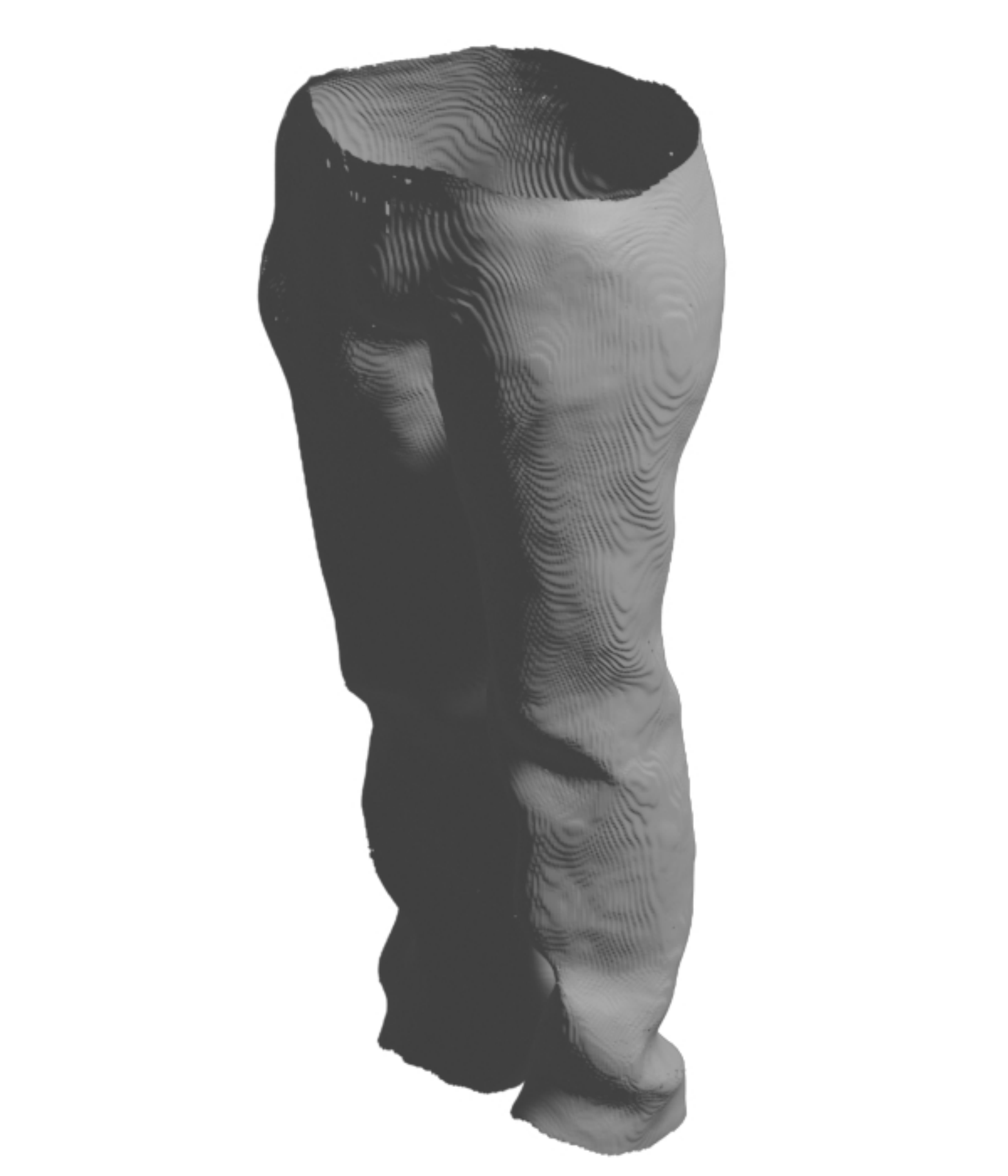}\\
    \includegraphics[width=.6\linewidth]{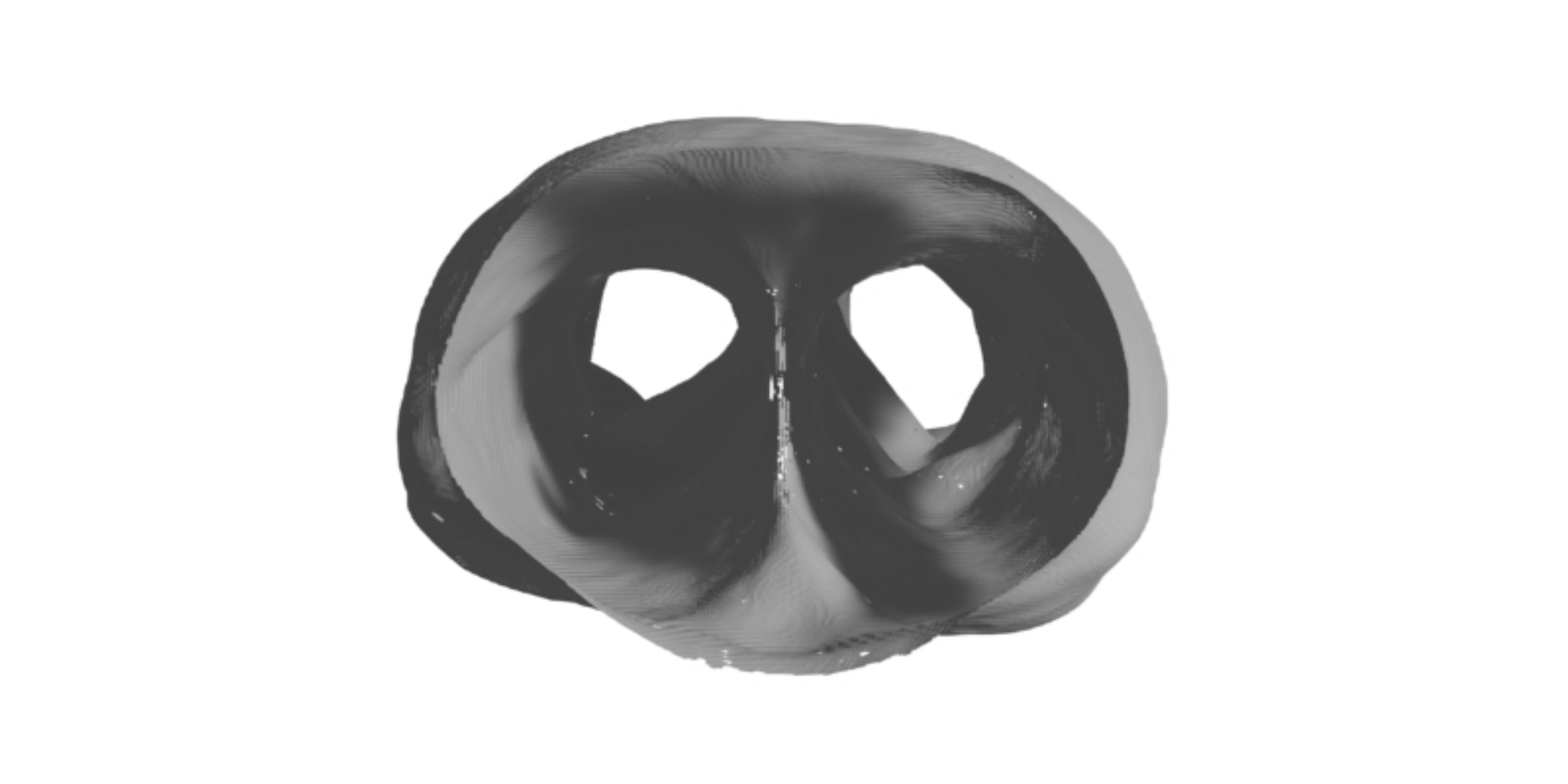}
\end{minipage}
\begin{minipage}[c]{.28\textwidth}
    \centering
    \includegraphics[width=.45\linewidth]{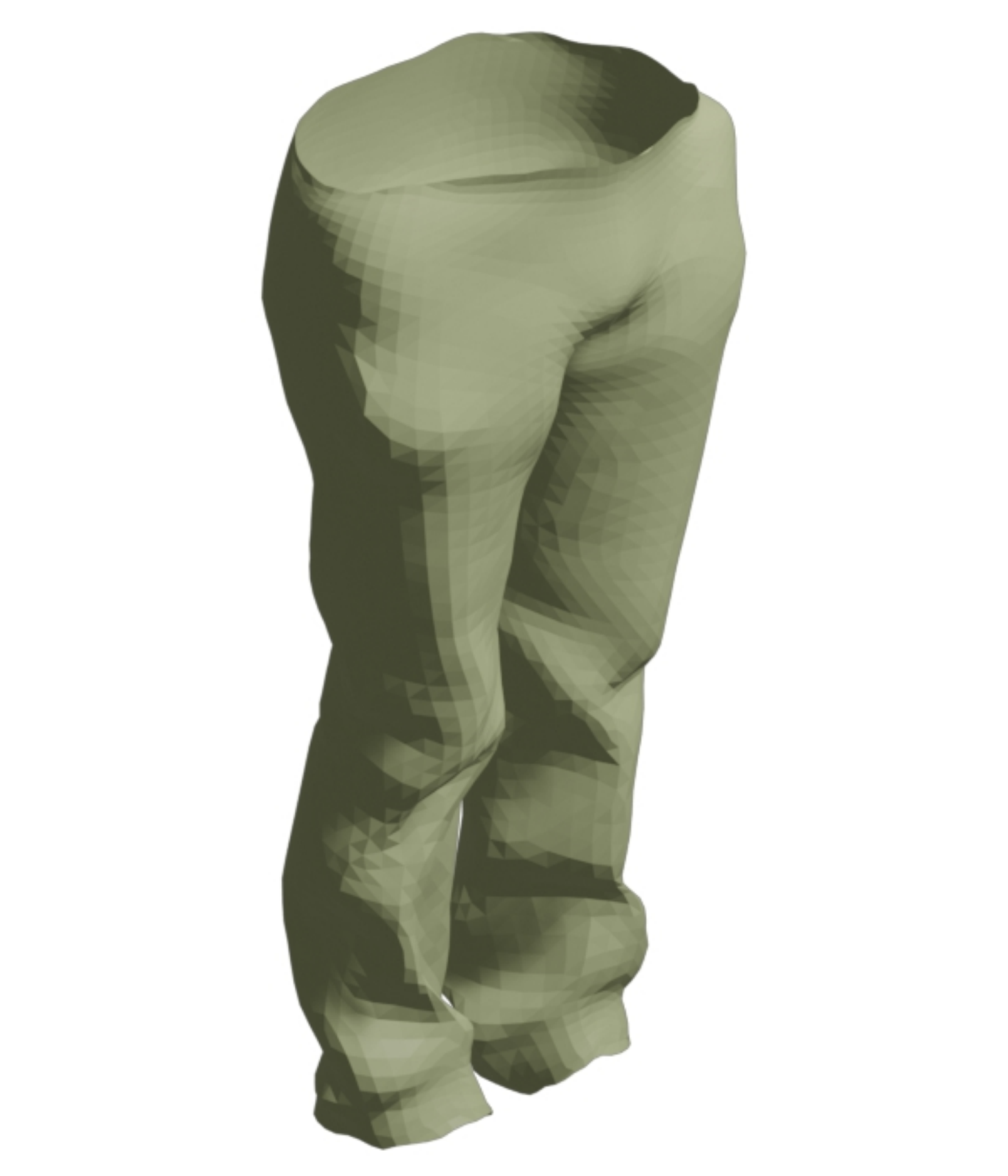}
    \includegraphics[width=.45\linewidth]{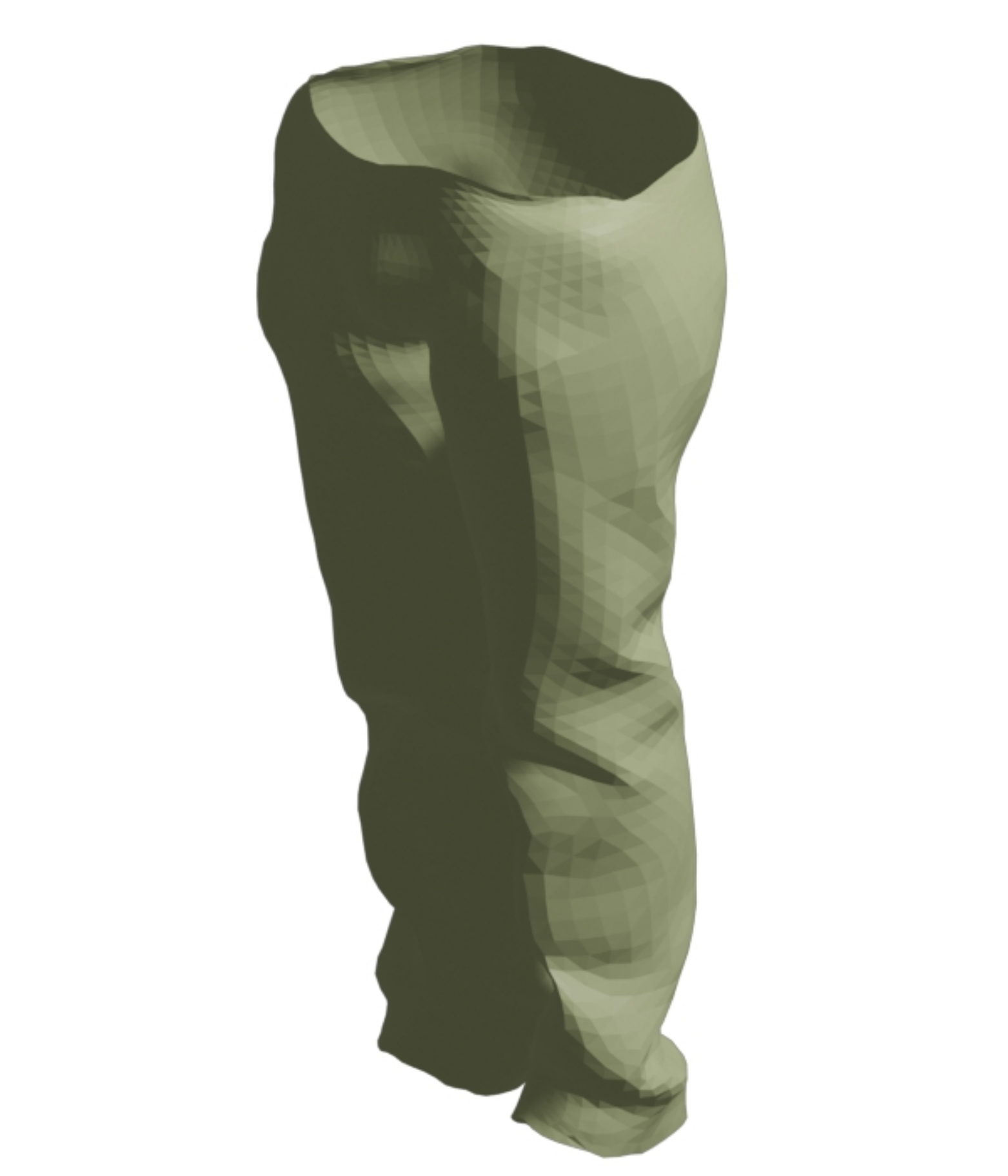}\\
    \includegraphics[width=.6\linewidth]{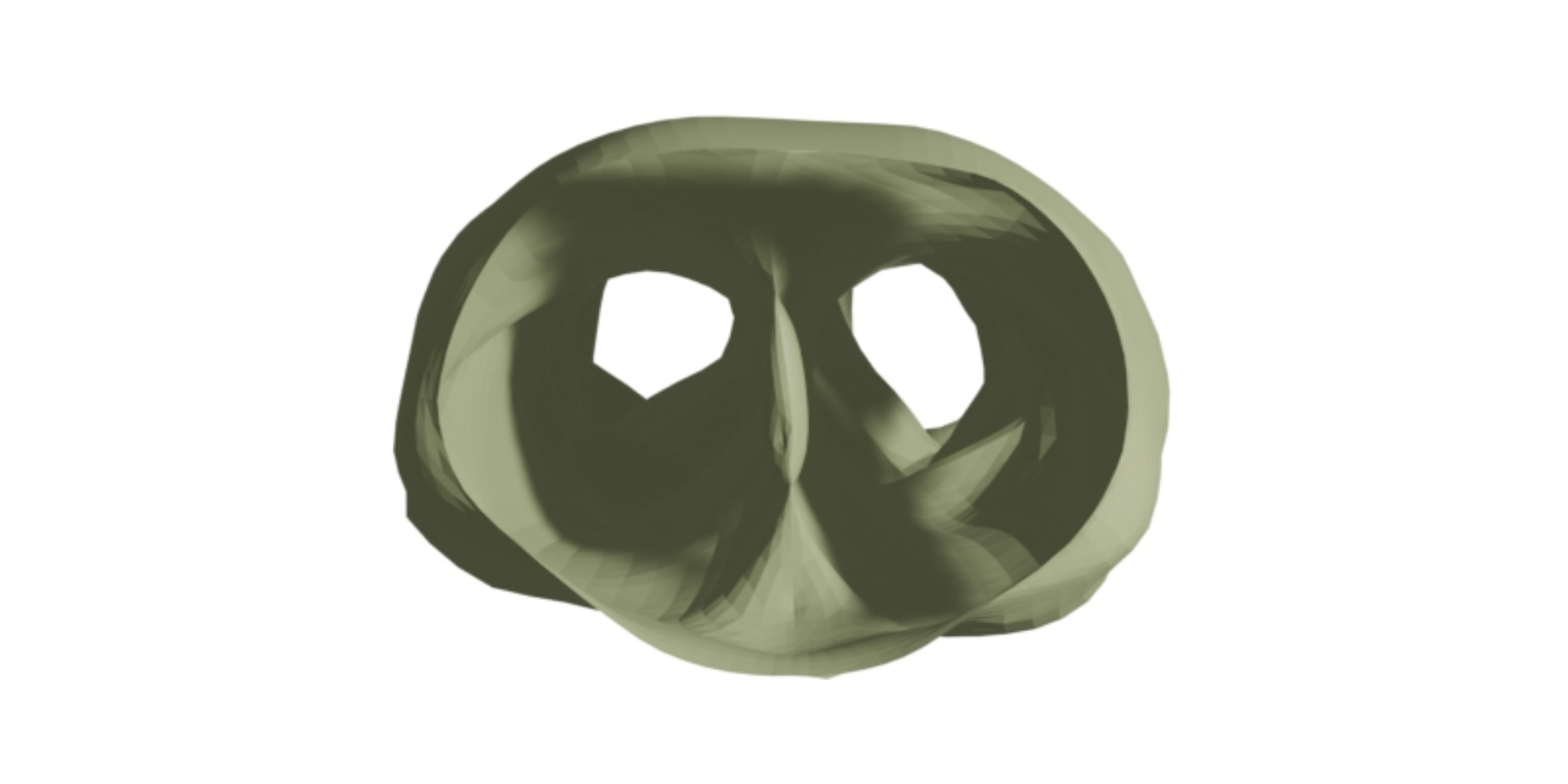}
\end{minipage}

\begin{minipage}[c]{.14\textwidth}
    \centering
    \includegraphics[width=1\linewidth]{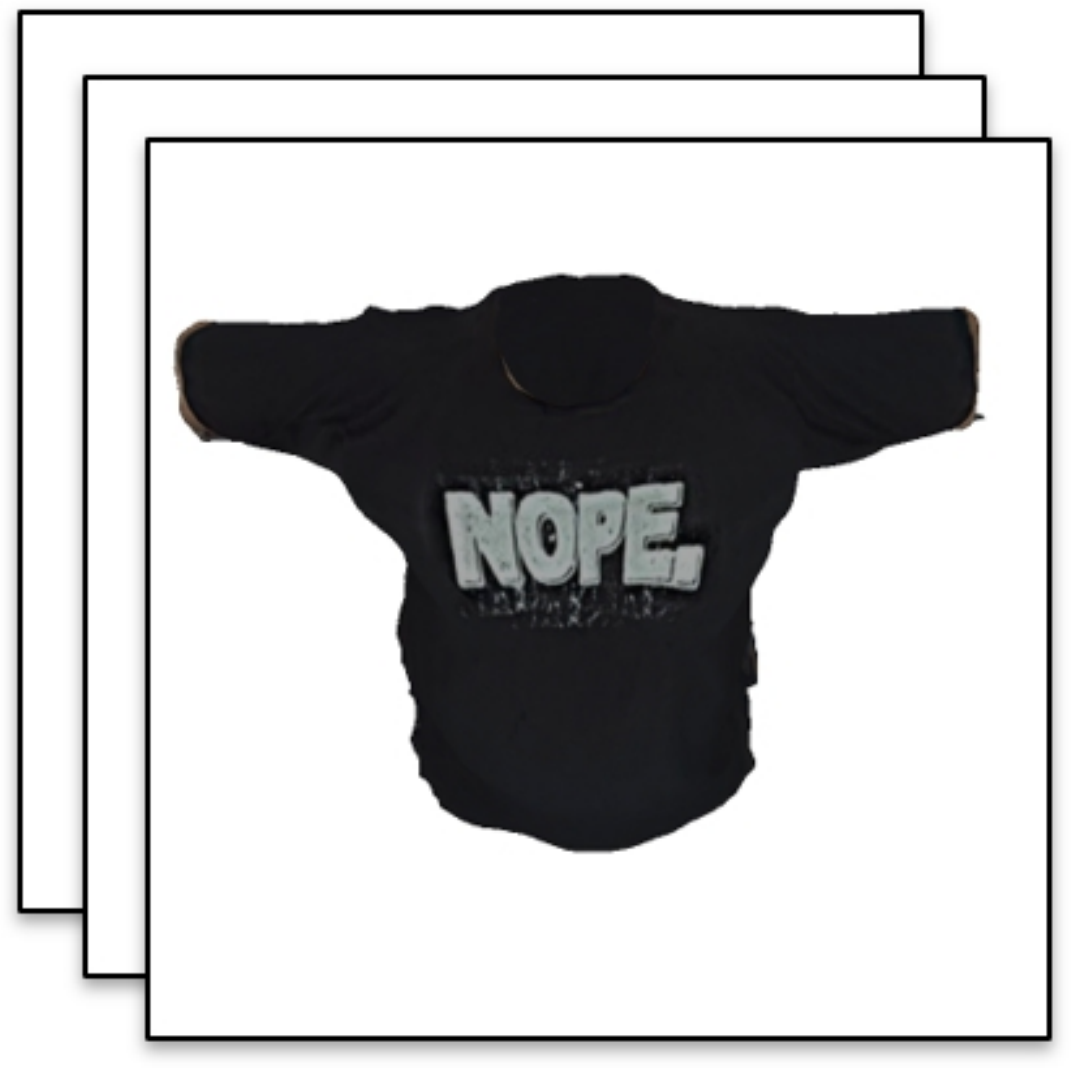}
\end{minipage}
\begin{minipage}[c]{.28\textwidth}
    \centering
    \includegraphics[width=.45\linewidth]{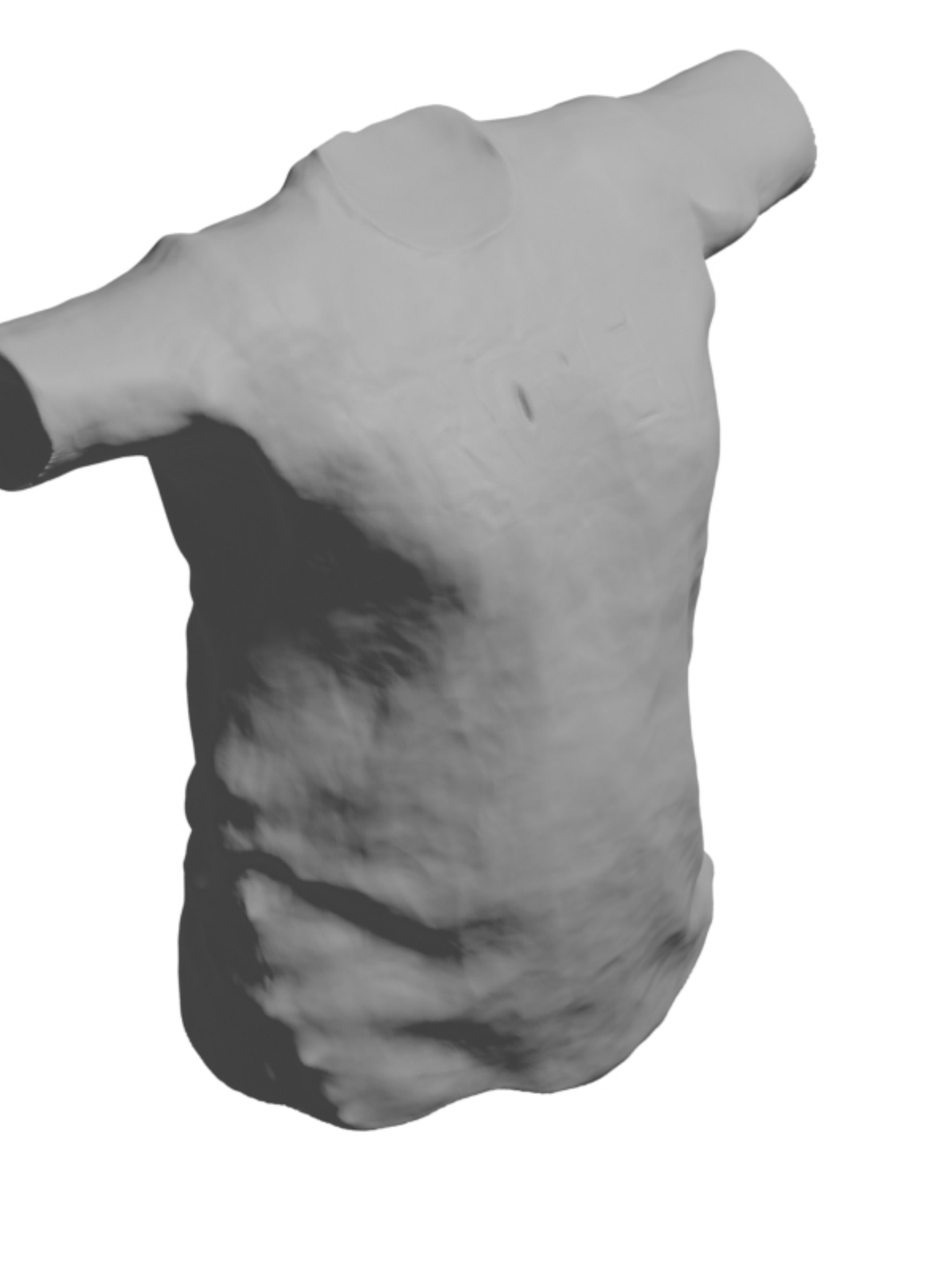}
    \includegraphics[width=.45\linewidth]{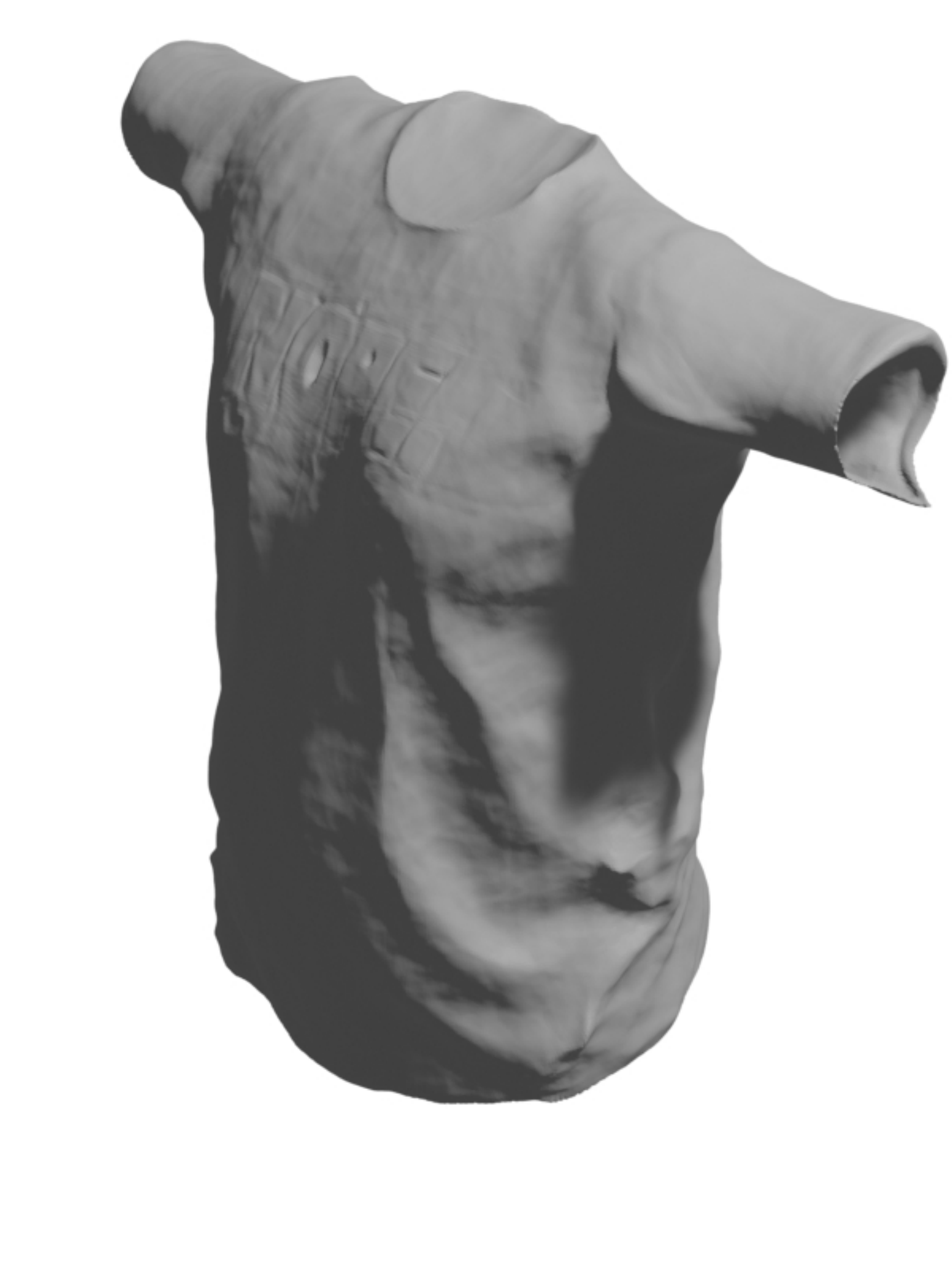}\\
    \includegraphics[width=.45\linewidth]{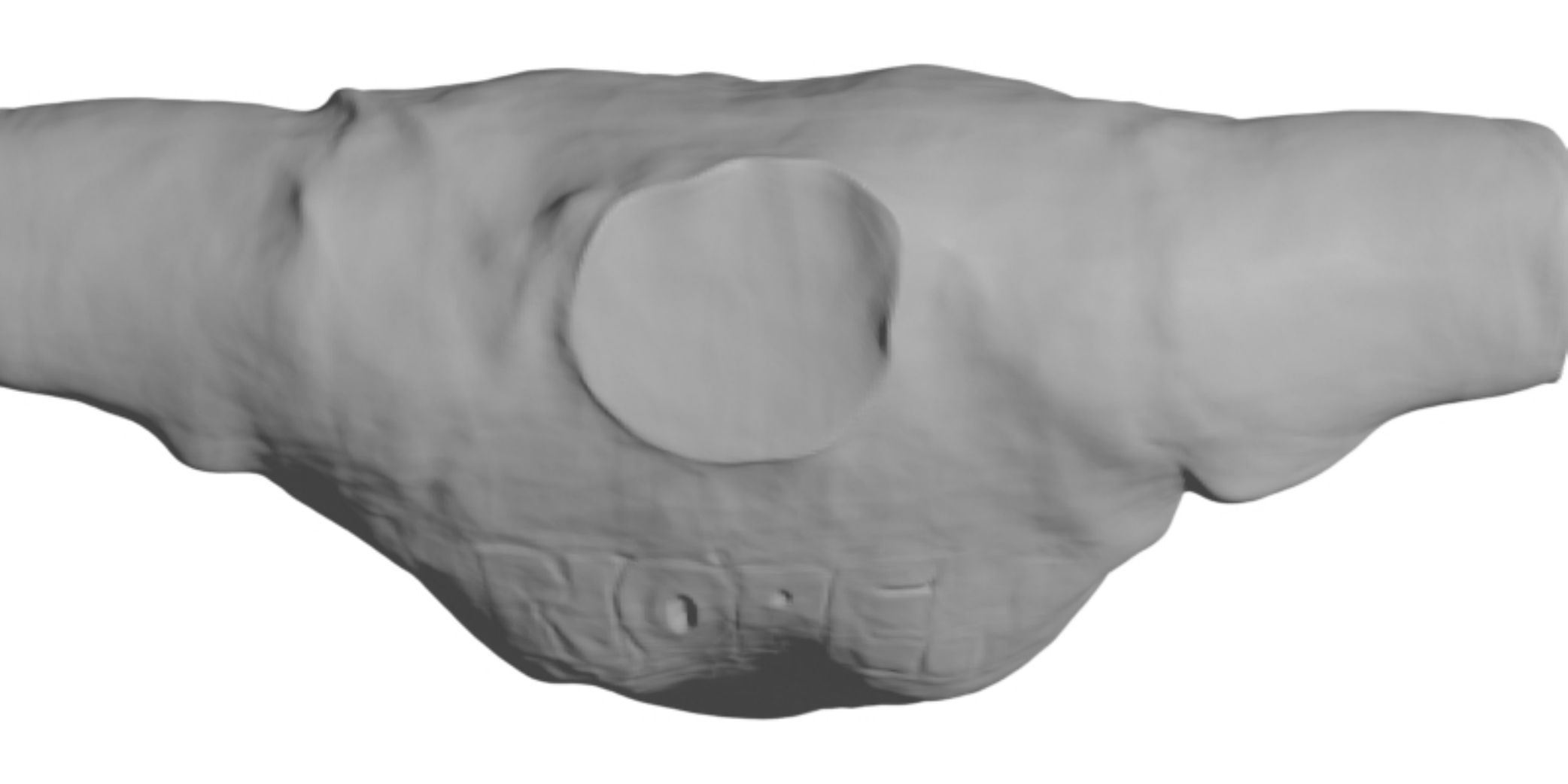}
\end{minipage}
\begin{minipage}[c]{.28\textwidth}
    \centering
    \includegraphics[width=.45\linewidth]{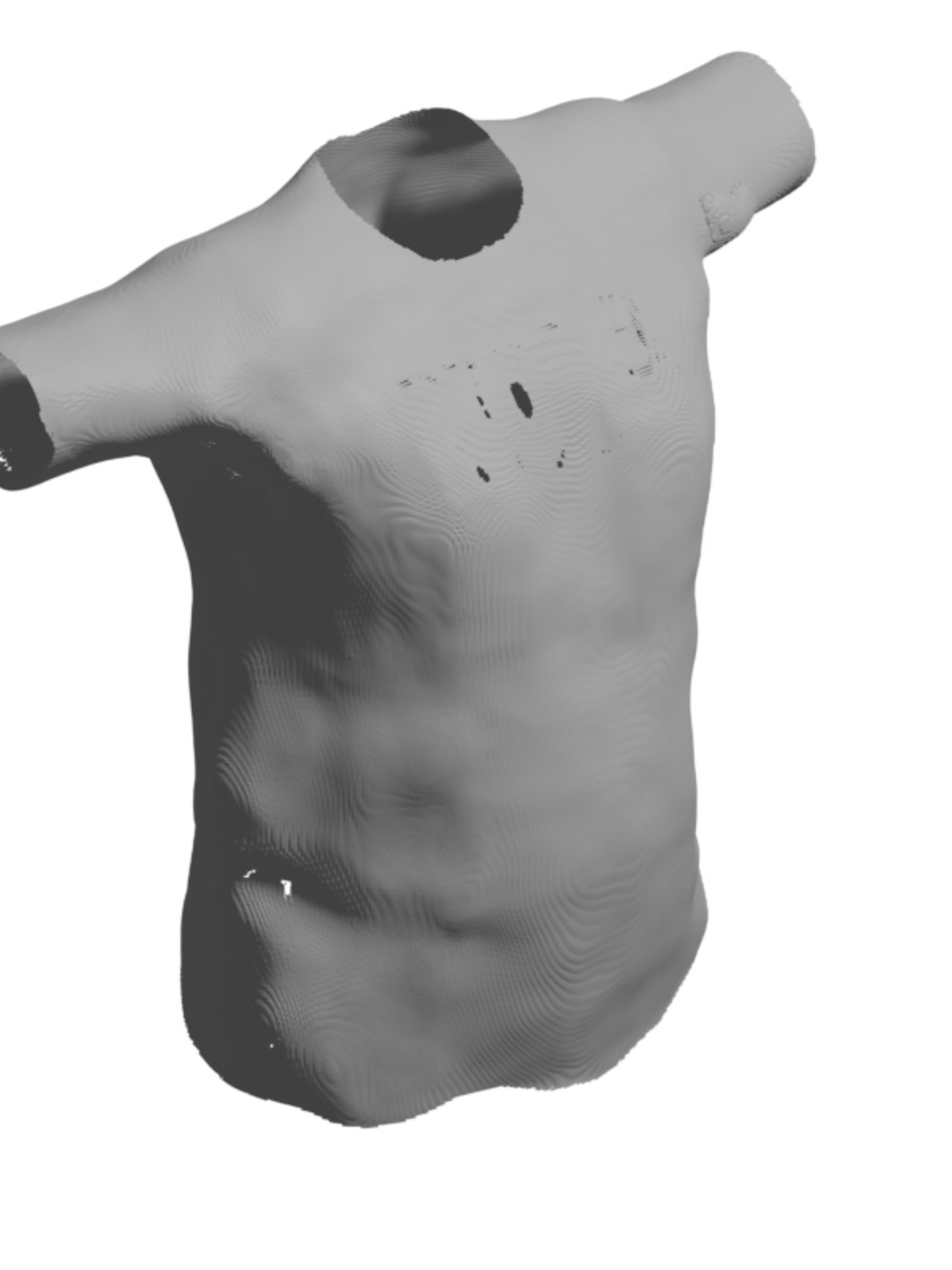}
    \includegraphics[width=.45\linewidth]{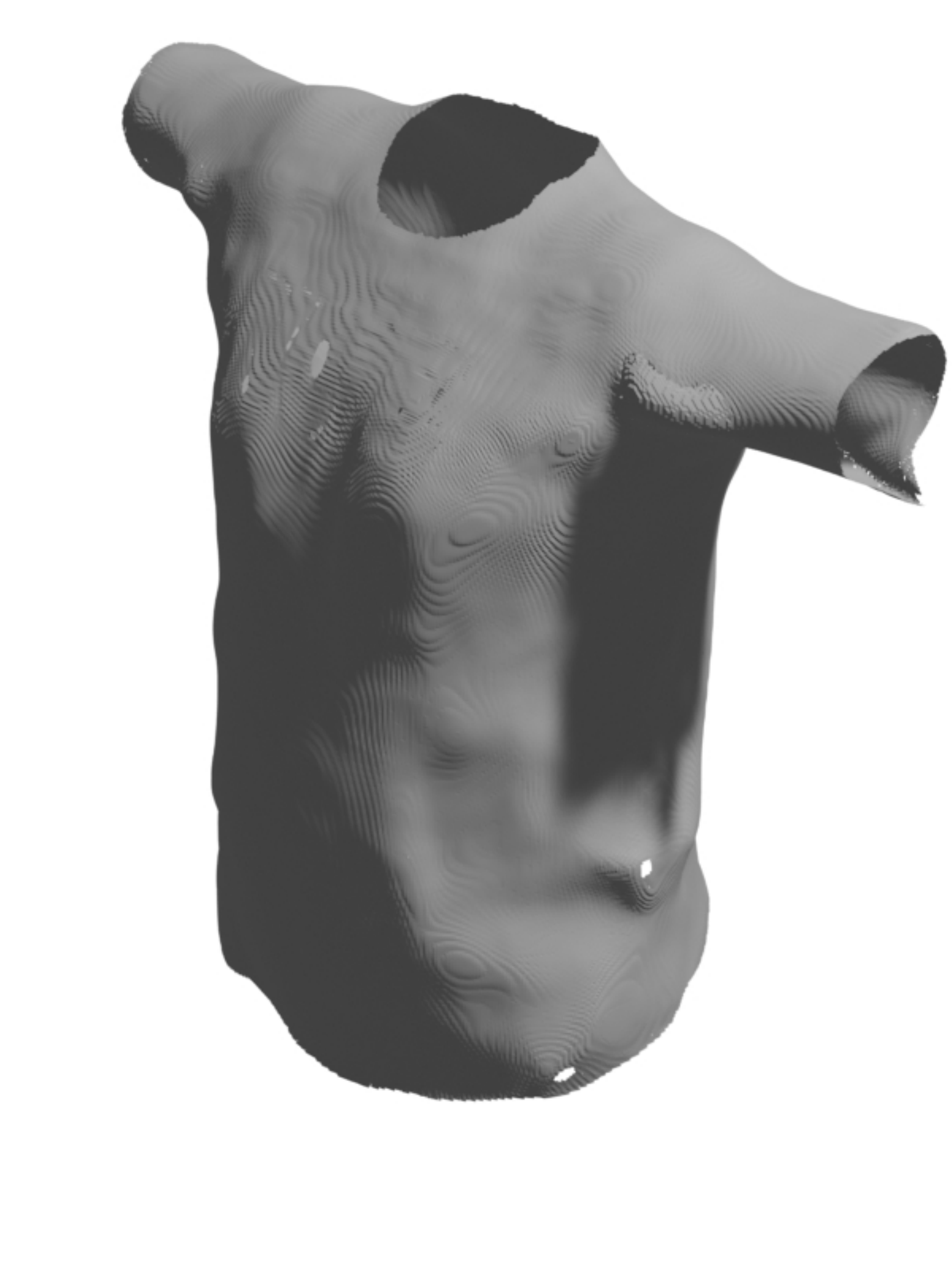}\\
    \includegraphics[width=.45\linewidth]{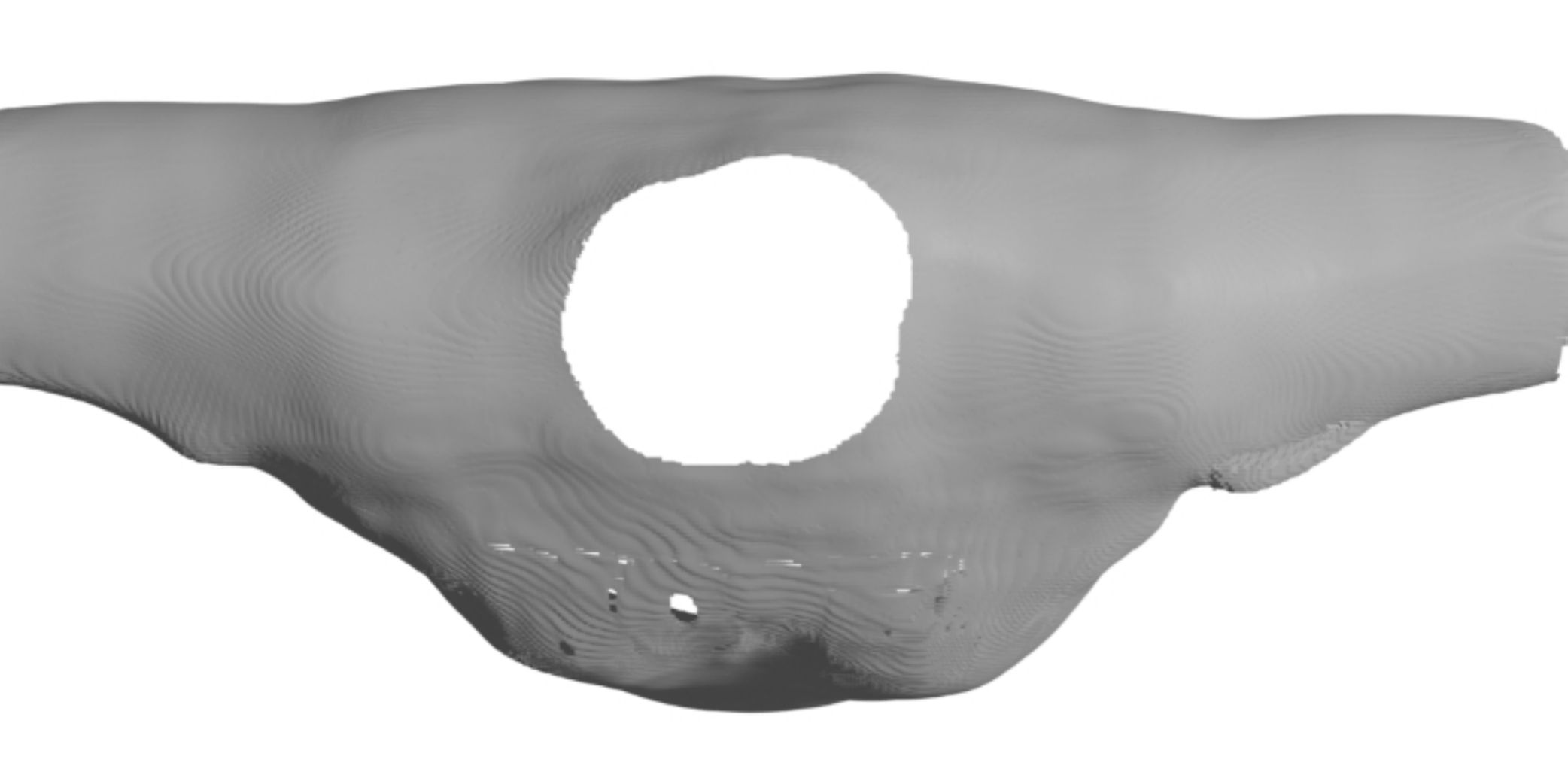}
\end{minipage}
\begin{minipage}[c]{.28\textwidth}
    \centering
    \includegraphics[width=.45\linewidth]{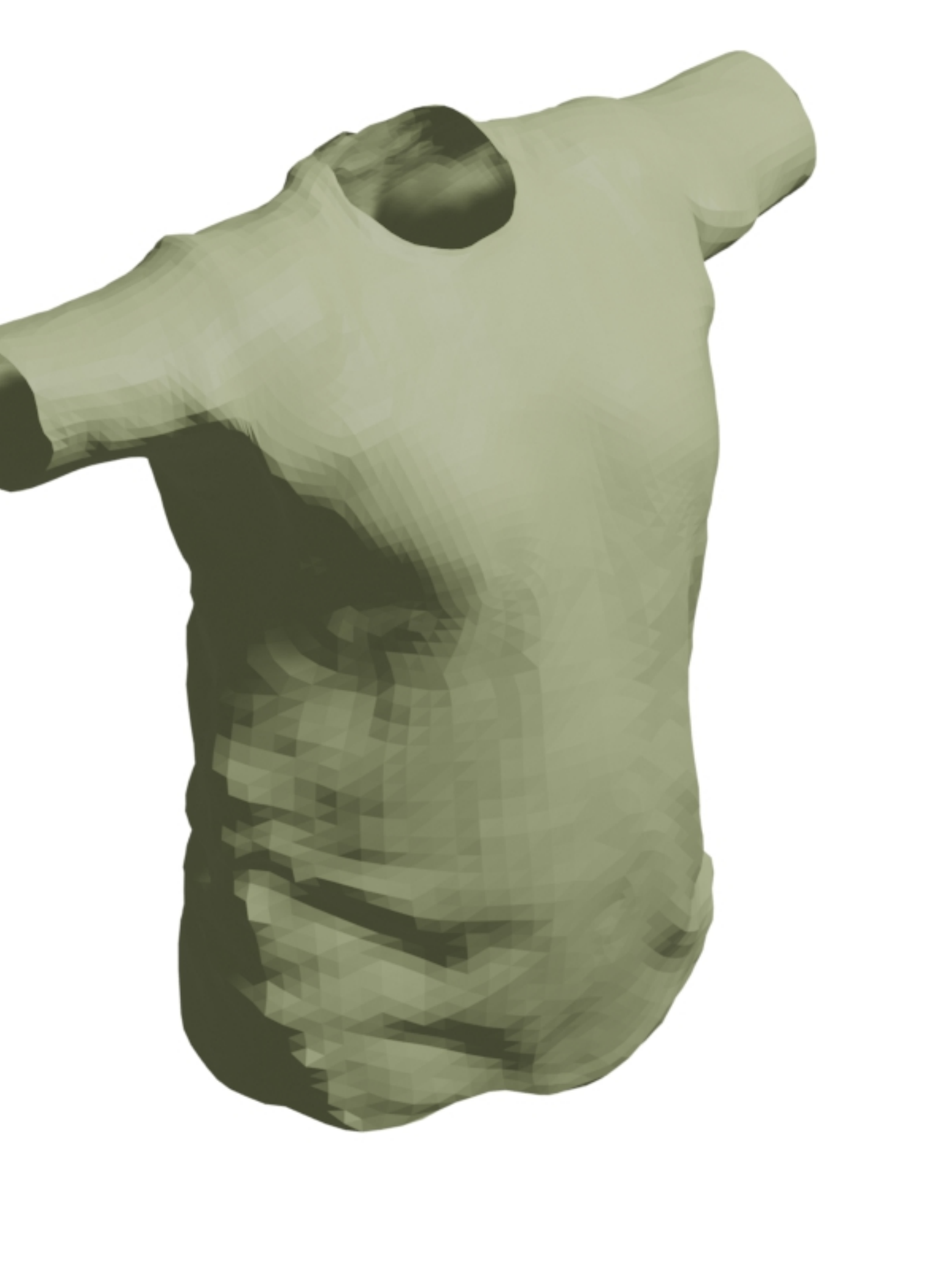}
    \includegraphics[width=.45\linewidth]{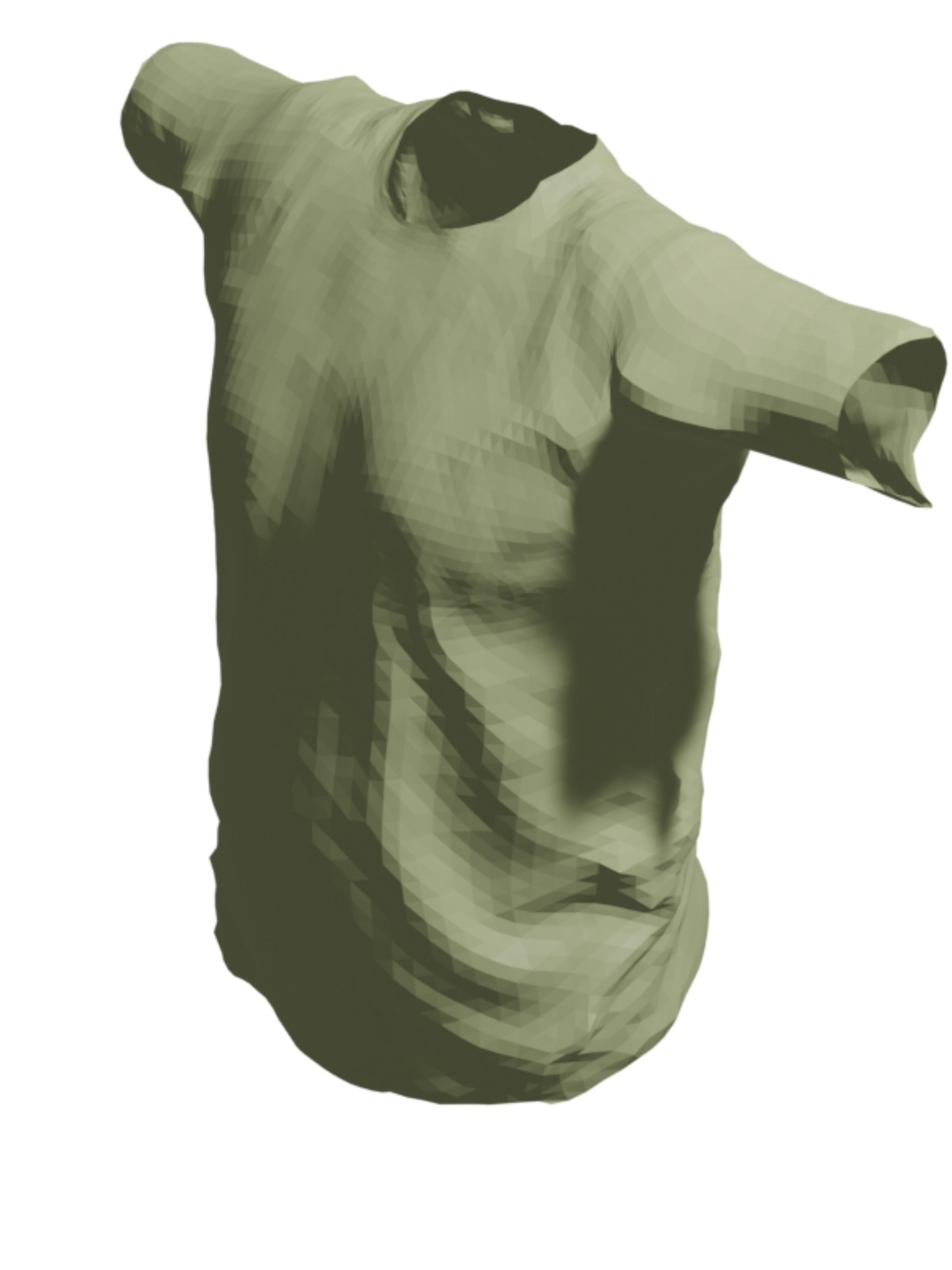}\\
    \includegraphics[width=.45\linewidth]{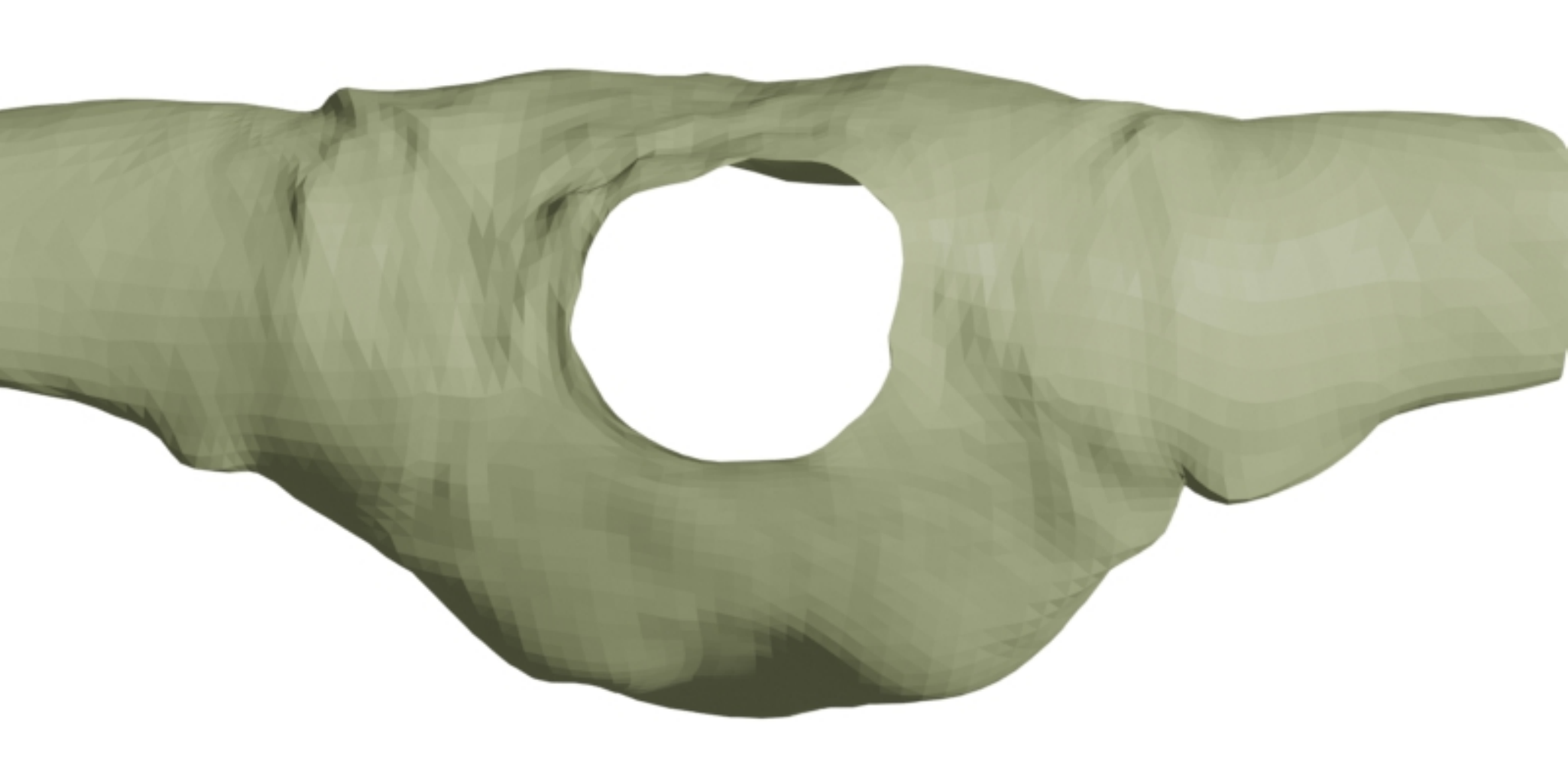}
\end{minipage}

\begin{minipage}[c]{.13\textwidth}
    \centering
    Input
\end{minipage}
\begin{minipage}[c]{.28\textwidth}
    \centering
    NeuS
\end{minipage}
\begin{minipage}[c]{.28\textwidth}
    \centering
    Ours
\end{minipage}
\begin{minipage}[c]{.28\textwidth}
    \centering
    Ground-truth
\end{minipage}

\caption{Additional results on the MGN~\cite{mgn} dataset with mask supervision.}
\label{supp_mgn}
\end{figure*}

\begin{figure*}[h]

\begin{minipage}[c]{.2\textwidth}
    \centering
    \includegraphics[width=1\linewidth]{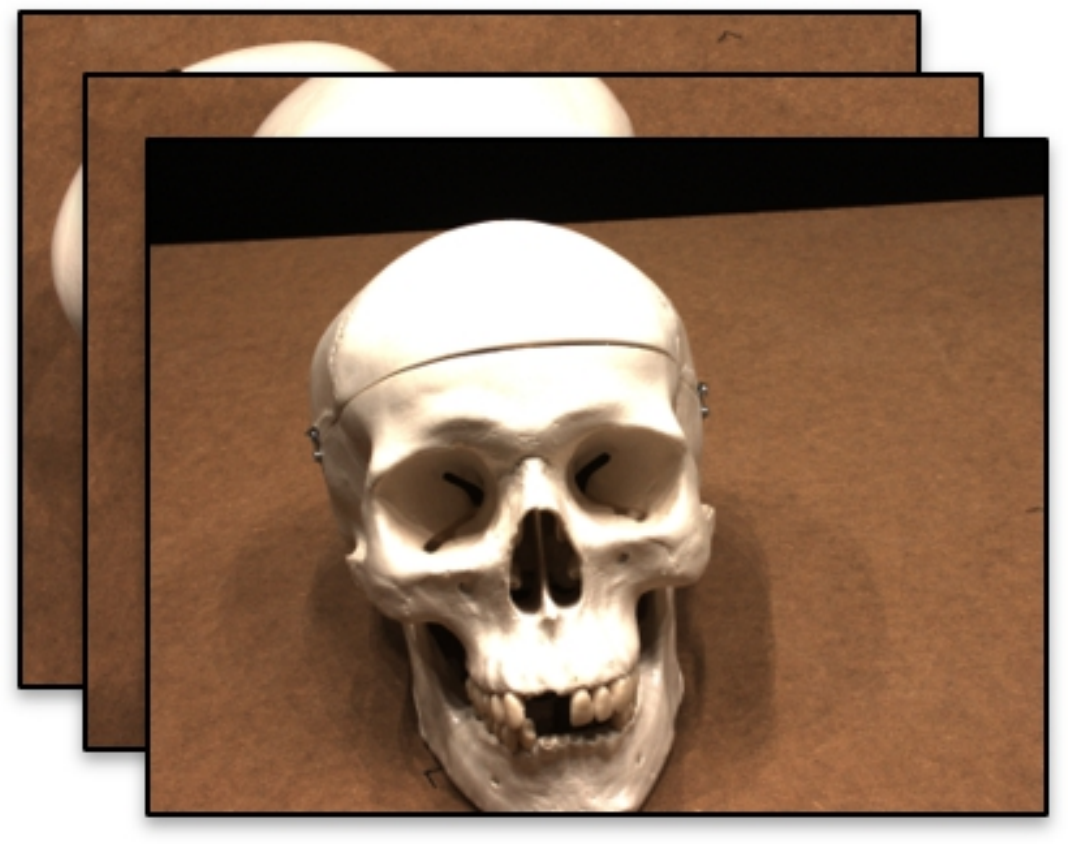}
\end{minipage}
\begin{minipage}[c]{.36\textwidth}
    \centering
    \includegraphics[width=1\linewidth]{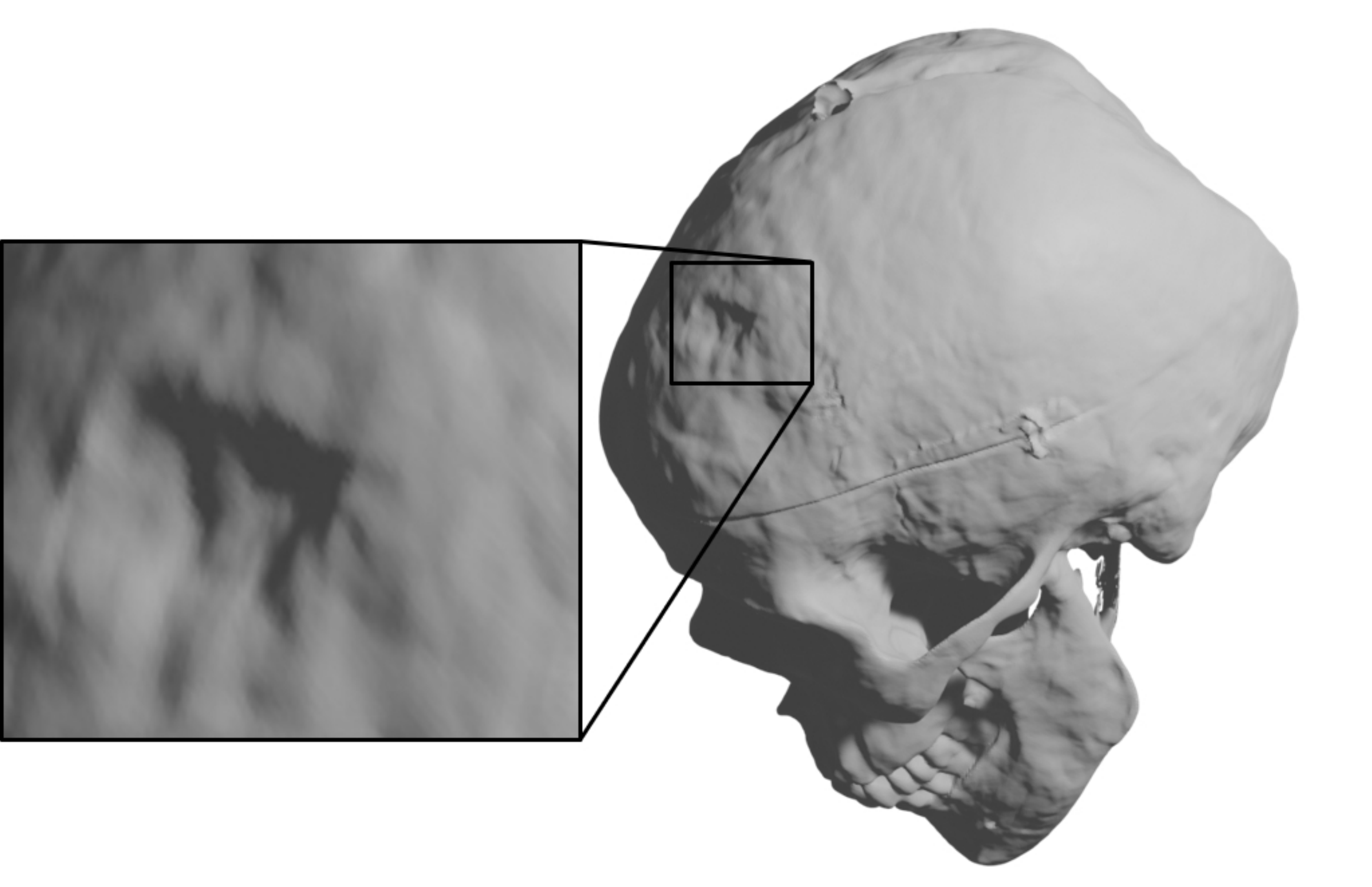}
\end{minipage}
\begin{minipage}[c]{.36\textwidth}
    \centering
    \includegraphics[width=1\linewidth]{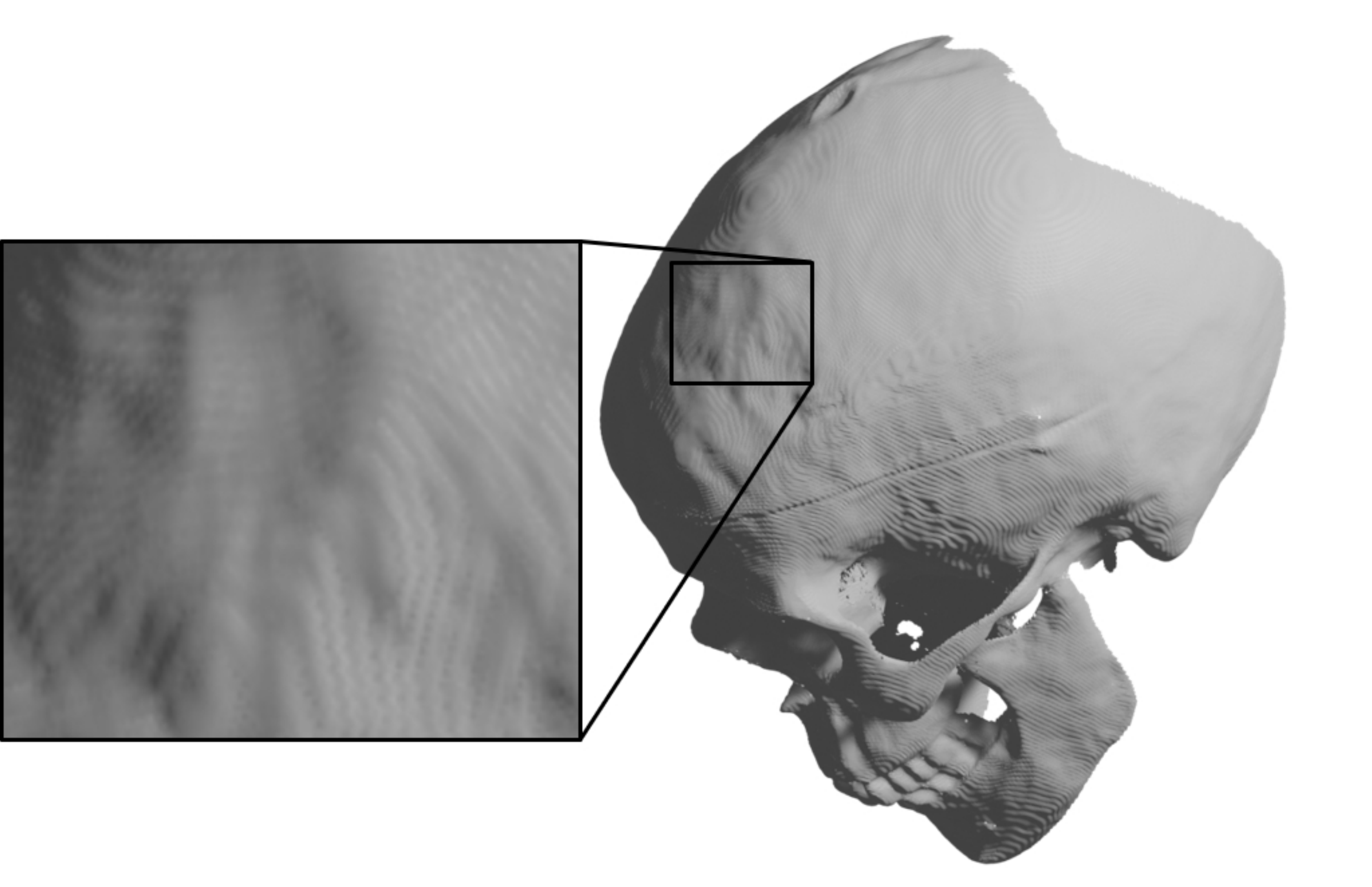}
\end{minipage}

\begin{minipage}[c]{.2\textwidth}
    \centering
    \includegraphics[width=1\linewidth]{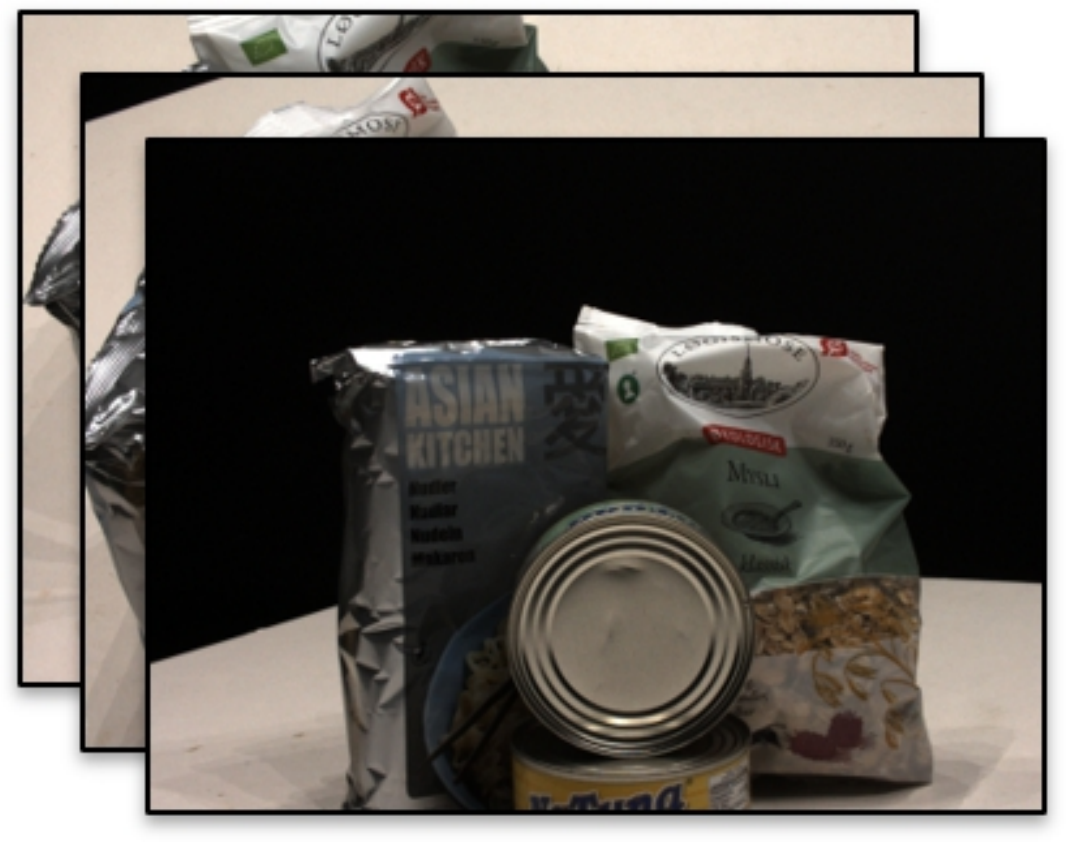}
\end{minipage}
\begin{minipage}[c]{.36\textwidth}
    \centering
    \includegraphics[width=1\linewidth]{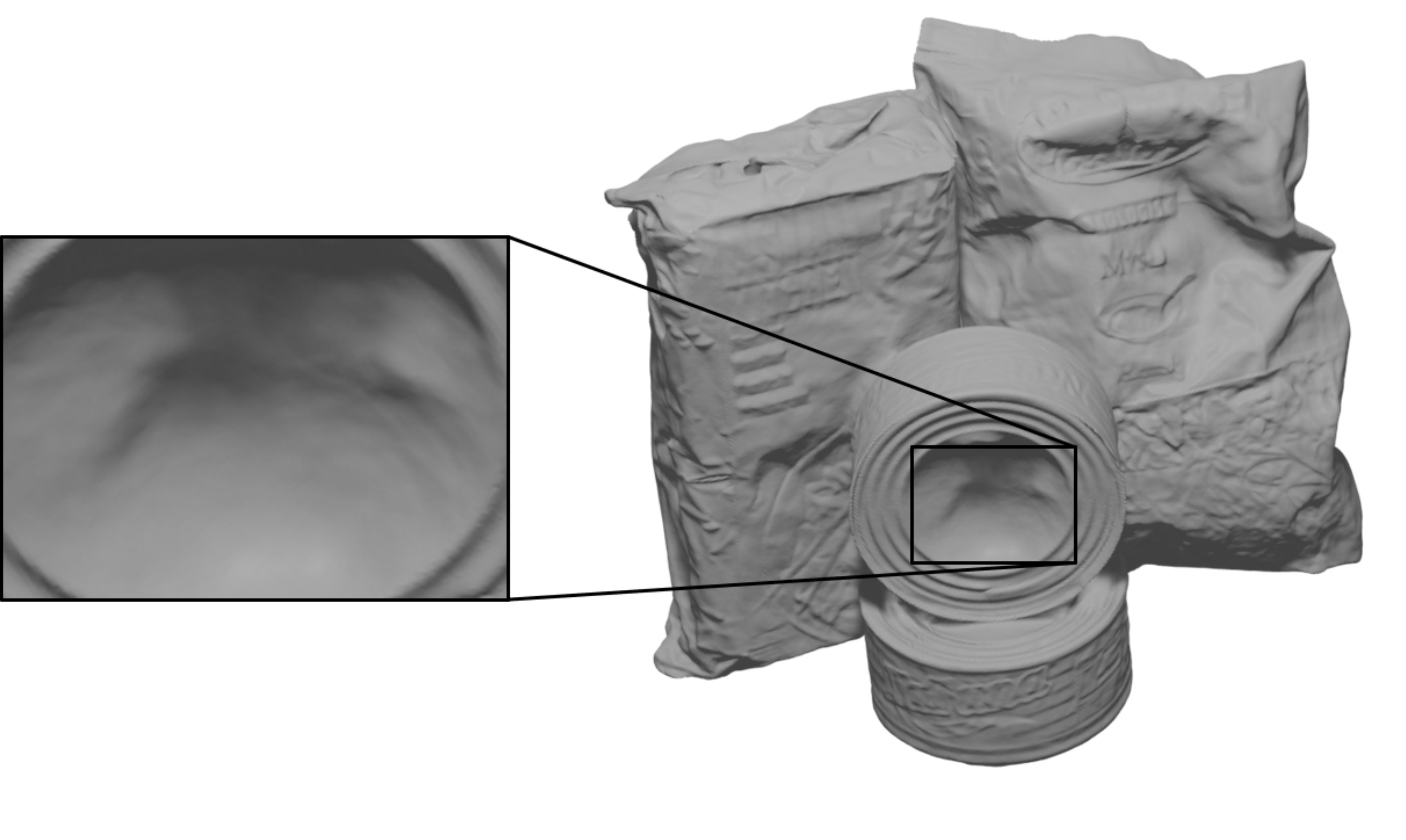}
\end{minipage}
\begin{minipage}[c]{.36\textwidth}
    \centering
    \includegraphics[width=1\linewidth]{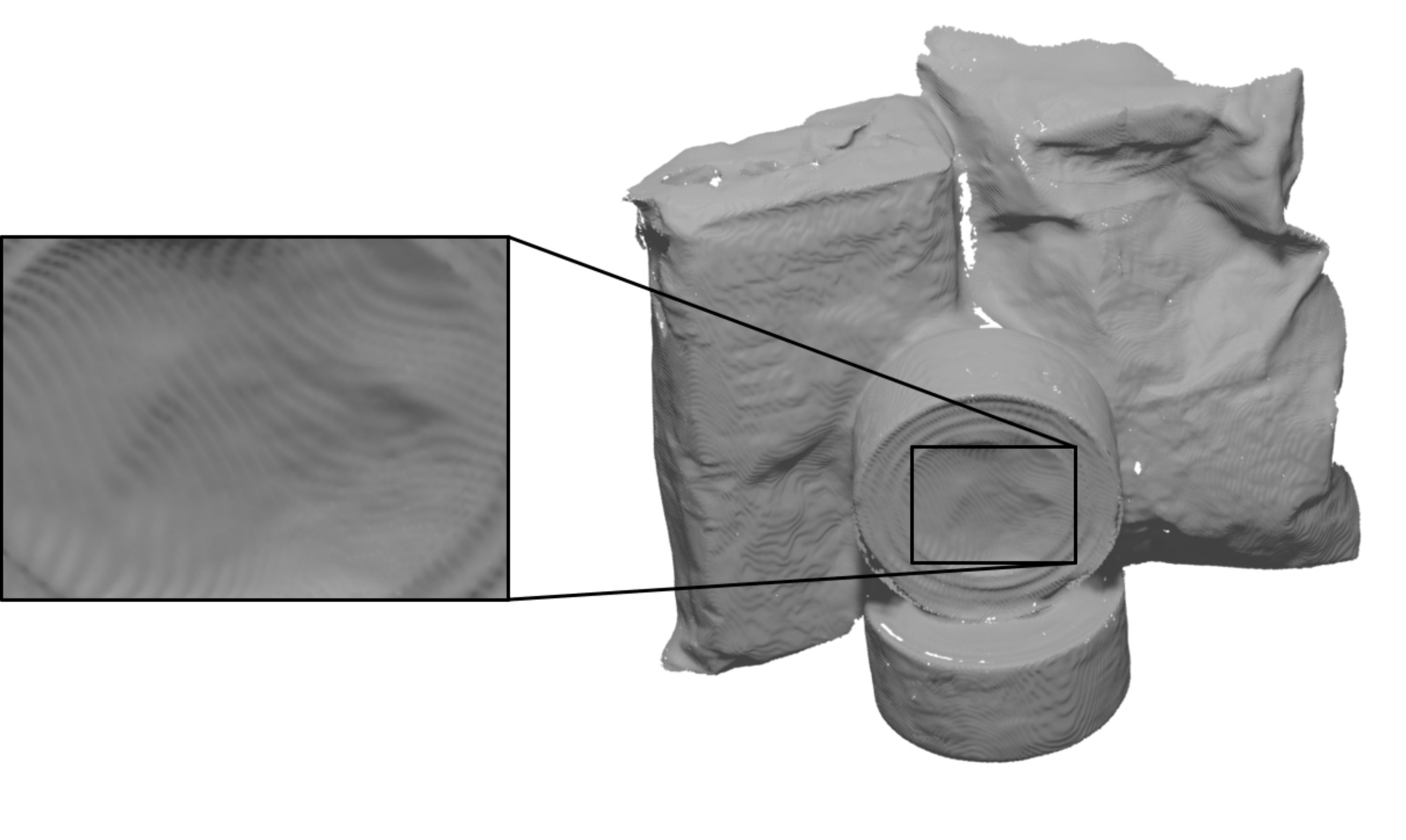}
\end{minipage}

\begin{minipage}[c]{.2\textwidth}
    \centering
    \includegraphics[width=1\linewidth]{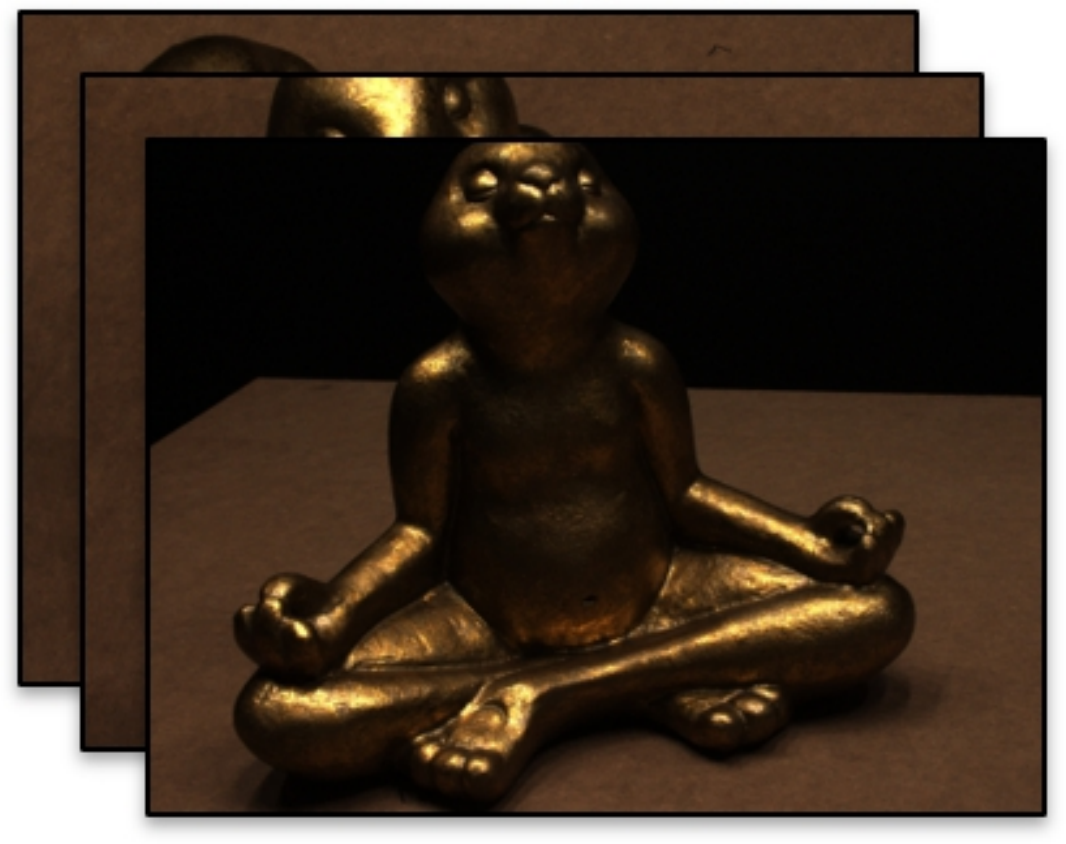}
\end{minipage}
\begin{minipage}[c]{.36\textwidth}
    \centering
    \includegraphics[width=1\linewidth]{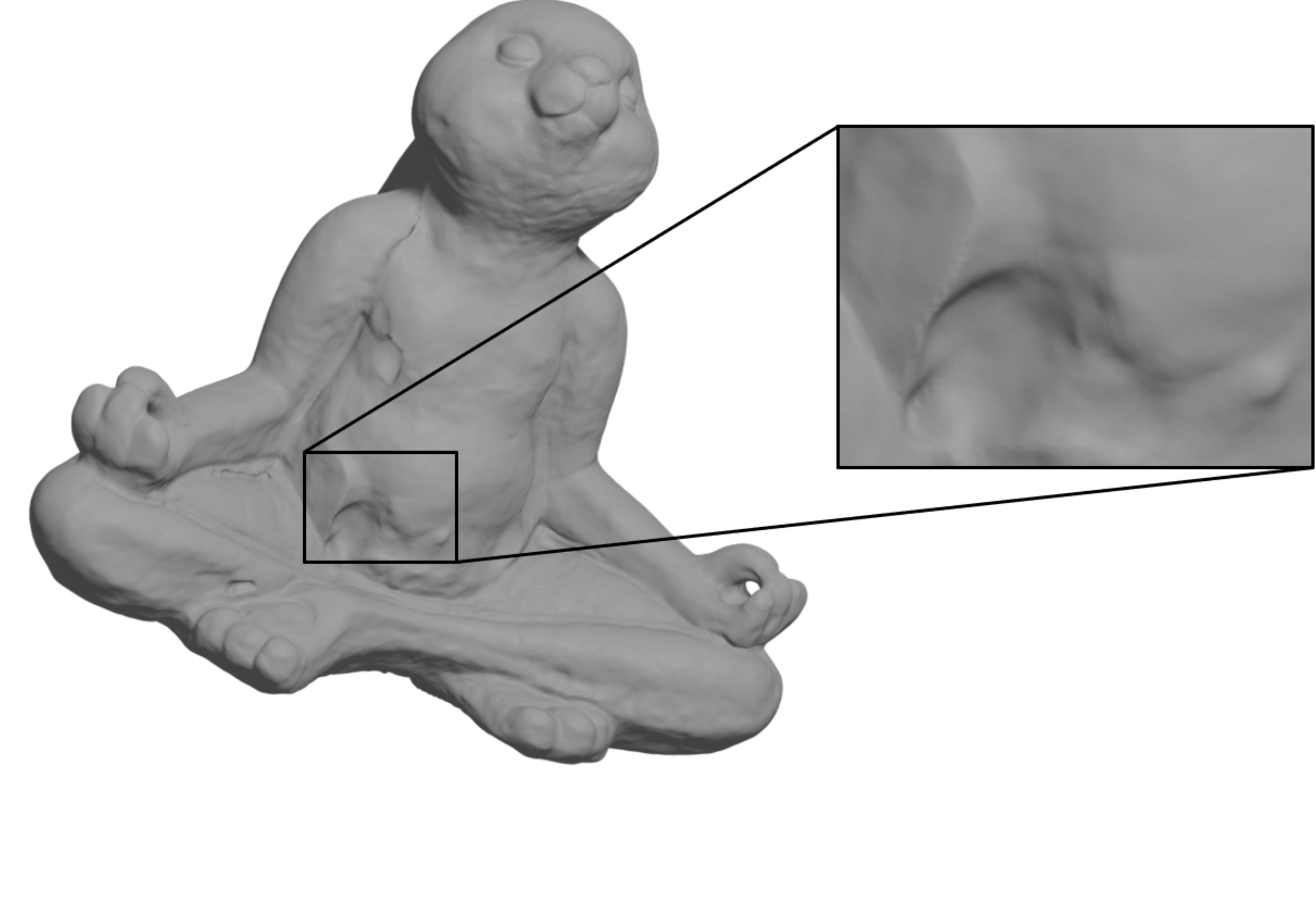}
\end{minipage}
\begin{minipage}[c]{.36\textwidth}
    \centering
    \includegraphics[width=1\linewidth]{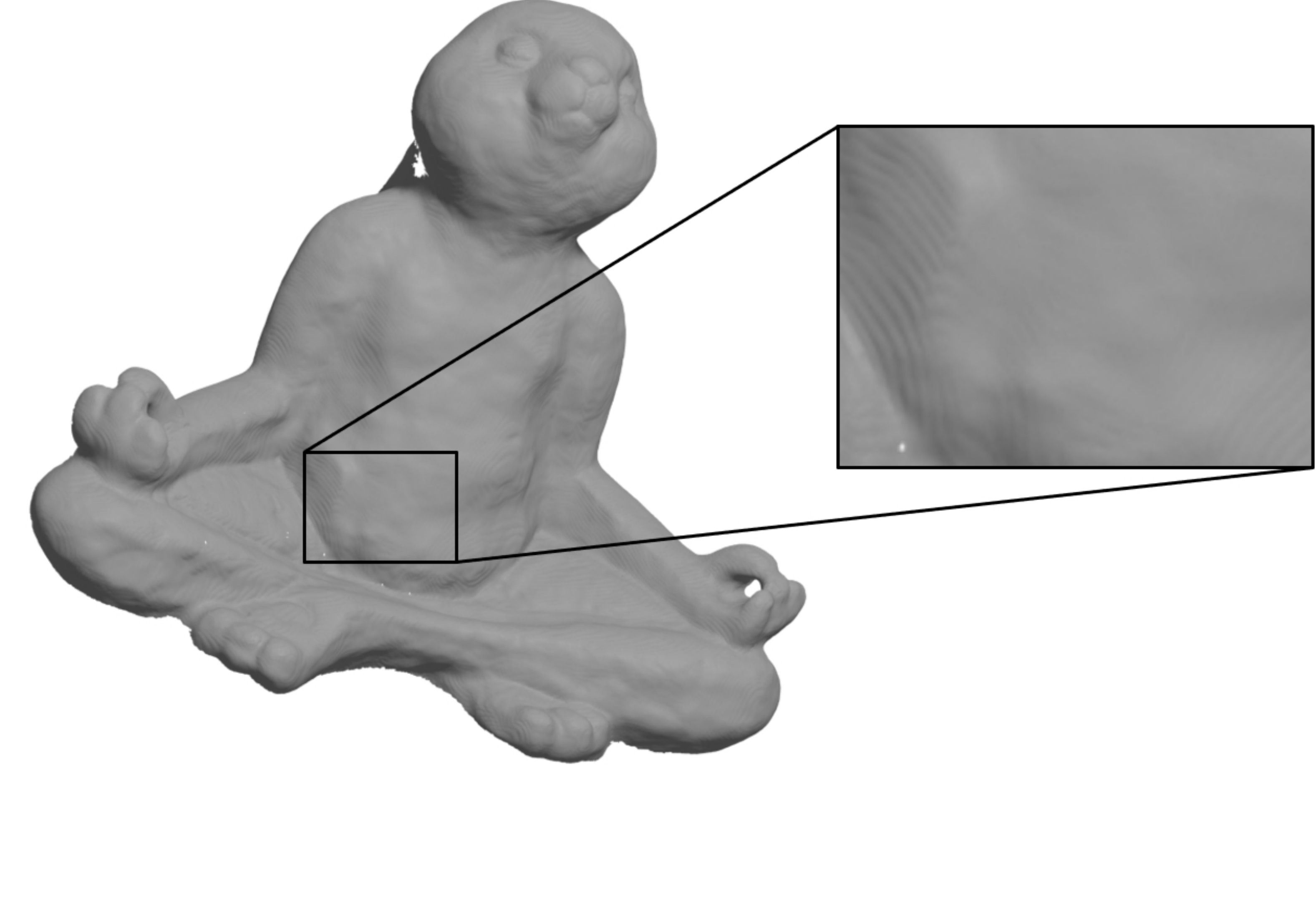}
\end{minipage}

\begin{minipage}[c]{.2\textwidth}
    \centering
    \includegraphics[width=1\linewidth]{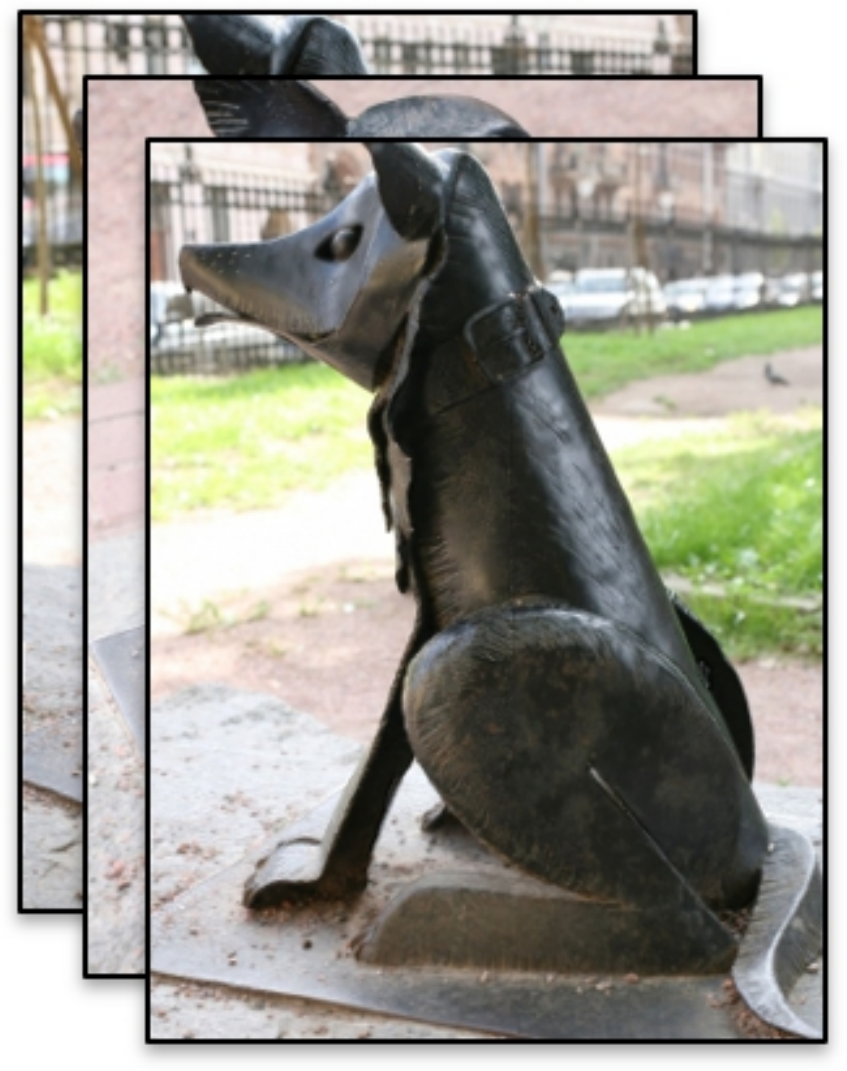}
\end{minipage}
\begin{minipage}[c]{.36\textwidth}
    \centering
    \includegraphics[width=1\linewidth]{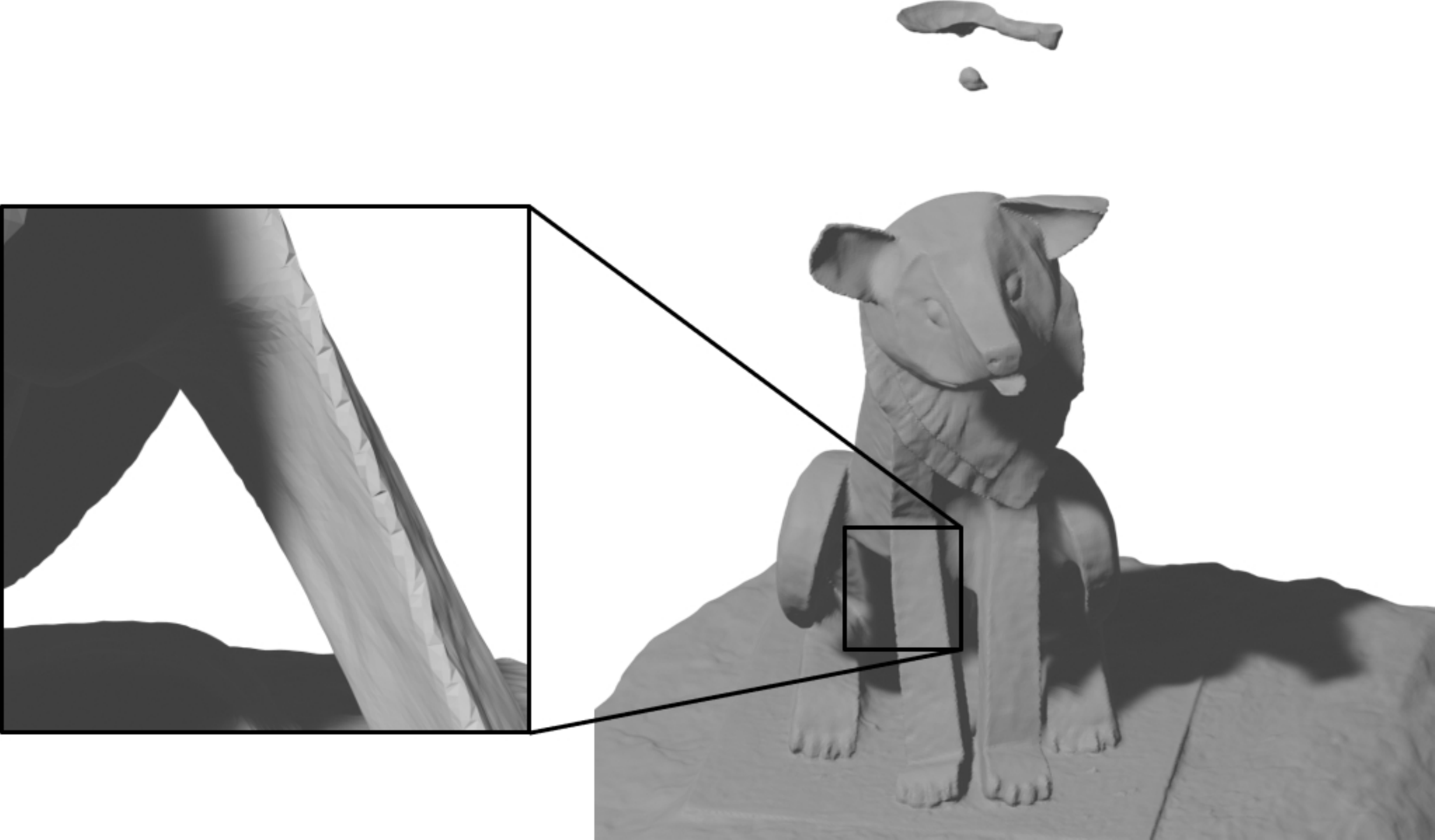}
\end{minipage}
\begin{minipage}[c]{.36\textwidth}
    \centering
    \includegraphics[width=1\linewidth]{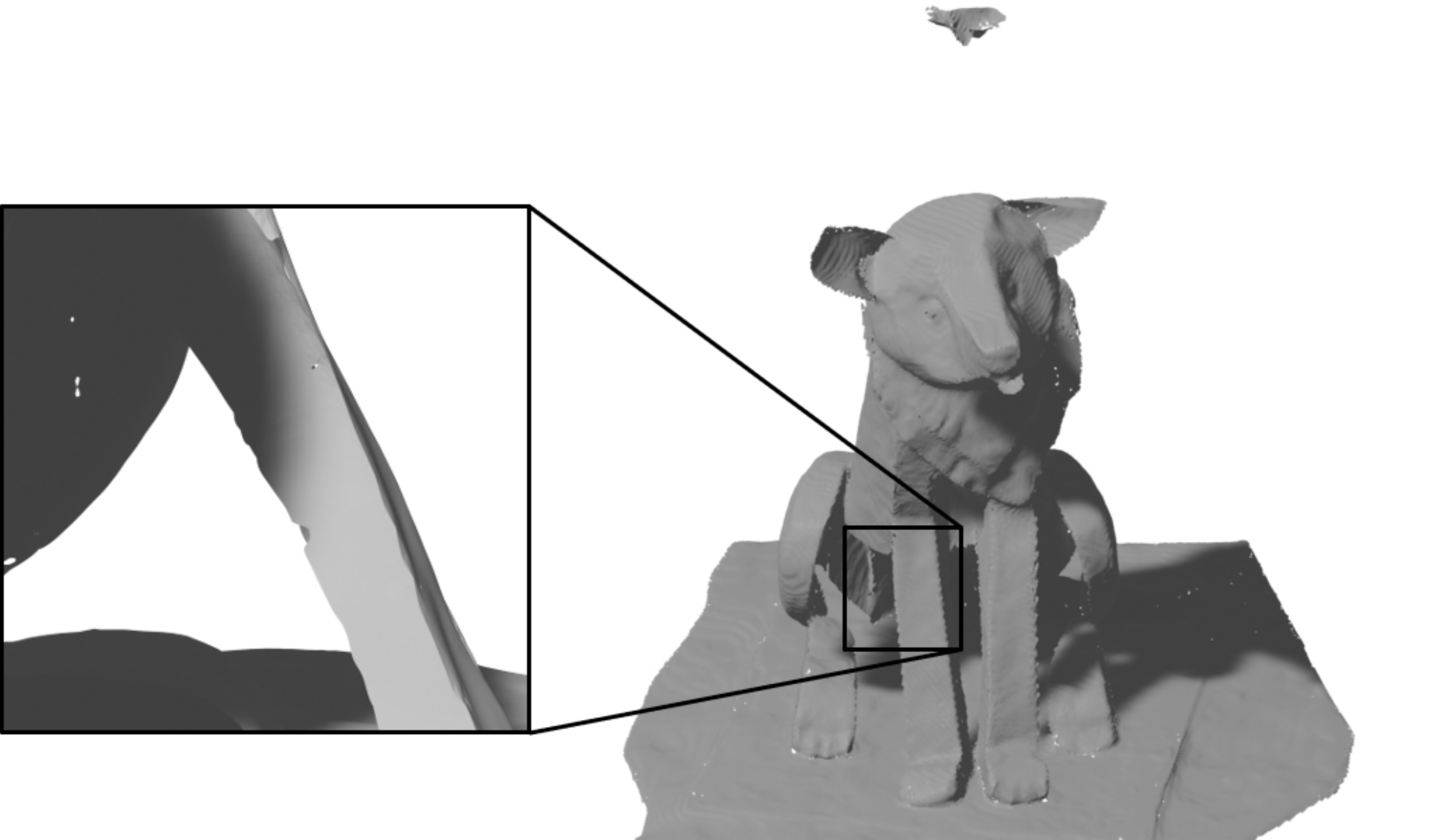}
\end{minipage}

\begin{minipage}[c]{.2\textwidth}
    \centering
    Input
\end{minipage}
\begin{minipage}[c]{.36\textwidth}
    \centering
    NeuS
\end{minipage}
\begin{minipage}[c]{.36\textwidth}
    \centering
    Ours
\end{minipage}

\caption{Additional results on the DTU~\cite{jensen2014large} dataset (the first three scenes) and BMVS~\cite{yao2020blendedmvs} dataset (the last one scene) with mask supervision.}
\label{supp_dtu}
\end{figure*}

\begin{figure*}[h]

\begin{minipage}[c]{.2\textwidth}
    \centering
    \includegraphics[width=1\linewidth]{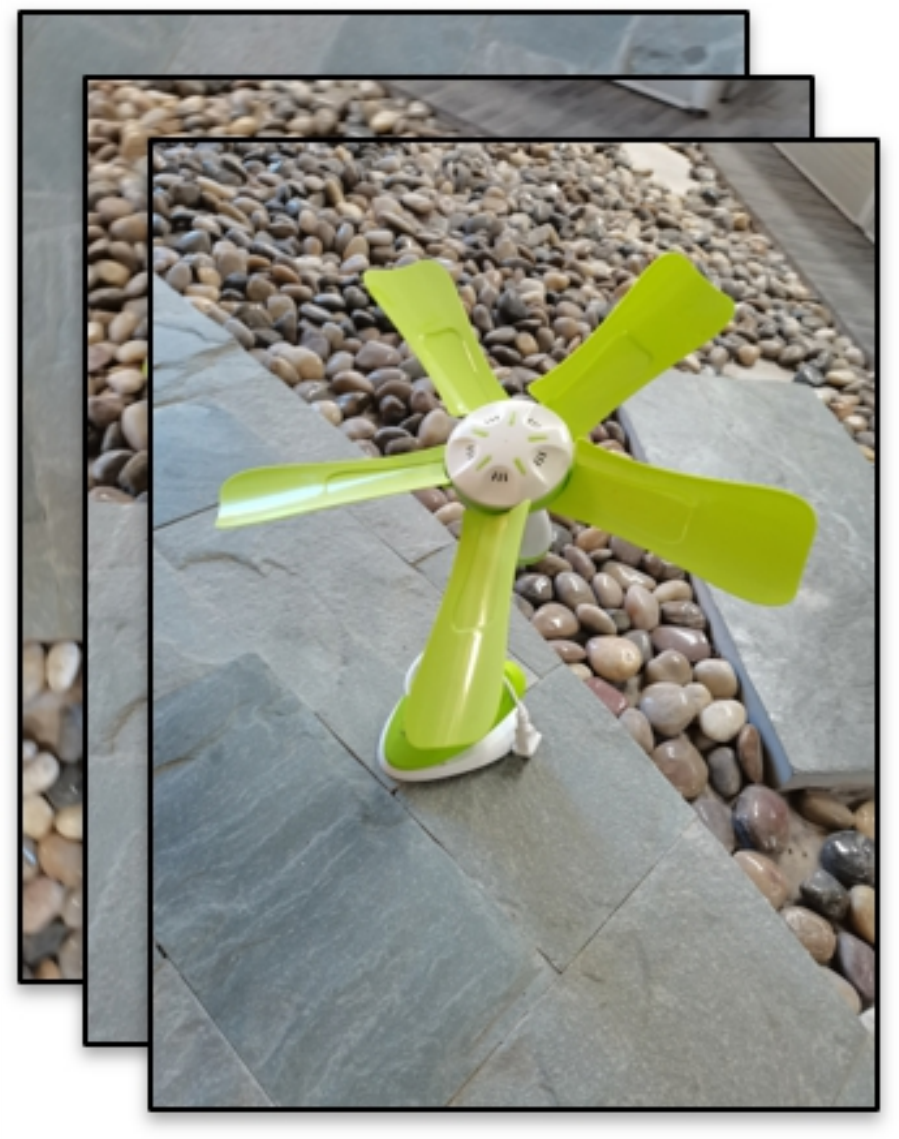}
\end{minipage}
\begin{minipage}[c]{.25\textwidth}
    \centering
    \includegraphics[width=1\linewidth]{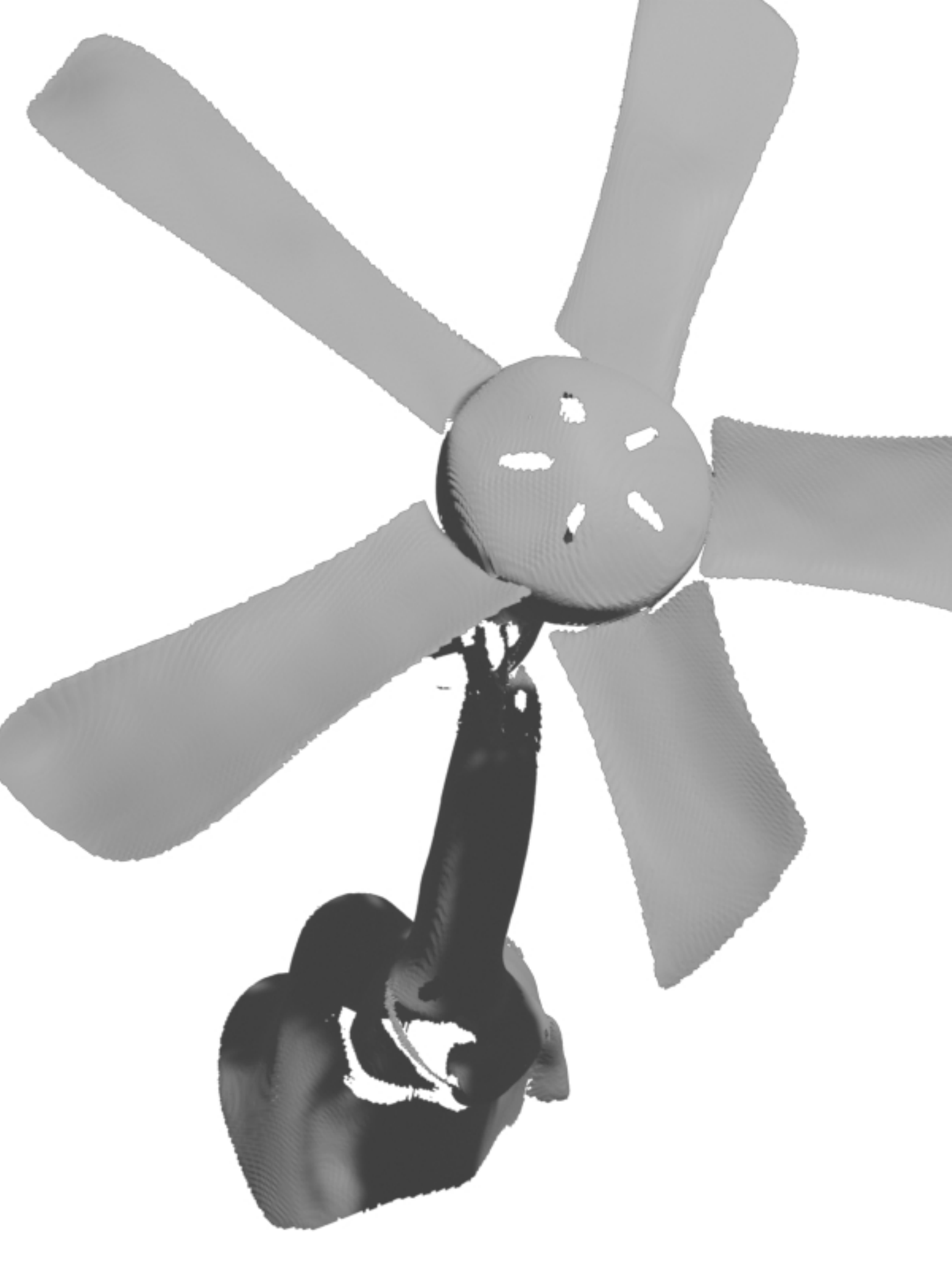}
\end{minipage}
\begin{minipage}[c]{.25\textwidth}
    \centering
    \includegraphics[width=1\linewidth]{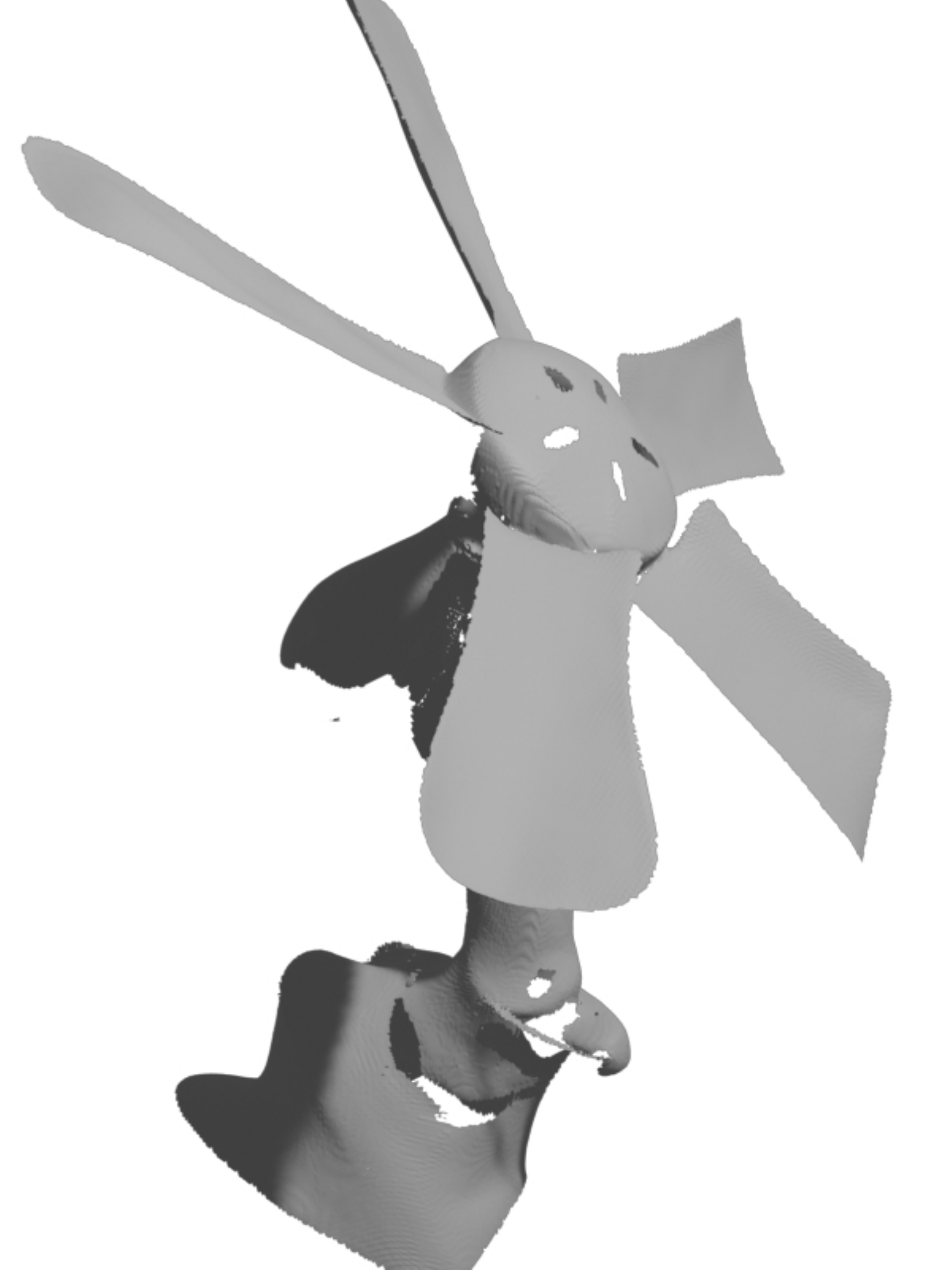}
\end{minipage}
\begin{minipage}[c]{.25\textwidth}
    \centering
    \includegraphics[width=1\linewidth]{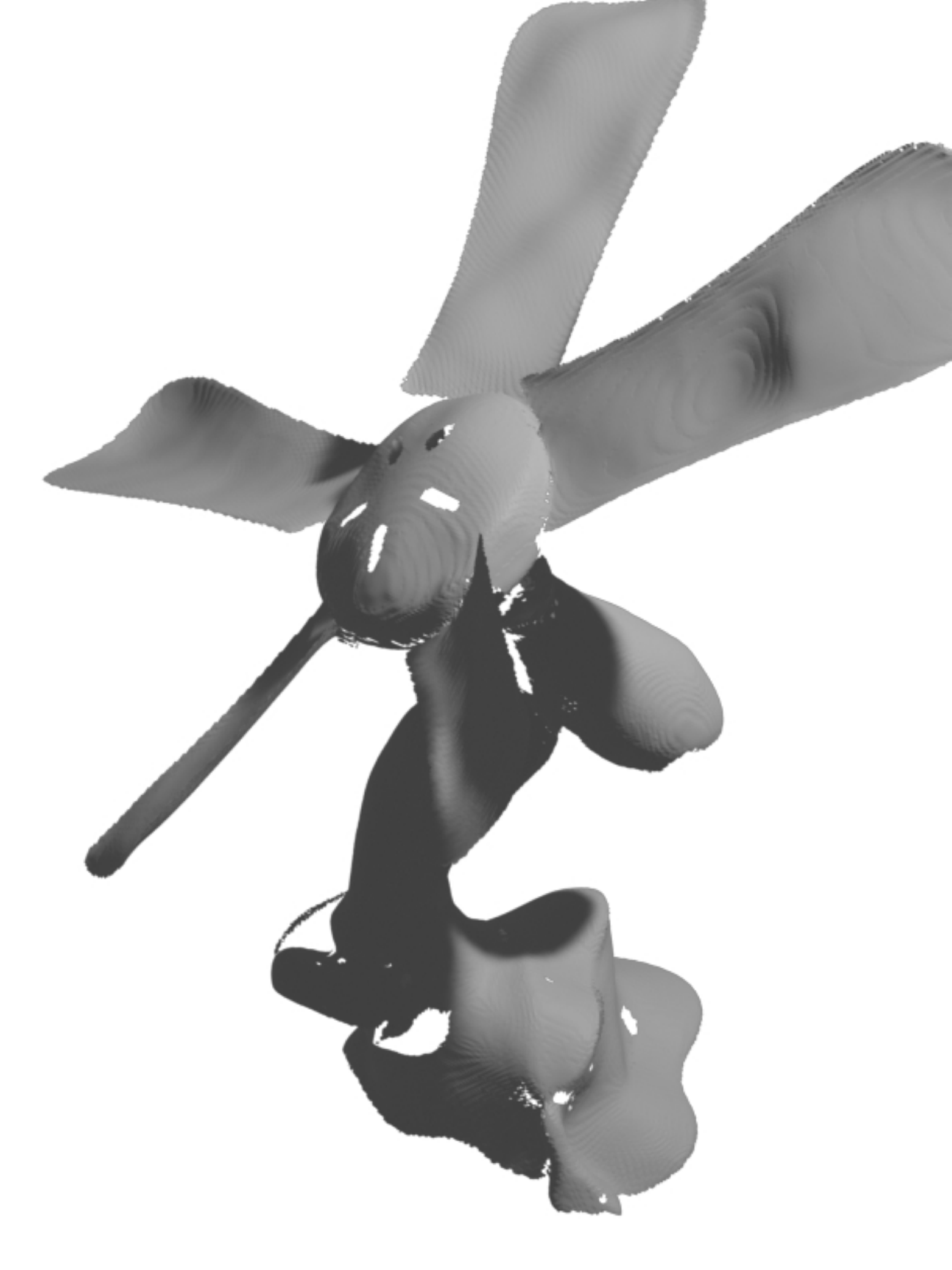}
\end{minipage}

\begin{minipage}[c]{.2\textwidth}
    \centering
    \includegraphics[width=1\linewidth]{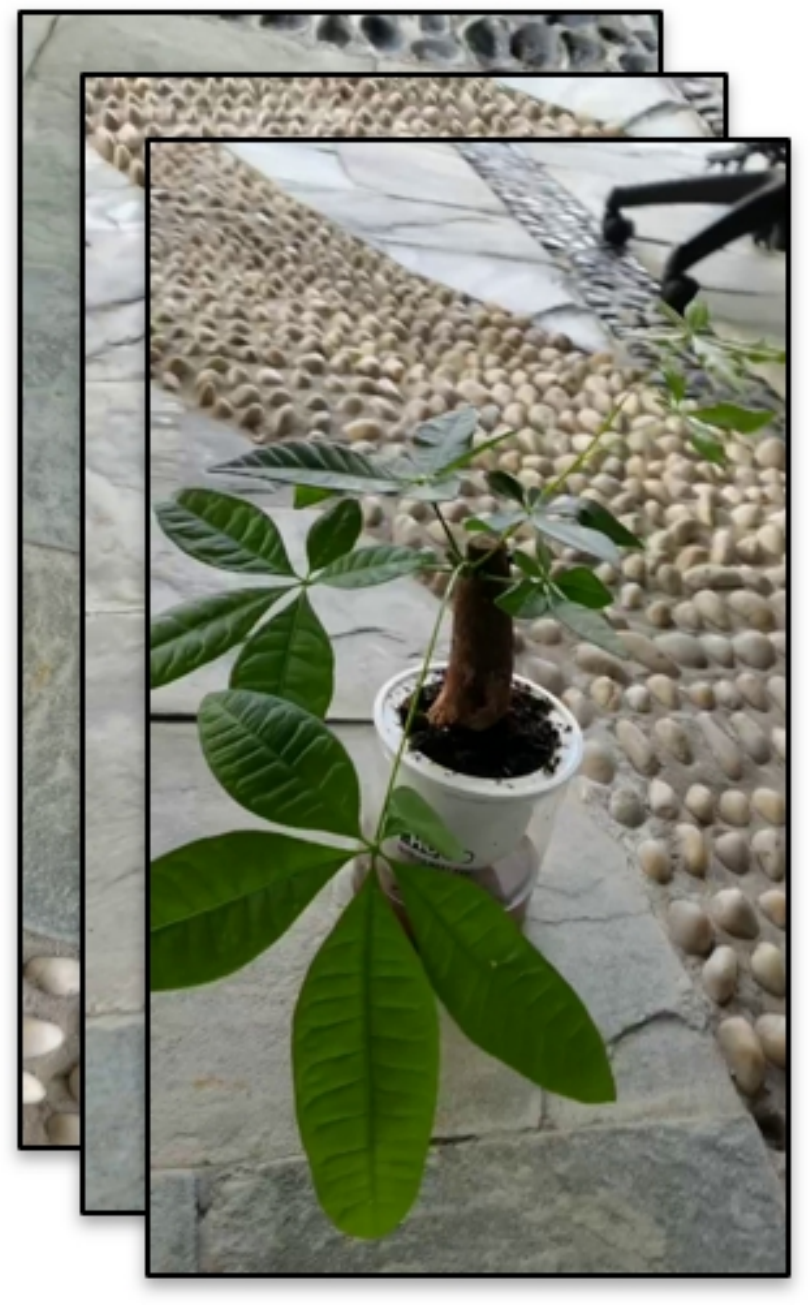}
\end{minipage}
\begin{minipage}[c]{.25\textwidth}
    \centering
    \includegraphics[width=1\linewidth]{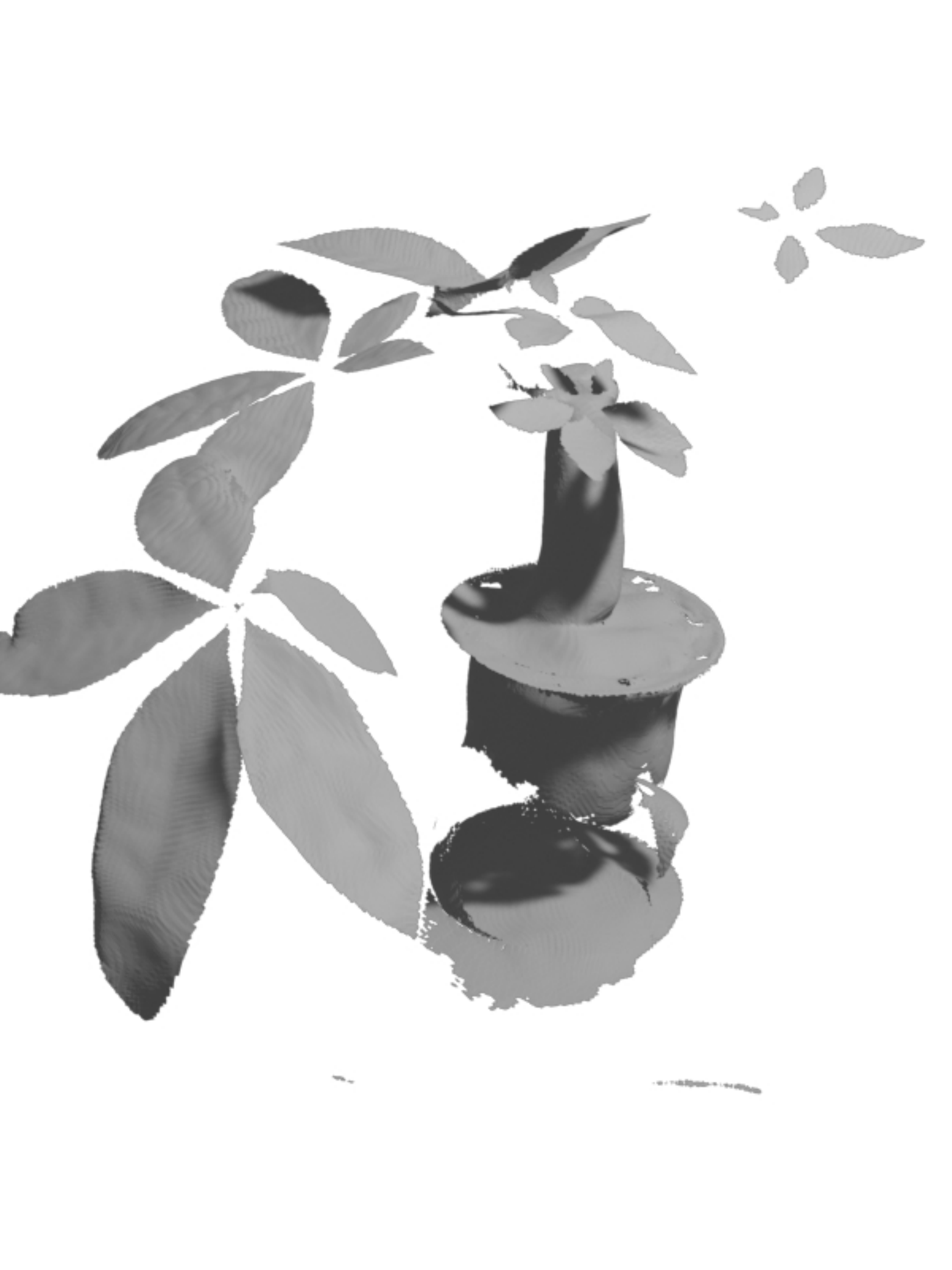}
\end{minipage}
\begin{minipage}[c]{.25\textwidth}
    \centering
    \includegraphics[width=1\linewidth]{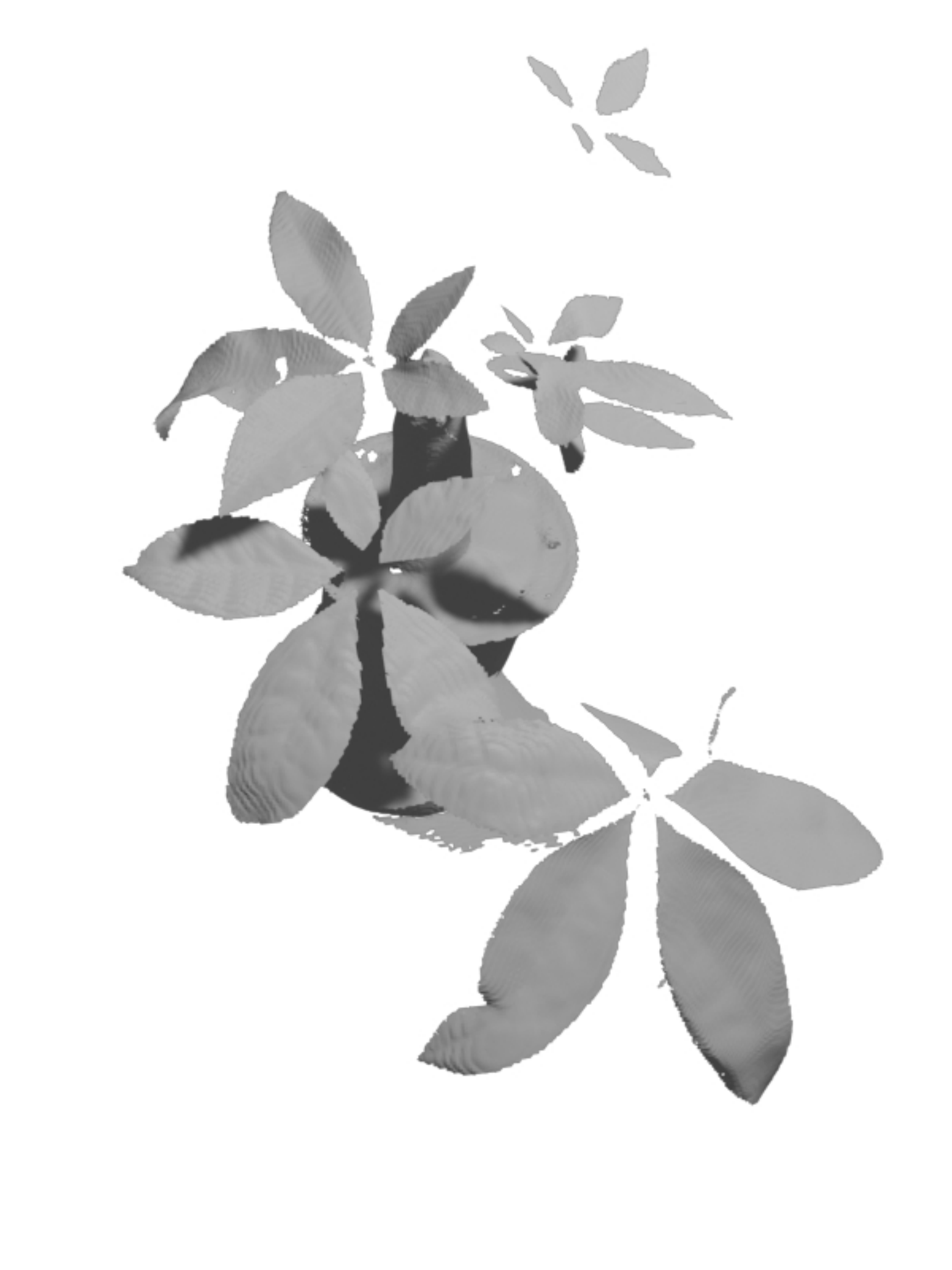}
\end{minipage}
\begin{minipage}[c]{.25\textwidth}
    \centering
    \includegraphics[width=1\linewidth]{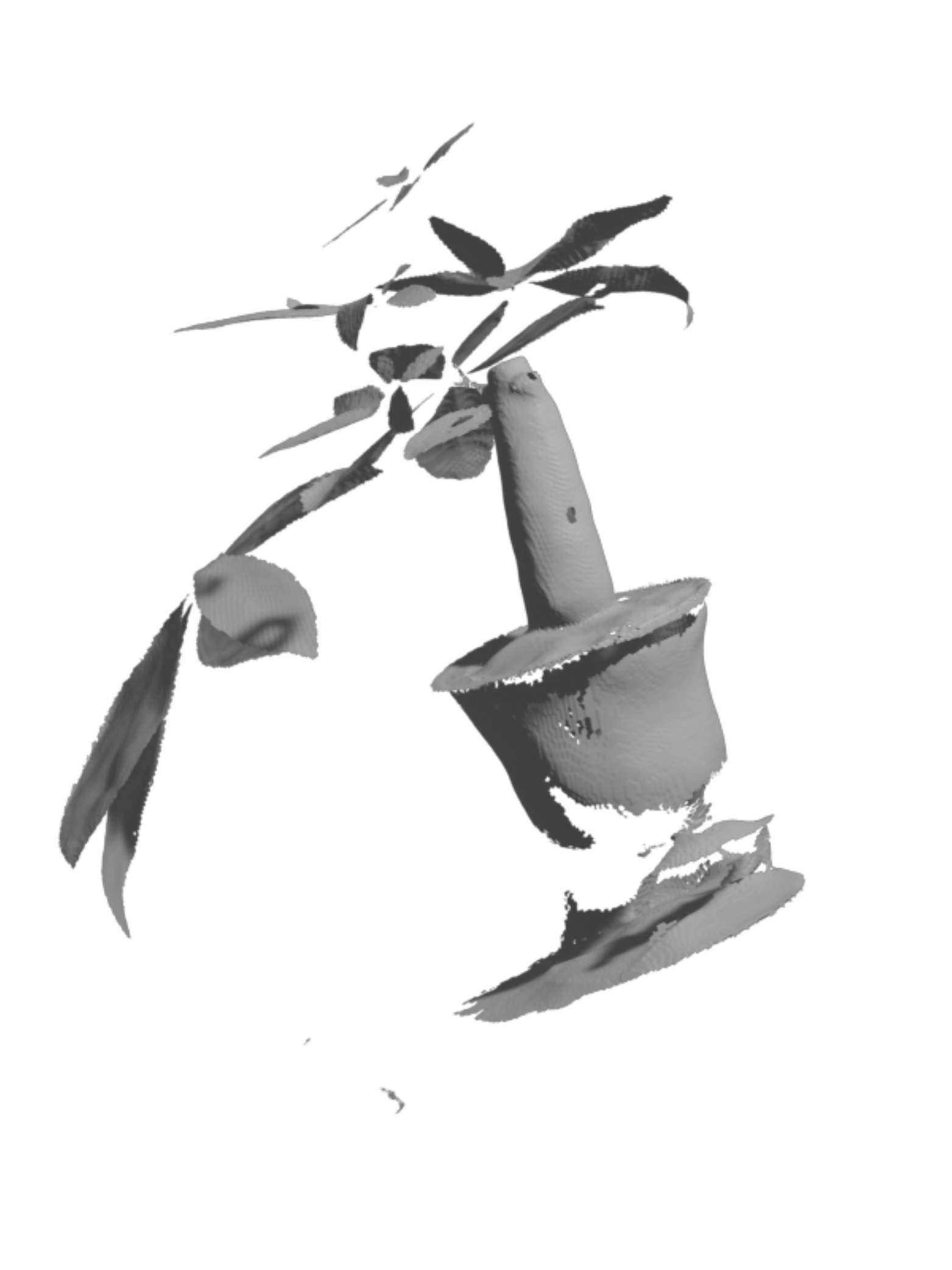}
\end{minipage}

\begin{minipage}[c]{.2\textwidth}
    \centering
    Input
\end{minipage}
\begin{minipage}[c]{.72\textwidth}
    \centering
    Ours
\end{minipage}

\caption{Additional results of the real-captured data without mask supervision.}
\label{supp_real}
\end{figure*}

We visualize more reconstruction results of NeUDF on DF3D~\cite{zhu2020deep}, MGN~\cite{mgn}, DTU~\cite{jensen2014large}, BMVS~\cite{yao2020blendedmvs} datasets and real-captured data. 
Fig.~\ref{supp_df3d_wo} shows the comparison with NeuS on the DF3D dataset without mask supervision. 
Fig.~\ref{supp_df3d} shows the comparison with NeuS on the DF3D dataset with mask supervision. 
Fig.~\ref{supp_mgn_wo} shows the comparison with NeuS on the MGN dataset without mask supervision. 
Fig.~\ref{supp_mgn} shows the comparison with NeuS on the MGN dataset with mask supervision. 
Fig.~\ref{supp_dtu} shows the comparison with NeuS on the DTU and BMVS datasets with mask supervision. 
Fig.~\ref{supp_real} shows the additional results of the real-captured scenes with open surfaces.


\end{document}